\documentclass[11pt, letterpaper]{article}
\usepackage{pdfpages}
\usepackage{amsmath}
\usepackage{amssymb}
\usepackage{multirow}
\usepackage{amsthm}
\usepackage{graphicx}
\usepackage{geometry}

\usepackage{url}
\usepackage{algorithm}
\usepackage{tocloft}
\usepackage{algorithmic}
\usepackage{caption}
\usepackage{subcaption}
\usepackage{booktabs}
\usepackage{rotating}
\usepackage{authblk}
\usepackage{appendix}
\usepackage[utf8]{inputenc}
\usepackage[T1]{fontenc}
\usepackage[english]{babel}
\usepackage{fancyhdr}
\usepackage{microtype}
\usepackage{parskip}
\usepackage{paralist, tabularx}
\usepackage{enumitem}
\usepackage{placeins}
\usepackage{float}
\usepackage{adjustbox}
\usepackage{placeins}
\usepackage{lscape}
\usepackage{ragged2e}
\usepackage{soul} 
\sethlcolor{green} 
\usepackage[style=numeric,backend=biber]{biblatex}
\usepackage{hyperref}
\usepackage{csquotes}

\hypersetup{
    colorlinks=true,
    linkcolor=blue,
    filecolor=magenta,      
    urlcolor=cyan,
    citecolor=red,
}

\definecolor{headercolor}{RGB}{0, 50, 100}
\newcommand{\fig}[4]{\begin{figure}[H]\begin{center}\includegraphics[#1]{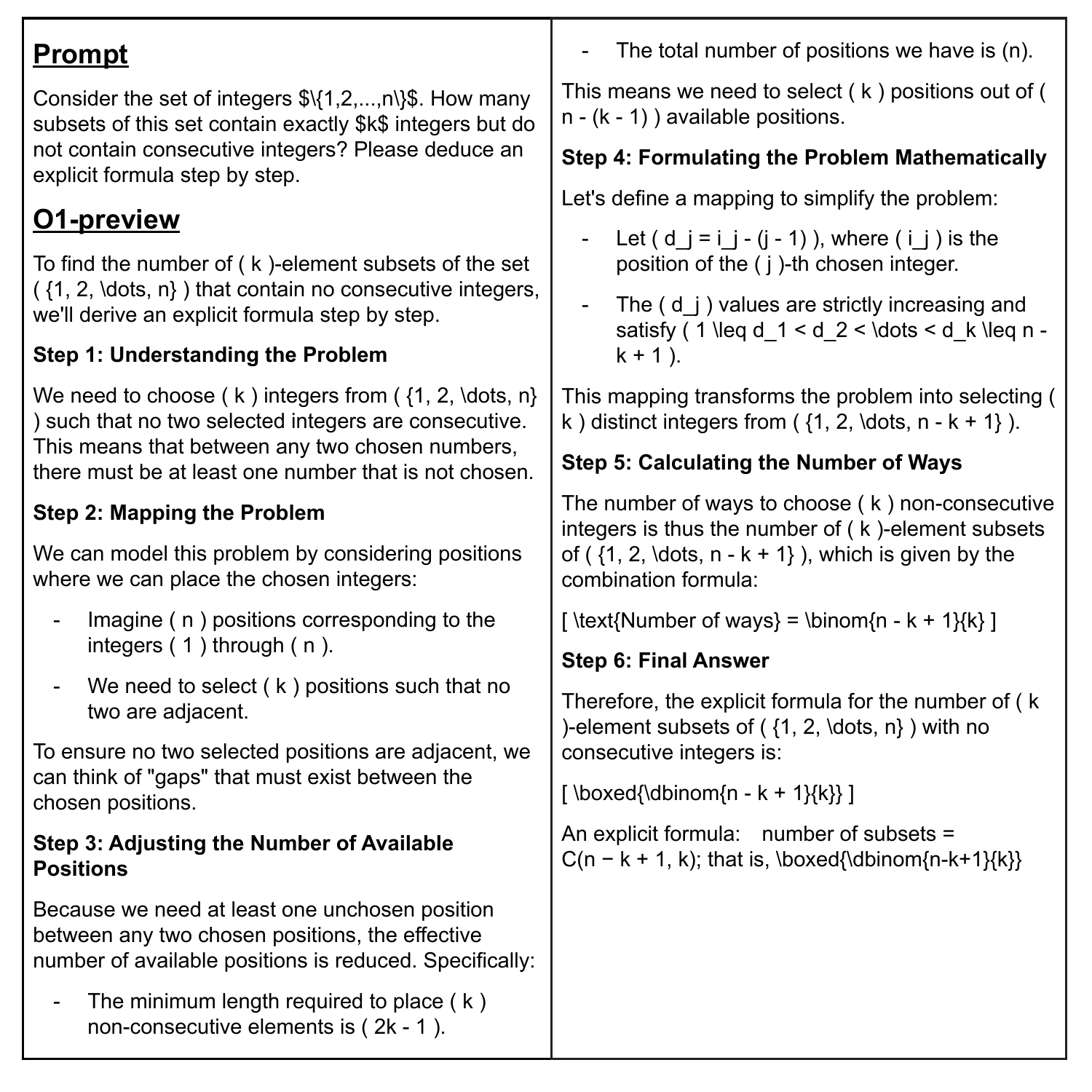}\end{center}\caption[#3]{\textbf{#3.} #4}\label{fig:#2}\end{figure}}
\title{Evaluation of OpenAI o1: Opportunities and Challenges of AGI}

\date{}

\newcommand*\samethanks[1][\value{footnote}]{\footnotemark[#1]}
\author[1]{Tianyang Zhong \thanks{Co-first authors.}}
\author[2]{Zhengliang Liu \samethanks}
\author[2]{Yi Pan \samethanks}
\author[3]{Yutong Zhang \samethanks}
\author[6]{Zeyu Zhang \samethanks}
\author[4,40]{Yifan Zhou \samethanks}
\author[5]{Shizhe Liang \samethanks}
\author[2]{Zihao Wu \samethanks}
\author[6]{Yanjun Lyu \samethanks}
\author[2]{Peng Shu \samethanks}
\author[6]{Xiaowei Yu \samethanks}
\author[6]{Chao Cao \samethanks}
\author[2]{Hanqi Jiang \samethanks}
\author[7]{Hanxu Chen \samethanks}

\author[2]{Yiwei Li \thanks{Co-second authors.}}
\author[2]{Junhao Chen \samethanks}
\author[8]{Huawen Hu \samethanks}
\author[9]{Yiheng Liu \samethanks}
\author[2]{Huaqin Zhao \samethanks}
\author[2]{Shaochen Xu \samethanks}
\author[2]{Haixing Dai \samethanks}
\author[2]{Lin Zhao \samethanks}
\author[10]{Ruidong Zhang \samethanks}
\author[11,12,13]{Wei Zhao \samethanks} 
\author[14]{Zhenyuan Yang \samethanks}
\author[15]{Jingyuan Chen \samethanks}
\author[15]{Peilong Wang \samethanks}
\author[2]{Wei Ruan \samethanks}
\author[16]{Hui Wang \samethanks}
\author[17]{Huan Zhao \samethanks}

\author[6]{Jing Zhang \thanks{Co-third authors.}}
\author[18]{Yiming Ren \samethanks}
\author[18]{Shihuan Qin \samethanks}
\author[6]{Tong Chen \samethanks}
\author[2]{Jiaxi Li \samethanks}
\author[19]{Arif Hassan Zidan \samethanks}
\author[19]{Afrar Jahin \samethanks}
\author[6]{Minheng Chen \samethanks}
\author[9]{Sichen Xia \samethanks} 
\author[15]{Jason Holmes \samethanks}
\author[6]{Yan Zhuang \samethanks} 
\author[8]{Jiaqi Wang \samethanks}
\author[20]{Bochen Xu \samethanks}
\author[21,22]{Weiran Xia \samethanks}
\author[2]{Jichao Yu \samethanks} 
\author[22]{Kaibo Tang \samethanks}
\author[23]{Yaxuan Yang \samethanks}
\author[24]{Bolun Sun \samethanks}
\author[2]{Lifeng Chen \samethanks}

\author[25]{Tao Yang}
\author[26]{Guoyu Lu}
\author[27]{Xianqiao Wang}
\author[28]{Lilong Chai}
\author[29]{He Li} 
\author[2]{Jin Lu}
\author[3]{Xin Zhang}
\author[20]{Bao Ge}
\author[9]{Xintao Hu}
\author[18]{Lian Zhang}
\author[30]{Hua Zhou}
\author[31]{Lu Zhang} 
\author[8]{Shu Zhang}
\author[2]{Zhen Xiang}
\author[32]{Yudan Ren}
\author[11,12]{Jun Liu}  
\author[17]{Xi Jiang}  
\author[33]{Yu Bao}
\author[19]{Wei Zhang}
\author[34]{Xiang Li}
\author[22]{Gang Li}
\author[15]{Wei Liu}
\author[35,36]{Dinggang Shen}
\author[37]{Andrea Sikora}
\author[38,39]{Xiaoming Zhai}
\author[6]{Dajiang Zhu}
\author[9]{Tuo Zhang}
\author[2,39]{Tianming Liu}

\affil[1]{Department of Mathematical and Statistical Sciences, University of Alberta, Edmonton, Canada}
\affil[2]{School of Computing, University of Georgia, GA, USA}
\affil[3]{Institute of Medical Research, Northwestern Polytechnical University, Xi'an, China}
\affil[4]{College of Arts and Sciences, University of Georgia, Athens, USA}
\affil[5]{Institute of Plant Breeding, Genetics \& Genomics, University of Georgia, Athens, GA, USA}
\affil[6]{Department of Computer Science and Engineering, University of Texas at Arlington, TX, USA}
\affil[7]{The Lamar Dodd School of Art, University of Georgia, GA, USA}
\affil[8]{School of Computer Science, Northwestern Polytechnical University, Xi'an, China}
\affil[9]{School of Automation, Northwestern Polytechnical University, Xi'an, China}
\affil[10]{University of California, Los Angeles, CA, USA}
\affil[11]{Department of Radiology, The Second Xiangya Hospital, Central South University, Changsha, China}
\affil[12]{Clinical Research Center for Medical Imaging in Hunan Province, Changsha, China}
\affil[13]{Institute of Biomedical and Health Engineering, Shenzhen Institutes of Advanced Technology, Chinese Academy of Sciences, Shenzhen, China}
\affil[14]{Guanghua School of Management, Peking University, Beijing, China}
\affil[15]{Department of Radiation Oncology, Mayo Clinic, Phoenix, Arizona, USA}
\affil[16]{Second Language Acquisition and Teaching, University of Arizona, Tucson, AZ, USA}
\affil[17]{School of Life Science and Technology, University of Electronic Science and Technology of China, Chengdu, China}
\affil[18]{The First Hospital of Hebei Medical University, Shijiazhuang, Hebei, China}
\affil[19]{School of Computer and Cyber Sciences, Augusta University, Augusta, GA, USA}
\affil[20]{School of Physics and Information Technology, Shaanxi Normal University, Xi'an, China}
\affil[21]{School of Future Technology, South China University of Technology, Guangzhou, China}
\affil[22]{Department of Radiology and BRIC, University of North Carolina at Chapel Hill, NC, USA}
\affil[23]{Department of Educational Psychology, University of Georgia, Athens, GA, USA}
\affil[24]{Johns Hopkins University, Baltimore, MD, USA}
\affil[25]{School of Architecture, Tsinghua University, Beijing, China}
\affil[26]{College of Engineering, University of Georgia, Athens, USA}
\affil[27]{School of Environmental, Civil, Agricultural and Mechanical Engineering, College of Engineering, University of Georgia, Athens, GA, USA}
\affil[28]{Department of Poultry Science, College of Agricultural and Environmental Sciences, University of Georgia, Athens, GA, USA}
\affil[29]{School of Chemical, Materials and Biomedical Engineering, University of Georgia, Athens, USA}
\affil[30]{Fielding School of Public Health, University of California, Los Angeles, Los Angeles, CA, USA}
\affil[31]{Department of Computer Science, Indiana University Indianapolis, IN, USA}
\affil[32]{School of Information Science \& Technology, Northwest University, Xi'an, China}
\affil[33]{Department of Graduate Psychology, James Madison University, Harrisonburg, VA, USA}
\affil[34]{Department of Radiology, Massachusetts General Hospital and Harvard Medical School, MA, USA}
\affil[35]{School of Biomedical Engineering, ShanghaiTech University, and Shanghai Clinical Research and Trial Center, Shanghai, China}
\affil[36]{Shanghai United Imaging Intelligence Co., Ltd.}
\affil[37]{College of Pharmacy, University of Georgia, GA, USA}
\affil[38]{Department of Mathematics, Science, and Social Studies Education, University of Georgia, Athens, GA, USA}
\affil[39]{AI4STEM Education Center, University of Georgia, Athens, GA, USA}

\begin{document}

\maketitle

\newpage

\begin{abstract}
This comprehensive study evaluates the performance of OpenAI's o1-preview large language model across a diverse array of complex reasoning tasks, spanning multiple domains, including computer science, mathematics, natural sciences, medicine, linguistics, and social sciences. Through rigorous testing, o1-preview demonstrated remarkable capabilities, often achieving human-level or superior performance in areas ranging from coding challenges to scientific reasoning and from language processing to creative problem-solving. Key findings include:

\begin{itemize}
\item 83.3\% success rate in solving complex competitive programming problems, surpassing many human experts.
\item Superior ability in generating coherent and accurate radiology reports, outperforming other evaluated models.
\item 100\% accuracy in high school-level mathematical reasoning tasks, providing detailed step-by-step solutions.
\item Advanced natural language inference capabilities across general and specialized domains like medicine.
\item Impressive performance in chip design tasks, outperforming specialized models in areas such as EDA script generation and bug analysis.
\item Remarkable proficiency in anthropology and geology, demonstrating deep understanding and reasoning in these specialized fields.
\item Strong capabilities in quantitative investing. O1 has comprehensive financial knowledge and statistical modeling skills.
\item Effective performance in social media analysis, including sentiment analysis and emotion recognition.
\end{itemize}

The model excelled particularly in tasks requiring intricate reasoning and knowledge integration across various fields. While some limitations were observed, including occasional errors on simpler problems and challenges with certain highly specialized concepts, the overall results indicate significant progress towards artificial general intelligence. This evaluation not only highlights o1-preview's current strengths and limitations but also identifies crucial areas for future development, including multi-modal integration, domain-specific validation, and ethical considerations for real-world applications. The findings provide valuable insights into the potential of large language models in numerous fields and pave the way for further advancements in AI research and application.
\end{abstract}
\newpage
\tableofcontents


\listoffigures

\newpage

\section{Introduction}\label{sec:introduction}
large language models (LLMs) have rapidly advanced in recent years, demonstrating impressive capabilities across a wide range of tasks~\cite{liu2023summary,zhou2023comprehensive,zhao2023brain,liu2023holistic}. This study aims to comprehensively evaluate the performance of OpenAI's \textbf{o1} model, widely perceived as the model with the highest reasoning capabilities in history so far, on complex reasoning tasks spanning multiple disciplines.
O1 represents a significant leap forward in AI reasoning capabilities. According to OpenAI's blog post~\cite{o12024}, o1 demonstrates remarkable proficiency across various domains, including competitive programming, advanced mathematics, and PhD-level scientific problem-solving. It ranks in the 89th percentile on competitive programming questions, places among the top 500 students in the US in a qualifier for the USA Math Olympiad, and surpasses human PhD-level accuracy on a benchmark of physics, biology, and chemistry problems. These achievements underscore o1's potential to revolutionize AI applications in science, coding, mathematics, and related fields.

While standard benchmarks have shown promising results~\cite{o12024}, they can be manipulated and are often not comprehensive enough to fully assess the model's capabilities \cite{mcintosh2024inadequacies,mizrahi2024state}. By rigorously testing the model's abilities across a wide range of domains with real world tasks, we seek to provide a more holistic evaluation of its progress towards artificial general intelligence and identify areas for further development.

\subsection{Background: What is New with o1}
\label{subsec:background}
LLMs, built on the Transformer \cite{vaswani2017attention} architecture, have evolved from early work such as BERT~\cite{devlin2018bert} and GPT~\cite{radford2018improving} to more advanced models such as GPT-3 and GPT-4~\cite{brown2020language,achiam2023gpt}. These models, trained on vast corpora of text data, have shown significant proficiency in understanding context, generating human-like text, and performing complex reasoning tasks~\cite{liu2023summary,zhou2023comprehensive,zhao2023brain,liu2023holistic,lyu2024gp,liu2024understanding,holmes2023evaluating,zhao2023ophtha,lee2023multimodality,liu2024radiation,wang2024large,sun2024trustllm,liu2024surviving,tan2023promises}.

\subsubsection{OpenAI o1 \& Chain-of-Thought Reasoning}
Chain-of-thought reasoning \cite{wei2022chain} is a recent advancement in LLMs that enables models to break down complex problems into intermediate steps, mirroring human-like problem-solving processes. This approach has shown particular promise in enhancing performance on tasks requiring multi-step reasoning or mathematical problem-solving. By explicitly generating a series of coherent thoughts leading to a conclusion, models can tackle more sophisticated problems and provide more interpretable outputs.

Compared to its predecessor GPT-4, o1 explicitly incorporates chain-of-thought into its inference process~\cite{o12024}. While GPT-4 could utilize chain-of-thought reasoning when prompted, o1 integrates this approach as a fundamental part of its architecture. This allows o1 to "think before it answers," producing a long internal chain of thought before responding to the user. This integration enables o1 to handle more complex reasoning tasks and provide more transparent explanations of its problem-solving process, potentially surpassing GPT-4's capabilities in areas requiring deep, multi-step reasoning.

\subsubsection{OpenAI o1 \& Reinforcement Learning}
Reinforcement Learning from Human Feedback (RLHF) is a powerful technique that has significantly advanced the capabilities of large language models. It combines reinforcement learning principles with human preferences to fine-tune models, aligning their outputs more closely with human expectations. RLHF typically involves supervised fine-tuning, reward modeling based on human preferences, and policy optimization through reinforcement learning. For a more detailed discussion of RLHF, see Section~\ref{rlhf_relatedwork}.

O1 employs advanced reinforcement learning techniques that significantly evolve beyond traditional RLHF methods. According to OpenAI, o1's performance consistently improves with more reinforcement learning (train-time compute) and with more time spent thinking (test-time compute)~\cite{o12024}. This novel approach likely incorporates Chain-of-Thought reasoning into its reinforcement learning framework, which allows it to generate and evaluate multiple reasoning paths before producing a final output. 

Unlike traditional models that primarily spend compute during training, o1 scales its performance with increased compute during inference. This suggests a form of online learning or search that occurs at test time, which could involve real-time exploration and refinement of reasoning strategies. The model potentially rewards not just the final answer, but also the quality and effectiveness of its reasoning steps. O1 appears to have mechanisms for self-reflection and improvement, which implement a form of self-supervised learning where the model's thoughts become training data for further enhancement~\cite{o12024,metadocsOpenAIWorks,interconnectsReverseEngineering}. 

It might generate multiple candidate reasoning paths in parallel, which use reinforcement learning to score and select the most promising paths, similar to the Quiet-STaR method~\cite{zelikman2024quiet,metadocsOpenAIWorks}. The model's ability to improve with more thinking time'' suggests a continuous learning loop, which possibly implements a form of meta-learning that adapts its reasoning strategies based on the specific task. 

These advanced techniques represent a paradigm shift which focuses on enhancing the model's reasoning capabilities during inference, rather than solely aligning its outputs with human preferences during training.

\subsection{Motivation}\label{subsec:motivation}
As LLMs grow more sophisticated, there is a critical need to understand their true capabilities and limitations beyond standard benchmarks. This study is driven by the goal of assessing o1-preview's ability to handle complex, multi-disciplinary tasks that require deep reasoning and knowledge integration. Such a comprehensive evaluation provides valuable insights into the current state of LLM technology and its potential for real-world applications, which is of significant interest to the scientific community.

\begin{figure}[H]
\centering
  \includegraphics[width = \textwidth]{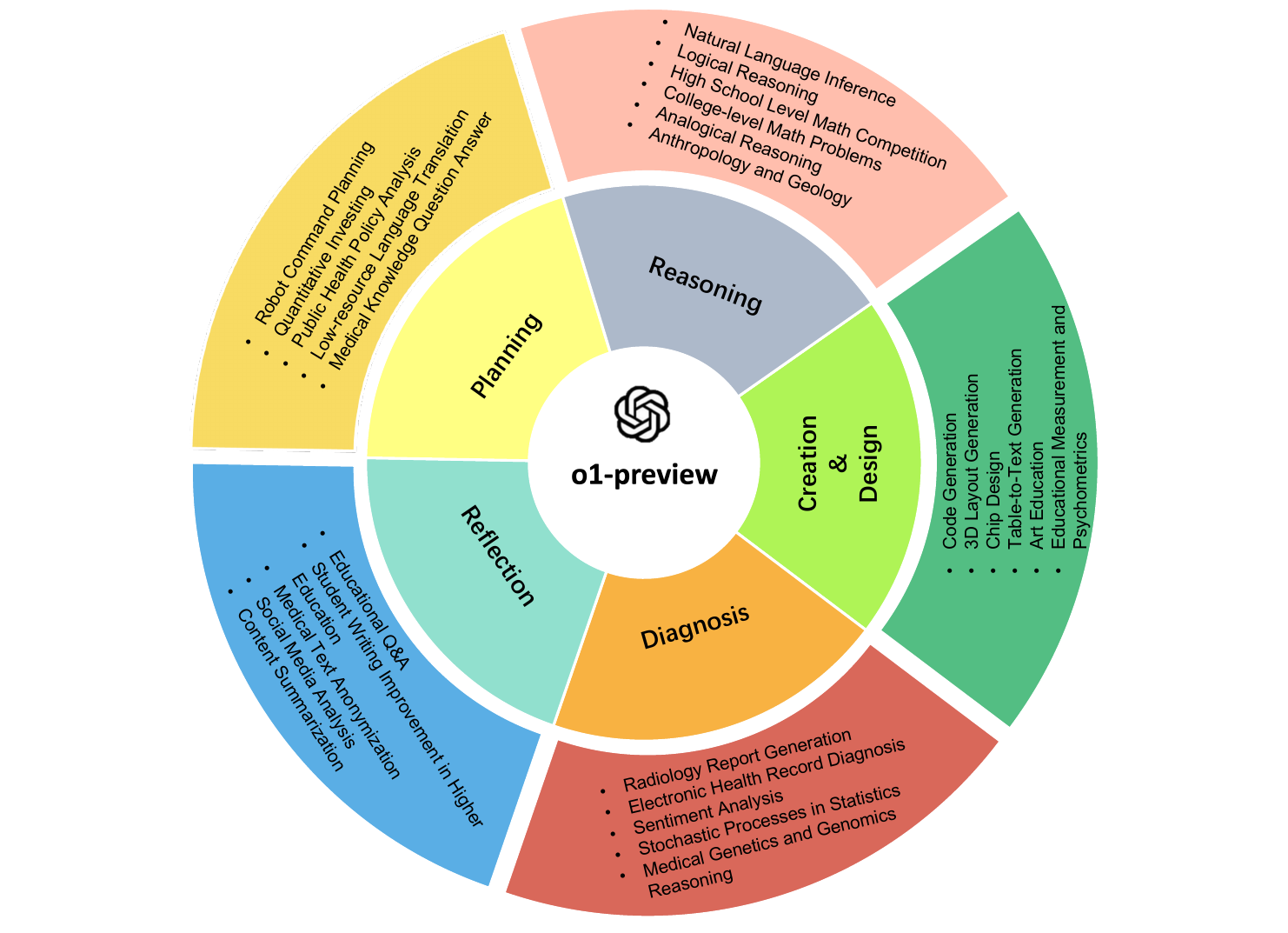}

\caption[Schematic Overview of the Evaluation Methodology]{\textbf{Schematic Overview of the Evaluation Methodology.} This diagram illustrates the five major evaluation domains for o1-preview: Creation and Design, Planning, Reasoning, Diagnosis, and Reflection. Each domain is tested through relevant tasks. The 27 distinct tasks evaluate the model's adaptability and effectiveness across a diverse set of cognitive and real-world challenges.}
\label{fig:Overview}      
\end{figure}

To comprehensively assess the capabilities of o1-preview, we structured our evaluation around five major domains: Creation and Design, Planning, Reasoning, Diagnosis, and Reflection (see Figure \ref{fig:Overview}). Each domain encompasses a set of relevant tasks that test specific aspects of the model's performance. In total, 27 tasks were designed to evaluate the model's adaptability and effectiveness across a diverse array of cognitive and real-world challenges.

\subsection{Key Findings}
Our comprehensive evaluation of o1-preview across various domains revealed several main insights:
\begin{itemize}
\item \textbf{Advanced Reasoning Capabilities:} o1-preview demonstrated exceptional logical reasoning abilities in multiple fields, including high school mathematics, quantitative investing, and chip design. It showed a strong capacity for step-by-step problem-solving and the ability to handle complex, multi-layered tasks.
\item \textbf{Domain-Specific Knowledge:} The model exhibited impressive knowledge breadth across diverse fields such as medical genetics, radiology, anthropology, and geology. It often performed at a level comparable to or exceeding that of graduate students or early-career professionals in these domains.

\item \textbf{Creative and Practical Applications:} In areas such as 3D layout generation and art education, o1-preview showed creativity and practical application skills, generating functional designs and structured lesson plans. However, it still lacks the flexibility and adaptability of human experts in these fields.

\item \textbf{Natural Language Understanding:} The model excelled in tasks requiring nuanced language understanding, such as sentiment analysis, social media analysis, and content summarization. It demonstrated the ability to capture complex expressions like irony and sarcasm, though it still struggles with very subtle emotional nuances.

\item \textbf{Scientific and Medical Reasoning:} o1-preview showed strong capabilities in medical diagnosis, radiology report generation, and answering complex medical exam questions. While it performed well in these areas, its reasoning process sometimes differed from that of trained medical professionals.

\item \textbf{Limitations and Areas for Improvement:} Despite its impressive performance, o1-preview showed limitations in handling extremely abstract logical puzzles, adapting to real-time dynamic situations, and consistently performing well on the most complex tasks in fields like advanced mathematics and stochastic processes.

\item \textbf{Potential for Real-World Applications:} The model's performance suggests significant potential for applications in various fields, from educational support and medical assistance to financial analysis and scientific research. However, further refinement and validation are necessary before deployment in critical real-world scenarios.

\end{itemize}

\subsection{AGI-Benchmark 1.0}

To contribute to the field of AI research and evaluation, we are introducing \textbf{AGI-Benchmark 1.0}, a comprehensive collection of the complex reasoning tasks used in this study to evaluate o1-preview. Unlike existing language model benchmarks such as MMLU~\cite{hendrycks2020measuring}, which primarily focus on question-answering and multiple-choice formats, AGI-Benchmark 1.0 is designed to assess a model's ability to tackle intricate, multi-step reasoning problems across a diverse set of domains.

This benchmark encompasses tasks from \textbf{27 distinct categories}, grouped into five major cognitive faculties:

\begin{itemize}
    \item \textbf{Reasoning:}
    \begin{itemize}
        \item Natural Language Inference
        \item Logical Reasoning
        \item High School Level Math Competition
        \item College-level Math Problems
        \item Analogical Reasoning
        \item Anthropology and Geology
    \end{itemize}

    \item \textbf{Planning:}
    \begin{itemize}
        \item Robot Command Planning
        \item Quantitative Investing
        \item Public Health Policy Analysis
        \item Low-resource Language Translation
        \item Medical Knowledge Question Answering
    \end{itemize}

    \item \textbf{Creation \& Design:}
    \begin{itemize}
        \item Code Generation
        \item 3D Layout Generation
        \item Chip Design
        \item Table-to-Text Generation
        \item Art Education
        \item Educational Measurement and Psychometrics
    \end{itemize}

    \item \textbf{Diagnosis:}
    \begin{itemize}
        \item Radiology Report Generation
        \item Electronic Health Record Diagnosis
        \item Sentiment Analysis
        \item Stochastic Processes in Statistics
        \item Medical Genetics and Genomics Reasoning
    \end{itemize}

    \item \textbf{Reflection:}
    \begin{itemize}
        \item Educational Q\&A
        \item Student Writing Improvement in Higher Education
        \item Medical Text Anonymization
        \item Social Media Analysis
        \item Content Summarization
    \end{itemize}
\end{itemize}

These tasks reflect the complexity of real-world problems and challenge models to demonstrate not just factual knowledge, but the ability to reason through novel contexts, engage in multi-step problem-solving, and exhibit creativity. \textbf{AGI-Benchmark 1.0} resists manipulation and memorization, providing a more authentic evaluation of a model's reasoning capabilities.

By releasing AGI-Benchmark 1.0 to the public, we aim to foster transparency, reproducibility, and collaborative progress in the pursuit of artificial general intelligence. This benchmark will be an invaluable resource for researchers and developers, guiding advancements in AI systems capable of solving real-world, complex problems across multiple domains. This benchmark will be available at \url{https://github.com/UGA-CAID/AGI-Bench}.

\section{Scope of the Study and Used Public Datasets}
\label{sec:scope}

In our study, we aim to explore and evaluate o1-preview's capabilities and limitations across various domains and complex tasks. Below is a comprehensive list of the domains and tasks we have included in our research.

\subsection{Code Generation}
\label{subsec:Coding-Generation}
Following the initial evaluation of o1-preview's coding capabilities, we extended our assessment by testing its performance in a competitive programming environment, specifically within Leetcode contests.

Leetcode is a widely recognized platform for coding challenges, designed to test and improve problem-solving skills in a variety of programming languages. It offers a range of problems covering topics such as algorithms, data structures, dynamic programming, and system design. Leetcode's contests are time-bound competitive events that challenge participants to solve multiple algorithmic problems of varying difficulty within a limited timeframe. These contests provide an ideal benchmark for evaluating the real-world coding abilities of language models, as they require not only syntactic correctness but also efficient problem-solving and optimization skills. Leetcode has hosted 415 weekly contests and 139 biweekly contests, each consisting of four problems designed to comprehensively cover the domain of data structures and algorithms. These contests serve as a thorough test of coding knowledge, spanning topics such as sorting, dynamic programming, graph theory, and more.

For the purpose of evaluating o1-preview, we specifically test its performance on Leetcode Weekly Contests 414 and 413, as well as Biweekly Contest 138. In each contest, o1-preview is given three submission attempts per problem. A problem was considered successfully solved if any of the submissions passed Leetcode's automated system judgement, which evaluates the correctness and efficiency of the code against predefined test cases. This evaluation framework allows for a detailed assessment of o1-preview's ability to reason through complex algorithmic challenges and produce accurate solutions under typical coding contest conditions.

\subsection{Radiology Report Generation}
\label{subsec:Radiology-Report-Generation}

OpenAI's next-generation large language model, o1-preview, has exhibited considerable potential in medical report generation. To evaluate its capabilities in this domain, we performed an assessment using the Chinese radiology report dataset SXY~\cite{zhong2023chatradiovaluerchatlargelanguage,zhang2024potentialmultimodallargelanguage} to test o1-preview's effectiveness in generating medical reports.

The SXY dataset, sourced from The Second Xiangya Hospital, Central South University, is a private Chinese radiology report dataset designed for training and evaluating models in radiology report generation. It includes 317,339 radiological reports from five categories (94,097 chest reports, 64,550 abdominal reports, 46,092 musculoskeletal reports, 69,902 head reports, and 42,698 maxillofacial \& neck reports). These reports offer comprehensive documentation of patients' radiological imaging analyses, encompassing a wide spectrum of pathological findings and providing diverse opportunities for model training and evaluation. Leveraging the SXY dataset, researchers can assess a model's performance in producing accurate, clinically relevant radiology reports, particularly its proficiency in understanding and generating complex medical terminology. This dataset serves as a valuable benchmark for radiology report generation tasks, spanning various medical imaging domains, and provides an ideal resource for validating model performance across different clinical scenarios.

To assess the performance of o1-preview, we randomly selected 10 radiology reports from the SXY dataset for evaluation. During the testing process, we conducted several experiments to determine the optimal prompt phrasing, ensuring that the prompts remained consistent across all trials. For the model-generated outputs, we employed the ROUGE metric to measure the degree of correspondence between the radiology reports produced by the large language models and the reference reports authored by medical professionals. Specifically, this study utilized three evaluation metrics: ROUGE-1 (R-1), ROUGE-2 (R-2), and ROUGE-L (R-L), as defined in \textbf{Eq.~\eqref{eq1}}.

\begin{equation} \label{eq1}
ROUGE-N=\frac{\sum_{S\in\lbrace ReferenceSummaries}{\sum_{{gram}_n\in S}{Count_{match}(gram_n)}}}{\sum_{S\in\lbrace ReferenceSummaries}{\sum_{{gram}_n\in S}{Count(gram_n)}}}
\end{equation}

\subsection{Robot Command Planning}
\label{subsec:Robot-Command-Planning}
o1-preview can analyze real-time sensor data and adapts to dynamic environments, providing flexible, intelligent control solutions. Its ability to generate robot control commands and control code tailored to various robotic platforms reduces manual intervention, allowing developers to optimize control algorithms on the fly. o1-preview can potentially refine its strategies, enhancing autonomy and resilience across industrial, household, and autonomous vehicle applications.

In this section, we evaluate the performance of o1-preview on the ROS official code repository dataset, ROS~\cite{fernandez2015learning}, which contains the official ROS code's usage. The task involves analyzing code snippets and determining their functionality and correctness, structured as a classification task where the given code either performs as expected, contains logical errors, or has undefined behavior. Code understanding tasks require advanced technical comprehension and reasoning to identify functional correctness and are widely used to evaluate AI models in software engineering contexts~\cite{li2024aldm}. Some domain-specific code datasets not only demand sophisticated reasoning but also assess the model's understanding of domain-specific programming principles, providing a more robust evaluation of its potential for real-world development scenarios. For detailed analysis and results, please see Section \ref{Robot examples}.

\subsection{Nature Language Inference}
\label{subsec:Nature-Language-Inference}
In this section, we evaluate o1-preview on the natural language inference (NLI) task. The NLI task involves determining the logical relationship between two sentences, structured as a classification task where the second sentence either logically follows from the first, contradicts the first, or is neutral (possibly true). NLI tasks require advanced language understanding and reasoning to analyze logical coherence and are widely used to evaluate LLMs. Some domain-specific NLI datasets not only demand advanced reasoning but also assess the domain-specific knowledge of LLMs, providing a more comprehensive evaluation of their potential for real-world applications.

Here, we evaluate o1-preview using data samples from five NLI datasets: MNLI, ANLI, QNLI, MedNLI, and RadQNLI~\cite{williams2017broad, nie2019adversarial, wang2018glue, romanov2018lessons, wu2023exploring}. Table~\ref{nli_bmk} presents a summary of these datasets with detailed descriptions, covering different formats and domains, ensuring a thorough assessment of o1-preview's reasoning capabilities.

\begin{table}[ht]
\centering
\caption{Summary of NLI datasets used for o1-preview evaluation.}
\label{nli_bmk}
\small
\begin{tabularx}{1\textwidth}{@{}lccX@{}}
\toprule
Dataset & Size & Text Source & Description\\
\midrule
MNLI \cite{williams2017broad}  &433k & 10 distinct genres &Determines the relationship in sentence pairs as entailment, neutral, or contradiction. Includes a broad range of diverse sources, making it ideal for evaluating model generalizability across unseen text.\\
ANLI  \cite{nie2019adversarial}  &169k & Wikipedia &Employs an iterative and adversarial Human-And-Model-in-the-Loop Entailment Training process, in which annotators craft increasingly complex examples to challenge models across three rounds.\\
QNLI \cite{wang2018glue}  &115k & Wikipedia &Given question-context sentence pairs with high lexical overlap, determines whether the context sentence contains the answer to the question.\\
MedNLI \cite{romanov2018lessons}  &14k & MIMIC-III \cite{johnson2016mimic} &Follows the MNLI schema, but sourced from the medical domain, specifically the MIMIC-III dataset.\\
RadQNLI \cite{wu2023exploring}  &10k & MIMIC-CXR \cite{johnson2019mimic} & Adapts the QNLI schema to the radiology domain, using data from MIMIC-CXR.\\
\bottomrule
\end{tabularx}
\end{table}

\subsection{Quantitative Investing}
\label{subsec:Quantitative-Investing}
The stock-trading-QA \cite{yym2023stocktradingqa} dataset offers several distinct advantages that set it apart from other financial question-and-answer datasets. First and foremost, it provides deep insights into trading strategies, financial models, and market analysis techniques, which are essential components of quantitative trading. This dataset covers a wide range of topics, from statistical models used in market forecasting to the role of automation in real-time trading, as well as how fundamental analysis can be integrated with technical signals to create more robust trading strategies.

One key advantage of this dataset is its focus on quantitative investment, a crucial aspect of modern finance. While many other financial QA datasets may address broader economic questions or general investment advice, this dataset narrows its scope to topics specifically relevant to algorithmic and quantitative trading. This allows it to dive deeper into concepts such as ARIMA models, machine learning techniques for market prediction, event-driven trading strategies, and other technical aspects that are fundamental to quantitative finance. By emphasizing these specialized topics, the dataset is highly applicable to traders, quants, and researchers focused on developing automated trading systems.

Another significant strength of the stock-trading-QA dataset is its ability to highlight the reasoning and numerical computation skills of models, especially those designed for quantitative reasoning. Unlike standard instruction-following datasets, which primarily test a model's ability to comprehend and follow simple commands or guidelines, this dataset is structured to challenge models with complex, domain-specific queries that require logical reasoning, numerical understanding, and advanced financial knowledge. For example, questions about optimizing trading algorithms, statistical model selection, or implementing automation in trading systems require much more than rote instruction following—they demand quantitative problem-solving and a deep understanding of financial markets.

To rigorously test and measure the performance of models using this dataset, a comprehensive evaluation framework is employed. Models are assessed based on their ability to accurately solve quantitative problems, generate coherent and logically sound responses, and correctly apply advanced financial concepts. Key performance metrics include accuracy for classification tasks, mean squared error (MSE) and root mean squared error (RMSE) for regression analyses, and precision and recall for information retrieval tasks. Additionally, the F1-score is utilized to provide a balanced measure of a model's precision and recall capabilities. 

Beyond quantitative metrics, qualitative assessments are conducted to evaluate the models' reasoning processes. This involves analyzing the logical flow of their solutions, the correctness of numerical computations, and the appropriateness of the financial methodologies applied. Models are also benchmarked against established financial theories and real-world market data to ensure their outputs are not only theoretically sound but also practically relevant. Human expert evaluations further validate the models' performance by comparing their answers to those provided by experienced professionals in the field of quantitative finance.

This multifaceted evaluation approach ensures that the models are not merely performing superficial computations but are genuinely understanding and engaging with complex financial concepts. It allows researchers and practitioners to identify specific areas where models excel or need improvement, thereby facilitating the development of more advanced and reliable quantitative trading systems.

In summary, the stock-trading-QA dataset excels in offering insights into critical areas of quantitative finance and trading models. Its highly specialized focus on topics like statistical modeling, automation, and signal integration makes it a valuable resource for testing the abilities of models in a quantitative context. By emphasizing model reasoning and numerical computation over simple instruction-following tasks, this dataset provides a robust platform for assessing the performance of AI models in handling complex financial queries.

o1-preview's mathematical reasoning capabilities enable it to perform complex, multi-factor, real-time quantitative model analysis, going beyond mere memorization. o1-preview can dynamically analyze numerous market variables, adjusting predictions and strategies in response to evolving conditions. This adaptability is crucial in quantitative finance, where models must factor in diverse data points such as price trends, trading volumes, and macroeconomic indicators. o1-preview’s ability to integrate these factors in real time allows for the continuous refinement of models and decision-making processes. Additionally, its capacity for multi-dimensional analysis ensures that it can identify nuanced correlations and patterns that simpler, rule-based systems might overlook, enhancing the accuracy and robustness of its quantitative trading strategies.

\subsection{Low-Resource Language Translation}
\label{subsec:Low-Resource Language Translation}

Low-resource translation is particularly challenging for Transformer-based models, like the previous GPT-4o, due to their reliance on large amounts of high-quality training data to learn language patterns effectively. Our study will evaluate o1-preview model's ability to handle low-resource language translation by using data from the Cherokee Corpus Main Section of the Cherokee-English Dictionary (CED) project \cite{cherokeedictionary}. The Cherokee Corpus is a collection of 1776 Cherokee sentences paired with corresponding English translations as ground truth. This resource will serve as a valuable benchmark to evaluate how effectively the model can handle translation tasks between Cherokee and English, particularly in the context of limited parallel data availability for the Cherokee language.

Throughout the experimental phase, o1-preview generated translations and grammatical breakdowns for Cherokee sentences, focusing on both word meanings and sentence structure. The model was instructed to translate five Cherokee sentences and provide detailed analyses of key components, such as noun phrases and verb conjugations. Each analysis covered the subject, action, and contextual meaning of the sentence, demonstrating the model's ability to handle translations for a low-resource language.

The experimental results show that o1-preview can successfully translate common phrases and identify grammatical structures such as plural nouns and verb tenses in Cherokee. However, due to Cherokee being a low-resource language, the model sometimes fails to recognize certain words, leading to incomplete or inaccurate translations. Despite this, the model is able to provide reasonable guesses for unknown words, ensuring a degree of consistency in the overall translation. This ability to infer meaning allows the model to generate plausible translations, even when faced with unfamiliar vocabulary. However, expert intervention is still often required to refine these guesses and ensure full accuracy. This indicates the need for more detailed linguistic data and expert guidance to improve the model’s performance in low-resource language translation tasks.

\subsection{Educational Q\&A}
\label{subsec:Educational-Q-A}
In the field of educational science, our study utilized the SciQ dataset~\cite{welbl2017crowdsourcing}, consisting of 13,679 scientifically oriented questions across various disciplines, including physics, biology, chemistry, and earth sciences. We selected the SciQ dataset to specifically evaluate o1-preview's capabilities in understanding and reasoning about scientific education knowledge. The dataset, with its wide range of scientifically oriented questions, allows us to assess the model's ability to comprehend complex concepts, make logical inferences, and generate accurate, well-explained answers across various scientific disciplines such as physics, biology, chemistry, and earth sciences. 

The results demonstrated that o1-preview performed exceptionally well on this dataset, showcasing its robust ability to select the correct answer despite the presence of distractor options. The model exhibited a strong understanding of key scientific concepts and was able to navigate through misleading distractors to consistently choose the correct answer. This performance highlights o1-preview model's exceptional capacity for discerning relevant information and ignoring irrelevant or incorrect options, further underscoring its potential in educational applications.

In the field of education, o1-preview holds significant potential to transform the way students learn and teachers deliver instruction. By effectively understanding and reasoning through complex scientific concepts, o1-preview can assist educators in creating more personalized learning experiences, allowing students to receive tailored feedback and guidance. As technology continues to evolve, o1-preview could play a pivotal role in alleviating the workload on teachers, enabling them to focus on higher-level pedagogical tasks. Furthermore, it has the potential to contribute to a more balanced distribution of educational resources. With continued advancements, o1-preview is poised to become an indispensable tool in modern education, facilitating more efficient, scalable, and equitable learning environments.

\subsection{Student Writing Improvement in Higher Education}
\label{subsec: slat}

This section aims to evaluate o1-preview’s potential to enhance student writing in higher education. Effective writing in higher education requires appropriate language conventions, coherent structure, rhetorical awareness, etc, which can be supported by o1-preview's advanced language capabilities. While current LLMs predominantly focus on grammar checking, we offer a more comprehensive assessment, covering linguistic accuracy, coherence, outline generation, citation management, and creativity/personalization. The student writing samples for this evaluation are sourced from Corpus \& Repository of Writing (CROW) \cite{Staples2018}, a large-scale collection of student drafts from various writing courses at three universities in the US, providing diverse scenarios and levels for a thorough analysis of o1-preview's capabilities.

\subsection{3D Layout Generation}
\label{subsec:3D Layout Generation}
In the domain of computer vision and 3D scene understanding, our study utilized the 3D-FRONT dataset~\cite{fu20213d}, which comprises a large collection of high-quality 3D indoor scenes with detailed room layouts and furniture arrangements. We selected the 3D-FRONT dataset to specifically evaluate o1-preview's capabilities in generating realistic and functional 3D room layouts. The dataset, with its extensive variety of room types, objects, and spatial configurations, allows us to assess the model's ability to comprehend complex spatial relationships, adhere to design principles, and produce layouts that are both aesthetically pleasing and functionally sound.

The results demonstrated that o1-preview performed exceptionally well on this dataset, showcasing its robust ability to generate coherent and realistic 3D layouts despite the complexity of the scenes. The model exhibited a strong understanding of spatial constraints and design guidelines, effectively placing objects within rooms while avoiding overlaps and ensuring accessibility. This performance highlights o1-preview's exceptional capacity for spatial reasoning and adherence to design constraints, further underscoring its potential in applications such as interior design and virtual environment creation.

In the field of 3D layout generation, the o1-preview holds significant potential to transform the way virtual spaces are designed and visualized. By effectively understanding spatial relationships and design principles, o1-preview can assist designers in creating more efficient and appealing layouts, allowing for rapid prototyping and customization. As technology continues to evolve, o1-preview could play a pivotal role in automating aspects of interior design, enabling professionals to focus on more creative and complex tasks. Furthermore, it has the potential to contribute to more immersive virtual environments in gaming and virtual reality applications. With continued advancements, o1-preview is poised to become an indispensable tool in the field of 3D layout generation, facilitating more efficient, scalable, and high-quality spatial design solutions.

\subsection{Chip Design}
\label{subsec:Chip Design}
The intersection of LLMs and MLLMs with chip design is poised to transform the semiconductor industry, offering capabilities that surpass traditional methods in efficiency, precision, and scalability. In a field where the margins between success and failure are razor-thin, the introduction of LLMs into chip design holds profound significance. The complex workflows, intricate trade-offs, and multi-dimensional challenges inherent to chip design make it an ideal candidate for AI-driven innovation. LLMs and MLLMs, with their ability to process vast amounts of data, perform high-level reasoning, and optimize processes, are uniquely positioned to revolutionize this domain.

\subsubsection{Transformative Potential of LLMs in Chip Design}
The chip design process is a delicate balance of performance optimization, power efficiency, and manufacturability, requiring advanced techniques to manage the trade-offs between these factors. Today’s semiconductor industry is tasked with creating ever-smaller, more efficient chips, all while maintaining lower costs and faster time-to-market. Traditional engineering tools and processes, while highly sophisticated, often fall short in providing the speed and depth of analysis required to stay ahead in this competitive environment. Herein lies the transformative potential of LLMs and MLLMs.

LLMs can rapidly process and analyze vast datasets, including prior chip designs, performance reports, and error logs, generating insights that might elude even the most experienced human engineers. This not only accelerates the design process but also improves its outcomes by ensuring more optimal circuit layouts, better power management, and early error detection. Moreover, MLLMs, with their ability to process multiple data forms—text, images, and simulations—offer a broader range of applications, enabling AI to assist across the full spectrum of chip design and manufacturing.

One particularly powerful application of LLMs lies in error prediction and mitigation. By analyzing historical manufacturing data, these models can identify patterns and foresee potential design flaws long before they reach the fabrication stage, reducing costs associated with defective chips. Furthermore, LLMs can also optimize the logistical supply chains in semiconductor production, minimizing material delays and ensuring timely production cycles, which are critical in an industry where time-to-market is a key differentiator.

\subsubsection{Experimenting with o1-preview in Chip Design}
As part of our exploration into the capabilities of LLMs, we are testing o1-preview in the domain of chip design. Given the complexity of this task and the absence of standardized chip design datasets, we draw inspiration from approaches like those found in ChipNeMo~\cite{liu2023chipnemo}, which pioneered the adaptation of LLMs for chip design. However, our evaluation assess the o1-preview’s versatility and efficacy in this domain.

The test will focus on three critical areas: Engineering Assistant Chatbot, EDA Script Generation, and Bug Summary and Analysis. These tasks represent key stages in the chip design and manufacturing process and serve as proxies to gauge the model’s ability to address complex, real-world engineering challenges.

The first task, Engineering Assistant Chatbot, will evaluate the o1-preview’s ability to serve as a technical consultant for engineers, providing expert guidance on complex design questions. This test will explore whether the model can synthesize information from technical documentation, past designs, and performance metrics to deliver high-quality responses in real time. In chip design, where decisions need to be made quickly and with high precision, such a tool could significantly enhance the productivity of engineers by reducing time spent on research and troubleshooting.

In the second task, EDA Script Generation, the o1-preview will generate code for Electronic Design Automation (EDA) tools based on specific design requirements. EDA tools (i.e., specifically Verilog in this experiment) are essential for automating various steps in the design and verification of electronic systems, and generating efficient code is critical for ensuring that chips meet performance and power requirements. This phase of the experiment will evaluate the quality and efficiency of the generated scripts, testing them in real-world scenarios to compare the performance of AI-generated scripts against human-designed ones.

Finally, the Bug Summary and Analysis task will focus on the model’s ability to analyze error logs and simulation data, summarizing the root causes of design errors and suggesting potential solutions. Chip design is highly susceptible to errors, especially at the simulation stage, and early detection and resolution of bugs can save significant time and resources. By testing o1-preview’s ability to automate this process, we aim to evaluate whether the model can reduce the burden on human engineers and provide clearer, more actionable bug reports.

\subsubsection{A Leap Toward AGI and the Future of Chip Design}
The success of o1-preview model in chip design could signify a major breakthrough, not only for semiconductor engineering but also for the pursuit of AGI. o1-preview’s advanced reasoning capabilities and ability to handle complex, multi-step workflows mark it as a potential game-changer in the development of AI systems that can operate with human-level understanding in specialized domains. Chip design, with its demanding balance of technical precision and high-stakes decision-making, serves as an ideal proving ground for the model’s capabilities.

What distinguishes o1-preview from earlier models is its enhanced ability to engage with intricate, domain-specific tasks, such as circuit optimization, bug analysis, and EDA script generation. These are not just routine engineering problems; they require deep understanding, real-time analysis, and adaptive problem-solving—skills that align closely with the core requirements for AGI. By successfully performing these tasks, o1-preview demonstrates its potential to transcend narrow AI applications and begin tackling more generalized, high-level intellectual challenges.

Furthermore, if o1-preview can meaningfully enhance chip design processes, reducing both time and error rates while optimizing overall design performance, it will not only revolutionize the semiconductor industry but also signal a profound step forward in AI development. This success would show that o1-preview is capable of the kind of holistic problem-solving and adaptive reasoning that is critical for the realization of AGI. Unlike traditional LLMs, which often struggle with highly specialized or technical tasks, o1-preview is proving itself capable of functioning in a variety of high-complexity domains, each with its own unique set of challenges.

By bridging the gap between task-specific expertise and general-purpose reasoning, o1-preview is laying the groundwork for a new generation of AI systems capable of solving a wide range of problems across diverse fields. Its potential success in chip design would be an important milestone on the path to achieving AGI, demonstrating that AI can not only learn from vast datasets but also apply that knowledge in novel, practical, and high-stakes scenarios.

In conclusion, The introduction of o1-preview into the field of chip design marks a significant step forward in AI’s potential to revolutionize this highly technical domain. By successfully handling critical tasks such as circuit layout optimization, EDA script generation, and bug summary analysis, o1-preview showcases its ability to manage the intricate balance of performance, power, and manufacturability that defines chip design.

More importantly, o1-preview’s capabilities represent a major advancement toward the realization of AGI. Unlike earlier LLMs that were confined to narrow tasks, o1-preview’s ability to reason through complex, technical workflows in chip design points to its potential as a general-purpose problem-solver. As it continues to demonstrate success in high-stakes, specialized environments, o1-preview is paving the way for future AI systems capable of tackling diverse and complex intellectual challenges across industries. This success will not only transform the semiconductor industry but also bring us closer to the era of AGI, where AI systems can perform at a human-like level across a wide range of tasks and domains.

\subsection{Logical Reasoning}
\label{subsec:Logical Reasoning}
We attempted to explore the performance of o1-preview in the field of logical reasoning. We defined five types of logical reasoning for manual instance classification, namely categorical reasoning, sufficient condition reasoning, necessary condition reasoning, disjunctive reasoning, and conjunction reasoning. o1-preview holds numerous application advantages in the field of logical reasoning. Firstly, it can efficiently handle a large amount of data. Whether it is complex text, images, or other forms of data, it can conduct rapid analysis in an extremely short time. Moreover, it has parallel processing capabilities and can simultaneously handle multiple logical reasoning tasks, greatly enhancing work efficiency. Secondly, it has high accuracy. Unaffected by human emotions, fatigue, and subjective biases, it always maintains objective and accurate reasoning. Based on sophisticated algorithms and extensive data analysis techniques, it can establish an accurate logical model to ensure reliable reasoning results. Furthermore, it has repeatability and consistency. For the same logical problem and input data, it can generate consistent reasoning results every time. The quality is stable. No matter when and where it is used, it can maintain a high-level reasoning ability. Additionally, it has strong learning and adaptability. Through continuous improvement and optimization based on new data and feedback, it can refine the logical reasoning model, adapt to different problem types and changing situations, and can also be applied in multiple fields by learning knowledge and logical rules in different fields. Finally, it can assist human decision-making, provide objective logical analysis and suggestions for human decision-makers, help make wiser decisions, and at the same time undertake some repetitive and cumbersome logical reasoning work, allowing humans to focus on more creative and strategic tasks. 

Multiple choice questions from the \href{https://github.com/lgw863/LogiQA-dataset}{LogiQA dataset}~\cite{Lgw} were used for testing because logical reasoning problems do not rely heavily on complex contexts. This dataset consists of 8,678 QA instances.(Train:7376; Eval:651; Test:651)The files is divided into English version: Train, Eval, Test., and Chinese version: zh\_train, zh\_eval, zh\_test.Each 8 lines constitute an example of a problem. (8,678 * 8 = 69,424) 
 
 Moreover, in the form of multiple choice questions, the average paragraph length of each sample is relatively small, and correct answers are provided. It is reported that the LogiQA dataset collects questions about logical understanding from public questions in the Chinese civil service examination, aiming to test candidates' critical thinking and problem-solving abilities. These raw data collected from the official website form 8678 question - answer data sample pairs. 

The results show that o1-preview performs extremely well on this dataset, maintaining high accuracy even after multiple recalculations. When faced with interference factors in the options, it demonstrates strong anti-interference ability. Through repeated consideration of problem descriptions and attempts at different options, it always selects the correct answer in multiple tests, highlighting the model's overall excellent performance in logical reasoning instead of being limited to a certain independent type of logical reasoning problem.

\subsection{Table-to-Text Generation}
\label{subsec:Table-to-Text Generation}
In this section, we aim to evaluate the effectiveness of o1-preview in performing table-to-text generation tasks for medical datasets. Specifically, we utilize data samples from the Alzheimer's Disease Neuroimaging Initiative (ADNI) dataset~\cite{Petersen2010ADNI}. The objective is to determine how well the model can convert structured tabular data into comprehensive and coherent natural language descriptions, focusing on the medical domain where accuracy and clarity are paramount.

\begin{figure}[t]
\includegraphics[width=\columnwidth]
{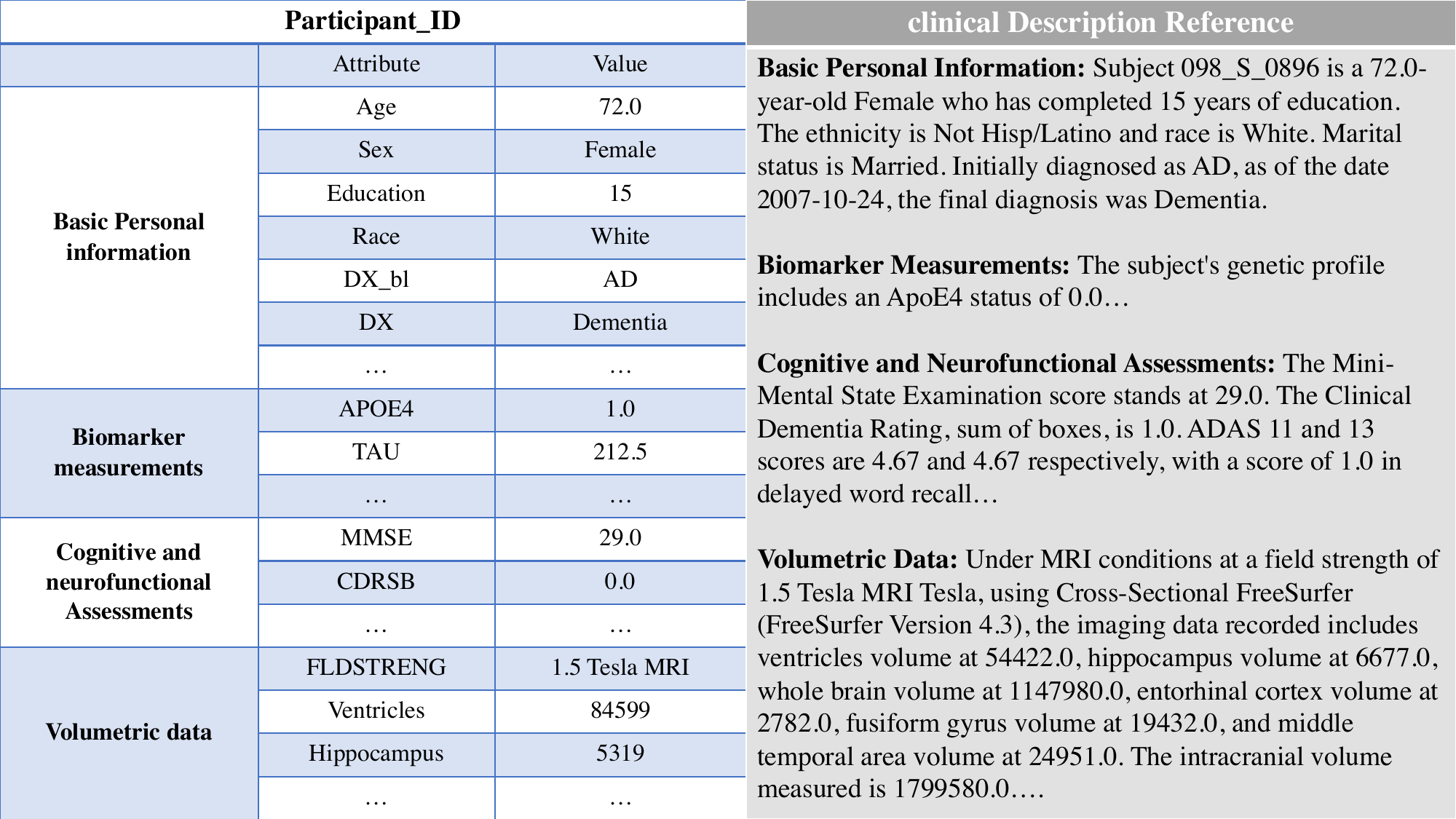}
  \caption{An example of a patient table and its corresponding clinical description.}
  \label{ExampleTable}
\end{figure}

The Alzheimer's Disease Neuroimaging Initiative (ADNI) is a large-scale, longitudinal study that was launched in 2004 with the primary aim of identifying biomarkers for the early detection and progression tracking of Alzheimer’s disease (AD). ADNI collects a wide range of data from participants, including clinical, imaging, genetic, and biomarker information from individuals with normal cognitive function, mild cognitive impairment (MCI), and Alzheimer’s disease. By leveraging neuroimaging techniques such as MRI and PET scans, ADNI seeks to deepen the understanding of the disease and support the development of new therapies. The collaborative nature of ADNI has enabled significant data sharing, which has greatly accelerated research in the field of Alzheimer’s disease. An example of a patient table and its corresponding clinical description is shown in Figure~\ref{ExampleTable}. In this work, we harness o1-preview to transform the ADNI dataset’s tabular data into fluent and accurate diagnostic reports, offering a powerful tool for medical professionals.

\subsection{High-School-Level Math Competition}
\label{subsec:High School Level Math Competition}
Recent efforts have extensively explored the math problem-solving abilities of the ChatGPT and its derivative models, with promising results achieved in multiple areas of mathematical reasoning and problem solving~\cite{10.1007/978-3-031-64779-6_45, app13106039, ai4040048, 10.1007/978-3-031-40113-8_8}. Some studies have even examined the potential for ChatGPT to serve as a tool for teaching and learning mathematics~\cite{wardat_chatgpt_2023}. With the key improvements in the o1-preview over previous versions of ChatGPT is its enhanced logical reasoning capability, we aim to assess o1-preview’s ability to solve mathematical problems with a specific focus on its reasoning skills. Mathematics, as a highly structured and logic-driven discipline, provides an ideal testing ground for evaluating this reasoning ability. To investigate o1-preview's performance, we designed a series of tests covering various difficulty levels. We begin with high school-level math competition problems in this section, followed by college-level mathematics problems in the next section, allowing us to observe the model’s logical reasoning across varying levels of complexity.

In this section, we selected two primary areas of mathematics: algebra and counting and probability in this section. We chose these two topics because of their heavy reliance on problem-solving skills and their frequent use in assessing logical and abstract thinking~\cite{hendrycksmath2021}. The dataset used in testing is from the MATH dataset~\cite{hendrycksmath2021}. The problems in the dataset cover a wide range of subjects, including Prealgebra, Intermediate Algebra, Algebra, Geometry, Counting and Probability, Number Theory, and Precalculus. Each problem is categorized based on difficulty, ranked from level 1 to 5, according to the Art of Problem Solving (AoPS). The dataset mainly comprises problems from various high school math competitions, including the American Mathematics Competitions (AMC) 10 and 12, as well as the American Invitational Mathematics Examination (AIME), and other similar contests. Each problem comes with detailed reference solutions, allowing for a comprehensive comparison of o1-preview's solutions. 

In addition to evaluating the final answers produced by o1-preview, our analysis delves into the step-by-step reasoning process of the o1-preview's solutions. By comparing o1-preview’s solutions with the dataset’s solutions, we assess its ability to engage in logical reasoning, handle abstract problem-solving tasks, and apply structured approaches to reach correct answers. This deeper analysis offers insights into o1-preview’s overall reasoning capabilities, using mathematics as a reliable indicator for logical and structured thought processes.


\subsection{College-level Math Problems}
\label{subsec:College level Math Problems}

We further investigate o1-preview's ability to solve college-level mathematics problems. Recently, numerous studies have focused on evaluating and enhancing the ability of LLMs to solve math problems \cite{ai4040048, hendrycksmath2021, li2024evaluatingmathematicalreasoninglarge, frieder2024mathematical, 10.1007/978-3-031-40113-8_8, imani2023mathpromptermathematicalreasoningusing, mishra2024mathcampsfinegrainedsynthesismathematical, poesia2024learningformalmathematicsintrinsic, doi:10.1098/rsta.2022.0044, heyueya2023solvingmathwordproblems, li2022lemmabootstrappinghighlevelmathematical, Poola23}. 
These models have shown potential in symbolic computation, mathematical reasoning and automated theorem proving. However, there remain significant challenges in evaluating the depth of their mathematical understanding, particularly when it comes to handling problems requiring long reasoning chains, abstract concepts, and intricate symbolic manipulation. Testing o1-preview's ability to solve such problems offers an opportunity to assess its capacity for mathematical reasoning and identify areas where it excels or struggles, providing valuable insights for future advancements in AI-driven mathematics education and problem-solving.

The problems presented in this section are manually created by the authors, drawing inspiration from various sources \cite{bilibili, berkeley, lang, AoPS}. They are particularly challenging due to their high level of abstraction and the complexity of reasoning required. These problems can be broadly categorized as follows: \begin{compactitem}
    \item Basic discrete mathematics problems.
    \item Advanced discrete mathematics problems.
    \item Calculus problems.
    \item Proofs of advanced theorems.
\end{compactitem}
We use discrete mathematics problems to evaluate o1-preview's ability to recognize discrete relations and patterns. Calculus problems are designed to test the model's comprehension of continuous concepts. Finally, by asking o1-preview to reproduce proofs of advanced mathematical theorems, we assess its capacity to manage abstract concepts and extended chains of reasoning.

Compared to the high school-level math problems tested in the previous subsection, college-level math problems typically involve higher levels of abstraction, more complex notations and longer reasoning chains.
By challenging o1-preview with these advanced mathematical problems and analyzing its outputs in details, we can gain a deeper understanding of its strengths and weaknesses in mathematical reasoning tasks. 
Although the problem set is relatively small, it provides a comprehensive representation of college-level mathematics, and the results reveal several intriguing aspects of o1-preview’s intrinsic problem-solving workflow.
Specifically, we empirically observed that o1-preview can easily solve basic discrete math problems. However, when confronted with more challenging problems, o1-preview tends to exhibit various types of mistakes.
See the detailed discussion in Section \ref{subsec:college math discussion}.


\subsection{Electronic Health Record Diagnosis}
\label{subsec:Electronic Health Record Diagnosis}
Electronic health records (EHRs) have become an integral component of modern healthcare, revolutionizing the way patient data is stored, accessed and used \cite{kruse2018use, hayrinen2008definition, li2024scoping}. EHRs serve as digital repositories that contain comprehensive patient medical histories, ranging from clinical notes to diagnostic images, making them invaluable resources to improve patient care and medical decision making. In this article, we explore the diagnostic capabilities of o1-preview by leveraging EHR data for patient diagnosis, aiming to assess how effectively o1-preview can interpret and analyze complex medical information.

We use the OHSUMED dataset (Download available online\footnote{\href{https://disi.unitn.it/moschitti/corpora/ohsumed-first-20000-docs.tar.gz}{https://disi.unitn.it/moschitti/corpora/
ohsumed-first-20000-docs.tar.gz}})~\cite{hersh1994ohsumed,hersh1994use} , a specialized subset of the MEDLINE database focused on biomedical literature, to evaluate the diagnostic capabilities of o1-preview. The OHSUMED dataset, curated by William Hersh and his team at Oregon Health Science University~\cite{tan2007large}, was designed for interactive retrieval evaluation and research on large text collections. OHSUMED consists of medical abstracts drawn from 270 medical journals over five years (1987-1991), comprising 13,929 documents. Each document is categorized into one of 23 cardiovascular disease categories, which presents a challenge for classification due to the large number of categories. To test o1-preview, we randomly selected 10 abstracts of biomedical texts related to different types of human cardiovascular disease from the MEDLINE database as our test cases.

By presenting each EHR text to o1-preview as a prompt and asking it to provide a diagnosis, the model has demonstrated the potential to support medical decision making. Notably, it tends to perform better with shorter texts, where diagnostic accuracy is higher. In longer and more complex records, while o1-preview still identifies key patterns, significant challenges remain in evaluating its performance. In addition, we observed that o1-preview not only provides diagnostic results but also offers reasoning based on the input text. This is evident in the model's ability to generate explanations and assess the relevance of the information to specific medical categories, demonstrating its capacity for inference. Testing o1-preview’s ability to analyze EHR texts provides a valuable opportunity to assess its diagnostic reasoning capabilities. By evaluating its ability to interpret complex medical data, identify relevant conditions, and deliver accurate diagnoses, we can gain crucial insights to advance AI-driven diagnostic tools and enhance their application in real-world healthcare settings.

\subsection{Stochastic Processes in Statistics}
Initial tests of o1-preview have already demonstrated promising results in mathematical domains.  However, to further explore its reasoning capabilities, it is essential to evaluate its performance in statistics, a field that requires more nuanced and complex decision-making.  By expanding testing into the domain of statistics, we aim to understand how well the model can handle probabilistic reasoning, which is crucial for applications in both theoretical and applied sciences.

The domain of stochastic processes was chosen for this evaluation because it is one of the most challenging areas within statistics.  Stochastic processes involve systems that evolve over time under the influence of randomness, making them ideal for testing a model’s ability to handle uncertainty, temporal dynamics, and predictive reasoning.  Problems in this area often require deep, step-by-step thought processes, demanding an advanced capacity for reasoning and understanding of random systems, which directly aligns with o1-preview's enhanced chain-of-thought capabilities.

For this study, we selected problems from the well-known textbook "Stochastic Processes" by Sheldon Ross (Download available online\footnote{\href{https://www.academia.edu/download/56739884/stochastic-processes-ross\_2.pdf}{https://www.academia.edu/download/56739884/stochastic-processes-ross\_2.pdf}})~\cite{ross1995stochastic}, a staple resource in senior undergraduate and graduate-level courses. The problems were chosen specifically for their complexity, covering key topics such as Markov chains, Poisson processes, and renewal theory. These problems require not only mathematical precision but also a deep conceptual understanding, providing a robust test of o1-preview's statistical reasoning abilities.

This assessment is critical because it evaluates o1-preview’s potential to solve complex, real-world problems where randomness and uncertainty play a central role. By comparing its performance against human-level reasoning in these advanced topics, we can better understand the practical applicability of the model in academic and professional contexts within statistics.

\subsection{Medical Text Anonymization}
\label{subsec: Medical Text Anonymization}

Most medical texts, such as clinical notes, are highly private because they contain a large amount of sensitive information \cite{zhang2024leveraging}. These texts are often highly technical, context-specific, and embedded with domain-specific terminology, requiring expertise for meaningful extraction. Due to the inherent challenges of completely anonymizing these texts, researchers encounter significant obstacles in accessing valuable medical text resources, all while adhering to strict ethical rules. Leveraging LLMs as powerful automatic tools for anonymizing medical texts to expand medical data resources has become increasingly urgent in current medical LLM development.

In this section, we utilize the latest o1-preview to anonymize medical texts. We use the 2014 i2b2/UTHealth de-identification challenge dataset~\cite{stubbs2015annotating}, for which a data use agreement must be signed to gain access to the dataset. This dataset has been annotated to highlight privacy features, including names, professions, locations, ages, dates, contracts, and IDs. We have implemented scripts to extract information from  XML files and store them as the model's inputs in text files. 

The objective is to employ o1-preview to detect and remove all privacy-related information from the given content. Three prompt styles, ranging from coarse to fine, have been designed to evaluate o1-preview's anonymization capabilities at different levels~\cite{liu2023deid}. The diversity of the annotated dataset helps avoid bias, and varying prompt levels ensure a comprehensive assessment of the model's performance. Based on our testing, o1-preview demonstrates an efficient ability to identify privacy features. Although performance may vary depending on the prompt style, the results remain robust, with most privacy information accurately detected and removed.

\subsection{Social Media Analysis}
\label{subsec:Social Media Analysis}
Social media analysis plays a critical role in modern business and societal contexts by providing valuable insights into public opinion, market trends, and consumer behavior. 
With the rise of platforms like Twitter, Instagram, and Facebook, individuals and organizations generate vast amounts of data that can be analyzed to understand patterns, sentiment, and influence. 
Testing the capability of large language models for social media analysis is essential for improving their ability to interpret complex human language, detect trends, and understand public sentiment at scale~\cite{rathje2024gpt}.
It also plays a key role in detecting and addressing biases in social media content, ensuring more balanced and fair analyses while driving the automation of key tasks like sentiment analysis and trend forecasting.

In this section, we will test the performance of o1-preview on four tasks in the field of social media analysis based on existing benchmarks~\cite{barbieri2020tweeteval}: \textit{sentiment analysis}, \textit{irony detection}, \textit{emotion recognition}, and \textit{offensive language identification}.
By testing o1-preview's performance on social media analysis data, we were able to further explore its potential for text analysis in fields such as psychology and the social sciences.

For sentiment analysis, we use a dataset from SemEval-2017 Task 4~\cite{rosenthal-etal-2017-semeval}, which contains tweets from Twitter in English and Arabic. The task requires predicting the sentiment of tweets in three categories: \textit{positive}, \textit{neutral}, and \textit{negative}.
Irony detection is performed on the Semeval-2018 dataset~\cite{van2018semeval}. 
note that this dataset includes two subtasks: subtask A is a binary irony detection task, while subtask B is a multi-class irony classification task that identifies whether the text contains a specific type of irony or no irony. And  we focus solely on subtask A in this paper.
In short, we define this task as a binary classification task, that is, we need to predict whether a tweet is ironic or not.
We use the most participated SemEval-2018, "Affect in Tweets"~\cite{mohammad-etal-2018-semeval}, to analyze the emotion recognition ability of o1-preview.
Given the limited number of tweets with a single label, we selected the four most common emotions—\textit{anger}, \textit{joy}, \textit{sadness}, and \textit{optimism}—as our labels.
As for offensive language identification, the task involves identifying whether a tweet contains some form of offensive language. Our test relies on the SemEval-2019 OffensEval dataset~\cite{zampieri-etal-2019-semeval}.
The above datasets are all reorganized by~\cite{barbieri2020tweeteval} (Publicly available\footnote{\href{https://github.com/cardiffnlp/tweeteval}{https://github.com/cardiffnlp/tweeteval}}).

\subsection{Analogical Reasoning}
\label{subsec:Analogical Reasoning}
Recent studies have shown that LLMs can match human performance across a range of analogical reasoning tasks, particularly those involving semantic structure mapping \cite{webb2023emergent, hu2023context, mitchell2021abstraction, lewis2024using}. This capability is not limited to abstract symbol manipulation, but also includes semantically meaningful symbols, thus providing a more nuanced understanding of analogical reasoning. To evaluate the analogical reasoning ability of o1-preview, we tested its performance on a dataset specifically designed for evaluating semantic structure mapping ability. This dataset comprises 13 distinct task types and a total of 136 samples, and provides a comprehensive evaluation of the model's analogical reasoning capability \cite{musker2024semanticstructuremappingllmhuman}.

Each of the 13 task types in this dataset is uniquely challenging, ranging from simple relational analogies to more complex tasks involving multiple layers of semantic content and structure. The 136 samples were carefully selected to encompass a wide range of semantic relations, ensuring a comprehensive evaluation of the model's analogical reasoning ability.

To test o1-preview's analogical reasoning capability, we employed tasks that require the model to transfer semantic structure and content from one domain to another. These tasks involve identifying and mapping semantic relations between words and phrases in a manner similar to human cognition and language acquisition. The goal of this evaluation is to determine whether o1-preview can perform analogical reasoning in a human-like manner by accurately aligning the relational structure between the source and target domains.

\subsection{Sentiment Analysis}
\label{subsec:Sentiment Analysis}
Recent studies have demonstrated that LLMs can achieve human-level performance in various sentiment analysis tasks. This capability extends beyond simple sentiment classification to include nuanced understanding of contextual and aspect-based sentiments, providing a more comprehensive perspective on natural language processing. To evaluate the sentiment analysis ability of o1-preview, we tested its performance on three benchmark datasets: IMDB \cite{maas2011learning}, SemEval-2014 Task 4 \cite{pontiki-etal-2014-semeval}, and SemEval-2015 Task 12 \cite{pontiki-etal-2015-semeval}. These datasets collectively offer a diverse and thorough assessment of the model's capabilities in sentiment analysis across different domains and text types.

The IMDB dataset comprises 50,000 movie reviews divided evenly between positive and negative sentiments. It is widely used for binary sentiment classification tasks, challenging models to accurately interpret sentiment in lengthy and varied user-generated content. This dataset tests the model's ability to handle informal language, sarcasm, and nuanced opinions expressed in movie reviews.

SemEval-2014 Task 4 focuses on aspect-based sentiment analysis in the domains of restaurants and laptops. It requires models not only to determine the overall sentiment but also to identify sentiments toward specific aspects within the text, such as service quality or battery life. This provides a more granular evaluation of the model's ability to understand and analyze sentiments at the aspect level.

SemEval-2015 Task 12 extends the challenge by including sentiment analysis in tweets. This dataset presents unique difficulties due to the informal language, use of slang, abbreviations, and character limitations inherent in social media content. It tests the model's robustness in handling noisy and brief text snippets, which is essential for real-world applications where data is unstructured and varied.

To assess o1-preview's sentiment analysis capabilities, we employed tasks that require the model to classify sentiment polarity, identify aspect-specific sentiments, and interpret informal and condensed language. These tasks involve understanding and analyzing sentiments expressed in different contexts and formats, paralleling the complexity of human language understanding. The goal of this evaluation is to determine whether o1-preview can perform sentiment analysis in a human-like manner by accurately interpreting and classifying sentiments across diverse datasets and textual styles.

\subsection{Anthropology and Geology}
\label{subsec:Anthropology and Geology}

LLMs have demonstrated significant reasoning capabilities~\cite{guo2023leveraging}  and a wealth of accumulated knowledge~\cite{ruixue2023insights}, enabling it to address questions across various disciplines. However, certain limitations persist, such as providing overly generalized answers, lacking specificity, and ambiguity in key areas~\cite{10290719}. To evaluate the performance of o1-preview, we posed questions related to paleoanthropology and geology. This was done to assess its understanding of these relatively niche disciplines.

Given the scarcity of test sets in the fields of geology and paleoanthropology, and to prevent the inadvertent reuse of test sets that may have been utilized during the training phase, we collaborated with experts in these areas to create new, specialized questions. Although the final test consists of only five questions, each is crafted with considerable difficulty, aiming to evaluate o1-preview's capabilities from multiple perspectives. Following this, the experts carefully assessed the quality of o1-preview's responses. This approach offers a more comprehensive reflection of o1-preview's reasoning abilities.

In the field of paleoanthropology (Figure~\ref{fig:paleo_geo_1}), we tasked o1-preview with answering questions related to ancient human genetics, including issues such as genetic contamination, genetic changes over time, and exploratory questions. For example, we posed a question about how discovering ancient human hair could be used to study various aspects of their civilization, living environment, and migration patterns. o1-preview's responses closely resembled those of an industry expert, progressing from broad conceptual explanations to more specific and in-depth answers. The model also demonstrated the ability to refine its responses based on interactive input, continuously deepening its understanding of the questions and providing increasingly precise and professional answers. This capability is remarkably similar to engaging with a domain expert, as it handles questions of varying complexity with ease, showcasing an impressive level of expertise.

To prevent the o1-preview model from having already seen the geology dataset during training, we had experts create new questions for o1-preview to answer. The experts then evaluated these responses to test the model's true capabilities.In the field of geology (Figure~\ref{fig:paleo_geo_7}), we examined o1-preview's ability to respond to true/false questions by describing a particular rock phenomenon and asking about its origin. Not only did o1-preview correctly identify the cause of the phenomenon, but it also provided supporting evidence, analyzing the observed features in a manner akin to that of a geologist. The model demonstrated the capacity to approach such questions with expert-level insight, engaging in in-depth discussions on topics like geology or paleoanthropology, much like a specialist in the field. This ability to reason and converse on complex scientific matters showcases its potential for facilitating expert-level discourse.

\subsection{Educational Measurement and Psychometrics}
\label{subsec: Educational Measurement and Psychometrics}
Educational measurement and psychometrics refer to the science and practice of developing the most appropriate test to accurately measure human characteristics, such as student knowledge, skills, abilities, and interests~\cite{bandalos2018measurement, kubiszyn2024educational}. Since these characteristics cannot be directly measured, researchers often rely on samples of behaviors, such as test responses or observations, to make inferences about these characteristics. Specifically, measurement involves assigning numerical values to observed events based on predefined rules. For example, in a test, each correct response may be assigned one point, and the total score represents the sum of these points that serves as a measure of the knowledge of the student~\cite{kubiszyn2024educational}. Measurement in social sciences involves developing valid and reliable instruments, designing effective data collection procedures, analyzing and evaluating the quality of the data, and communicating the results to multiple stakeholders, such as educators, policymakers, students, and parents. The key goals of educational measurement are (1) to draw valid conclusions about the abilities, achievements, or attitudes of the students, (2) to assess progress towards educational objectives, and (3) to improve teaching and learning~\cite{kubiszyn2024educational}.

Therefore, educational measurement and psychometrics is an interdisciplinary field closely intertwined with education, psychology, and statistics~\cite{bandalos2018measurement}. However, it faces notable challenges, particularly due to limited research and development compared to fields like computer science. This scarcity is due in part to sluggish enrollment and slow growth in graduate education programs. From 2007 to 2016, the number of conferred doctoral degrees in areas like "educational evaluation and research," "educational statistics and research methods," and "educational assessment, testing, and measurement" ranged from a low of 128 in 2012 to a high of 157 in 2015~\cite{leventhal2021surveying}. This restricted academic growth highlights potential limitations in training datasets for large language models. As such, exploring the performance of LLMs like o1-preview in the context of educational measurement and psychometrics becomes increasingly valuable for the development of the field. Notably, the dataset we used for this task was selected from multiple representative quiz questions in an introduction-level course named \textit{Measurement Theory} based on Bandalos' textbook~\cite{bandalos2018measurement} for first-year graduate students. This dataset was built by content experts who had more than 10 years of research and teaching experience for a graduate-level course at James Madison University.

\subsection{Public Health Policy Analysis}
\label{subsec:Public Health Policy Analysis}
The intersection of public health and LLMs represents a unparalleled opportunity to enhance public health surveillance, healthcare administration, and health policy-making. In this section, we assess o1-preview’s ability to analyze and evaluate the Affordable Care Act (ACA), one of the most significant healthcare reforms in recent U.S. history.

Our focus is on the significant aspects of the ACA, such as the expansion of insurance coverage, improvements in access to care, and the broader public health impacts of these changes. The evaluation follows the Q\&A format, where the prompts derived from the article named The Affordable Care Act at 10 Years: Evaluating the Evidence and Navigating an Uncertain Future\cite{aca_evaluation_2021}. It is important to note that the dataset used for this task is limited in size and scope, consisting of key questions focused on ACA findings. Thus, our evaluation emphasizes depth of reasoning, factual accuracy, and consistency in how the model addresses complex health policy. For further evaluations, we suggest using larger datasets, such as the ACA state dataset\cite{aca_state_data_2024}, to provide more comprehensive training for the model.

The evaluation involved 10 distinct prompts, each addressing critical areas of the ACA’s public health impact, including insurance coverage expansion, Medicaid expansion, surgical outcomes, preventive services, and healthcare disparities. These prompts challenged o1-preview to demonstrate nuanced reasoning and provide answers based on real-world health outcomes. The generated responses compared against expert insights from the article to determine their accuracy, depth, and relevance.

Overall, this study determine how effectively o1-preview can process and reason about health policy, particularly its performance in analyzing the ACA. Although the current dataset size is limited, this evaluation offers insight into o1-preview’s great potential for policy analysis and its capability to support public health decision-making processes.

\subsection{Medical Genetics and Genomics Reasoning}
\label{subsec: Genetic-medical QA and genomics Reasoning}

Genetics and diseases are closely related, and understanding the cause-and-effect relationship between them is crucial for the biomedical field. From a view of the knowledge base, the Inferential information can be used to establish deep-learning frameworks to reason the mutual relationship between genes and proteins. Several studies utilize the Gene Ontology (GO) knowledgebase to design neural network architectures that simulate gene or protein interactions within cells~\cite{gene2023gene, yu2016translation, ma2018using}. At the same time, using the bio-text data along with a large language model approach to infer the relationship between genomics and biomedical entities has become more prevalent. For example, A precise GPT model can significantly aid genomics research by minimizing the time interdisciplinary researchers, who may have limited genomics expertise, spend searching for and retrieving information from reliable sources. GeneGPT~\cite{jin2024genegpt} use a language model as an agent to connect NCBI Web APIs, which use in-context learning and an augmented decoding algorithm to conduct automatic API calls. Recent work, GP-GPT~\cite{lyu2024gp} exhibits proficiency in accurately retrieving medical genetics information and executing common genomics analysis tasks, including information retrieval and relationship determination. 

In this section, we seek to evaluate the model's ability to reason through problems related to medical genetics and genomics. To assess the model's genomics and genetics medical reasoning ability, we designed a set of experiments that focused on genomics questions and answers, where the model was required to generate reasonable answers towards the pre-defined genomics questions. These questions are about the relationship predictions and corresponding explanations. The tasks were created to test various aspects of genomics reasoning using a dataset comprising 20 question-and-answer (QA) tasks from the GenTunring benchmarks dataset~\cite{hou2023geneturing}. GeneTuring is a comprehensive  QA database used to assess the performance of GPT models in genomics (Methods), which is based on bio-texts. It consists of twelve modules, encompassing a total of 600 question-answer pairs, which can be categorized into four main groups. These modules reflect tasks commonly encountered in genomics research. The experimental tasks were designed based on the GeneTuring benchmark. The delicate selected gene-disease-related QA terms from the GeneTuring database have been reviewed and significantly extended by manually adding the corresponding gene/disease information from OMIM~\cite{amberger2015omim}. One extended term is utilized to build one full testing instance, which contains the instruction prompt, the gene's description and the disease's description. The model needed to follow the prompt instructions to respond to the gene-disease questions, in the meantime, the model was asked to give inference details and explanations step by step provided with the extended information.

\subsection{Medical Knowledge Question Answer}
In this section, we evaluate o1-preview’s performance on the Medical Knowledge question answer (QA), Medical knowledge QA involves tackling real-world examination datasets from complex medical disciplines, including pharmacology, medicine, and surgery. These tasks demand a deep and comprehensive understanding of the field, requiring knowledge that is comparable to that of human experts. Such QA systems must not only grasp foundational medical concepts but also integrate interdisciplinary knowledge, clinical reasoning, and problem-solving skills, making them particularly challenging in medical AI development.

In this section, we will utilize the MedMCQA dataset~\cite{pal2022medmcqa} to assess o1-preview's capability in answering medical knowledge questions. MedMCQA is a newly developed, large-scale Multiple-Choice Question Answering (MCQA) dataset, specifically designed to tackle real-world medical entrance exam questions. The dataset comprises 194,000 high-quality multiple-choice questions (MCQs) from the medical domain, covering 2,400 healthcare topics and 21 distinct medical subjects, making it a robust and diverse benchmark for evaluation. Each question is accompanied by the correct answer(s), alternative options, and additional auxiliary data, most notably detailed explanations of the solutions. This rich dataset allows for an in-depth assessment of model performance in the medical field.

Specifically, we extracted 10 questions from the MedMCQA dataset, covering various medical knowledge areas such as Anatomy, Pathology, Pharmacology, and more. These questions are categorized into two levels of difficulty: easy and difficult. Easy questions typically consist of straightforward knowledge-based queries, while difficult questions require scenario-based inference (see Section \ref{subsec: Medical Knowledge Question Answer} for more details). All questions are multiple-choice and include relevant explanations, allowing us to directly assess the accuracy of o1-preview's responses and the validity of its explanations. We will analyze the model’s selected options and explanations to provide a systematic and comprehensive evaluation.

\subsection{Art Education}
\label{subsec:Art Education}
This study evaluates the performance of the o1-preview model in art education, focusing on its ability to assist art educators, contribute to curriculum theory, and support educational standards. By engaging with diverse prompts—including art lesson development, creative writing exercises, reflections on artistic processes, and explanations of key educational concepts—the research examines the model’s capacity to assist in both practical teaching strategies and deeper curriculum design.

We develop our testing data from a set of prompts and scenarios covering tasks such as curriculum development, lesson planning, artistic writing, and engagement with educational theories. The questions were structured around real-world classroom challenges and reflective exercises, such as designing a cardboard assemblage art activity \cite{Cardboard}, creating a dramatic monologue from the perspective of a non-human object \cite{ArtB-Based-Research}, and exploring high school students' identities under the influence of artist Nick Cave’s \textit{Soundsuits} \cite{persona}. Key aspects of educational theory, including William Pinar's concept of \textit{currere} \cite{pinar2019currere}, inclusive education, such as cross-disciplinary connections with disability studies \cite{derby2011disability}, were incorporated to test the model's ability to engage with both practical strategies and theoretical concepts.

Grounded in educational theory, particularly Pinar's \textit{currere}\cite{pinar2019currere}, the study explores how o1-preview can contribute to reflective learning by interpreting and explaining complex ideas within curriculum theory and educational practice. This involves defining concepts like \textit{currere} and inclusive education and adapting them to specific educational contexts, requiring the model to comprehend theoretical ideas and apply them in practical scenarios such as art activity planning and student engagement.

The research also examines the model's ability to support reflective practices, encouraging students and educators to consider how personal experiences shape their understanding of art and culture. It analyzes o1-preview's capacity to assist art educators in setting objectives, aligning lessons with educational standards, and engaging students creatively. Furthermore, the study assesses whether the model can provide coherent, contextually relevant responses that support comprehensive curriculum development in art education, including suggestions on representation, inclusivity, and interdisciplinary connections to improve instructional quality.

Overall, the study assesses o1-preview’s potential as a tool for art educators in guiding artistic writing, planning lessons, and fostering inclusive and reflective learning environments. It explores whether the model can enhance curricula by contributing to personalized, culturally responsive, and critically reflective educational practices.

\subsection{Content Summarization}
\label{subsec:Content Summarization}
Automatic summarization, one of the central problems in Natural Language Processing, is the process of creating a short, concise summary of a longer text document. The goal is to distill the most important information from the original text into a shorter version that retains the essence of the content. It obtains high requirements for language models in context relevance, text content understanding and text generation. 

In this section, we evaluate o1-preview’s performance on the Journalism Content Summarizing task, which focuses on summarizing news articles in one sentence. We use a dataset called \textit{XSum}, from the article named \textit{Don’t Give Me the Details, Just the Summary! Topic-Aware Convolutional Neural Networks for Extreme Summarization}~\cite{Narayan2018DontGM}. This dataset consists of BBC articles and accompanying one single sentence summary. It includes 226,711 articles ranging from 2010 to 2018 and covers a wide range of categories (including sports, technology, business, family, art, health, education, etc.)~\cite{Narayan2018DontGM}.

In this section, we design several considerable judgment points for testing o1-preview’s performance on textual content summarization. One is whether the summary's content is consistent with the original text. Another is whether the summary generated by the model is too detailed or too general. Moreover, the length of the summary is also considered in our research.

\section{Related Work}

\subsection{Foundation Models}

The Transformer architecture \cite{vaswani2017attention} has quickly become a foundational framework for both language and vision-based models. First introduced in the field of Natural Language Processing (NLP), it revolutionized the handling of long sequences through the use of a self-attention mechanism, setting new benchmarks in machine translation. This architectural breakthrough laid the groundwork for models like BERT and GPT \cite{devlin2018bert,radford2018improving}, which significantly improved performance across a wide range of NLP tasks. As datasets expanded and models grew more complex, Transformer-based models such as RoBERTa, GPT-2, and GPT-3 \cite{liu2019roberta,radford2019language,brown2020language} emerged, further pushing the limits of what was possible in natural language understanding.

OpenAI's ChatGPT and GPT-4 \cite{openaiIntroducingChatGPT, openai2023gpt4} have been particularly noteworthy, demonstrating impressive language comprehension and reasoning capabilities. These models, thanks to vast training data and techniques like Reinforcement Learning from Human Feedback (RLHF) \cite{ouyang2022training}, perform effectively on a variety of tasks without the need for specific fine-tuning. Meanwhile, open-source models such as LLaMA and Mistral \cite{touvron2023llama,jiang2023mistral} have also shown strong performance, contributing to ongoing research into fine-tuning large language models for specialized domains, including medicine, education, robotics, and more \cite{liu2023radiology,wang2024large}.

The success of Transformers in NLP has naturally extended to the realm of computer vision. The Vision Transformer (ViT) \cite{dosovitskiy2020image} adapted the architecture for image processing, and subsequent models like DeiT, Swin Transformer, and Masked Autoencoders (MAE) \cite{touvron2021training,liu2021swin,he2022masked} have driven further advancements in visual tasks. As with NLP, larger datasets and increased parameters have enabled the development of even more powerful models like ViT-22B \cite{dehghani2023scaling}, leading to significant improvements in image recognition and analysis.

\subsection{Prompt Engineering}
With the rapid advancement of artificial intelligence technology, particularly in the field of natural language processing, the applications of large language models, such as OpenAI’s GPT series models \cite{brown2020language}, are becoming increasingly widespread. These models have demonstrated significant capabilities in tasks such as text generation, translation, and dialogue systems. However, effectively guiding these models to generate outputs that meet expectations and avoiding erroneous or inappropriate content has become a pressing challenge. prompt engineering has emerged in response to this need. Its goal is to maximize the performance and practical value of LLMs by designing precise and effective input prompts that guide the models toward generating desired results.

The importance of prompt engineering is reflected on multiple levels. First, it serves as a crucial link between user needs and model capabilities. While LLMs possess powerful generative abilities, they may also produce content that deviates from the topic or contains bias \cite{sheng2021revealing}. Through carefully designed prompts, users can better control the direction and quality of the model’s outputs. Additionally, in the context of few-shot learning and zero-shot learning, prompt engineering can significantly enhance model performance \cite{gao2021making}. When large amounts of labeled data are lacking, well-crafted prompts can tap into the model’s latent knowledge, enabling it to perform satisfactorily on new tasks. Furthermore, prompt engineering helps reduce model biases and errors, improving the accuracy and reliability of generated content, thereby enhancing user trust in the model \cite{bender2021dangers}.

There are two main approaches to prompt engineering: manual design and automated optimization. Manual prompt design is the most intuitive and commonly used method. Researchers or practitioners craft appropriate prompts based on their understanding of task requirements and model characteristics to guide the model in generating the desired output \cite{radford2019language}. For instance, in a question-answering system, including clear questions and necessary contextual information in the prompt can improve the accuracy of the model’s responses. The advantage of this method lies in its flexibility, allowing designers to make adjustments according to specific situations. However, its disadvantage is the significant time and effort required, and the quality of the design heavily depends on the expertise and experience of the designer.

To improve efficiency and effectiveness, automated prompt optimization methods have gradually gained traction. Among these, Prompt Tuning \cite{lester2021power} and Prefix Tuning \cite{li2021prefix} are representative techniques. Prompt Tuning adds trainable prompt vectors before the model input, enabling fine control over model behavior with minimal parameter adjustment. Prefix Tuning, on the other hand, inserts trainable prefixes into the model’s hidden layers to influence its internal representations and generation process. The common feature of these methods is that they achieve excellent performance on specific tasks by fine-tuning only a small portion of the model’s parameters, avoiding the need for large-scale model retraining and saving computational resources.

Despite the significant results achieved in practice, prompt engineering still faces several challenges. First, there is the issue of interpretability and transparency. Automated prompts often involve high-dimensional continuous vectors, making it difficult to intuitively understand their specific impact on model outputs \cite{qin2021learning}. Secondly, the generalizability of prompts across different tasks and domains is limited. A prompt effective for a specific task may not be directly applicable to other tasks, requiring redesign or adjustment \cite{vu2021spot}. Moreover, controlling model bias and undesirable outputs, while avoiding the introduction of subjective or discriminatory content in prompts, is also an important area of research \cite{liang2022holistic}.

Looking ahead, prompt engineering is expected to integrate with other techniques, such as human-computer interaction and reinforcement learning, to further improve the efficiency and effectiveness of prompt design. At the same time, developing more intelligent prompt generation tools that leverage the model’s own capabilities to assist in prompt design is a promising direction. Additionally, strengthening the theoretical foundation of prompt engineering and gaining a deeper understanding of the relationship between prompts and model behavior will help elevate the level of research in this field. As a critical component in the application of large language models, prompt engineering holds significant theoretical and practical value. Through the careful design and optimization of prompts, models can be effectively guided to generate high-quality, expectation-aligned content that meets diverse task requirements. Although challenges remain, continued research and technological development will undoubtedly enable prompt engineering to play an increasingly important role, leading to new breakthroughs and innovations in the field of AI.

\subsection{Chain of Thought}
Chain-of-Thought (CoT) prompting is a powerful technique that enhances the reasoning capabilities of large language models. Specifically, CoT prompting enables models to solve complex problems by generating intermediate reasoning steps instead of directly providing the final answer. This approach is particularly effective for multi-step tasks, such as mathematical problem-solving and logical reasoning~\cite{wei2022chain}. CoT aims to overcome the limitations of traditional prompting methods. In traditional prompting, models are typically given input-output examples and are expected to produce a final answer directly. However, this method often falls short for tasks that require multi-step reasoning~\cite{rae2021scaling}. CoT addresses this issue by instructing the model to explicitly articulate each step in the reasoning process.

Following the initial work on CoT, subsequent research on annotation methods can be categorized into manual, automatic, and semi-automatic approaches~\cite{chu-etal-2024-navigate}. Manual CoT relies on human-crafted prompts to guide the reasoning process. For instance, the Prompting-based Arithmetic Learner (PAL) framework uses programming-language rationales to generate programs, manually constructing prompts for arithmetic tasks~\cite{gao2023pal}. On the other hand, automatic prompting techniques like zero-shot CoT and Auto-CoT generate reasoning steps automatically~\cite{kojima2022large, zhang2023automatic}. Some research has focused on semi-automatic methods that combine automatic generation with limited human supervision. For example, AutoMate CoT~\cite{shum-etal-2023-automatic} and BoostedPrompt~\cite{pitis2023boosted} use a small set of human-annotated rationales to bootstrap the process, enabling the generation of higher-quality reasoning chains even in the absence of extensive labeled data.

Beyond basic CoT prompting, various extensions have been developed to tackle specific tasks or challenges. For instance, techniques like Self-Refine~\cite{madaan2023selfrefine} and CoSelfRevisions~\cite{le2024codechain} allow models to revise or verify their reasoning processes as they generate solutions. Other advanced frameworks, such as Tree of Thought (ToT)\cite{yao2023tree} and Graph of Thought (GoT)\cite{besta2024graph}, further extend the CoT structure by exploring multiple reasoning paths or incorporating network-based structures.

\subsection{Multi-modal Large Language Models}
Recent years have witnessed remarkable advancements in multi-modal large language models (MLLMs), which have demonstrated unprecedented capabilities in understanding and generating content across various modalities. These models have significantly impacted numerous domains, including computer vision, natural language processing, and human-computer interaction.

The evolution of MLLMs can be traced back to early work on combining visual and textual information. \cite{lu2019vilbert} proposed ViLBERT, one of the first models to jointly process visual and linguistic inputs using a two-stream architecture. Building upon this foundation, \cite{li2020oscar} introduced Oscar, which utilized object tags as anchor points to align image regions with text tokens, significantly improving vision-language pre-training. These early efforts laid the groundwork for more sophisticated multi-modal models. A significant breakthrough came with the introduction of CLIP by \cite{radford2021learning}, which demonstrated the effectiveness of contrastive learning in aligning visual and textual representations at scale. This work not only achieved impressive zero-shot performance across various tasks but also paved the way for subsequent research into scaling multi-modal models to billions of parameters. Following this trend, \cite{alayrac2022flamingo} presented Flamingo, a visual language model capable of few-shot learning across a wide range of vision and language tasks. The model's architecture, combining a vision encoder with a language model through cross-attention layers, has become a blueprint for many subsequent MLLMs. As the field progressed, researchers began to focus on enhancing the cognitive abilities of MLLMs. In 2023, \cite{li2023cogvlm} introduced CogVLM, a large vision-language model that achieves strong performance across various vision-language tasks while maintaining high efficiency. The model's cognitive training approach enables it to handle complex reasoning tasks involving both visual and textual inputs, representing a significant step forward in multi-modal understanding. Concurrently, efforts were made to improve the versatility and applicability of MLLMs. \cite{zhang2023mplug} proposed mPLUG-Owl, an MLLM that demonstrates impressive capabilities in multi-modal dialogue and image editing. The model's unique architecture allows for efficient fine-tuning and adaptation to downstream tasks, making it particularly valuable for real-world applications. This work highlighted the potential of MLLMs to bridge the gap between research and practical implementation. More recently, \cite{chen2024mm1} introduced MM1, a multi-modal model that achieves state-of-the-art performance on a wide range of vision and language tasks. MM1's novel pre-training strategy, which incorporates large-scale web-crawled image-text pairs, enables the model to develop a deep understanding of the relationships between visual and textual content. This research underscores the importance of diverse and extensive training data in developing robust MLLMs. As the capabilities of MLLMs continue to expand, researchers have increasingly focused on addressing key challenges in the field. One such challenge is the need for more efficient architectures and compression techniques, given the growing size of these models. \cite{wang2023efficient} made significant strides in this direction by proposing a novel pruning method that substantially reduces the computational requirements of MLLMs without sacrificing performance. This work exemplifies the ongoing efforts to make MLLMs more accessible and deployable in resource-constrained environments. Another important trend in MLLM research is the emphasis on enhancing multi-modal reasoning capabilities. \cite{yang2024multimodal} contributed to this area by introducing a new benchmark for evaluating complex reasoning tasks that require the integration of visual and textual information. This work not only provides valuable tools for assessing MLLM performance but also highlights the increasing sophistication of tasks that these models are expected to handle. As MLLMs become more powerful and widely adopted, the research community has also turned its attention to addressing ethical concerns such as bias, fairness, and privacy. \cite{johnson2024ethical} made a significant contribution in this regard by proposing a framework for evaluating and mitigating biases in multi-modal models. This research underscores the importance of responsible AI development and the need to consider the societal implications of advancing MLLM technology.

In conclusion, the field of multi-modal large language models has seen rapid and multifaceted progress in recent years. From foundational architectures to state-of-the-art models with sophisticated reasoning capabilities, MLLMs have evolved to become powerful tools for understanding and generating content across multiple modalities. As research continues to advance, we can expect further improvements in model performance, efficiency, and ethical considerations. These developments promise to pave the way for more powerful, versatile, and responsible AI systems that can seamlessly integrate information from diverse sources and modalities.

\subsection{Fine-tuning Large Language Models} 

Fine-tuning in deep learning adapts pre-trained models to specific tasks by adjusting their parameters, offering significant advantages over training from scratch. This technique not only improves model performance while reducing training time and data requirements, but also allows for specialization in targeted domains. There are two main approaches: full parameter fine-tuning, which adjusts all model parameters, and parameter-efficient fine-tuning, which modifies only a subset. Fine-tuning is particularly beneficial for specialized tasks, enabling models to learn domain-specific vocabulary and contextual nuances crucial in fields like medicine or law. It enhances the model's ability to produce structured outputs consistently, adhere to specific guidelines, and consider ethical implications relevant to particular applications. Whether through full or parameter-efficient methods, fine-tuning is essential for tailoring large language models to specialized use cases, significantly improving their accuracy, reliability, and practical applicability in targeted domains while balancing performance and efficiency.

Full parameter fine-tuning~\cite{zhang2023instruction} is a comprehensive adaptation method for pre-trained models to new tasks. In this process, every layer and parameter of the model is updated and optimized, allowing the model to better learn the features of the new task and improve its performance. However, due to the large number of parameters to be adjusted, full parameter fine-tuning often requires significant computational resources and time. More crucially, it can lead to "catastrophic forgetting"~\cite{liu2024more}, where the model forgets the knowledge it has previously learned when learning new tasks. This phenomenon is particularly prominent when the model is large and the tasks are highly dissimilar. Therefore, full parameter fine-tuning is generally suitable for scenarios with sufficient data, abundant computational resources, and a strong correlation between the new and old tasks. To mitigate catastrophic forgetting, techniques such as parameter-efficient fine-tuning and knowledge distillation can be considered.

Parameter-efficient fine-tuning~\cite{han2024parameter} is a technique that adapts large pre-trained models to new tasks without significantly increasing the number of model parameters. The core idea is to achieve transfer learning across different tasks by fine-tuning specific parts of the model or introducing additional modules while maintaining model efficiency. Common parameter-efficient fine-tuning methods include partial fine-tuning, adapters, prompt tuning, prefix tuning, and LoRA~\cite{hu2021lora}. Partial fine-tuning only adjusts the top or a few layers of the model, preserving the general knowledge acquired during pre-training; adapters insert small adapter modules into each layer of the model and only train the parameters of these modules; prompt tuning guides the model to generate the desired output by adding specific prompts to the input text; prefix tuning adds learnable prefixes to the beginning of the input sequence and controls the model's output by adjusting these prefixes; LoRA performs low-rank decomposition of the model's attention matrices and only trains the decomposed matrices. The advantages of parameter-efficient fine-tuning include reduced computational cost, faster training, alleviated overfitting, and improved generalization. The choice of parameter-efficient fine-tuning method depends on the specific task, dataset, and available computational resources.

\subsection{Large Language Model Agent and Retrieval-Augmented Generation}

Artificial Intelligence agents are autonomous entities capable of perceiving their environment, reasoning about it, and executing actions to achieve specific goals. They have made significant advancements in various fields, including robotics, finance, and healthcare. In robotics, AI agents have enabled autonomous navigation and manipulation, exemplified by the development of self-driving vehicles in recent years \cite{badue2019selfdrivingcarssurvey}. Advancements in deep reinforcement learning have led to AI agents capable of complex decision-making in dynamic environments \cite{vinyals2019grandmaster}. These agents operate without continuous human intervention, leading to increased efficiency in tasks such as data analysis and real-time decision-making. They learn from interactions with their environment, allowing them to adapt to new situations and improve performance over time. The development of model-based reinforcement learning algorithms has been crucial in this regard \cite{wang2019benchmarking}. However, designing AI agents involves complex algorithms and substantial computational resources, which can be cost-prohibitive. Many AI agents are task-specific and struggle to generalize knowledge to different domains. Additionally, autonomous decision-making raises issues related to accountability, transparency, and potential biases encoded in the agents. Recent studies have highlighted that AI systems can perpetuate existing biases, leading to unfair outcomes in applications such as hiring and lending \cite{mehrabi2022surveybiasfairnessmachine}. In finance, AI agents are used in algorithmic trading, risk assessment, and fraud detection, improving efficiency and accuracy. Research has demonstrated that AI-driven high-frequency trading can influence market dynamics and liquidity \cite{bansal2021impact}. In healthcare, AI agents assist in patient monitoring, diagnostic support, and personalized treatment planning. Deep learning models have been developed for medical image analysis, aiding in the early detection of diseases such as cancer \cite{li2019deep}.

Retrieval-augmented generation (RAG) combines large language models with information retrieval systems to generate responses grounded in external data sources. This approach enhances the factual accuracy and relevance of generated content. By accessing up-to-date information during generation, RAG models produce more accurate and contextually relevant outputs, overcoming the limitations of static training data. Lewis et al. \cite{lewis2021retrievalaugmentedgenerationknowledgeintensivenlp} introduced the RAG model, which integrates a pre-trained language model with a neural retriever to improve performance on knowledge-intensive tasks. Subsequent research has built upon this framework to enhance the capabilities of language models. For example, Guu et al. \cite{guu2020realmretrievalaugmentedlanguagemodel} proposed REALM, which enables language models to retrieve and reason over large corpora. The effectiveness of RAG models heavily relies on the quality and reliability of the retrieved data. The retrieval process adds latency and requires additional computational resources. Integrating retrieval mechanisms with generative models introduces challenges in system design and optimization. Despite these challenges, RAG models have been successfully applied in various domains. In customer support, they power intelligent chatbots that provide accurate and context-specific assistance to users \cite{feng2021surveydataaugmentationapproaches}. In education, they generate personalized learning content by retrieving relevant materials tailored to individual learner needs \cite{essa2023personalized}.

Despite the advancements brought by AI agents and RAG, limitations persist. AI agents often lack the ability to generalize across tasks, and RAG models can be constrained by the quality of their data sources and computational demands. The new o1-preview model emerges as a response to these challenges, aiming to combine the autonomy of AI agents with the knowledge integration of RAG while mitigating their respective drawbacks. The o1-preview model is anticipated to enhance generalization by leveraging advanced learning architectures, enabling effective performance across diverse tasks without extensive retraining. It is designed to optimize performance by reducing computational overhead, allowing faster inference even when accessing external data. Additionally, it aims to improve reliability by incorporating robust mechanisms for data verification to ensure that outputs are accurate and trustworthy. The o1-preview model holds promise in advancing the field of AI by addressing the limitations of existing approaches and providing a more versatile and reliable framework for autonomous agents and knowledge integration.

\subsection{Large Language Models \& Reasoning}
In recent years, large language models have demonstrated remarkable capabilities in natural language understanding and generation. However, their proficiency in complex reasoning tasks remains limited. Challenges arise particularly when dealing with multi-step logical deductions, abstract reasoning, and the integration of knowledge across various domains. Enhancing the reasoning capabilities of LLMs has thus become a crucial objective in artificial intelligence research. Based on the training mechanisms and cognitive frameworks of LLMs, their deficiencies in reasoning are evident. Specifically, addressing intricate problems necessitates deliberate and meticulous contemplation. However, due to the limited guidance provided during the training phase, LLMs are often constrained by token-level, left-to-right decision-making processes during inference~\cite{bengio2019systematic}. When faced with complex reasoning challenges, humans typically employ diverse cognitive abilities and interact with tools, external knowledge, and environmental information to accomplish tasks. Replicating this level of cognitive function in LLMs remains a significant challenge~\cite{lake2017building}.

In response to these limitations, several approaches have been proposed to emulate human reasoning processes. One foundational method is the \textit{Chain-of-Thought} (CoT) prompting~\cite{wei2022chain}, which involves providing LLMs with examples that include detailed intermediate reasoning steps. By explicitly modeling the reasoning process, CoT enables LLMs to tackle arithmetic, commonsense reasoning, and problem-solving tasks more effectively. Building upon CoT, the \textit{Tree-of-Thought} (ToT) framework~\cite{yao2023tree} models the reasoning process as a search through a tree of possible thought sequences. This approach allows the model to explore multiple reasoning paths simultaneously, enhancing its ability to handle ambiguous or complex tasks by considering a broader range of possibilities. The \textit{Self-Consistency} technique~\cite{wang2022self} improves reasoning by generating multiple reasoning paths and selecting the most consistent answer across these paths. This method reduces the impact of reasoning errors in any single chain and improves overall answer accuracy.

Another significant advancement is the integration of external tools into the reasoning process. The \textit{ReAct} framework~\cite{yao2022react} combines reasoning and acting by enabling LLMs to interact with external environments and tools, such as calculators or knowledge bases. This integration allows LLMs to perform computations, retrieve up-to-date information, and validate reasoning steps, thereby enhancing factual accuracy and reducing hallucinations. By leveraging the strengths of both internal reasoning processes and external resources, ReAct represents a promising approach to bridging the gap between LLMs and human-like reasoning capabilities.

The evolution of reasoning methods in LLMs reflects a shift from implicit, data-driven learning to approaches that incorporate explicit reasoning processes. Early LLMs relied on large-scale pre-training, hoping that reasoning abilities would emerge from vast amounts of data. However, this proved insufficient for tasks requiring logical deductions and complex problem-solving. The introduction of prompt engineering techniques, such as CoT, marked a significant advancement by leveraging the models' capabilities through carefully designed prompts that guide the reasoning process. Subsequent methods, such as ToT and Self-Consistency, focused on exploring multiple reasoning paths and selecting the most consistent answers, further enhancing the models' ability to handle ambiguous or complex tasks.

The integration of external tools and environments, as exemplified by the ReAct framework, represents a crucial step towards emulating human-like reasoning in LLMs. By combining internal reasoning processes with the ability to interact with external resources, these approaches more closely mimic the diverse cognitive abilities employed by humans when faced with complex challenges. As research in this field continues to progress, we can expect further advancements in the development of LLMs that exhibit increasingly sophisticated reasoning capabilities, bringing us closer to the goal of achieving truly intelligent and versatile artificial intelligence systems.

\subsection{Reinforcement Learning with Human Feedback}
\label{rlhf_relatedwork}

Generating responses that satisfy human expectations is a critical challenge in the development of large language models. Reinforcement Learning from Human Feedback~\cite{christiano2017deep, ziegler2019fine} has emerged as a promising solution to this challenge, enabling the creation of some of the most widely deployed AI models to date.

The development of RLHF can be traced back to the work of Schulman et al. \cite{schulman2017proximal}, who proposed a novel reinforcement learning approach that indirectly learns reward functions based on human preferences. This method simplified human involvement by having humans qualitatively compare the quality of two trajectories, achieving good results in simulated robotics and Atari games.
In the same year, OpenAI introduced the Proximal Policy Optimization (PPO) algorithm \cite{christiano2017deep}, which has been widely applied in the RLHF stage of LLMs. PPO optimizes the policy function to maximize cumulative returns while maintaining similarity between new and old policies, effectively avoiding instability and high variance problems during training.

Stiennon et al. \cite{stiennon2020learning} further advanced RLHF by replacing automated evaluation methods with human preferences as the training objective. Starting from an initial model fine-tuned with Supervised Fine-Tuning (SFT), they collected samples through human comparisons, built a reward model, and then optimized using reinforcement learning strategies, significantly improving the quality of summaries.

A significant milestone was reached with InstructGPT \cite{ouyang2022training}, which combined RLHF with GPT-3. This approach fine-tuned GPT-3 through human feedback to improve the model's ability to follow user intent, achieving performance improvements while reducing the number of parameters.

In the context of RLHF, we define:
\begin{itemize}
\item \textbf{States ($s$)}: The current context or prompt.
\item \textbf{Actions ($a$)}: The model's generated responses.
\item \textbf{Reward ($r$)}: Feedback based on human preferences.
\item \textbf{Policy ($\pi$)}: The strategy the model uses to generate responses.
\end{itemize}
The standard RLHF fine-tuning process typically consists of three stages:
\begin{itemize}
\item \textbf{Supervised Fine-Tuning (SFT)}: This initial stage adapts the pre-trained language model to generate outputs that align with human-written examples of desired behavior.
\item \textbf{Reward Modeling (RM)}: A reward model is trained on pairs of model outputs, where humans have indicated a preference for one output over the other. This model learns to predict these preferences, enabling it to generate reward signals for new, unseen model outputs.
\item \textbf{Reinforcement Learning (RL)}: The language model is fine-tuned using reinforcement learning to maximize the rewards predicted by the reward model.
\end{itemize}

Within this framework, various algorithms have been developed to implement and optimize the reinforcement learning process. Two prominent approaches are Proximal Policy Optimization (PPO) and Direct Preference Optimization (DPO):

\textbf{Proximal Policy Optimization (PPO)} \cite{schulman2017proximal}:
\begin{itemize}
\item Implements the RM and RL stages separately.
\item Uses an explicit reward model trained on human preferences.
\item Updates the policy by optimizing a clipped surrogate objective:
\end{itemize}
\begin{equation}
L^{CLIP}(\theta) = \mathbb{E}_t[\min(r_t(\theta)\hat{A}_t, \text{clip}(r_t(\theta), 1-\epsilon, 1+\epsilon)\hat{A}_t)]
\end{equation}
where $r_t(\theta)$ is the probability ratio between the new and old policy, and $\hat{A}_t$ is the estimated advantage.

PPO's key contribution is its ability to perform stable policy updates while maintaining good sample efficiency. The "proximal" aspect of PPO is particularly important in the context of language models, as it helps maintain semantic consistency by ensuring that policy updates do not deviate too far from the previous policy.

\textbf{Direct Preference Optimization (DPO)} \cite{rafailov2024direct}:
\begin{itemize}
\item Combines the RM and RL stages into a single optimization process.
\item Bypasses the need for an explicit reward model.
\item Directly optimizes the policy based on preferred responses:
\end{itemize}
\begin{equation}
L^{DPO}(\theta) = \mathbb{E}{(x,y_w,y_l)\sim\mathcal{D}}[\log\sigma(\beta(r\theta(y_w|x) - r_\theta(y_l|x)))]
\end{equation}
where $y_w$ and $y_l$ are the winning and losing responses, respectively, and $r_\theta$ represents the policy's preference between responses.

DPO's key contribution is simplifying the RLHF process by eliminating the need for a separate reward model and policy optimization step, potentially leading to more efficient training.

The application of RLHF has expanded beyond text-only models to include multi-modal large models and text-to-image generation models. GPT-4 \cite{achiam2023gpt} demonstrated the effectiveness of RLHF in multi-modal contexts, while numerous works have leveraged RLHF to enhance model performance in text-to-image generation \cite{lee2023aligning, fan2024reinforcement, black2023training}, showing good results in meeting human preferences, mitigating biases, and improving text-image alignment.

\subsection{Evaluation Complex Reasoning Tasks}

o1-preview, a recent entrant in the pantheon of LLMs, has emerged as a paradigm shift in AI, particularly in its adeptness at tackling complex reasoning tasks. This model's innovative approach, which integrates a "chain-of-thought" mechanism \cite{wei2023chainofthoughtpromptingelicitsreasoning}, sets it apart from its predecessors by enabling a more nuanced and methodical problem-solving process. Unlike traditional LLMs that often excel in pattern recognition and natural language processing, o1-preview's iterative reasoning allows it to dissect complex problems, evaluate intermediate steps, and synthesize solutions with a level of precision and logical coherence that was previously unattainable. By explicitly modeling the reasoning process, o1-preview demonstrates a significant advancement in the field of AI, bringing us closer to achieving human-like cognitive capabilities in machines.

The o1-preview's capabilities are most pronounced in domains that demand sustained, multi-step reasoning. For instance, in scientific research, where the model is tasked with analyzing data, formulating hypotheses, and drawing conclusions, it has demonstrated an impressive ability to conduct detailed analyses that are both accurate and insightful. Similarly, in the realm of programming, the model's capacity to understand and manipulate syntactic structures has been instrumental in tasks such as code debugging and optimization. Its integrated fact-checking feature is particularly valuable in high-stakes domains like legal analysis and medical diagnostics, ensuring that the information it processes and the conclusions it draws are grounded in verified data. These capabilities highlight the potential of o1-preview to revolutionize various industries by automating complex decision-making processes and augmenting human expertise.

However, o1-preview is not without its limitations. In scenarios that require rapid responses or where the complexity of reasoning is not paramount, the model's meticulous approach can be a liability. The iterative nature of its reasoning process can lead to slower output times compared to models optimized for speed, such as GPT-4 \cite{achiam2023gpt}, in tasks that do not necessitate deep analytical capabilities. This trade-off between depth and speed is a critical consideration in applications where time sensitivity is paramount. It is essential to recognize these limitations and to deploy o1-preview in contexts where its strengths can be fully leveraged, while relying on alternative models for tasks that prioritize speed over depth.

The selection of evaluation domains for o1-preview was strategic, aiming to encompass a broad spectrum of complex reasoning tasks. These domains, ranging from computational mathematics to legal analysis, serve as a crucible for the model's capabilities, testing its adaptability and generalizability across different types of cognitive challenges \cite{liu2023holisticevaluationgpt4vbiomedical}. By subjecting o1-preview to these diverse and demanding tasks, we seek to gain a comprehensive understanding of its potential and its limitations, which is essential for guiding its future development and application. The insights gleaned from this evaluation will not only inform the refinement of o1-preview but also contribute to the broader field of AI research, providing valuable lessons for the development of future models.

The implications of this evaluation extend beyond academic interest into the practical realms of technology and industry \cite{liu2024understandingllmscomprehensiveoverview}. By delineating the strengths and weaknesses of o1-preview, we provide a roadmap for the integration of advanced AI into fields that stand to benefit from enhanced reasoning capabilities. The insights gleaned from this study will inform the development of future LLMs, potentially revolutionizing sectors such as finance, healthcare, and education by automating complex decision-making processes and augmenting human expertise. Furthermore, this work contributes to the broader goal of advancing artificial general intelligence, pushing the boundaries of what machines can achieve and how they can collaborate with humans to solve some of the most pressing challenges of our time.

In conclusion, the evaluation of o1-preview in complex reasoning tasks is a critical step towards harnessing the full potential of AI in transforming various sectors of society. Through a rigorous and comprehensive assessment, we aim to unlock new possibilities for innovation and progress, while also identifying areas for improvement and further research. The o1-preview's performance in these evaluations will not only inform its immediate applications but also shape the trajectory of AI development, ensuring that future systems are better equipped to tackle the complex challenges that define our era. As we continue to push the boundaries of what is possible with AI, the insights gained from studying models like o1-preview will be instrumental in guiding us towards a future where machines and humans can work together seamlessly to solve the most pressing problems facing our world.

\newpage
\section{Experiments and Observation}
\label{Exp}
\subsection{Test Procedure}




In the testing phase, we conducted a rigorous evaluation of the o1-preview model, deploying it across a diverse array of domain-specific tasks to comprehensively assess its performance and reasoning capabilities. The primary goal was to determine the model’s ability to generate coherent, contextually appropriate, and logically consistent responses when confronted with challenges across various specialized fields. This was achieved by using domain-specific prompts and comparing the model’s responses to pre-established benchmarks, thereby evaluating how effectively o1-preview could adapt to the nuanced demands of different domains.

Our testing approach followed a unified, systematic procedure designed to evaluate o1-preview’s performance across several fields, including medicine, education, robotics, mathematics, etc. Each domain was chosen to highlight different aspects of the model’s capabilities, ranging from multi-step reasoning and technical problem-solving to real-time decision-making and knowledge application. For example, in the medical domain, we used structured prompts that simulated real-world clinical scenarios, testing the model’s ability to interpret patient data and suggest plausible diagnoses or treatments. In robotics, the model was tasked with generating code or instructions for various robotic tasks, emphasizing its ability to apply logic to dynamic, real-world operations. Similarly, the mathematics prompts required the model to solve intricate problems that necessitated a deep understanding of formulas, theorems, and multi-step problem-solving processes.

For each domain, we meticulously curated the datasets to ensure that they were representative of the types of challenges encountered by professionals in those fields. The structured prompts were designed not only to test the model’s ability to retrieve factual information but also to assess how well it could adapt its reasoning to domain-specific contexts. In each case, we analyzed o1-preview’s ability to generalize across tasks while maintaining coherence, accuracy, and relevance.

The evaluation went beyond simply recording o1-preview’s correctness. Each response was critically analyzed for logical coherence, contextual relevance, and domain appropriateness. In domains like medicine, where incorrect responses could have serious real-world implications, we carefully assessed the model’s ability to synthesize data from complex scenarios and provide plausible, evidence-based solutions. The same level of scrutiny was applied in mathematics, where multi-step reasoning is essential, and in robotics, where real-time task generation demands precision.

Moreover, we ensured that the evaluation was exhaustive by including multi-layered analyses—ranging from granular error breakdowns to broader assessments of how o1-preview handled domain-specific intricacies. For instance, when assessing its performance in education, we examined how effectively the model could generate accurate, didactic responses that were pedagogically sound. The detailed analysis provided insights into the model’s ability to not only answer questions but also facilitate learning.

One of the key outcomes of this testing process was the identification of areas where o1-preview excelled, as well as those where further refinement is required. Its capacity to adapt to the unique demands of each domain while producing coherent and logically sound responses highlights its potential as a tool for a wide range of professional applications. However, the testing also revealed certain limitations, particularly in tasks requiring more intricate domain-specific reasoning, suggesting areas for further improvement.

\newpage

\subsection{Code Generation}
To evaluate the code generation capabilities of o1-preview, we assessed its performance based on the number of correct solutions and the total points it accumulated in each contest. Table \ref{coding_results} presents the detailed results of o1-preview’s performance in our coding evaluation. Each problem within the contest is assigned a score proportional to its difficulty, with more challenging problems receiving higher scores. The maximum total score achievable in each contest is 21 points.

\begin{table}[H]
\caption{Number of passed questions and corresponding points o1-preview has obtained in our coding test. o1-preview is given three attempts. A problem is considered successfully solved by o1-preview if it passes Leetcode's automated judgement system within any of the three allowed submission attempts.}
\begin{tabular}{ccccc}
\hline
                              & Contest 414 & Contest 413 & Biweekly Contest 138 & Total  \\ \hline
Number of passed solution     & 4           & 3           & 3                    & 10     \\
Percentage of passed solution & 100\%       & 75\%        & 75\%                 & 83.3\% \\
Score of point                & 21          & 15          & 14                   & 50     \\ \hline
\end{tabular}

\label{coding_results}
\end{table}

As presented in Table \ref{coding_results}, o1-preview successfully passed 10 out of 12 problems, yielding a high passing rate of 83.3\%. The model only failed one problem in Weekly Contest 413 and another in Biweekly Contest 138. Notably, both of these problems were categorized as "hard" level. Despite taking several minutes to generate solutions, o1-preview was unable to pass these challenges within the three submission attempts.

For the failed problem in Weekly Contest 413, o1-preview initially generated a solution that was correct but exceeded the time limit. Subsequent submissions in the second and third attempts resulted in incorrect answers. Similarly, for the problem in Biweekly Contest 138, o1-preview provided incorrect solutions for all three attempts.

Despite these challenges, o1-preview demonstrated comparable or faster code generation times relative to top human competitors. This indicates that o1-preview possesses a strong reasoning capability and can effectively handle the majority of coding tasks. However, in particularly complex or computationally intensive cases, the model's performance remains limited, as observed in these difficult problems.

\begin{figure}[H]
    \centering
    \includegraphics[width = \textwidth]{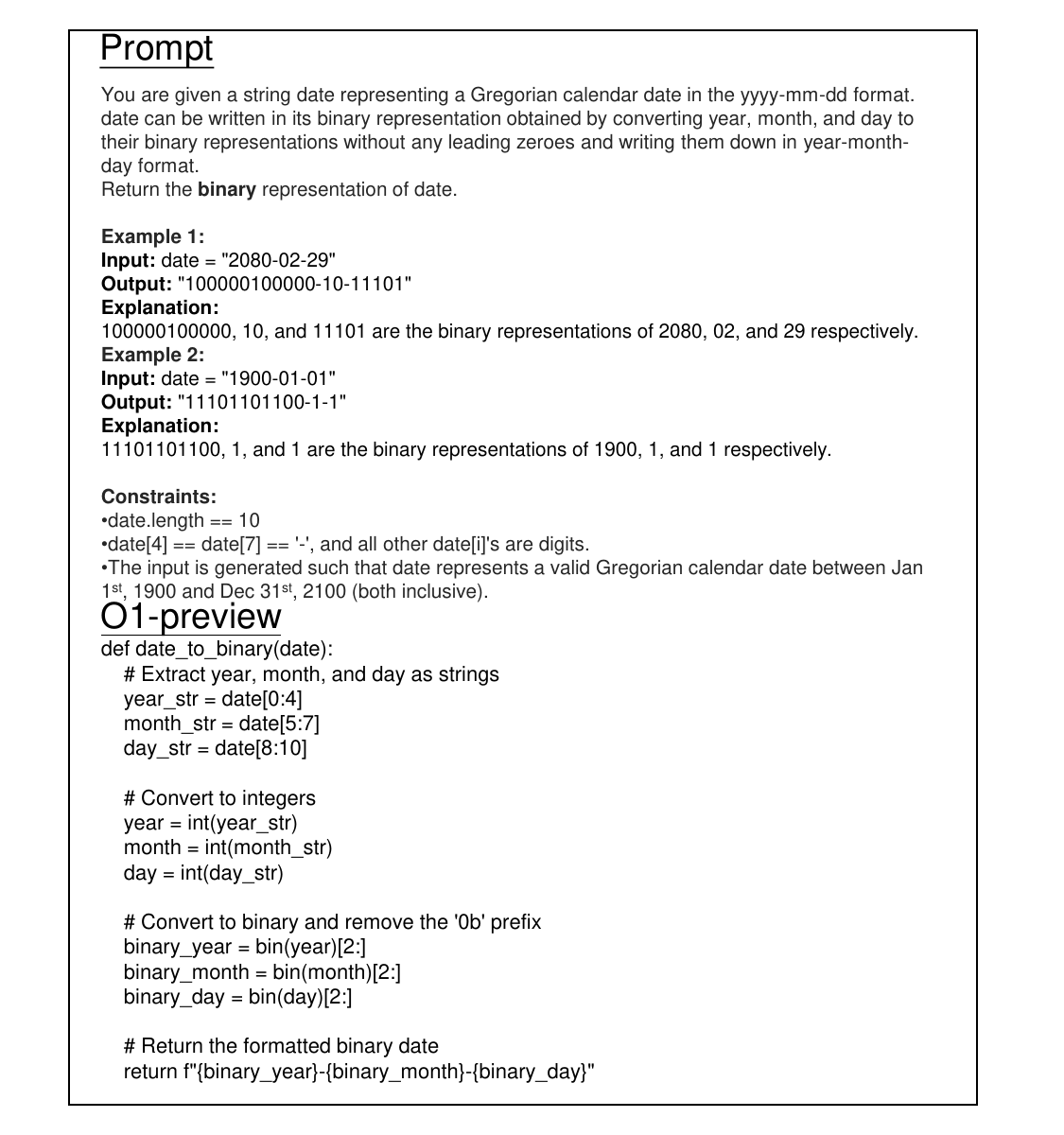}
    \caption[Code Generation: Case 1]{\textbf{Coding: Case 1.} A demonstration case from coding testing of a easy level problem on o1-preview solution. The generated code passes the Leetcode automated judgement system for the first attempt.}
    \label{fig:coding_1}
\end{figure}

\begin{figure}[H]
    \centering
    \includegraphics[width = \textwidth]{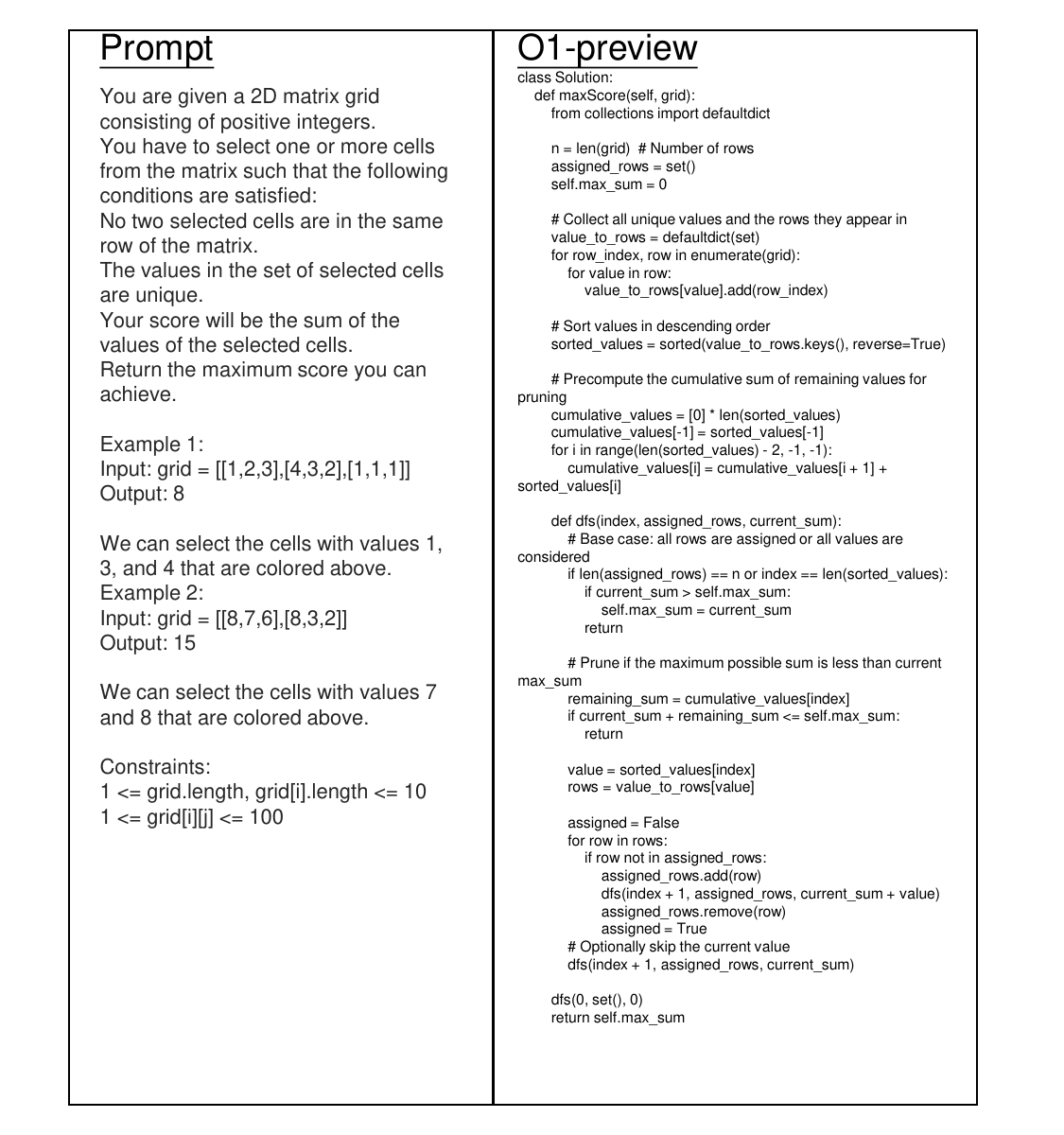}
    \caption[Code Generation: Case 2]{\textbf{Coding: Case 2.} In this case, o1-preview generates the correct code. However, this code has exceeded the time limit. For the following second and third attempt, o1-preview keeps generating the code with wrong answer. Therefore, we regard this problem as a failure case.}
    \label{fig:coding_2}
\end{figure}

Figures \ref{fig:coding_1} and \ref{fig:coding_2} illustrate two examples from our coding evaluation. In Figure \ref{fig:coding_1}, o1-preview demonstrates its ability to efficiently solve easy-level problems, completing the task with minimal computation time and successfully passing all test cases. However, as shown in Figure \ref{fig:coding_2}, the model encounters difficulties with a hard-level problem. In this failure case, o1-preview initially generated a correct solution, but the solution exceeded the time limit. Following the first submission, the model became trapped in a cycle of attempting to optimize the code’s time complexity, which resulted in incorrect solutions in subsequent attempts. This example highlights the model’s struggle with complex problems requiring significant optimization, where its efforts to improve efficiency led to repeated errors. Additional test cases are provided in the Appendix~\ref{appendix:coding}
\newpage

\newpage


\subsection{Radiology Report Generation}

We evaluated the medical report generation capability of the o1-preview by comparing its ROUGE metrics against baseline models such as gpt-4-turbo, gpt-4o, and others. Table \ref{tab:Results_for_Compared_LLMs} provides a detailed comparison of the performance of o1-preview with five other models, with the best-performing results highlighted. Additionally, we calculated the average generation time for medical reports produced by each model to further assess their overall efficiency and performance.

As shown in Table \ref{tab:Results_for_Compared_LLMs}, the ROUGE scores for the reports generated by o1-preview compared to those written by doctors are: R-1: 0.3019, R-2: 0.0448, and R-L: 0.2841, ranking highest among the six models. Notably, o1-preview also has the longest average report generation time at 15.051 seconds, while gpt-3.5-turbo is the fastest at 2.024 seconds, though it has the lowest ROUGE scores. This difference can be attributed to o1-preview's extensive inference verification processes aimed at ensuring result accuracy, which gpt-3.5-turbo lacks. In terms of generation time, both o1-preview and gpt-4-turbo took longer than the other models but delivered stronger ROUGE results.

\begin{table}[!ht]
\centering
\caption{Comparing different models with R-1, R-2, and R-L scores and average time, including performance indicators.}
\label{tab:Results_for_Compared_LLMs}
\resizebox{0.8\textwidth}{!}{%
\begin{tabular}{lcccc}
\toprule
\multirow{2}{*}{Models} & \multicolumn{3}{c}{SXY} & \multirow{2}{*}{Average Time(s)} \\ \cmidrule(lr){2-4}
                                 & R-1  & R-2  & R-L   &                              \\ \midrule
o1-preview                       & \textbf{0.3019} & 0.0448 & \textbf{0.2841} & 15.051                       \\ 
o1-mini                          & 0.2582 & 0.0586 & 0.2449 & 5.793                        \\ 
gpt-4o                           & 0.2855 & 0.0615 & 0.2552 & 4.764                        \\ 
gpt-4o-mini                      & 0.1346 & 0.0154 & 0.1269 & \textbf{1.629} \\ 
gpt-4-turbo                      & 0.2999 & \textbf{0.0656} & 0.2667 & 13.942               \\ 
gpt-3.5-turbo                    & 0.0600   & 0.0154 & 0.0600   & 2.024      \\ \bottomrule
\end{tabular}%
}
\end{table}

\begin{figure}[H]
    \centering
    \includegraphics[width = 0.9\textwidth]{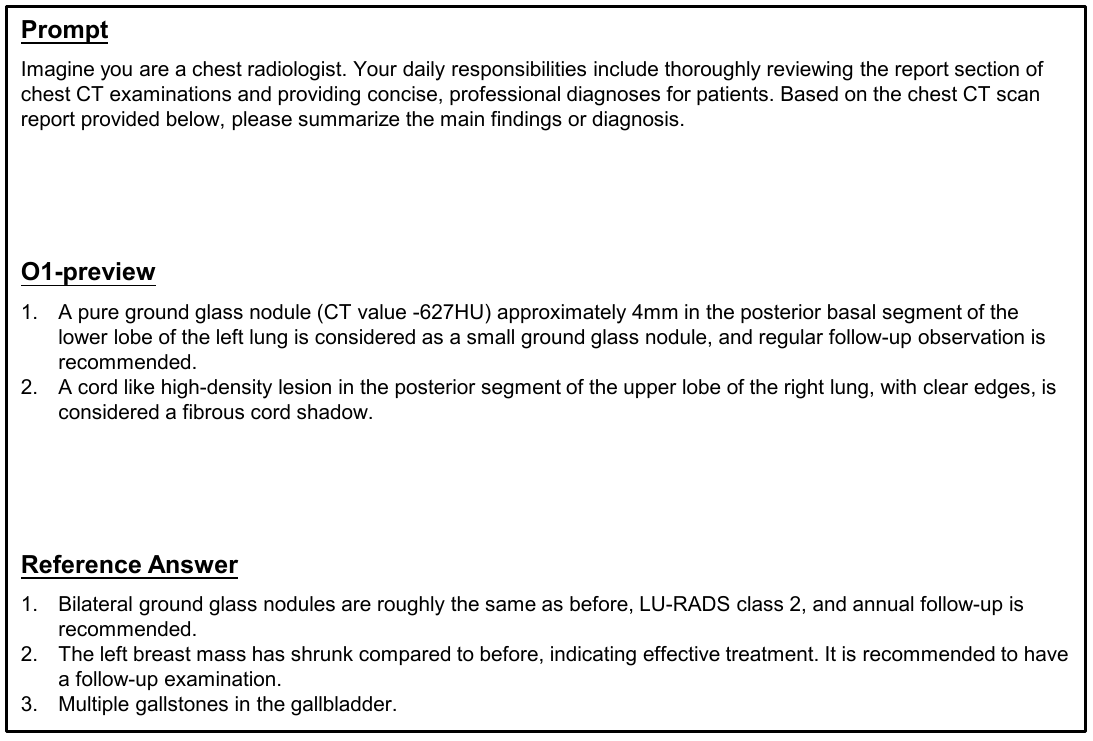}
    \caption[Radiology Report Generation: Case 1]{\textbf{Radiology: Case 1.} We observed that the o1-preview closely aligns with human writing patterns, exhibiting clear organization. To enhance reader accessibility, the radiological report generated in the zero-shot o1-preview task has been translated from Chinese to English.}
    \label{fig:Radiology_1}
\end{figure}

\begin{figure}[H]
    \centering
    \includegraphics[width = 0.9\textwidth]{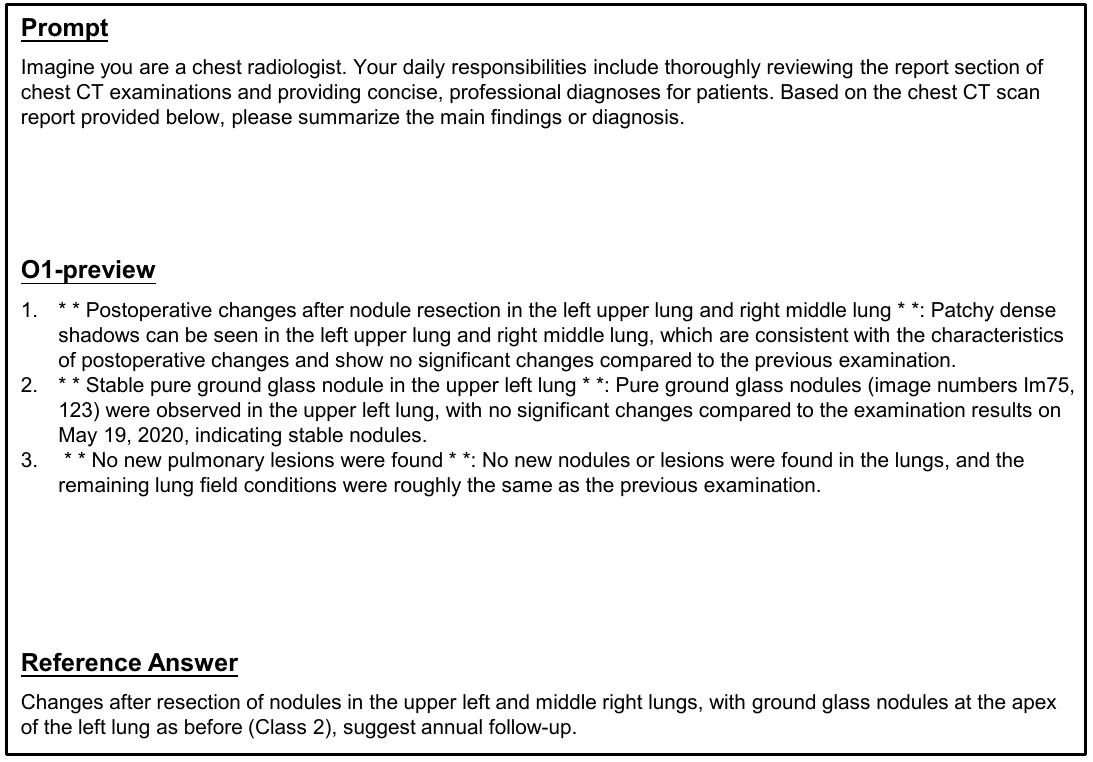}
    \caption[Radiology Report Generation: Case 2]{\textbf{Radiology: Case 2.} We observed that the o1-preview closely aligns with human writing patterns, exhibiting clear organization and concise language. To enhance reader accessibility, the radiological report generated in the zero-shot o1-preview task has been translated from Chinese to English.}
    \label{fig:Radiology_2}
\end{figure}

Figures \ref{fig:Radiology_1} and \ref{fig:Radiology_2} present two examples illustrating the evaluation of our radiology report generation. During the evaluation, o1-preview demonstrated the ability to generate reliable radiological reports with zero-shot samples in less time, highlighting its strong reasoning and knowledge transfer capabilities, as well as its potential in the medical field. We observed that o1-preview aligns well with human writing patterns, featuring clear organization and concise language. While other models did not achieve the highest similarity scores, most were able to follow instructions and complete the tasks. Although these large language models have the potential to reduce doctors' workloads, alleviate the strain on limited medical resources, and foster collaboration between clinical practitioners and AI technology, substantial improvements and rigorous validation are necessary before clinical deployment. Additional test cases are provided in the Appendix~\ref{appendix:Radiology}

\newpage
\subsection{Robot Command Planning}
\label{Robot examples}
The Robot Operating System (ROS) is widely recognized as one of the most commonly utilized frameworks for robotic control and development. It integrates key functionalities such as node communication, parallel control, and behavior trees—features that are essential in robotics but rarely found in traditional software systems. Due to the complexity and specialized nature of these control mechanisms, mastering ROS is considered highly challenging, even among seasoned programmers. This difficulty is further compounded by the fact that many large language models (LLMs) struggle to generate high-quality ROS control code efficiently, primarily because of the nuanced and domain-specific knowledge required to handle the intricate interactions between its components. Consequently, ROS development remains a specialized skill that is not easily automated or replicated by general-purpose AI models.

\begin{figure}[H]
    \centering
    \includegraphics[width = 0.9\textwidth]{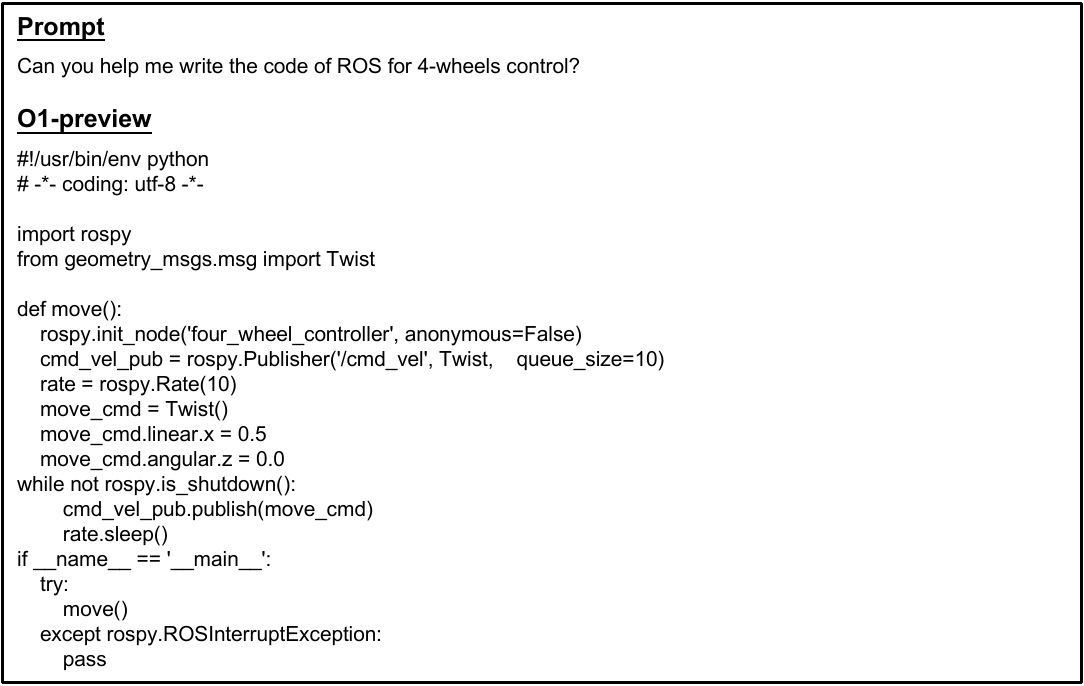}
    \caption[Robot Command Planning: Case 1]{\textbf{Robot Command Planning: Case 1.} ROS code generated by o1-preview for controlling a four-wheel robot, featuring Twist message handling to control forward movement and angular velocity.}
    \label{fig:robot1}
\end{figure}

The results, as depicted in Figure~\ref{fig:robot1}, clearly illustrate o1-preview's profound comprehension of robot command structures. The generated code not only aligns accurately with the given instructions but also demonstrates a high degree of precision and reliability. This suggests that o1-preview possesses an exceptional ability to interpret and synthesize complex commands within the context of robotic control. The accuracy of the output further underscores the model's robustness in handling intricate tasks, which are critical in real-world applications where precise control and execution are paramount.

\newpage

Moreover, this performance highlights o1-preview's potential to play a significant role in advancing automation and robotics, as it can efficiently translate abstract instructions into executable commands. The model’s capabilities extend beyond simple task interpretation, offering a deeper integration of machine learning in robotics. As the field moves toward more sophisticated and autonomous systems, the ability of models like o1-preview to adapt and deliver highly accurate code suggests promising advancements in both industrial and research-based robotic applications.

This demonstration affirms that o1-preview is not only a powerful tool for natural language processing but also a versatile system that can be leveraged to enhance control mechanisms in robotics, paving the way for innovative approaches in machine learning-driven automation.

\newpage


\subsection{Nature Language Inference} 
We randomly selected two test samples from each dataset's test set, conducting a total of 10 test cases. Table \ref{nli_case1} presents an example for each dataset, where o1-preview accurately analyzed the logical relationships, demonstrating advanced reasoning ability and domain-specific knowledge. This underscores its potential for real-world applications in various complex scenarios. Additional examples can be found in Appendix \ref{appx_nli}.

\begin{sidewaystable}[thp]
\centering
\caption[Nature Language Inference: Case 1]{Examples of o1-preview responses on NLI task.\\ For NLI task, the prompt is: Given a premise sentence and a hypothesis sentence, predict whether the premise entails the hypothesis (entailment), contradicts the hypothesis (contradiction), or neither (neutral).\\
For QNLI task, the prompt is: Determine whether the context sentence contains the answer to the question.}
\label{nli_case1}
\scriptsize 
\begin{tabularx}{\textwidth}{@{}>{\hsize=.1\hsize}X>{\hsize=.4\hsize}X>{\hsize=.4\hsize}X>{\hsize=0.7\hsize}X>{\hsize=0.2\hsize}X@{}}
\toprule
Dataset & Premise / Context & Hypothesis / Question & o1-preview Response & Label \\
\midrule
MNLI &
Tom is the winner of a year's supply of Turtle Wax, and he will receive his prize just as soon as the Shopping Avenger figures out how much Turtle Wax actually constitutes a year's supply. &
There are no winners of the one year supply of Turtle wax. &
The premise explicitly states that Tom is the winner, while the hypothesis claims there are no winners. These statements directly oppose each other, making the relationship a contradiction. Answer: \hl{Contradiction}&
Contradiction\\
\addlinespace[5pt]

ANLI &
The recent 14\% hike in third class postage rates, accompanied by simultaneous double-digit paper price increases, has hit smaller catalogers especially hard. &
Smaller catalogers are hit harder than bigger ones. &
The premise states that smaller catalogers have been hit especially hard by the increase in postage rates and paper prices. The term "especially hard" implies that smaller catalogers are more adversely affected compared to others in the industry. Although the premise does not explicitly mention bigger catalogers, the use of "especially" suggests a comparison, indicating that smaller catalogers are hit harder than their larger counterparts. Therefore, the premise entails the hypothesis. Answer: \hl{Entailment} &
Entailment\\
\addlinespace[5pt]

QNLI &
Stagecoach is the primary operator in the city proper, with cross-city services mainly between both the West and East ends via the city centre with some services extending out to the MetroCentre, Killingworth, Wallsend and Ponteland. &
How many major bus stations are in the city of Newcastle? &
\hl{No}, the context sentence does not provide the number of major bus stations in the city of Newcastle. &
Not entailment\\
\addlinespace[5pt]

\midrule

MedNLI &
EKG showed T-wave depression in V3-5, with no prior EKG for comparison. &
Patient has a normal EKG. &
The premise states that the EKG revealed T-wave depression in leads V3-5, which is an abnormal finding indicating potential cardiac issues such as ischemia or other heart conditions. A normal EKG would not show such abnormalities. Therefore, the hypothesis that the patient has a normal EKG directly contradicts the information provided in the premise. Answer: \hl{Contradiction} &
Contradiction\\
\addlinespace[5pt]

RadQNLI &
The heart size is probably normal, but the cardiac silhouette is partially obscured by a moderate left pleural effusion. &
Do we find any unilateral or bilateral pleural effusion? &
\hl{Yes}, the context sentence contains the answer. It mentions a moderate left pleural effusion, which indicates a unilateral pleural effusion. &
Entailment\\
\bottomrule
\end{tabularx}
\end{sidewaystable}

\newpage


\subsection{Quantitative Investing}
\begin{figure}[H]
    \centering
    \includegraphics[width = \textwidth]{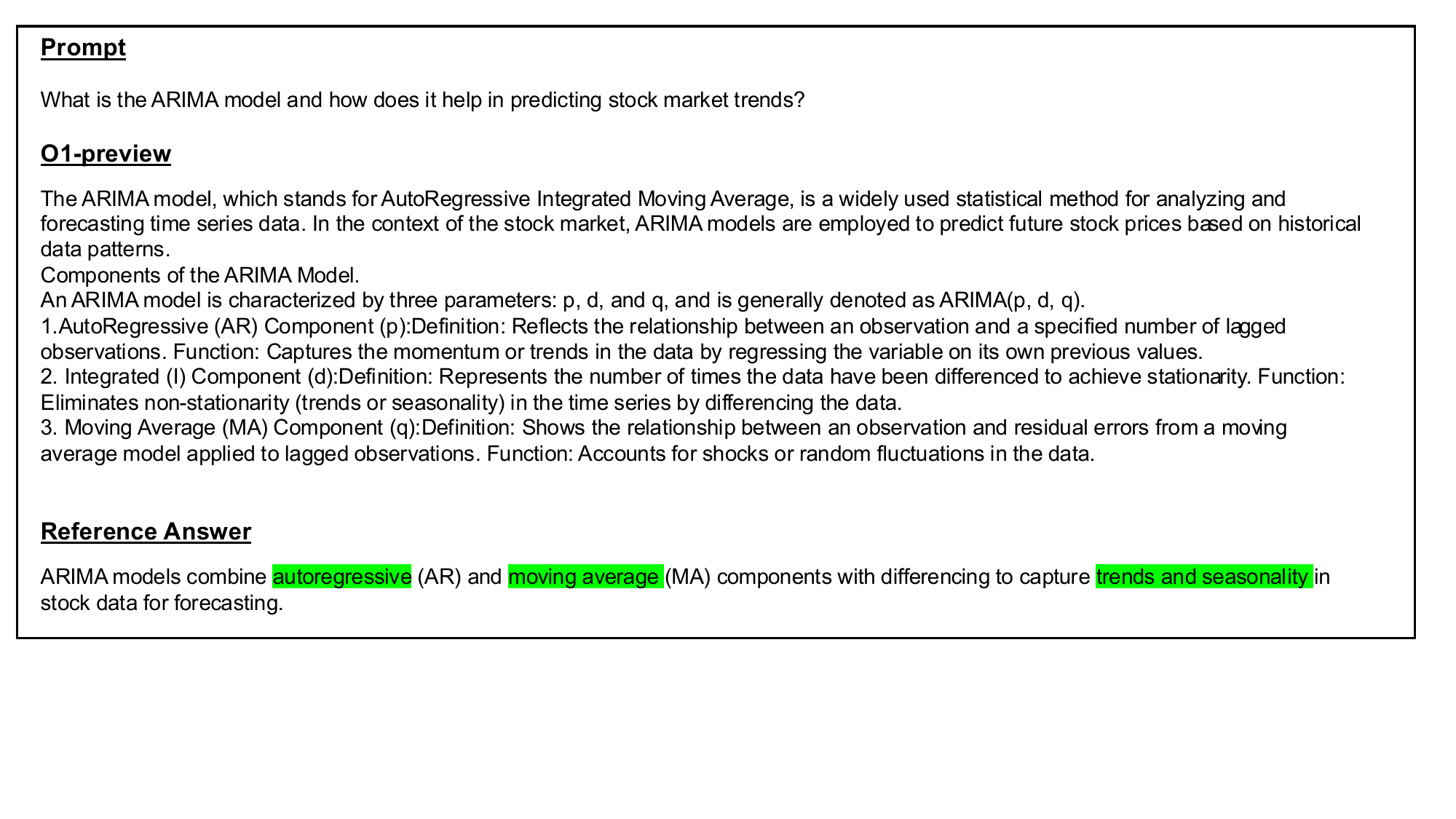}
    \caption[Quantitative Investing: Case 1]{\textbf{Quantitative Investing: Case 1.} o1-preview successfully explained the concept of ARIMA, described its functionality, and detailed its working mechanism, demonstrating a thorough understanding of the statistical model. The dataset for this task is sourced from \cite{yym2023stocktradingqa}.}
    \label{fig:Qsample1}
\end{figure}

o1-preview has solid financial knowledge and statistical modeling skills, and can deeply understand and clearly explain complex concepts such as mean reversion (from Figure \ref{fig:Qsample3}), ARIMA models (from Figure \ref{fig:Qsample1}), stochastic oscillators(from Figure \ref{fig:Qsample5}), and the role of autocorrelation (from Figure \ref{fig:Qsample3}) in market forecasting. In addition, o1-preview has demonstrated the ability to comprehensively apply trend, seasonality (from Figure \ref{fig:Qsample4}), and residual analysis in market forecasting, and has the ability to think systematically and apply practically. Overall, o1-preview has performed well in the fields of financial analysis and market forecasting, and has the ability to conduct comprehensive analysis and application.

\begin{figure}[H]
    \centering
    \includegraphics[width = \textwidth]{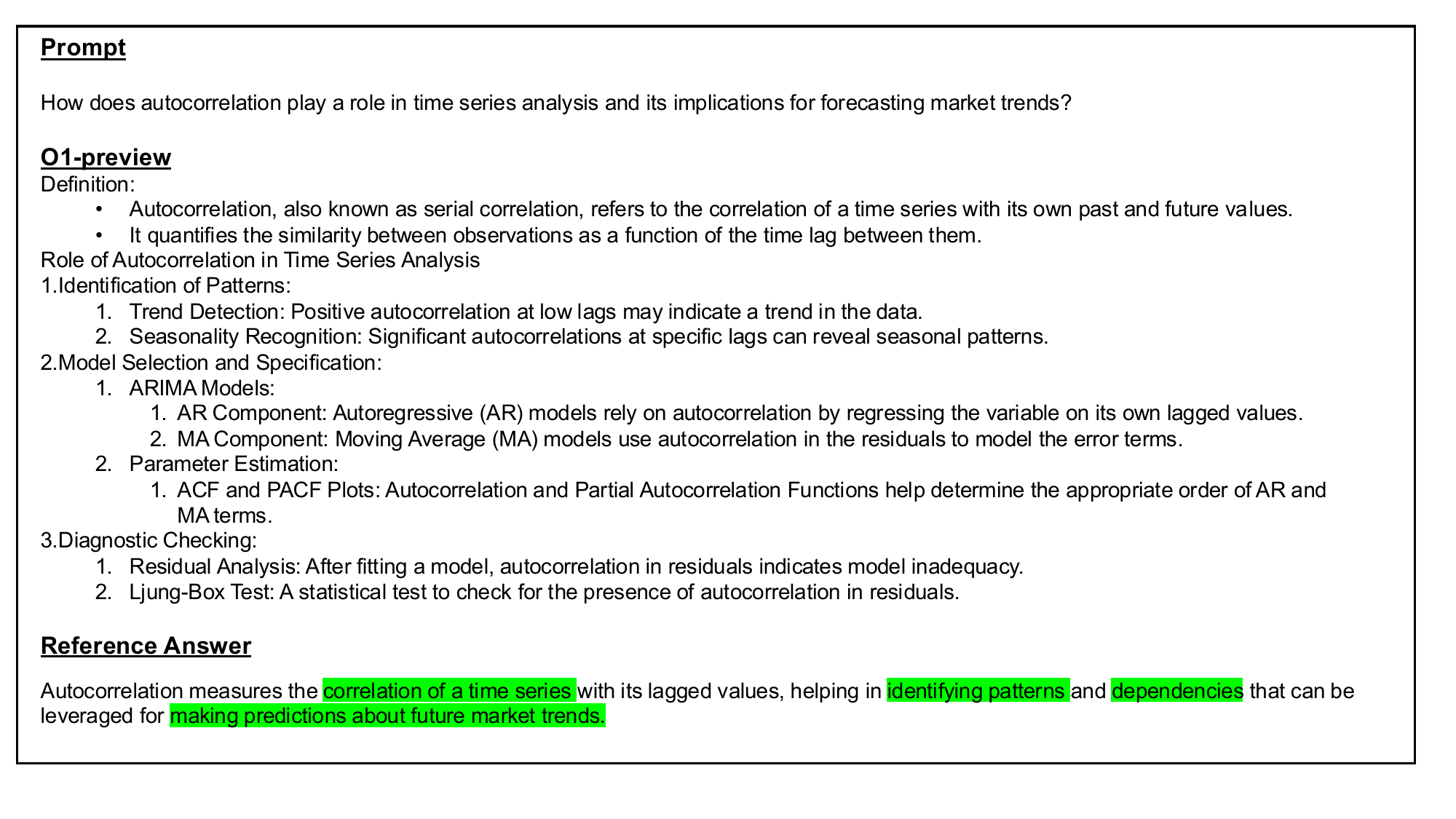}
    \caption[Quantitative Investing: Case 2]{\textbf{Quantitative Investing: Case 2.} o1-preview successfully identified the role of autocorrelation in market forecasting, explained its working mechanism, and highlighted its importance, demonstrating an accurate understanding of the underlying process.}
    \label{fig:Qsample2}
\end{figure}

\newpage

\subsection{Low-Resource Language Translation}
\begin{figure}[H]
    \centering
    \includegraphics[width = \textwidth]{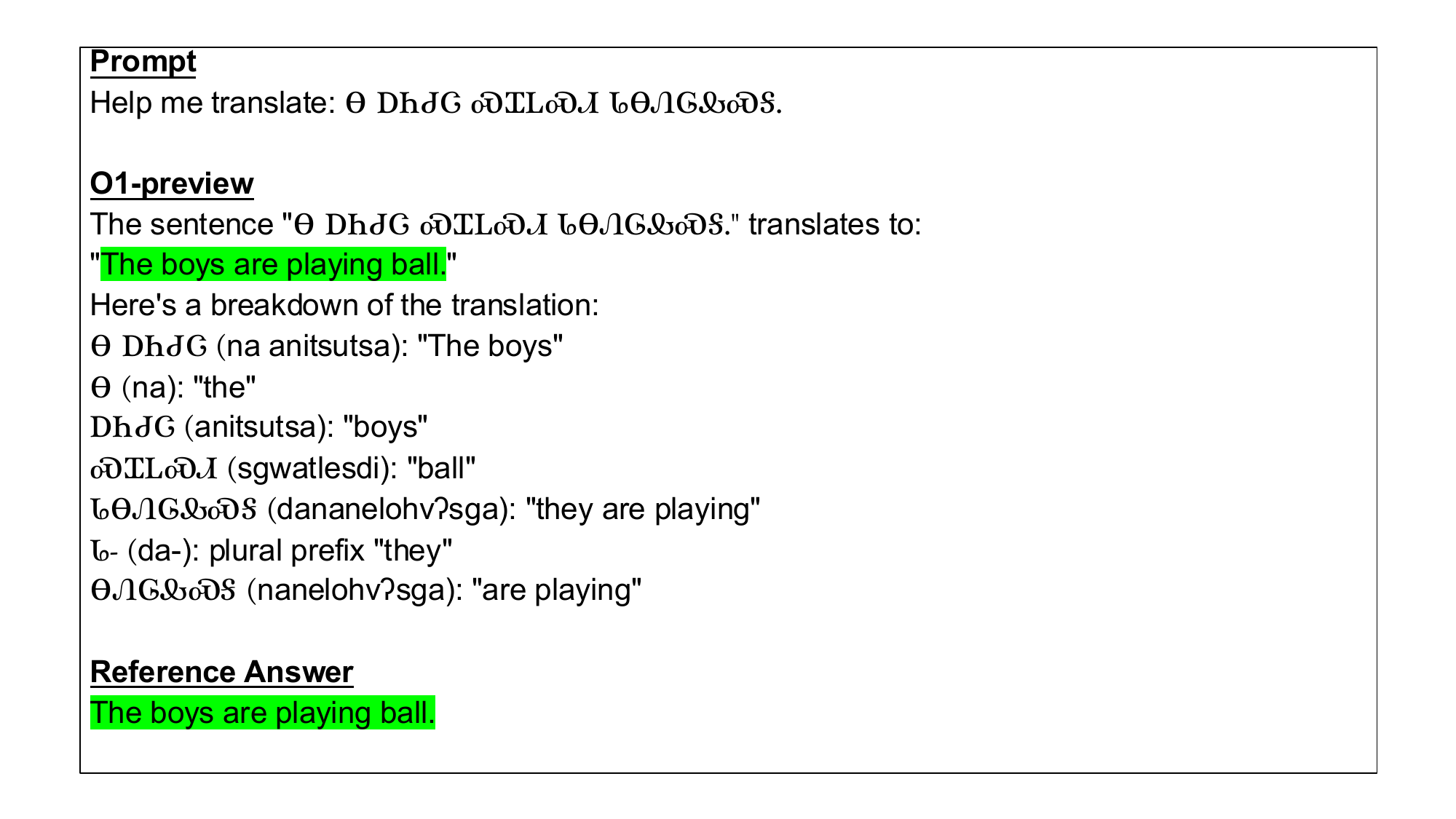}
    \caption[Low-Resource Language Translation: Case 1]{\textbf{Low-Resource Language Translation: Case 1.} In this translation sample, the model accurately translates this Cherokee sentence as "The boys are playing ball", which is exactly the same with reference answer. It also offers a detailed breakdown of each component of the sentence, demonstrating its ability to handle both grammar and words in this translation.}
    \label{fig:cherokee_1}
\end{figure}

\begin{figure}[H]
    \centering
    \includegraphics[trim=200 0 200 0,clip,width = \textwidth]{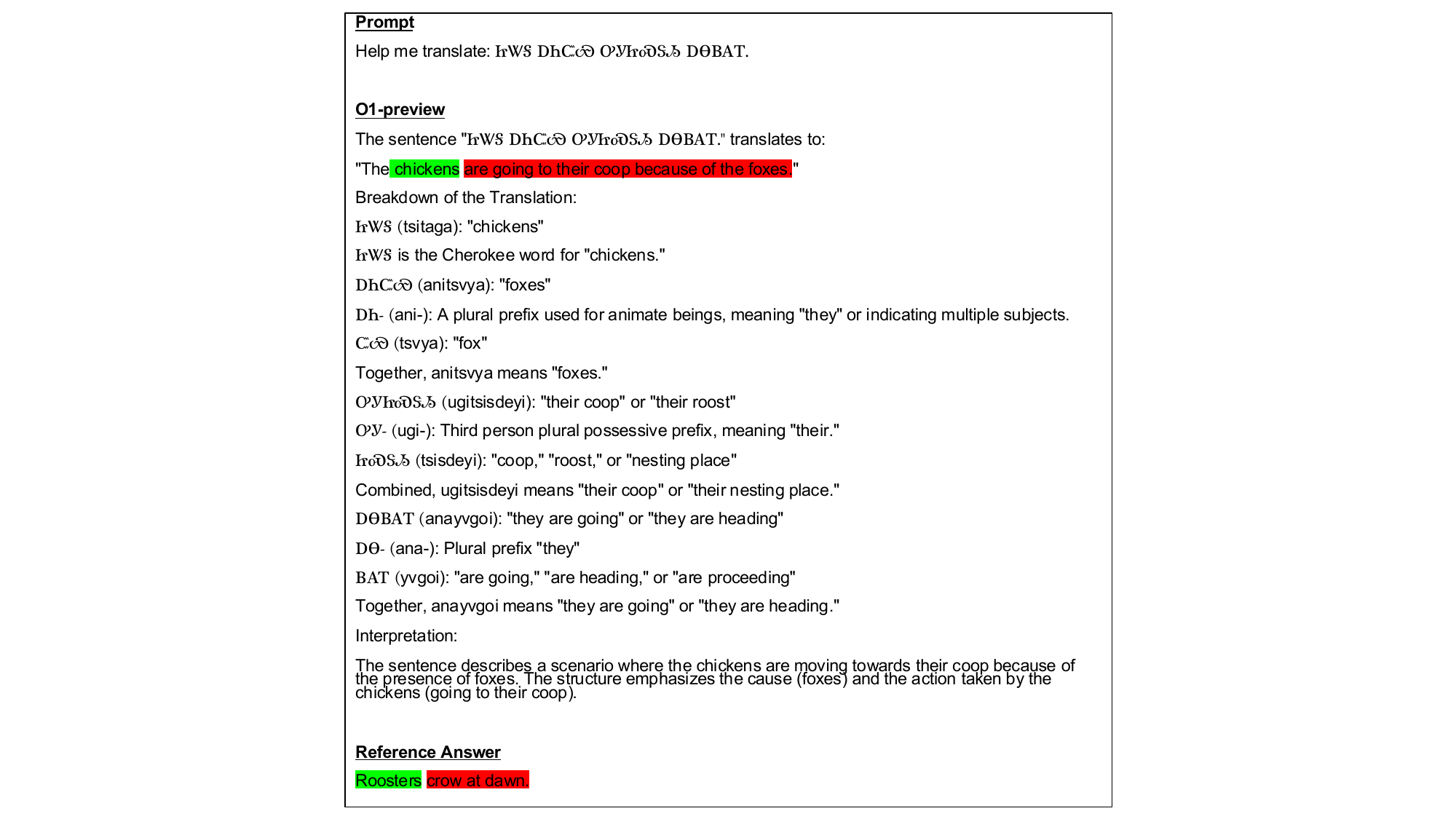}
    \caption[Low-Resource Language Translation: Case 2]{\textbf{Low-Resource Language Translation: Case 2.} This translation correctly translate the word chicken. But o1-preview model is unable to translate the rest words, like male, crow and dawn, in this sentence.}
    \label{fig:cherokee_2}
\end{figure}

In the evaluation of low-resource language translations by the o1-preview model, two Cherokee sentence cases were examined. In the first scenario, as shown in Figure \ref{fig:cherokee_1}, the sentence "The boys are playing ball" was accurately translated. The model identified all key elements, including the plural form of "boys" and the action verb "playing," while also offering a detailed breakdown of the grammatical structure. Similarly, in the second case, as shown in Figure \ref{fig:cherokee_2}, the o1-preview model translated the sentence into "The chickens are going to their coop because of the foxes." While it correctly identified "chickens," it has the some interpretations, for example, "foxes" and "coop", differed from the reference translation, which described "roosters crowing at dawn." These instances demonstrated the model’s ability to handle word recognition and basic grammar in low-resource language translation. However, while the model accurately faced unknown words in each case, it struggled with contextual nuances and precision in more complex scenarios.

\newpage

\newpage

\subsection{Educational Q\&A}
\begin{figure}[H]
    \centering
    \includegraphics[width = \textwidth]{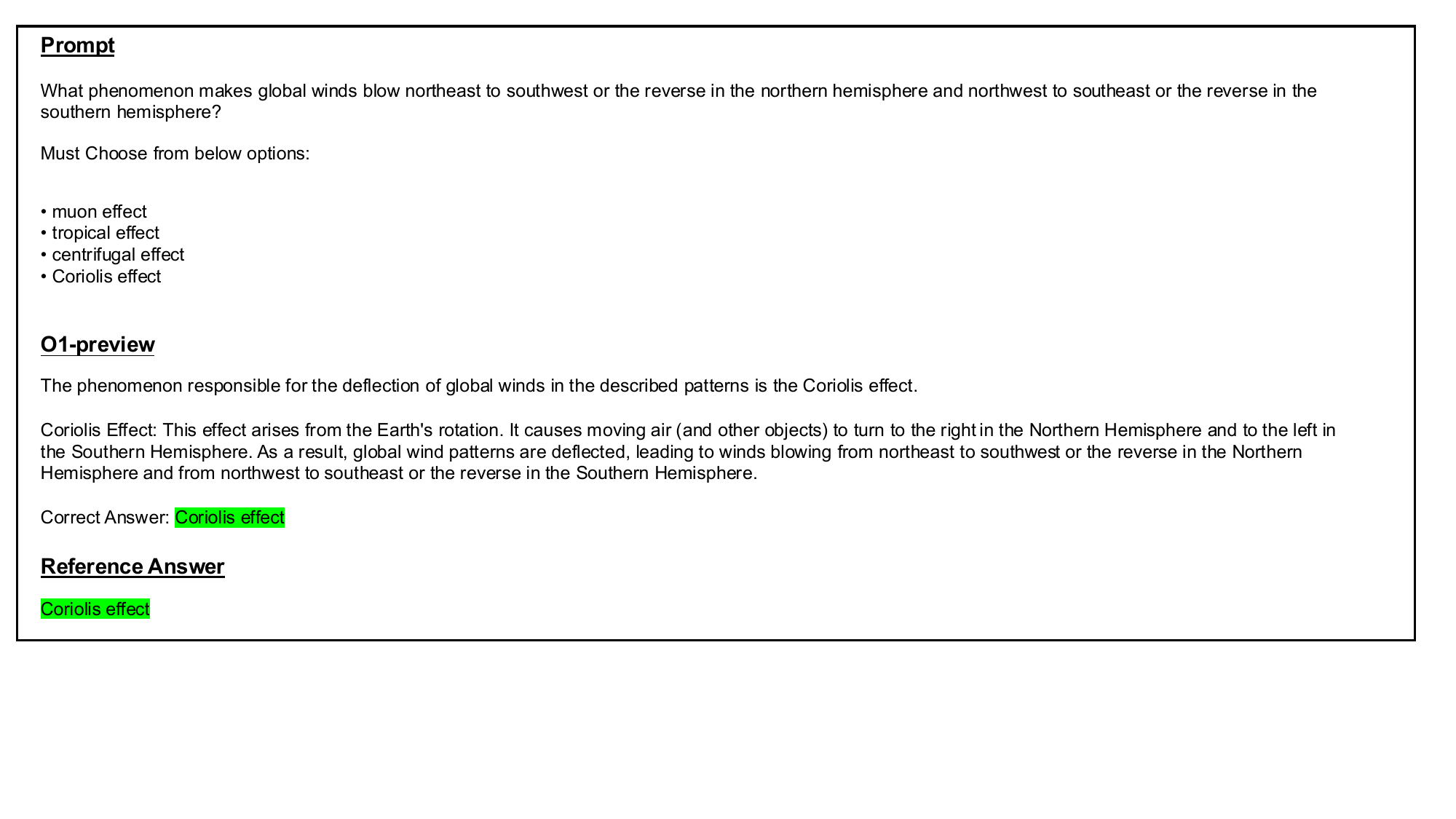}
    \caption[Educational Q\&A: Case 1]{\textbf{Educational Q\&A: Case 1.} o1-preview successfully identified the correct answer, Coriolis effect, for the phenomenon responsible for global wind deflection, demonstrating accurate understanding of the underlying process. The dataset for this task is sourced from \cite{welbl2017crowdsourcing}.}
    \label{fig:education_qa_1}
\end{figure}

\begin{figure}[H]
    \centering
    \includegraphics[width = \textwidth]{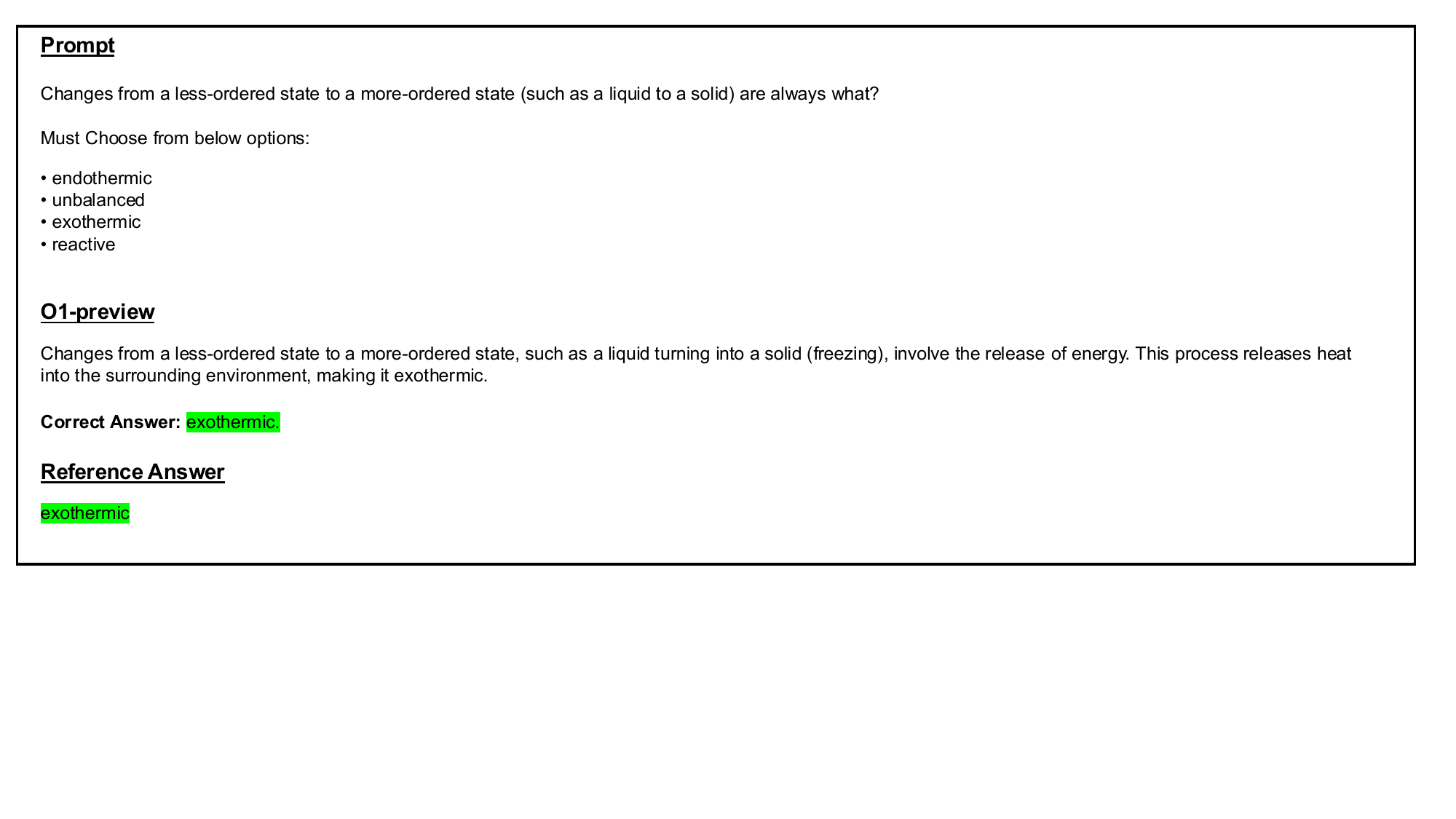}
    \caption[Educational Q\&A: Case 2]{\textbf{Educational Q\&A: Case 2}  o1-preview successfully identified the correct answer, exothermic, for the process of changes from a less-ordered state to a more-ordered state, such as a liquid turning into a solid, demonstrating accurate understanding of the energy release involved. The dataset for this task is sourced from \cite{welbl2017crowdsourcing}.}
    \label{fig:education_qa_2}
\end{figure}

In this educational test using o1-preview, the model successfully identified the correct answers for two distinct scientific concepts. In the first Figure \ref{fig:education_qa_1}, o1-preview correctly recognized that the phenomenon responsible for global wind deflection is the Coriolis effect, which results from the Earth's rotation and affects wind patterns across hemispheres. In the second Figure \ref{fig:education_qa_2}, o1-preview accurately identified exothermic as the process where changes from a less-ordered state to a more-ordered state, such as a liquid turning into a solid, involve the release of heat. These results demonstrate o1-preview’s capability to accurately interpret and respond to scientifically framed questions.

\newpage

\newpage

\subsection{Student Writing Improvement in Higher Education}
\label{test: slat}
\begin{figure}[H]
    \centering
    \includegraphics[width = \textwidth]{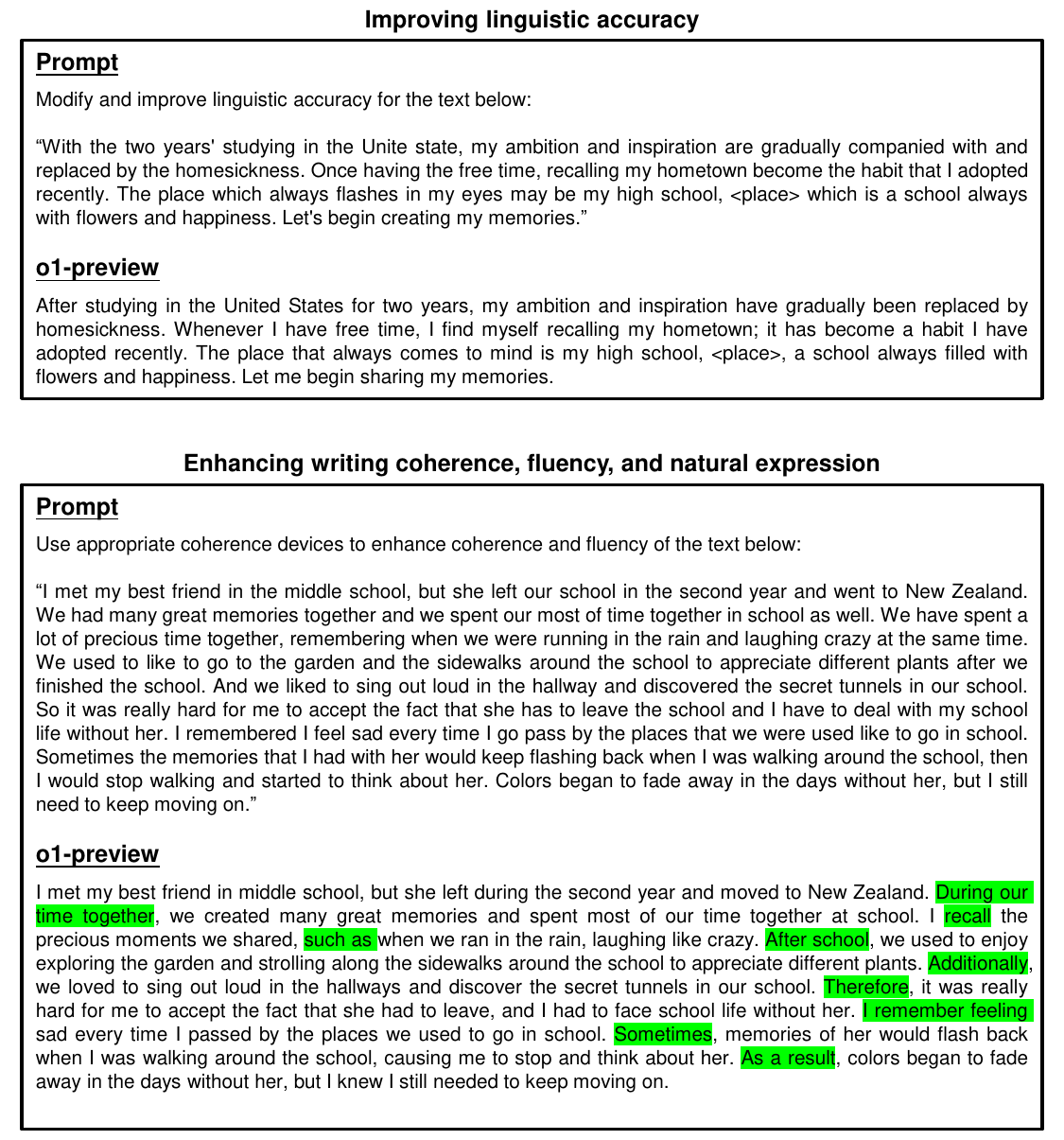}
    \caption[Academic Writing Improvement: Case 1]{\textbf{Academic Writing Improvement: Case 1.}  Example of general improvements in higher education writing.}
    \label{slat_case1}
\end{figure}

\begin{figure}[H]
    \centering
    \includegraphics[width = \textwidth]{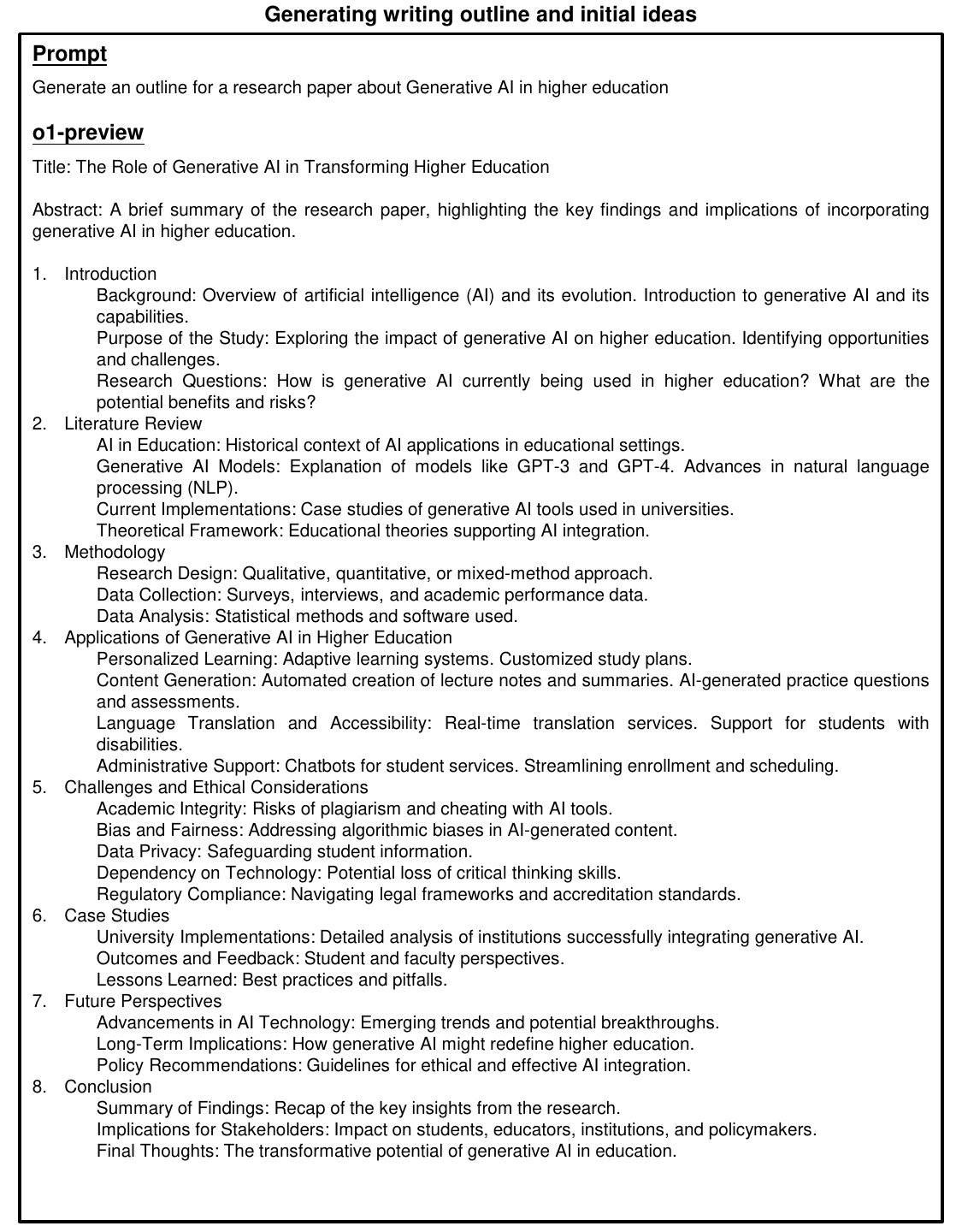}
    \caption[Academic Writing Improvement: Case 2]{\textbf{Academic Writing Improvement: Case 2}  Example of writing preparation support.}
    \label{slat_case2}
\end{figure}

\begin{figure}[H]
    \centering
    \includegraphics[width = \textwidth]{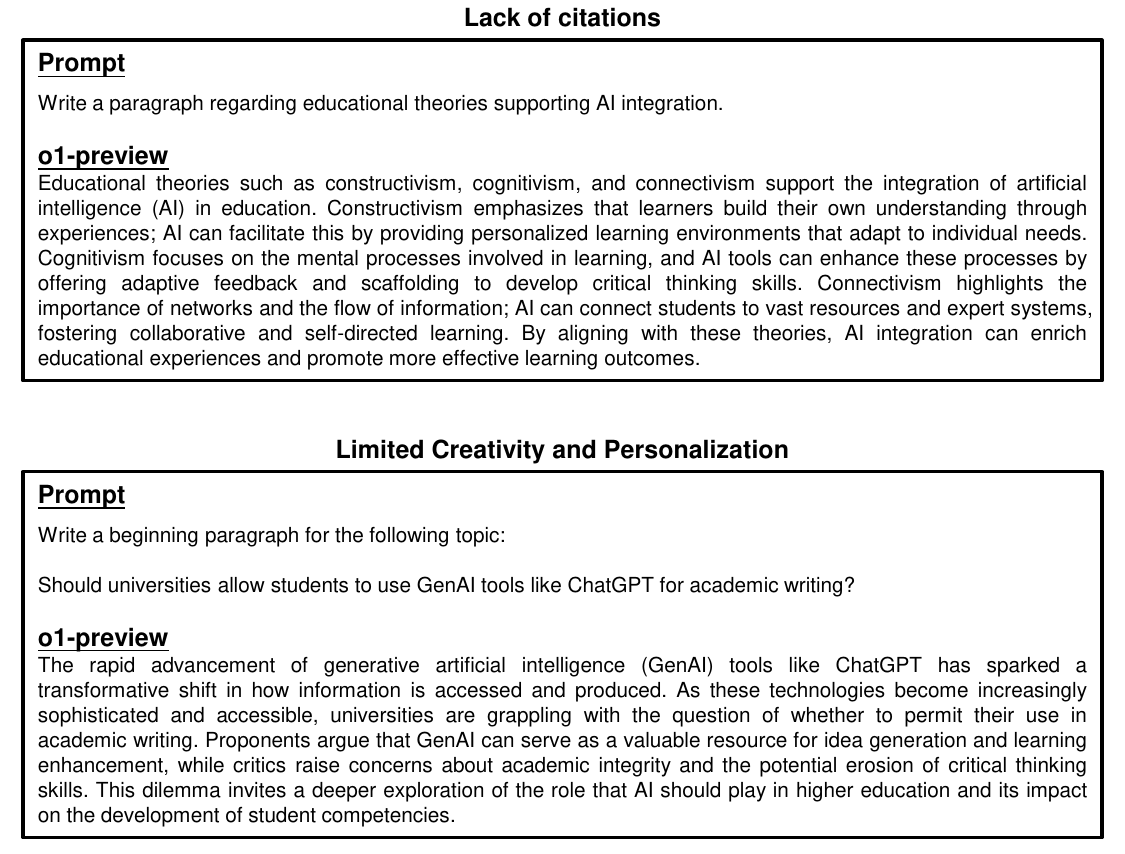}
    \caption[Academic Writing Improvement: Case 3]{\textbf{Academic Writing Improvement: Case 3}  Example of limitations.}
    \label{slat_case3}
\end{figure}

As educators explore new technologies to enhance student learning, o1-preview may serve as an important teaching and learning assistant in helping students improve their writing skills. Qualitative analysis of the o1-preview model's application in student writing in Figures \ref{slat_case1}, \ref{slat_case2}, \ref{slat_case3} has shown promising results in several areas. For example, the above screenshots reveal that o1-preview appears to be able to provide effective writing support, including improving linguistic accuracy, enhancing writing coherence, and generating writing outlines and initial ideas. However, while these functions seem to contribute positively, the overall impact may vary depending on the specific needs of learners and the contexts in which the tool is used. The initial analysis found that the AI-generated responses sometimes follow a rigid, machine-like template and lack citations, which may lead students to passively adopt these structures, ultimately limiting the effectiveness and creativity in their writing. Therefore, it is important for educators to understand that while GenAI can be a valuable supplement, it cannot replace the essential human guidance in writing education. Most importantly, the current AI model is still under development, which means its limitations must be carefully considered in educational contexts \cite{wang2023seeing,wang2024system,latif2023ai}. We hope that future research will further investigate and examine o1's functions and its effectiveness in higher education, providing deeper insights into how it can be best utilized across various learning contexts.

\newpage

\newpage

\subsection{3D Layout Generation}
In this part, we evaluated the performance of model o1-preview in generating room layouts. We provided the model with a carefully designed prompt aimed at guiding it to effectively arrange objects within a given room type. We use the prompt from LLplace \cite{yang2024llplace3dindoorscene}. The specific prompt is as follows:

You are a skilled room layout designer. Your task is to arrange [Objects] within a given [Room Type] effectively. Follow these guidance to complete your design:

1. Extract the [Room Type], [Objects], and [Bounding Box Size] from the provided JSON data.
2. Analyze the spatial relationships among [Objects] within the specified [Room Type]. Pay special attention to avoiding overlap and consider other spatial factors like accessibility and aesthetics.
3. Determine and design the precise location of all [Objects] ensuring that their bounding boxes do not overlap and that the layout is functional and visually appealing.
4. I prefer objects to be placed at the edge (the most important constraint) of the room if possible, which makes the room look more spacious.
5. The objects are usually \textbf{aligned}.
6. Chairs must be placed near to the table/desk and face to the table/desk.
7. The last design output token is the [/Task Output] and only one.
8. Report your design with detailed 3D space coordinates and rotation angles for each object in JSON format, as follows:

{ “object”: “object”, “coordinates”: [ { “x”: x, “y”: y, “z”: z } ], “rotate”: [ { “angle”: r } ] }

The centroid of the room is { “x”: 0.00, “y”: 0.00, “z”: 0.00 }.

First carefully read this example:

[Example Room Type] Bedroom [/Example Room Type]

[Example Objects and Bounding Box Size] /* A fixed example is put here to show the input format */ [/Example Objects and Bounding Box Size]

[Example Output] /* A fixed example is put here to show the output format */ [/Example Output]

Now, please proceed with the design task as outlined and provide only the JSON formatted output of your design:

[Task Room Type] /* Input room type */ [/Task Room Type]

[Task Objects \& Bounding Box Size] /* The JSON format input of objects description and bounding box size */ [/Task Objects \& Bounding Box Size]

Through experiments, we found that model \textit{o1-preview} performed exceptionally well in the task of room layout generation. The model was able to accurately extract the room type, objects, and bounding box sizes from the provided JSON data, ensuring the correct interpretation of fundamental information. It conducted an in-depth spatial relationship analysis of the objects within the specified room type, successfully avoiding overlap between objects in most cases, while also considering spatial factors such as accessibility and aesthetics.

In the layout design task, the model strictly adhered to the guidelines, precisely determining the positions of all objects and ensuring that their bounding boxes did not overlap. The resulting layouts were highly functional and visually pleasing. Notably, the model prioritized placing objects at the edges of the room, a strategy that effectively increased the sense of spaciousness. Objects in the layout were generally aligned, and chairs were correctly placed near tables or desks and faced them, meeting both functional and aesthetic requirements.

\begin{figure}[H]
    \centering
    \includegraphics[width = 0.85\textwidth]{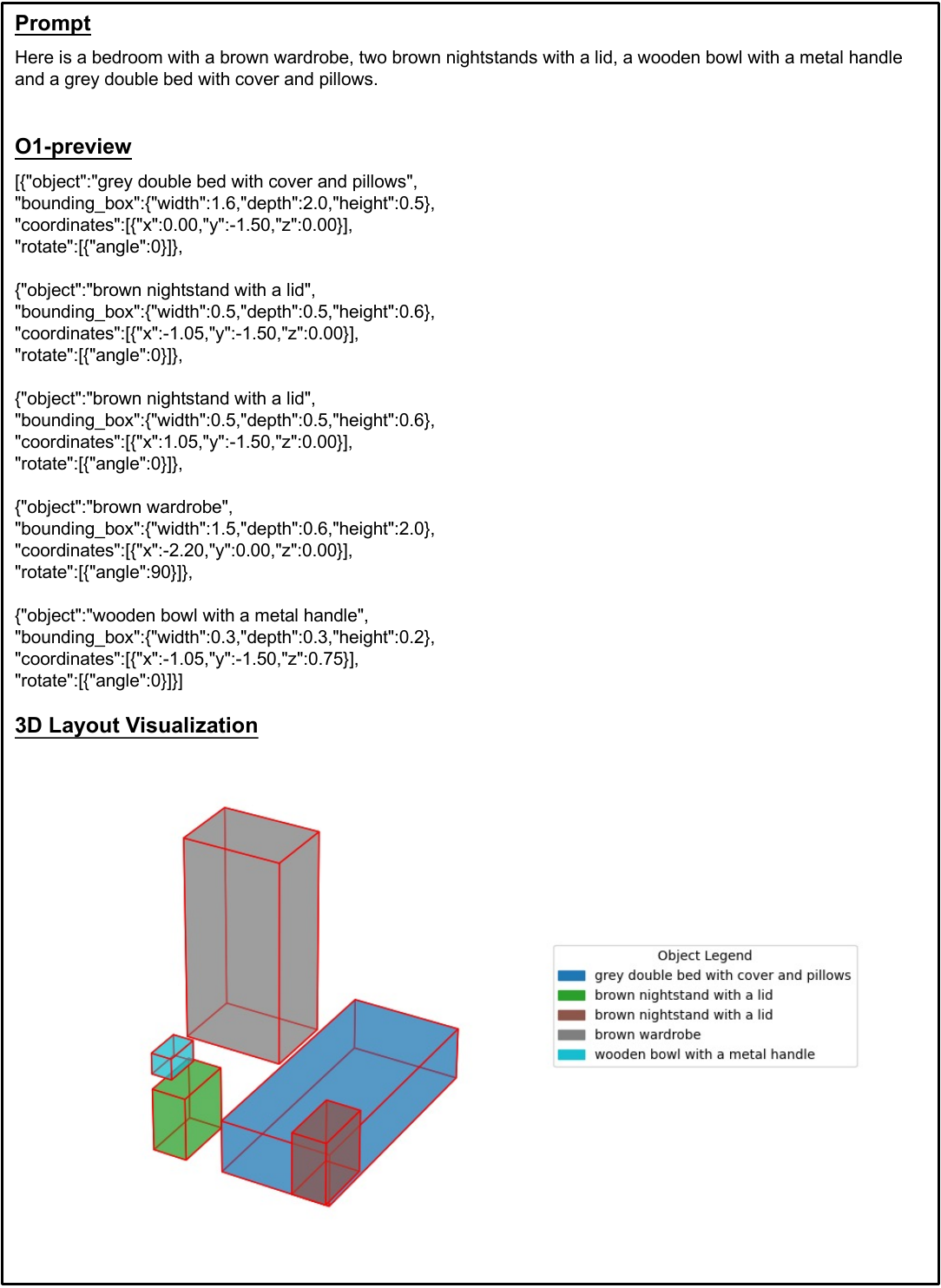}
    \caption[3D Layout Generation: Case 1]{\textbf{3D Layout Generation: Case 1.} Generation of a bedroom layout.}
    \label{fig:Layout_1}
\end{figure}

\begin{figure}[H]
    \centering
    \includegraphics[width = 0.85\textwidth]{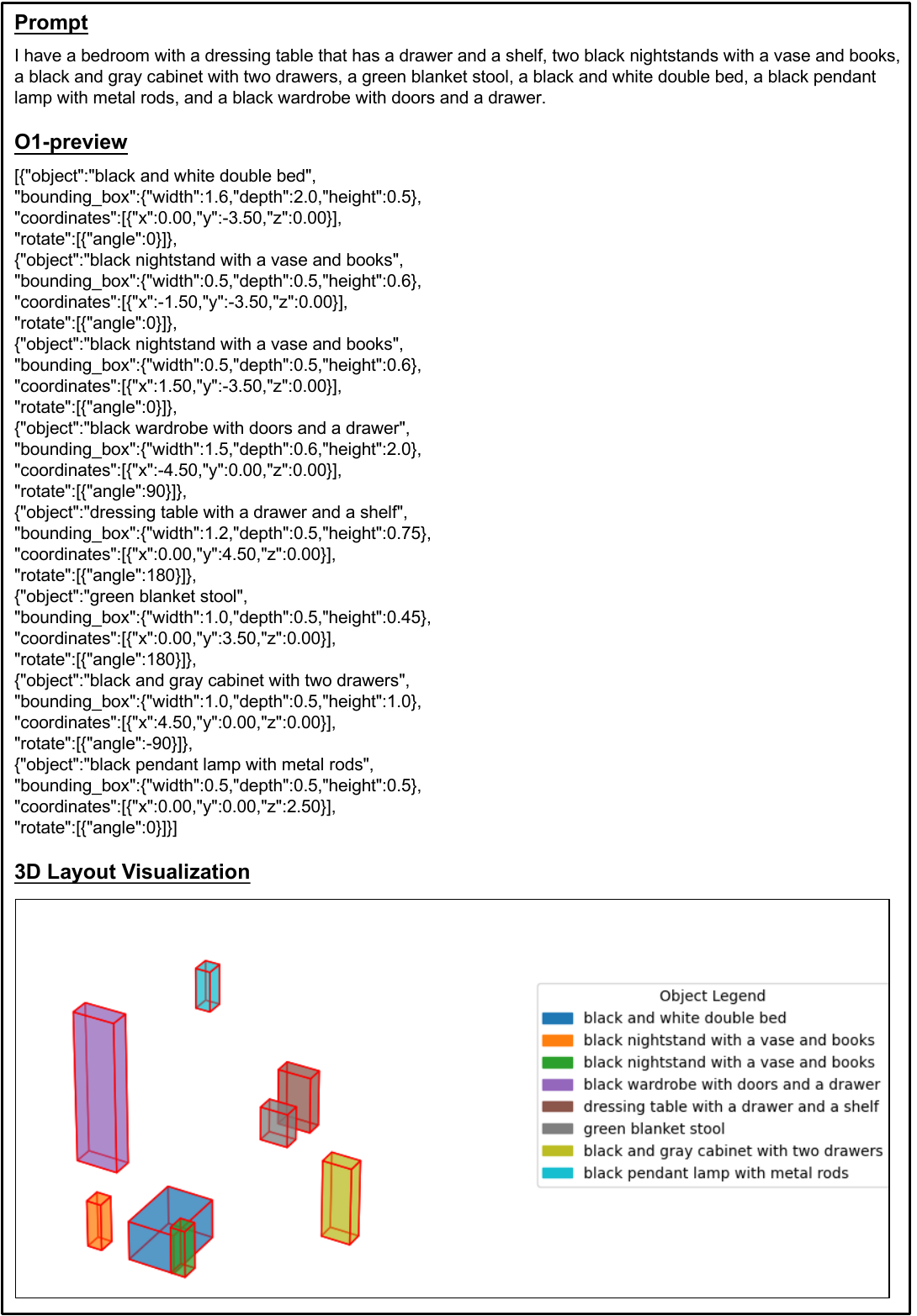}
    \caption[3D Layout Generation: Case 2]{\textbf{3D Layout Generation: Case 2.} Generation of a bedroom layout.}
    \label{fig:Layout_2}
\end{figure}

\begin{figure}[H]
    \centering
    \includegraphics[width = 0.85\textwidth]{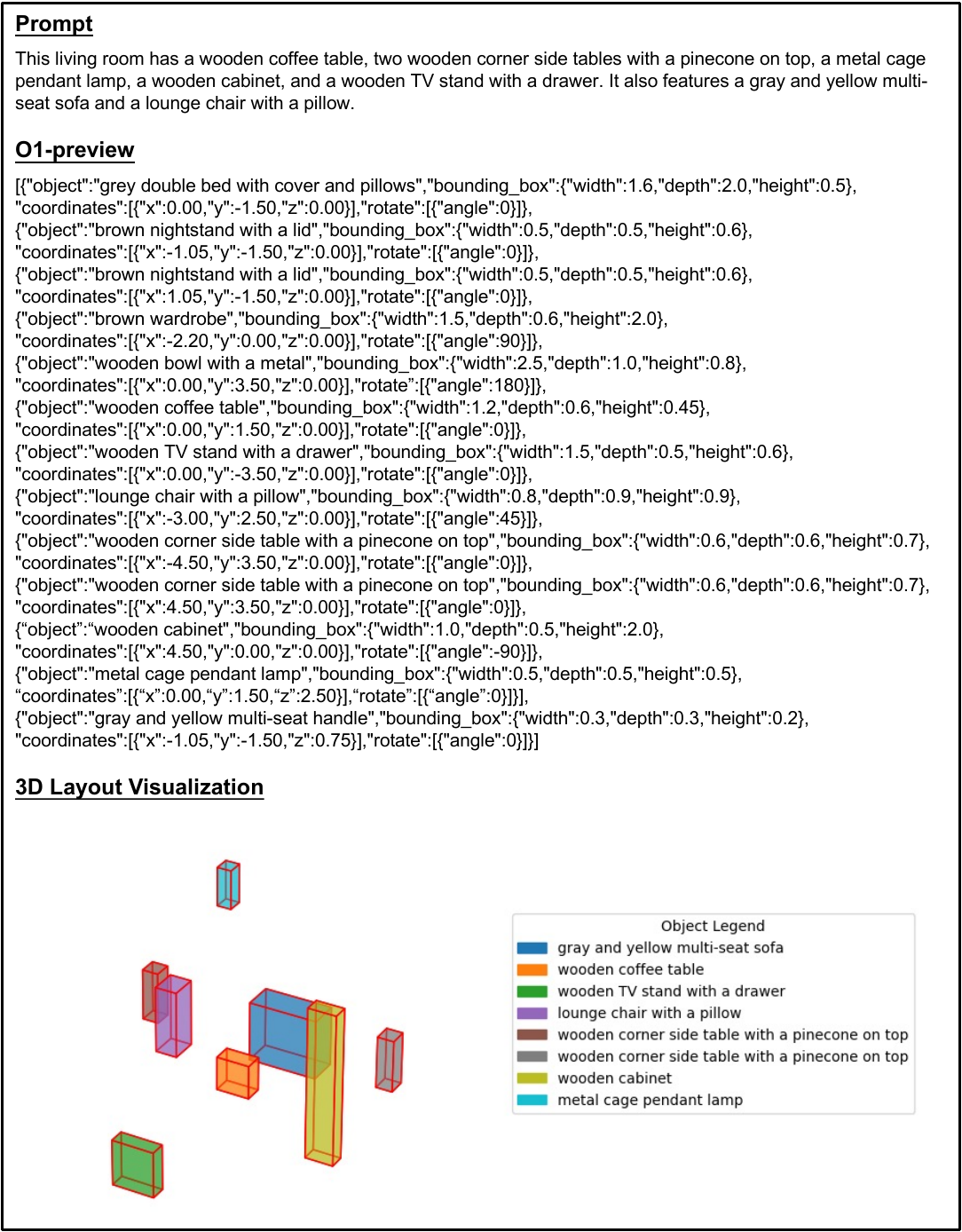}
    \caption[3D Layout Generation: Case 3]{\textbf{3D Layout Generation: Case 3.} Generation of a living room layout.}
    \label{fig:Layout_3}
\end{figure}

\begin{figure}[H]
    \centering
    \includegraphics[width = 0.85\textwidth]{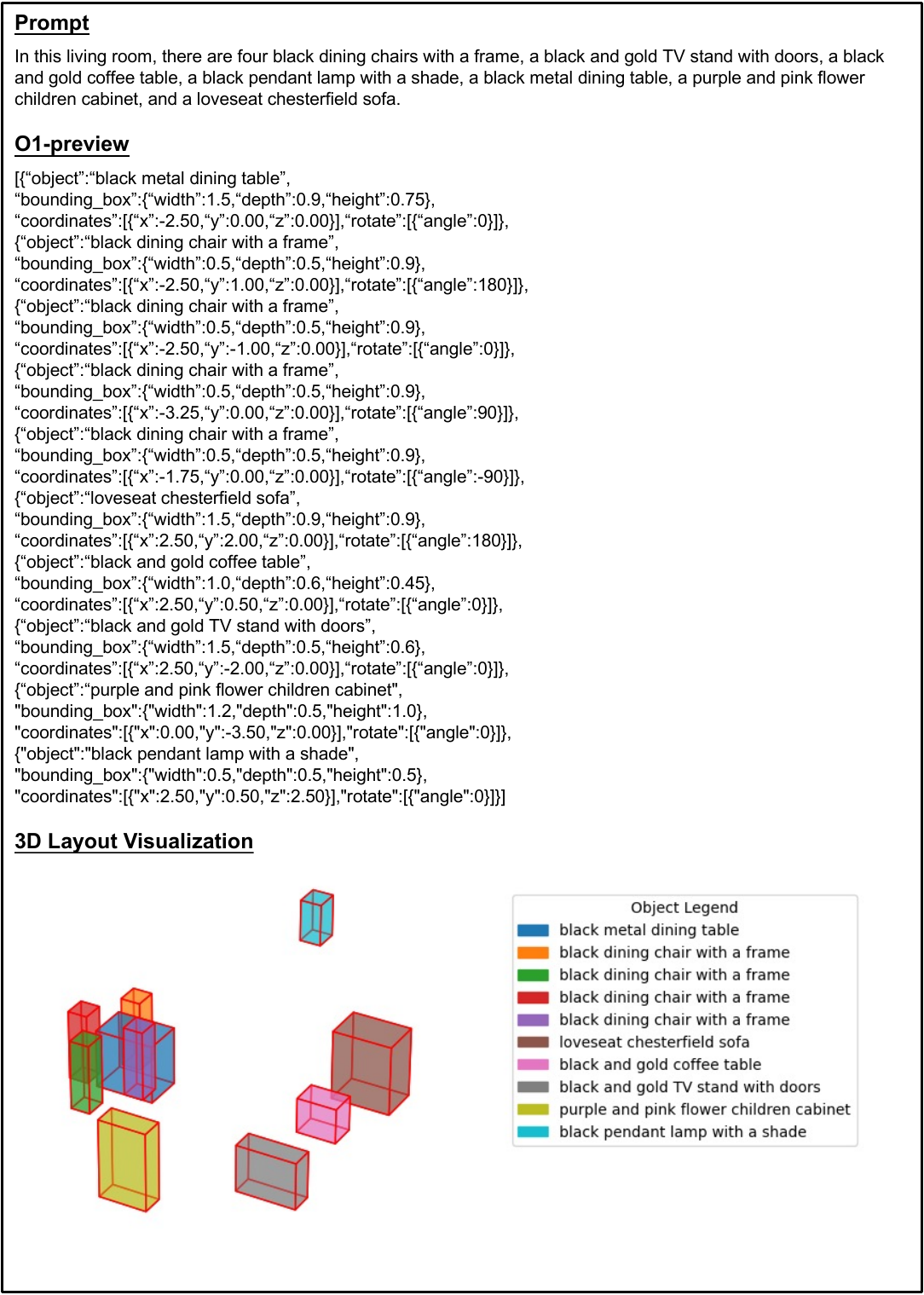}
    \caption[3D Layout Generation: Case 4]{\textbf{3D Layout Generation: Case 4.} Generation of a living room layout.}
    \label{fig:Layout_4}
\end{figure}

However, it is necessary to rigorously point out that in a small number of examples, there were instances of object overlap in three-dimensional space, indicating that the results were not entirely accurate. To comprehensively evaluate the model’s performance, we conducted three-dimensional visualizations of each generated layout; specific details are shown in Figures \ref{fig:Layout_1}, \ref{fig:Layout_2}, \ref{fig:Layout_3}, \ref{fig:Layout_4}. Additionally, we reported the three-dimensional spatial coordinates and rotation angles of each object in the required detailed JSON format. 

In summary, \textit{o1-preview} demonstrated strong capabilities in the task of room layout generation. Although it accurately understood the requirements of the prompt and reflected a high level of functionality and aesthetics in the generated layouts, there were still some minor inaccuracies, such as overlaps in certain examples, indicating room for improvement. The model has broad application prospects in fields such as interior design and virtual reality, but future work should focus on enhancing its accuracy in complex layouts.

\newpage

\newpage

\newpage

\newpage

\newpage

\subsection{Chip Design}
In this study, we conducted a comprehensive evaluation of OpenAI’s o1-preview model within the context of chip design, a domain where LLMs show considerable promise but have yet to be fully explored. Our focus was on testing the model’s performance across three critical dimensions: Engineering Assistant Chatbot, Electronic Design Automation (EDA) Script Generation, and Bug Summary \& Analysis. These dimensions represent pivotal tasks in semiconductor design and production, requiring deep technical expertise, problem-solving abilities, and precision—all areas where advanced LLMs could potentially offer transformative benefits.

Given the lack of standardized datasets specific to chip design, we adopted the ChipNeMo~\cite{liu2023chipnemo} test sample, a pioneering example of domain-adapted LLMs in this field, as a reference framework. This allowed us to rigorously assess o1-preview’s capabilities while making meaningful comparisons between its performance and that of ChipNeMo, a model specifically adapted for chip design.

\subsubsection{Engineering Assistant Chatbot}
\begin{figure}[H]
    \centering
    \includegraphics[width = \textwidth]{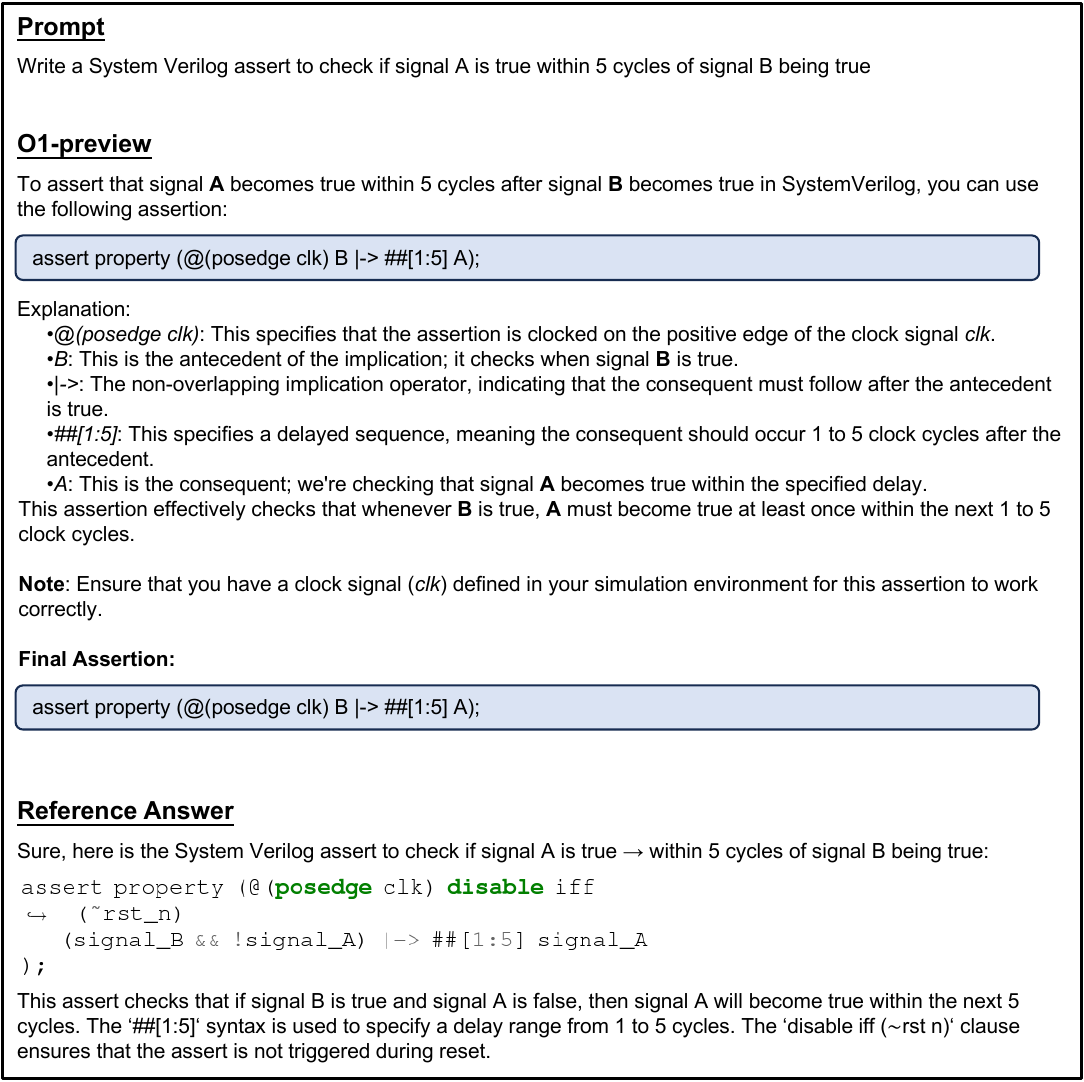}
    \caption[Chip Design-Engineering Assistant Chatbot: Case 1]{\textbf{Chip Design-Engineering Assistant Chatbot: Case 1.} The comparison between ChipNeMo and o1-preview in generating a SystemVerilog assertion to verify signal A within 5 cycles of signal B. o1-preview’s response offered additional context, addressing clocking environments and real-world scenarios, making it more adaptable to complex design verification processes.}
    \label{fig:chip1}
\end{figure}

In this section, we evaluate o1-preview's performance as an engineering assistant chatbot. In the first example (Figure~\ref{fig:chip1}), the prompt asks how to run multiple seeds of random tests using the testgen framework. ChipNeMo’s response was functional but relatively basic, offering a clear, step-by-step approach to configuring the random test seeds. However, it lacked deeper insights into how this process could be optimized, especially when scaling for large simulations—a crucial factor in chip design where extensive random testing is essential for validating design robustness. On the other hand, o1-preview’s response was more comprehensive. In addition to covering the basic configuration steps, it offered a deeper explanation on how to parallelize the test process, ensuring that the tests were not only random but also scalable. This detailed approach reflects a higher level of expertise, as it anticipates potential challenges such as random number generation integrity and error handling during large-scale simulations, which are critical for efficient testing in high-stakes environments like semiconductor design. From a professional chip design perspective, o1-preview’s response demonstrates superior problem-solving skills, particularly in managing the intricacies of large-scale test simulations, making it the more valuable solution for real-world engineering scenarios.

\begin{figure}[H]
    \centering
    \includegraphics[width = .95\textwidth]{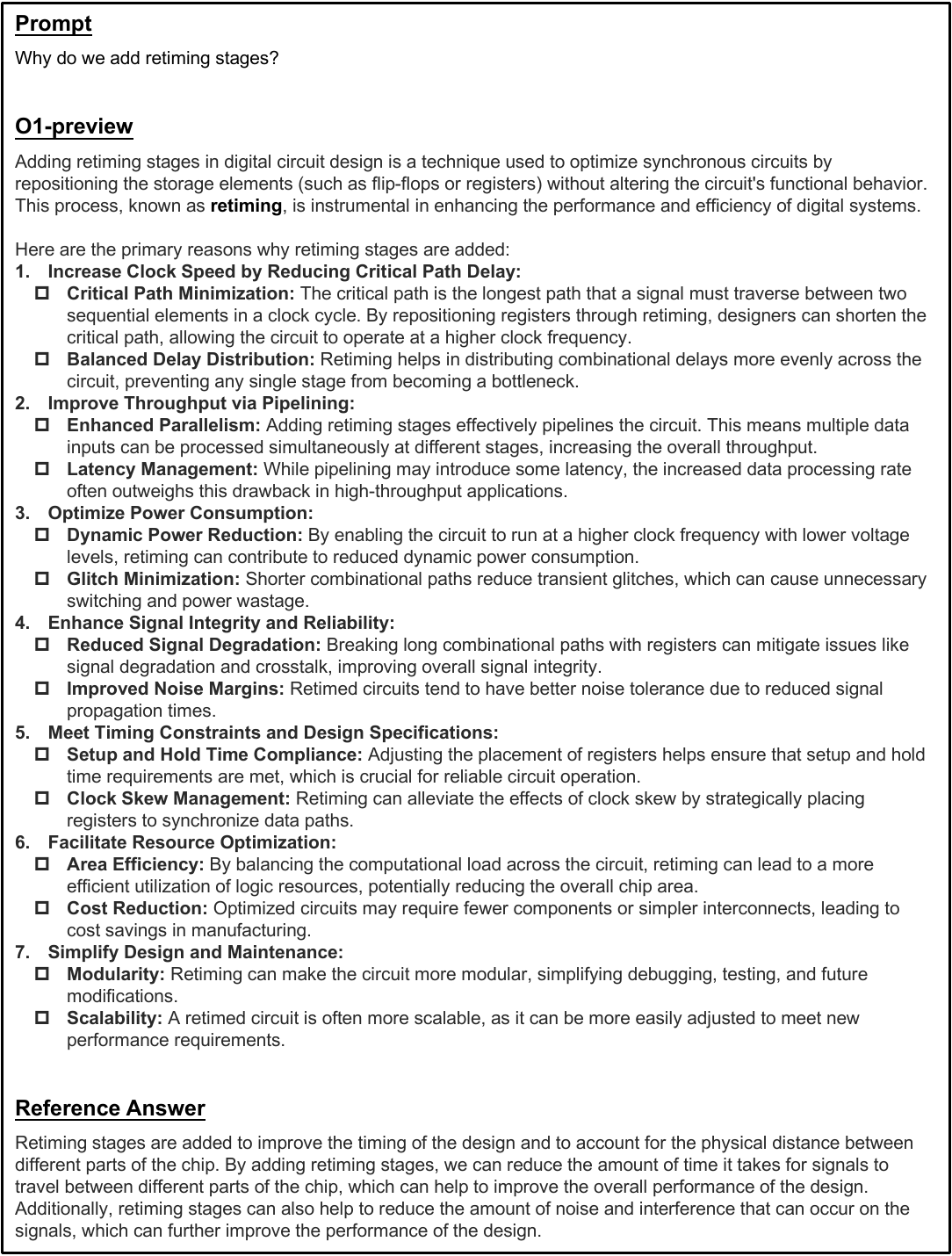}
    \caption[Chip Design-Engineering Assistant Chatbot: Case 2]{\textbf{Chip Design-Engineering Assistant Chatbot: Case 2.} This figure illustrates the responses from ChipNeMo and o1-preview on the purpose of adding retiming stages. o1-preview provided a comprehensive discussion on the trade-offs and broader context, offering a more expert-level analysis suitable for optimizing performance and resource utilization in high-frequency digital designs. }
    \label{fig:chip2}
\end{figure}

In the second example (Figure~\ref{fig:chip2}), the task was to write a SystemVerilog assertion to check if signal A becomes true within five clock cycles of signal B being true. ChipNeMo’s response was technically correct, providing a solid and clear assertion that could be implemented effectively. However, while the explanation was accurate, it did not delve into any advanced verification challenges, such as handling asynchronous inputs or potential pitfalls that might arise in different clocking environments. o1-preview, in contrast, not only provided a correct assertion but also took the opportunity to explain how this assertion could behave in different scenarios, such as varying clock domains and frequency mismatches. Additionally, o1-preview discussed how engineers could modify the assertion for more complex verification setups, ensuring that it accounted for edge cases that are common in real-world chip designs. This level of detail and foresight is indicative of a deeper understanding of the verification process, making o1-preview’s response significantly more useful to engineers dealing with complex, time-sensitive verification tasks in chip design.

\begin{figure}[H]
    \centering
    \includegraphics[width = .98\textwidth]{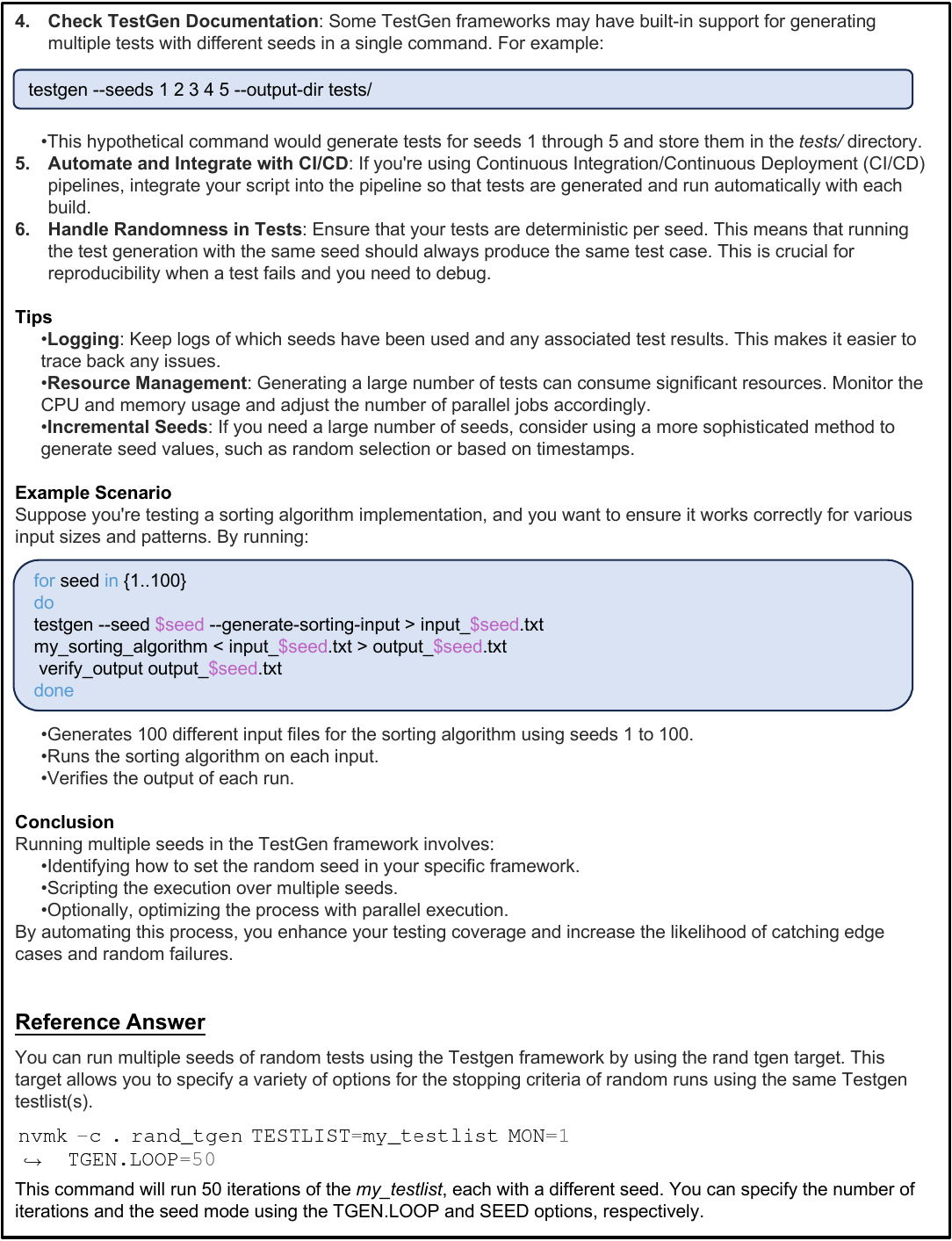}
    \caption[Chip Design-Engineering Assistant Chatbot: Case 3]{\textbf{Chip Design-Engineering Assistant Chatbot: Case 3.} The comparison between ChipNeMo and o1-preview in addressing how to run multiple seeds of random tests using the testgen framework. o1-preview offered a detailed, scalable solution, highlighting its ability to handle large-scale random testing, a critical requirement in design verification.   }
    \label{fig:chip3}
\end{figure}

In the third example (Figure~\ref{fig:chip3}), the query was about retiming stages and their purpose in digital circuit design. ChipNeMo provided a thorough response, outlining the key benefits of retiming, such as reducing critical path delays, improving throughput through pipelining, and optimizing power consumption. While this explanation was comprehensive, it presented a list of advantages in a somewhat generic manner without exploring the more intricate trade-offs involved in retiming decisions. o1-preview, however, offered a more nuanced response. In addition to explaining the same core benefits, it discussed the potential drawbacks of retiming, such as increased latency in certain paths or complications with meeting tight setup and hold time requirements. Furthermore, o1-preview offered advice on how to balance retiming with other optimization techniques, such as logic restructuring, to achieve the best overall design outcome. This consideration of trade-offs and broader optimization strategies demonstrates a more expert-level understanding of the circuit design process. From an engineering standpoint, o1-preview’s ability to not only highlight the advantages but also address the potential downsides of retiming provides a far more insightful and practical response.

\subsubsection{EDA Script Generation}
In the context of EDA Script Generation, both o1-preview and ChipNeMo were tested on their ability to generate scripts for EDA tools, a critical task in chip design automation. These scripts are pivotal in automating various stages of chip development, such as logic verification and circuit optimization, which require both precise coding and a deep understanding of the chip design environment.

\begin{figure}[H]
    \centering
    \includegraphics[width = \textwidth]{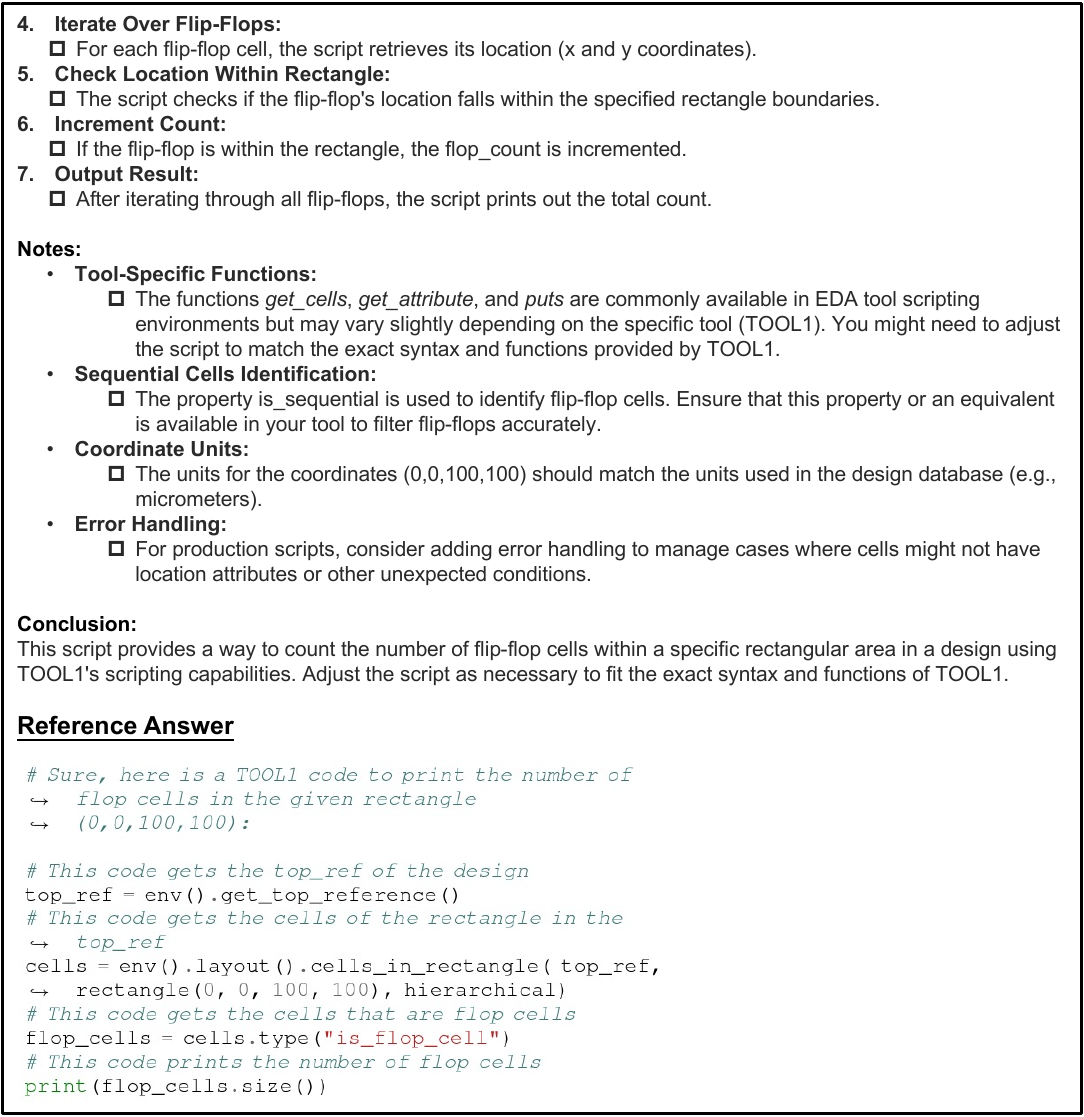}
    \caption[Chip Design-EDA Script Generation: Case 1]{\textbf{Chip Design-EDA Script Generation: Case 1.} This figure demonstrates the responses of ChipNeMo and o1-preview when tasked with generating a TOOL1 code to count flip-flop cells in a defined region. o1-preview extended the solution with performance optimization techniques for large-scale designs, offering deeper insight into real-world physical design challenges.}
    \label{fig:chip4}
\end{figure}

In Figure~\ref{fig:chip4}, one such prompt tasked the models with writing a TOOL1 code to count the number of flip-flop cells within a given rectangular boundary, a common operation in physical design when determining logic distribution and optimizing layout. ChipNeMo’s response provided a basic solution, covering the mechanics of how to traverse the design hierarchy and count the flip-flops in the specified region. While it delivered the correct structure and functionality, the response was somewhat limited in its flexibility. The code did not explore advanced features like handling edge cases (such as overlapping regions or boundary conditions) or optimizing the search algorithm for larger designs, both of which are essential considerations in production environments where efficiency is crucial.

In contrast, o1-preview’s response offered a more nuanced solution. The model not only provided the core code but also discussed potential optimization strategies, such as how to efficiently traverse large designs and how to handle boundary conditions more gracefully. Moreover, o1-preview included comments and suggestions for extending the functionality, such as adding the ability to count other types of cells or integrating the script into a larger automation flow for design rule checking. This response demonstrated a deeper understanding of the complexities involved in real-world EDA environments, where flexibility and scalability are paramount. o1-preview was able to anticipate potential issues that engineers might face when deploying this script in large-scale designs, making its response significantly more applicable for practical use in a professional setting.

\begin{figure}[H]
    \centering
    \includegraphics[width = \textwidth]{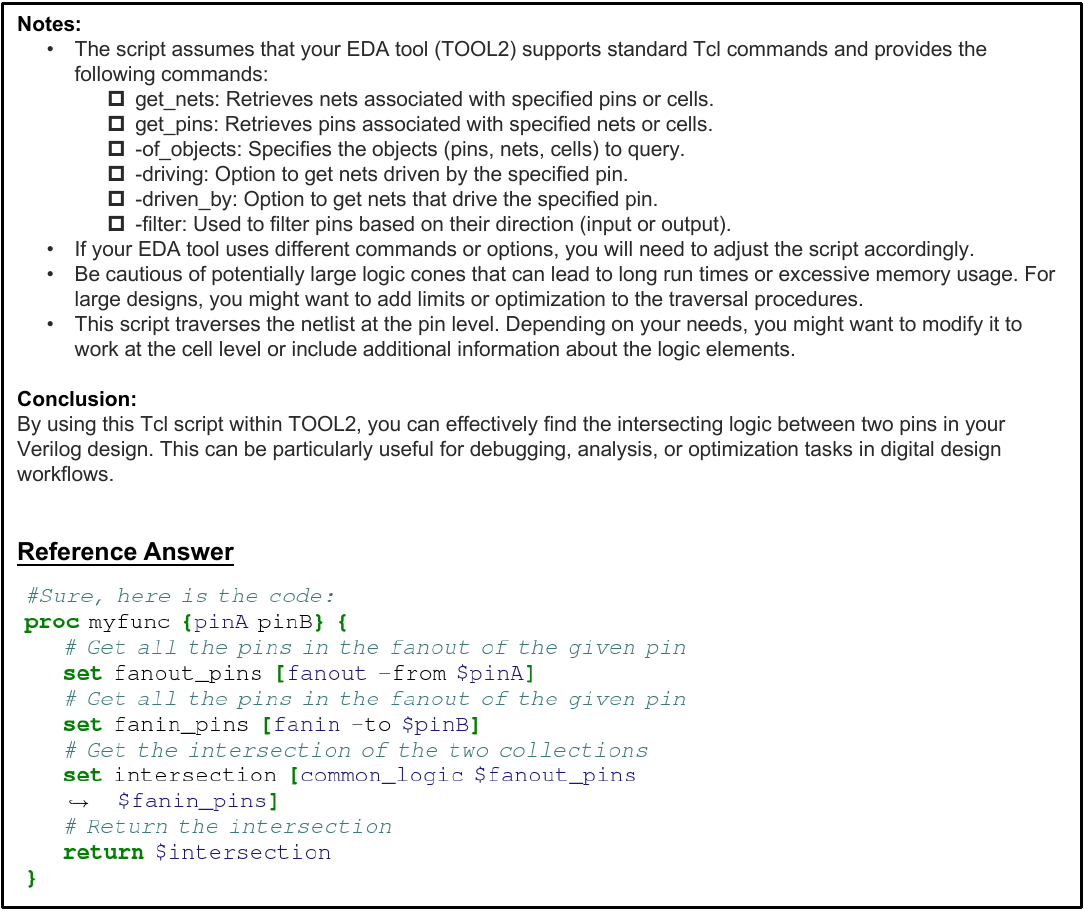}
    \caption[Chip Design-EDA Script Generation: Case 2]{\textbf{Chip Design-EDA Script Generation: Case 2.} The figure shows the generated TCL procedures from ChipNeMo and o1-preview for finding intersecting logic between two pins. o1-preview provided an optimized approach for handling large designs and discussed edge cases, making it more applicable to real-world circuit analysis tasks.}
    \label{fig:chip5}
\end{figure}

Shown in Figure~\ref{fig:chip5}, another query asked the models to write a TOOL2 TCL procedure to find intersecting logic between two pins, a task that typically involves traversing the netlist and analyzing connections between logic elements. ChipNeMo’s response was straightforward, generating a functional TCL procedure that correctly identified the logic elements between two pins. However, the response lacked depth in terms of optimizing the logic traversal for large-scale designs or handling more complex netlist structures with multiple levels of hierarchy, which are often encountered in modern chip design.

o1-preview, on the other hand, delivered a more comprehensive response by not only generating the procedure but also discussing the nuances of handling different types of netlist structures, such as flat versus hierarchical designs. The model provided suggestions on optimizing the traversal algorithm to minimize runtime in large designs, an essential consideration in EDA where scripts often need to handle millions of logic elements. Furthermore, o1-preview anticipated potential edge cases, such as how to handle pins with multiple fan-outs or fan-ins, and proposed extensions to the procedure for handling such scenarios. This level of detail illustrates a greater understanding of the complexities involved in chip design and the need for adaptable, efficient scripts in EDA environments.

\subsubsection{Bug Summary \& Analysis}
\begin{figure}[H]
    \centering
    \includegraphics[width = \textwidth]{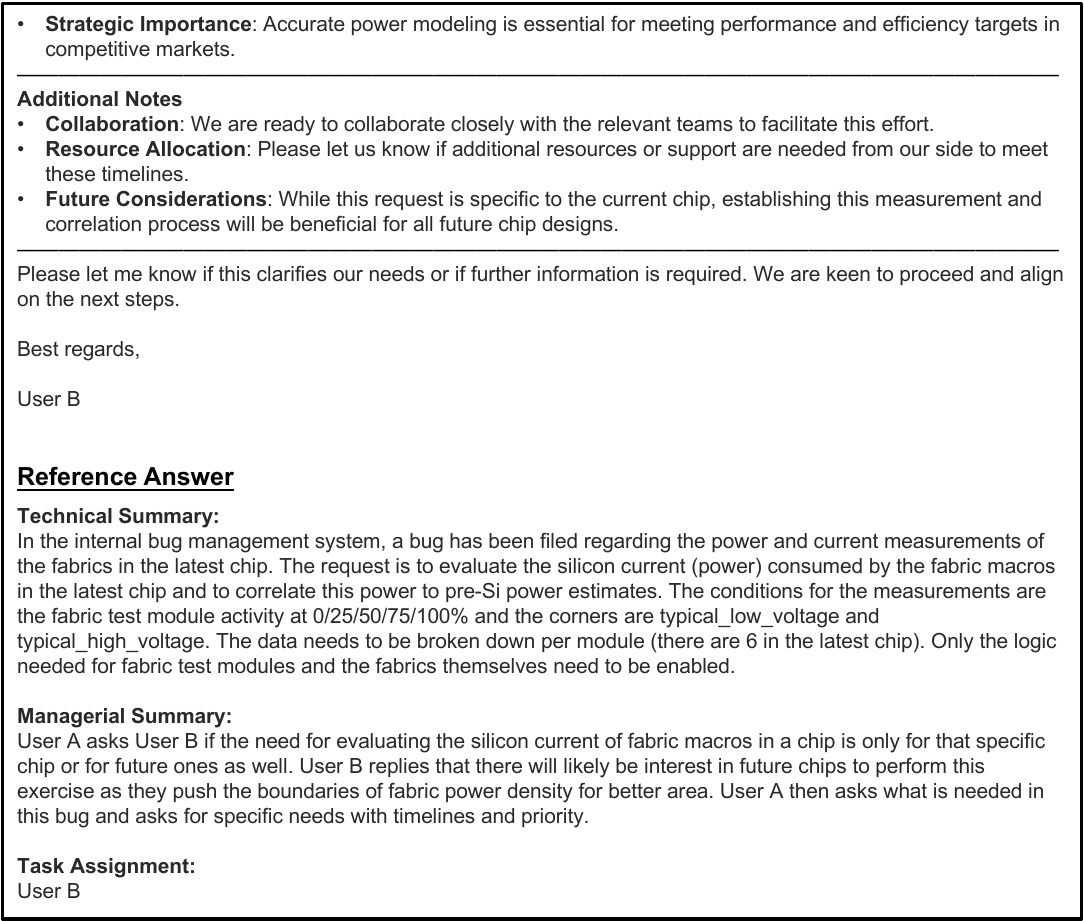}
    \caption[Chip Design-Bug Summary \& Analysis: Case 1]{\textbf{Bug Summary \& Analysis: Case 1.} Bug report analysis comparing ChipNeMo and o1-preview for evaluating power consumed by fabric macros. While ChipNeMo provided a standard summary, o1-preview offered deeper analysis with strategic recommendations for power isolation and task prioritization, essential for power optimization in silicon designs. }
    \label{fig:chip6}
\end{figure}

In the context of Bug Summary \& Analysis, both o1-preview and ChipNeMo were tested on their ability to analyze bug reports related to silicon current measurement in chip fabrics, a crucial task in chip power optimization shown in Figure~\ref{fig:chip6}. Bug analysis in semiconductor design involves not only identifying the root cause of the problem but also summarizing the issue in a way that allows teams to prioritize and resolve it efficiently.

For the given bug, the report concerns a request to measure the power consumption of fabric macros in a chip under varying levels of activity and voltage conditions. This is a typical challenge during the design-for-test (DFT) phase, where engineers need to evaluate how closely the silicon’s power consumption aligns with pre-silicon estimates. The goal is to isolate the power consumed by specific parts of the chip to improve the accuracy of power models and optimize power efficiency.

ChipNeMo’s response was technically sound but somewhat formulaic in structure. It provided both a technical summary and a managerial summary, effectively breaking down the core components of the issue. The technical summary outlined the conditions for power measurement, such as test module activity percentages and the required voltage corners. However, the managerial summary simply relayed the back-and-forth discussion between users without delving into any deeper insights or offering a clear action plan. ChipNeMo’s response focused heavily on summarizing the immediate facts of the bug report without providing much context or additional analysis on the implications of the power measurement for future designs or suggesting strategies for correlating silicon data with pre-silicon estimates.

In contrast, o1-preview’s response offered a much more thorough analysis of the bug. While covering the same technical and managerial aspects as ChipNeMo, o1-preview also addressed the broader implications of the power measurement. It analyzed how the fabric power data could influence future chip designs, particularly in optimizing power density and improving area efficiency—factors that are critical in modern chip development. Furthermore, o1-preview provided a more in-depth interpretation of the request, offering suggestions on how to approach the power measurement with more refined isolation techniques and data breakdowns to ensure more accurate comparisons with pre-silicon estimates. It also provided a clearer task assignment framework, ensuring that the engineering teams could prioritize the tasks with timelines and specific milestones for isolating and measuring the fabric macros’ power consumption.

From a bug analysis perspective, o1-preview not only responded to the immediate needs of the bug report but also demonstrated a deeper understanding of the challenges posed by such tasks in real-world chip development. Its response took into account both the technical and strategic dimensions of the issue, offering a more valuable solution for engineers tasked with optimizing chip power consumption. By anticipating potential challenges and offering concrete suggestions on data analysis and task delegation, o1-preview demonstrated superior problem-solving capabilities compared to ChipNeMo.

In conclusion, across the three critical tasks—Engineering Assistant Chatbot, EDA Script Generation, and Bug Summary \& Analysis—o1-preview consistently demonstrated superior problem-solving capabilities, depth of analysis, and practical relevance compared to ChipNeMo. In the Engineering Assistant Chatbot task, o1-preview excelled by offering more comprehensive solutions, anticipating edge cases, and providing optimization insights that reflected real-world engineering challenges. In EDA Script Generation, o1-preview again showed a greater understanding of scalability and flexibility, delivering solutions that could be adapted to complex, large-scale designs. Its ability to optimize code and handle intricate netlist structures demonstrated advanced reasoning that would be invaluable to engineers working on chip automation tasks. Finally, in Bug Summary \& Analysis, o1-preview’s thorough analysis and strategic foresight allowed it to offer not just a summary of the bug but also suggestions for power optimization and task prioritization. This comprehensive approach underscores o1-preview’s capacity to engage with both the technical and managerial aspects of chip design, making it a powerful tool in handling complex semiconductor development tasks. Ultimately, o1-preview’s consistent performance highlights its potential to play a pivotal role in advancing chip design automation and enhancing engineering workflows.
\newpage

\newpage
\newpage
\begin{figure}[H]
    \centering
    \includegraphics[width = .98\textwidth]{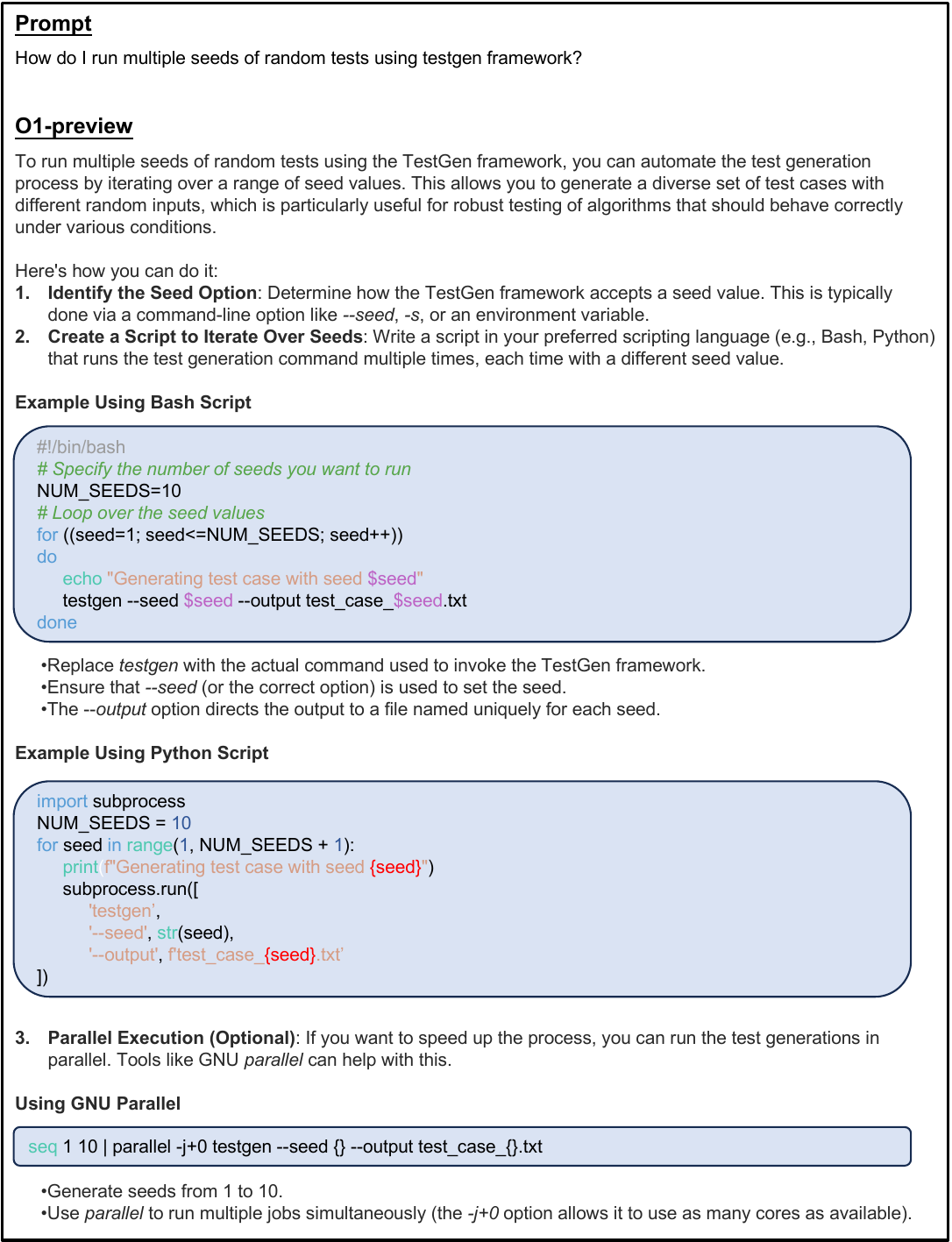}
    \label{fig:chip3-1}
\end{figure}
\newpage
\begin{figure}[H]
    \centering
    \includegraphics[width = \textwidth]{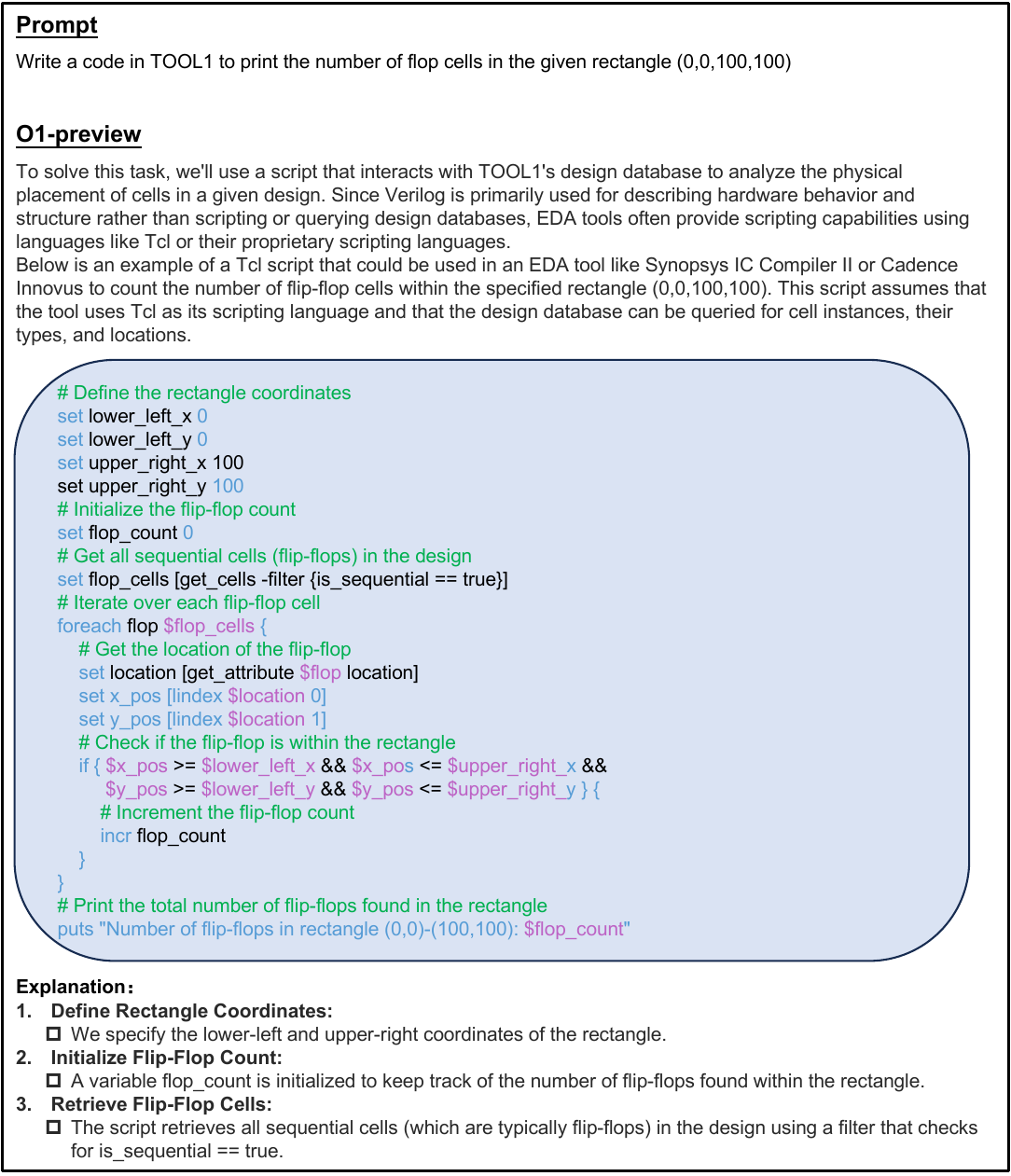}
    \label{fig:chip4-1}
\end{figure}
\newpage
\begin{figure}[H]
    \centering
    \includegraphics[width = \textwidth]{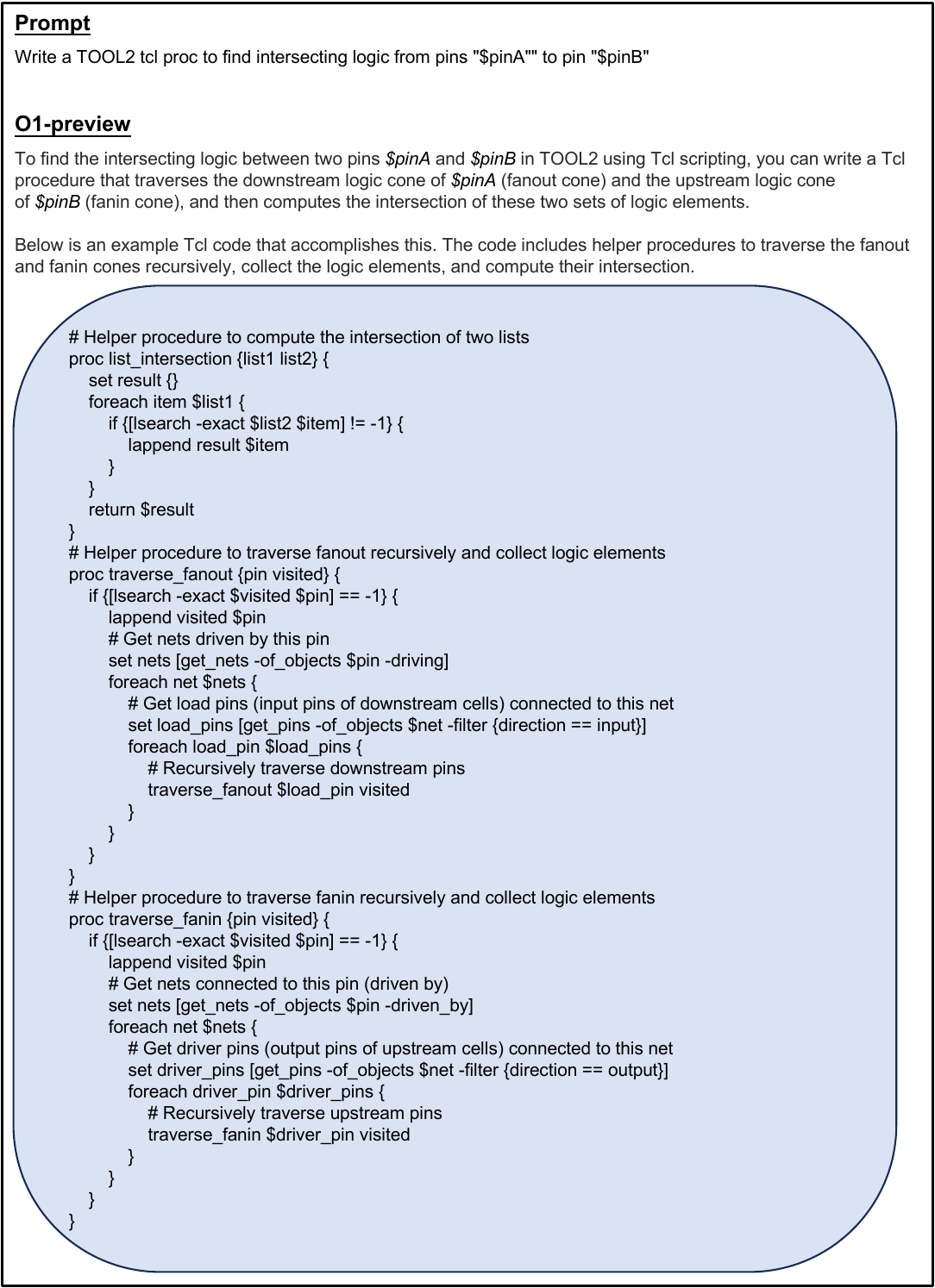}
    \label{fig:chip5-1}
\end{figure}
\begin{figure}[H]
    \centering
    \includegraphics[width = \textwidth]{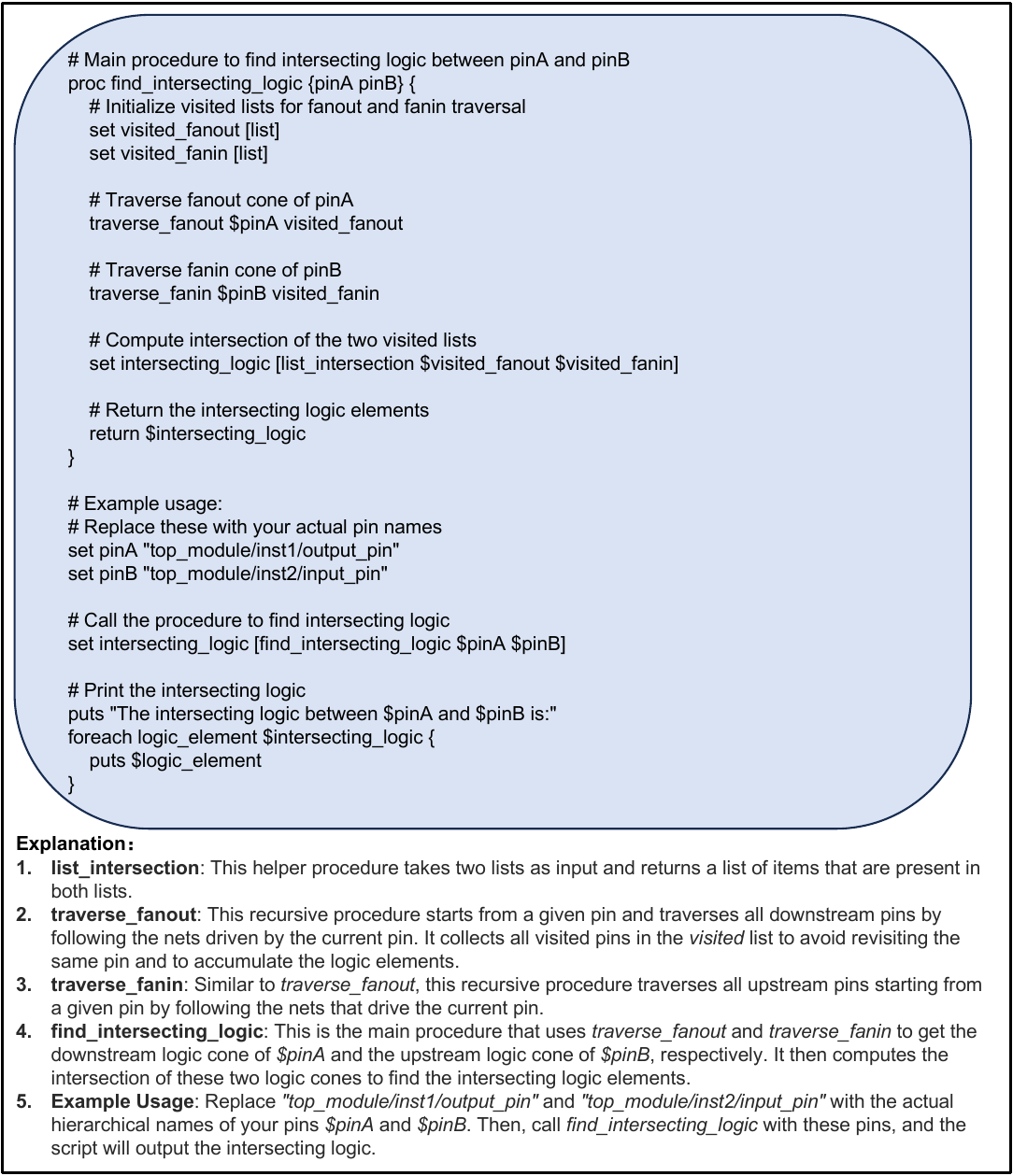}
    \label{fig:chip5-2}
\end{figure}
\newpage
\begin{figure}[H]
    \centering
    \includegraphics[width = \textwidth]{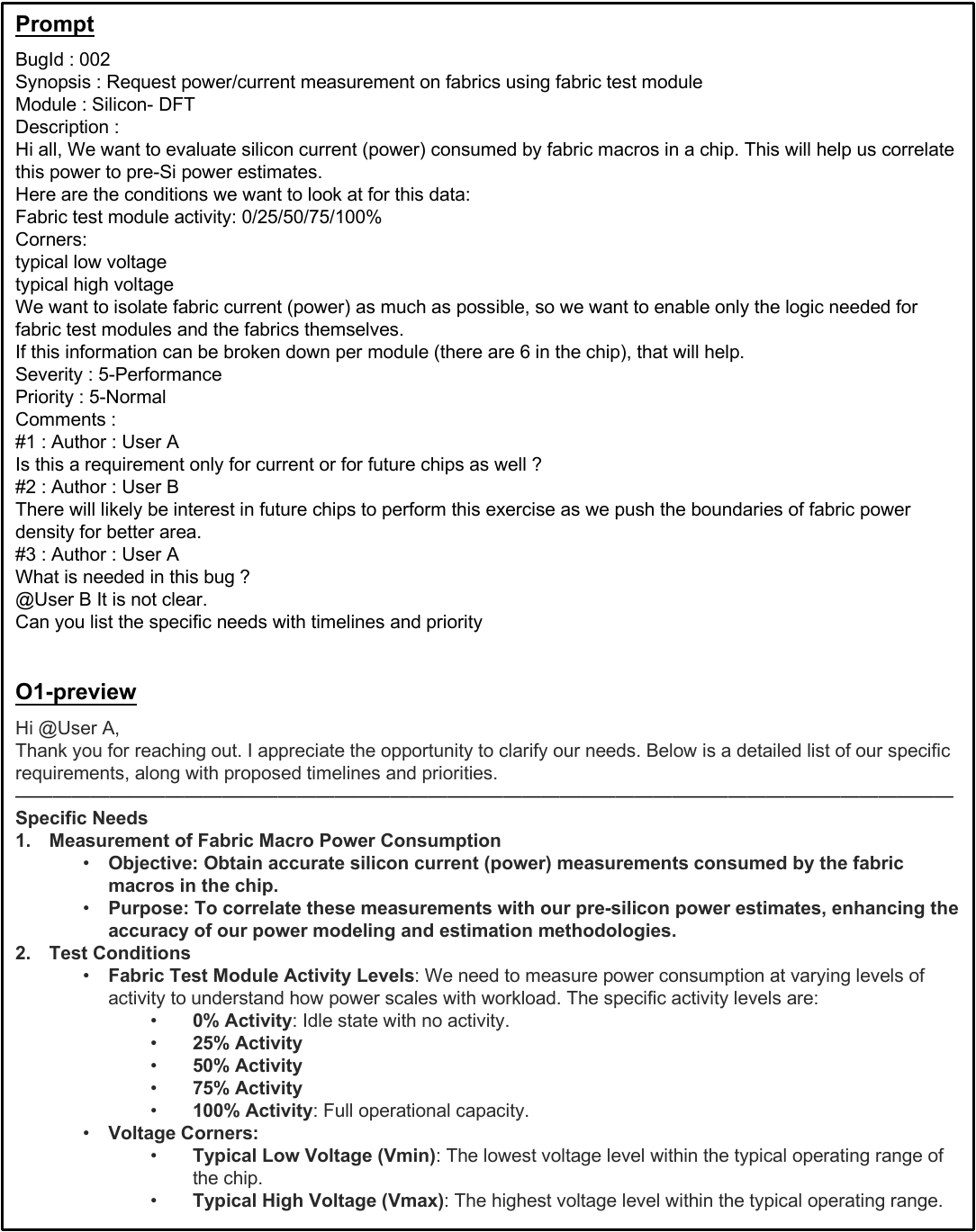}
    \label{fig:chip6-1}
\end{figure}
\begin{figure}[H]
    \centering
    \includegraphics[width = \textwidth]{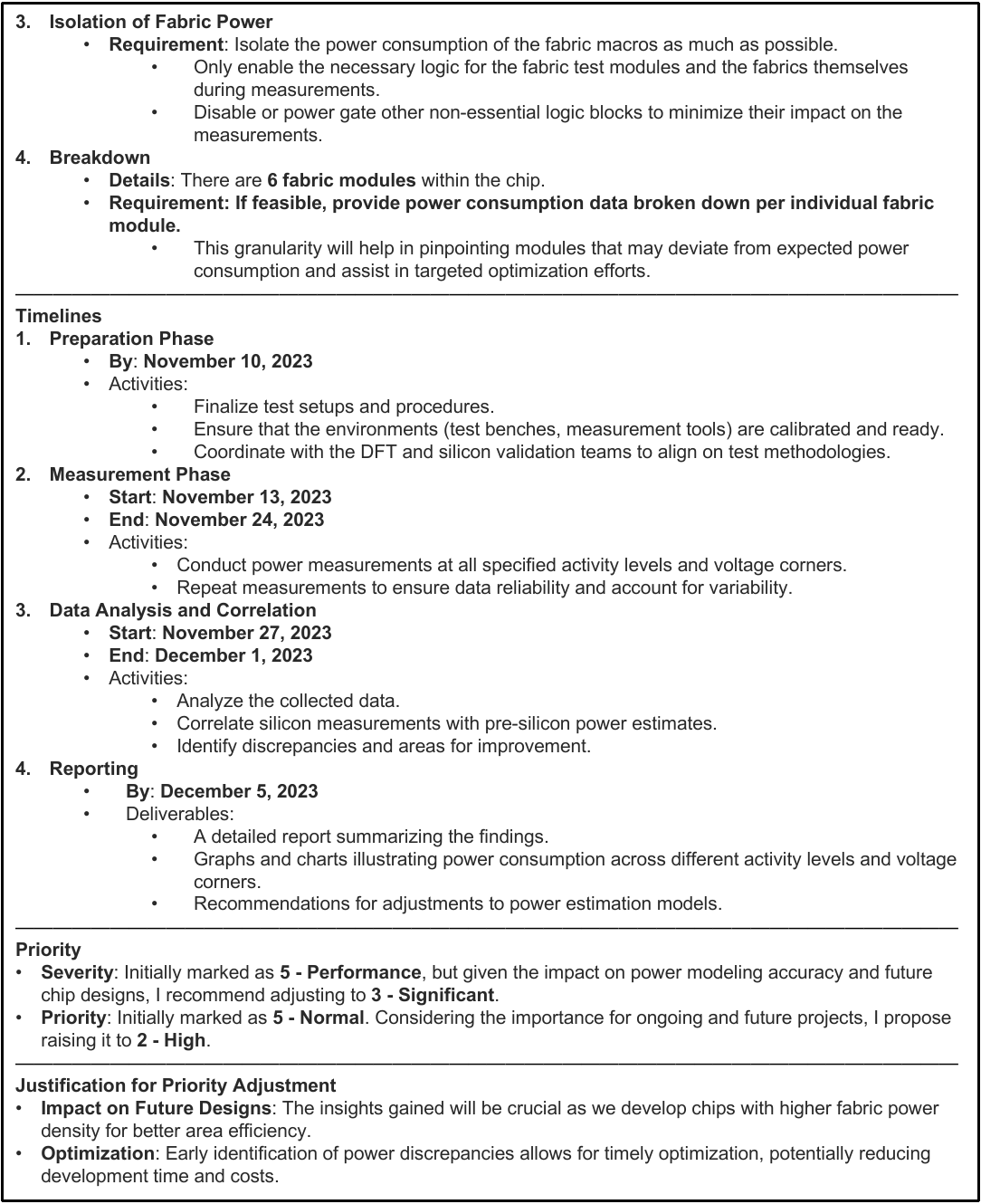}
    \label{fig:chip6-2}
\end{figure}
\newpage

\subsection{Logical Reasoning}
o1-preview has demonstrated powerful capabilities in the tests of logical reasoning problems. It has abundant knowledge related to logical reasoning. Judging from the test results, whether it is a regular logical reasoning question or a complex logical reasoning problem with a certain degree of difficulty, such as those involving multi - level nested reasoning or multi - condition comprehensive judgment, it can provide very accurate and well - organized analysis processes and answers, and its analysis ability is extremely outstanding. 

\begin{figure}[H]
    \centering
    \includegraphics[width = \textwidth]{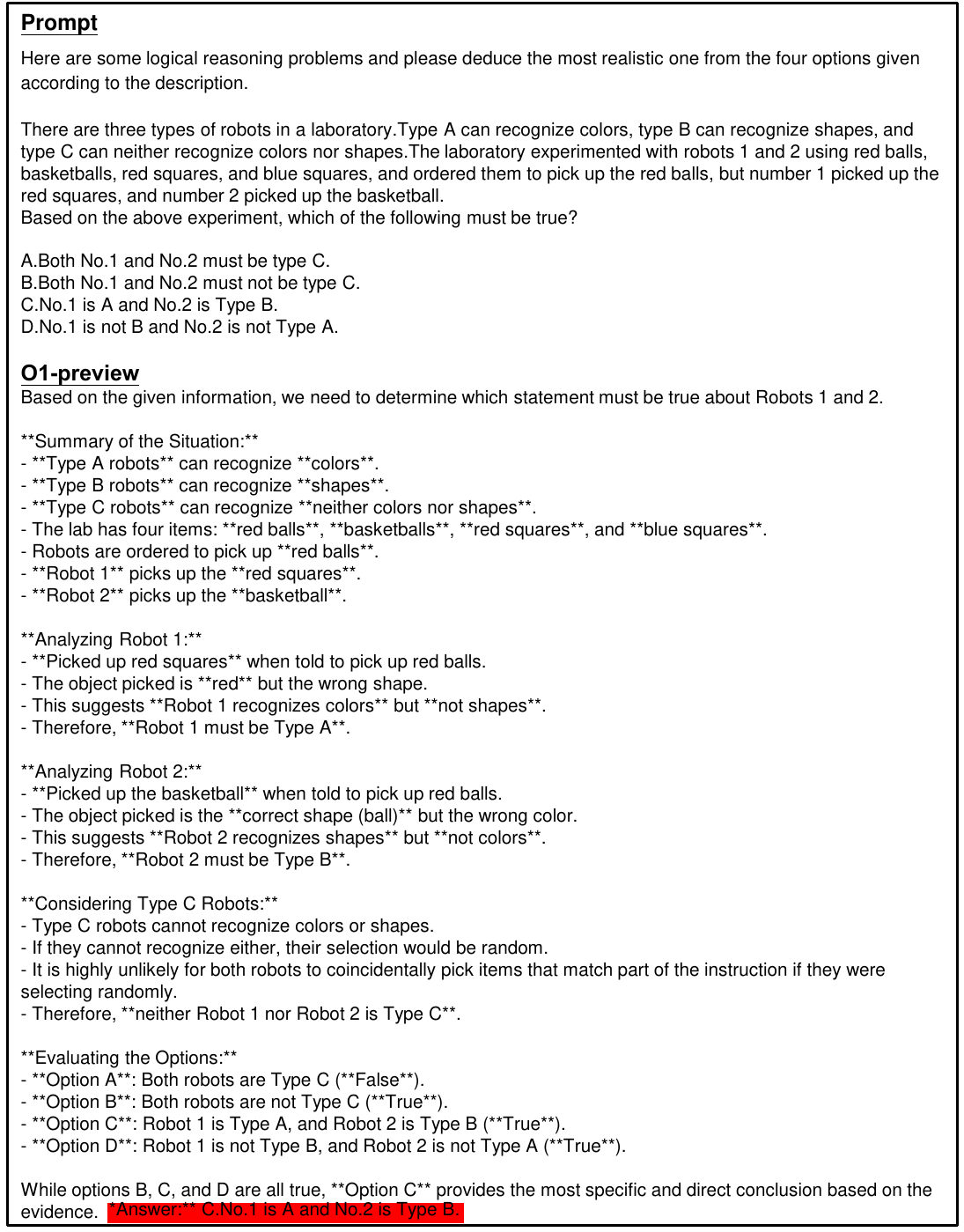}
        \caption[Logical Reasoning: Case 2]{\textbf{Logical Reasoning: Case 2.} There are also test results that are different from the answers provided on the official website. From the perspective of the standard answers of the test questions, these are failed samples.  }
    \label{fig:Logical Reasoning: Case 2}
\end{figure}

In the process of in-depth research and testing on the reasoning ability of o1-preview, we have discovered a noteworthy phenomenon. The reasoning of o1-preview cannot be completely mapped accurately to the given standard answer. This implies that in some specific problem scenarios, o1-preview may not be able to provide results that are entirely consistent with the traditional standard answers, such as those in Figures~\ref{fig:Logical Reasoning: Case 2},~\ref{fig:Logical Reasoning: Case 5},~\ref{fig:Logical Reasoning: Case 8},~\ref{fig:Logical Reasoning: Case 9},~\ref{fig:Logical Reasoning: Case 10}. However, despite this shortcoming, the thinking process demonstrated by o1-preview is astonishing. When confronted with various complex reasoning tasks, o1-preview can rapidly analyze and break down problems. It will attempt to examine problems from different angles, just like a perceptive thinker exploring unknown realms. Whether dealing with logical puzzles, making situational inferences, or analyzing complex relationship networks, o1-preview can display its unique way of thinking.For instance, when facing a complex logical reasoning problem, o1-preview may not directly give the conclusion in the standard answer. Instead, it will analyze each element of the problem step by step and propose some novel thinking paths and possibilities. This exploratory thinking process provides us with a new perspective and prompts us to re-examine the essence and solution methods of problems.

\begin{figure}[H]
    \centering
    \includegraphics[width = \textwidth]{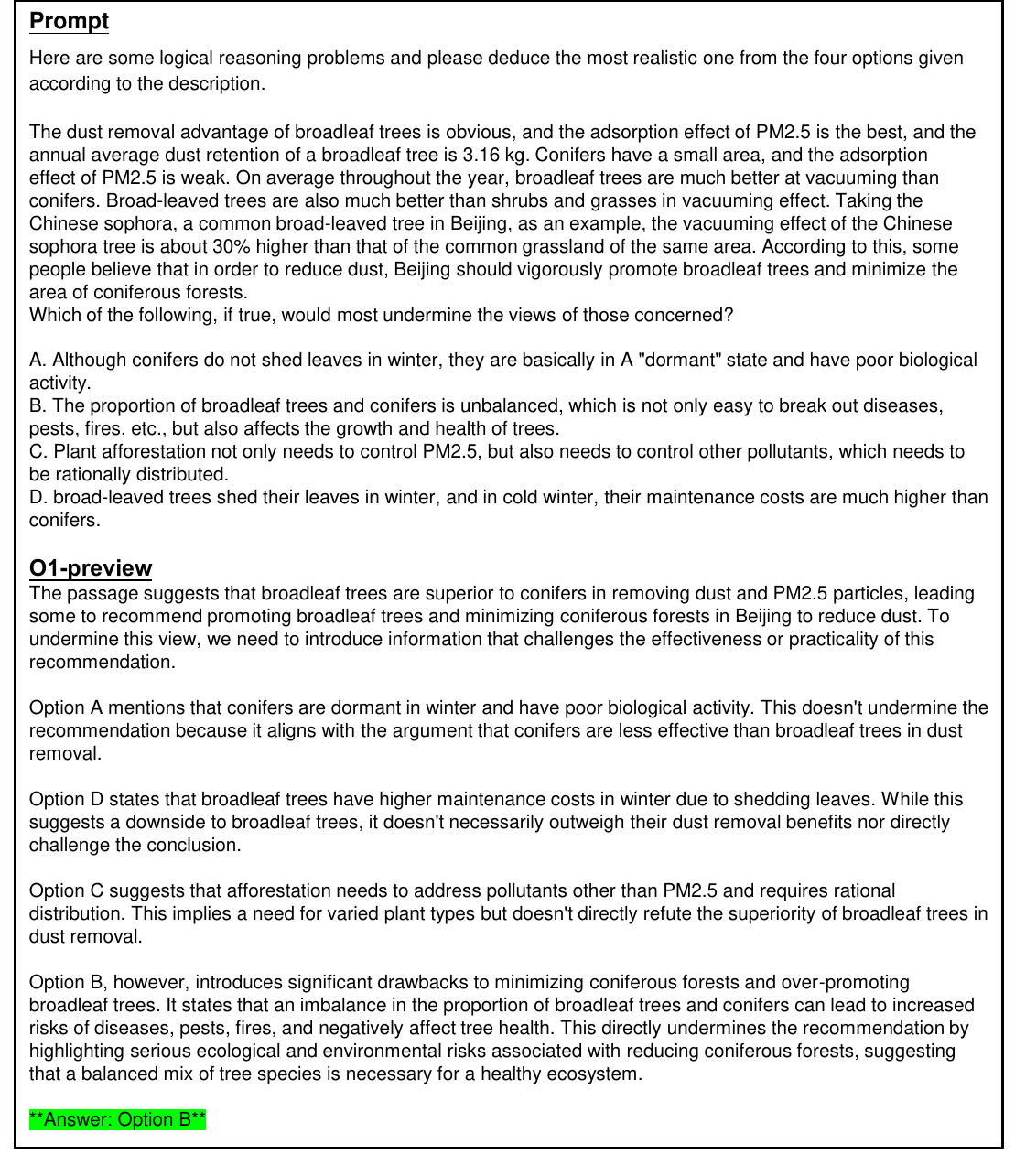}
    \caption[Logical Reasoning: Case 1]{\textbf{Logical Reasoning: Case 1.} In this case, In this sample, the o1-preview model demonstrated excellent logical analysis performance and obtained the answer consistent with the expectation. }
    \label{fig:Logical Reasoning: Case 1}
\end{figure}

It can analyze complex problems, break them down into smaller components, and systematically work through them to arrive at accurate conclusions. Whether it's solving mathematical puzzles, deducing the next step in a sequence, or evaluating the validity of an argument,o1-preview demonstrates a high level of intelligence and precision. See Figures~\ref{fig:Logical Reasoning: Case 1},~\ref{fig:Logical Reasoning: Case 3},~\ref{fig:Logical Reasoning: Case 4},~\ref{fig:Logical Reasoning: Case 6},~\ref{fig:Logical Reasoning: Case 7}. Its ability to handle multiple variables and conditions simultaneously is particularly impressive. It can consider different scenarios, weigh the pros and cons of each option, and make informed decisions. This not only showcases its logical acumen but also its adaptability to various types of problems.

The thinking process of o1-preview also reflects its strong learning and adaptability. Although it may deviate from the standard answer on some problems, by continuously receiving new data and feedback, o1-preview can gradually optimize its reasoning model and continuously approach the correct answer. This continuous improvement ability brings great potential for its future applications. It can present its thought process in a way that is easy to understand, making it a valuable tool for education, research, and problem-solving in various fields. However, although its error rate is relatively low, there are still some cases of mistakes. For example, on some problems that require the exploration of very subtle logical relationships, it may occasionally have deviations in reasoning. For most logical reasoning tests, only the basic information of the problem needs to be provided without particularly detailed guidance, and o1-preview can successfully complete the reasoning analysis. Moreover, it can quickly process a large number of different types of logical reasoning problems. Nevertheless, when facing extremely complex problems involving the intersection of multiple different logical systems, o1-preview may encounter bottlenecks. This is a limitation of o1-preview in handling complex logical reasoning tasks. 

What we would like to point out is that the official standard answer is not necessarily accurate. Instead, the analysis and reasoning process offered by o1-preview is truly convincing. From our perspective, this is far more gratifying than relying on the standard answer.

\newpage

\newpage

\subsection{Table-to-Text Generation}

\begin{figure}[H] 
    \centering
\includegraphics[width=\textwidth, height=0.9\textheight, keepaspectratio]{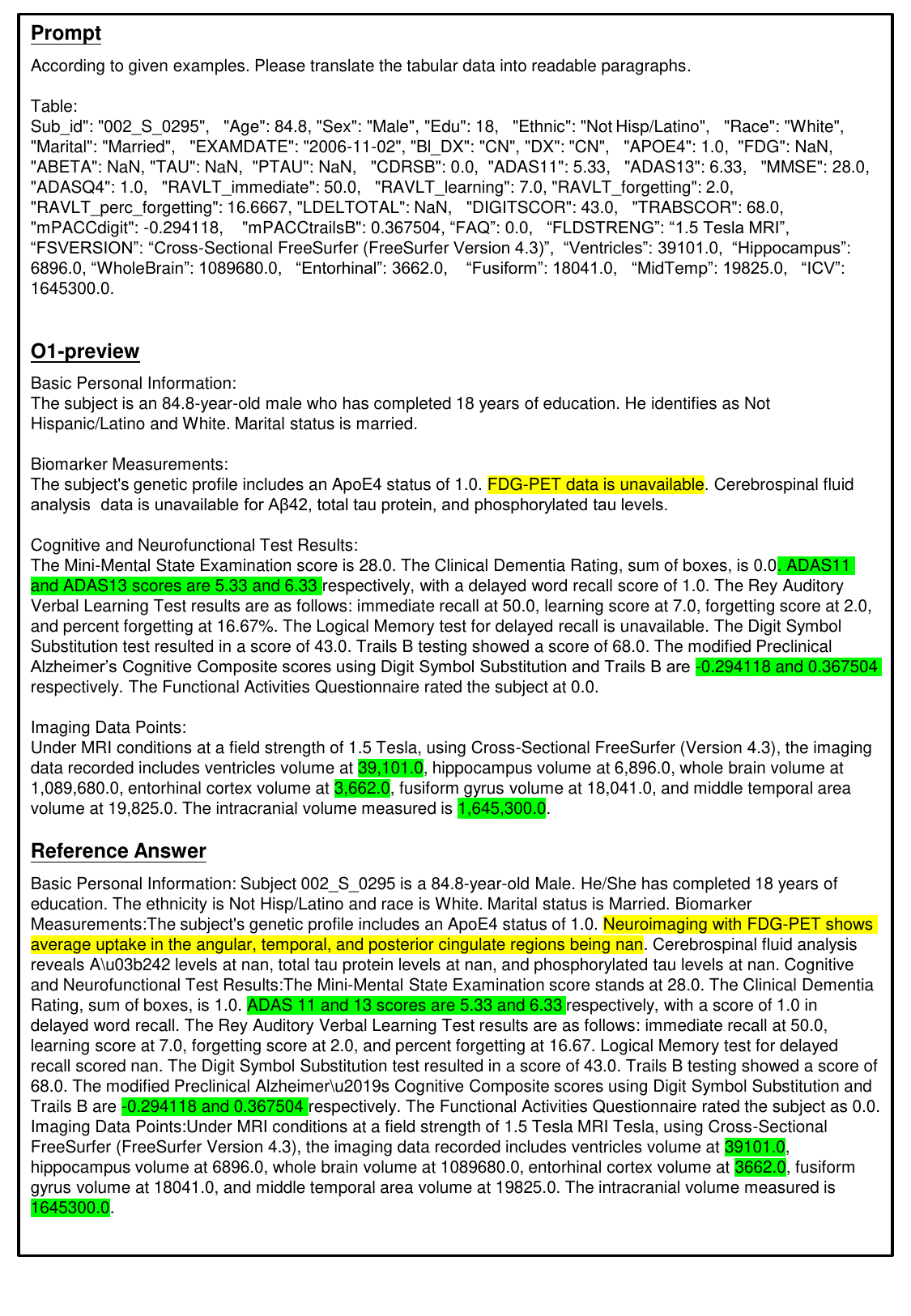}
    \caption[Table-to-Text Generation: Case 1]{\textbf{Table-to-Text: Case 1.} A demonstration case of ADNI tabular data and its corresponding generated report. Green highlights indicate that o1-preview precisely repeats the measurements, while yellow suggests that o1-preview may have its own interpretation of the tabular data during text generation.}
    \label{ADT1}
\end{figure}

\begin{figure}[H]
    \centering
\includegraphics[width=\textwidth, height=0.9\textheight, keepaspectratio]{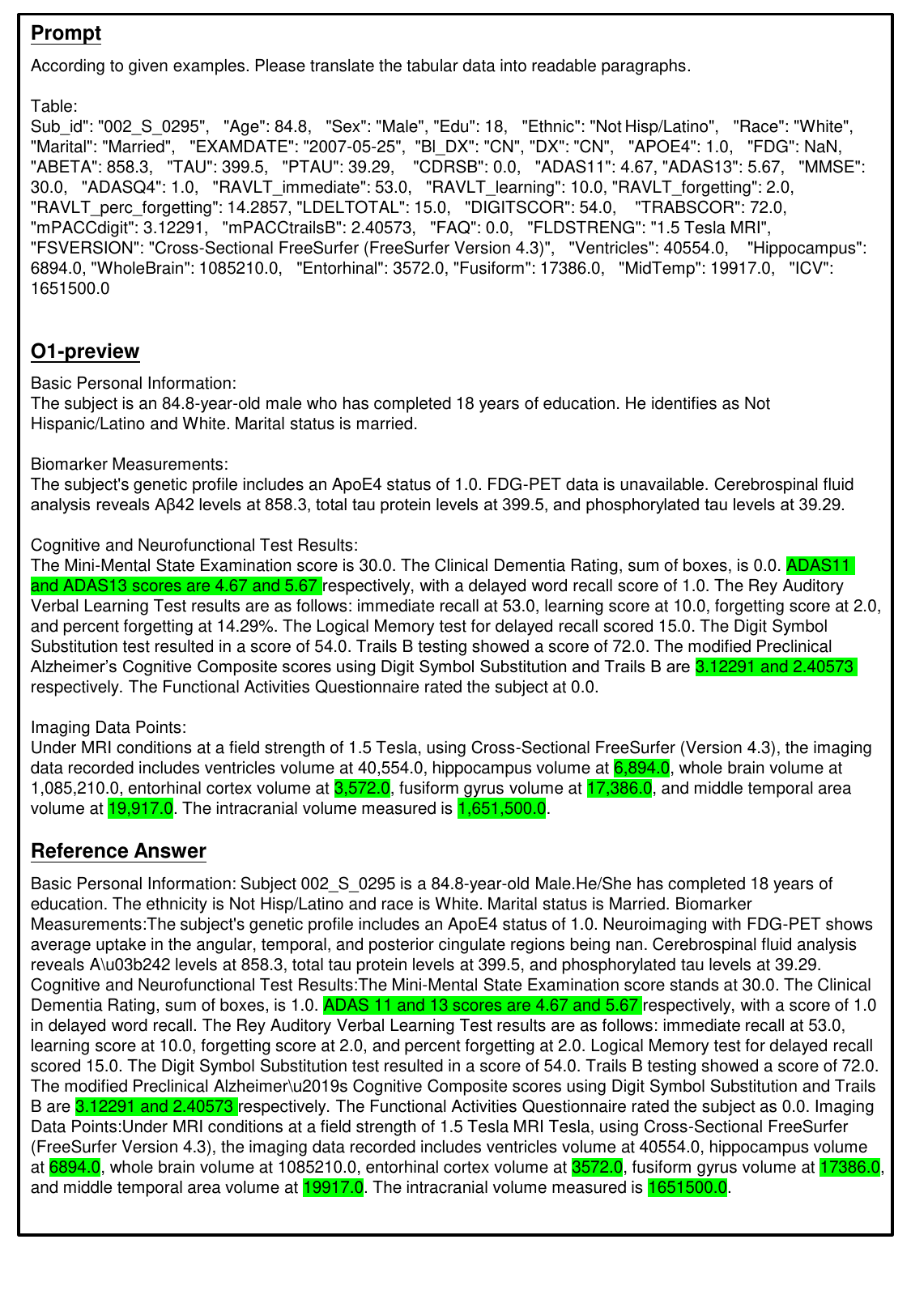}
    \caption[Table-to-Text Generation: Case 2]{\textbf{Table-to-Text: Case 2.} A demonstration case of ADNI tabular data and its corresponding generated report.}
    \label{ADT2}
\end{figure}

The o1-preview model has demonstrated strong performance on the table-to-text generation task using the ADNI dataset, particularly in transforming clinical measurements into coherent and accurate narratives. The examples of table-to-text generation for the ADNI datasets are shown in (Figure~\ref{ADT1}) and (Figure~\ref{ADT2}). Its ability to interpret and present complex clinical data—such as patient demographics, cognitive scores, and imaging biomarkers—has proven both reliable and precise, offering valuable insights for medical research and reporting. The model’s output not only captures the nuances of the data but also maintains the integrity of the clinical information, making it a promising tool for enhancing data-driven storytelling in healthcare. Based on the experimental results, the model provides well-organized and accurate analyses, demonstrating exceptional capabilities in summarizing and interpreting clinical data.

During our in-depth testing of o1-preview’s table-to-text abilities, we observed an intriguing phenomenon. In cases where certain clinical measurements, such as a patient's data, were marked as "NaN" (Not a Number), o1-preview didn’t simply output this label. Instead, it intelligently inferred that the data was missing or unavailable, offering a more meaningful interpretation. This highlights the model’s ability to not only reflect the raw data but also apply a deeper understanding of its context. Such features suggest o1-preview could be instrumental in automating comprehensive and insightful reports, further elevating the potential of AI in medical documentation and analysis.

\subsection{High School Level Math Competition}
To effectively assess the logical reasoning capability of the o1-preview, we conducted 10 test cases, 5 algebra problems and 5 counting and probability problems, spanning difficulty levels from 1 to 5. The o1-preview demonstrated outstanding performance, achieving 100\% accuracy across all 10 test cases. This consistent performance on problems of varying difficulty demonstrates its ability to handle both simple and more advanced mathematics problems. Beyond just providing correct final answers, the o1-preview’s solutions have also demonstrated detailed, step-by-step explanations which include the key steps found in the reference solutions. This level of detailed solutions suggest that the model is approaching a human-like capability in logical reasoning within the domain of mathematics.

\begin{figure}[H]
    \centering
    \includegraphics[width = \textwidth]{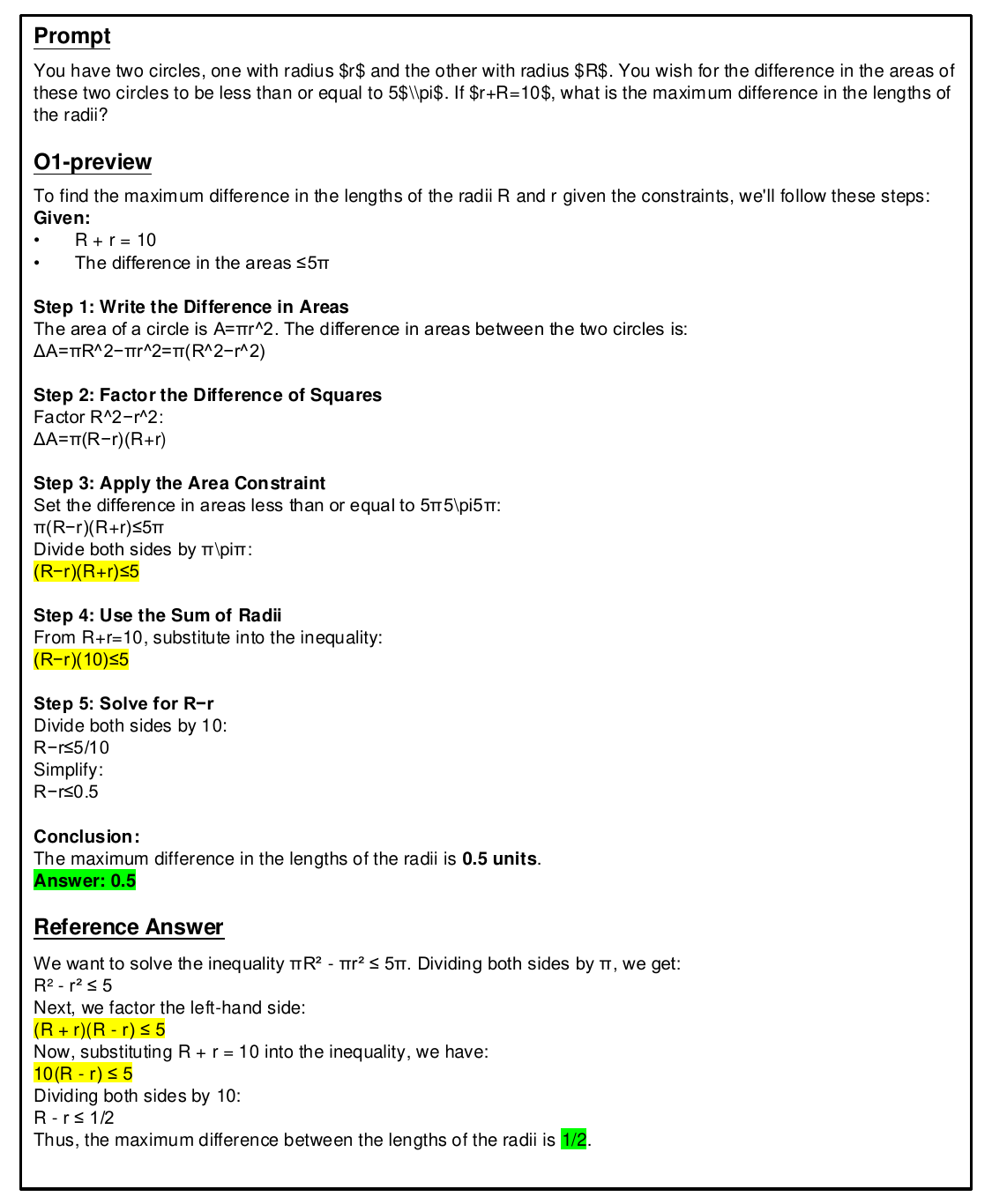}
    \caption[High School Level Math Competition: Case 1]{\textbf{High School Level Math Competition: Case 1: Algebra.} This level 5 algebra problem involves determining the maximum difference between the radii of two circles given an area constraint. The yellow highlights indicate shared steps in both the the o1-preview output and the reference answer, while the green highlights mark the shared final answers.}
    \label{fig:high_school_math_1}
\end{figure}

In case 1 (Figure~\ref{fig:high_school_math_1}), a level 5 algebra problem was presented, asking the model to determine the maximum difference between the radii of two circles given a constraint on the difference in their areas. The problem involved both geometric and algebraic reasoning, as the model needed to consider the difference of squares in the area equation and apply the given constraints to determine the maximum difference between the length of the radii. The o1-preview accurately identified the essential steps of utilizing the area difference formula and solving the inequality, eventually reaching the final answer of 0.5. The solution was detailed and the key steps aligned with the reference answer, highlighting the model’s capability to handle complex geometric problems with logical steps. 

\begin{figure}[H]
    \centering
    \includegraphics[width = \textwidth]{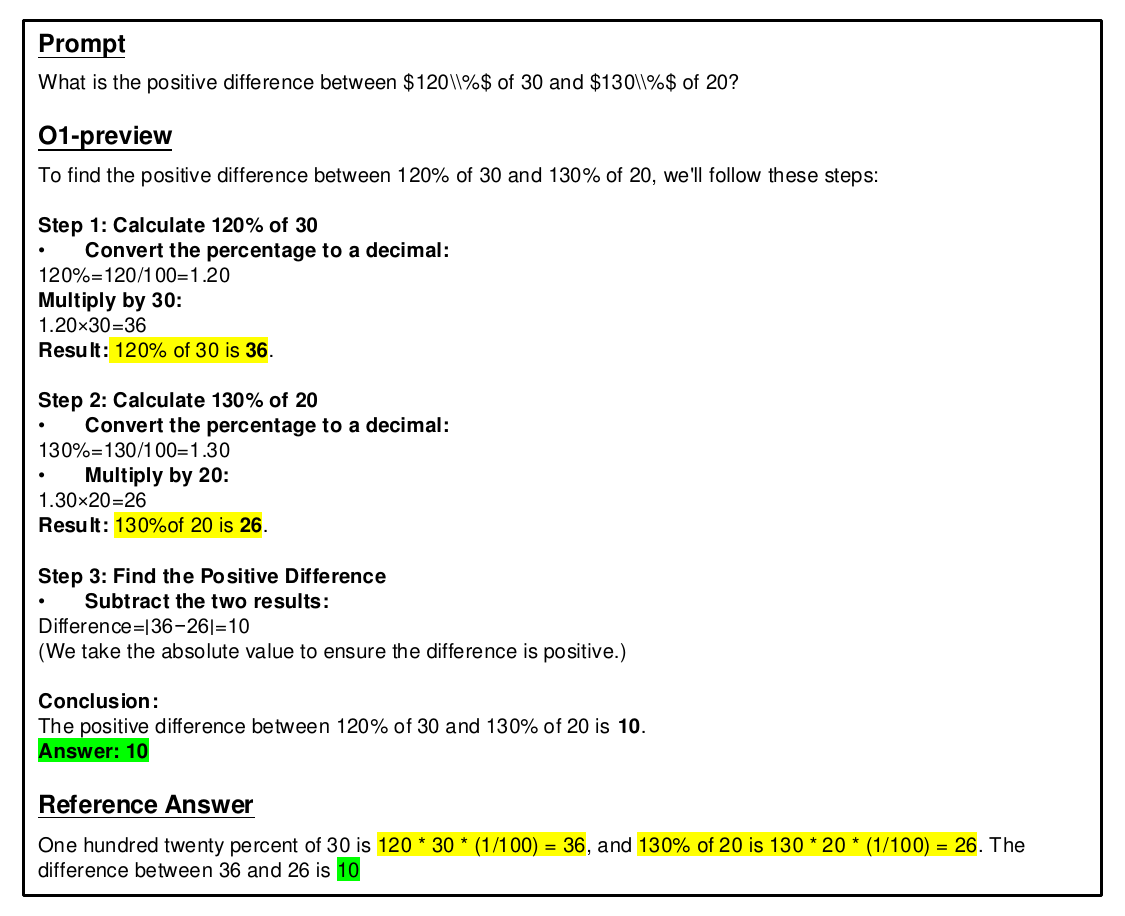}
    \caption[High School Level Math Competition: Case 2]{\textbf{High School Level Math Competition: Case 2: Algebra.} This level 1 algebra problem involves calculating the positive difference between two percentage values. The yellow highlights mark the shared intermediate steps shared in both the o1-preview's output and the reference answer, while the green highlights emphasize the correct final answers. The o1-preview efficiently completed this problem in 4 seconds. }
    \label{fig:high_school_math_2}
\end{figure}

In case 2 (Figure~\ref{fig:high_school_math_2}), the problem involved calculating the positive difference between 120\% of 30 and 130\% of 20. Although this was only a level 1 algebra problem involving basic percentage calculations, it could still display the o1-preview's logical approach if it could clearly present the essential steps. In the o1-preview's solution, it converted percentages into decimals, multiplied them by the appropriate values, and computed the absolute difference between the results. The o1-preview calculated the correct answer of 10, with the key steps shared with the reference answer. 

\newpage

\begin{figure}[H]
    \centering
    \includegraphics[width = .9\textwidth]{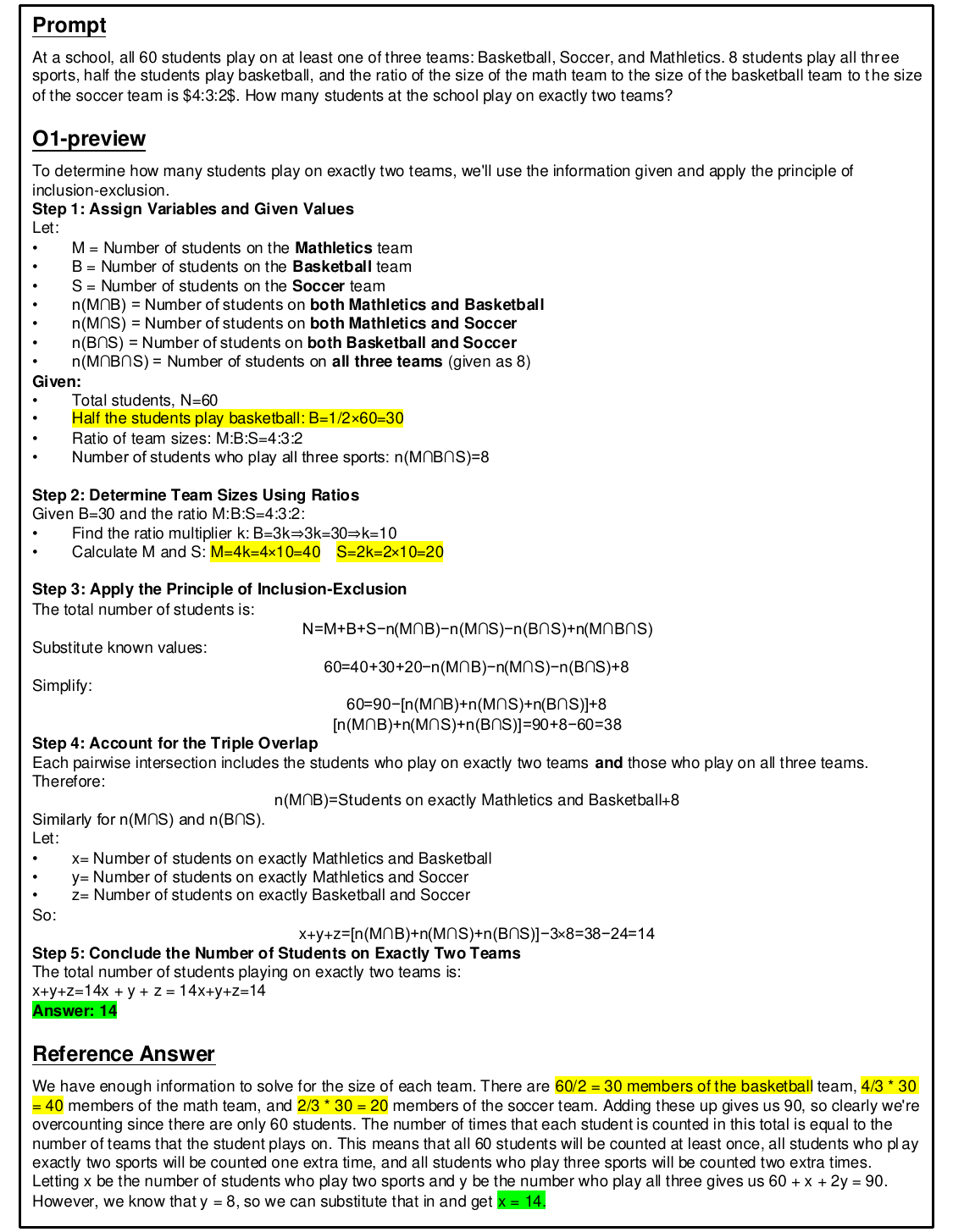}
    \caption[High School Level Math Competition: Case 3]{\textbf{High School Level Math Competition: Case 3: Counting and Probability.} This level 5 counting and probability problem involves determining how many students play on exactly two sports teams. The yellow highlights mark shared steps in both o1-preview's output and the reference answer, showing the team size calculations. The green highlights indicate the shared final answer. The o1-preview completed this high level problem in an extended time of 52 seconds.}
    \label{fig:high_school_math_3}
\end{figure}

Case 3 (Figure~\ref{fig:high_school_math_3}) presented a combinatorial problem involving the principle of counting and probability to determine how many students at a school played on exactly two sports teams. The o1-preview adeptly handled this problem by first defining variables for each team and calculating the team sizes using the given ratios. The o1-preview then applied the inclusion-exclusion principle to account for the overlaps between teams, particularly focusing on the number of students playing on exactly two teams. The final solution of the o1-preview matched the reference answer of 14 and the o1-preview's solution demonstrated its capability in handling the counting problems, such as overcounting and managing complex combinatorial reasoning tasks.

\begin{figure}[H]
    \centering
    \includegraphics[width = .9\textwidth]{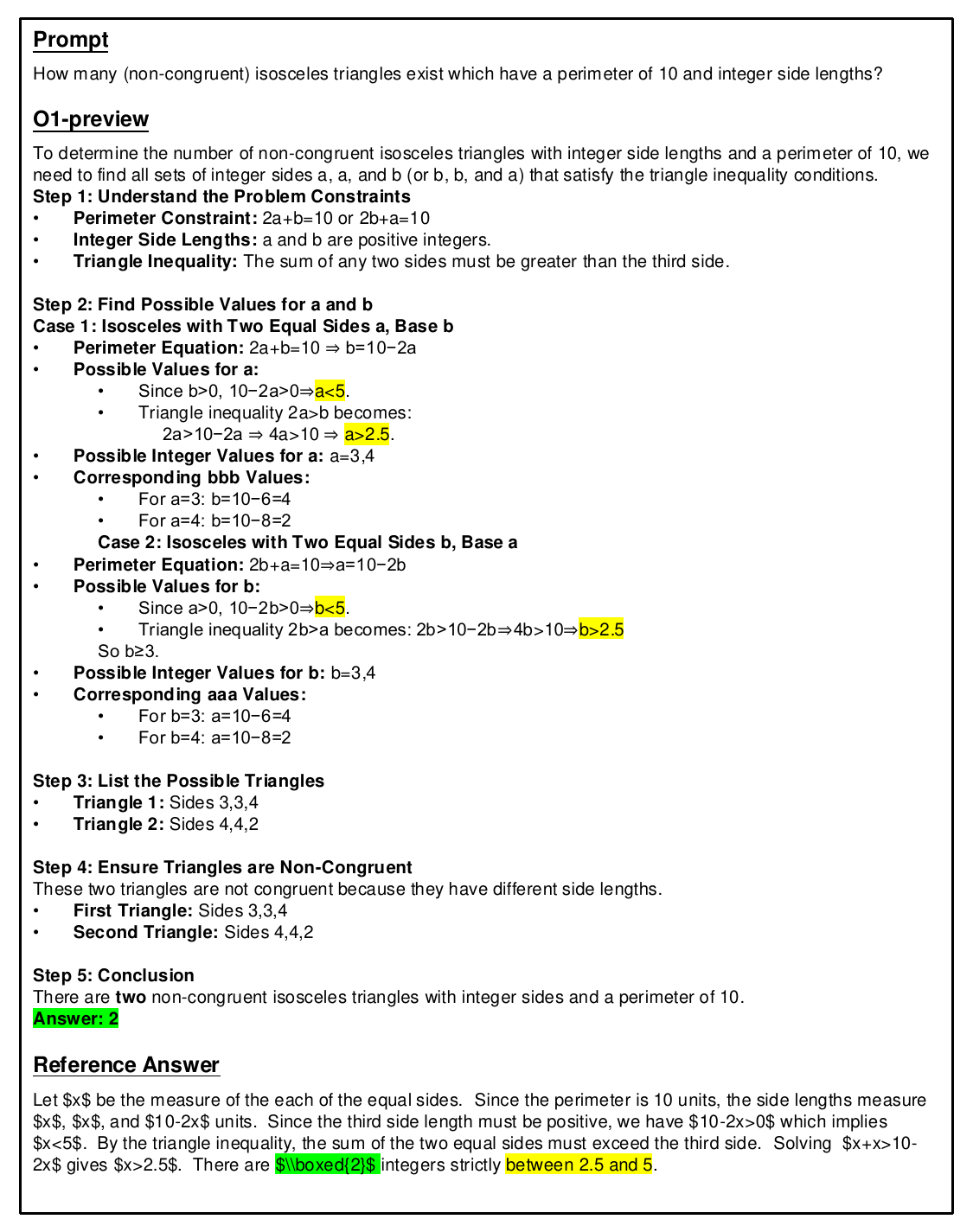}
    \caption[High School Level Math Competition: Case 4]{\textbf{High School Level Math Competition: Case 4: Counting and Probability.} This level 3 counting and probability problem involves determining the number of non-congruent isosceles triangles with integer side lengths and a perimeter of 10. The yellow highlights indicate key steps shared in both the o1-preview output and the reference answer, while the green highlights mark the shared final answer.}
    \label{fig:high_school_math_4}
\end{figure}

In case 4 (Figure~\ref{fig:high_school_math_4}), the o1-preview was asked to solve a algebraic problem that involved determining how many non-congruent isosceles triangles exist with integer side lengths and a perimeter of 10. The problem required the use of the triangle inequality theorem to validate potential side lengths while ensuring that no two triangles were congruent. The o1-preview systematically considered all possible combinations of side lengths, verified each combination using the triangle inequality, and correctly identified that two such triangles satisfied the conditions. The key steps in o1-preview's solution mirrored those in the reference answer, with the green-highlighted final answer and yellow-highlighted steps indicating the shared reasoning between the model's solution and the reference. This case further illustrates the o1-preview's strength in handling algebraic and combinatorial problems, showcasing its ability to systematically integrate algebraic and geometric reasoning.

Another notable observation is the speed at which the o1-preview generated solutions. For most cases, the model produced correct answers in just a few seconds, reflecting its efficiency in processing and reasoning. This instant response time is promising for real-time applications in mathematical problem-solving. However, for some of the more challenging level 5 problems, the o1-preview took up to 50 seconds to generate solutions. This prolonged thinking time may indicate that the o1-preview is engaging in more intricate reasoning, similar to how humans approach difficult problems. While this observation suggests the model may exhibit human-like thought processes under higher difficulty, further investigation is necessary to draw definitive conclusions about its human-like thinking process in these scenarios.

\newpage

\newpage

\newpage

\newpage


\subsection{College-level Math Problems}\label{subsec:college math discussion}

To evaluate o1-preview's ability to solve college-level math problems, we presented the model with 12 manually chosen problems. These problems cover topics including discrete math, Calculus and proofs of advanced math theorems. The answers generated by the o1-preview model were manually evaluated by the authors and categorized as correct, questionable, or incorrect. Due to complexity of the problems, we do not provide reference answers in our Figures. The test results are summarized in Table \ref{tab:college math}. 

\begin{table}[H]
    \centering
    \caption{Number of correct, questionable and incorrect answers generated by the o1-preview model for each type of math problem. The correctness of the answers was manually determined by the authors.}
    \begin{tabular}{c c c c c c}
        \hline
        & Basic discrete & Advanced discrete & Calculus & Proofs & Total\\
        \hline
        correct answers       & 2 & 0 & 1 & 4 & 7\\
        questionable answers  & 0 & 0 & 1 & 0 & 1\\
        incorrect answers     & 0 & 3 & 0 & 1 & 4\\
        \hline
        Total                   & 2 & 3 & 2 & 5 & 12\\
        \hline
    \end{tabular}

    \label{tab:college math}
\end{table}

Now, we provide a detailed analysis of each problem and the answer generated by o1-preview. In particular, we provide in-depth analysis for each of the questionable or incorrect answers.


Question 1 (Figure \ref{fig:Q1_o1}) and 2 (Figure \ref{fig:Q2_o1}) are basic discrete math problems from the field of Combinatorics. 
Both are presented as counting problems involving variables, asking for explicit formulas expressed in terms of those variables. o1-preview solved these successfully, indicating that the model can understand basic discrete patterns and perform symbolic computation effectively.


Question 3 (Figure \ref{fig:Q3_o1}), 4 (Figure \ref{fig:Q4_o1}), 5 (Figure \ref{fig:Q5_o1}) are advanced discrete math problems from the field of elementary number theory. 
The model struggled the most with this set of problems, failing to solve all three.

To take a closer look into these problems, Question 3 involves proving an identity about two sequences $\{a_i\}_{i=1}^n$ and $\{b_i\}_{i=1}^n$ of positive integers provided that they satisfy certain conditions. 
During test time, after first noticing some easy identities satisfied by the sequences, the model introduced a creative step (highlighted in yellow in Figure \ref{fig:Q3_o1}), 
proposing that if a specific relation between $a_i$ and $b_i$ for each $i$ could be proved, then the problem would be easily solved. 
This suggestion is incorrect in the first place, as the suggested relation is not true in general. 
What is more concerning is the model's subsequent steps to follow through on its own suggestion.
Instead of proving that the relation holds universally for any valid sequences $\{a_i\}_{i=1}^n$ and $\{b_i\}_{i=1}^n$, the model constructed a specific pair of sequences that satisfies the relation (also highlighted in yellow). 
For this particular pair, the required identity trivially holds. 
The model then erroneously generalized from this specific case, concluding that the relation must hold in all cases (highlighted in red) and, therefore, claiming to have proven the identity.
This approach clearly violates logical principles, as a single example cannot be used to prove a general statement. This error, known as ``proof by example'', which is a type of inappropriate generalization, indicates a gap in the model's reasoning abilities: while the model may generate creative insights, it lacks the rigor required to validate general claims through proper logical reasoning and proof construction.
The experiment was repeated a few more times, but the o1-preview model was not able to generate a correct solution.


Question 4 asks for a proof that there always exists a particular ordering of the first $n$ integers so that when concatenated into a single integer, the resulting number is divisible by 7. 
One example ($n=2$) is included in the prompt to help the model understand the question. 
Initially, the model proposed several ideas, none of which were particularly useful for solving the problem. 
It eventually resorted to using the pigeonhole principle. This approach was incorrect, as counterexamples can easily be constructed. 
The authors followed up by highlighting this mistake and providing counterexamples. 
Although the model acknowledged its mistakes, it ultimately responded that it was unable to solve the problem.


Question 5 is the famous Problem 6 from the 1988 International Mathematical Olympiad (IMO), which requires showing that whenever an integer $n$ can be written in a specific form, it must be a perfect square (i.e. $n = k^2$ for some $k$). 
Given the popularity of this problem, the authors initially expected the model to solve it with ease, but it did not.
The model did demonstrate some insight by recognizing that proving the discriminant of specific quadratic is a perfect square would solve the problem.
In the subsequent steps, however, the model claimed to observe a pattern by testing small integer values. However, misled by specific test values, the model committed another case of inappropriate generalization, assuming that the observed pattern holds in general. 
Moreover, this observed pattern is of the form ``$n = b^2$'', which overlaps with what it was asked to prove (highlighted in red in Figure \ref{fig:Q5_o1}). 
This constitutes ``circular reasoning'', where the model’s assumption essentially repeats the problem's conclusion. 
After a few more attempts, the model kept making similar mistakes and was unable to generate a correct proof. 
These logical flaws suggest that while the o1-preview model can identify patterns from small examples, it is prone to being misled and making inappropriate generalizations. Additionally, in cases requiring longer reasoning chains, the model may lose track of its initial assumptions and fall into circular reasoning.


Question 6 (Figure \ref{fig:Q6_o1}) and 7 (Figure \ref{fig:Q7_o1}) are advanced Calculus problems. o1-preview performed reasonably well on both. Specifically, Question 6 requires proving a bound for a continuously differentiable function defined on $[0, 1]$. Solving this problem requires applying the Fundamental Theorem of Calculus and the Cauchy-Schwartz inequality. The model solved it correctly and efficiently, suggesting that it possesses adequate knowledge of classical theorems and is able to apply them when the clues are obvious. 

Question 7 involves testing the convergence of an infinite series. Since such problems are classical and many examples are available on the internet, the prompt specifically asked for two different methods to ensure that the model would not rely on a memorized solution. The model provided two methods, but neither was completely satisfactory. 
The first method was simply incorrect. The model attempted to compare the given series to a specific subseries of the harmonic series $\sum_{n=1}^\infty \frac{1}{n}$, concluding the divergence of the given series based on the divergence of the harmonic series (highlighted in red in Figure \ref{fig:Q7_o1}). This reasoning is flawed, as a subseries of a divergent series can still converge.
The second method, while correct, relied on advanced techniques from graduate-level Real Analysis (highlighted in yellow).
Given the overall inadequacy of the answers, particularly the incorrect first method and the overly advanced second method, we deemed the model's responses to be questionable.


The last five problems are all advanced theorem proof problems.

Question 8 (Figure \ref{fig:Q8_o1}) and 8.5 (Figure \ref{fig:Q8.5_o1}) come from the field of Real Analysis, with Question 8.5 being a follow-up to Question 8. Both involve verifying certain topological properties of infinite dimensional spaces. The model successfully solved both problems. 

Question 9 (Figure \ref{fig:Q9_o1}), which was picked from a textbook on Functional Analysis \cite{lang}, proved to be challenging to the model. o1-preview failed to solve it particularly due to a misunderstanding of linear independence in the context of infinite dimensional spaces.

Question 10 (Figure \ref{fig:Q10_o1}) and 10.5 (Figure \ref{fig:Q10.5_o1}) are on Hilbert's Nullstellensatz, one of the most famous theorems from the field of Algebraic Geometry. 
The two questions are essentially the same, but each allows the model to assume a different intermediate result. 
This design ensures that the model pursues distinct approaches, rather than relying on a memorized solution. The model performed well on both questions, providing two different proofs that align with their respective intermediate results.


To close this subsection, we provide a summary of the strengths and limitations of o1-preview on college-level mathematical reasoning tasks.

Regarding the strengths, we have empirically observed that: \begin{itemize}
    \item Overall, o1-preview possesses comprehensive knowledge of college-level mathematical terminologies and classical theorems. Even when problems involve advanced concepts like the function space $C[0,1]$ or topological interior and closure, the model had no difficulty understanding the problem statements. Moreover, when the problem can be solved by directly applying one or two theorems, no matter how advanced they are, the model can effectively identify and apply the appropriate theorems. This in particular applies to Question 6, Question 7 (second method), Question 8, and Question 10 and 10.5.
    \item o1-preview shows a spark of creative reasoning ability, which involves making insightful observations, and sometimes constructing small examples to aid observation-making, as in Question 3 and Question 5. However, this ability seems to be in its early stages and is incomplete.
    \item Some of the mistakes made by the o1-preview model are remarkably human-like. Unlike earlier models, such as GPT-4o, whose incorrect responses were often vague, repetitive, and filled with irrelevant details, the errors made by the new model are more relatable and understandable. In fact, the types of mistakes resemble those a college student might make, particularly in Question 4, Question 7 (first method), and Question 9. For comparison purposes, the response to Question 7 generated by GPT-4o is illustrated in Figure \ref{fig:Q7_4o}. Note the repetition as highlighted in yellow, and two occurrences vague reasoning as highlighted in red.
\end{itemize}

For the limitations, our empirical observations indicate that: \begin{itemize}
    \item While o1-preview demonstrates the ability to make insightful observations and identify patterns from examples it constructs, it can also be misled by its own observations, leading to inappropriate generalizations—mistakenly assuming that coincidences apply more broadly. This applies to Question 3 and Question 5. In general, there is no evidence to support that the model possesses a genuine understanding of logical principles. When faced with extended reasoning chains, where it may even fall into circular reasoning.
    \item It seems the o1-preview model lacks an awareness of problem difficulty and often prioritizes efficiency over appropriateness. For instance, in Question 7, it applied overly advanced techniques to solve a moderately challenging problem.
    \item Overall, the model tends to perform better on problems with a high level of abstraction that require advanced theorems but involve shorter reasoning chains, such as the last five problems, rather than on problems with straightforward statements that require less advanced knowledge but involve much longer reasoning processes, such as the three advanced discrete math problems.
\end{itemize}


\fig{width=\textwidth}{Q1_o1}{College-Level Math: Case 1}{A demonstration case of solving a basic discrete math problem. The generated answer is manually evaluated as correct.}
\newpage

\fig{width=\textwidth}{Q5_o1}{College-Level Math: Case 5}{A demonstration case of solving a advanced discrete math problem. The generated answer is manually evaluated as incorrect.}
\newpage

\fig{width=\textwidth}{Q7_o1}{College-Level Math: Case 7}{A demonstration case of solving a Calculus problem. The generated answer is manually evaluated as questionable.}
\newpage


\fig{width=\textwidth}{Q7_4o}{College-Level Math: Case 7 (GPT-4o)}{Answer to the same Calculus problem as in Figure \ref{fig:Q7_o1} generated by GPT-4o.}


\subsection{Electronic Health Record Diagnosis}
\begin{figure}[H]
    \centering
    \includegraphics[width = 0.85\linewidth]{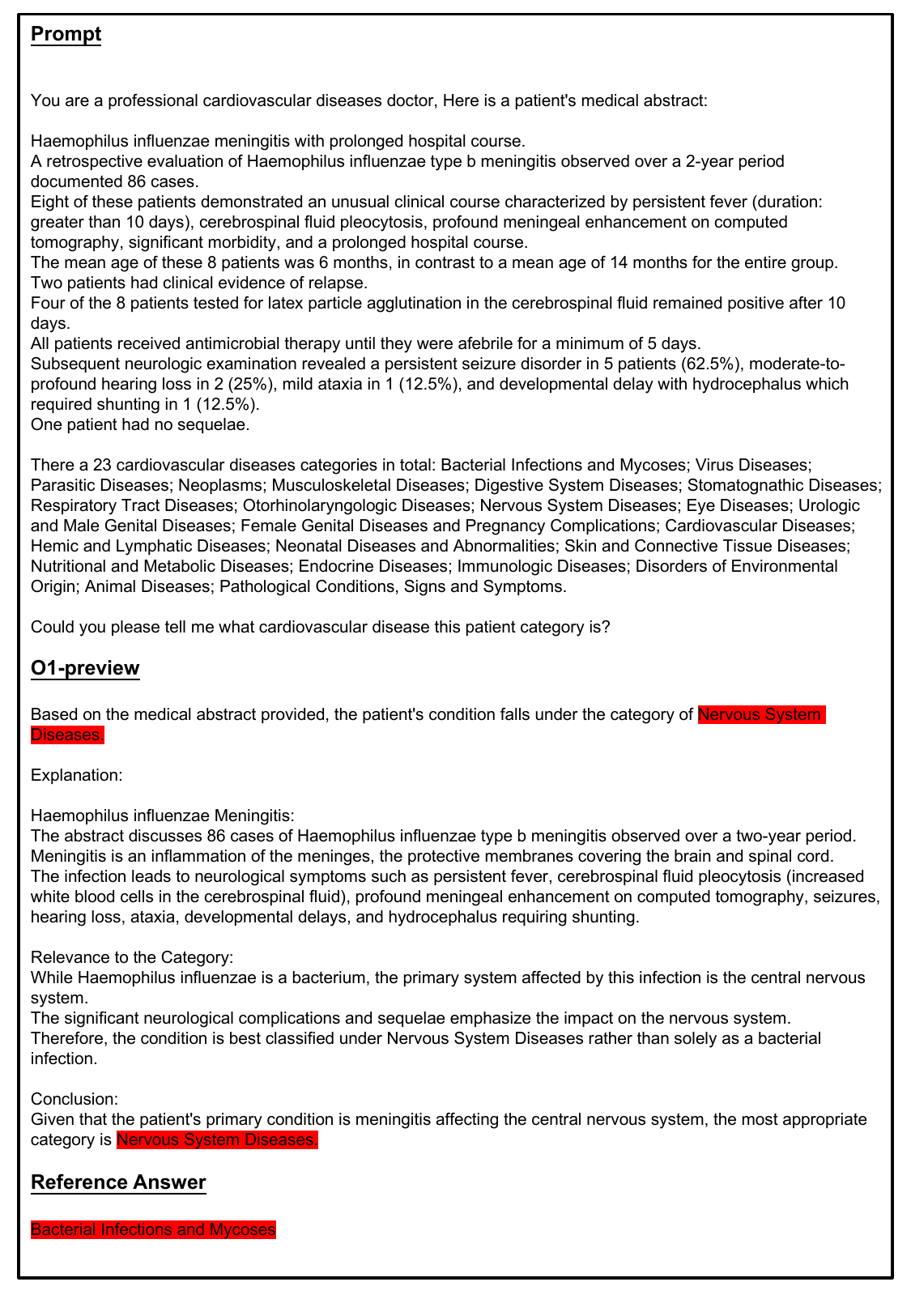}
    \caption[Electronic Health Record Diagnosis: Case 1.]
    {\textbf{Electronic Health Record Diagnosis: Case 1.} One specific examples of a demonstration cases for EHR diagnosis involves the use of the o1-preview model. Red indicates that the model's output is completely inconsistent with the reference diagnosis, while green signifies complete consistency.}
    \label{fig:EHR-1}
\end{figure}

\begin{figure}[H]
    \centering
    \includegraphics[width = 0.85\linewidth]{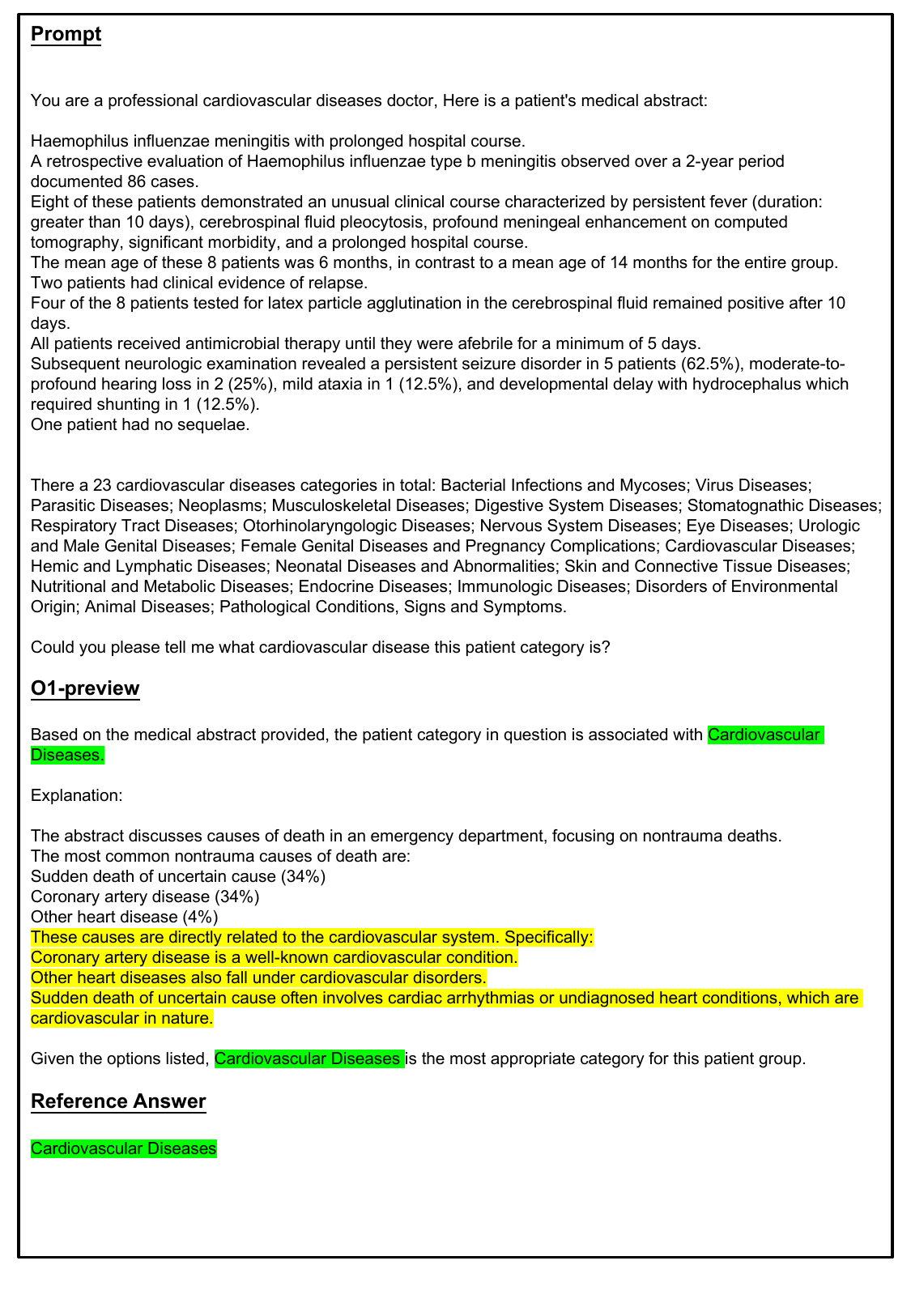}
    \caption[Electronic Health Record Diagnosis: Case 2.]
    {\textbf{Electronic Health Record Diagnosis: Case 2.} One specific examples of a demonstration cases for EHR diagnosis involves the use of the o1-preview model. Red indicates that the model's output is completely inconsistent with the reference diagnosis, while green signifies complete consistency.}
    \label{fig:EHR-2}
\end{figure}

\begin{figure}[H]
    \centering
    \includegraphics[width = 0.85\linewidth]{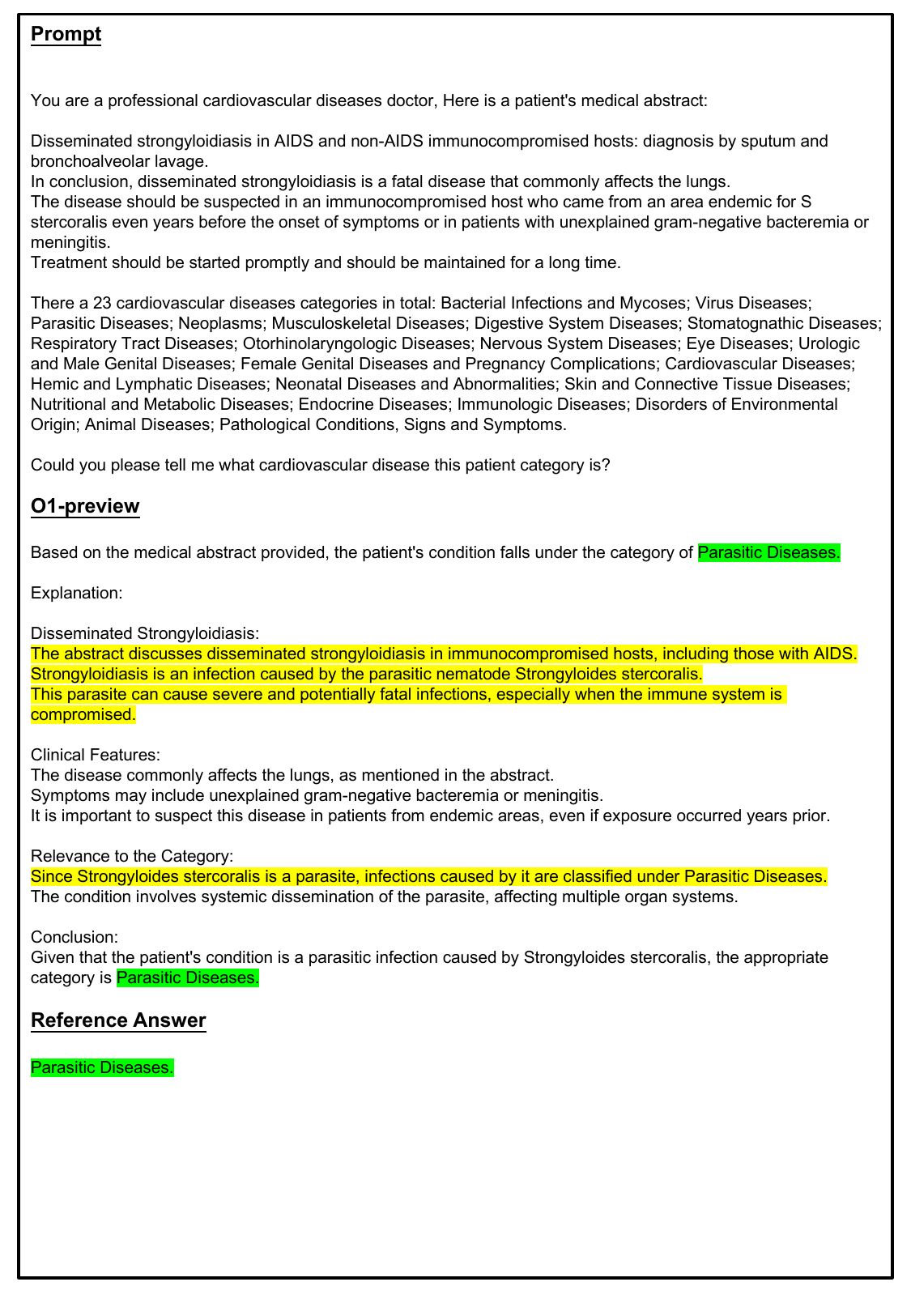}
    \caption[Electronic Health Record Diagnosis: Case 3.]
    {\textbf{Electronic Health Record Diagnosis: Case 3.} One specific examples of a demonstration cases for EHR diagnosis involves the use of the o1-preview model. Red indicates that the model's output is completely inconsistent with the reference diagnosis, while green signifies complete consistency.}
    \label{fig:EHR-4}
\end{figure}

\begin{figure}[H]
    \centering
    \includegraphics[width = 0.85\linewidth]{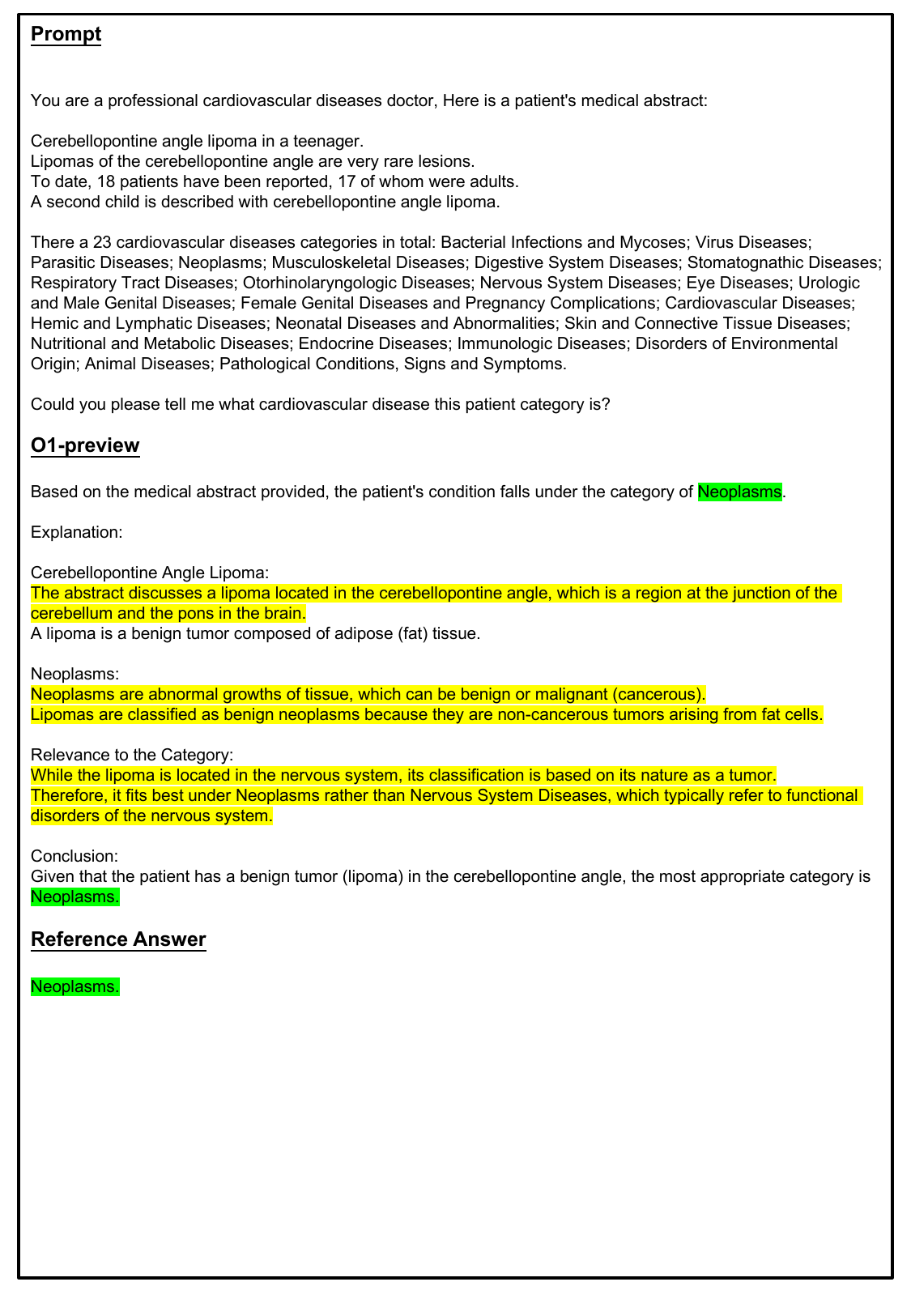}
    \caption[Electronic Health Record Diagnosis: Case 4.]
    {\textbf{Electronic Health Record Diagnosis: Case 4.} One specific examples of a demonstration cases for EHR diagnosis involves the use of the o1-preview model. Red indicates that the model's output is completely inconsistent with the reference diagnosis, while green signifies complete consistency.}
    \label{fig:EHR-5}
\end{figure}

To evaluate the diagnostic capabilities of the o1-preview model using EHRs, we tested it with 10 manually selected cases, each representing a different cardiovascular disease category. The model's responses were manually reviewed by the authors and categorized as either correct (marked in green) or incorrect (marked in red). In addition to identifying key medical patterns, the o1-preview model also provided diagnostic explanations and reasoning based on the information provided, demonstrating strong inferential capabilities.

o1-preview has demonstrated notable potential in supporting medical diagnosis through its ability to interpret EHR data and provide relevant diagnoses. While the model performs exceptionally well with shorter, more concise texts (Figure \ref{fig:EHR-2}, Figure \ref{fig:EHR-4}, Figure \ref{fig:EHR-5}), challenges persist when dealing with longer and more complex records, which may impact its accuracy (Figure \ref{fig:EHR-1}). Nevertheless, the model's ability to offer reasoning and diagnostic justifications indicates that with future fine-tuning, particularly through reinforcement learning tailored to EHR data, its diagnostic capabilities could be significantly enhanced. Additional examples can be found in (Appendix \ref{appendix:EHR}).

Overall, the o1-preview's capabilities to understand and analyze EHRs show great potential for the development of AI-driven healthcare tools, though further refinement is needed to improve its effectiveness in real-world medical applications.

\newpage
\newpage

\newpage

\newpage

\newpage

\subsection{Stochastic Processes in Statistics}
The performance of the o1-preview on stochastic process problems is summarized as follows. Overall, the model exhibited a strong ability to not only understand the majority of stochastic process problems but also to provide step-by-step, detailed solutions. In many cases, the explanations were thorough, though at times overly detailed, which occasionally led to somewhat redundant or verbose answers. This tendency, while not detrimental, suggests that the model's solution clarity could benefit from a more concise presentation in certain situations.

Despite its success with most problems, the o1-preview encountered difficulties with more complex Markov chain problems, particularly in accurately determining the steady state of the Markov chain, which resulted in errors. This indicates that while the model performs well on simpler tasks, there is room for improvement in its deep understanding and handling of more intricate, advanced topics. 

\begin{figure}[H]
    \centering
    \includegraphics[width = 0.9\textwidth]{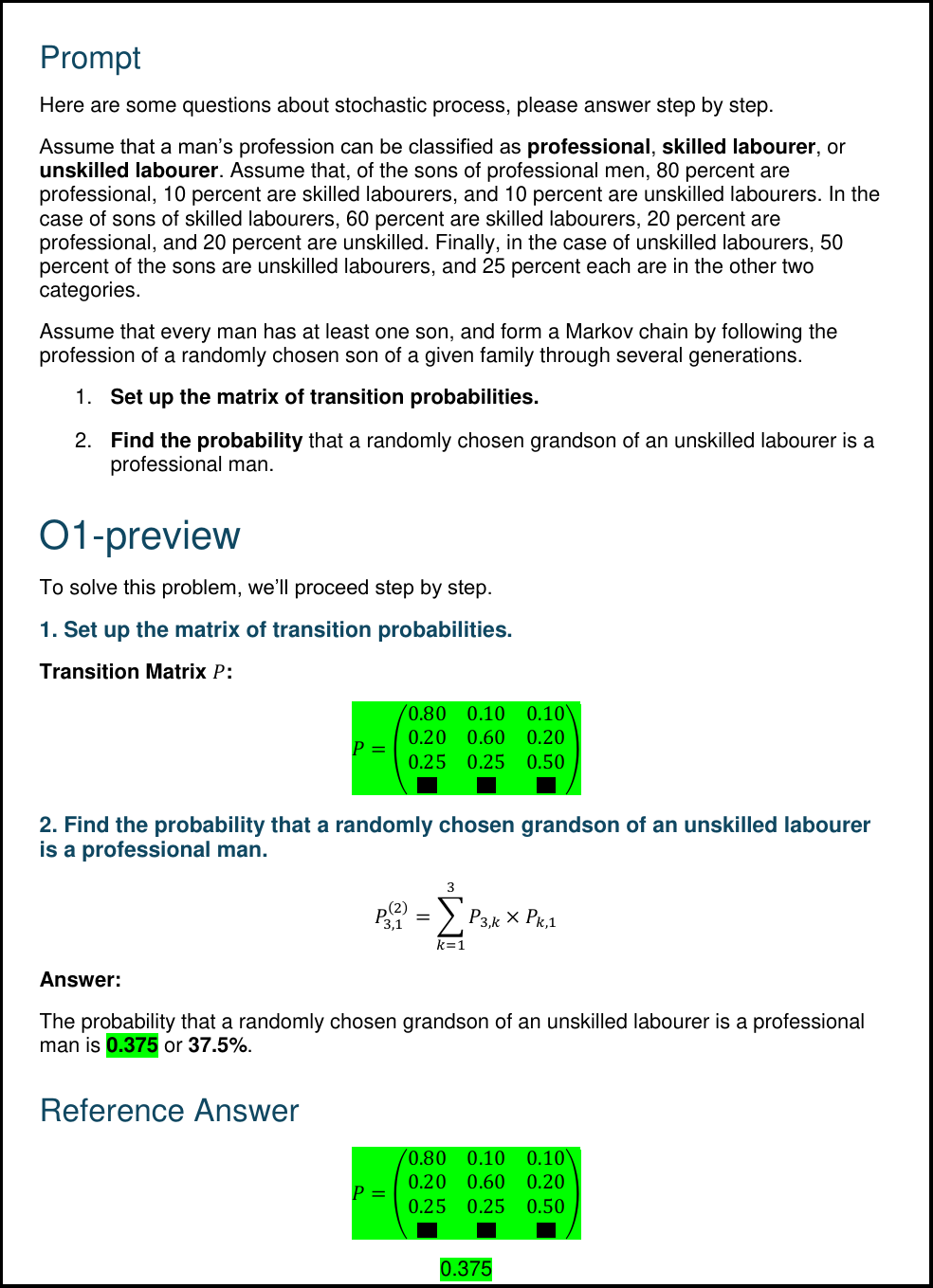}
    \caption[Stochastic Processes in Statistics: Case 2]{\textbf{Stochastic Processes in Statistics: Case 2.} A demonstration case of Stochastic Process problem solving. The o1-preview failed to solve the problem.}
    \label{fig:Stochastic_2}
\end{figure}

As shown in Appendix \ref{appendix:Stochastic}, the model achieved an 80\% accuracy rate on basic problems. However, as evidenced in Figure \ref{fig:Stochastic_2}, it struggled with more challenging cases. Additionally, the model was evaluated on some classic mathematical problems that large language models typically find challenging, such as comparing 9.11 and 9.8. In these instances, the o1-preview performed exceptionally well, as shown in Figure \ref{fig:Stochastic_6}.

\newpage

\begin{figure}[H]
    \centering
    \includegraphics[width = 0.9\linewidth]{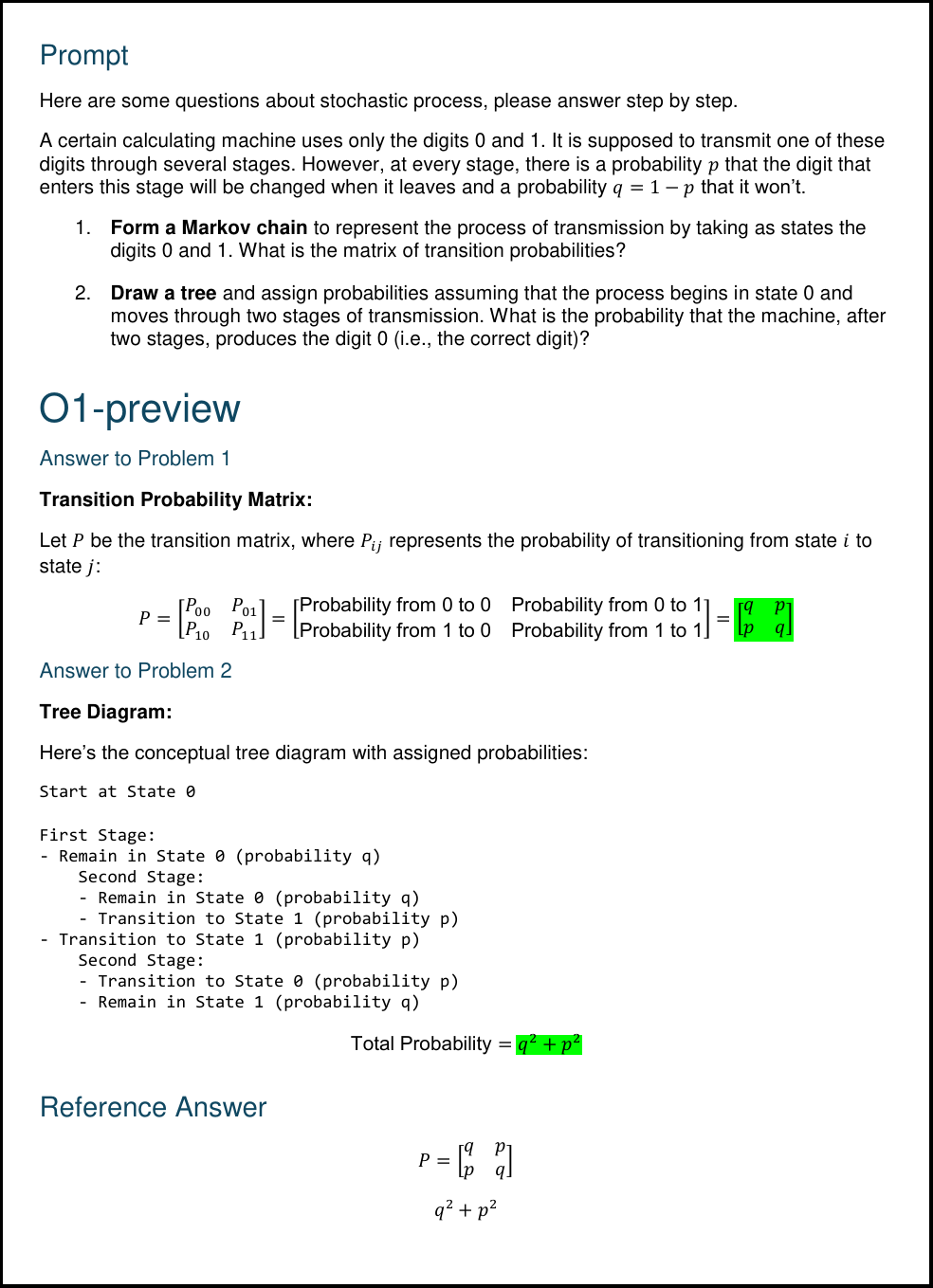}
    \caption[Stochastic Processes in Statistics: Case 1]{\textbf{Stochastic Processes in Statistics: Case 1.} A demonstration case of Stochastic Process problem solving. The o1-preview successfully solved the problem.}
    \label{fig:Stochastic_1}
\end{figure}
\newpage

\newpage

\subsection{Medical Text Anonymization}
To assess o1-preview's effectiveness in medical text anonymization, we crafted three customized prompts with varying degrees of specificity. These prompts were structured from broad to detailed, allowing us to gauge the model's performance across different levels of granularity. The first, the coarse prompt, provides minimal guidance, simply directing the model to identify private information without offering further specifics. The medium prompt introduces a more comprehensive list of privacy-related elements, while the fine prompt supplies detailed instructions, specifying the properties of each privacy feature. Table~\ref{tab: MAT_prompt} outlines this experiment's complete set of prompts.

\begin{table}[h]
\centering
\caption{Prompt Table for Medical Text Anonymization.}
\label{tab: MAT_prompt}
\begin{tabular}{p{3cm}p{4cm}p{8cm}}
\toprule
\textbf{Implicit Prompt} & \textbf{General Prompt} & \textbf{Explicit Prompt}\\
\midrule
Please anonymize the following text. Replace all the private information with the term “[redacted]”.: & What are the names? Who are the professions? What are the locations? What are ages? What are dates? What are contacts? What are IDs? What are the phone numbers? Replace the matching characters with [redacted]: & Please anonymize the following clinical note. Specifically, replace all the following information with the term “[redacted]”: redact any strings that might be a name or acronym or initials, redact any strings separated by the \verb|\| \verb|/| symbol, redact patients' names, doctors' names and the strings in front of M.D. or after Dr., redact pager names and medical staff names, redact any strings that look like something years old or age 37, redact any dates and IDs and numbers and record dates, redact locations and addresses and clinic names, redact professions and ages and contacts, redact any acronyms and initials.: \\
\bottomrule
\end{tabular}
\end{table}

\begin{figure}
    \centering
    \includegraphics[width=0.85\linewidth]{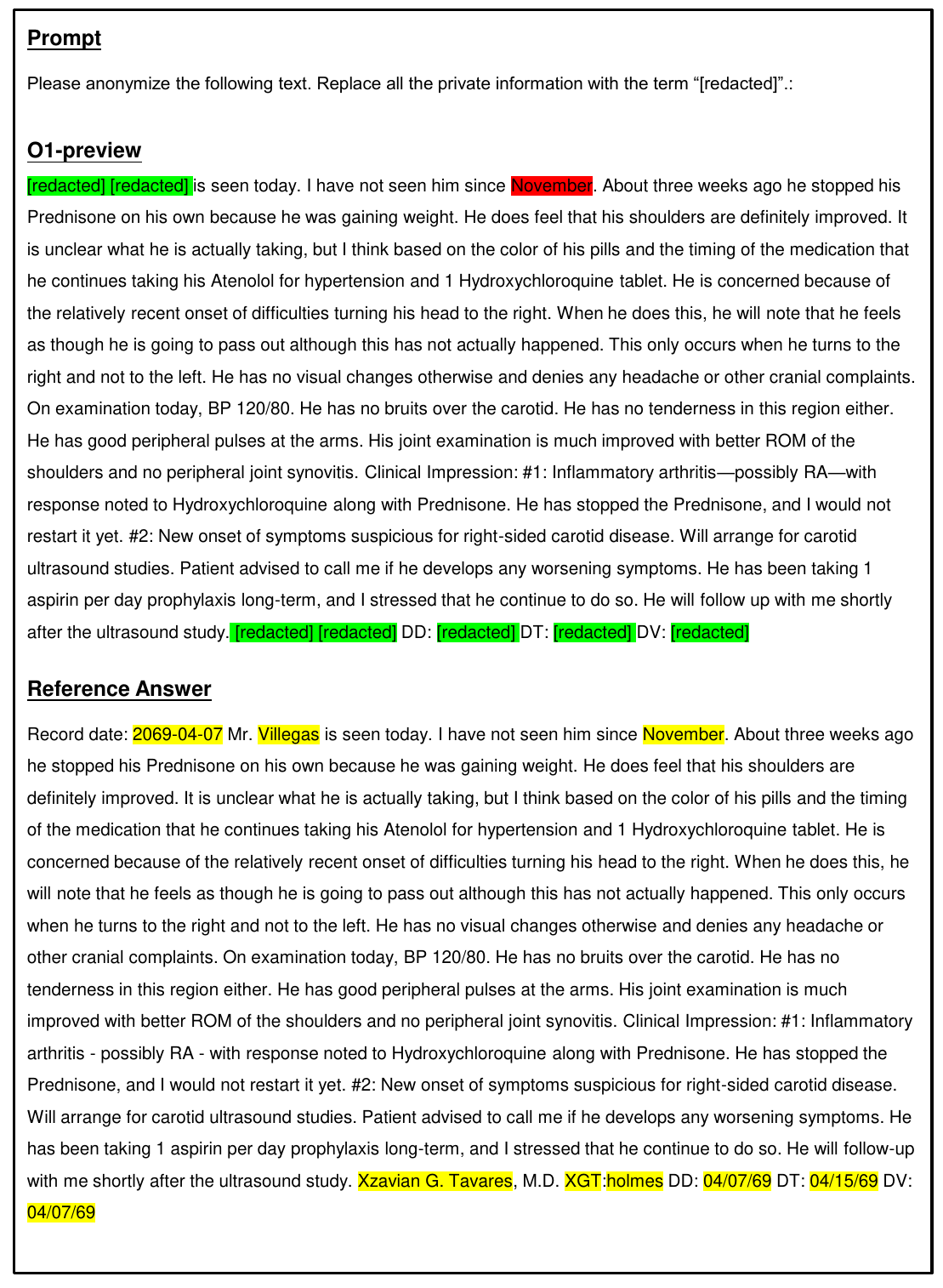}
    \caption[Medical Text Anonymization: Case 1]{\textbf{Medical Text Anonymization: Case 1.} An example showing the capability of o1-preview to anonymize medical texts with the implicit prompt. Yellow highlights the annotations made by human experts. Green marks the sensitive privacy information detected by o1-preview. Red indicates modifications beyond the human-marked annotation. }
    \label{fig:MTA-case1-1}
\end{figure}

\begin{figure}
    \centering
    \includegraphics[width=0.85\linewidth]{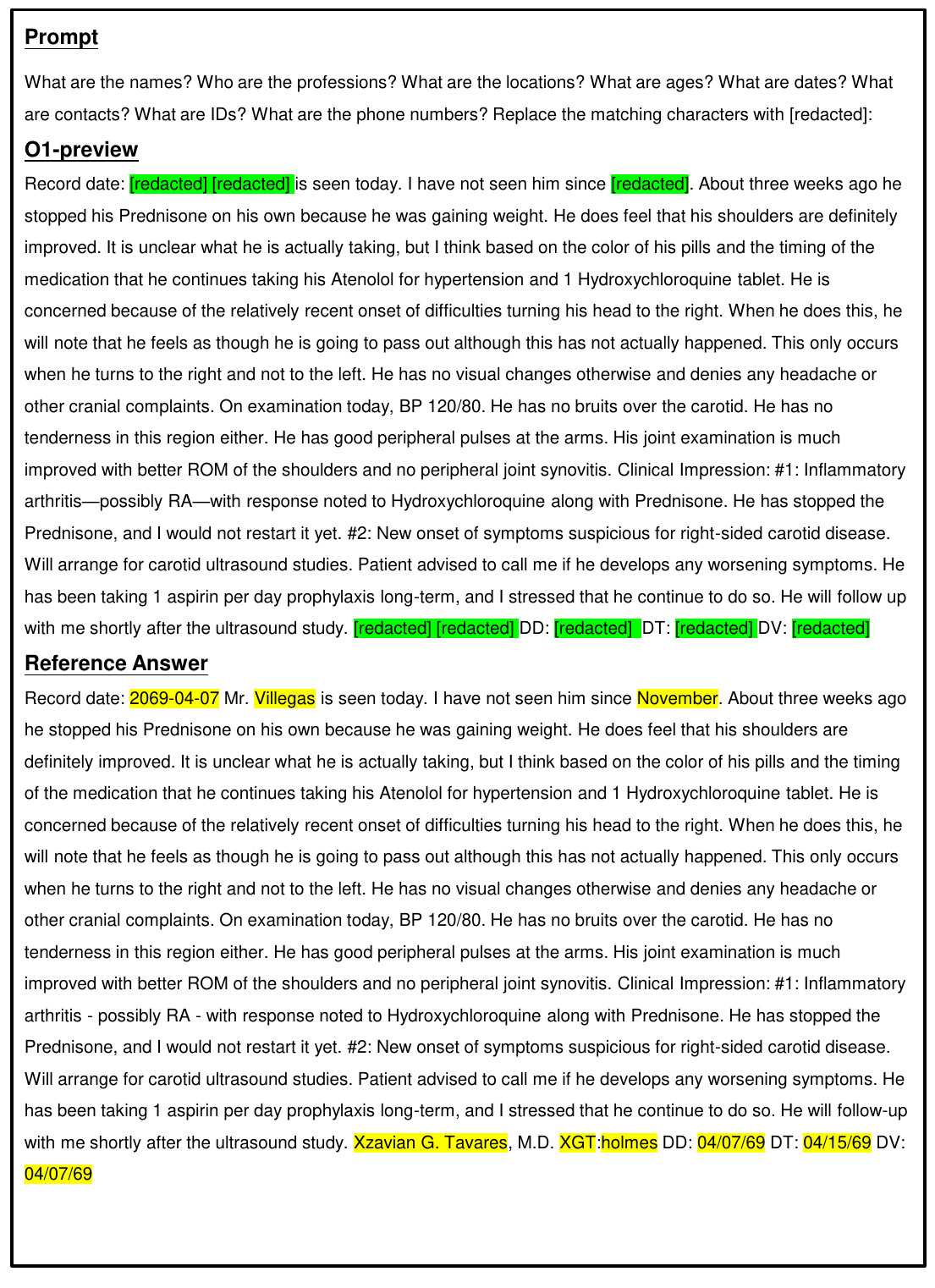}
    \caption[Medical Text Anonymization: Case 2]{\textbf{Medical Text Anonymization: Case 2.} An example showing the capability of o1-preview to anonymize medical texts with the general prompt. Yellow highlights the annotations made by human experts. Green marks the sensitive privacy information detected by o1-preview. Red indicates additional modifications beyond the human-marked annotation.}
    \label{fig:MTA-case1-2}
\end{figure}

\begin{figure}
    \centering
    \includegraphics[width=0.85\linewidth]{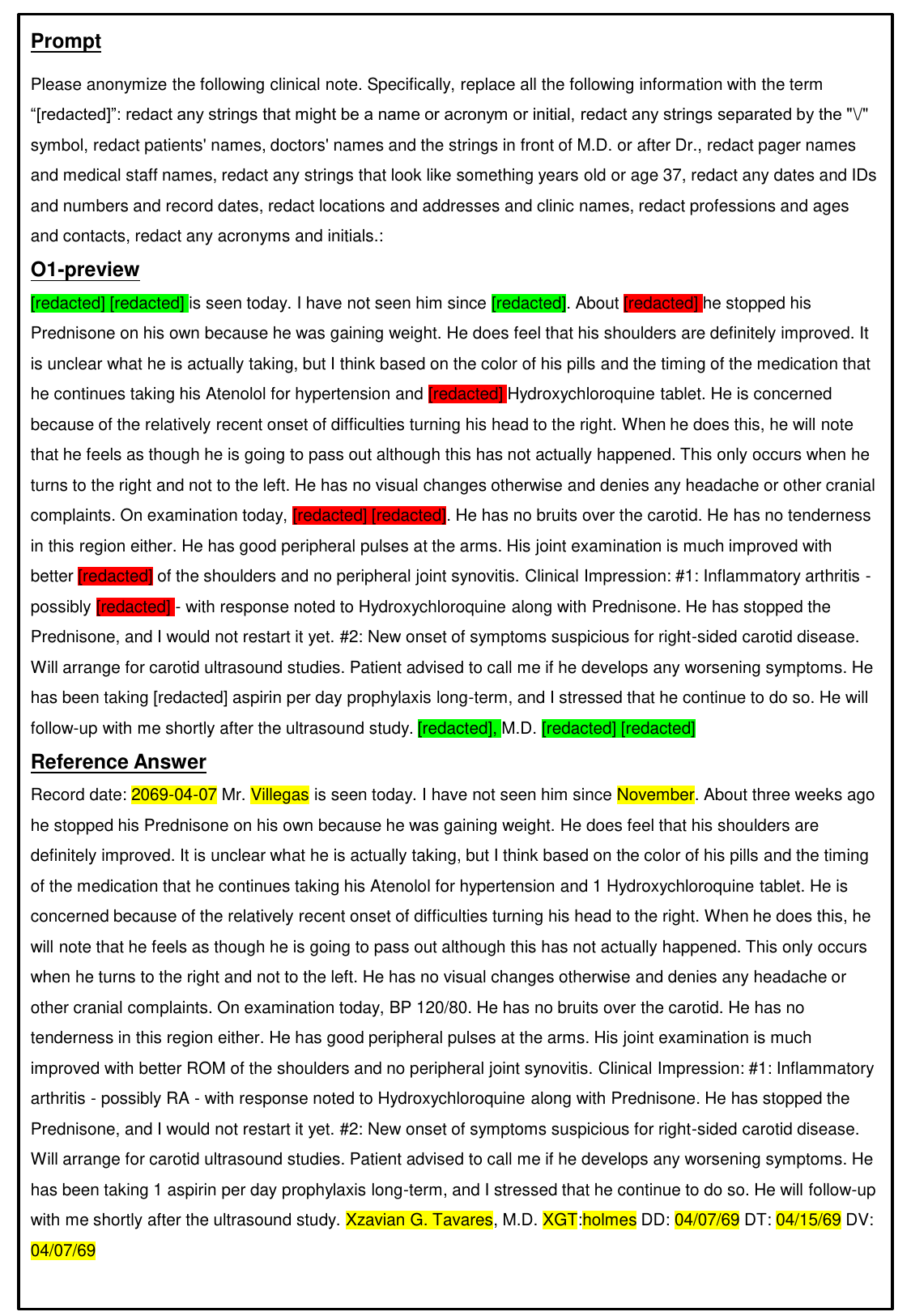}
    \caption[Medical Text Anonymization: Case 3]{\textbf{Medical Text Anonymization: Case 3.} An example showing the capability of o1-preview to anonymize medical texts with the explicit prompt. Yellow highlights the annotations made by human experts. Green marks the sensitive privacy information detected by o1-preview. Red indicates modifications beyond the human-marked annotation. }
    \label{fig:MTA-case1-3}
\end{figure}

As the results (Figures~\ref{fig:MTA-case1-1},~\ref{fig:MTA-case1-2},~\ref{fig:MTA-case1-3}) present, o1-preview is capable of detecting nearly all sensitive information, even when processing lengthy content. It can efficiently handle long texts, analyze the material from multiple perspectives, and generate results in a short amount of time. Although different words are redacted across the three prompts, each effectively removed essential privacy information, demonstrating the impressive general capabilities of o1-preview. The results have proved its potential to be a powerful automatic tool to anonymize professional medical texts.

o1-preview accurately detects personal names and timestamps, but slight variations can be observed when using different prompts. With implicit prompts, o1-preview focuses primarily on the overall structure of the content, identifying sensitive information in fixed positions while preserving abbreviations that aid in localization. For general prompts, due to the strong alignment between the annotation features and the prompts, o1-preview detects all sensitive information without any additional effort. However, with explicit prompts, the abundance of instructions causes the o1-preview to evaluate each word individually. While it successfully identifies all sensitive information, it occasionally removes more meaningful words than expected. In summary, the effectiveness of o1-preview largely depends on the appropriateness of the prompt.

\newpage

\subsection{Social Media Analysis}
\begin{figure}[H]
    \centering
    \includegraphics[width = 0.85\textwidth]{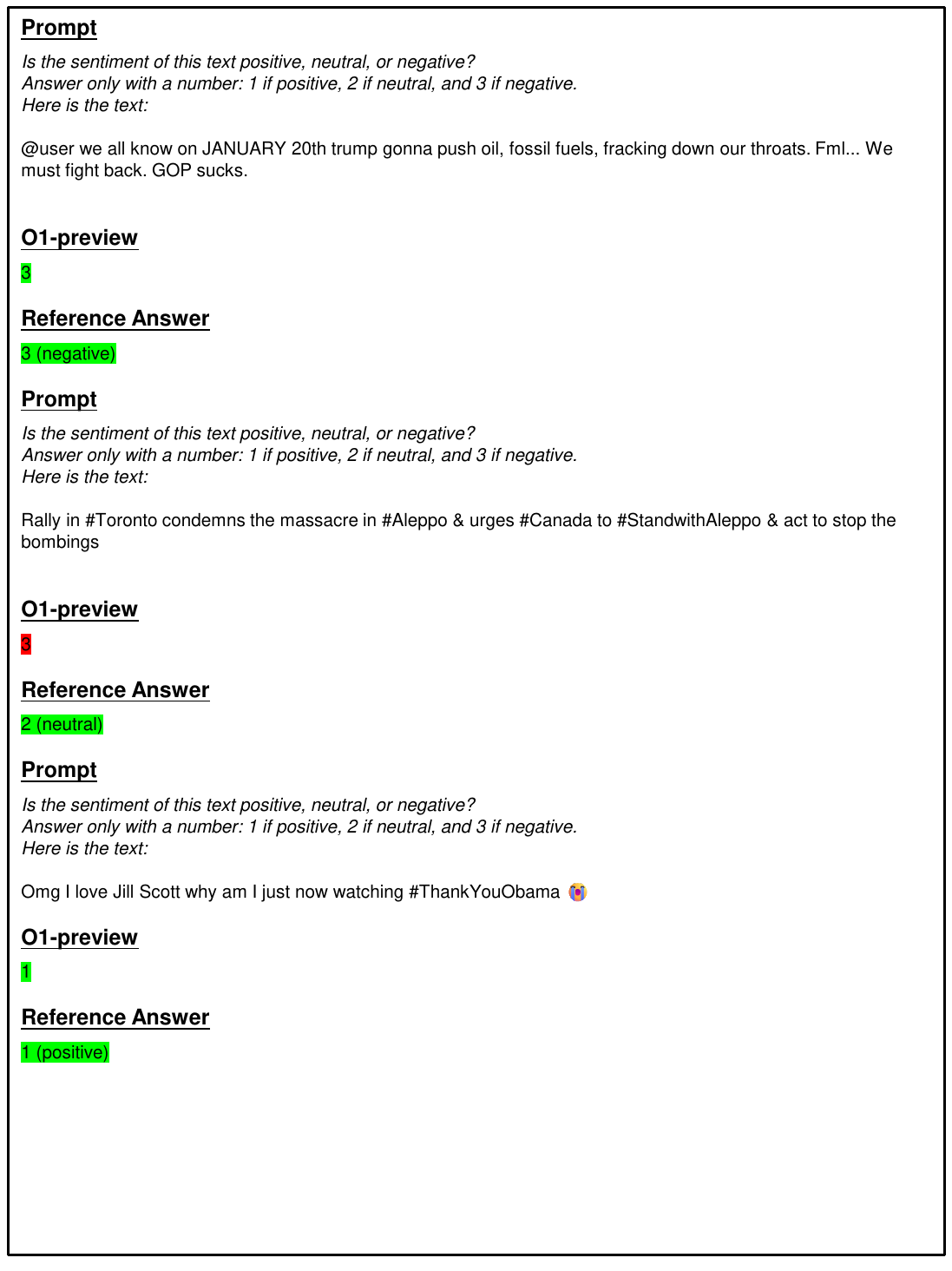}
    \caption[Social Media Analysis: Case 1-3]{\textbf{Social Media Analysis: Case 1-3.} Demonstration cases for sentiment analysis. Red indicates that the output of the o1-preview model is completely inconsistent with the reference answer, while green signifies full consistency. }
    \label{fig:sma_1-3}
\end{figure}

\begin{figure}[H]
    \centering
    \includegraphics[width = 0.85\textwidth]{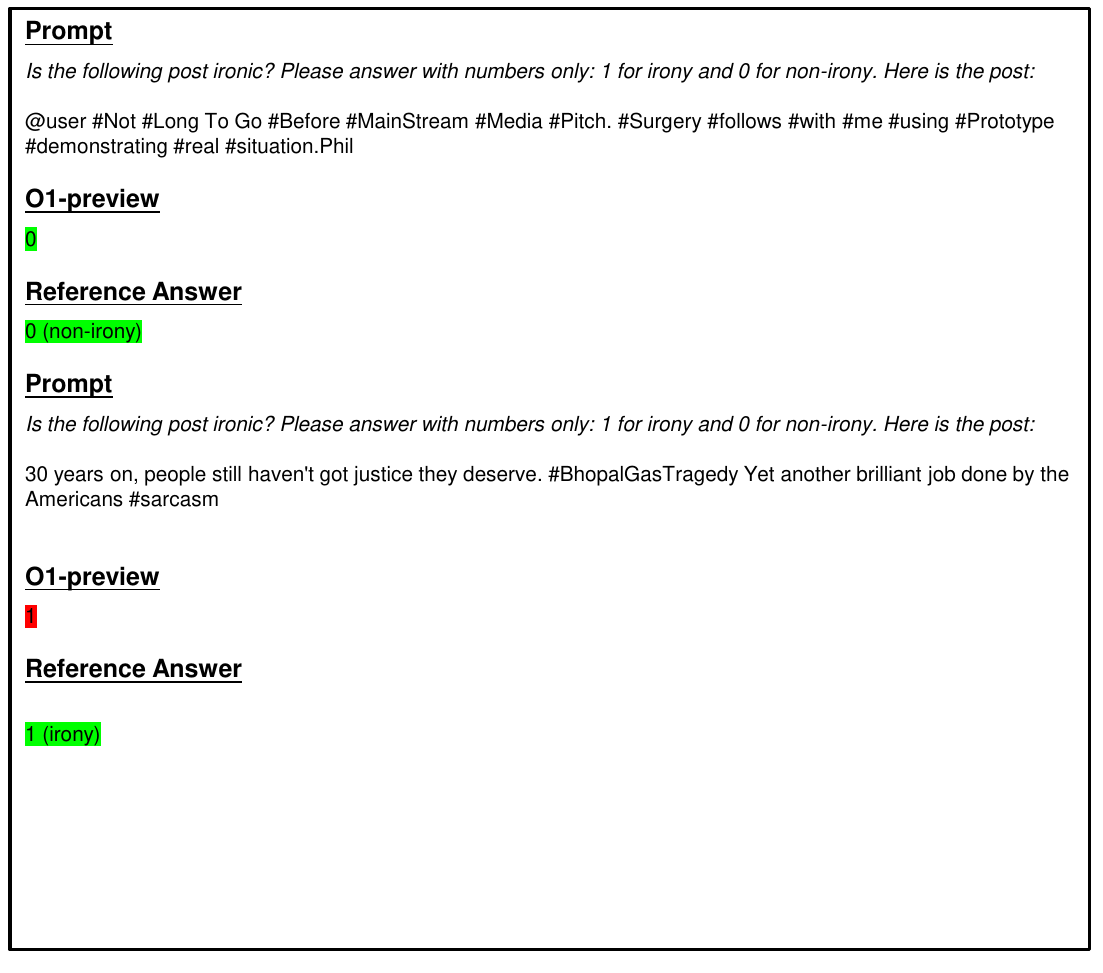}
    \caption[Social Media Analysis: Case 4-5]{\textbf{Social Media Analysis: Case 4-5.} Demonstration cases for offensive language identification. }
    \label{fig:sma_4-5}
\end{figure}

To evaluate the capabilities of o1-preview in social media analysis, we used four common tweet datasets and reported results across 11 cases. The prompt templates used throughout the testing process are presented in Table~\ref{tab:SMA_prompt}.
Among these 11 cases, three cases are sentiment analysis (Figure~\ref{fig:sma_1-3}), two cases are offensive language detection (Figure~\ref{fig:sma_4-5}), four are emotion recognition(Figure~\ref{fig:sma_6-9}), and two are irony identification (Figure~\ref{fig:sma_10-11}).
\begin{table}[h]
\centering
\caption{Prompt table for social media analysis.}
\label{tab:SMA_prompt}
\begin{tabular}{p{3.5cm}p{4.8cm}p{3.2cm}p{3.2cm}}
\toprule
\textbf{Sentiment analysis} & \textbf{Emotion detection} & \textbf{Offensiveness} &  \textbf{Irony detection} \\
\midrule
Is the sentiment of this text positive, neutral, or negative? Answer only with a number: 1 if positive, 2 if neutral, and 3 if negative. Here is the text: [Tweet text] & Which of these [number of] emotions—[list of emotions]—best represents the mental state of the person writing the following (Indonesian) text? Answer only with a number: 1 if [emotion1], 2 if [emotion2], [...]. Here is the text: [Tweet text] & Is the following post offensive? Answer only with a number: 1 if offensive, and 0 if not offensive. Here is the post: [Tweet text] &  Is the following post ironic? Please answer with numbers only: 1 for irony and 0 for non-irony. Here is the post:  [Tweet text] \\
\bottomrule
\end{tabular}
\end{table}

During the reasoning tests, we found that o1-preview performed very well in most cases, with its predictions aligning closely with the ground truth. However, there were some instances where it produced incorrect results. For example, such errors were observed in both sentiment analysis and emotion recognition tasks.
In the case of sentiment analysis, o1-preview mistakenly judged a tweet with neutral sentiment as containing negative sentiment.
When performing sentiment analysis on this example, o1-preview output was: \textit{I'm mapping out the sentiment expressed in the text, where it condemns a massacre and bombings, and advocates action for Aleppo, indicating both negative and positive sentiments}.
Although o1-preview ultimately gave the wrong answer, it was actually unsure during its reasoning whether the text had more positive or negative sentiments. Despite this uncertainty, it did not choose the neutral option that lies between positive and negative sentiments.
Similarly, this phenomenon was also observed in another error case on the emotion recognition task.

Although our test results are not comprehensive due to OpenAI's policy restrictions, we have roughly summarized some reasons that may cause o1-preview to analyze incorrect results in some cases.
First, these tasks require deep semantic understanding and the ability to interpret subtle emotional cues, such as sarcasm or tone, which models may struggle to capture accurately.
emotions often rely on implicit context, cultural background, or even non-verbal cues, which are difficult for text-based models to grasp. This suggests that while large language models excel in many areas, they still face significant limitations when it comes to understanding and interpreting human emotions.
Overall, the performance results of o1-preview on social media analysis tasks are positive. It captures complex expressions such as irony, sarcasm, and humor on most tasks.
\newpage

\newpage

\subsection{Analogical Reasoning}
To evaluate the analogical reasoning ability of o1-preview, we designed a set of experiments focused on semantic structure mapping, where the model needs to transfer semantic relations from one domain to another. These tasks were created to test various aspects of analogical reasoning. The complexity of these tasks varies, from basic relational analogies to more intricate challenges involving multiple layers of semantic content.

The experimental tasks were divided into two main categories:
\begin{enumerate}
    \item \textbf{Mapping of Semantic Structure:} These tasks involve identifying and transferring relational structures between semantically meaningful symbols. The model must align abstract relationships such as those between "dog" and "puppy" with non-linguistic representations.
    \item \textbf{Mapping of Semantic Content:} The model needed to transfer specific content-related attributes from one domain to another. For example, it must infer that the "dog" symbol typically has four legs, converting it into a non-semantic representation.
\end{enumerate}

The prompt template used in this experiment was: "We are conducting an experiment on general reasoning abilities. You will be presented with a series of words and drawings. Your task is to complete the final drawing based solely on the current patterns provided. Please respond concisely with only the final drawing. Question: [Question]".

o1-preview shows varying levels of success across different tasks. It demonstrates a strong ability to analyze and align underlying semantic structures in simple relational analogies. As the complexity of the tasks increases, especially those involving multiple layers of semantic content and structure, the model's performance remaines strong, correctly answering 10 test questions.

\begin{figure}[H]
    \centering
    \includegraphics[width = 0.85\textwidth]{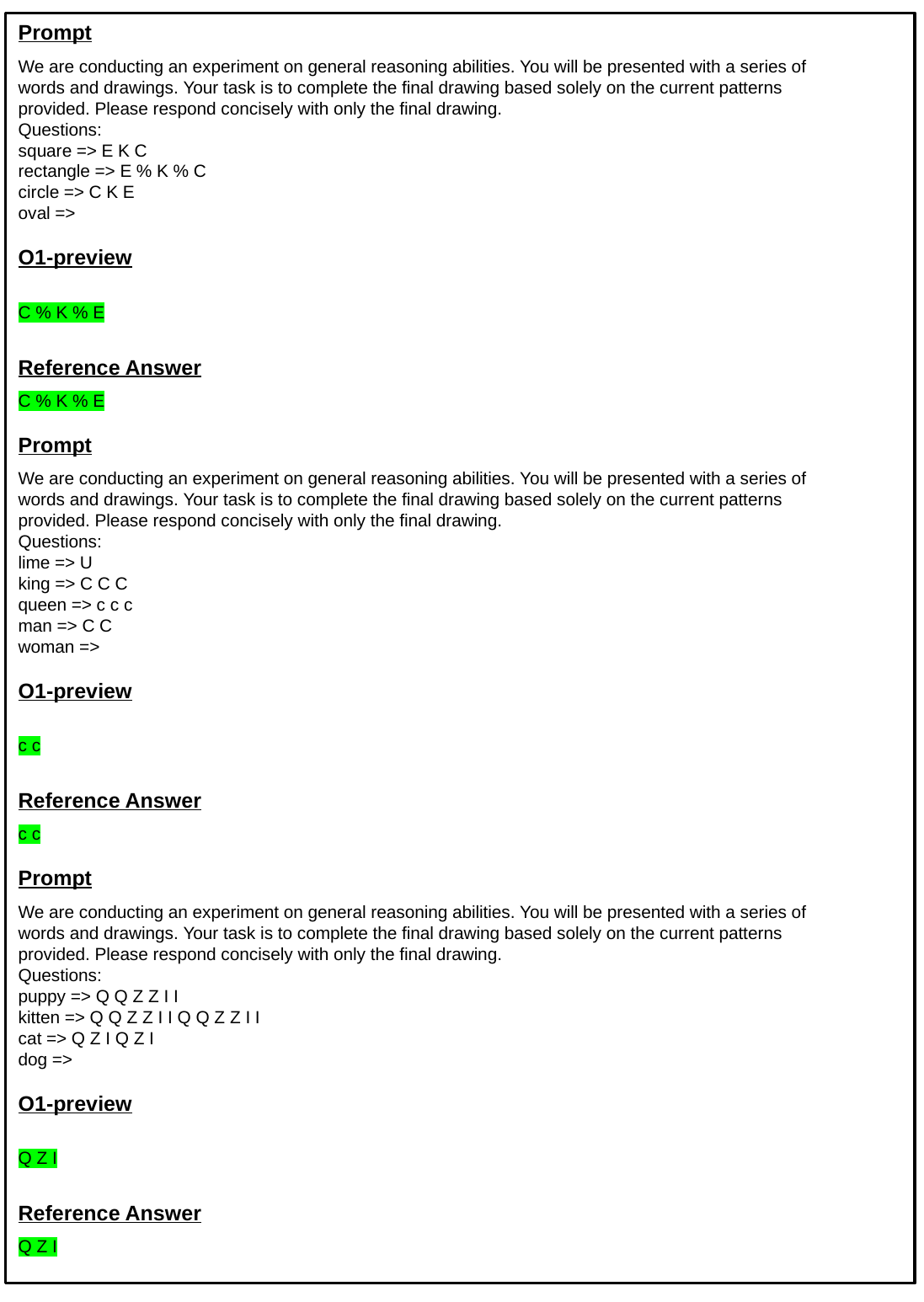}
    \caption[Analogical Reasoning: Case 1-3]{\textbf{Analogical Reasoning: Case 1-3.} A demonstration case for direct semantic structure mapping. Green denotes that the answers are correct. }
    \label{fig:AR_1-3}
\end{figure}

\begin{figure}[H]
    \centering
    \includegraphics[width = 0.85\textwidth]{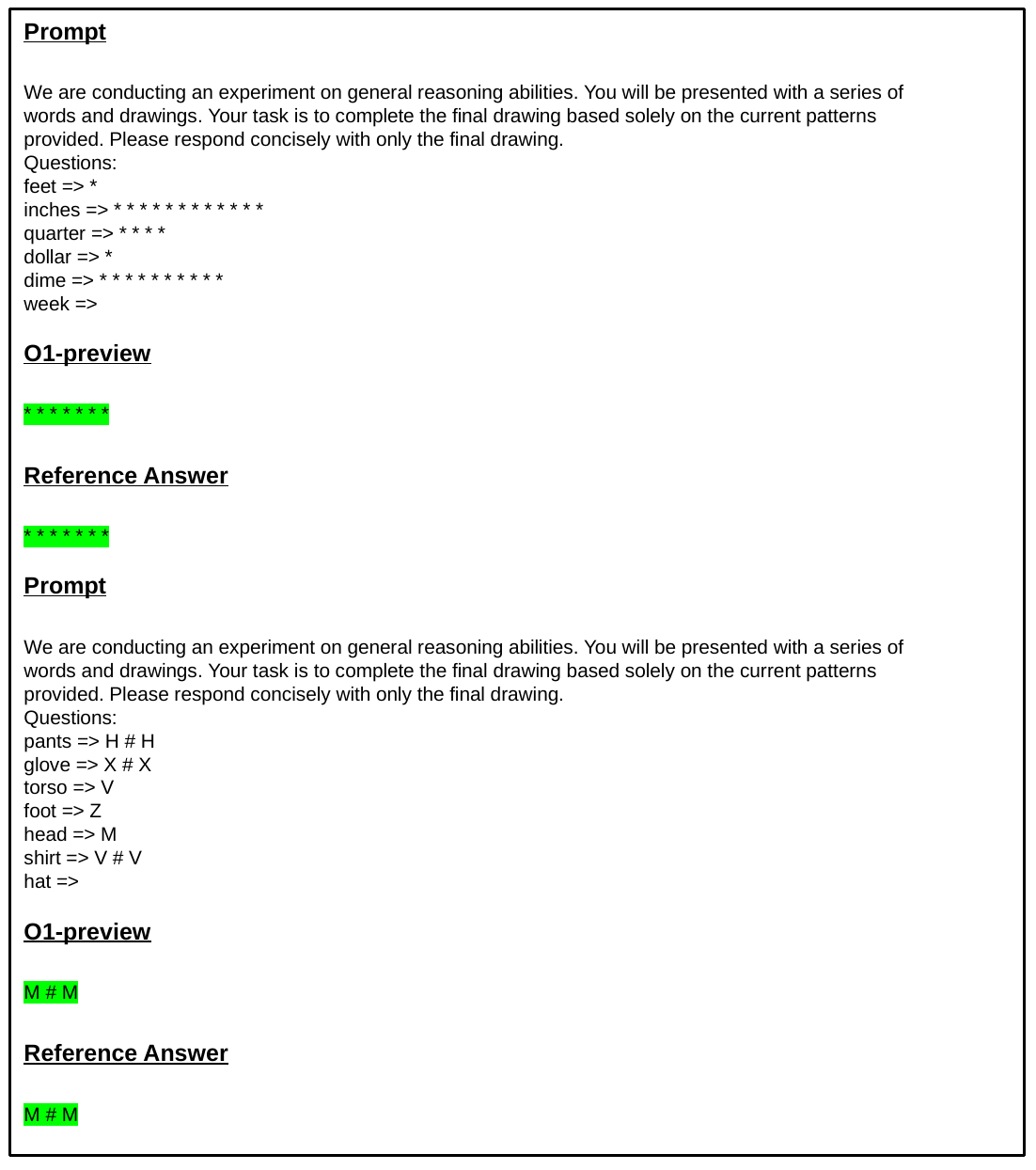}
    \caption[Analogical Reasoning: Case 4-5]{\textbf{Analogical Reasoning: Case 4-5.} A demonstration case for multiple layers of semantic content and structure. Green denotes that the answers are correct. }
    \label{fig:AR_4-5}
\end{figure}

Figure~\ref{fig:AR_1-3} and~\ref{fig:AR_4-5} illustrate this performance. Figure~\ref{fig:AR_1-3} shows examples of direct semantic structure mapping, both with and without distractors, while Figure~\ref{fig:AR_4-5} shows examples of complex analogies. In these tasks, o1-preview effectively eliminates distractions and extracts analogical features through reasoning analysis to obtain correct results.

The observations from these experiments indicate that o1-preview has robust analogical reasoning capabilities. It can derive correct reasoning outcomes in both direct semantic structure mapping and more complex analogies, demonstrating an ability to effectively identify and apply learned relational patterns.

\newpage

\newpage


\newpage

\subsection{Sentiment Analysis}
To evaluate the sentiment analysis capabilities of the o1-preview model, we designed a series of experiments focusing on testing the model's ability to classify and understand sentiments across different domains and text types. We employed three widely-used benchmark datasets: IMDB, SemEval-2014 Task 4, and SemEval-2015 Task 12. These datasets cover sentiment analysis tasks ranging from movie reviews to restaurants, electronics, and social media content, providing a comprehensive evaluation of the model's abilities.

The experimental tasks were divided into 2 main categories: 

\begin{enumerate} 

\item \textbf{Overall Sentiment Classification:} Using the IMDB dataset, the model was tasked with classifying long-form movie reviews as positive or negative. This dataset contains 50,000 user-generated reviews with diverse language styles, including sarcasm, metaphors, and complex linguistic phenomena. 

\item \textbf{Aspect-Based Sentiment Analysis:} Utilizing the SemEval-2014 Task 4 dataset, the model needed to identify sentiments toward specific aspects within the text, such as the service quality of a restaurant or the battery life of a laptop. This required the model to not only understand the overall sentiment but also to analyze sentiment tendencies toward specific attributes. 

\end{enumerate}

In our experiments, we provided the model with appropriate prompt templates, such as: "You are a sentiment analysis expert. Analyze the sentiment of the following text and provide the sentiment category (Positive, Negative, Neutral). Text: [Text Content]." and "You are a sentiment analysis expert. Please analyze all the emotions expressed in the following text. For each emotion, provide the sentiment category, the relevant text snippet, and a brief explanation. Text:[Text Content]."

The o1-preivew model performed excellently across different tasks. In the overall sentiment classification task, it accurately handled complex and lengthy movie reviews, correctly classifying the majority of test samples. In aspect-based sentiment analysis, the model successfully identified sentiments toward specific aspects within the text and could distinguish different sentiments toward different aspects within the same text. 

\begin{figure}[H]
    \centering
    \includegraphics[width = \textwidth]{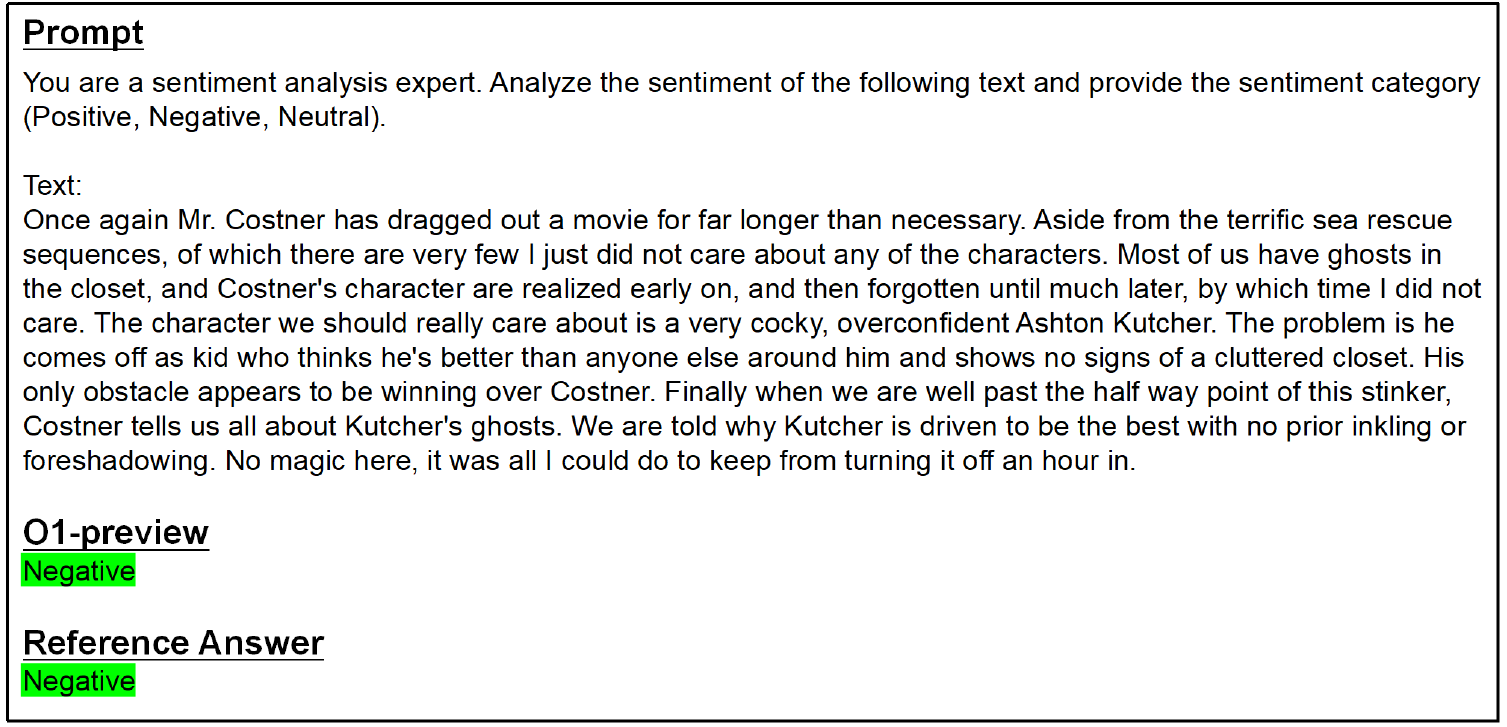}
    \caption[Sentiment Analysis: Case 1.]{\textbf{Sentiment Analysis: Case 1.} An example illustrating the capability of o1-preview in the field of sentiment analysis. }
    \label{fig:senti_1}
\end{figure}

\begin{figure}[H]
    \centering
    \includegraphics[width = \textwidth]{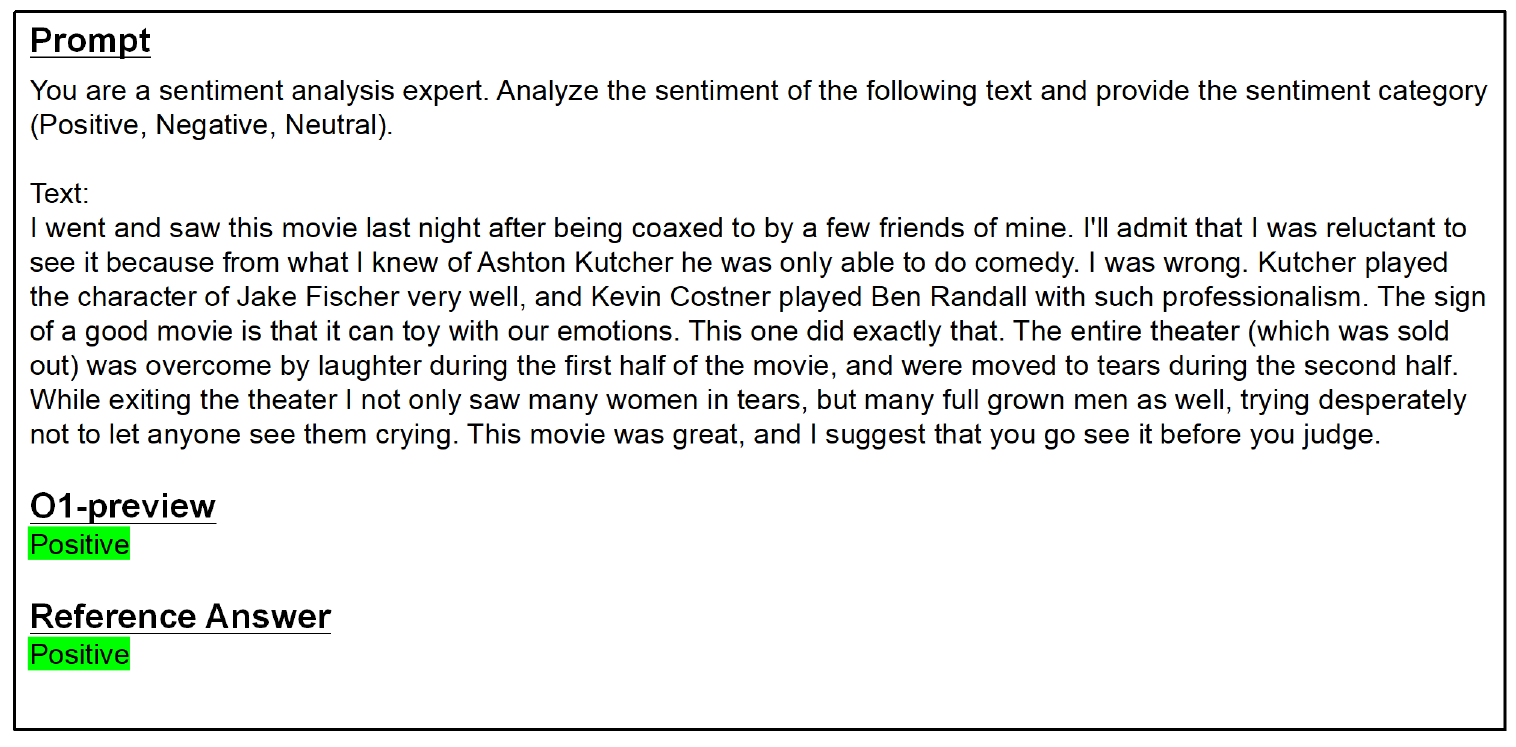}
    \caption[Sentiment Analysis: Case 2.]{\textbf{Sentiment Analysis: Case 2.} An example illustrating the capability of o1-preview in the field of sentiment analysis. }
    \label{fig:senti_2}
\end{figure}

\begin{figure}[H]
    \centering
    \includegraphics[width = \textwidth]{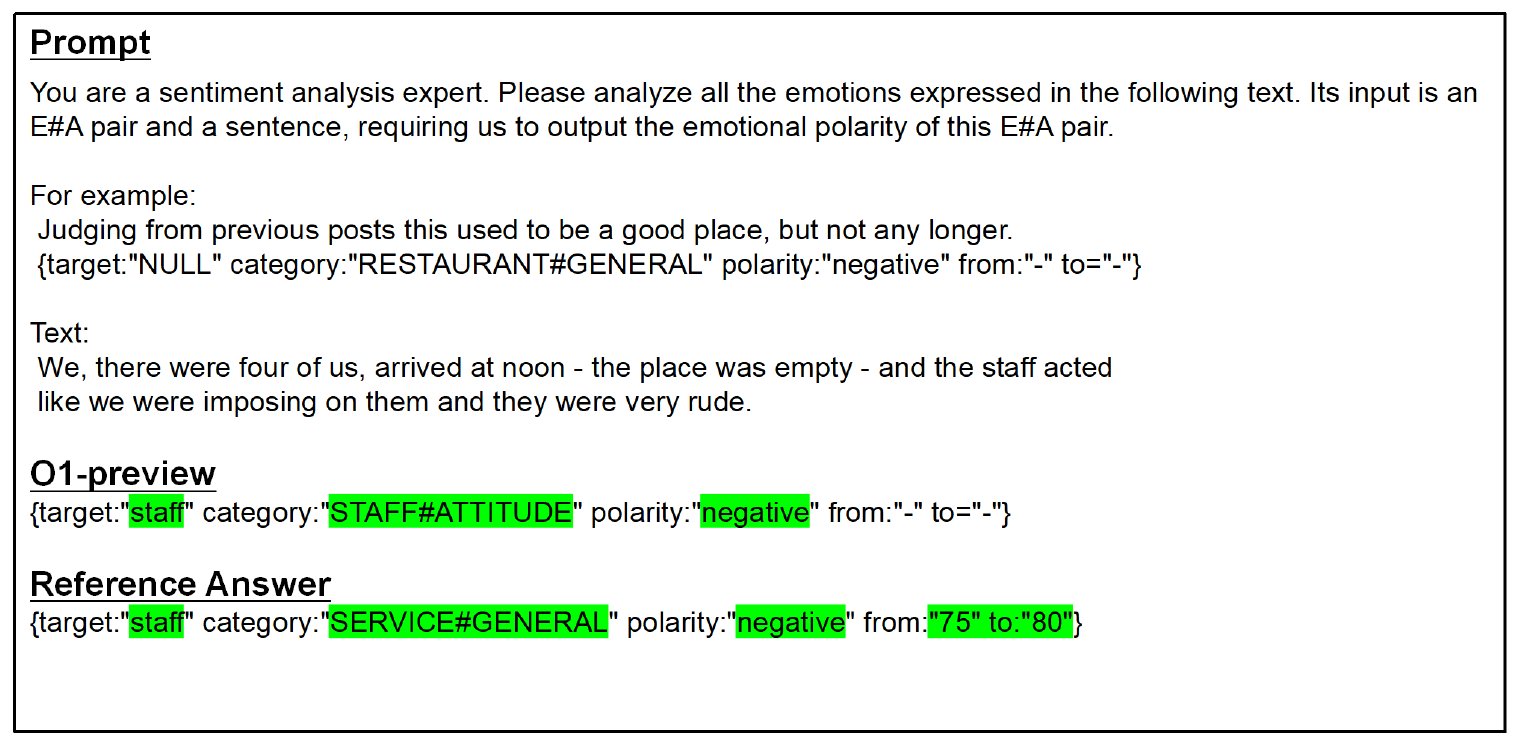}
    \caption[Sentiment Analysis: Case 3.]{\textbf{Sentiment Analysis: Case 3.} An example illustrating the capability of o1-preview in the field of sentiment analysis. }
    \label{fig:senti_3}
\end{figure}

\begin{figure}[H]
    \centering
    \includegraphics[width = \textwidth]{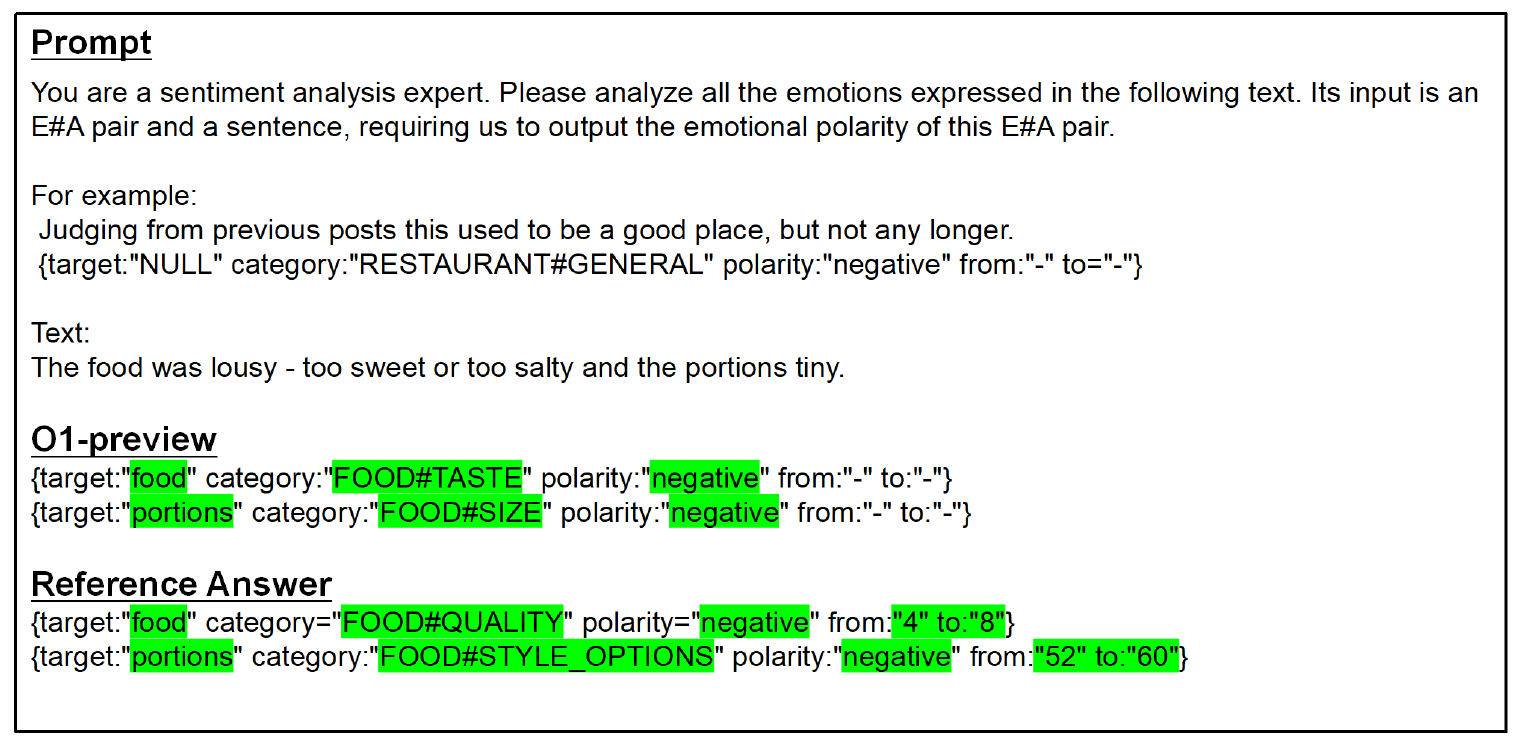}
    \caption[Sentiment Analysis: Case 4.]{\textbf{Sentiment Analysis: Case 4.} An example illustrating the capability of o1-preview in the field of sentiment analysis. }
    \label{fig:senti_4}
\end{figure}

These results are illustrated in Figures~\ref{fig:senti_1},~\ref{fig:senti_2},~\ref{fig:senti_3},~\ref{fig:senti_4},~\ref{fig:senti_5},~\ref{fig:senti_6}. From these experiments, it can be observed that the o1-preview model possesses strong sentiment analysis capabilities, accurately performing sentiment classification and analysis across different domains and text types. It excels in understanding complex linguistic structures and identifying implicit sentiments.

Overall, these experiments demonstrate that the o1-preview model has strong capabilities in sentiment analysis, achieving a high level of sentiment understanding across different domains and text types.

\newpage


\subsection{Anthropology and Geology}
\begin{figure}[H]
    \centering
    \includegraphics[width = \textwidth]{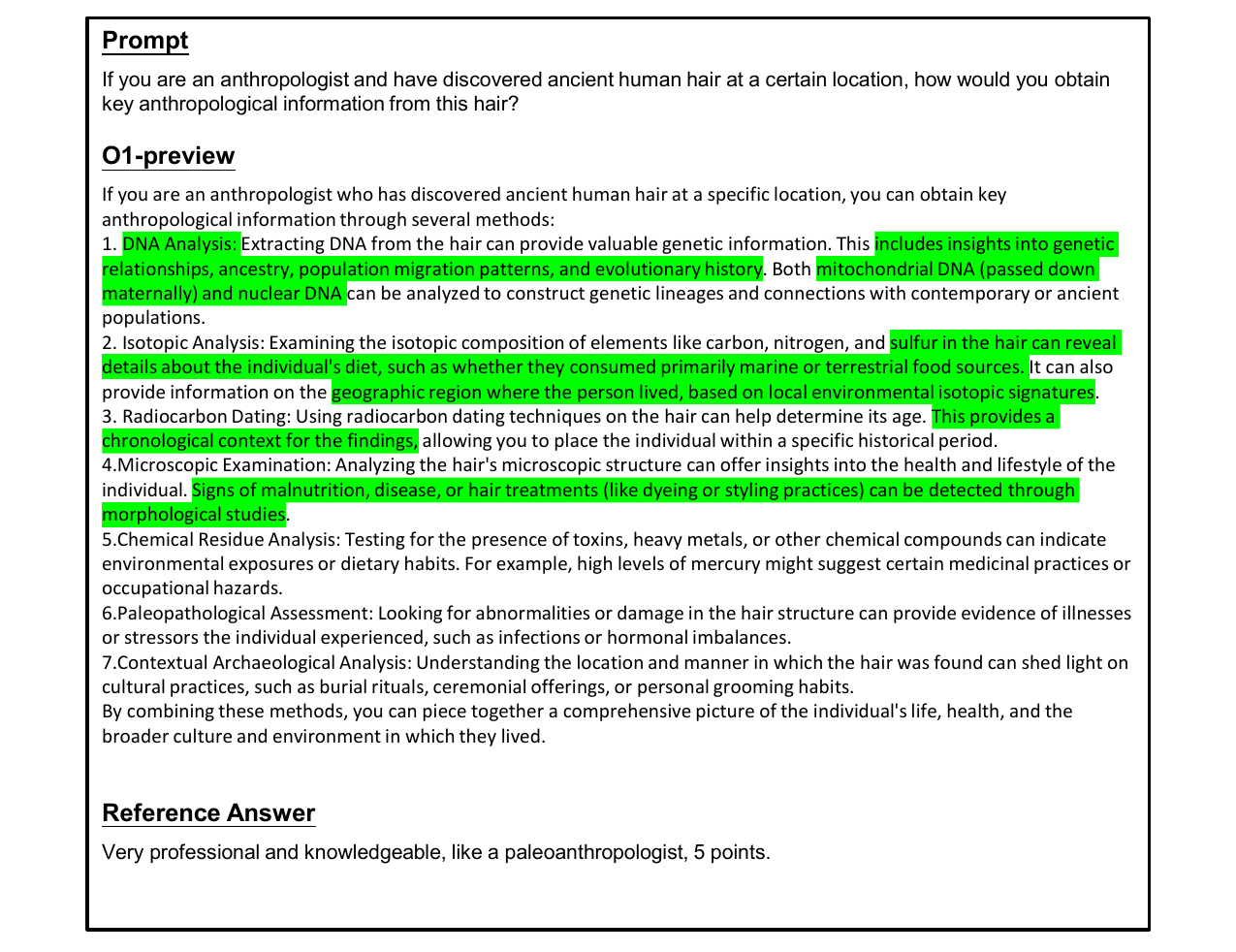}
      \caption[Anthropology Reasoning: Case 1.]{\textbf{Anthropology Reasoning: Case 1.} An example illustrating the capability of o1-preview to respond to specialized questions in the domain of paleoanthropology. Green denotes that the answers are correct. }
    \label{fig:paleo_geo_1}
\end{figure}

\begin{figure}[H]
    \centering
    \includegraphics[width = \textwidth]{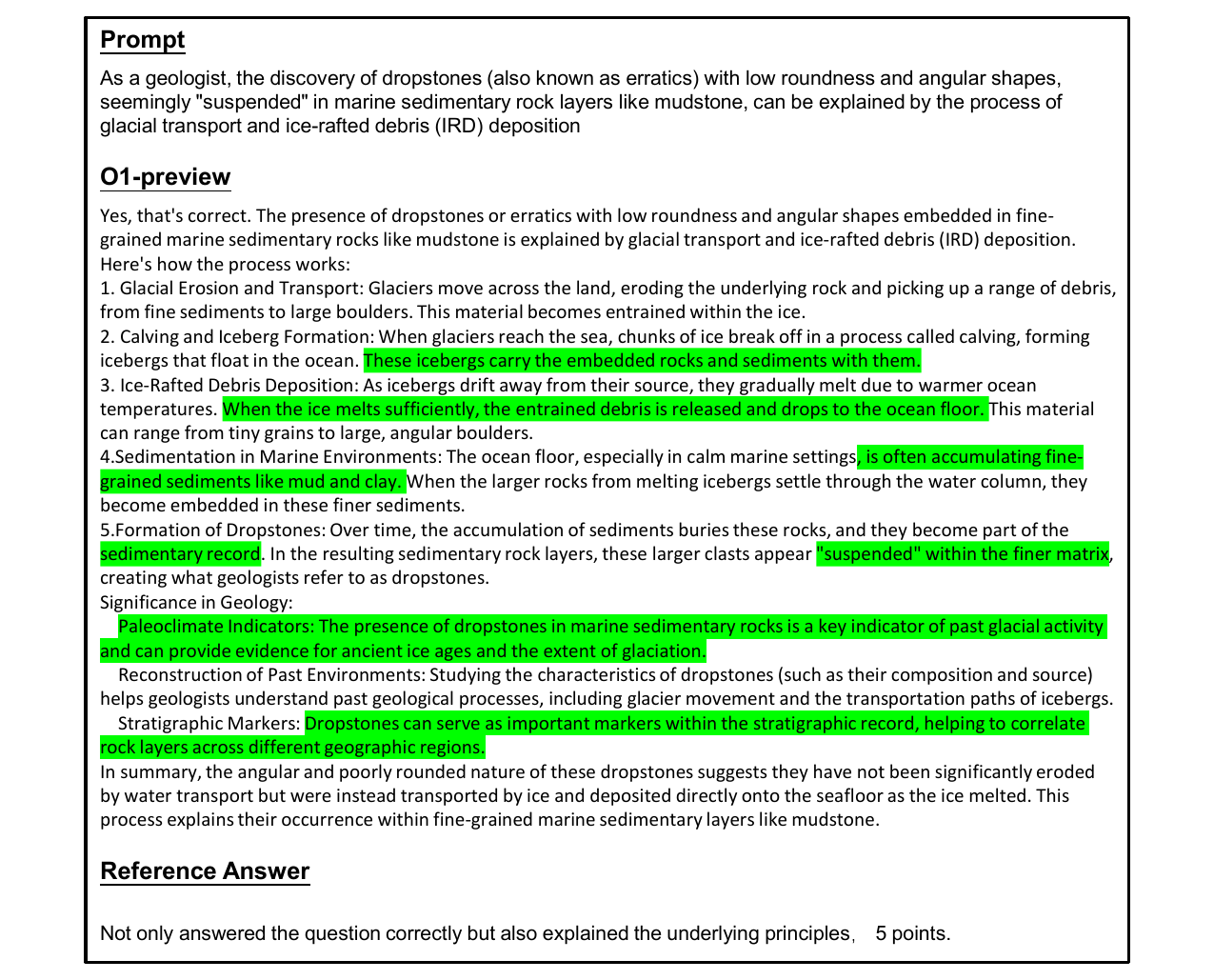}
     \caption[Geology Reasoning: Case 1.]{\textbf{Geology Reasoning: Case 1.} An example illustrating the capability of  o1-preview to respond to specialized questions in the domain of Geology. Green denotes that the answers are correct.}
    \label{fig:paleo_geo_7}
\end{figure}

Testing has shown that  o1-preview is currently capable of answering complex questions in paleoanthropology and geology. It can address detailed aspects of ancient genes in paleoanthropological research (Figure~\ref{fig:paleo_geo_1}) and has a deep understanding of geological issues(Figure~\ref{fig:paleo_geo_7}). As the questions become more specific,  o1-preview is able to provide increasingly detailed and in-depth responses, offering researchers in these fields relevant information, current research status, and commonly used methods. Additionally, it can engage in discussions on specific topics, much like a peer. However, it lacks creative thinking and cannot delve into the finer details of particular issues. Nonetheless, it serves as a valuable resource for providing quick, introductory guidance to a given field.

\newpage

\begin{figure}[H]
    \centering
    \includegraphics[width = .8\textwidth]{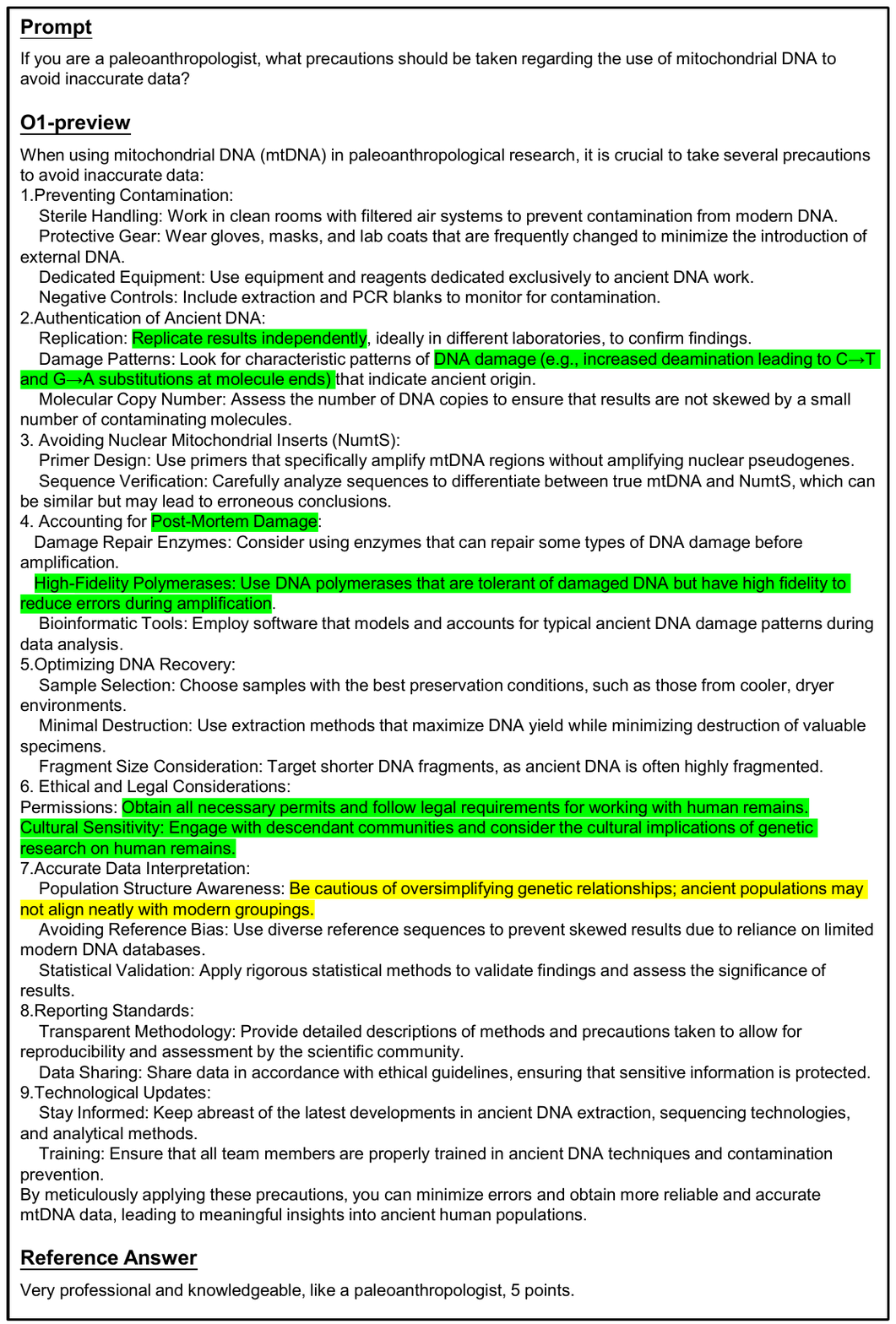}
    \caption[Anthropology Reasoning: Case 2.]{\textbf{Anthropology Reasoning: Case 2.} An example illustrating the capability of o1-preview to respond to specialized questions in the domain of paleoanthropology. Green denotes that the answers are correct. }
    \label{fig:paleo_geo_2}
\end{figure}
\newpage

\begin{figure}[H]
    \centering
    \includegraphics[width = \textwidth]{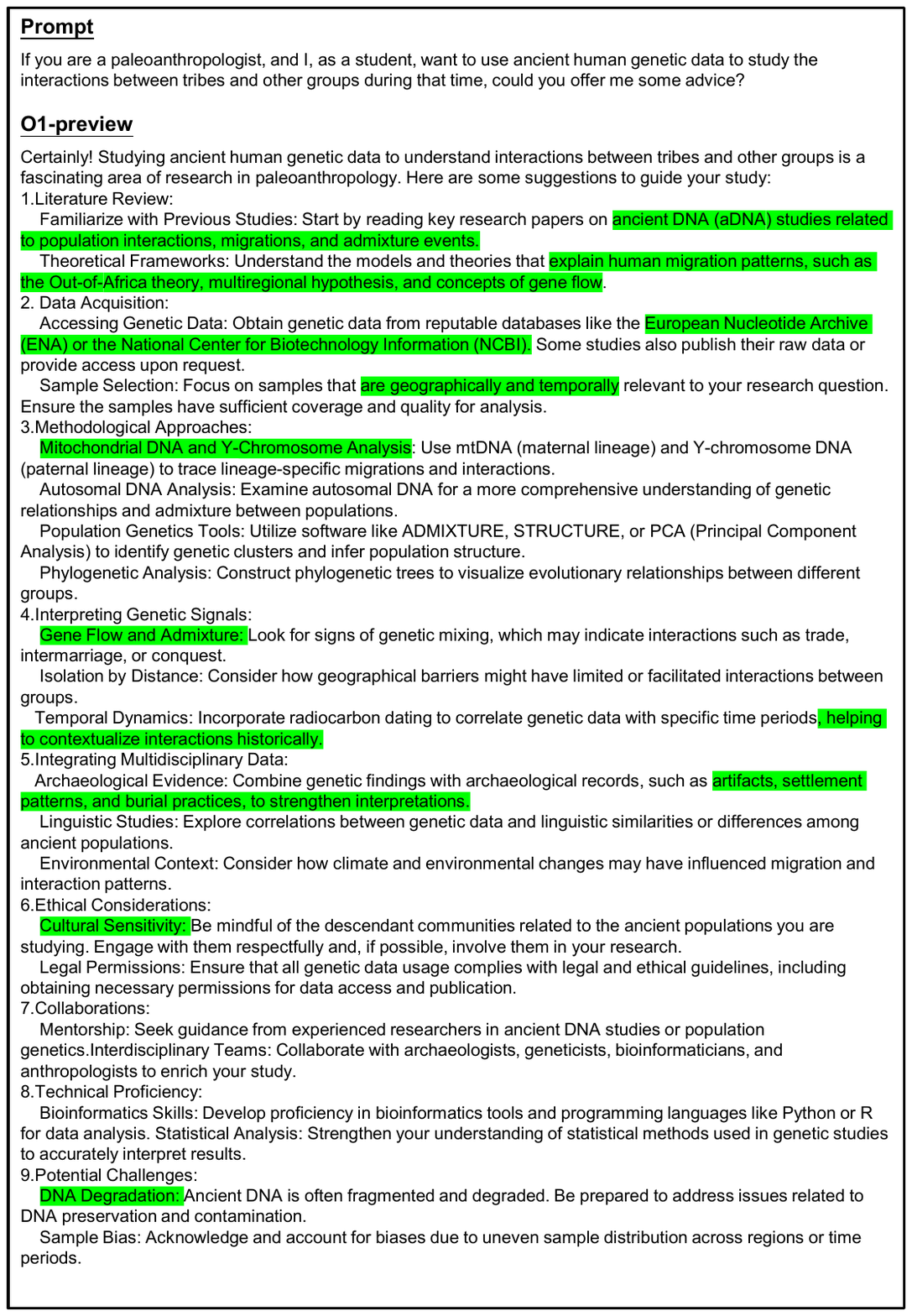}

    \label{fig:paleo_geo_3}
\end{figure}
\newpage

\begin{figure}[H]
    \centering
    \includegraphics[width = \textwidth]{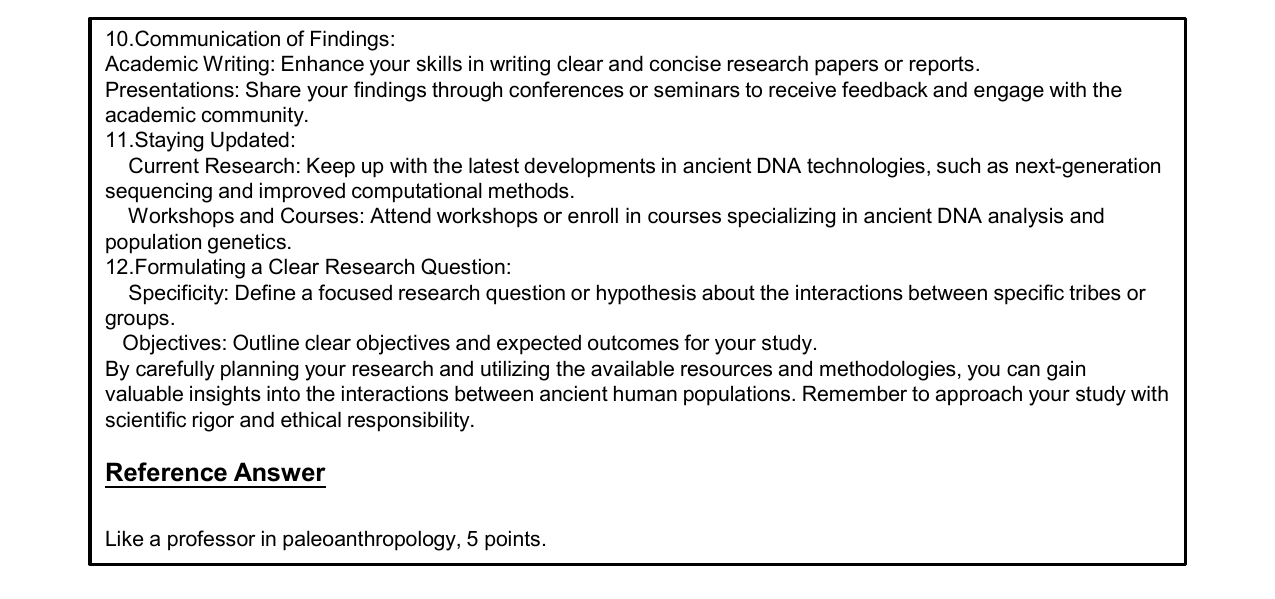}

   \caption[Anthropology Reasoning: Case 3.]{\textbf{Anthropology Reasoning: Case 3.} An example illustrating the capability of  o1-preview to respond to specialized questions in the domain of paleoanthropology. Green denotes that the answers are correct. }
    
    \label{fig:paleo_geo_4}
\end{figure}
\newpage

\begin{figure}[H]
    \centering
    \includegraphics[width = \textwidth]{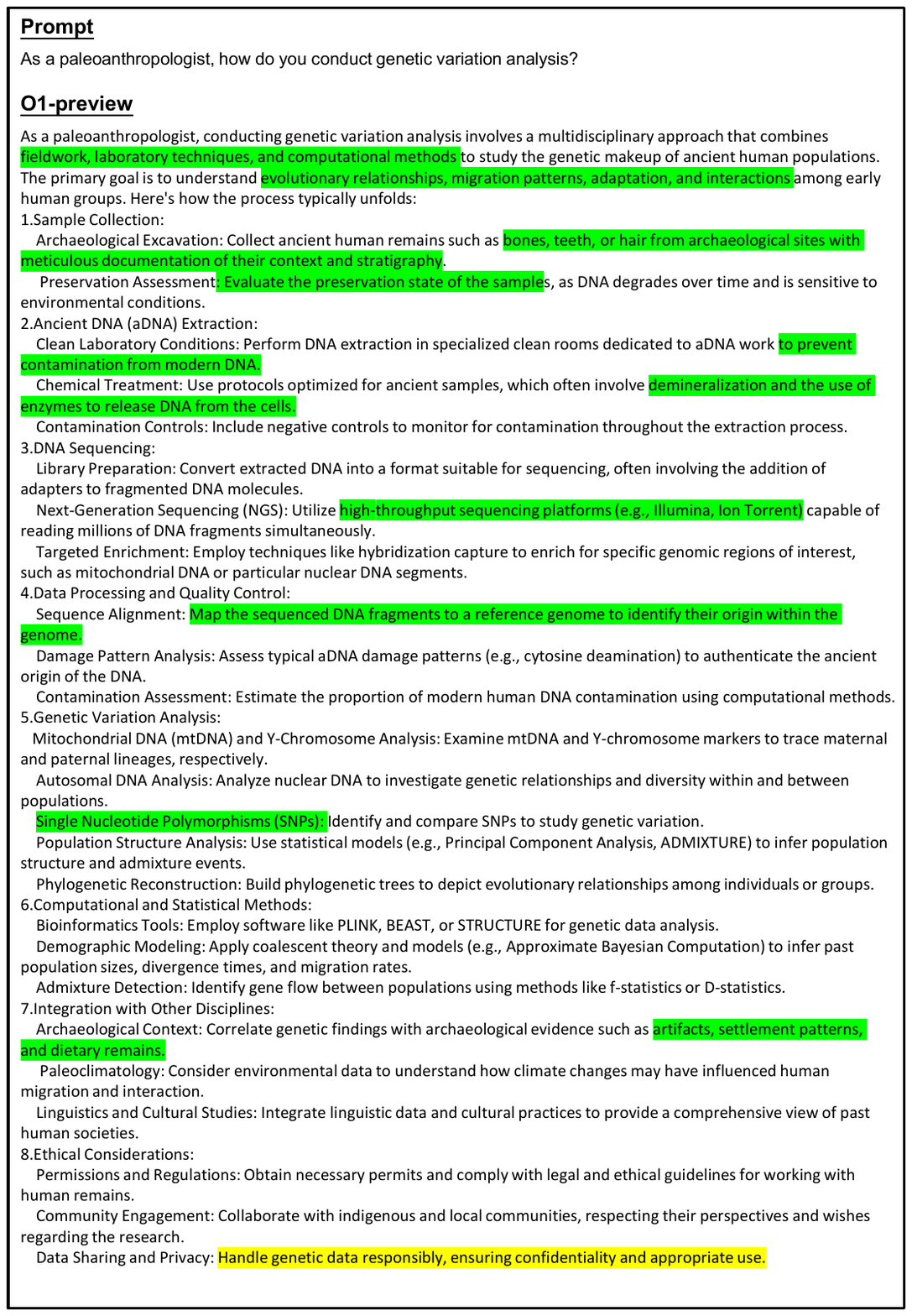}
    
    \label{fig:paleo_geo_5}
\end{figure}
\newpage

\begin{figure}[H]
    \centering
    \includegraphics[width = \textwidth]{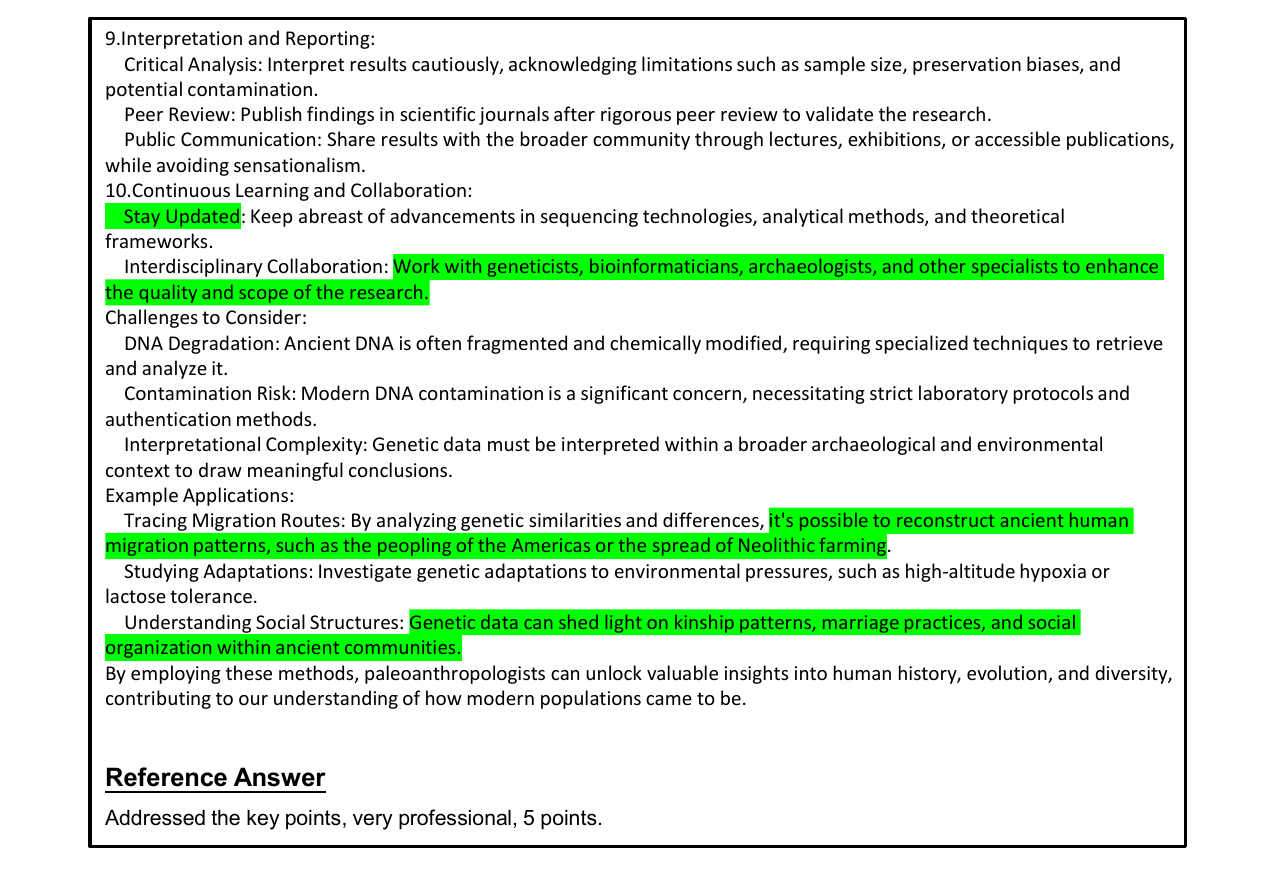}
    \caption[Anthropology Reasoning: Case 4.]{\textbf{Anthropology Reasoning: Case 4.} An example illustrating the capability of  o1-preview to respond to specialized questions in the domain of paleoanthropology. Green denotes that the answers are correct.}
    \label{fig:paleo_geo_6}
\end{figure}
\newpage

\newpage


\subsection{Educational Measurement and Psychometrics}
\begin{figure}[H]
    \centering
    \includegraphics[width = 0.85\linewidth]{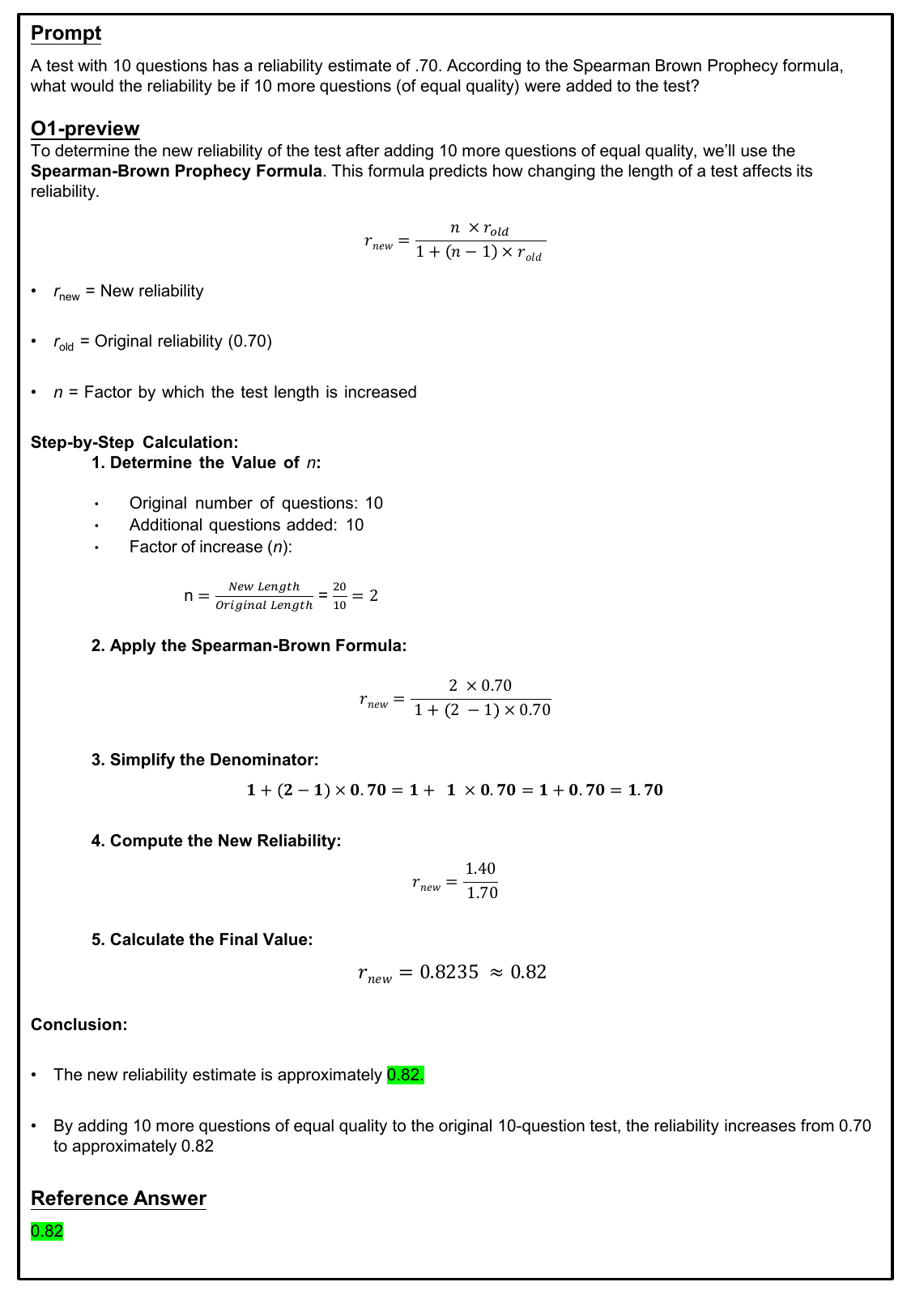}
    \caption[Educational Measurement and Psychometrics: Case 1]{\textbf{Educational Measurement and Psychometrics: Case 1.} An example illustrating the capability of o1-preview to respond to specialized questions of test reliability in the domain of educational measurement and psychometrics. Green indicates complete consistency.}
    \label{fig:emp_case1}
\end{figure}

In evaluating o1-preview's performance in educational measurement and psychometrics, we selected multiple representative quiz questions from an introduction-level course named \textit{Measurement Theory} based on Bandalos' textbook~\cite{bandalos2018measurement} for first-year graduate students. Our goal is to evaluate the performance of the o1-preview in addressing key concepts in a relatively narrow domain from both theoretical and practical perspectives, including test score reliability, standard error of measurement, and interrater reliability. Each prompt was designed to test the o1-preview's efficacy in providing concrete explanations, applying appropriate statistical techniques, and offering solutions that align with established psychometric frameworks. The following sections provide detailed descriptions of all testing scenarios:

For the first task in Figure \ref{fig:emp_case1}, which involved calculating a reliability coefficient using the Spearman-Brown Prophecy formula, o1-preview demonstrated impressive accuracy. The model not only applied the correct formula but also provided a detailed explanation of each step, making the process informative and easy to follow. This suggests that o1-preview is proficient in handling statistical calculations related to test reliability, which is an important area of educational psychometrics. Its ability to calculate and articulate the reasoning behind the steps reflects a strong understanding of this foundational area in the field.

\begin{figure}[H]
    \centering
    \includegraphics[width = 0.85\linewidth]{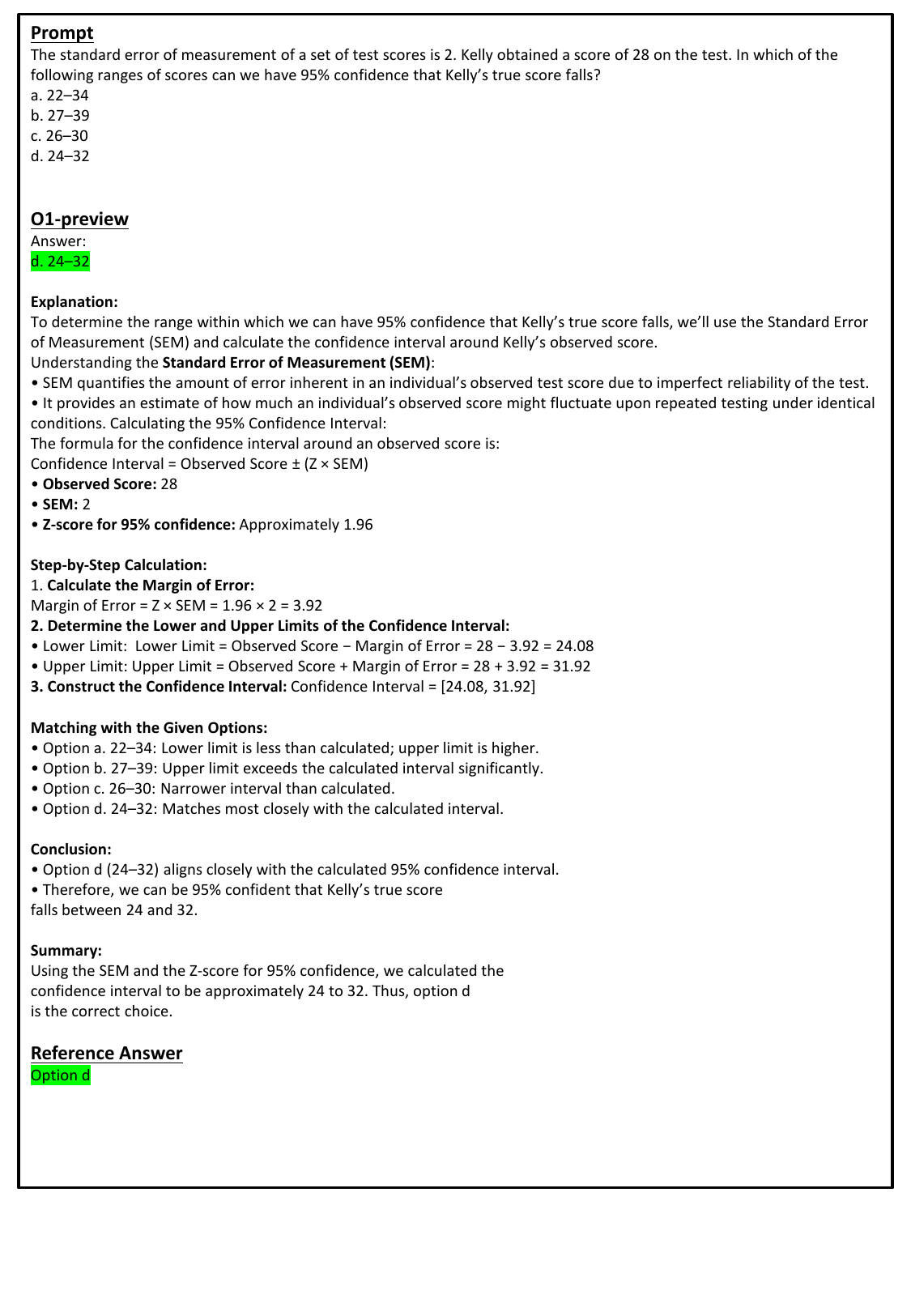}
    \caption[Educational Measurement and Psychometrics: Case 2]{\textbf{Educational Measurement and Psychometrics: Case 2.} An example illustrating the capability of o1-preview to respond to specialized questions of standard error of measurement in the domain of educational measurement and psychometrics. Green indicates complete consistency.}
    \label{fig:emp_case2}
\end{figure}

In the second task in Figure \ref{fig:emp_case2}, which assessed o1-preview understanding of the standard error of measurement and confidence intervals, the model once again delivered a solid response. It accurately calculated the correct range of scores, demonstrating its capability to identify and apply measurement concepts to real-world test scenarios. This further highlights o1-preview's ability to provide reliable and informed outputs in educational measurement tasks.

\begin{figure}[H]
    \centering
    \includegraphics[width = 0.85\linewidth]{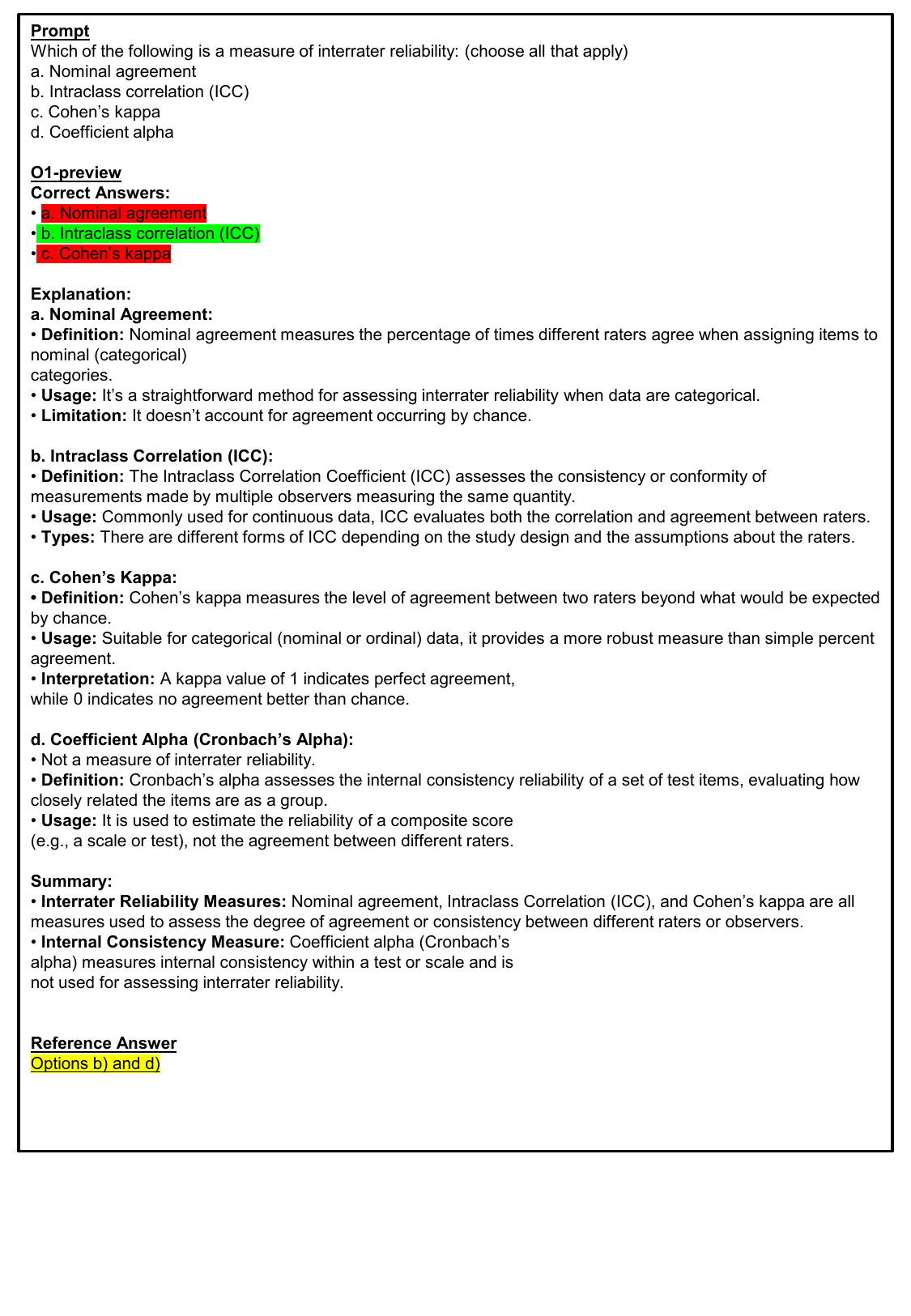}
    \caption[Educational Measurement and Psychometrics: Case 3] {\textbf{Educational Measurement and Psychometrics: Case 3.} An example illustrating the capacity of the o1-preview to respond to specialized questions of interrater reliability in the domain of educational measurement and psychometrics. Red represents inconsistency between the o1-preview response and the reference answer, yellow indicates partial consistency, while green signifies complete consistency.}
    \label{fig:emp_case3}
\end{figure}

However, o1-preview’s performance in the third task in Figure \ref{fig:emp_case3} exposed some limitations in disentangling similar concepts. The third task aims to solve a multiple choice question inculding more than one correct response that evaluates understanding concepts of interrater agreement and interrater reliability, which are two close concepts that measure the consistency of multiple raters' ratings on a performance assessment (e.g., an essay). The model incorrectly identified nominal agreement and Cohen’s kappa as measures of interrater reliability, when the appropriate measures should have been Intraclass Correlation (ICC) and Coefficient Alpha. However, the o1-preview offered the correct definition of the four measures together with the practical application scenarios. This response showcases o1-preview's lack of understanding of the connections between key concepts when the potential training set provides limited information. 

In sum, the o1-preview excelled in tasks requiring statistical computation and reasoning, demonstrating a solid grasp of standard psychometric concepts that are equivalent to or beyond a first-year master's or doctoral student's level in educational measurement and psychometrics. However, our tests revealed some weaknesses in its ability to distinguish closely related concepts even under simple questions. 

\newpage

\subsection{Public Health Policy Analysis}

Throughout the evaluation of Public Health Policy Analysis, we tested o1-preview’s capacity to analyze health policy with respect to the Affordable Care Act (ACA). The test targeted measuring the model's accuracy in responding to questions about the ACA's effects on insurance coverage, healthcare disparities, preventive services, and Medicaid expansion.

\begin{figure}[H]
        \centering
        \includegraphics[width=0.85\linewidth]{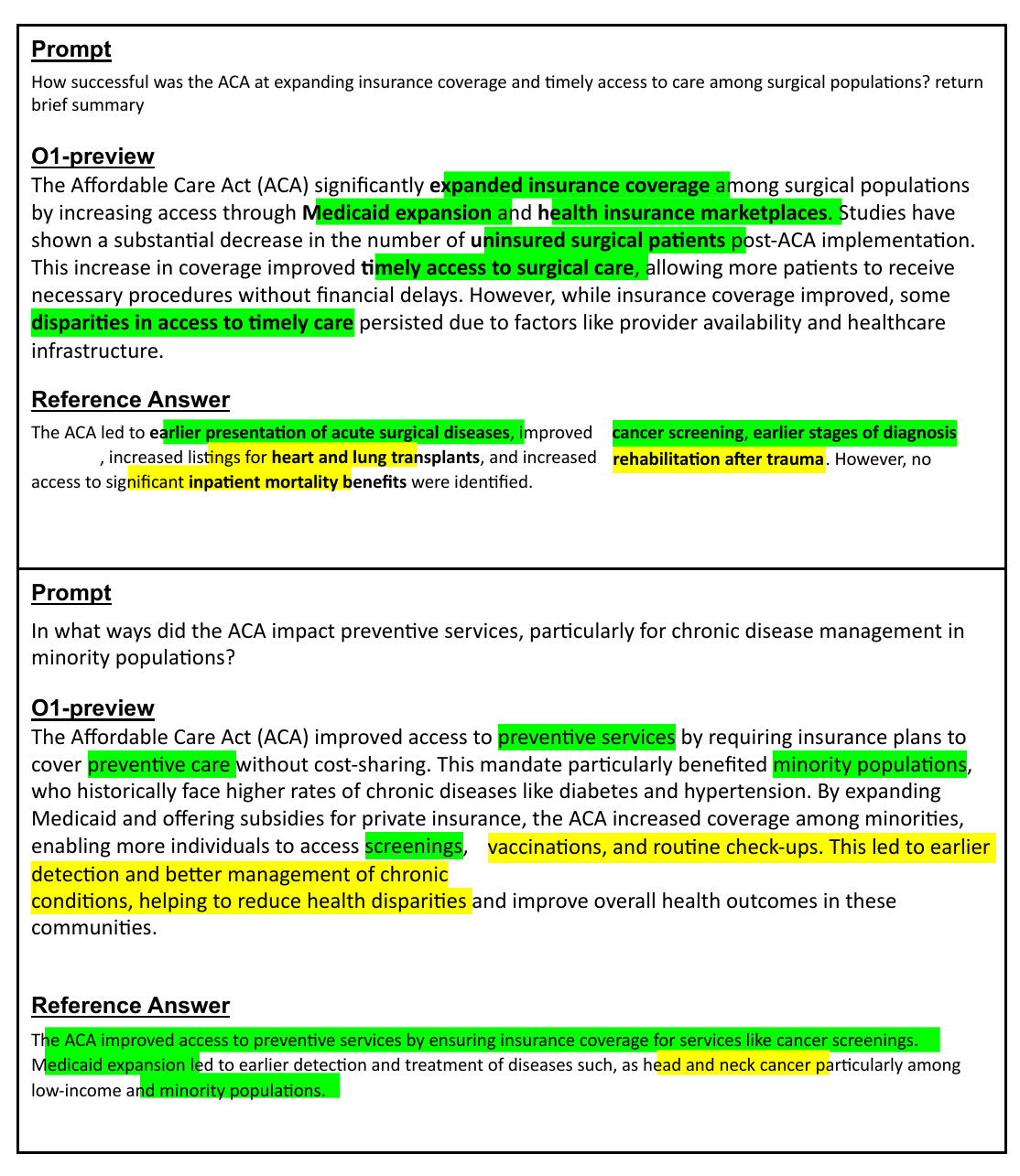}
        \caption[Public Health Policy Analysis: Case 1-2.]{{\textbf{Public Health Policy Analysis: Case 1-2.}} Two examples illustrating the capability of o1-preview to respond to specialized questions of health policy in the domain of Public Health. Yellow indicates partial consistency between the o1-preview response and the reference answer, while green signifies complete consistency.}
        \label{fig:PH_1-2}
    \end{figure}

Based on the test samples, o1-preview demonstrated strong alignment with reference answers, showing a clear grasp of the ACA's key impacts. For example, in \ref{fig:PH_1-2} Case 1, the model successfully identified the overall trend but overlooked specifics, such as the increased listings for heart and lung transplants. In \ref{fig:PH_1-2} Case 2, it provided precise insights into increased coverage and access to care, closely matching the reference answer.

\begin{figure}[H]
    \centering
    \includegraphics[width=0.85\linewidth]{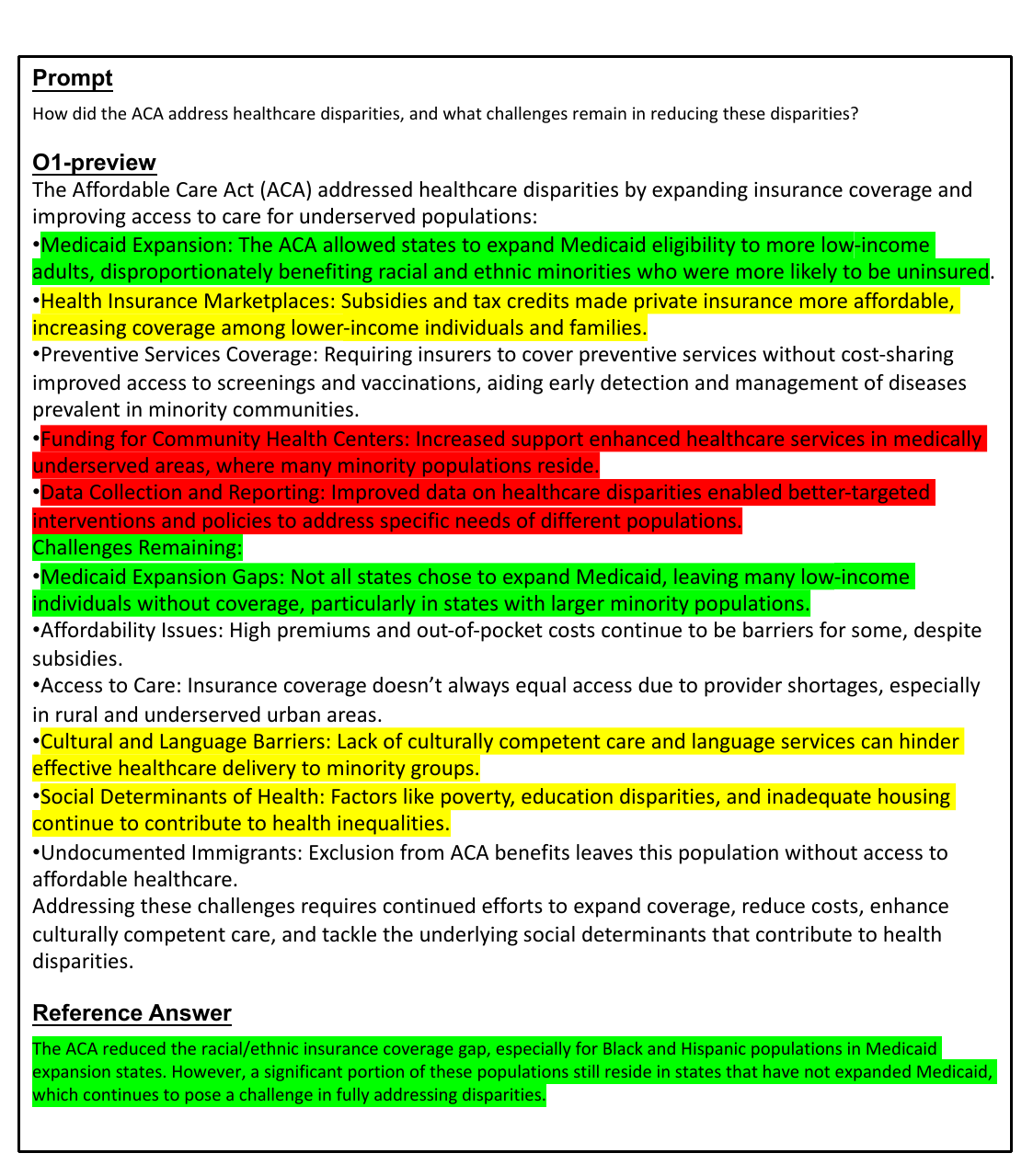}
    \caption[Public Health Policy Analysis: Case 3.]{{\textbf{Public Health Policy Analysis: Case 3.}} An example illustrating the capability of o1-preview to respond to specialized questions of health policy in the domain of Public Health. Yellow indicates partial consistency between the o1-preview's response and the reference answer,  and red indicates full consistency, while green signifies complete consistency.}
    \label{fig:PH_3}
\end{figure}

However, the o1-preview model occasionally lacked depth in addressing specific topics, as shown in \ref{fig:PH_3}, where key disease-specific examples were omitted. This limitation reduced its ability to capture the nuances of the ACA’s effects entirely on certain populations. This suggests that the model requires further refinement, especially in areas demanding advanced analysis and detailed specificity.

In this domain, o1-preview performed robustly in analyzing the Affordable Care Act and its broader public health impacts. The model effectively highlighted significant elements of health policy, such as Medicaid expansion, preventive services, and efforts to address healthcare disparities. Despite some limitations in providing detailed explanations, its reasoning aligned well with the ground truth across most cases. 

For future evaluations, we could consider using larger datasets, such as the \textit{Compilation of State Data on the Affordable Care Act}\cite{aca_state_data_2024}, which would allow for more thorough training and testing of o1-preview’s abilities. Additionally, incorporating more evaluation metrics, along with broader perspectives on policy evaluation, would offer a more comprehensive assessment of the model’s public health policy analysis capabilities. 

With further improvements in providing more specific examples and advanced analysis, o1-preview will exhibit significant potential for excelling in public health policy analysis and policy-making tasks. 
\newpage

\newpage

\subsection{Medical Genetics and Genomics Reasoning}
\begin{figure}[H]
        \centering
        \includegraphics[width=0.99\linewidth]{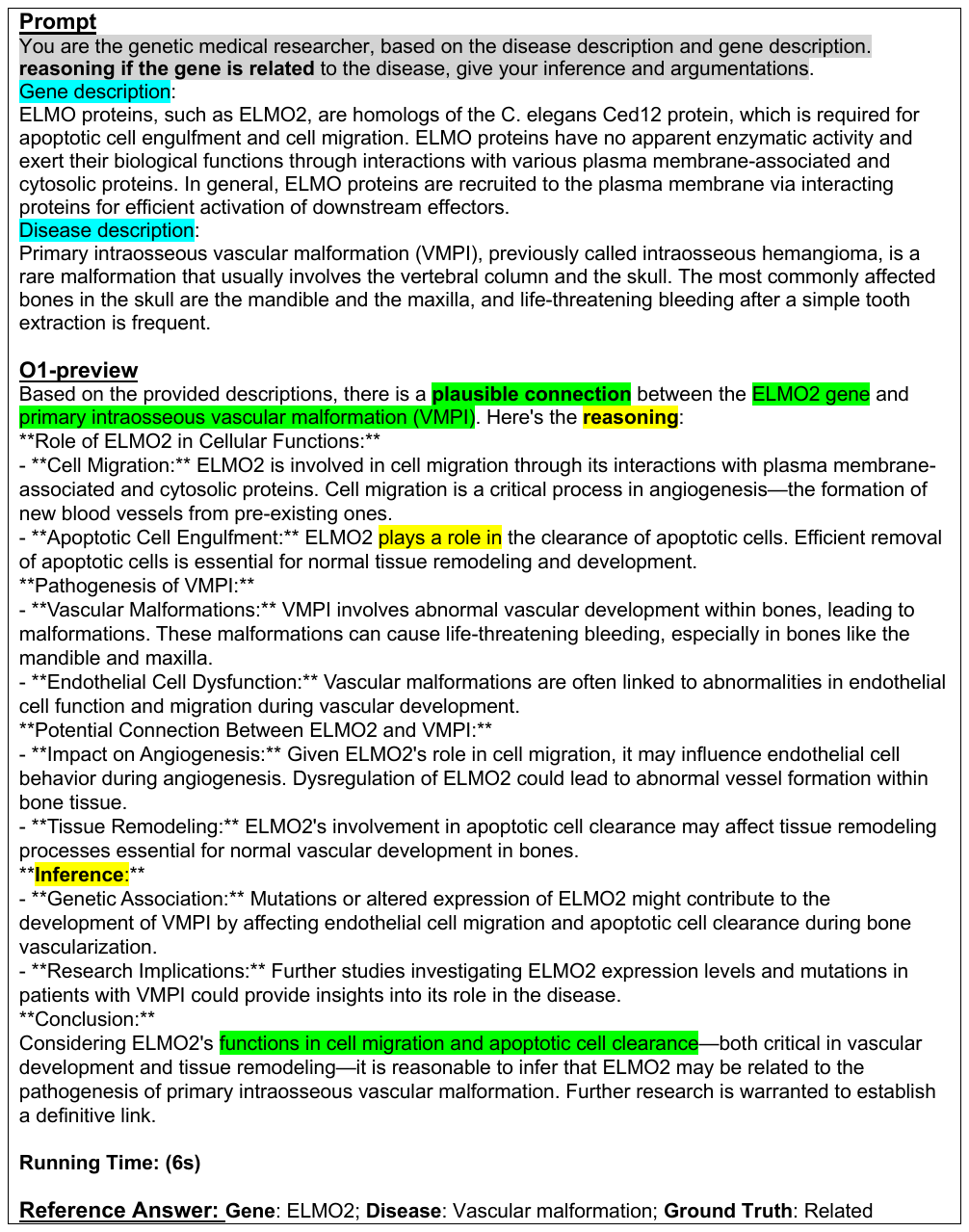}
        \caption[Medical Genetics and Genomics Reasoning: Case 1.]{\textbf{Medical Genetic and Genomics Reasoning: Case 1.} Positive example illustrating the capability of the o1-preview to respond to gene-disease-related questions. Grey indicates the prompt instructions. Blue indicates the gene description and disease description. Green indicates the correct inferences. Yellow indicates the detailed deduction steps of the o1-preview.}
        \label{fig:GD_3-1}
\end{figure}

\begin{figure}[H]
    \centering
    \includegraphics[width=0.99\linewidth]{ 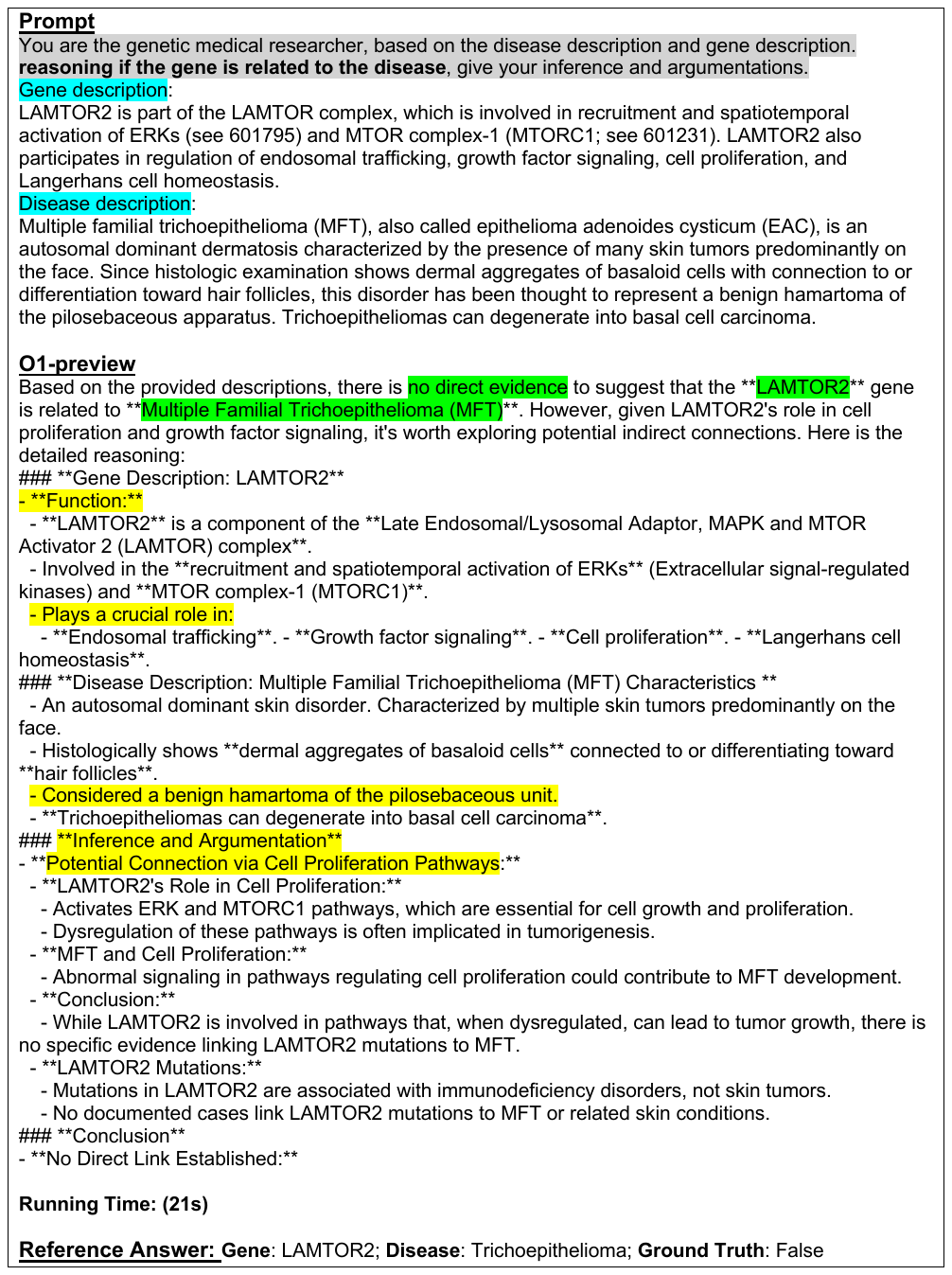 }
    \vspace{-1cm}
    \caption[Medical Genetics and Genomics Reasoning: Case 2.]{\textbf{Medical Genetics and Genomics Reasoning: Case 2.} Positive example illustrating the capability of o1-preview to respond to gene-disease-related questions. Grey indicates the prompt instructions. Blue indicates the gene description and disease description. Green indicates the correct inferences. Yellow indicates the detailed deduction steps of the o1-preview.}
    \label{fig:GD_3-2}
\end{figure}

To evaluate the model's reasoning ability in genomics and medical genetics, we designed a series of experiments centred around genomics-related questions, requiring the model to generate appropriate responses to predefined questions. These questions focused on relationship predictions and their corresponding explanations. The tasks aimed to assess different facets of genomics reasoning and utilized selected QA tasks from the GenTurning benchmarks~\cite{hou2023geneturing}. In all the tested 20 QA instances, o1-preview yields correct answers. Considering the related information we provided in the prompt, the o1-preview has the ability to generate reliable conclusions by following a correct chain-of-though which is derived from the truth. On observing the details of the reasoning steps, the o1-preview provided explanations for all aspects and showed clearly how it came to the conclusions. We show a true-positive case in (Figure ~\ref{fig:GD_3-1}) and a false-negative case in (Figure ~\ref{fig:GD_3-2}), which in both cases, o1-preview shows satisfactory results. In the view of inference time consumption, it is observed that for the negative terms, the o1-preview takes a significantly longer time to draw the conclusion. In our test, the average running time of a true-positive conclusion is 12.67s, while the average running time of a false-negative conclusion is 20.09s. Based on this observation, it is reasonable to speculate that the o1-preview takes a longer chain-of-though in attempting to reject a genomics hypothesis.  Additional examples can be found in (Appendix \ref{appendix:geno}).

\subsection{Medical Knowledge Question Answer}
\label{subsec: Medical Knowledge Question Answer}
\begin{figure}[H]
    \centering
    \includegraphics[width=0.95\linewidth]{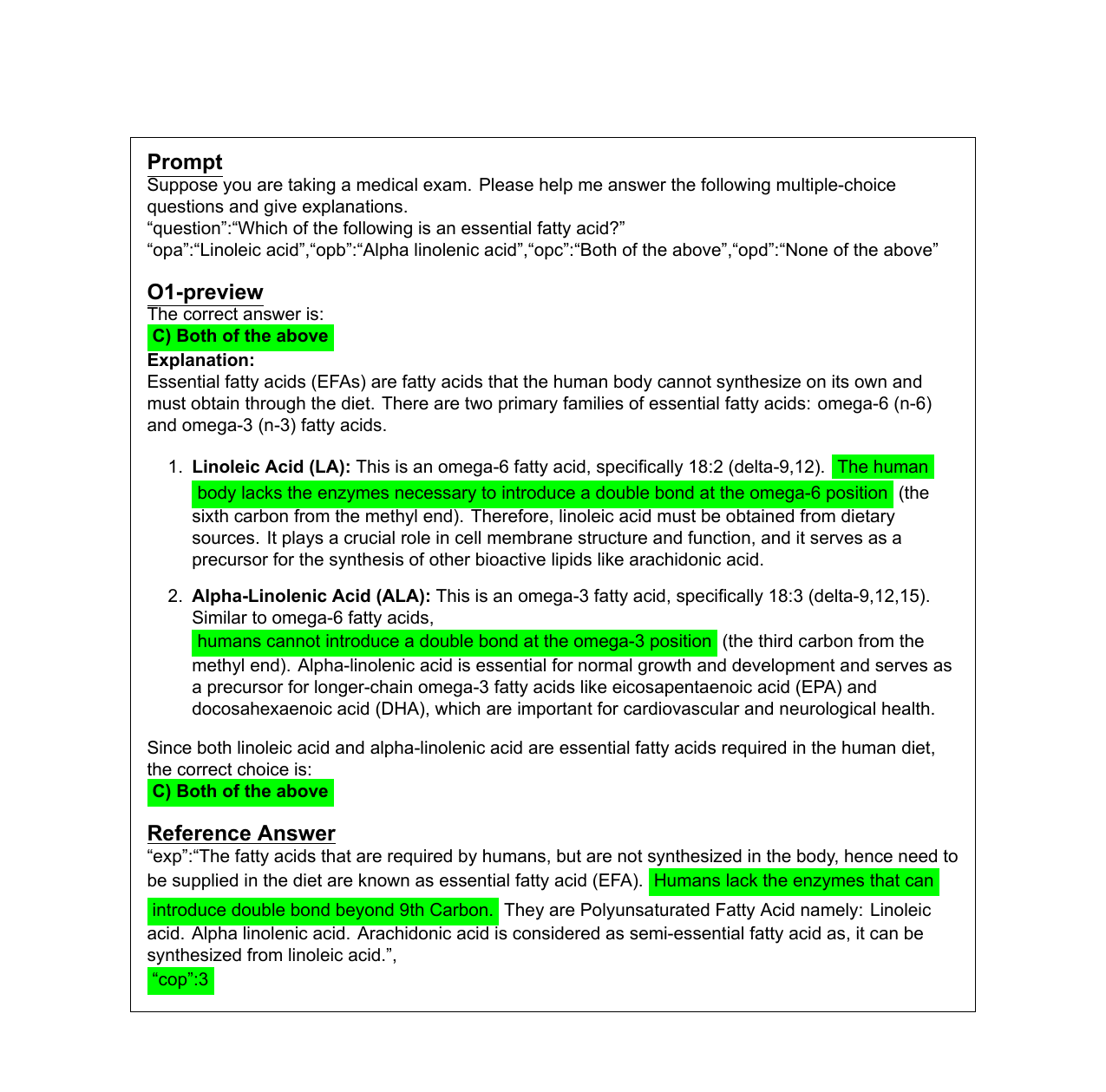}
    \caption[Medical Knowledge Question Answer: Case 1]{\textbf{Medical Knowledge Question Answer: Case 1}. A example illustrating the capability of o1-preview to respond to questions that arises in medical exams. Yellow indicates partial consistency between the o1-preview response and the reference answer, while green signifies complete consistency.}
    \label{fig:mkqa-case1}
\end{figure}

\begin{figure}[H]
    \centering
    \includegraphics[width=0.95\linewidth]{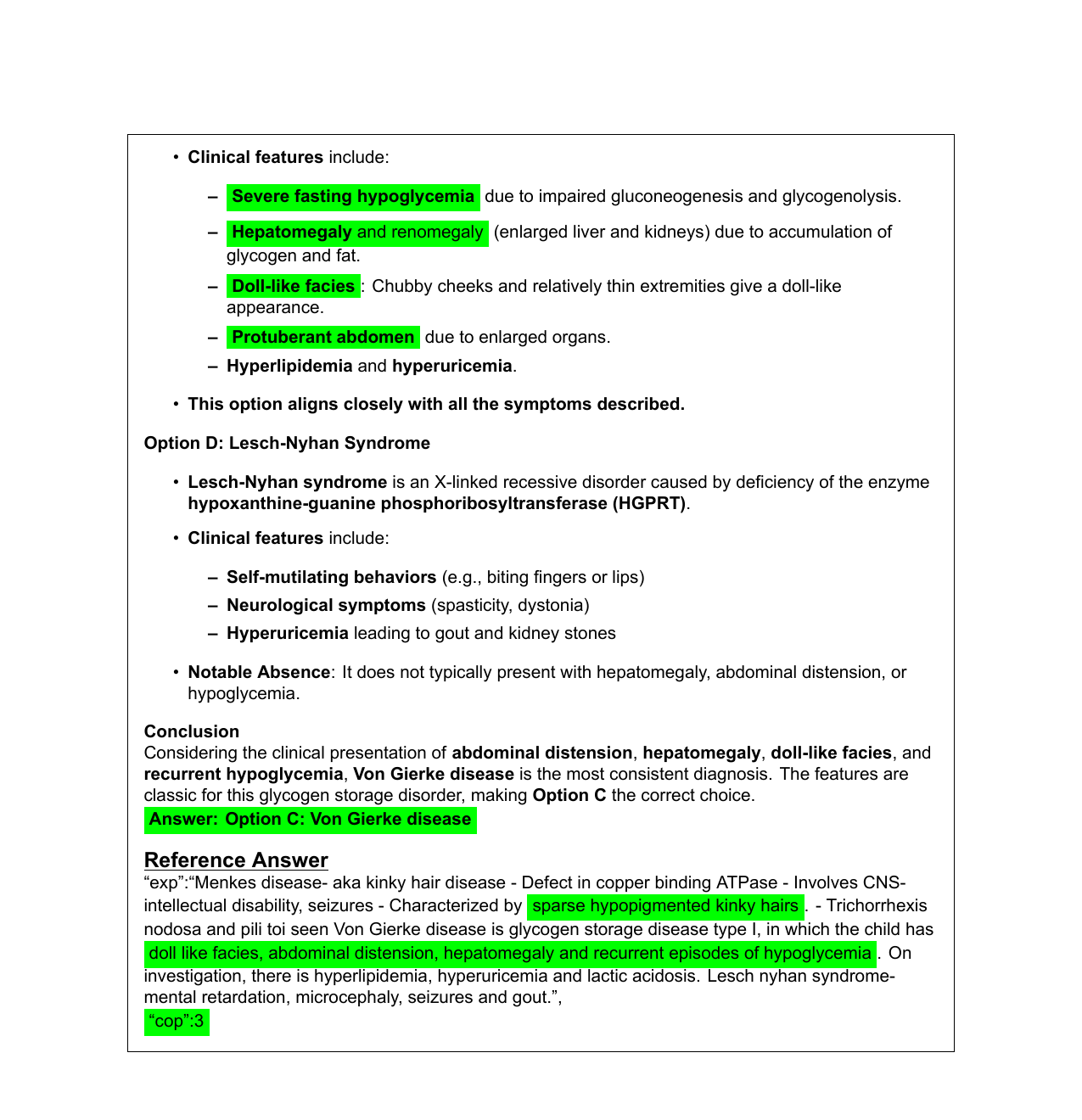}
        \caption[Medical Knowledge Question Answer: Case 2]{\textbf{Medical Knowledge Question Answer: Case 2}. A example illustrating the capability of o1-preview to respond to questions that arises in medical exams. Yellow indicates partial consistency between the o1-preview response and the reference answer, while green signifies complete consistency.}
    \label{fig:mkqa-case2}
\end{figure}

During the evaluation of o1-preview’s performance in Medical Knowledge Question Answer, we found that the model provided accurate and reasoned responses. The test aims to evaluate whether the model can select the correct options for medical knowledge questions and give reasonable explanations.

o1-preview demonstrated alignment with the reference answer. Upon selecting the correct option, it provided a reasonable explanation of the underlying knowledge, suggesting that the model comprehends the question's content and identifies the key elements required to solve the problem. For instance, in Figure~\ref{fig:mkqa-case1}, when confronted with a relatively simple knowledge-based question, the model chose the correct option and provided a coherent explanation of the functions and effects of several fatty acids. Likewise, in Figure~\ref{fig:mkqa-case2} , when addressing a more complex scenario-based inference question, o1-preview selected the correct answer and offered common symptoms and explanations for four diseases, expanding upon the question's scope. The explanation for the correct answer, Von Gierke disease, closely matched the standard answer explanation.

However, the model occasionally overlooks key information when dealing with specific topics, resulting in a lack of necessary logical judgment. Even when choosing the correct option, the model may miss key conditions when evaluating the wrong option. For example, in Figure \ref{fig:mkqa-case9}, it ignores the fact that ``autosomal recessive polycystic kidney disease'' cannot occur in a ``28-year-old male,'' indicating an inability to use logical thinking to process multiple pieces of information in parallel. Despite these minor flaws, o1-preview demonstrates strong reasoning capabilities, especially when answering questions using large medical knowledge datasets.

In summary, o1-preview performs well in medical knowledge question answering. The model effectively captures the key information in the questions and can provide accurate options and effective explanations through retrieval. Although some explanations exhibit illogical reasoning and information gaps, in most cases, the overall reasoning is consistent with the facts. In the current test, the model achieved 100\% accuracy. With further refinements, such as offering more specific examples and conducting deeper analyses, o1-preview shows great potential for medical knowledge question answering tasks.
\begin{figure}[H]
    \centering
    \includegraphics[width=0.95\linewidth]{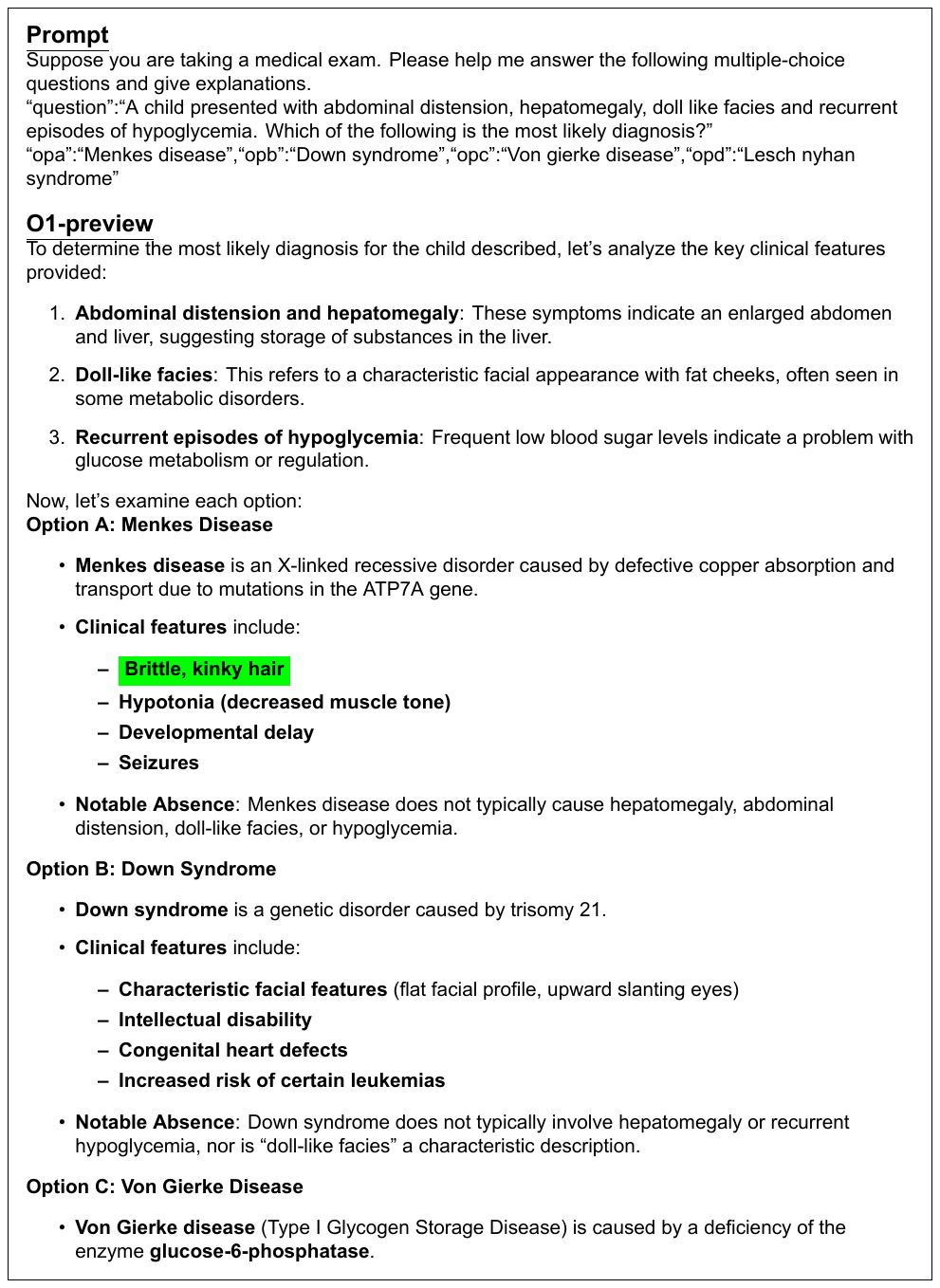}
\end{figure}

\newpage

\subsection{Art Education}

This study evaluates the performance of o1-preview in art education by comparing its responses to those of human experts. Although six diverse tasks were tested, detailed explanations are presented for these two key example tasks—explaining the concept of \textit{currere} in education~\cite{pinar2019currere} and designing a cardboard assemblage art activity for children~\cite{Cardboard}—to assess the model's ability to engage with both theoretical educational frameworks and practical curriculum design. The analysis concentrated on creativity, practical application, and depth of theoretical understanding.

\begin{figure}[H]
    \centering
    \includegraphics[width=0.75\linewidth]{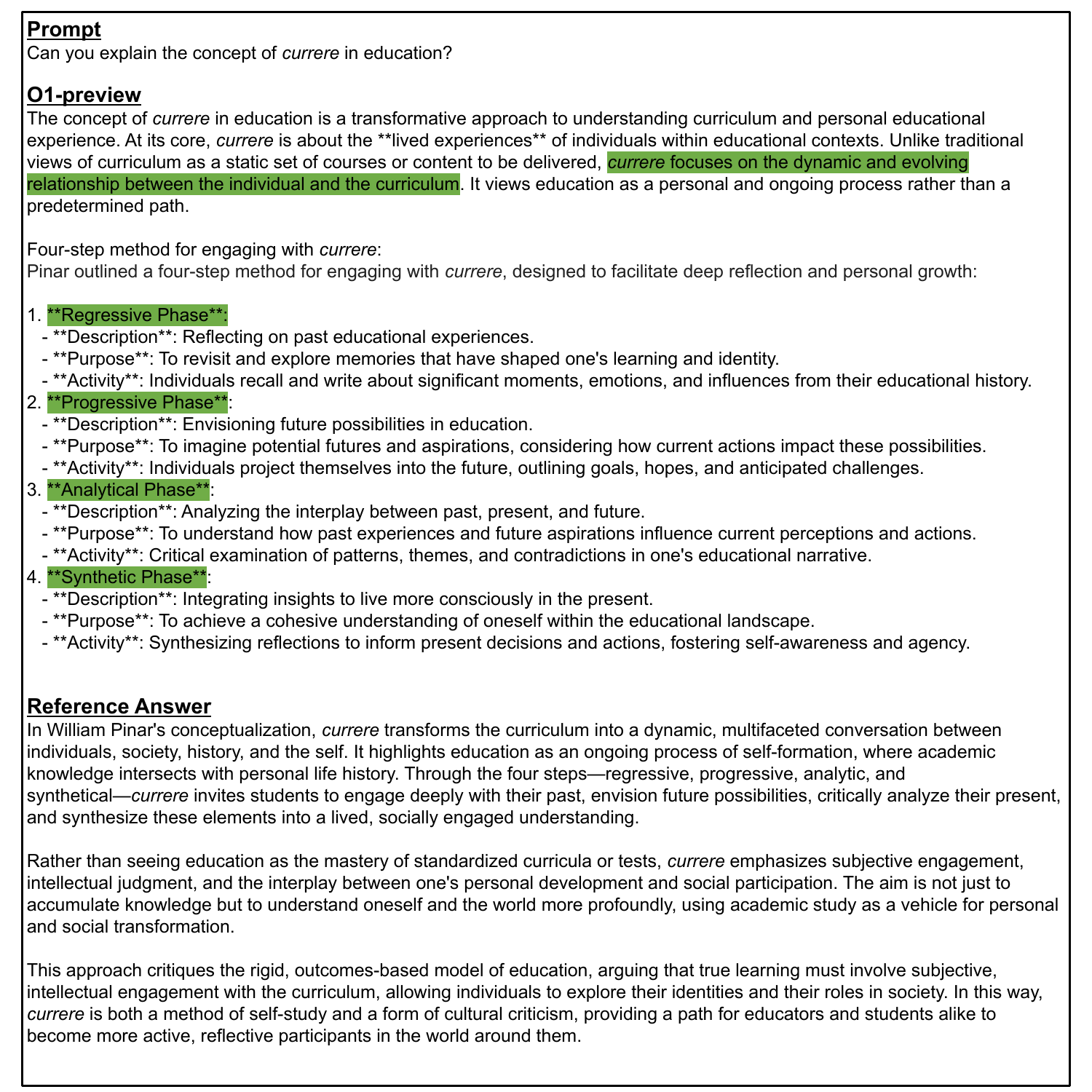}
        \caption[Art Education: Case 1]{\textbf{Art Education: Case 1}. A example illustrating the capability of  o1-preview to explain the concept of \textit{currere} \cite{pinar2019currere}.}
    \label{fig:Art Edu case 1}
\end{figure}

In the task requiring an explanation of \textit{currere} in education, o1-preview provided a well-organized and thoughtful response. It accurately described the four phases of \textit{currere}—regressive, progressive, analytical, and synthetic—and elaborated on how these stages enable individuals to reflect on their educational experiences. The model emphasized the importance of personal reflection, memory, and synthesizing past and future learning experiences, closely aligning with William Pinar’s original work on \textit{currere}~\cite{pinar2019currere}. This explanation showcased o1-preview’s capacity to engage with theoretical concepts and communicate their educational relevance clearly. As illustrated in Figure \ref{fig:Art Edu case 1}.

\begin{figure}[H]
    \centering
    \includegraphics[width=0.75\linewidth]{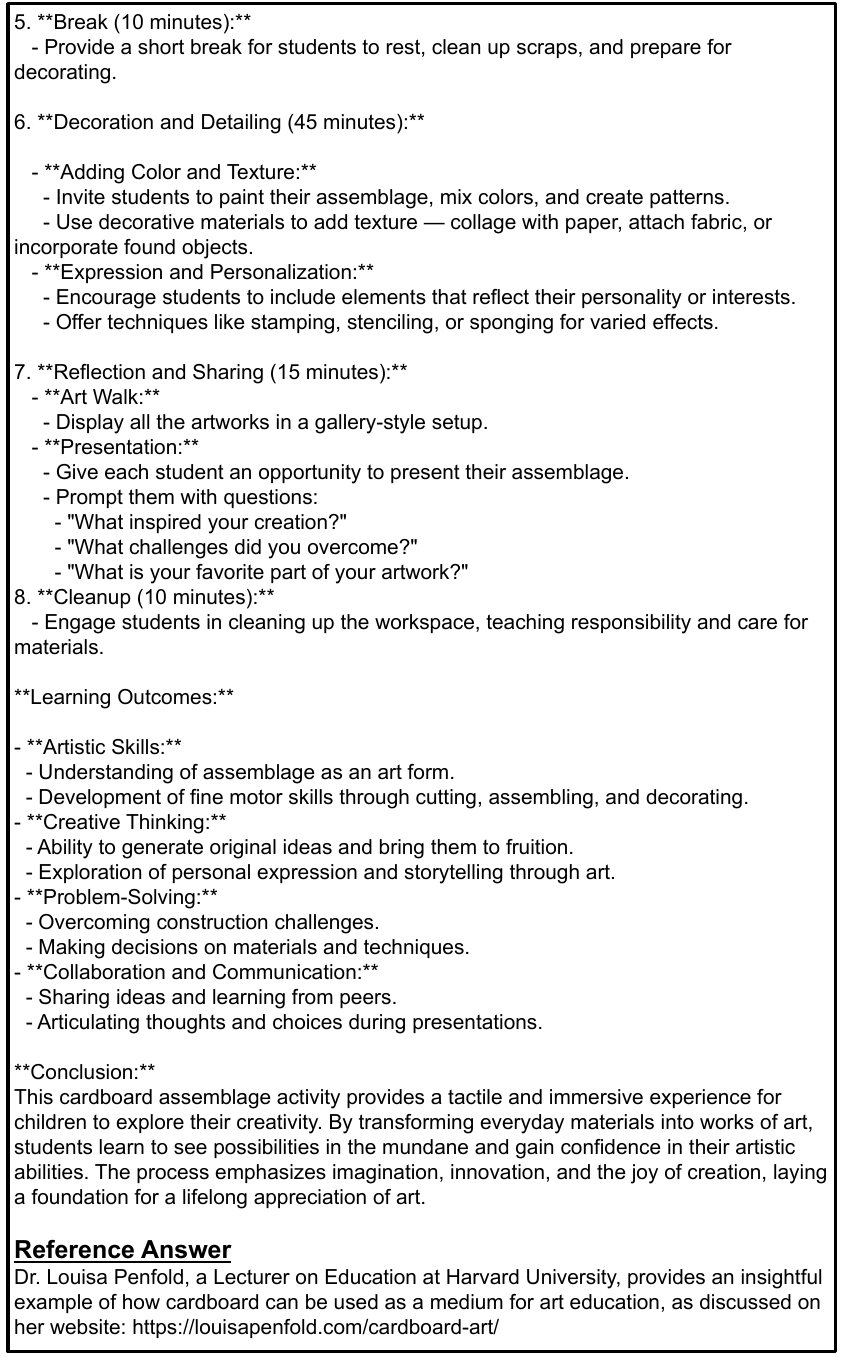}
        \caption[Art Education: Case 2]{\textbf{Art Education: Case 2}. A example illustrating the capability of o1-preview to design creative in-class activities for children.}\cite{Cardboard}
    \label{fig:Art Edu case 2}
\end{figure}

When tasked with designing a cardboard assemblage art activity, o1-preview demonstrated strong capabilities in creating a structured, step-by-step curriculum. It suggested a variety of materials, offered instructions that encouraged creativity and collaboration, and outlined clear learning objectives. However, when compared to Dr. Penfold’s approach—Harvard University's Art Education Coordinator—o1-preview’s lesson appeared more rigid and structured, lacking the flexibility and adaptive scaffolding essential for fostering children's open-ended exploration. Dr. Penfold emphasizes allowing children to freely explore materials, introducing techniques when appropriate to encourage spontaneity and imaginative problem-solving, both of which are essential to child-centered art education~\cite{Cardboard}. The more standardized curriculum model proposed by o1-preview highlights a gap between its procedural creativity and the responsive, individualized approach that human educators like Dr. Penfold implement. As illustrated in Figure \ref{fig:Art Edu case 2}.

In conclusion, while o1-preview excels in both practical curriculum planning and theoretical explanation, it lacks the depth of emotional insight and contextual sensitivity that human educators bring to art education. The model's versatility in handling both practical and theoretical aspects is evident, yet it falls short in adapting to children's varying needs in real-time and fostering the kind of open-ended exploration advocated by educators like Dr. Penfold. This highlights a key difference between AI models and human educators— o1-preview can be a valuable tool for generating ideas and structuring content, but human educators are essential for creating flexible, responsive learning environments that nurture children's creativity in a more personal and adaptive manner. 
\newpage

\newpage


\begin{figure}[H]
    \centering
    \includegraphics[width=0.75\linewidth]{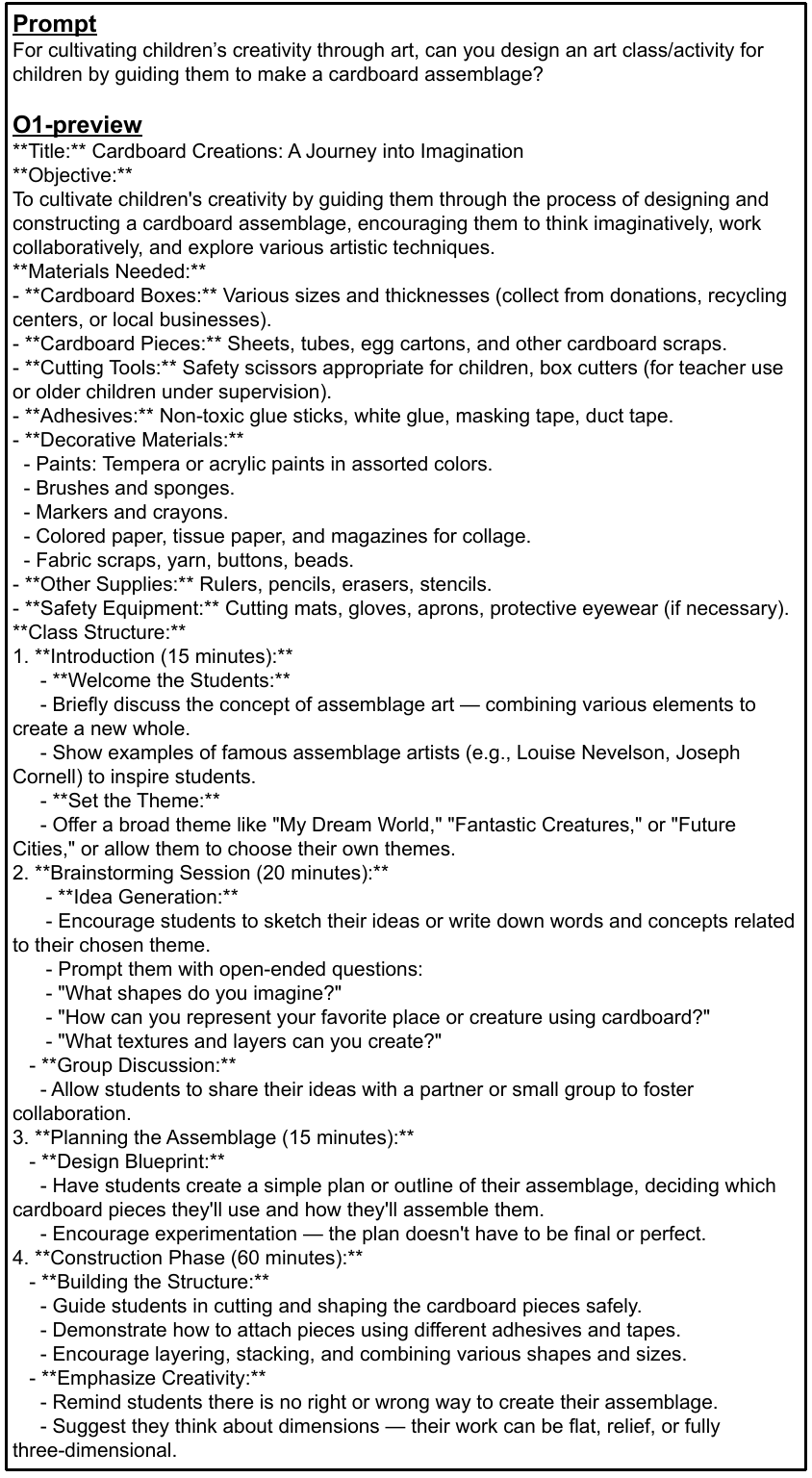}
\end{figure}

\newpage


\subsection{Content Summarization}

For the content summarization task, we designed the same prompt for all the news articles to evaluate the performance of the o1-preview. The prompt, “Please summarize the content in one sentence: ….”, highlights the length of the summary and limits the results to a sentence. This test aims to evaluate whether the o1-preview model can summarize the textual content with the ability of context relevance, text content understanding and text generation. We compare the summary of the o1-preview with the manual summary for the corresponding article from three perspectives: accuracy, degree of abstraction and length.

\begin{figure}[H]
    \centering
    \includegraphics[width=0.9\linewidth]{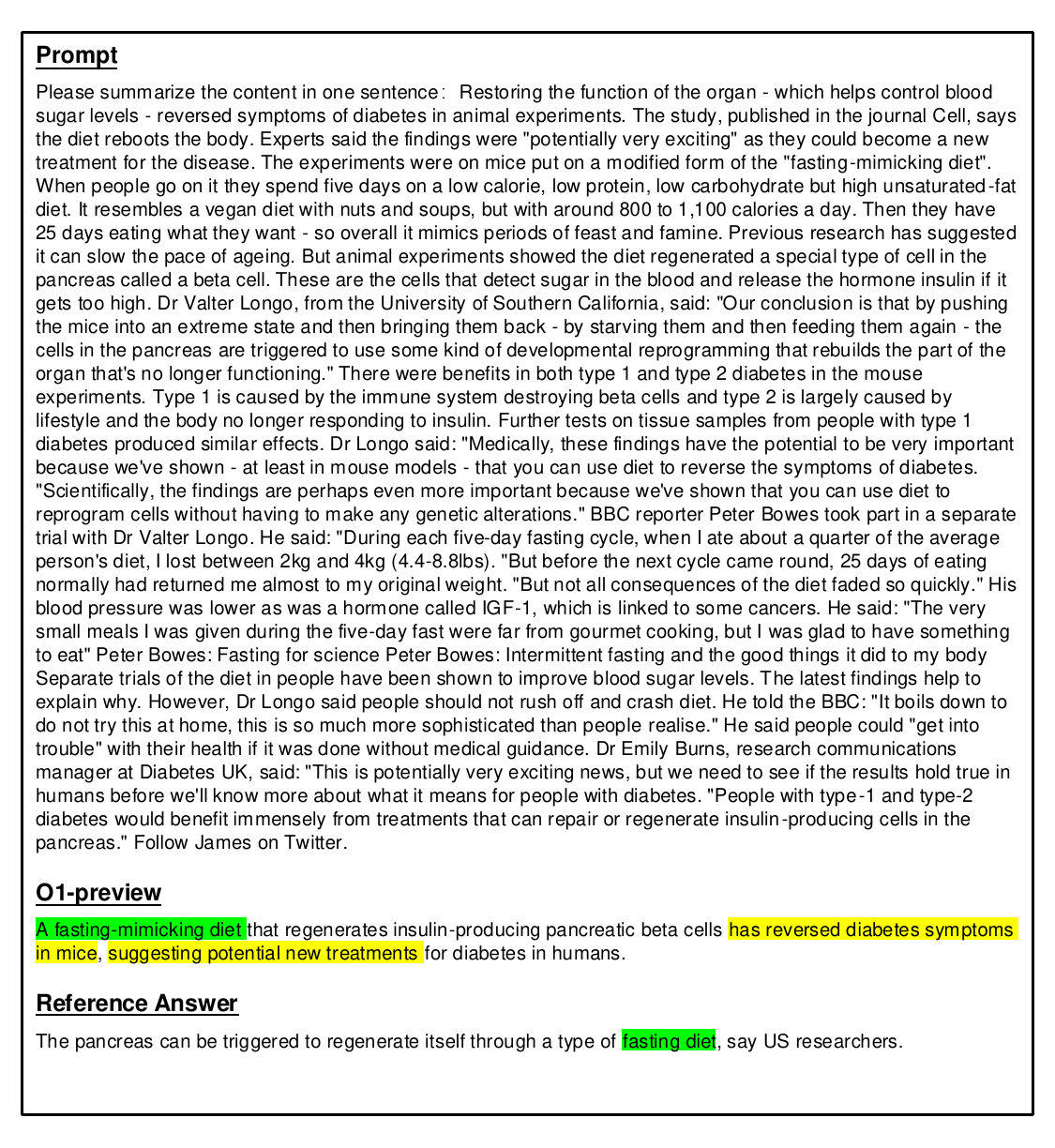}
    \caption[Content Summarization: Case 1.]{{\textbf{Content Summarization: Case 1.}} A example illustrating the capability of o1-preview to respond to summarize news content. Yellow indicates partial consistency between the o1-preview's response and the reference answer, while green signifies complete consistency.}
    \label{fig:summarization-case1}
\end{figure}

Summaries generated by o1-preview show high similarity with a reference summary and original content in some samples, although they have different expressions with the same meaning. For instance, in Figure \ref{fig:summarization-case1}, the summary of both human beings and o1-preview has the same theme, which is “a type of fasting diet can regenerate pancreas cells”.

\begin{figure}[H]
    \centering
    \includegraphics[width=0.9\linewidth]{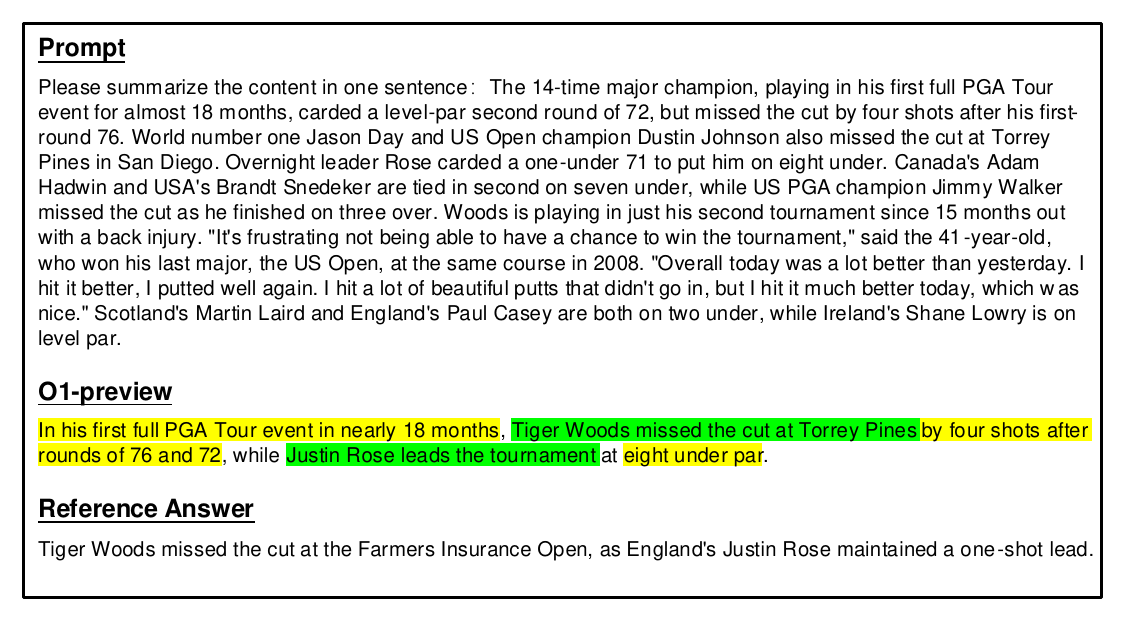}
    \caption[Content Summarization: Case 2.]{\textbf{Content Summarization: Case 2.} A example illustrating the capability of o1-preview to respond to summarize news content. Yellow indicates partial consistency between the o1-preview's response and the reference answer, while green signifies complete consistency.}
    \label{fig:summarization-case2}
\end{figure}

However, the o1-preview model tends to generate more detail, compared with manually summarizing. Human beings seem to be better at and more accustomed to a higher level of abstraction of text. For instance, in Figure \ref{fig:summarization-case2}, although both of the two summaries describe “Tiger Woods missed the cut at the Farmers Insurance Open”. Human beings tend to show the reason in a processed expression, which does not directly show the scores of the competition but shows a calculated and more abstractive result, while the o1-preview model lists several details about the competition. This kind of difference is shown not only in a single sample, this is common sense. In Figure \ref{fig:summarization-case3}, o1-preview model generated too much details so that the summary is not as easy to read as manual summary.

As for the length, summaries of o1-preview are longer than those of human beings in most samples. This is because summaries of o1-preview contain more details than manual summaries.

In summary, the o1-preview model is well performed in content summarization tasks. It can meet the basic needs of most summarization tasks, but compared with human summaries, it usually contains more details and is slightly inferior to humans in the ability to abstract information.
\newpage

\newpage

\newpage
\section{Discussion}\label{sec:discussion}
Our comprehensive evaluation of o1-preview reveals both impressive capabilities and notable limitations across various domains. In coding tasks, the model demonstrated strong performance in problem-solving and algorithm implementation, while struggling with optimizing solutions for complex problems under time constraints. This suggests a need for improved efficiency in the model's reasoning process for time-sensitive applications.

For mathematical reasoning, o1-preview excelled in solving high school-level problems but showed inconsistent performance in more advanced college-level mathematics. This indicates a potential gap in the model's abilities to understand logical principles and handle long reasoning processes, highlighting an area for future development.

The model's performance in scientific reasoning tasks was generally strong, particularly in integrating knowledge across different scientific disciplines. At times, it provided overly verbose explanations or failed to prioritize the most relevant information, suggesting a need for improved conciseness and information filtering mechanisms.

In language-related tasks, o1-preview demonstrated impressive multilingual capabilities and strong performance in natural language inference. Its performance in low-resource language translation revealed limitations in handling languages with limited training data. Future work should focus on improving the model's ability to generalize linguistic patterns across diverse language families.

For medical and healthcare-related tasks, the model showed promise in generating coherent medical reports and analyzing electronic health records. Occasional production of inaccurate or inconsistent diagnoses underscores the critical need for further refinement and extensive validation before any potential clinical application.

In creative and analytical tasks such as geology reasoning and financial analysis, o1-preview demonstrated a remarkable ability to understand complex requirements and generate detailed outputs. The quality and practicality of these outputs varied, suggesting a need for better grounding in real-world constraints and domain-specific knowledge.

Future research directions should focus on:
\begin{itemize}
\item Enhancing the model's efficiency in complex problem-solving under time constraints.
\item Improving performance in advanced mathematical reasoning and abstract concept manipulation.
\item Developing better mechanisms for concise and relevant information extraction and presentation.
\item Expanding capabilities in low-resource languages and improving cross-lingual generalization.
\item Refining domain-specific knowledge integration, particularly in specialized fields such as medicine, engineering and other sciences that require domain expertise~\cite{liu2023summary,lee2023multimodality,tian2024assessing,liu2023radonc,holmes2024benchmarking,tang2023policygpt,guan2023cohortgpt,rezayi2024exploring,zhang2024generalist,gong2024advancing,liu2024surviving,liu2024fine,xu2024reasoning,shu2024llms,zhao2024revolutionizing,wei2023chat2brain,rezayi2022clinicalradiobert,holmes2023benchmarking,shi2023mededit,dou2023towards,liu2023radiology,cai2023exploring}.
\item Improving the model's ability to handle multi-step reasoning tasks consistently across various domains, including sciences, engineering, medicine, public health, pharmacy, arts, humanities, education, agriculture, business, law, politics, and social studies. These domains can benefit from models that can reason well~\cite{liu2023summary,zhao2023brain,ma2024iterative,dai2023chataug,liu2023radiology,liu2023context,zhang2023biomedgpt,dai2023ad,holmes2023evaluating,liao2024mask,liao2023differentiating,wang2023prompt,liu2023holistic,rezayi2022agribert,liu2024llm,liu2023pharmacygpt,wang2024comprehensive,liu2023transformation,liu2023evaluating,li2023artificial}.
\end{itemize}

\subsection{LLM Agents and o1: Advancing Problem-Solving in Science and Engineering}

Large language model agents are designed to handle complex tasks requiring sequential reasoning, planning, and memory. These agents typically consist of four key components: a core language model (the "brain"), a planning mechanism, memory systems, and the ability to use external tools~\cite{wang2024survey}. By integrating these elements, LLM agents can break down complex problems, formulate strategies, and adapt their approaches based on feedback and new information.

o1 potentially changes the application potential of LLM agents. The combination of LLM agents with o1-preview's advanced capabilities presents exciting possibilities for tackling intricate science and engineering challenges. o1-preview has already demonstrated remarkable proficiency across various domains, from coding and mathematical reasoning to scientific analysis and creative problem-solving. When enhanced with the structured approach of LLM agents, o1 could potentially address even more complex, multi-step problems in scientific research and engineering design.

For instance, in scientific research, an o1-powered LLM agent could systematically approach experimental design by breaking down the process into subtasks such as literature review, hypothesis formation, methodology planning, and data analysis strategy. The agent's planning capabilities could help in outlining long-term research projects, while its memory systems could maintain coherence across extended investigations, recalling relevant past experiments or related findings.
In engineering, such an agent could assist in complex system design by managing interconnected components, predicting potential issues, and suggesting optimizations. The tool use capability of LLM agents could allow o1 to interface with specialized engineering software, running simulations or accessing specific databases as needed during the design process.

Moreover, the self-reflection and improvement capabilities of LLM agents could enhance o1's already impressive reasoning skills, allowing for continuous refinement of solutions in iterative design processes or long-term research projects. This could be particularly valuable in fields like materials science or drug discovery, where vast parameter spaces need to be explored systematically.

However, it's important to note that challenges remain. The current limitations of LLM agents, such as context constraints and difficulties with long-term planning, would need to be addressed. Additionally, ensuring the reliability and consistency of outputs in critical scientific and engineering applications would be paramount.

\newpage
\section{Conclusion}\label{sec:conclusion}
This study provides a comprehensive evaluation of OpenAI's o1-preview model across a diverse range of complex reasoning tasks. Our findings demonstrate that o1-preview represents a significant advancement in artificial intelligence, showing impressive capabilities in areas such as coding, scientific reasoning, language processing, and creative problem-solving. The model's ability to integrate knowledge across different domains and apply it to novel problems is particularly noteworthy, suggesting progress towards more general artificial intelligence.

Despite these advancements, our study also uncovers important limitations and areas for improvement, particularly in advanced mathematical reasoning, handling of specialized domain knowledge, and consistent performance across all task types. 

In conclusion, o1-preview is an unprecedented model that surpasses past landmarks in machine intelligence. This comprehensive study not only confirms its capabilities but also identifies critical pathways for future improvements toward AGI.

\newpage
\appendix
\section{Appendix}
\subsection{Code Generation}
\label{appendix:coding}
\begin{figure}[H]
    \centering
    \includegraphics[width = \textwidth]{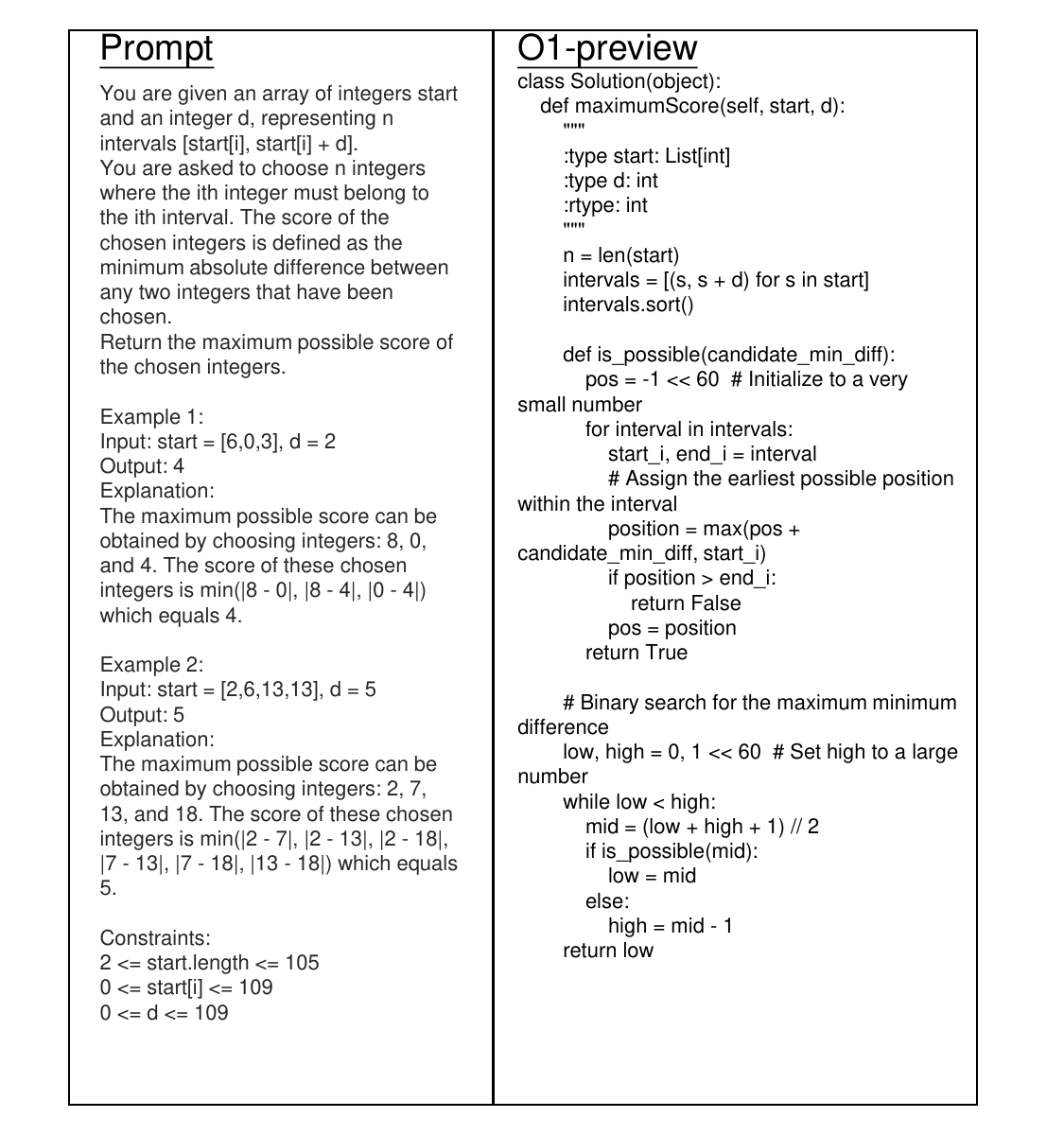}
    \caption[Code Generation: Case 3]{\textbf{Coding: Case 3.} A demonstration case from coding testing of a medium level problem on o1-preview solution. The generated code passes the Leetcode automated judgement system for the first attempt.}
    \label{fig:coding_3}
\end{figure}
\newpage

\begin{figure}[H]
    \centering
    \includegraphics[width = \textwidth]{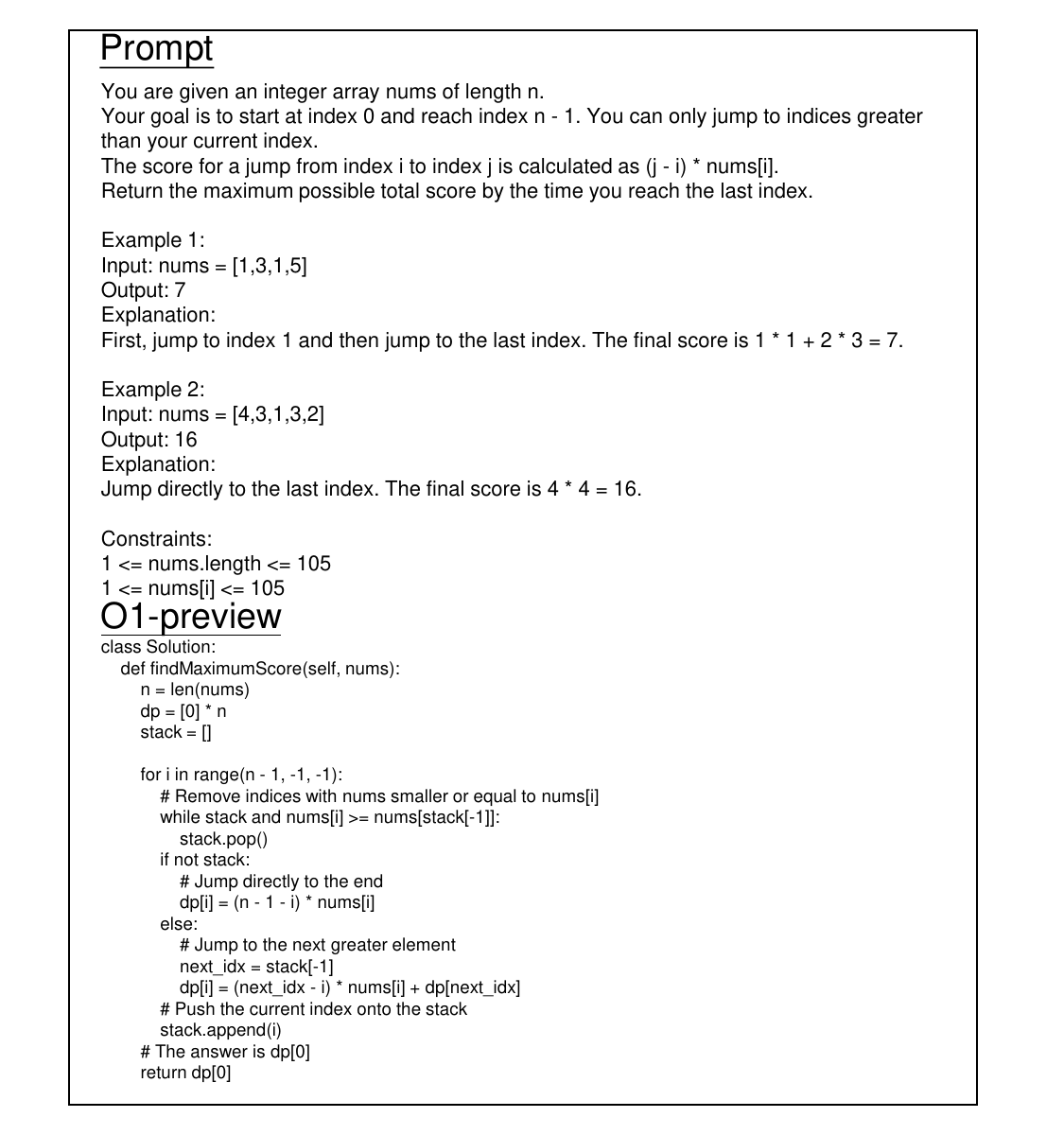}
    \caption[Code Generation: Case 4]{\textbf{Coding: Case 4.} An additional case from coding testing of a medium level problem on o1-preview solution. The generated code passes the Leetcode automated judgement system for the first attempt.}
    \label{fig:coding_4}
\end{figure}
\newpage

\begin{figure}[H]
    \centering
    \includegraphics[width = \textwidth]{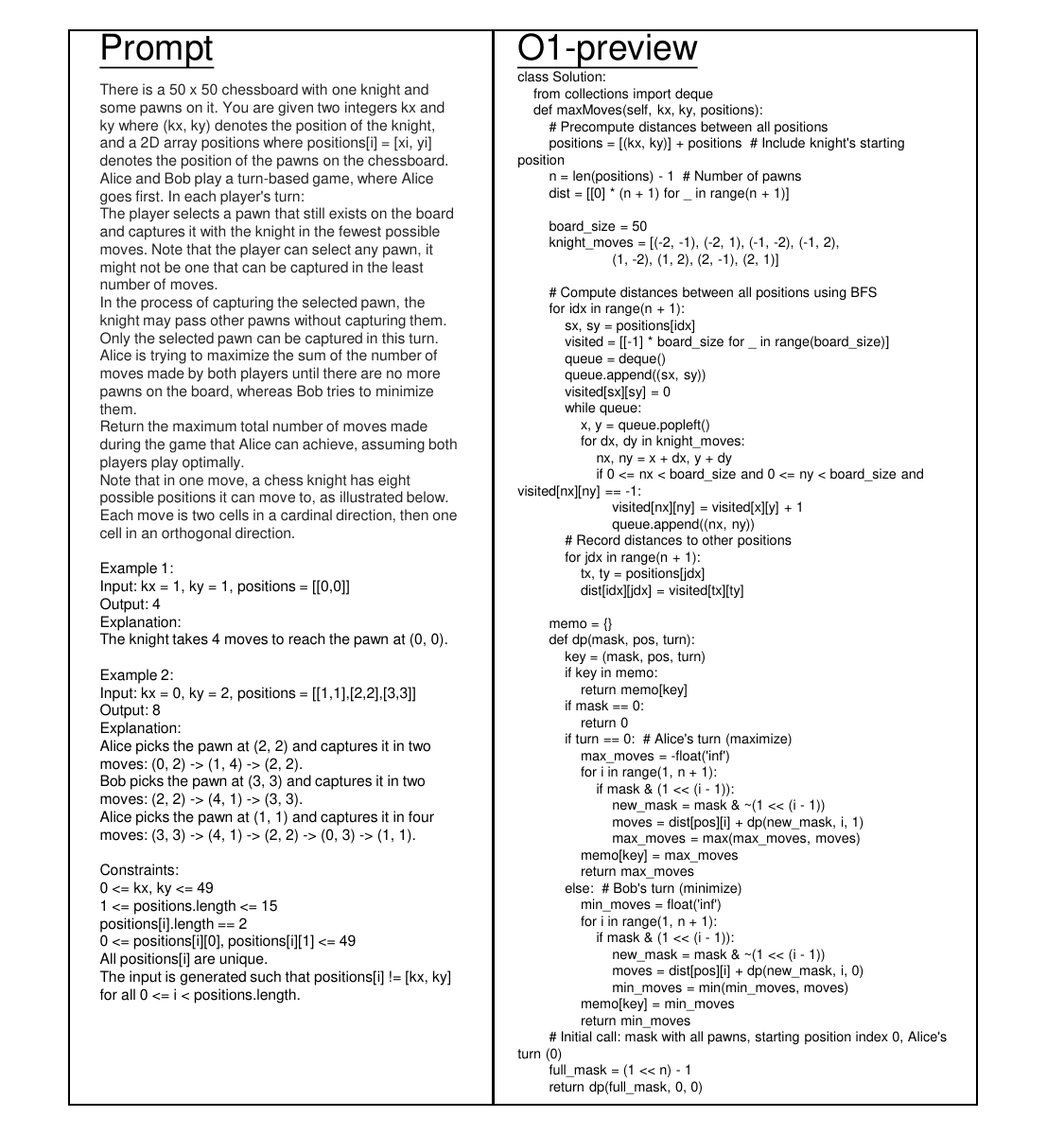}
    \caption[Code Generation: Case 5]{\textbf{Coding: Case 5.} A demonstration case from coding testing of a hard level problem on o1-preview solution. The generated code passes the Leetcode automated judgement system for the first attempt.}
    \label{fig:coding_5}
\end{figure}
\newpage

\begin{figure}[H]
    \centering
    \includegraphics[width = \textwidth]{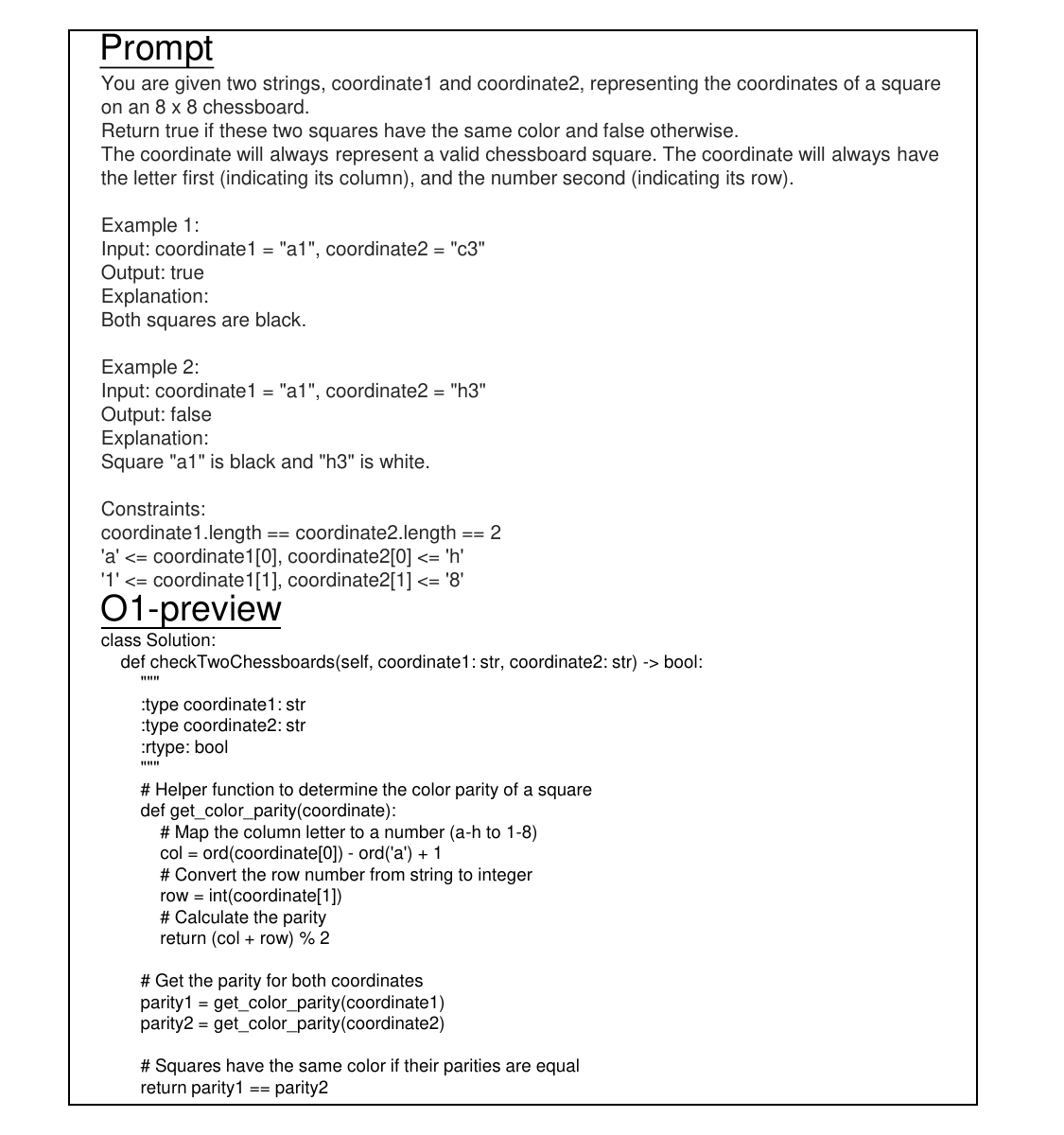}
    \caption[Code Generation: Case 6]{\textbf{Coding: Case 6.} A demonstration case from coding testing of a easy level problem on o1-preview solution. The generated code passes the Leetcode automated judgement system for the first attempt.}
    \label{fig:coding_6}
\end{figure}
\newpage

\begin{figure}[H]
    \centering
    \includegraphics[width = \textwidth]{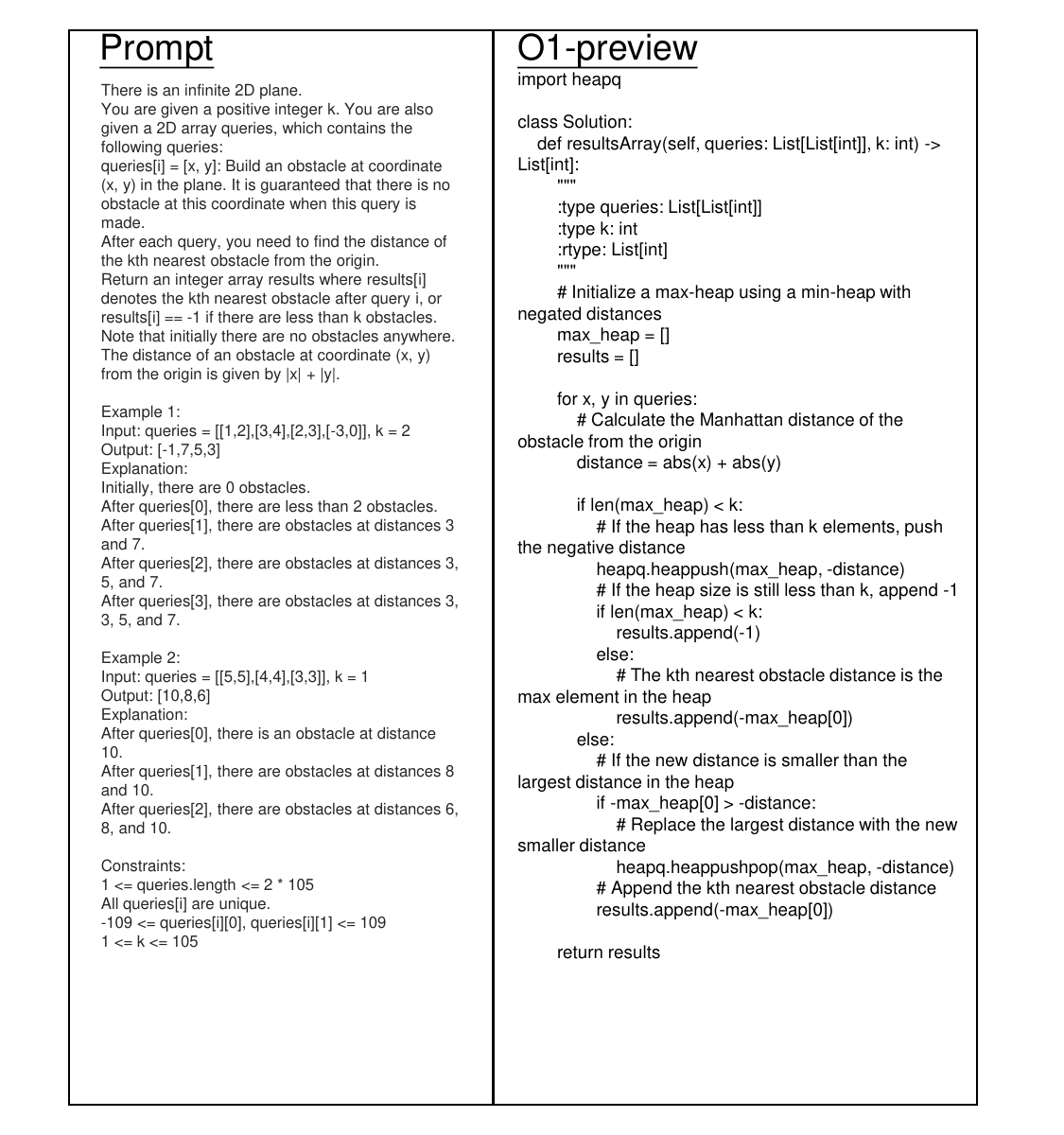}
    \caption[Code Generation: Case 7]{\textbf{Coding: Case 7.} A demonstration case from coding testing of a medium level problem on o1-preview solution. The generated code passes the Leetcode automated judgement system for the first attempt.}
    \label{fig:coding_7}
\end{figure}
\newpage

\begin{figure}[H]
    \centering
    \includegraphics[width = \textwidth]{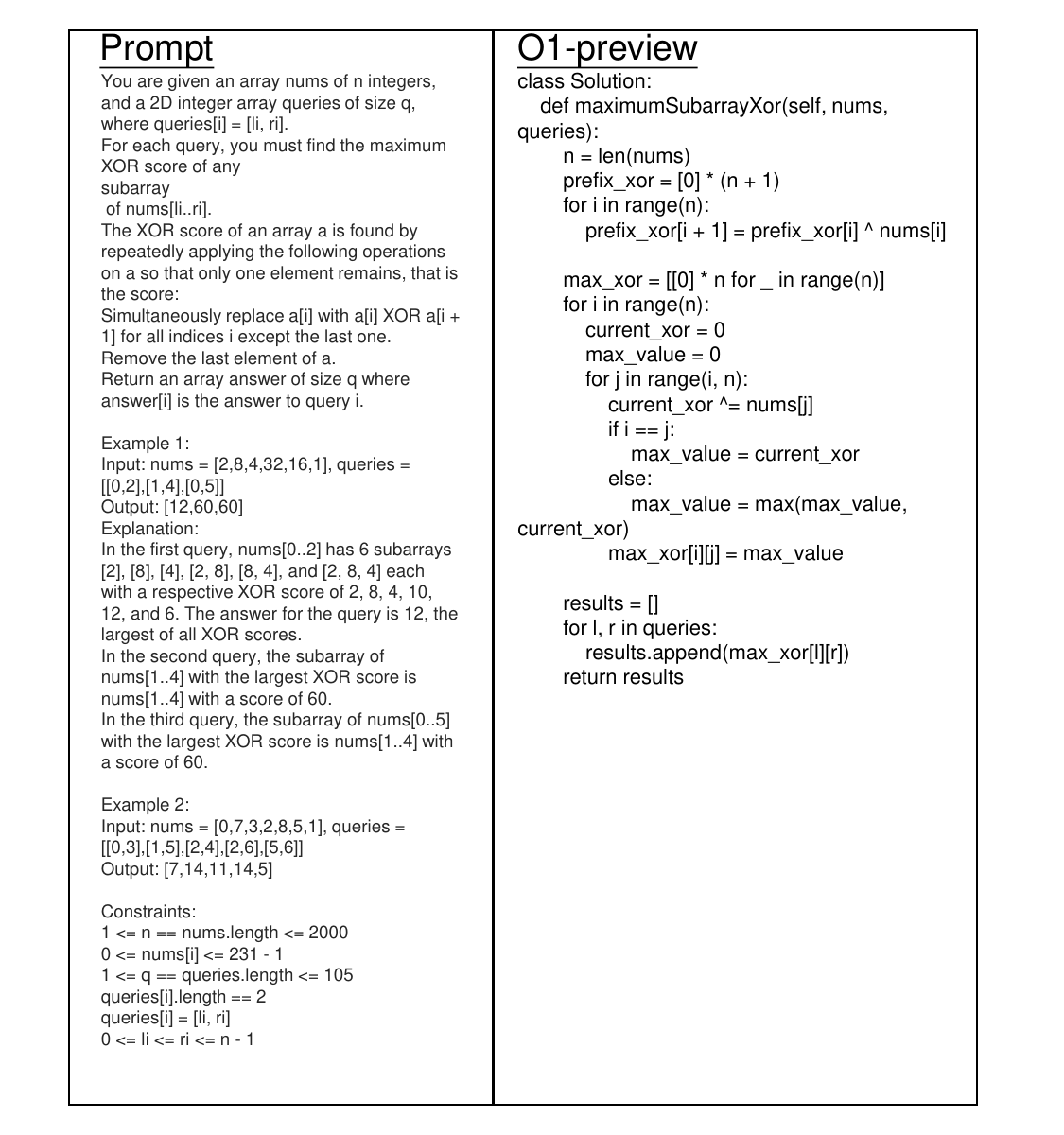}
    \caption[Code Generation: Case 8]{\textbf{Coding: Case 8.} A demonstration case from coding testing of a hard level problem on o1-preview solution. The generated code passes the Leetcode automated judgement system for the first attempt.}
    \label{fig:coding_8}
\end{figure}
\newpage

\begin{figure}[H]
    \centering
    \includegraphics[width = \textwidth]{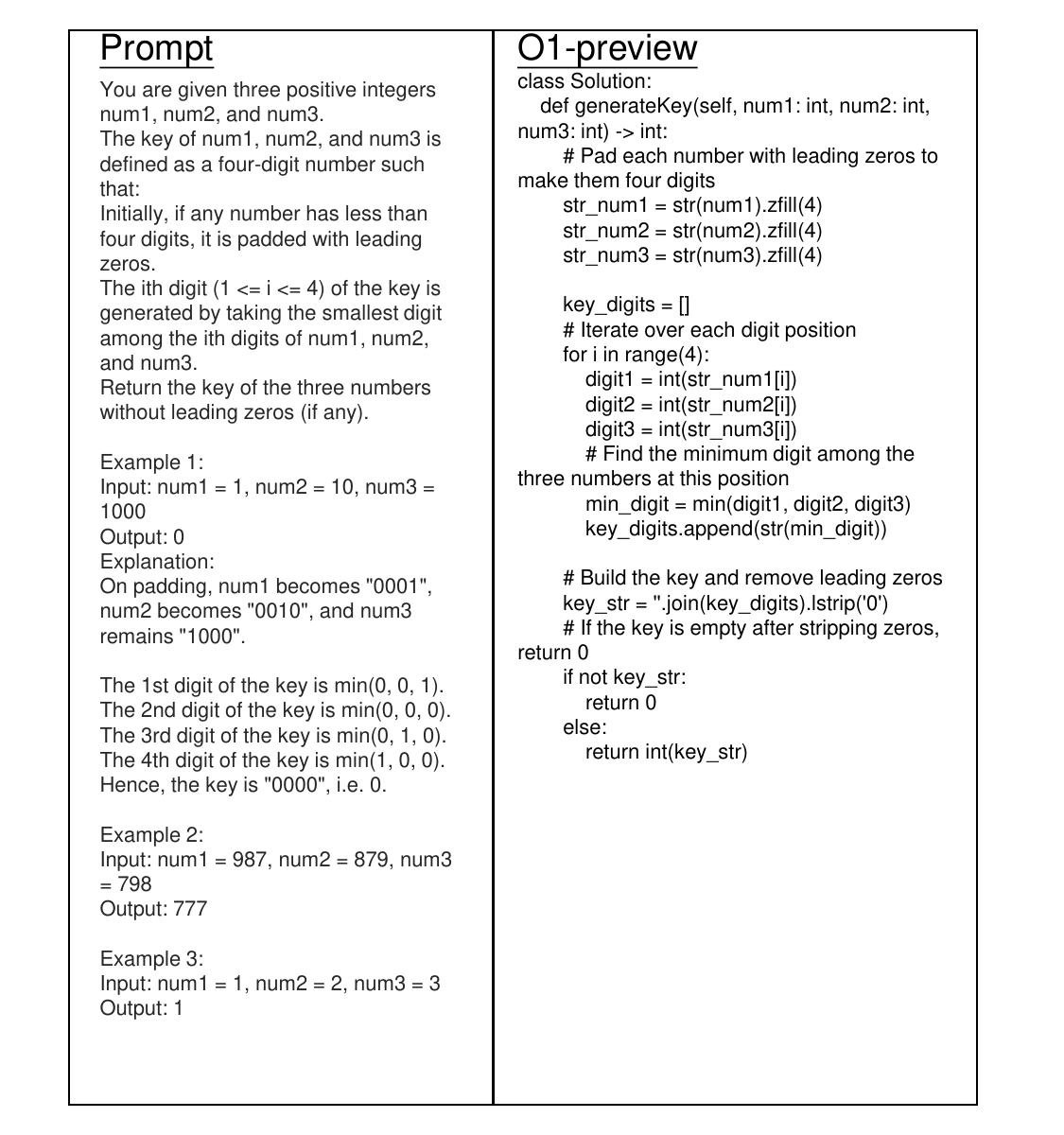}
    \caption[Code Generation: Case 9]{\textbf{Coding: Case 9.} A demonstration case from coding testing of a easy level problem on o1-preview solution. The generated code passes the Leetcode automated judgement system for the first attempt.}
    \label{fig:coding_9}
\end{figure}
\newpage

\begin{figure}[H]
    \centering
    \includegraphics[width = \textwidth]{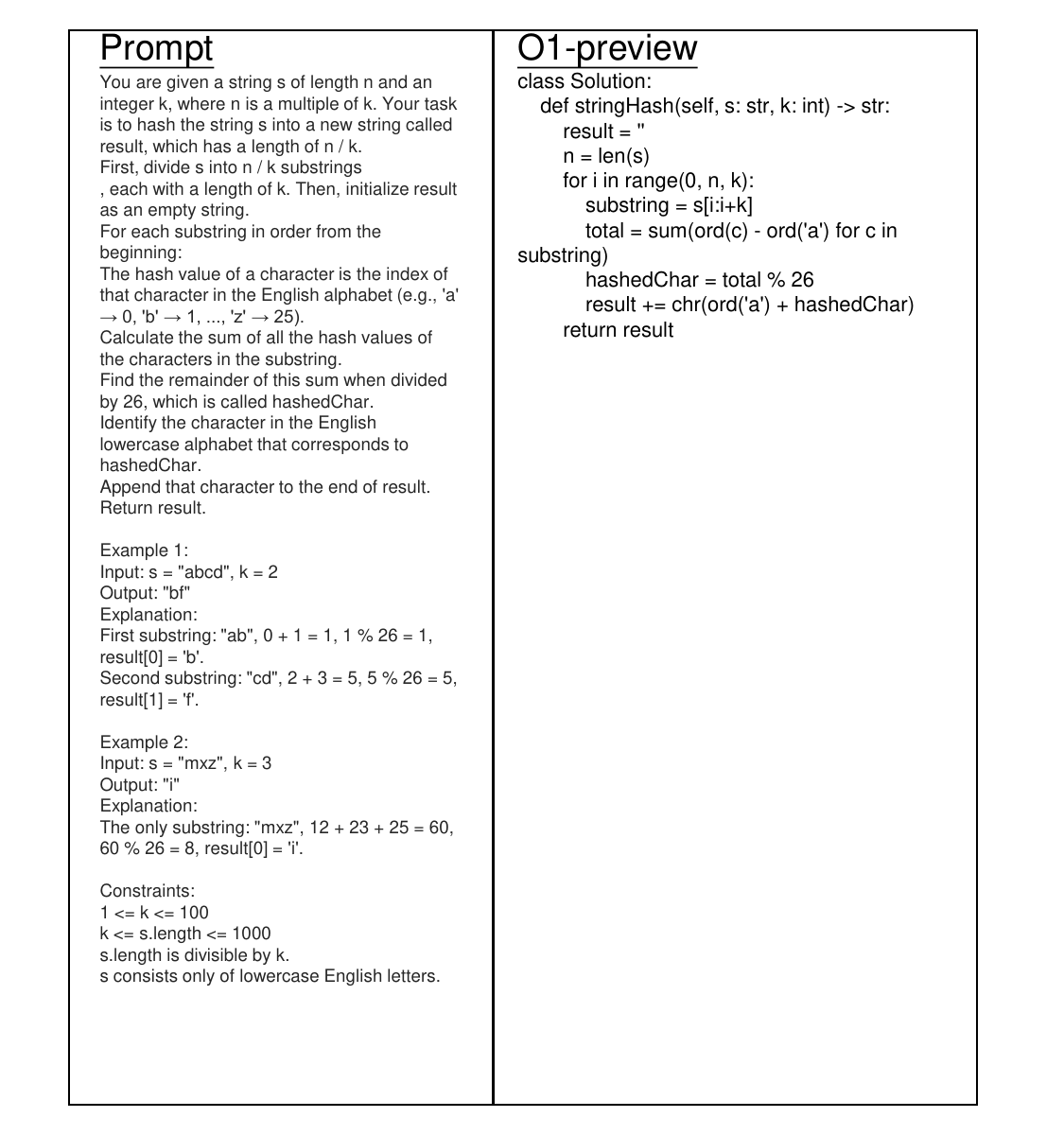}
    \caption[Code Generation: Case 10]{\textbf{Coding: Case 10.} A demonstration case from coding testing of a medium level problem on o1-preview solution. The generated code passes the Leetcode automated judgement system for the first attempt.}
    \label{fig:coding_10}
\end{figure}
\newpage

\begin{figure}[H]
    \centering
    \includegraphics[width = \textwidth]{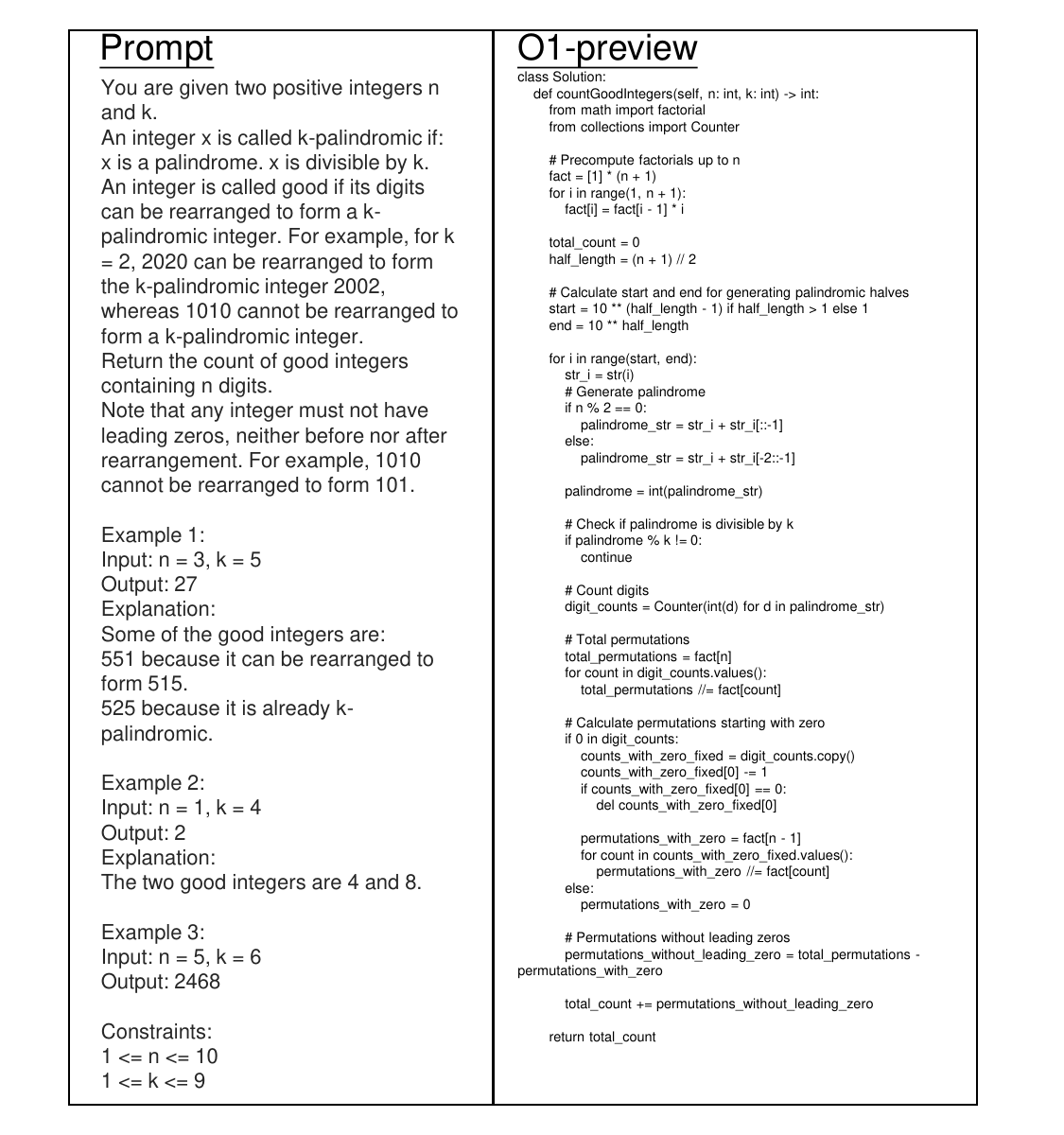}
    \caption[Code Generation: Case 11]{\textbf{Coding: Case 11.} A demonstration case from coding testing of a hard level problem on o1-preview solution. o1-preview fails to pass this test problem. All of three attempts generate wrong answer.}
    \label{fig:coding_11}
\end{figure}
\newpage

\begin{figure}[H]
    \centering
    \includegraphics[width = \textwidth]{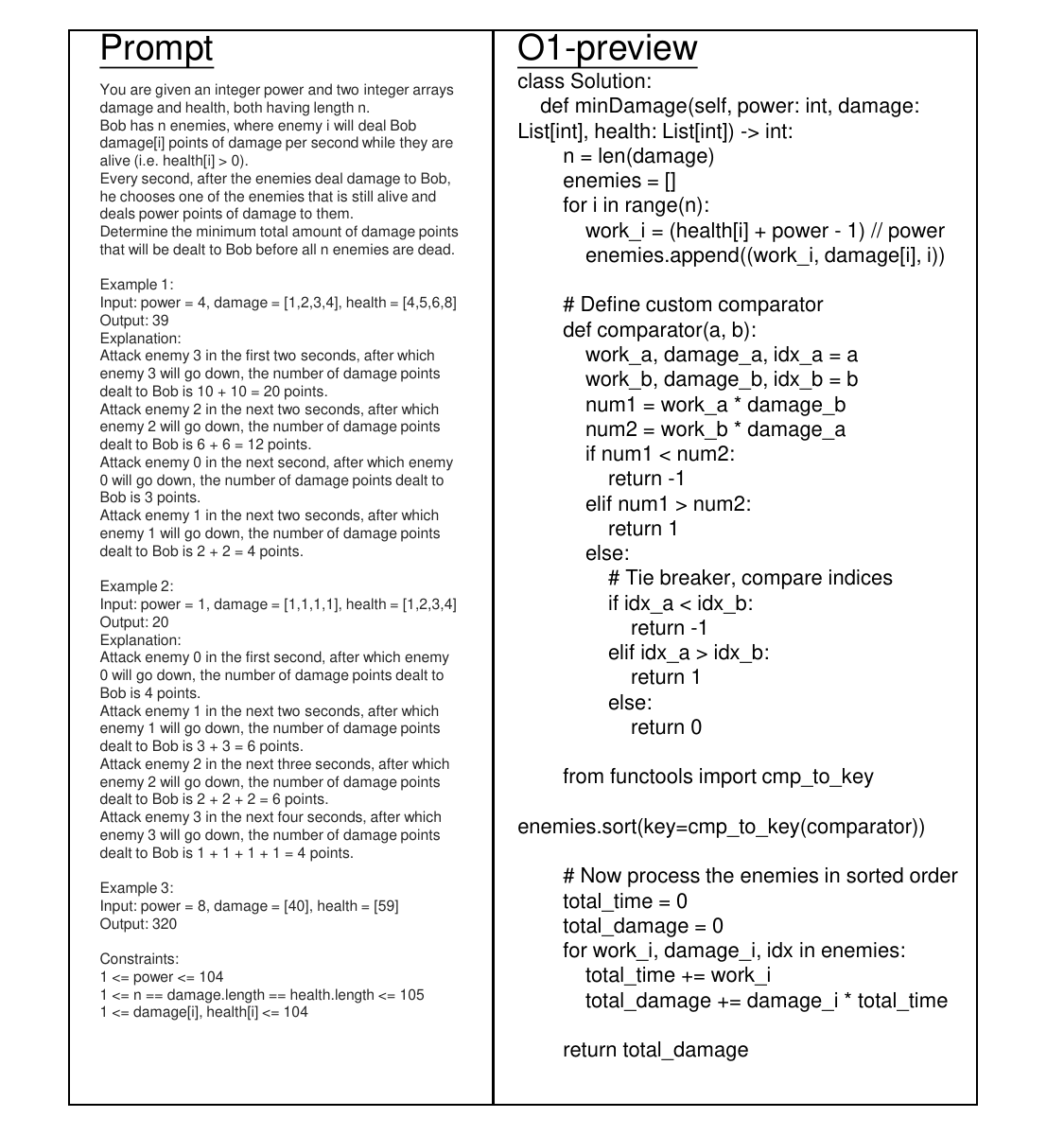}
    \caption[Code Generation: Case 12]{\textbf{Coding: Case 12.} A demonstration case from coding testing of a hard level problem and o1-preview solution. The generated code passes the Leetcode automated judgement system for the first attempt.}
    \label{fig:coding_12}
\end{figure}
\newpage


\subsection{Radiology Report Generation}
\label{appendix:Radiology}
\begin{figure}[H]
    \centering
    \includegraphics[width = \textwidth]{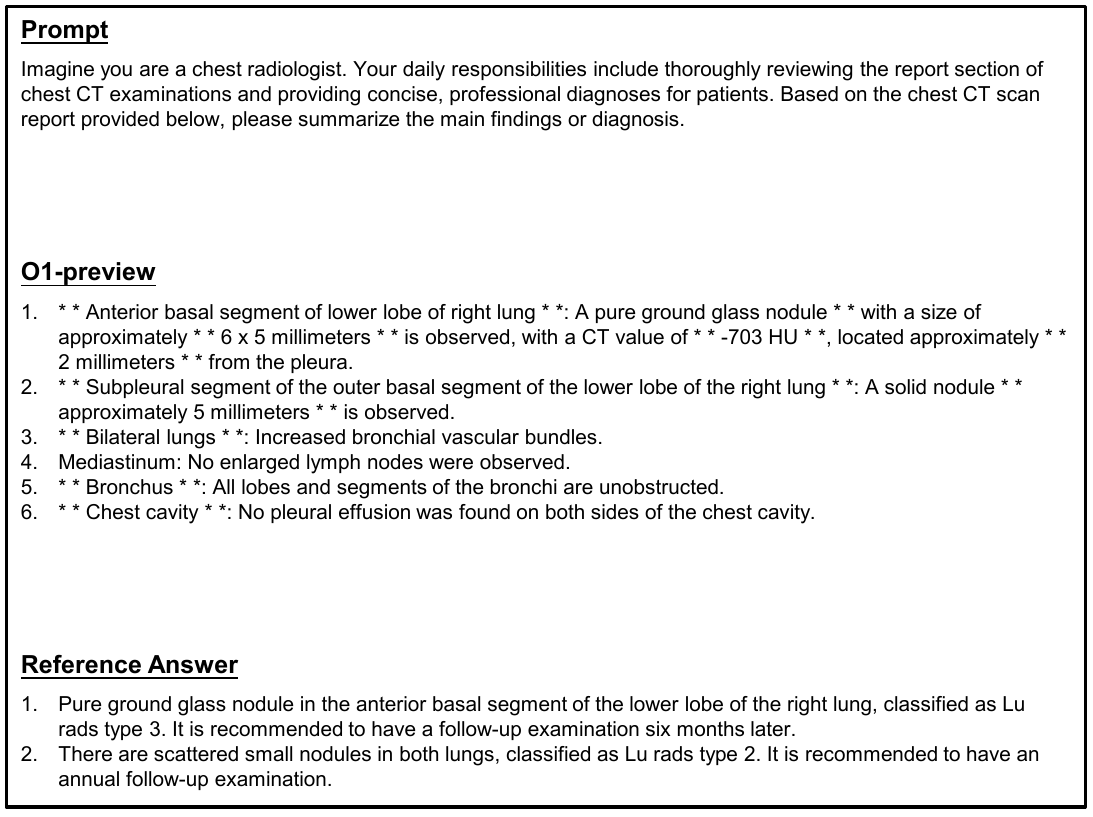}
    \caption[Radiology Report Generation: Case 3]{\textbf{Radiology: Case 3.} We observed that the o1-preview closely aligns with human writing patterns, exhibiting clear organization and concise language. To enhance reader accessibility, the radiological report generated in the zero-shot o1-preview task has been translated from Chinese to English.}
    \label{fig:Radiology_3}
\end{figure}
\newpage

\begin{figure}[H]
    \centering
    \includegraphics[width = \textwidth]{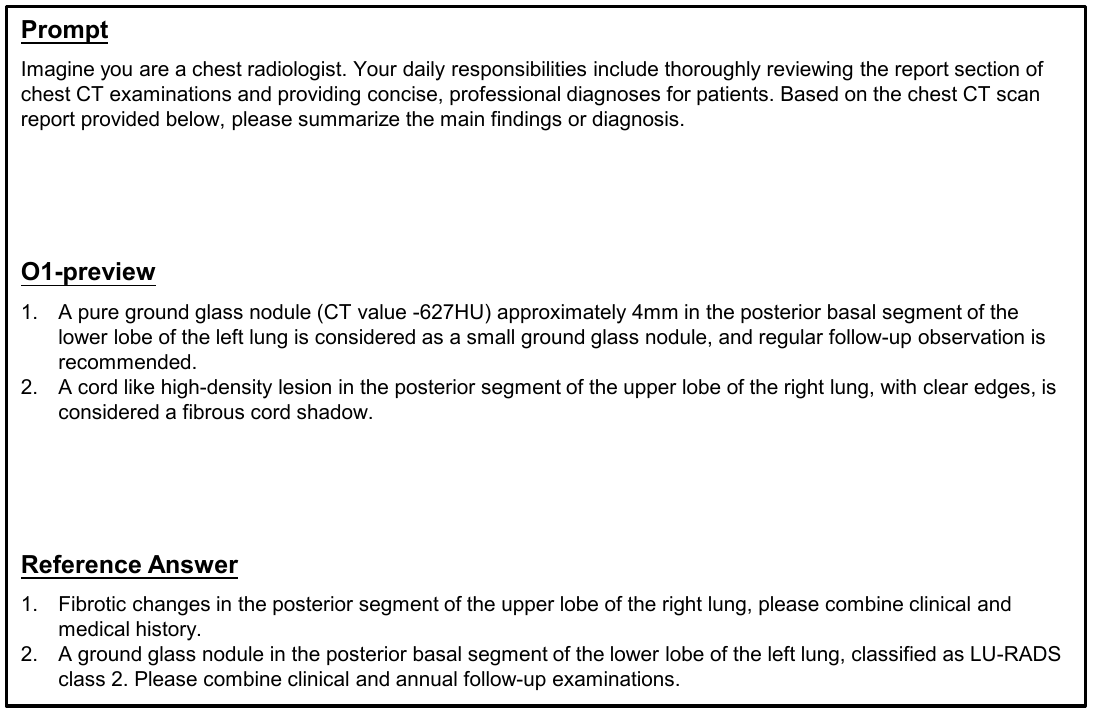}
    \caption[Radiology Report Generation: Case 4]{\textbf{Radiology: Case 4.} We observed that the o1-preview closely aligns with human writing patterns, exhibiting clear organization and concise language. To enhance reader accessibility, the radiological report generated in the zero-shot o1-preview task has been translated from Chinese to English.}
    \label{fig:Radiology_4}
\end{figure}
\newpage

\begin{figure}[H]
    \centering
    \includegraphics[width = \textwidth]{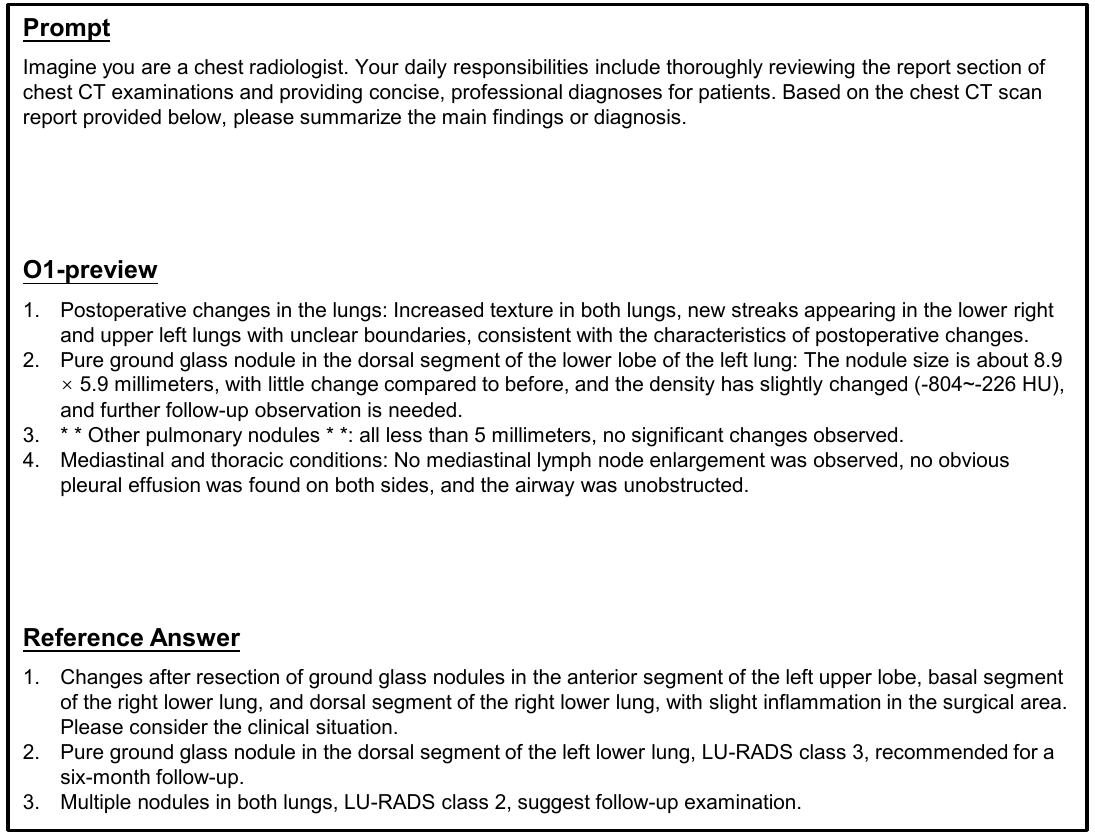}
    \caption[Radiology Report Generation: Case 5]{\textbf{Radiology: Case 5.} We observed that the o1-preview closely aligns with human writing patterns, exhibiting clear organization and concise language. To enhance reader accessibility, the radiological report generated in the zero-shot o1-preview task has been translated from Chinese to English.}
    \label{fig:Radiology_5}
\end{figure}
\newpage

\begin{figure}[H]
    \centering
    \includegraphics[width = \textwidth]{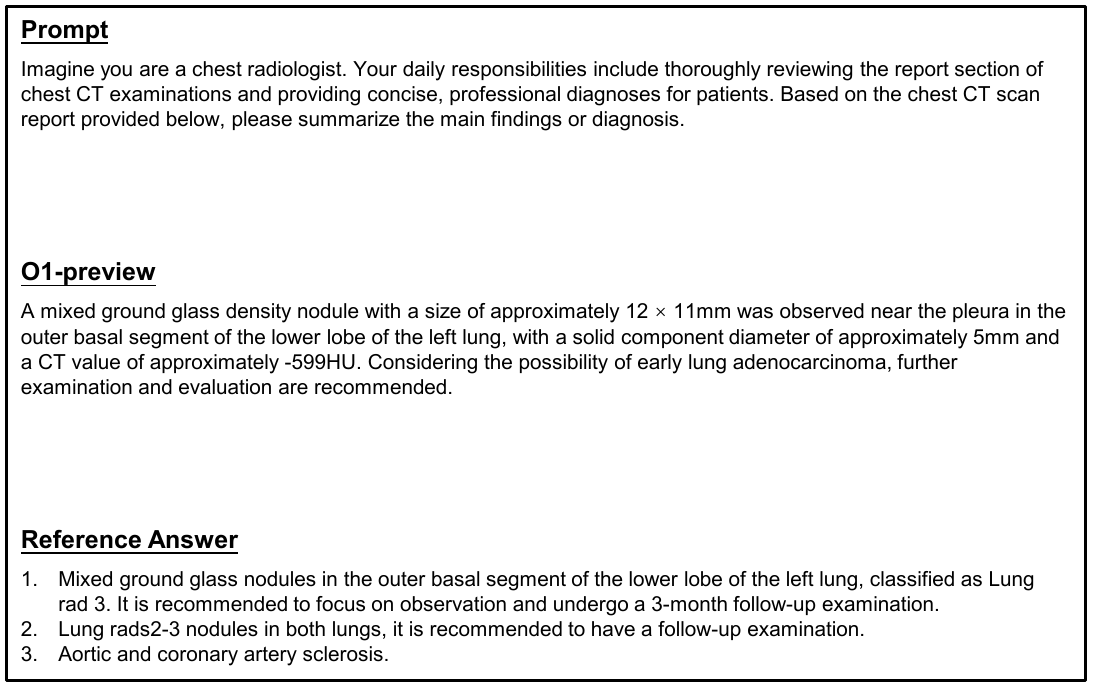}
    \caption[Radiology Report Generation: Case 6]{\textbf{Radiology: Case 6.} We observed that the o1-preview closely aligns with human writing patterns, exhibiting clear organization and concise language. To enhance reader accessibility, the radiological report generated in the zero-shot o1-preview task has been translated from Chinese to English.}
    \label{fig:Radiology_6}
\end{figure}
\newpage

\begin{figure}[H]
    \centering
    \includegraphics[width = \textwidth]{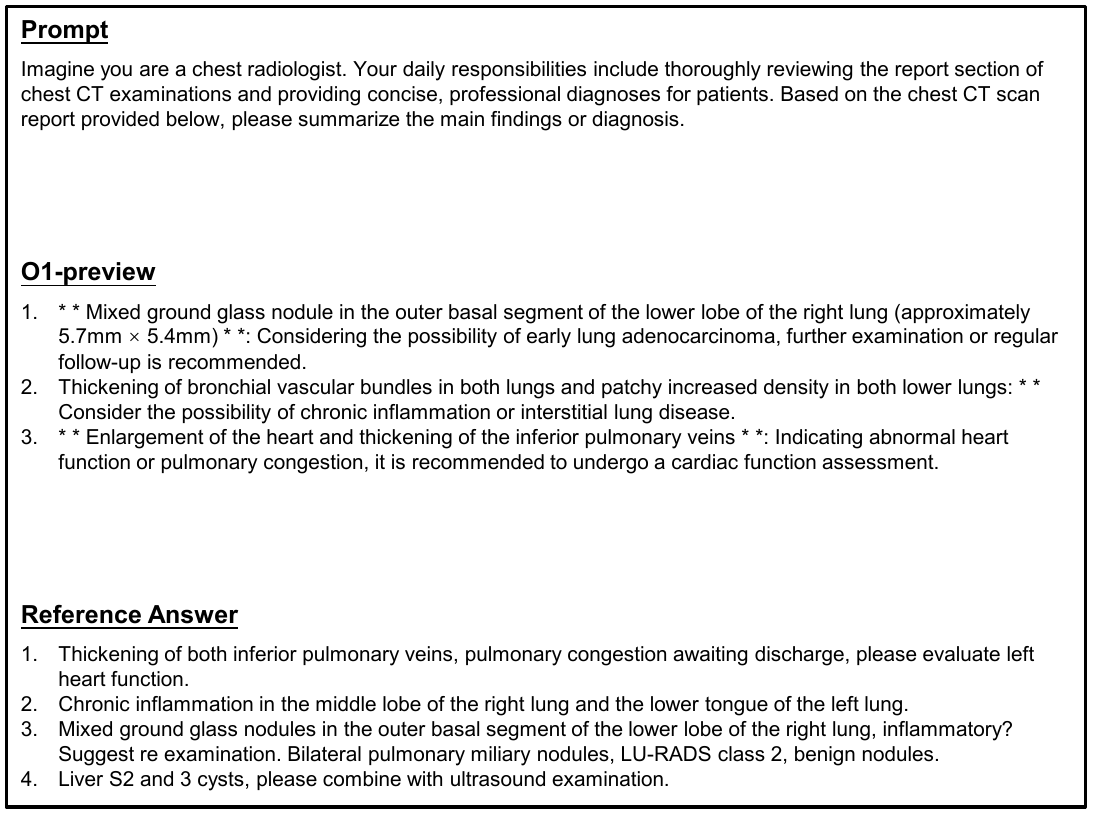}
    \caption[Radiology Report Generation: Case 7]{\textbf{Radiology: Case 7.} We observed that the o1-preview closely aligns with human writing patterns, exhibiting clear organization and concise language. To enhance reader accessibility, the radiological report generated in the zero-shot o1-preview task has been translated from Chinese to English.}
    \label{fig:Radiology_7}
\end{figure}
\newpage

\begin{figure}[H]
    \centering
    \includegraphics[width = \textwidth]{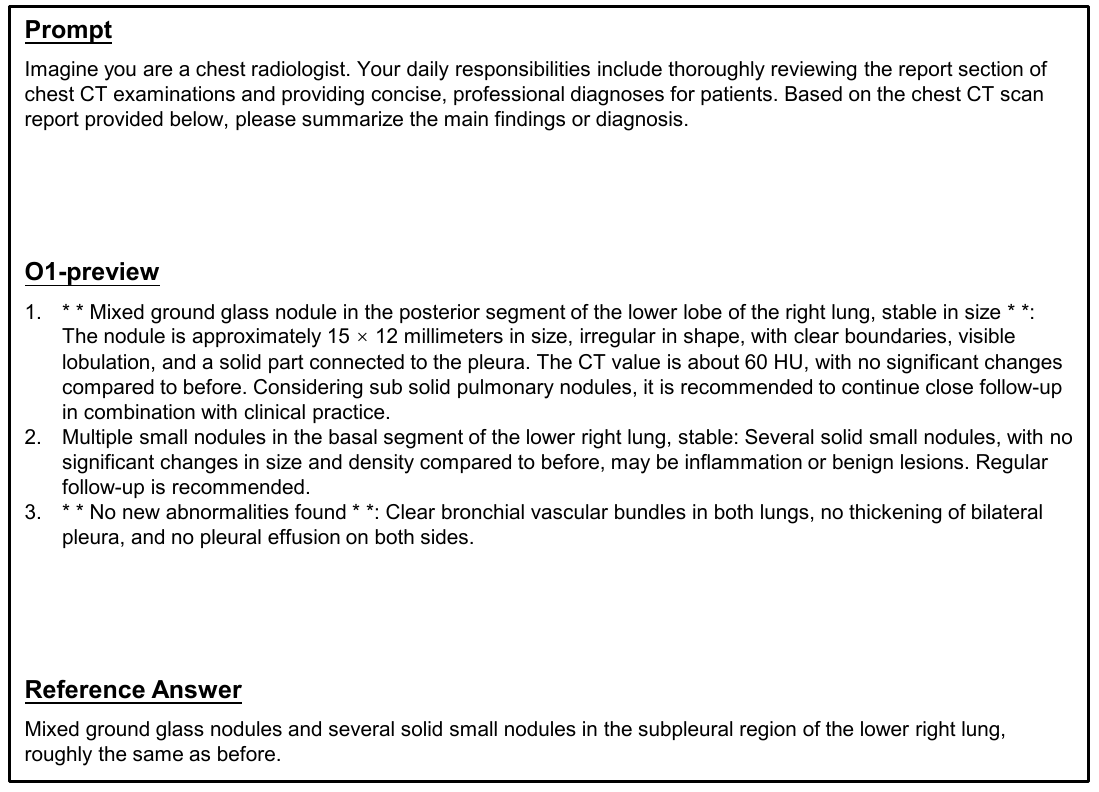}
    \caption[Radiology Report Generation: Case 8]{\textbf{Radiology: Case 8.} We observed that the o1-preview closely aligns with human writing patterns, exhibiting clear organization and concise language. To enhance reader accessibility, the radiological report generated in the zero-shot o1-preview task has been translated from Chinese to English.}
    \label{fig:Radiology_8}
\end{figure}
\newpage

\begin{figure}[H]
    \centering
    \includegraphics[width = \textwidth]{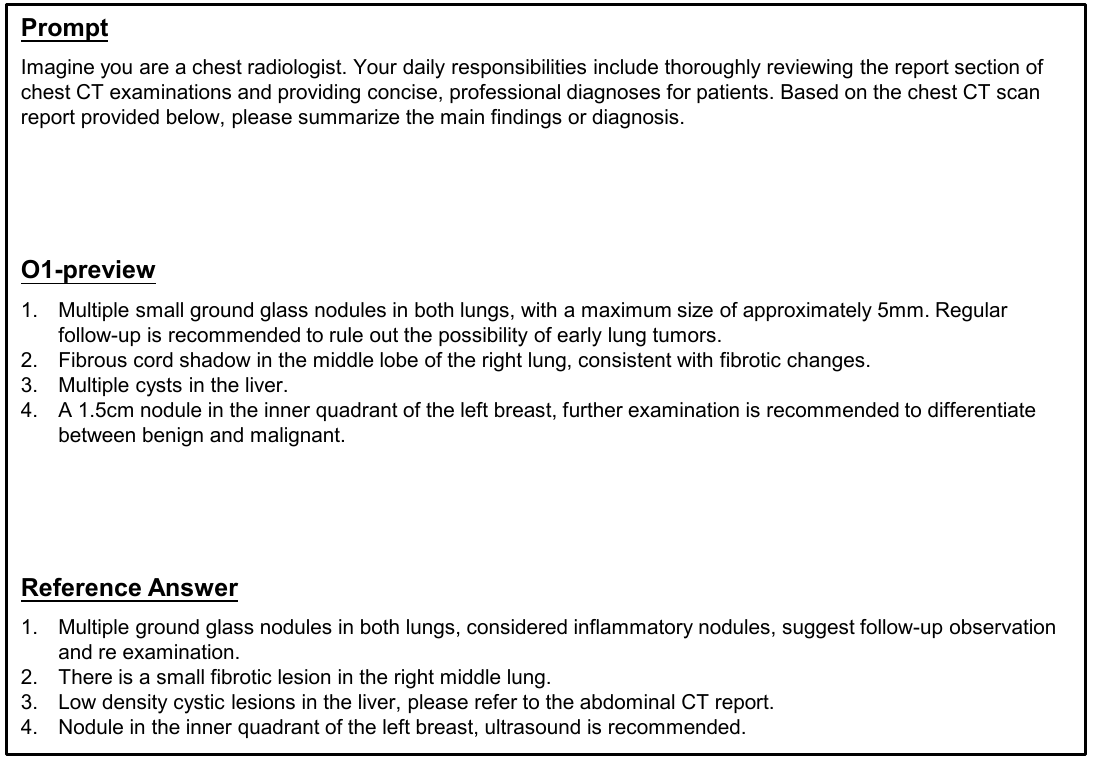}
    \caption[Radiology Report Generation: Case 9]{\textbf{Radiology: Case 9.} We observed that the o1-preview closely aligns with human writing patterns, exhibiting clear organization and concise language. To enhance reader accessibility, the radiological report generated in the zero-shot o1-preview task has been translated from Chinese to English.}
    \label{fig:Radiology_9}
\end{figure}
\newpage

\begin{figure}[H]
    \centering
    \includegraphics[width = \textwidth]{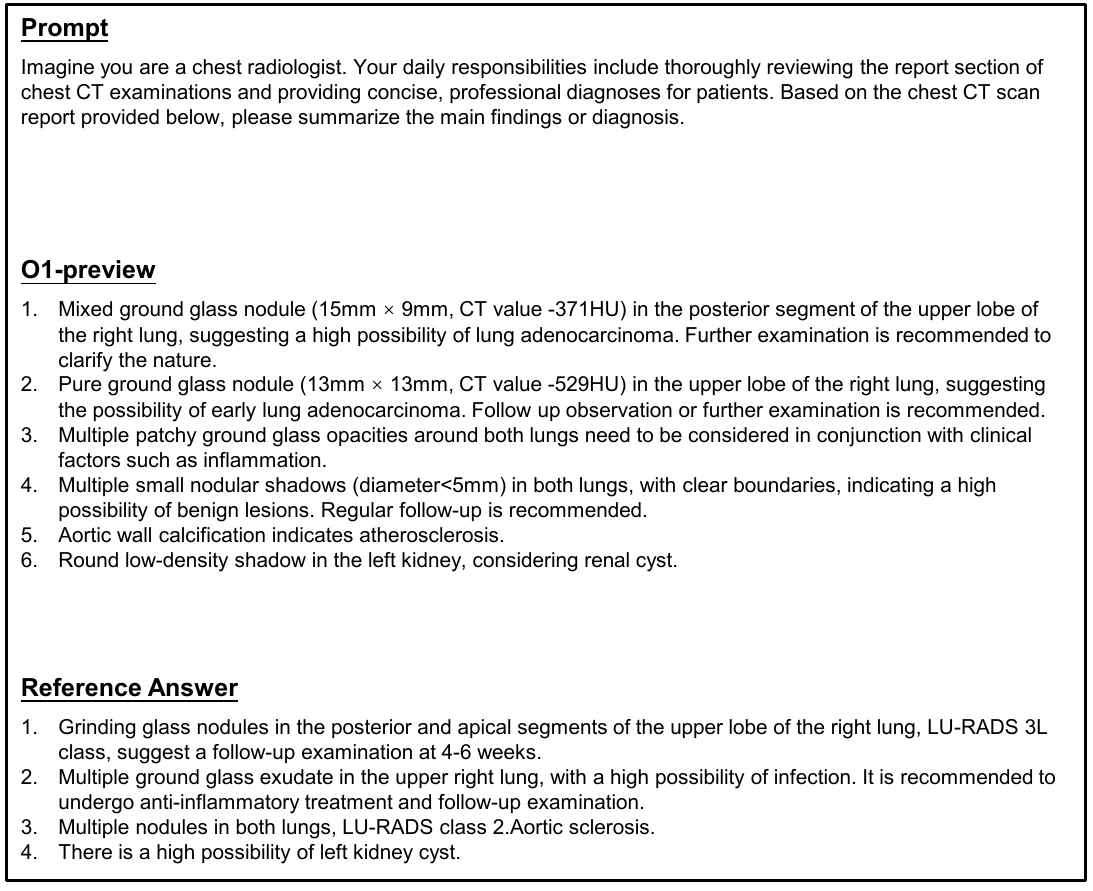}
    \caption[Radiology Report Generation: Case 10]{\textbf{Radiology: Case 10.} We observed that the o1-preview closely aligns with human writing patterns, exhibiting clear organization and concise language. To enhance reader accessibility, the radiological report generated in the zero-shot o1-preview task has been translated from Chinese to English.}
    \label{fig:Radiology_10}
\end{figure}
\newpage


\subsection{Robot Command Planning}
\label{appendix:rcp}
\begin{figure}[H]
\begin{center}
\includegraphics[width = 0.85\textwidth]{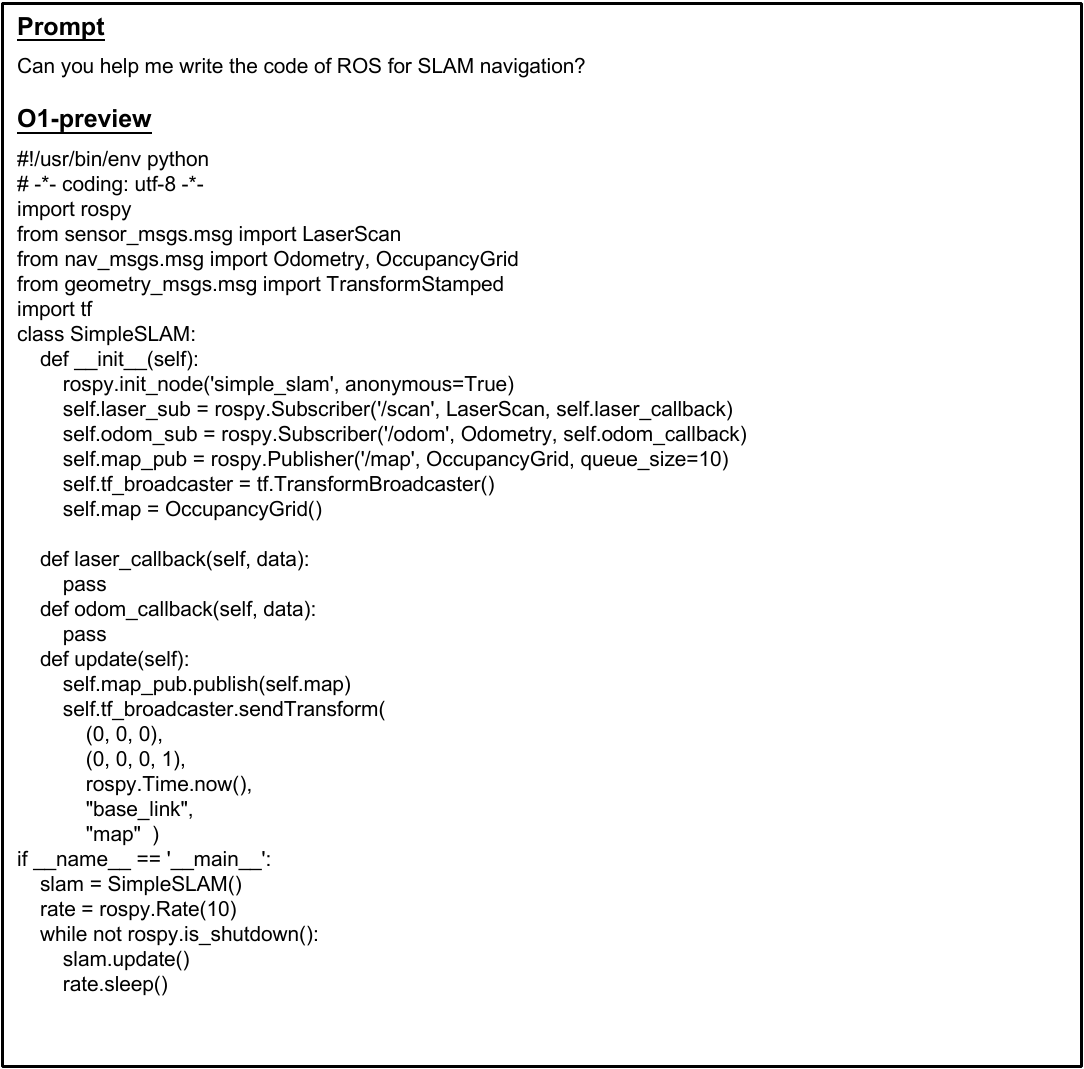}
    \caption[Robot Command Planning: Case 2]{\textbf{Robot Command Planning: Case 2.} o1-preview’s response demonstrates a ROS implementation for Simultaneous Localization and Mapping (SLAM), integrating laser scan and odometry data to build a dynamic map.}
    \label{fig:robot2}    
\end{center}
\end{figure}
\newpage

\begin{figure}[H]
\begin{center}
\includegraphics[width = 0.85\textwidth]{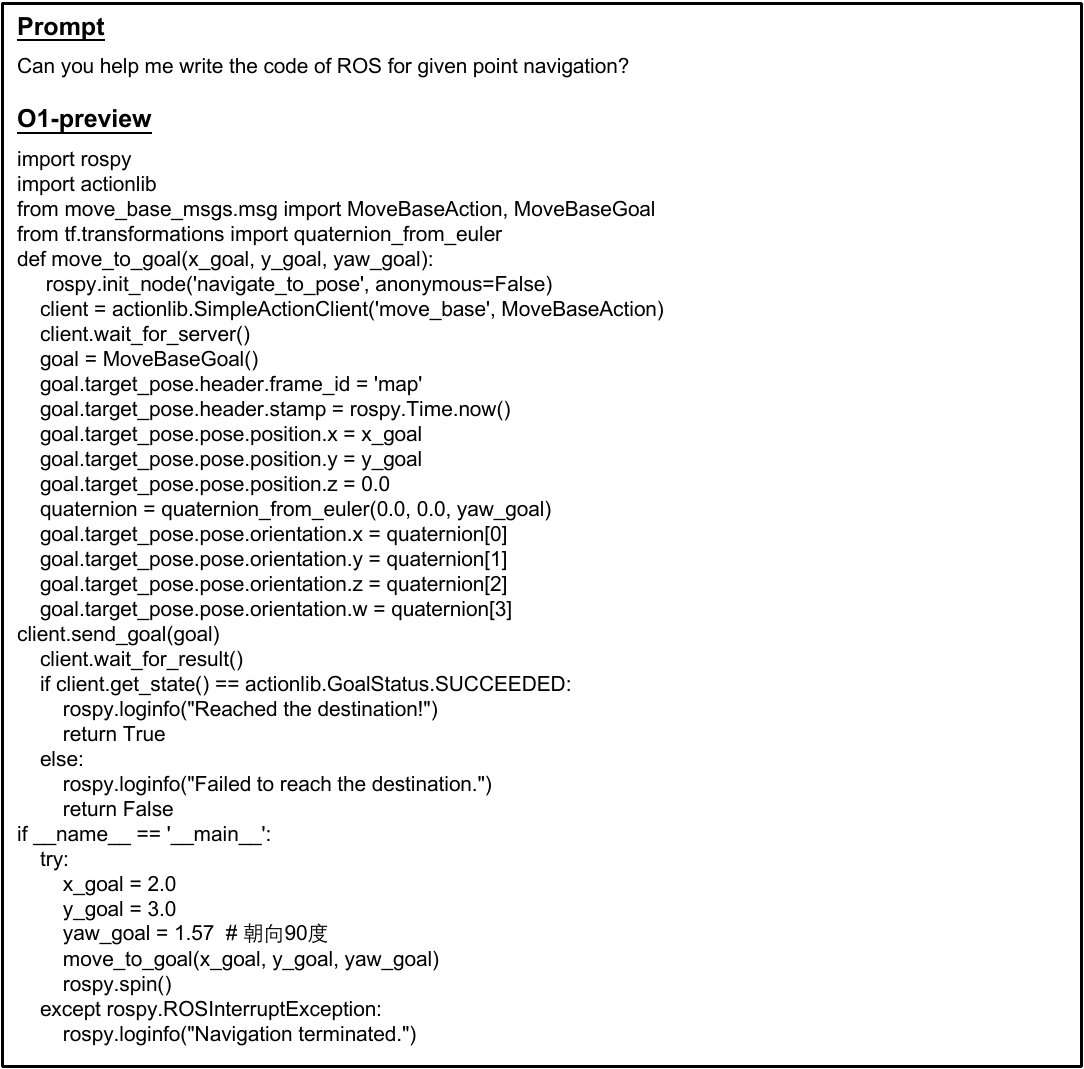}
    \caption[Robot Command Planning: Case 3]{\textbf{Robot Command Planning: Case 3.} The code generated by o1-preview implements point-to-point navigation using the ROS MoveBase action server, leveraging goal coordinates and quaternion-based orientation control. }
    \label{fig:robot3}    
\end{center}
\end{figure}
\newpage

\begin{figure}[H]
\begin{center}
\includegraphics[width = 0.85\textwidth]{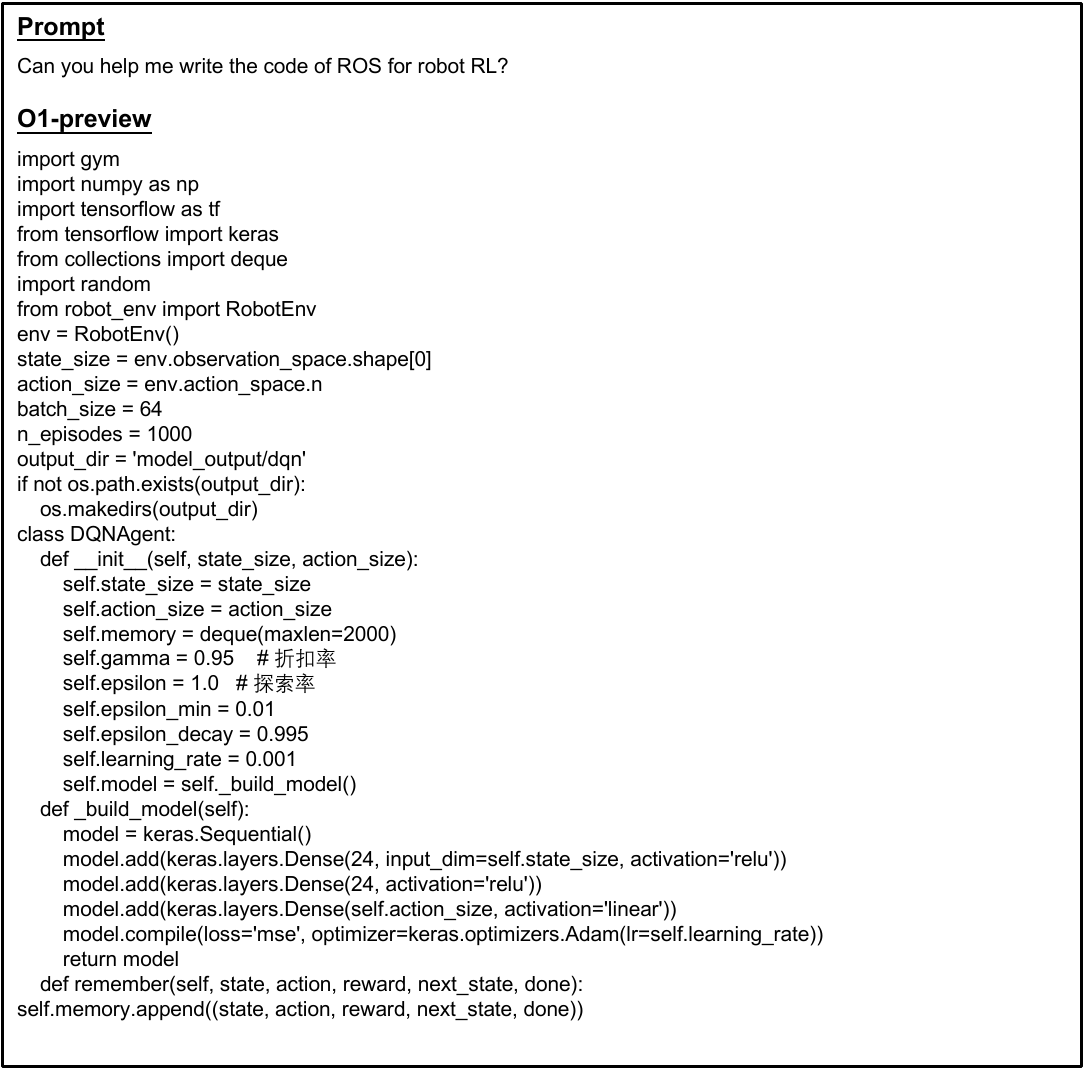}
    \caption[Robot Command Planning: Case 4]{\textbf{Robot Command Planning: Case 4.} A deep Q-learning agent for robot control is generated by o1-preview using ROS, with a neural network built in TensorFlow to optimize decision-making in a simulated environment.}
    \label{fig:robot4}    
\end{center}
\end{figure}
\newpage

\begin{figure}[H]
\begin{center}
\includegraphics[width = 0.85\textwidth]{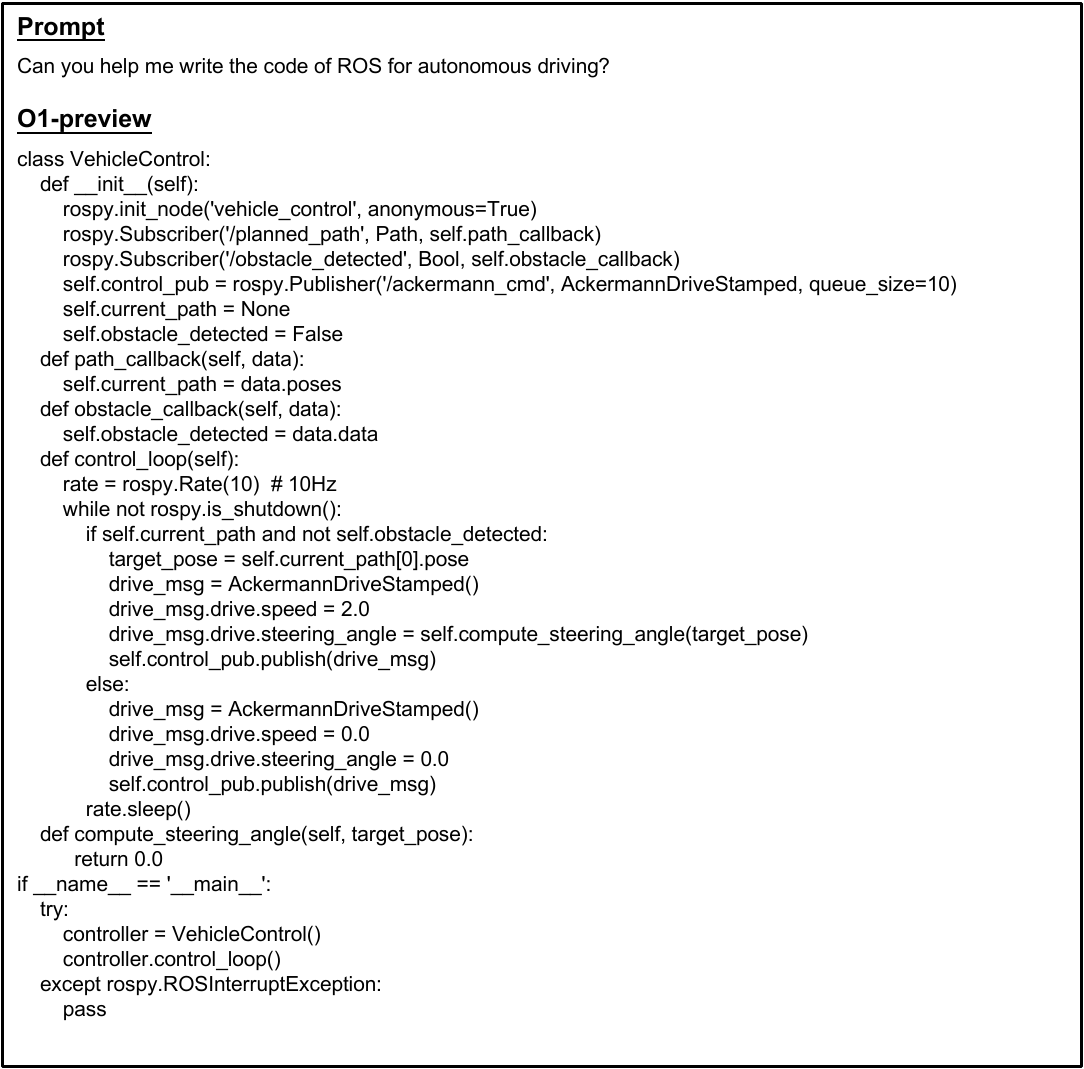}
    \caption[Robot Command Planning: Case 5]{\textbf{Robot Command Planning: Case 5.} o1-preview generates a ROS-based control loop for autonomous driving, processing planned paths and obstacle detection to adjust the vehicle’s speed and steering dynamically. }
    \label{fig:robot5}    
\end{center}
\end{figure}
\newpage


\subsection{Nature Language Inference}
\label{appx_nli}
Additional examples are shown in Table \ref{nli_case2}.

\begin{sidewaystable}[htbp]
\centering
\caption[Nature Language Inference: Case 2]{Additional examples of o1-preview responses on NLI task.}
\label{nli_case2}
\scriptsize 
\resizebox{0.9\textwidth}{!}{ 
\begin{tabularx}{\textwidth}{@{}>{\hsize=.1\hsize}X>{\hsize=.45\hsize}X>{\hsize=.4\hsize}X>{\hsize=0.6\hsize}X>{\hsize=0.2\hsize}X@{}}
\toprule
Dataset & Premise / Context & Hypothesis / Question & o1-preview Response & Label \\
\midrule
MNLI &
The tourist industry continued to expand, and though it became one of the top two income earners in Spain, a realization that unrestricted mass tourism was leading to damaging long-term consequences also began to grow. &
Tourist's caused damage to Spain. &
The premise indicates that mass tourism was leading to "damaging long-term consequences" in Spain. This suggests that the influx of tourists was causing harm to the country over time. Therefore, the hypothesis that "Tourists caused damage to Spain" is directly supported by the premise. This means the premise entails the hypothesis. Answer: \hl{Entailment}&
Entailment\\
\addlinespace[5pt]

ANLI &
Delhi Public School Megacity was established in 2004 at Kolkata belonging to the Delhi Public School Society. It is the only DPS other than Delhi Public School Newtown Kolkata to be affiliated to the Council for the Indian School Certificate Educations(CISCE). &
Delhi Public School Megacity is the only DPS affiliated with CISCE. &
The premise clearly states that there are two DPS schools affiliated with CISCE: Delhi Public School Megacity and Delhi Public School Newtown Kolkata. The hypothesis, however, claims that Delhi Public School Megacity is the only DPS affiliated with CISCE. This directly contradicts the information in the premise, which acknowledges the existence of another affiliated school. Therefore, the relationship between the premise and the hypothesis is a contradiction. Answer: \hl{Contradiction} &
Contradiction\\
\addlinespace[5pt]

QNLI &
In addition, "Super Bowl City" opened on January 30 at Justin Herman Plaza on The Embarcadero, featuring games and activities that will highlight the Bay Area's technology, culinary creations, and cultural diversity. &
What was the name of the event at the The Embarcadero that was held prior to Super Bowl 50 to help show off some of the things that San Francisco has to offer? &
\hl{Yes}, the context sentence contains the answer. The event was called "Super Bowl City." &
Entailment\\
\addlinespace[5pt]

\midrule

MedNLI &
Mother developed separation of symphysis pubis and was put in traction. &
She has orthopedic injuries. &
The premise states that the mother experienced a separation of the symphysis pubis—a joint located in the pelvis—and required traction. This condition is an injury to the musculoskeletal system, specifically involving bones and joints, which falls under the field of orthopedics. Therefore, the hypothesis that "She has orthopedic injuries" is directly supported by the information in the premise. Thus, the premise entails the hypothesis. Answer: \hl{Entailment} &
Entailment\\
\addlinespace[5pt]

RadQNLI &
LEFT UPPER EXTREMITY ULTRASOUND: [**Doctor Last Name **] scale and Doppler son[**Name (NI) 76**] of the left internal jugular, subclavian, axillary, basilic, cephalic, and brachial veins were performed. &
Are there any clots in the deep veins of the left upper extremity? &
\hl{No}, the context sentence does not contain the answer to the question. It mentions that an ultrasound was performed on specific veins of the left upper extremity but does not indicate whether any clots were found. &
Not entailment\\
\bottomrule
\end{tabularx}
}
\end{sidewaystable}
\newpage


\subsection{Quantitative Investing}

\begin{figure}[H]
    \centering
    \includegraphics[width = \textwidth]{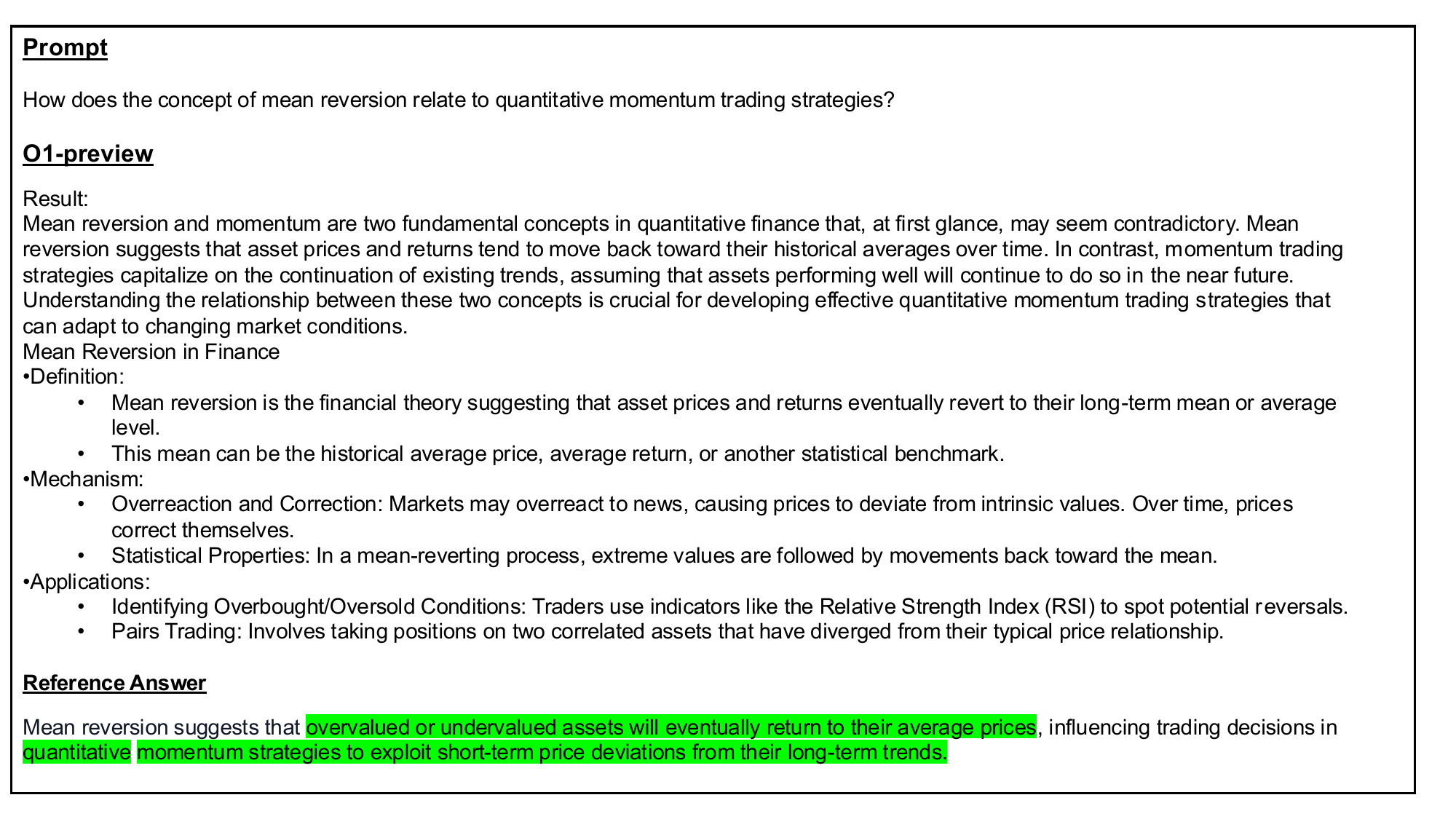}
    \caption[Quantitative Investing: Case 3]{\textbf{Quantitative Investing: Case 3.} o1-preview successfully explained the concept of mean reversion, detailing its working mechanism, and introduced its downstream tasks, demonstrating an accurate understanding of the underlying principles.}
    \label{fig:Qsample3}
\end{figure}
\newpage

\begin{figure}[H]
    \centering
    \includegraphics[width = \textwidth]{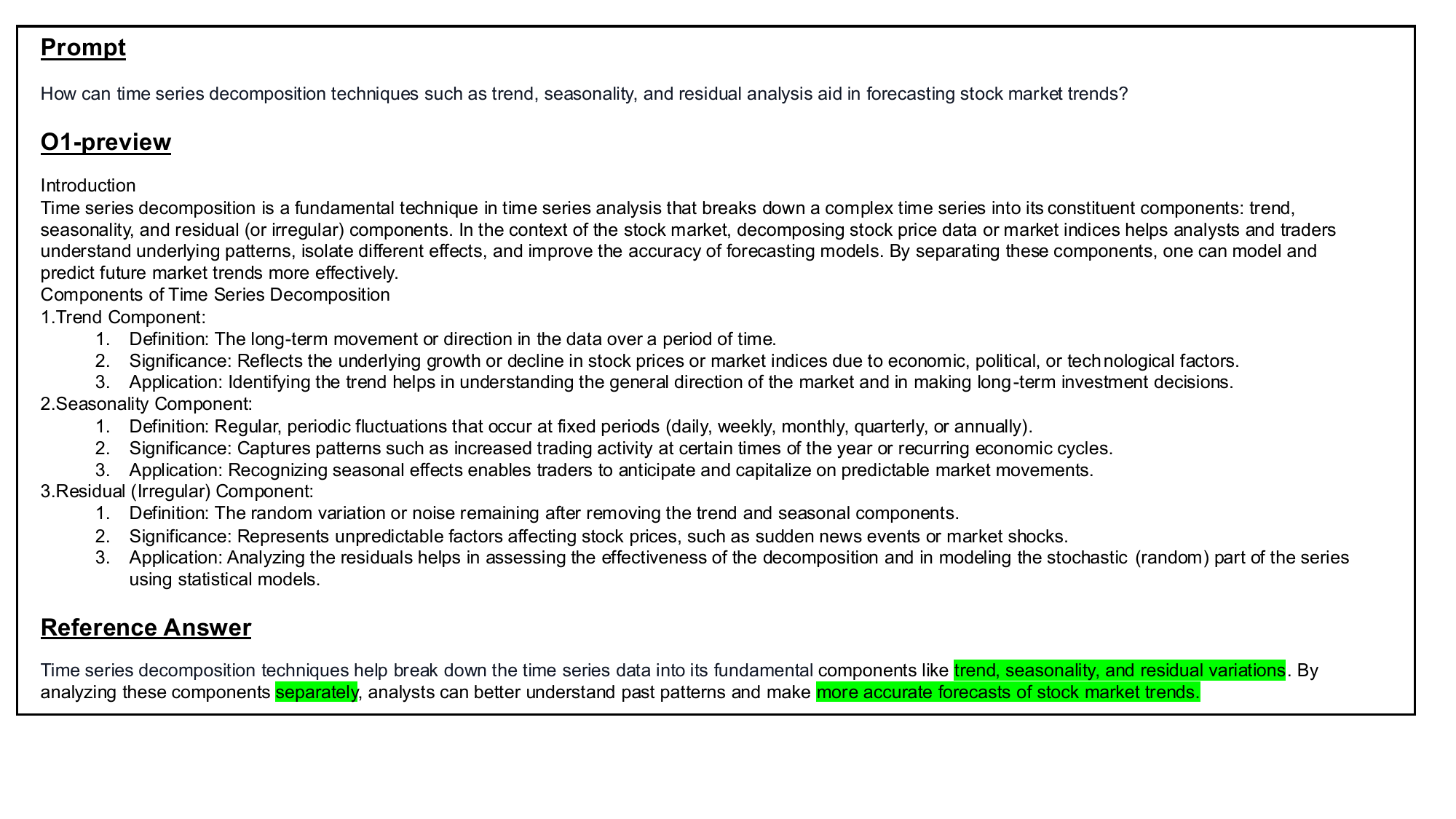}
    \caption[Quantitative Investing: Case 4]{\textbf{Quantitative Investing: Case 4.} o1-preview successfully analyzed the differences among trend, seasonality, and residual analysis in stock market forecasting, and explained how they work together, demonstrating a comprehensive understanding of their combined role in predicting market movements.}
    \label{fig:Qsample4}
\end{figure}
\newpage

\begin{figure}[H]
    \centering
    \includegraphics[width = \textwidth]{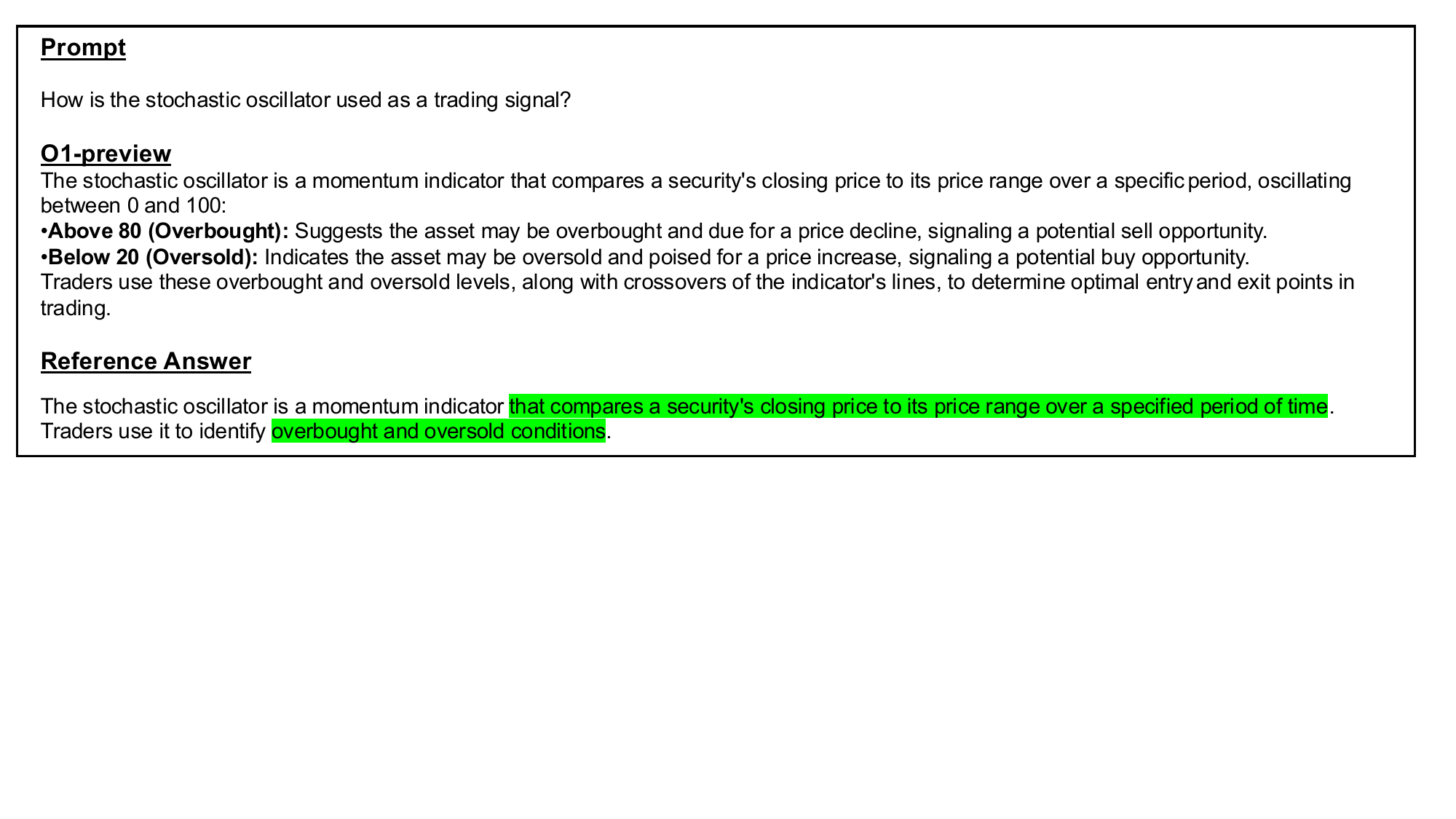}
    \caption[Quantitative Investing: Case 5]{\textbf{Quantitative Investing: Case 5.} o1-preview successfully explained the stochastic oscillator, provided examples, and demonstrated its application, showing an accurate understanding of the underlying concept.}
    \label{fig:Qsample5}
\end{figure}
\newpage


\subsection{Low-Resource Language Translation}

\begin{figure}[H]
    \centering
    \includegraphics[trim=140 0 140 0,clip,width = \textwidth]{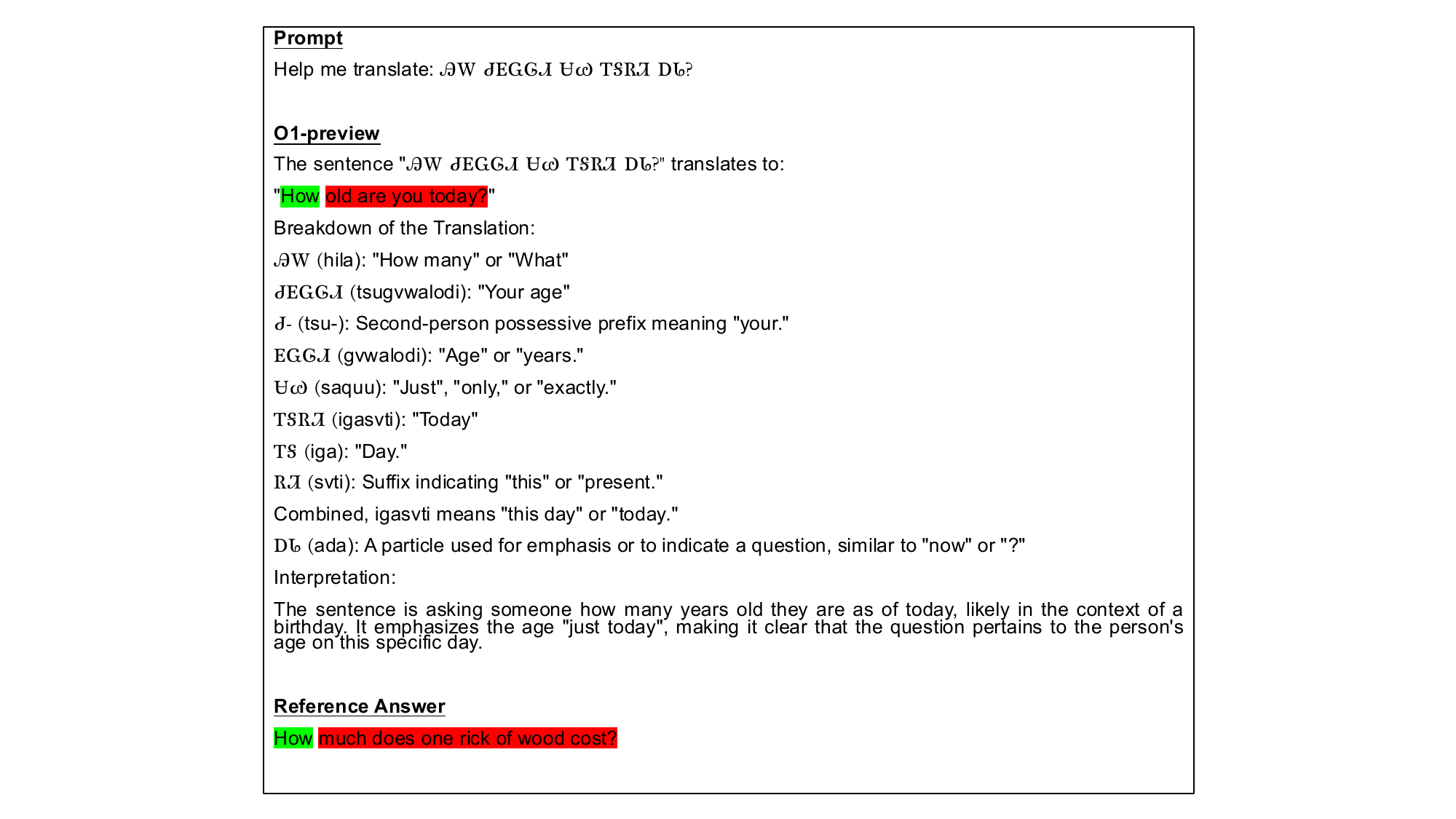}
    \caption[Low-Resource Language Translation: Case 3]{\textbf{Low-Resource Language Translation: Case 3.} This test evaluates the o1-preview model's ability to translate the Cherokee sentence.The model translates the sentence as "How old are you today?" and provides a detailed breakdown of the translation components, explaining the meaning of each Cherokee word and particle. However, the translation is totally different from the reference answer due to mistakenly translating the words in the sentence.}
    \label{fig:Cherokee_3}
\end{figure}
\newpage

\begin{figure}[H] 
    \centering
    \includegraphics[trim=260 5 260 10,clip,width = \textwidth]{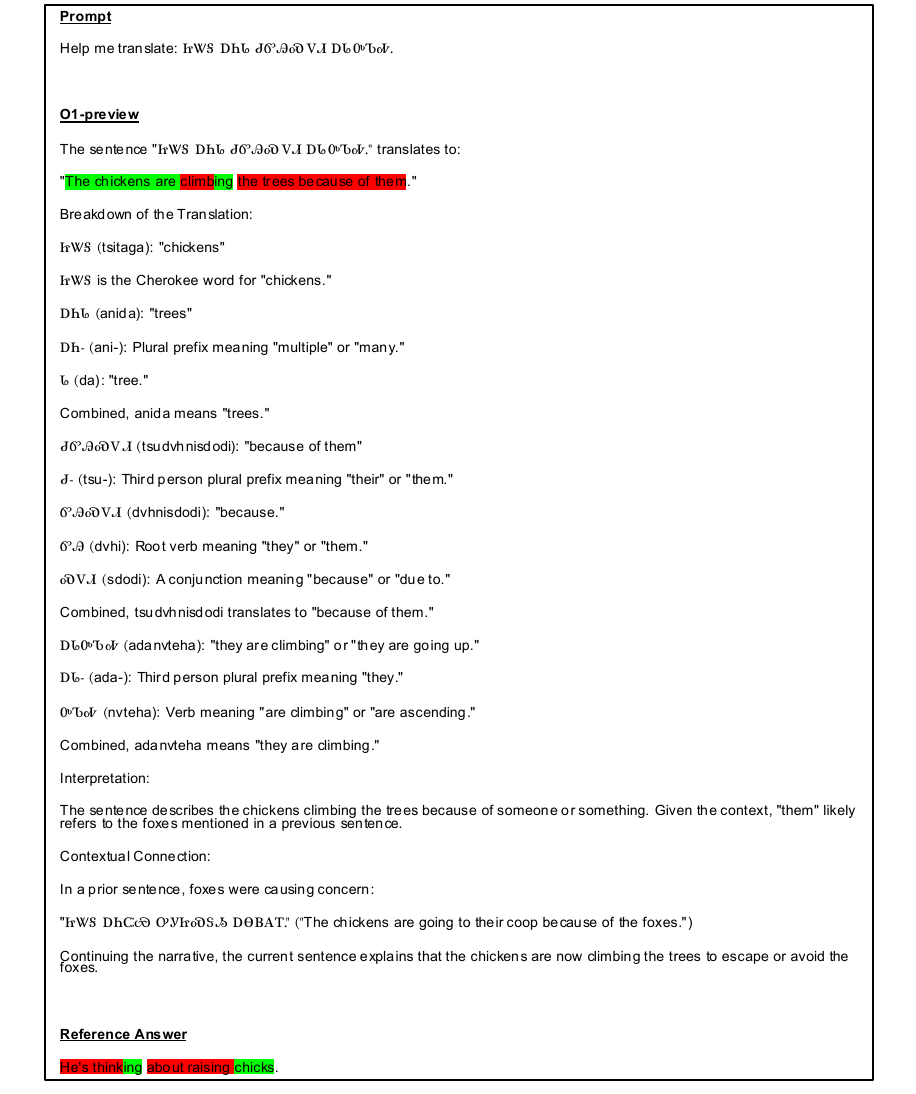}
    \caption[Low-Resource Language Translation: Case 4]{\textbf{Low-Resource Language Translation: Case 4.} The o1-preview model correctly translates the noun chickens and accurately captures the present progressive tense. The o1-preview model also attempts to establish a contextual connection, suggesting that the chickens are reacting to foxes mentioned in a previous sentence. However, this contextual interpretation is incorrect, as the reference answer, "He's thinking about raising chicks," is the ground truth, indicating a misalignment in the overall translation.}
    \label{fig:Cherokee_4}
\end{figure}
\newpage

\begin{figure}[H]
    \centering
    \includegraphics[trim=55 0 55 0,clip,width = \textwidth]{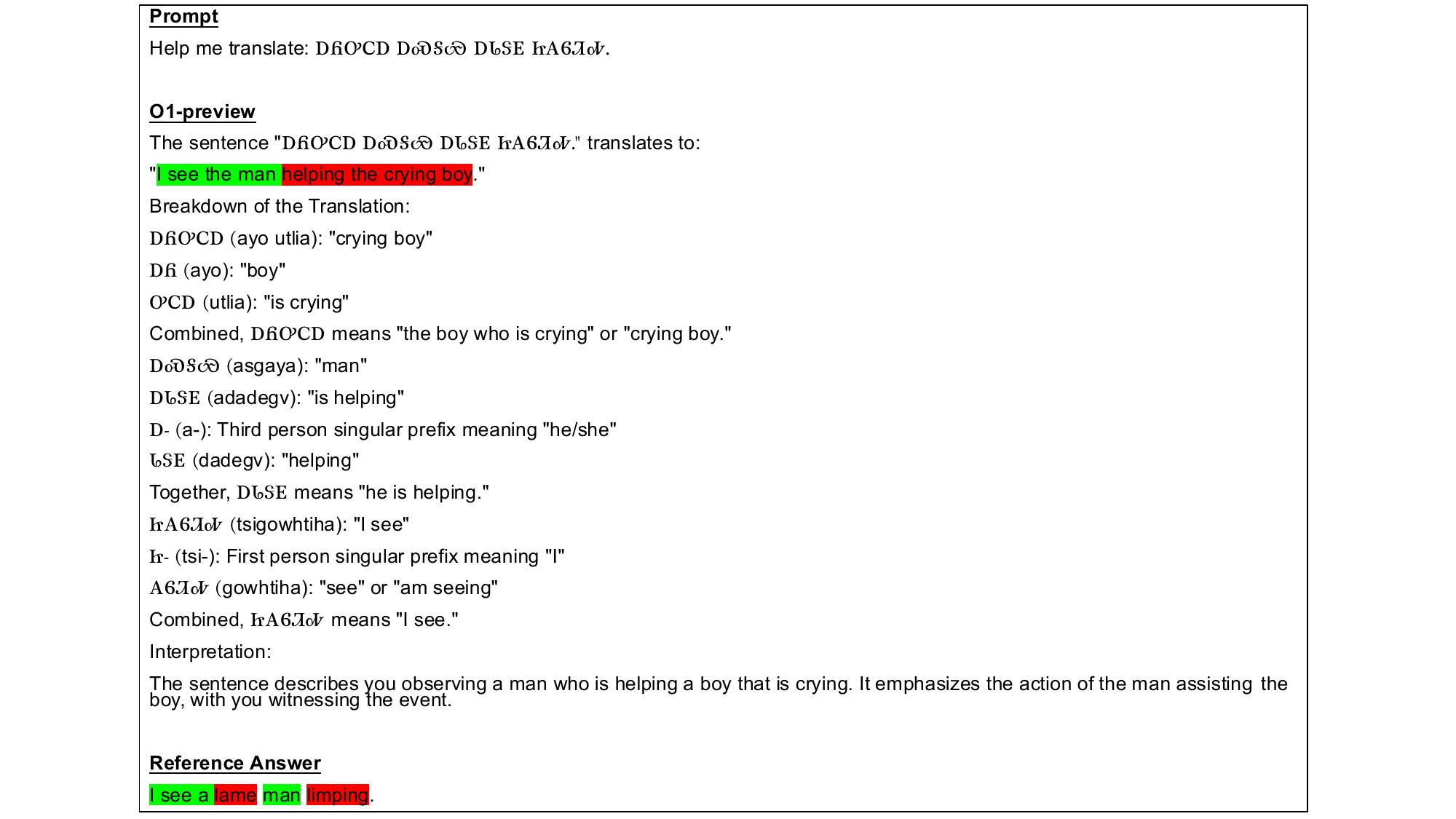}
   \caption[Low-Resource Language Translation: Case 5]{\textbf{Low-Resource Language Translation: Case 5.} The o1-preview model successfully translate the sentence into "I see a man doing something". The provided detailed breakdown of each component reveals a mismatch in the rest interpretation. While the model succeeds in grammatical analysis, the translation result does not align with the intended meaning.}
    \label{fig:Cherokee_5}
\end{figure}
\newpage


\subsection{Educational Q\&A}

\begin{figure}[H]
    \centering
    \includegraphics[width = 0.9\textwidth]{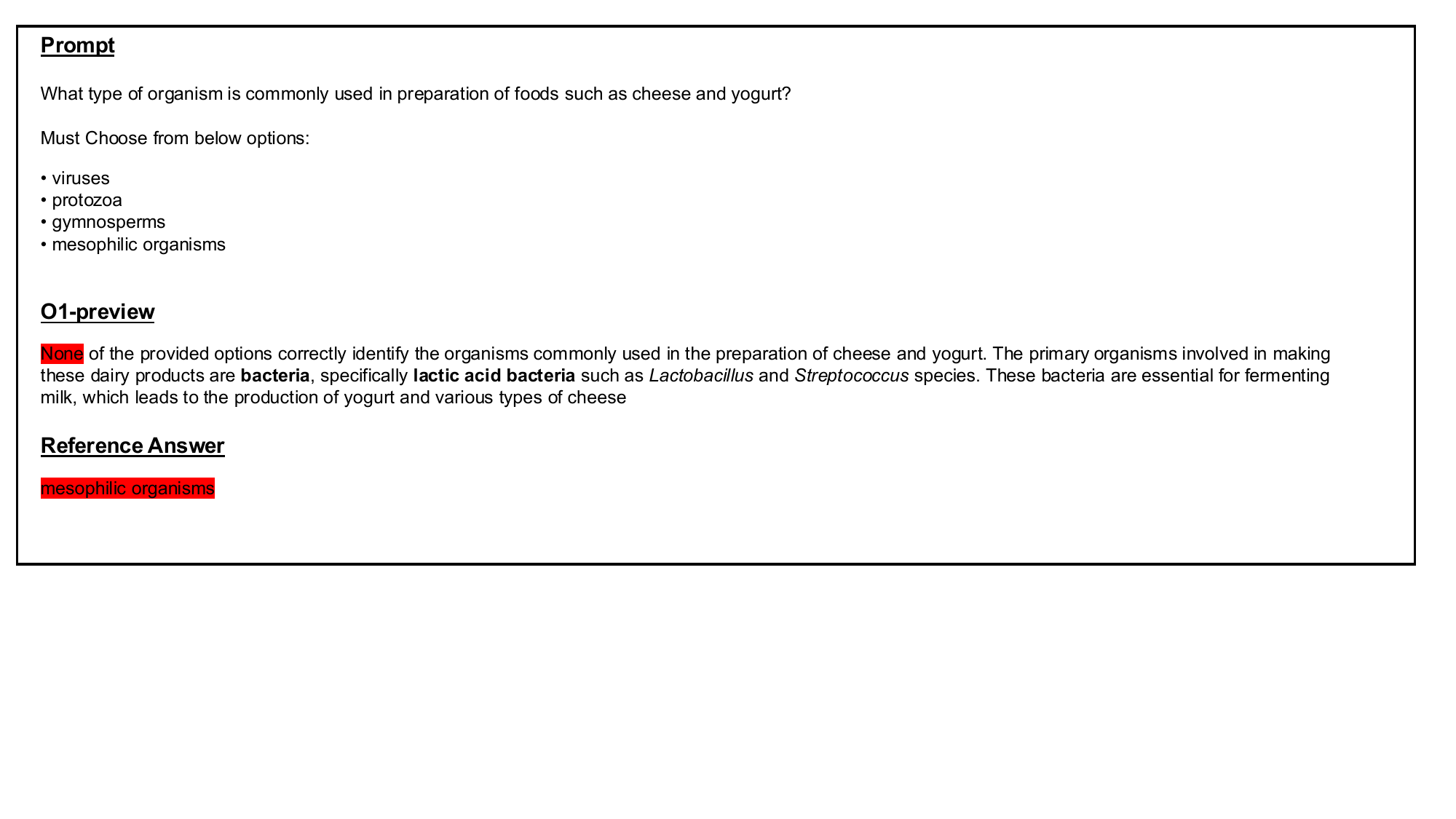}
    \caption[Educational Q\&A: Case 3]{\textbf{Educational Q\&A: Case 3.} o1-preview identified that none of the provided options were correct for the question regarding organisms used in the preparation of cheese and yogurt. Instead, it overthought the problem by providing an explanation about lactic acid bacteria (such as Lactobacillus and Streptococcus) being responsible for the fermentation process, despite the question’s simpler intent. This case demonstrates that o1-preview may sometimes overthink and fail to select from the given options. The dataset for this task is sourced from \cite{welbl2017crowdsourcing}.}
    \label{fig:eudcation_3}
\end{figure}
\newpage

\begin{figure}[H]
    \centering
    \includegraphics[width = 0.9\textwidth]{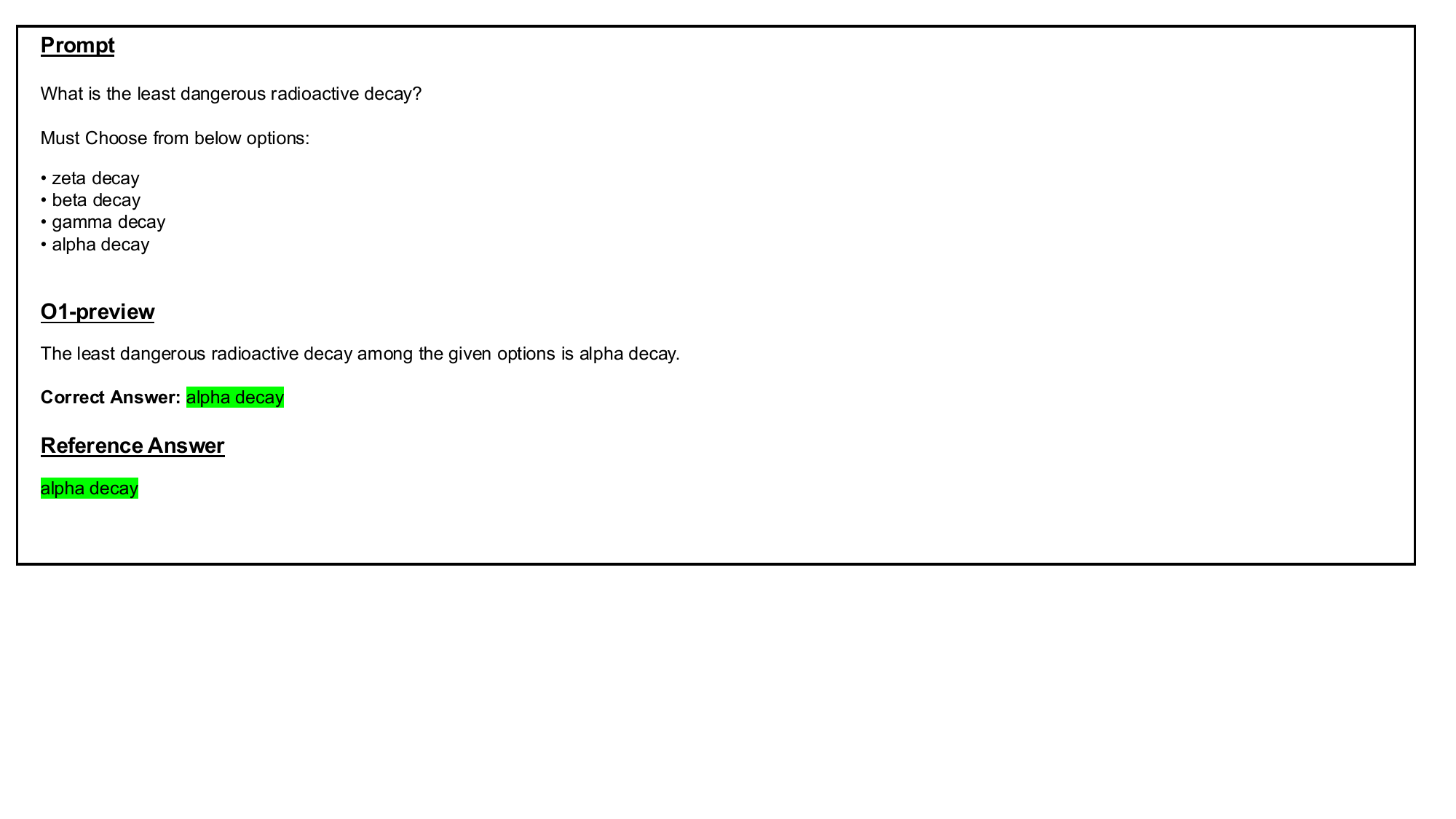}
    \caption[Educational Q\&A: Case 4]{\textbf{Educational Q\&A: Case 4.} o1-preview accurately determined that alpha decay is the least dangerous form of radioactive decay among the provided options, showcasing its ability to interpret and correctly answer safety-related questions about radioactive materials. The dataset for this analysis is sourced from \cite{welbl2017crowdsourcing}.}
    \label{fig:eudcation_4}
\end{figure}
\newpage

\begin{figure}[H]
    \centering
    \includegraphics[width = 0.9\textwidth]{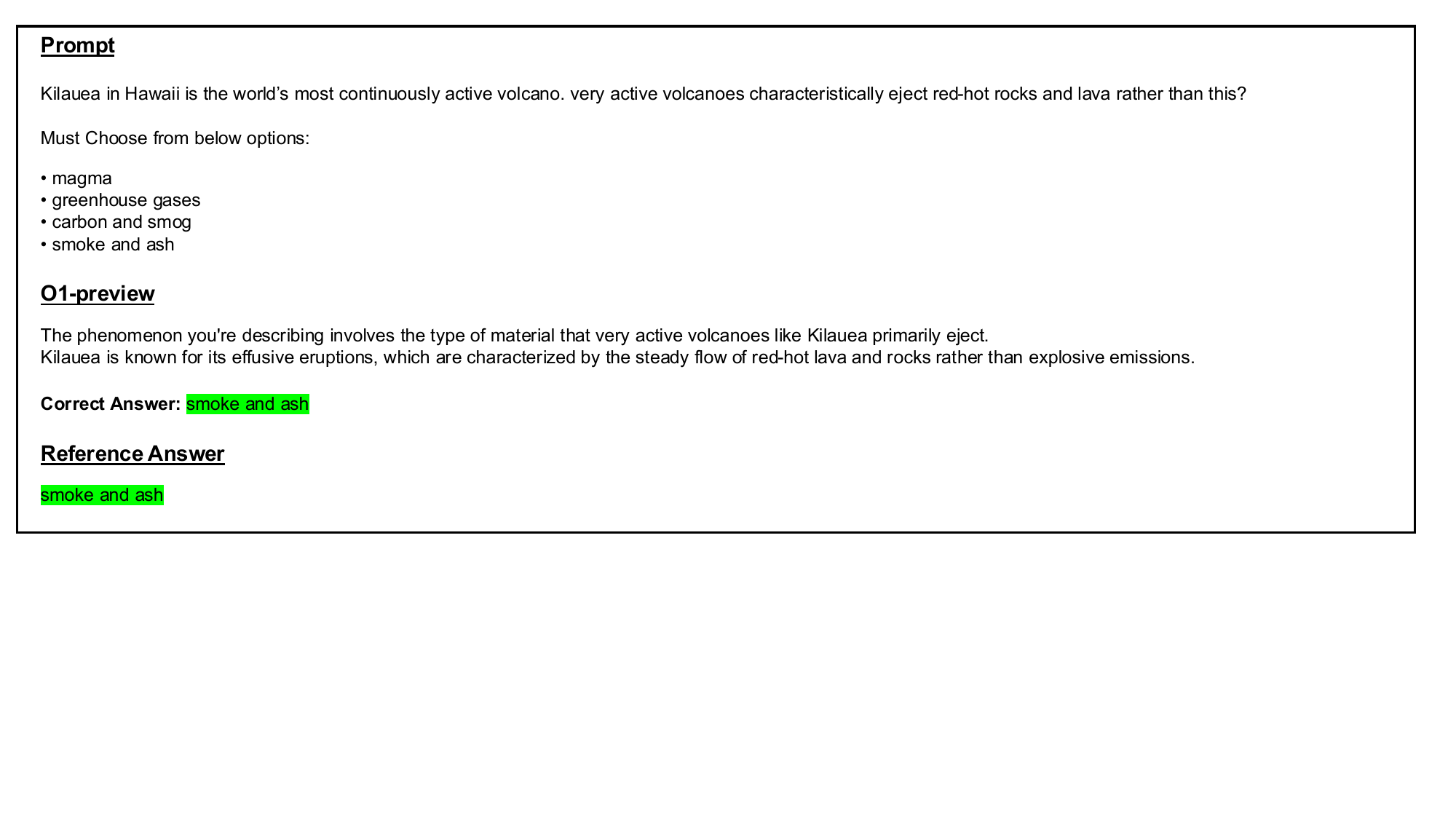}
    \caption[Educational Q\&A: Case 5]{\textbf{Educational Q\&A: Case 5.} o1-preview accurately determined that smoke and ash were the correct answer in a question about volcanic eruptions. The model identified the characteristics of Kilauea's effusive eruptions, showing that it understands the difference between material ejected in explosive versus effusive volcanic activities. The dataset for this task is sourced from \cite{welbl2017crowdsourcing}.}
    \label{fig:eudcation_5}
\end{figure}
\newpage


\subsection{Logical Reasoning}
 \begin{figure}[H]
     \centering
     \includegraphics[width = 0.85\textwidth]{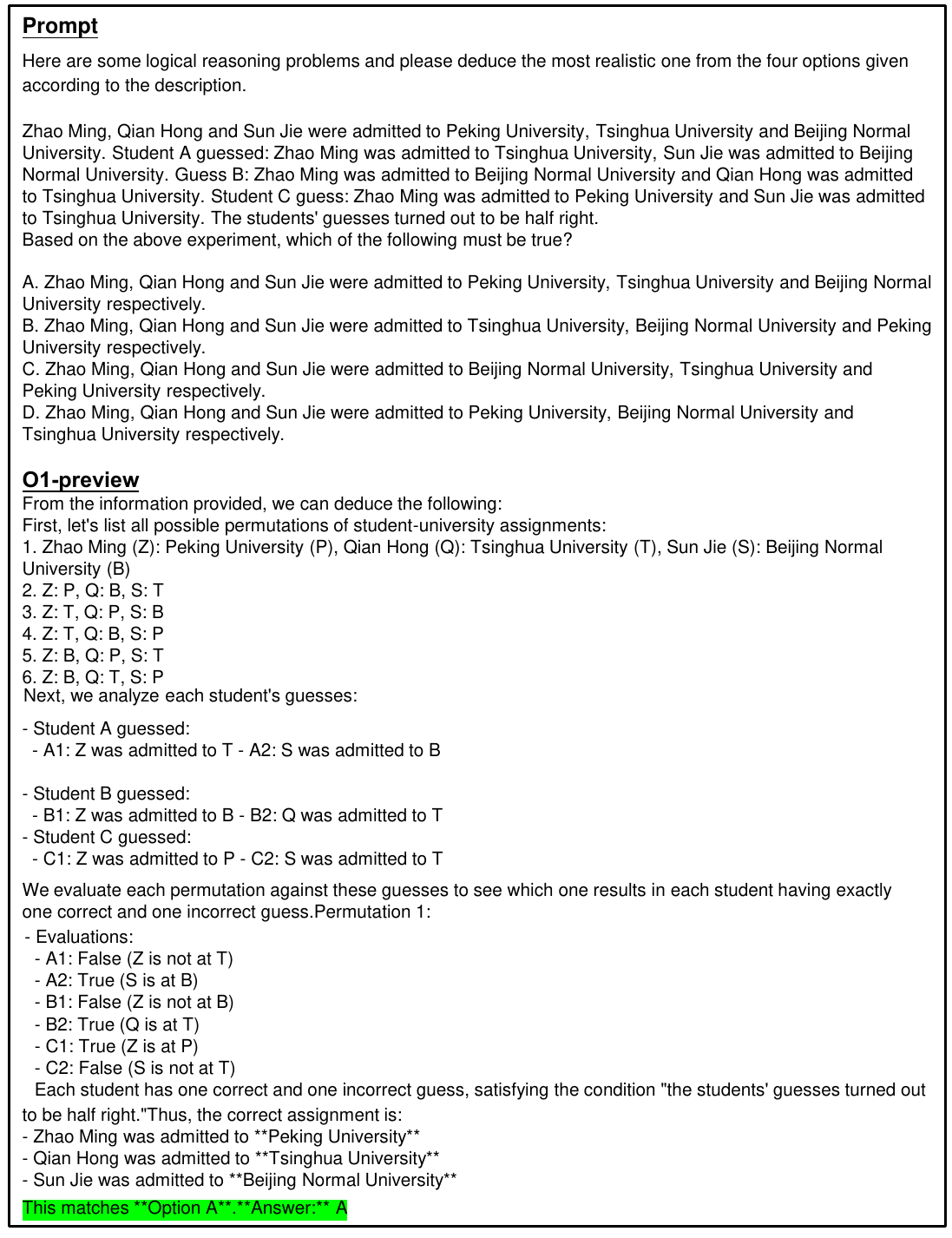}
     \caption[Logical Reasoning: Case 3]{\textbf{Logical Reasoning: Case 3.} Three students, Zhao Ming, Qian Hong, and Sun Jie, were admitted to Peking University, Tsinghua University, and Beijing Normal University. Student A, B, and C made guesses, but only half of each guess was correct. After analysis, it was determined that Zhao Ming was admitted to Peking University, Qian Hong to Tsinghua University. }
     \label{fig:Logical Reasoning: Case 3}
 \end{figure}
 \newpage

 \begin{figure}[H]
      \centering
     \includegraphics[width = 0.85\textwidth]{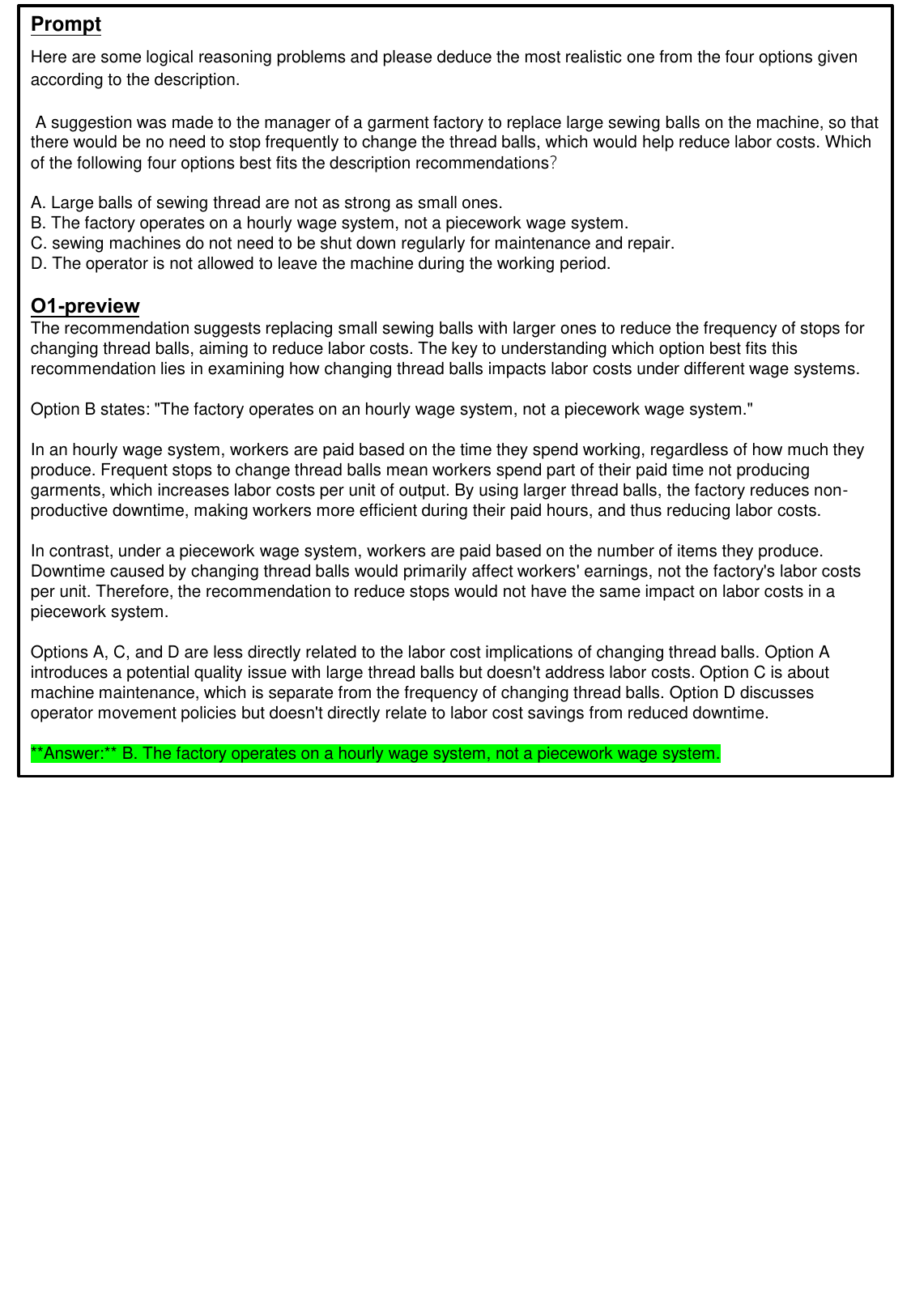}
     \caption[Logical Reasoning: Case 4]{\textbf{Logical Reasoning: Case 4.}  The analysis performance of the o1-preview model is remarkable. It accurately grasps the core purpose of the suggestion in the question stem, which is to reduce labor costs. By in-depth analysis of the relationship between each option and this purpose, it clearly points out the close connection between option B and reducing labor costs. In the analysis process, a detailed and reasonable explanation is provided for the impact of replacing sewing balls under different wage systems on labor costs.  }
     \label{fig:Logical Reasoning: Case 4}
 \end{figure}
 \newpage

 \begin{figure}[H]
      \centering
     \includegraphics[width = 0.85\textwidth]{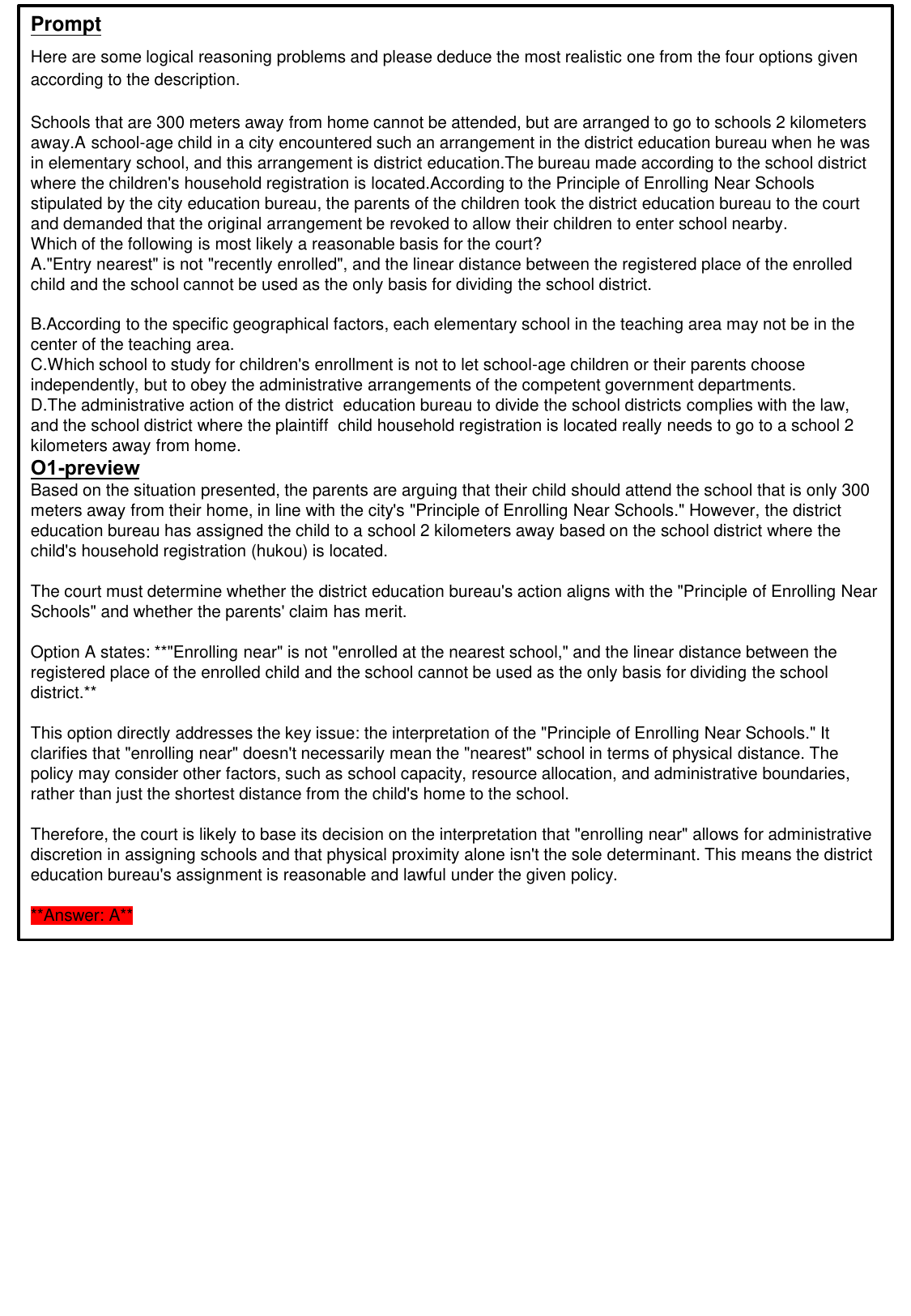}
     \caption[Logical Reasoning: Case 5]{\textbf{Logical Reasoning: Case 5.} It can closely focus on the core of the problem and deeply analyze the relationship between each option and the key issue of the case. The interpretation of option A is accurate and in place. It not only considers the specific meaning of the policy but also makes reasonable inferences based on the actual situation, providing a clear logical thinking for reaching the correct answer.  }
     \label{fig:Logical Reasoning: Case 5}
 \end{figure}
 \newpage

  \begin{figure}[H]
      \centering
     \includegraphics[width = 0.85\textwidth]{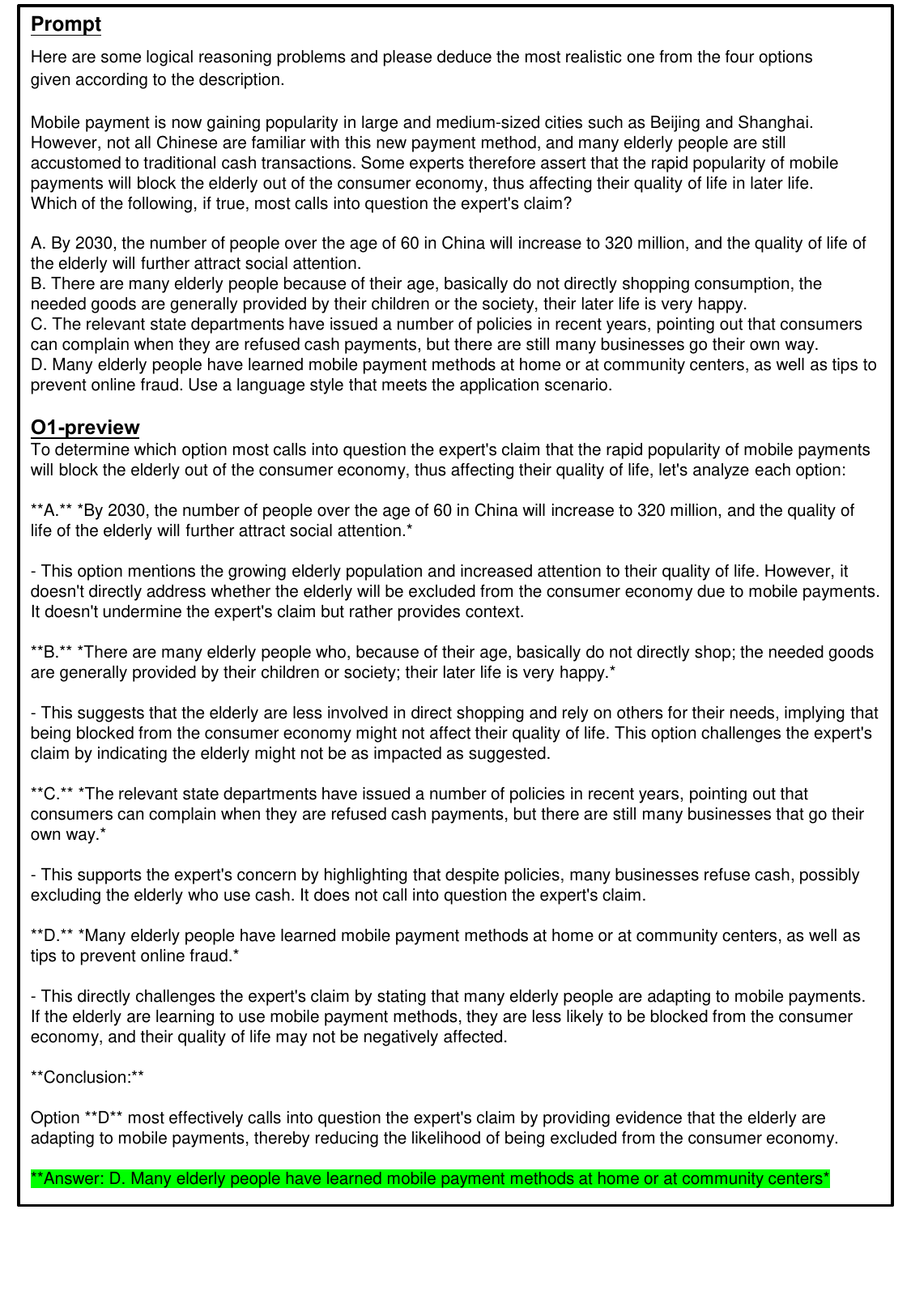}
     \caption[Logical Reasoning: Case 6]{\textbf{Logical Reasoning: Case 6.} The analysis performance of the o1-preview model is rather excellent. It conducts a meticulous analysis of each option, clearly expounds the relationship between each option and the expert's view. Through step-by-step reasoning, it accurately identifies the option that most effectively challenges the expert's view. In the analysis process, the logic is rigorous and comprehensive considerations are made, which can effectively help users understand the problem and reach the correct answer. }
     \label{fig:Logical Reasoning: Case 6}
 \end{figure}
 \newpage
 
\begin{figure}[H]
      \centering
     \includegraphics[width = 0.85\textwidth]{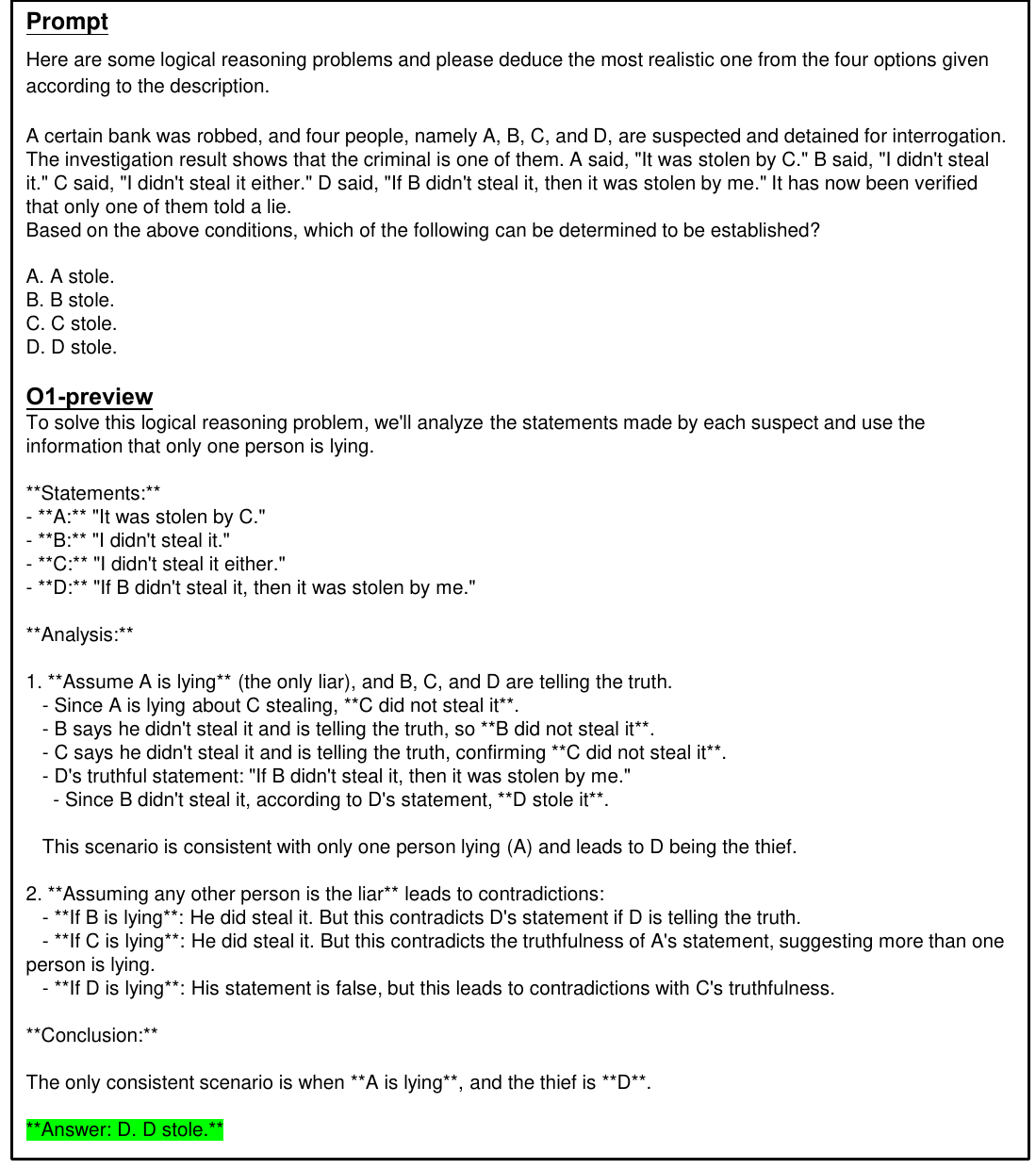}
     \caption[Logical Reasoning: Case 7]{\textbf{Logical Reasoning: Case 7.} The o1-preview model performs extremely well in analyzing the bank robbery problem. It closely adheres to the key point that only one person is lying. It meticulously analyzes each person's statement and conducts step-by-step reasoning through assumptions. This method is both organized and efficient. When delving into each assumption in combination with the problem's conditions, for example, when assuming that A is lying, it accurately deduces the result. The logic is strict and convincing. Additionally, the model fully considers various possibilities to avoid one-sided conclusions, providing users with comprehensive and accurate analysis results, which helps users better understand the problem and reach the correct answer. }
     \label{fig:Logical Reasoning: Case 7}
 \end{figure}
 \newpage
 
  \begin{figure}[H]
      \centering
     \includegraphics[width = 0.85\textwidth]{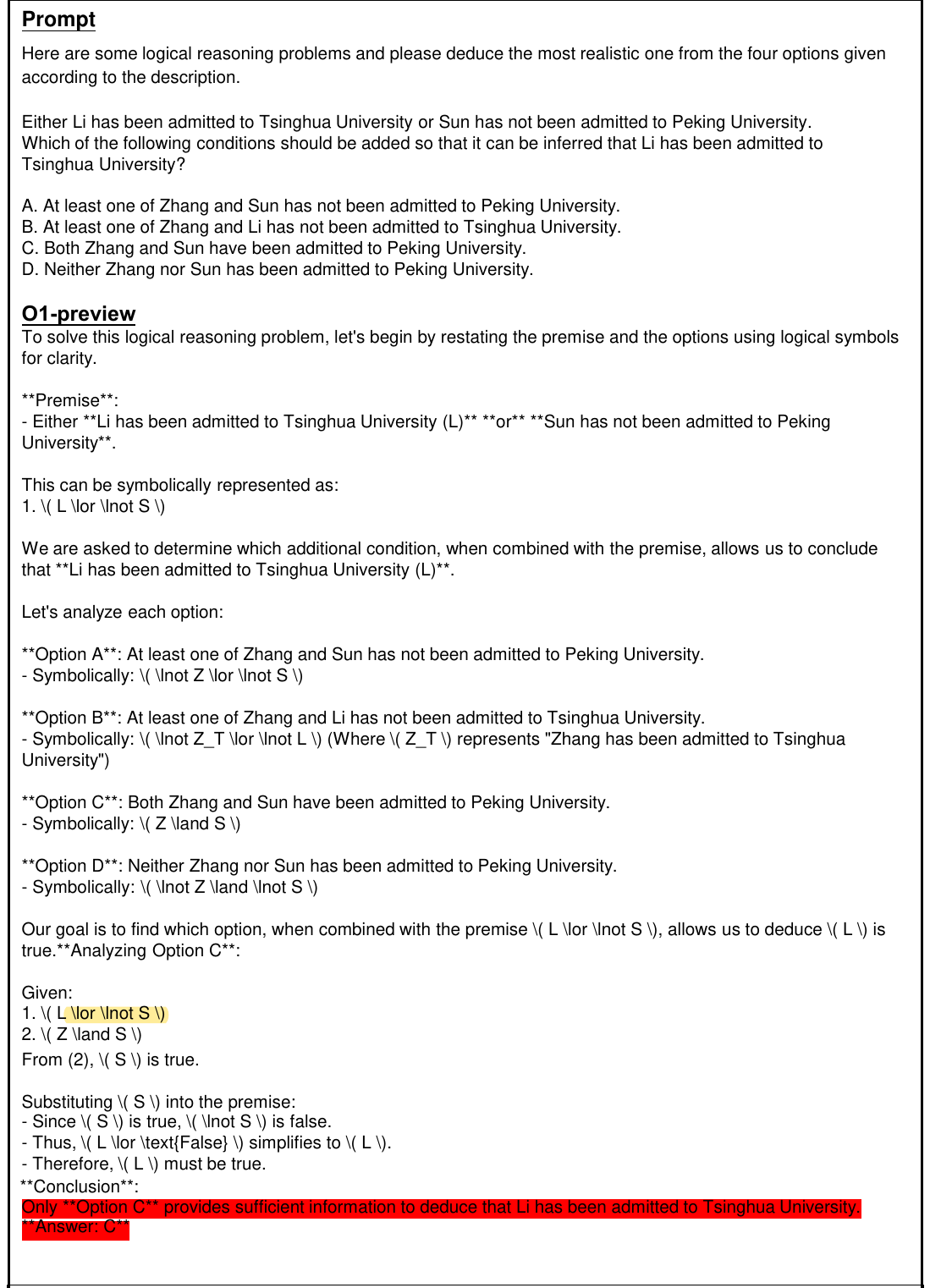}
     \caption[Logical Reasoning: Case 8]{\textbf{Logical Reasoning: Case 8.} Red indicates that the model's output is completely inconsistent with the reference diagnosis, The o1-preview model is highly effective in analyzing this logical reasoning problem. It presents the premise “Either Li has been admitted to Tsinghua University or Sun has not been admitted to Peking University” using logical symbols. When analyzing options, it closely integrates the problem content. For option C, starting from the given condition “Both Zhang and Sun have been admitted to Peking University,” it conducts accurate deduction, showing a tight logical process and helping users understand and find the correct answer. }
     \label{fig:Logical Reasoning: Case 8}
 \end{figure}
  \newpage
  
  \begin{figure}[H]
      \centering
     \includegraphics[width = 0.85\textwidth]{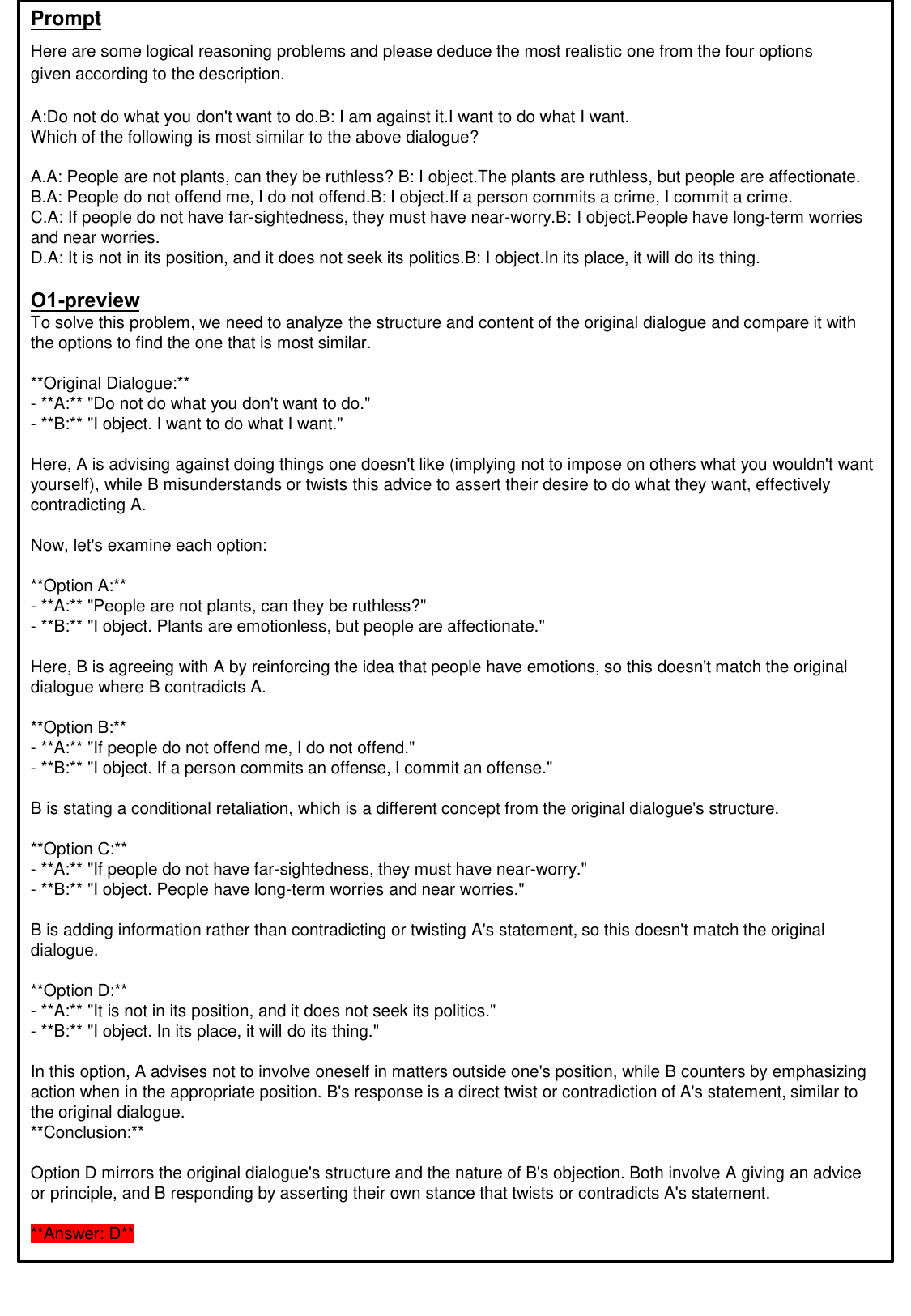}
     \caption[Logical Reasoning: Case 9]{\textbf{Logical Reasoning: Case 9.} The o1-preview model demonstrates excellent performance in analyzing this problem. When analyzing option D, it clearly points out that A's statement "If one is not in a particular office, one should not meddle in its affairs" and B's response "I object. If one is in a particular office, one will handle its affairs" are similar in structure to the original dialogue. In both cases, A presents a principle-based view, and B responds in a way that twists or rebuts. }
     \label{fig:Logical Reasoning: Case 9}
 \end{figure}
  \newpage
  
  \begin{figure}[H]
      \centering
     \includegraphics[width = 0.85\textwidth]{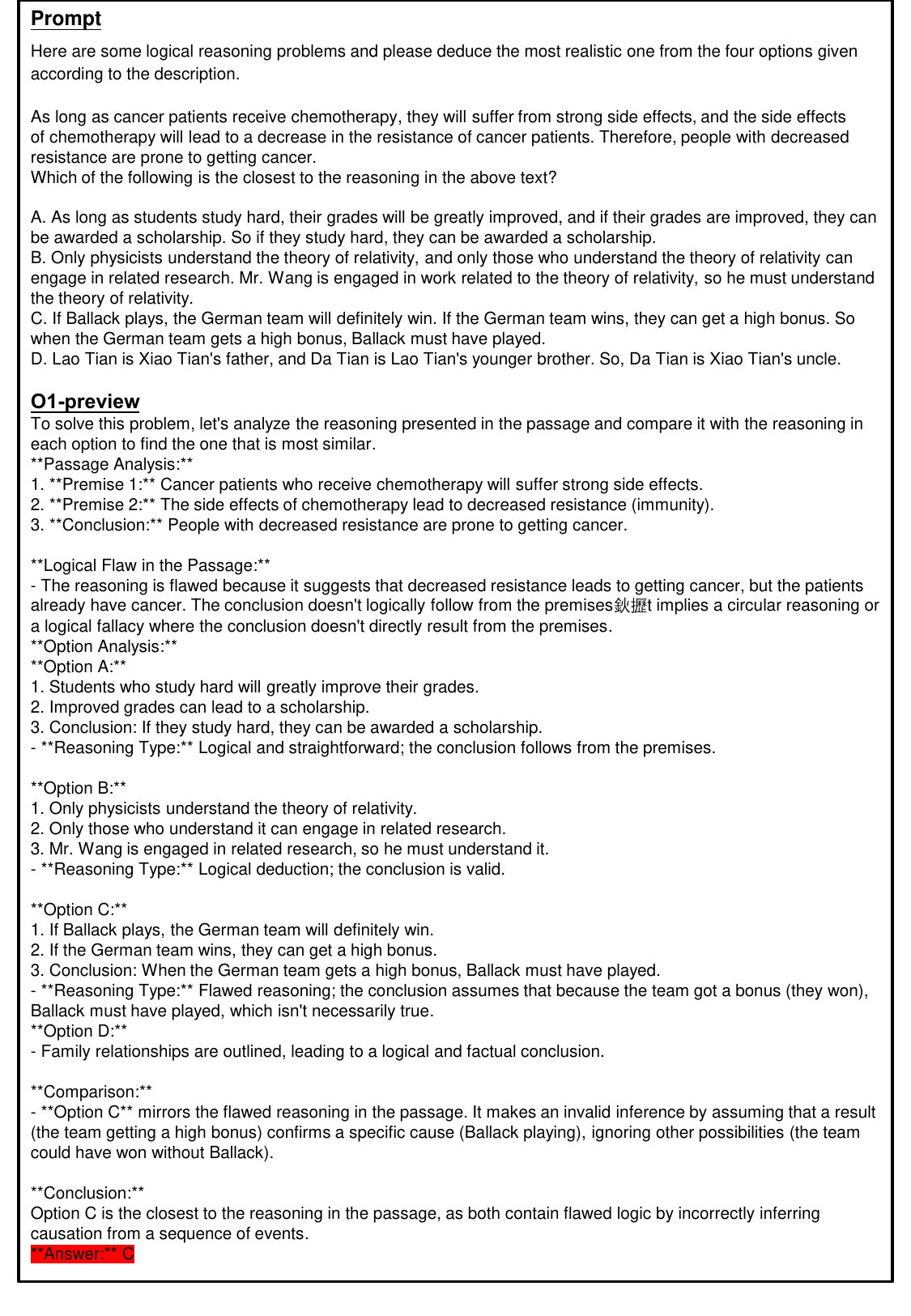}
     \caption[Logical Reasoning: Case 10]{\textbf{Logical Reasoning: Case 10.} It meticulously analyzes and compares the reasoning of each option, elaborating that Option A is logically reasonable, Option B's logical deduction is effective, Option C has a similar defect to the question stem, and Option D is a clear conclusion about family relationships. Through rigorous comparison, it finds Option C, which is closest to the question stem. The entire analysis process is logically strict and well-structured, providing users with problem-solving ideas and accurate answers.} 
     \label{fig:Logical Reasoning: Case 10}
 \end{figure}
  \newpage

\subsection{High School Level Math Competition}
\begin{figure}[H]
    \centering
    \includegraphics[width = \textwidth]{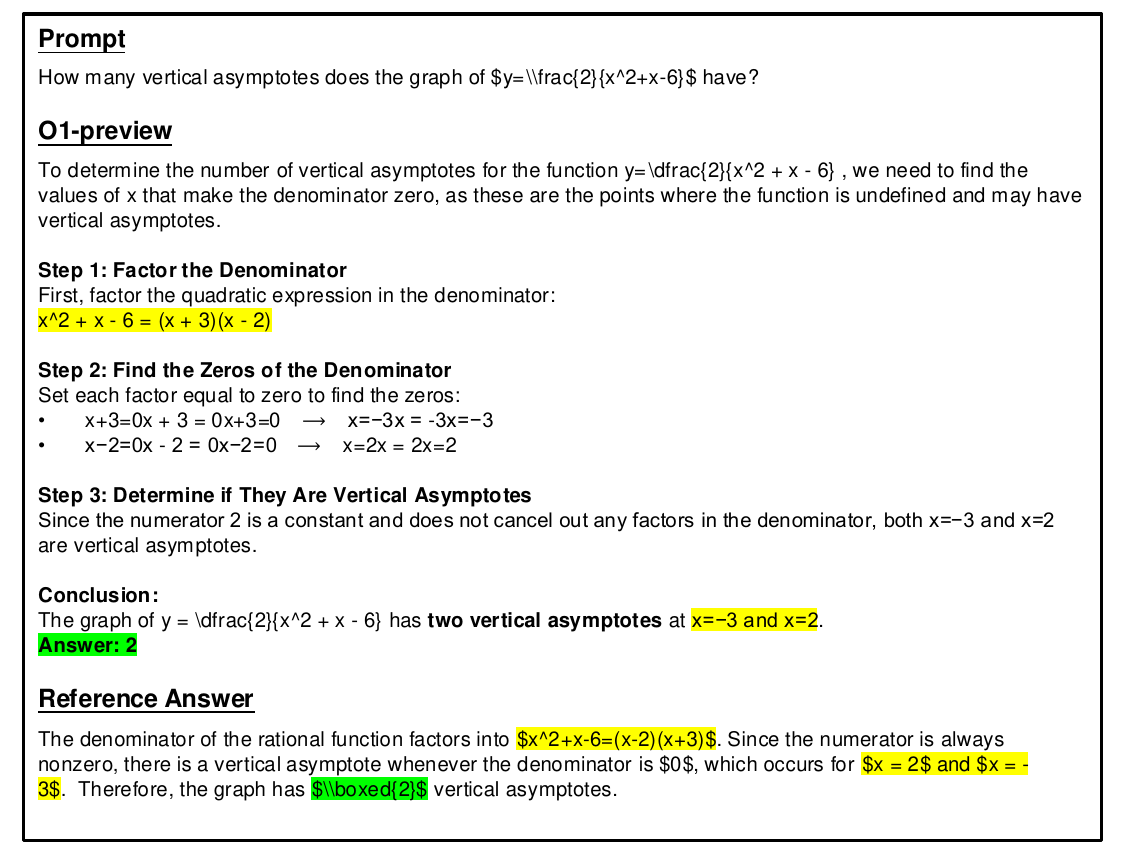}
    \caption[High School Level Math Competition: Case 5]{\textbf{High School Level Math Competition: Case 5: Algebra} This level 3 algebra problem involves determining the number of vertical asymptotes for the rational function. The yellow highlights indicate shared steps in both the o1-preview output and the reference answer, while the green highlights mark the shared final answer. }
    \label{fig:high_school_math_5}
\end{figure}
\newpage

\begin{figure}[H]
    \centering
    \includegraphics[width = \textwidth]{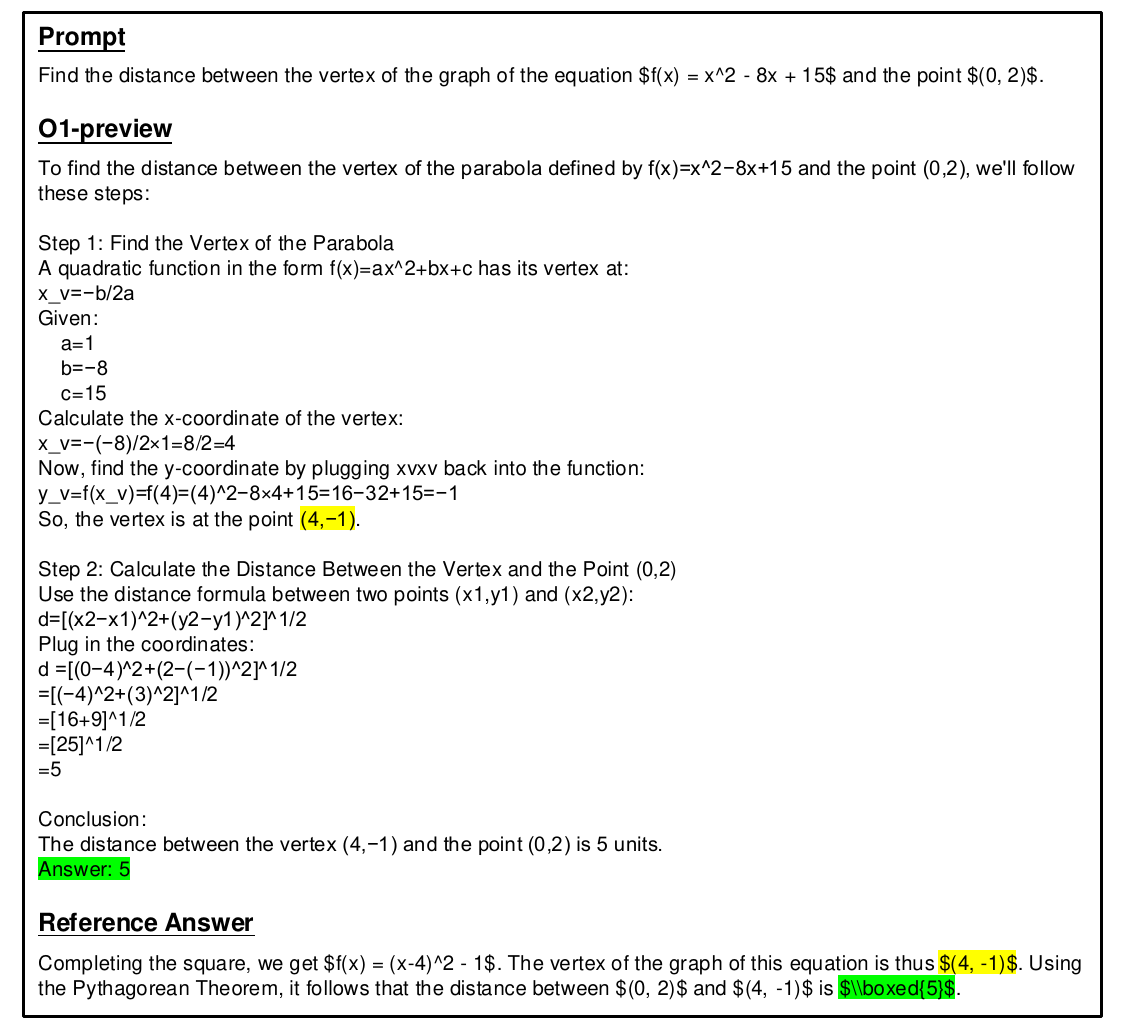}
    \caption[High School Level Math Competition: Case 6]{\textbf{High School Level Math Competition: Case 6: Algebra} This level 4 algebra problem involves determining the distance between the vertex of a parabola and a given point. The yellow highlights indicate shared steps in both the o1-preview output and the reference answer, while the green highlight marks the shared final answer. }
    \label{fig:high_school_math_6}
\end{figure}
\newpage

\begin{figure}[H]
    \centering
    \includegraphics[width = \textwidth]{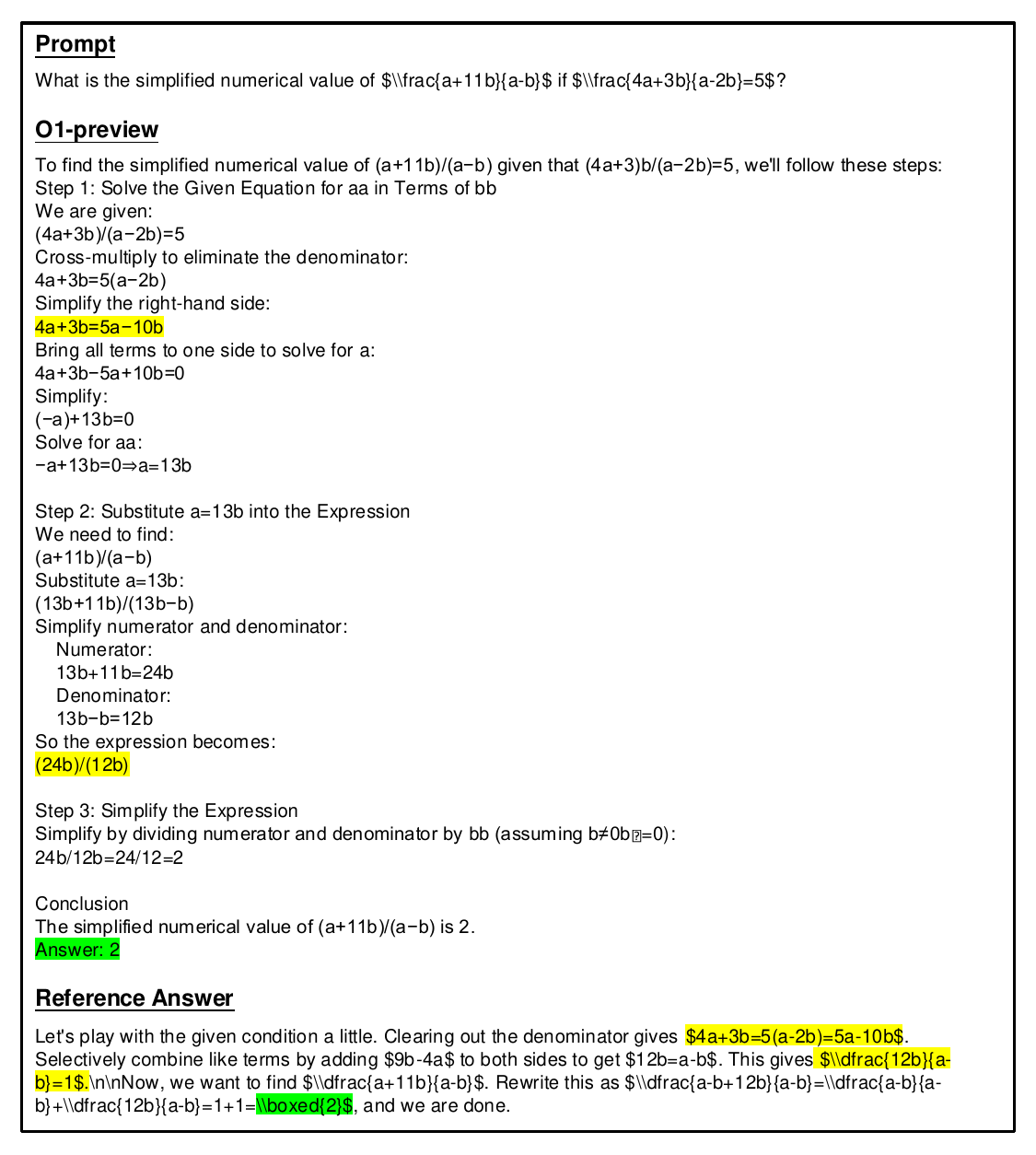}
    \caption[High School Level Math Competition: Case 7]{\textbf{High School Level Math Competition: Case 7: Algebra} This level 4 algebra problem requires simplifying a given expression based on a relationship between two variables. The yellow highlights indicate the shared steps in both the o1-preview output and the reference answer. }
    \label{fig:high_school_math_7}
\end{figure}
\newpage

\begin{figure}[H]
    \centering
    \includegraphics[width = \textwidth]{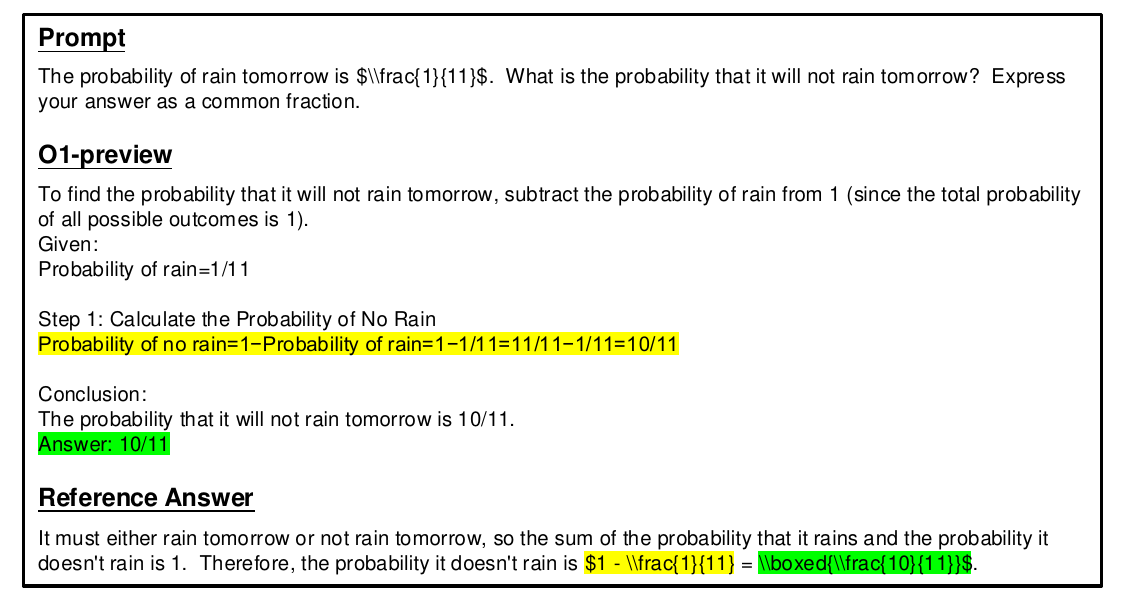}
    \caption[High School Level Math Competition: Case 8]{\textbf{High School Level Math Competition: Case 8: Counting and Probability} This level 1 probability problem involves calculating the likelihood that it will not rain tomorrow, given the probability of rain. The yellow highlights indicate shared steps in both the o1-preview output and the reference answer, with the green highlight marking the correct final answer. }
    \label{fig:high_school_math_8}
\end{figure}
\newpage

\begin{figure}[H]
    \centering
    \includegraphics[width = \textwidth]{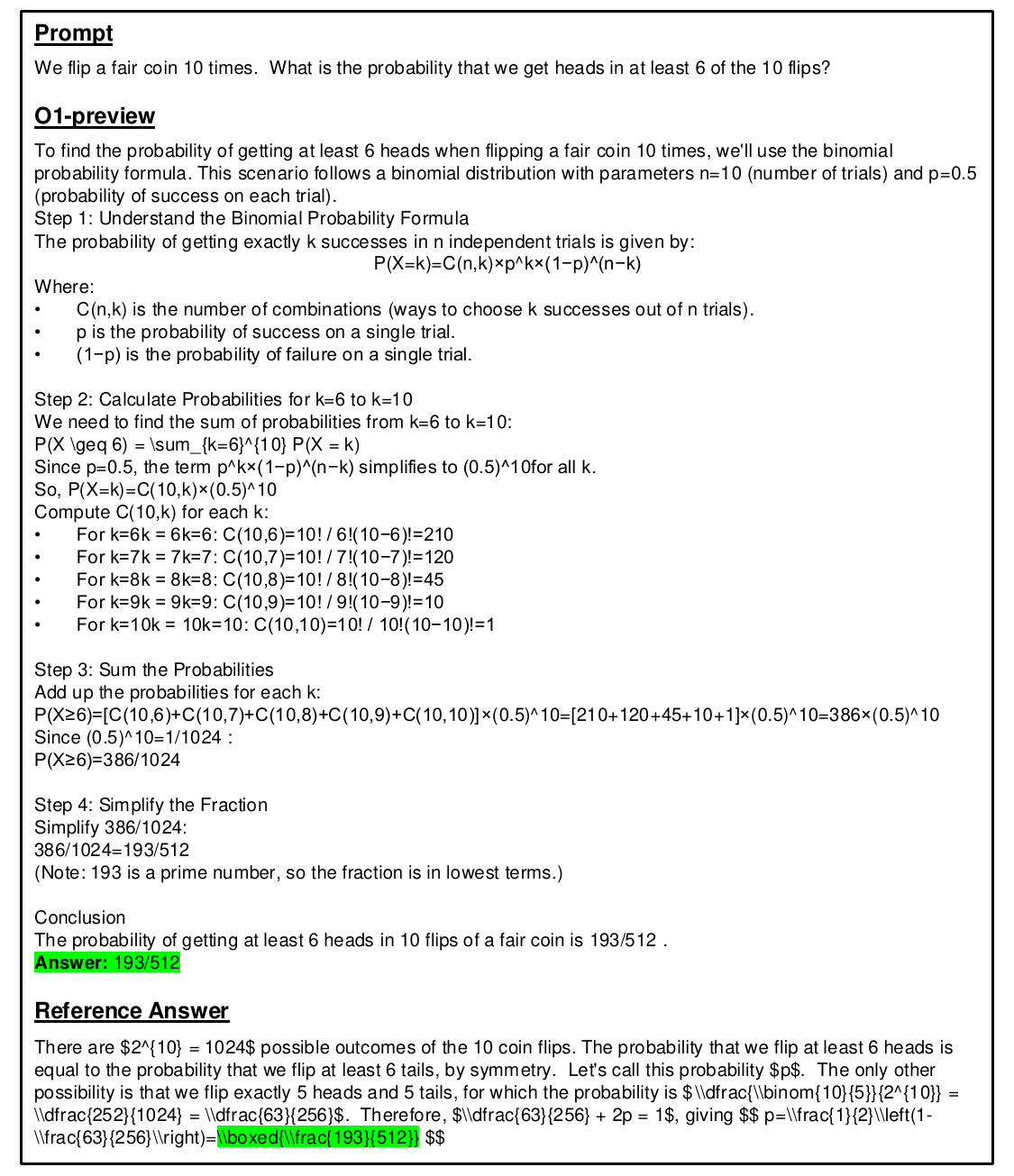}
    \caption[High School Level Math Competition: Case 9]{\textbf{High School Level Math Competition: Case 9: Counting and Probability} This level 4 probability problem involves finding the likelihood of getting heads in at least 6 out of 10 coin flips. The yellow highlights indicate the shared steps in both the o1-preview output and the reference answer, while the green highlights mark the shared final answer. }
    \label{fig:high_school_math_9}
\end{figure}
\newpage

\begin{figure}[H]
    \centering
    \includegraphics[width =.98 \textwidth]{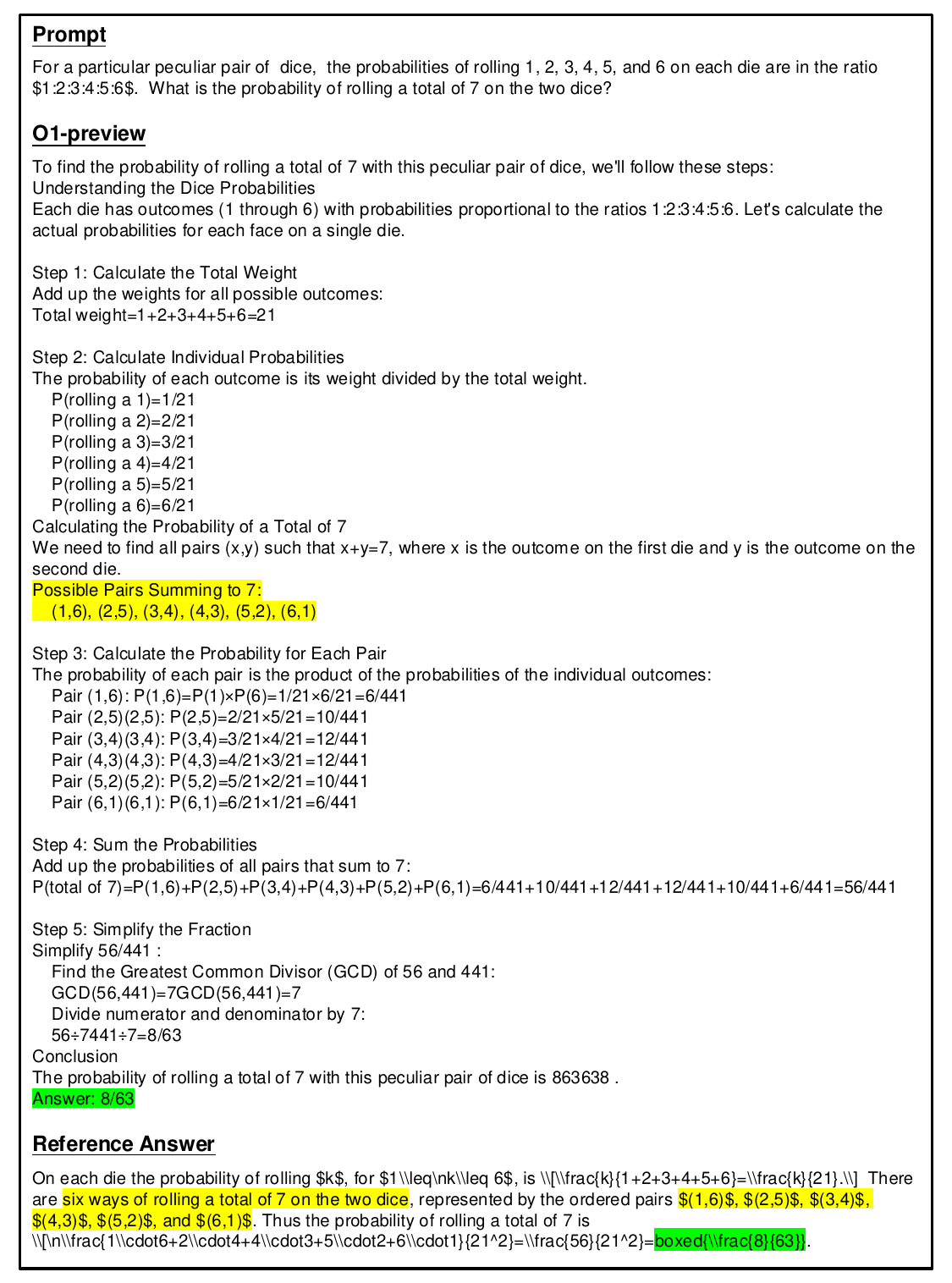}
    \caption[High School Level Math Competition: Case 10]{\textbf{High School Level Math Competition: Case 10: Counting and Probability} This is a level 4 probability problem.}
    \label{fig:high_school_math_10}
\end{figure}


\subsection{College-level Math Problems}
\label{appendix:college level math}

\fig{width=0.9\textwidth}{Q2_o1}{College-Level Math: Case 2}{A demonstration case of solving a basic discrete math problem. The generated answer is manually evaluated as correct.}
\newpage

\fig{width=0.9\textwidth}{Q3_o1}{College-Level Math: Case 3}{A demonstration case of solving an advanced discrete math problem. The generated answer is manually evaluated as incorrect.}
\newpage

\fig{width=0.9\textwidth}{Q4_o1}{College-Level Math: Case 4}{A demonstration case of solving an advanced discrete math problem. The generated answer is manually evaluated as incorrect.}
\newpage

\fig{width=0.9\textwidth}{Q6_o1}{College-Level Math: Case 6}{A demonstration case of solving a Calculus problem. The generated answer is manually evaluated as correct.}
\newpage

\fig{width=0.9\textwidth}{Q8_o1}{College-Level Math: Case 8}{A demonstration case of o1-preview reproducing the proof of an advanced math theorem. The generated answer is manually evaluated as correct.}
\newpage

\fig{width=0.9\textwidth}{Q8.5_o1}{College-Level Math: Case 8.5}{A demonstration case of o1-preview reproducing the proof of an advanced math theorem. The generated answer is manually evaluated as correct.}
\newpage

\fig{width=0.9\textwidth}{Q9_o1}{College-Level Math: Case 9}{A demonstration case of o1-preview reproducing the proof of an advanced math theorem. The generated answer is manually evaluated as incorrect.}
\newpage

\fig{width=0.9\textwidth}{Q10_o1}{College-Level Math: Case 10}{A demonstration case of o1-preview reproducing the proof of an advanced math theorem. The generated answer is manually evaluated as correct.}
\newpage

\fig{width=0.9\textwidth}{Q10.5_o1}{College-Level Math: Case 10.5}{A demonstration case of o1-preview reproducing the proof of an advanced math theorem. The generated answer is manually evaluated as correct.}
\newpage


\subsection{Electronic Health Record Diagnosis}
\label{appendix:EHR}
\begin{figure}[H]
    \centering
    \includegraphics[width = 0.82\linewidth]{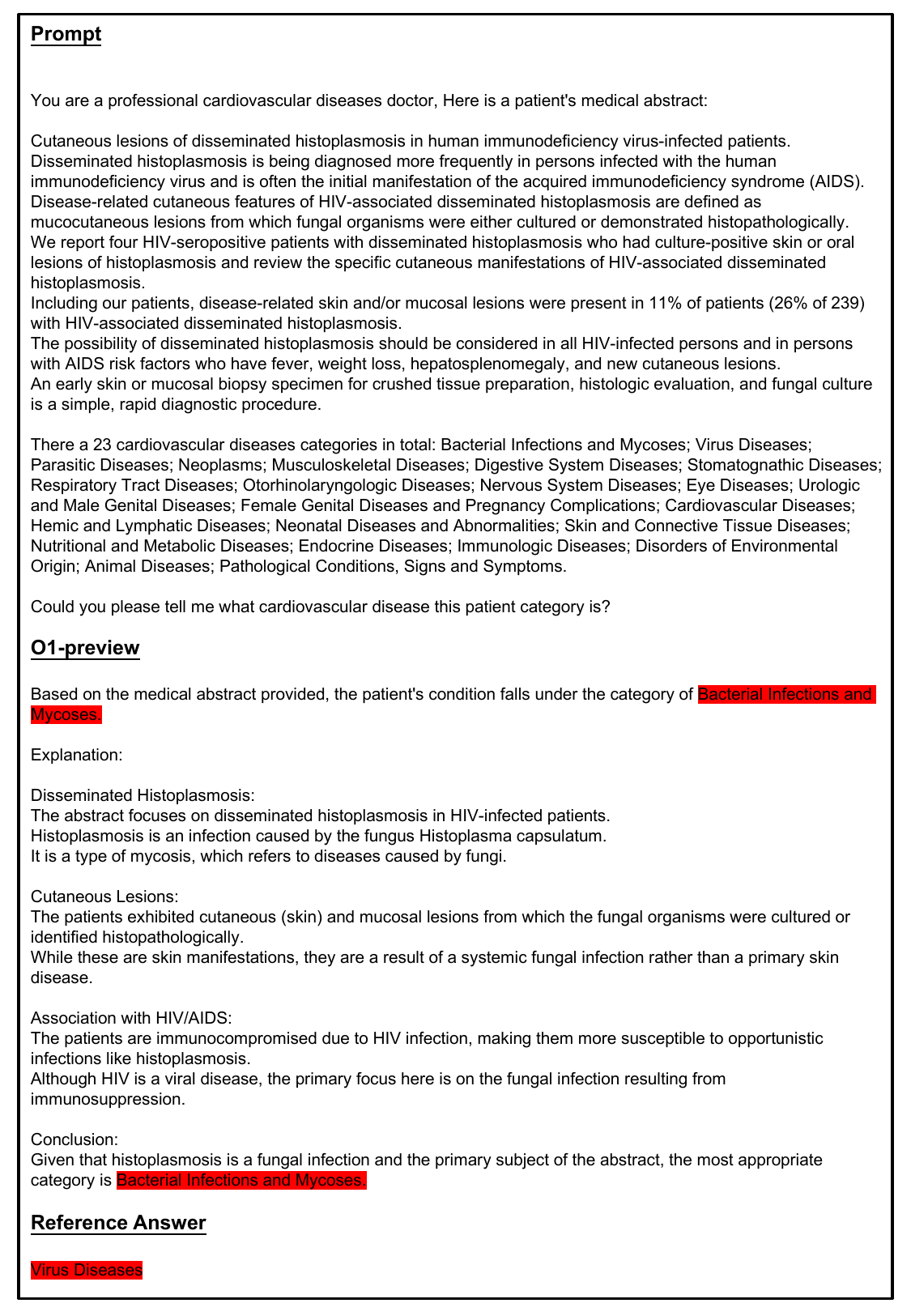}
    \caption[Electronic Health Record Diagnosis: Case 5.]
    {\textbf{Electronic Health Record Diagnosis: Case 5.} One specific examples of a demonstration cases for EHR diagnosis involves the use of the o1-preview model. Red indicates that the model's output is completely inconsistent with the reference diagnosis, while green signifies complete consistency.}
    \label{fig:EHR-3}
\end{figure}
\newpage

\begin{figure}[H]
    \centering
    \includegraphics[width = 0.82\linewidth]{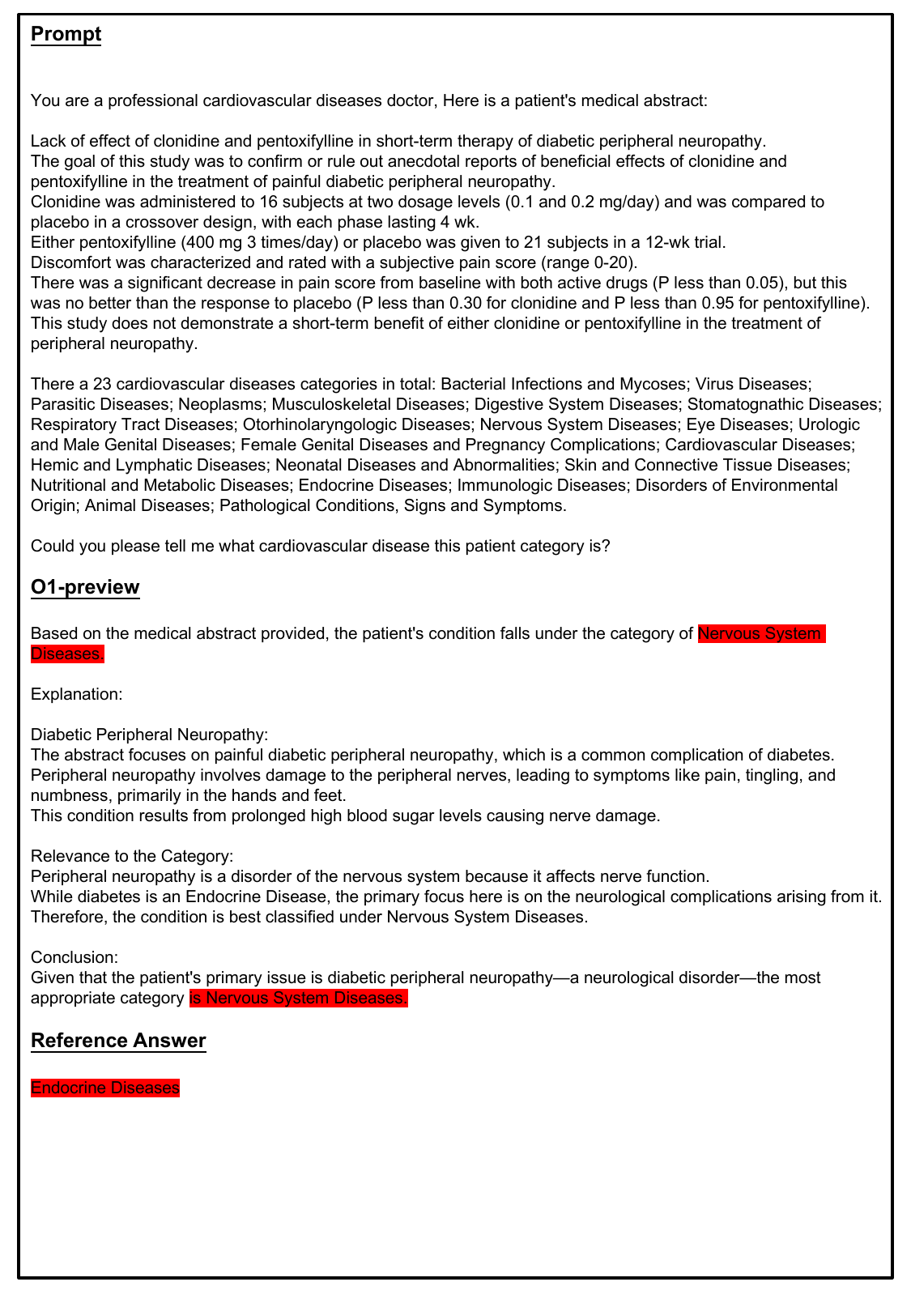}
    \caption[Electronic Health Record Diagnosis: Case 6.]
    {\textbf{Electronic Health Record Diagnosis: Case 6.} One specific examples of a demonstration cases for EHR diagnosis involves the use of the o1-preview model. Red indicates that the model's output is completely inconsistent with the reference diagnosis, while green signifies complete consistency.}
    \label{fig:EHR-6}
\end{figure}
\newpage

\begin{figure}[H]
    \centering
    \includegraphics[width = 0.82\linewidth]{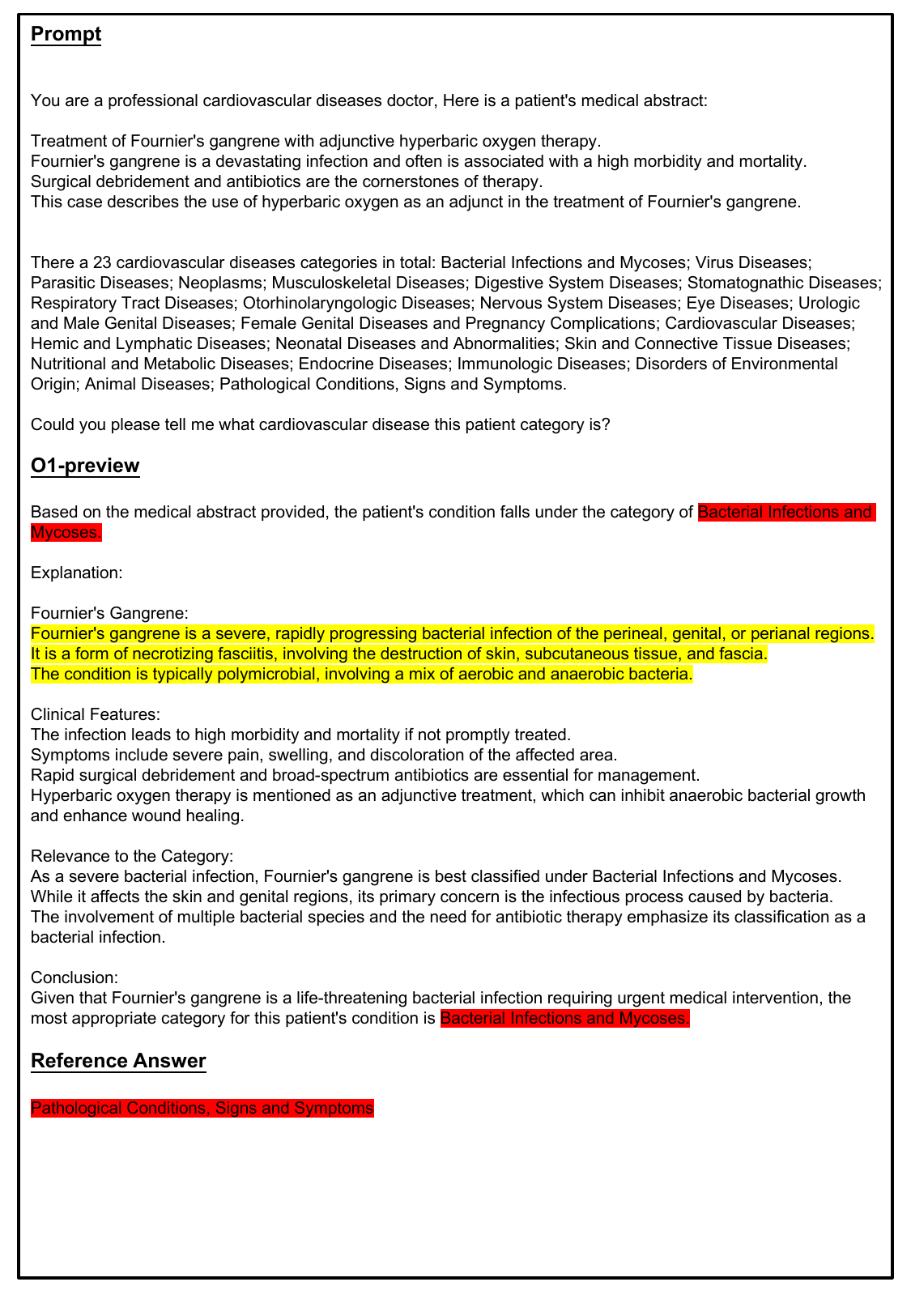}
    \caption[Electronic Health Record Diagnosis: Case 7.]
    {\textbf{Electronic Health Record Diagnosis: Case 7.} One specific examples of a demonstration cases for EHR diagnosis involves the use of the o1-preview model. Red indicates that the model's output is completely inconsistent with the reference diagnosis, while green signifies complete consistency.}
    \label{fig:EHR-7}
\end{figure}
\newpage

\begin{figure}[H]
    \centering
    \includegraphics[width = 0.82\linewidth]{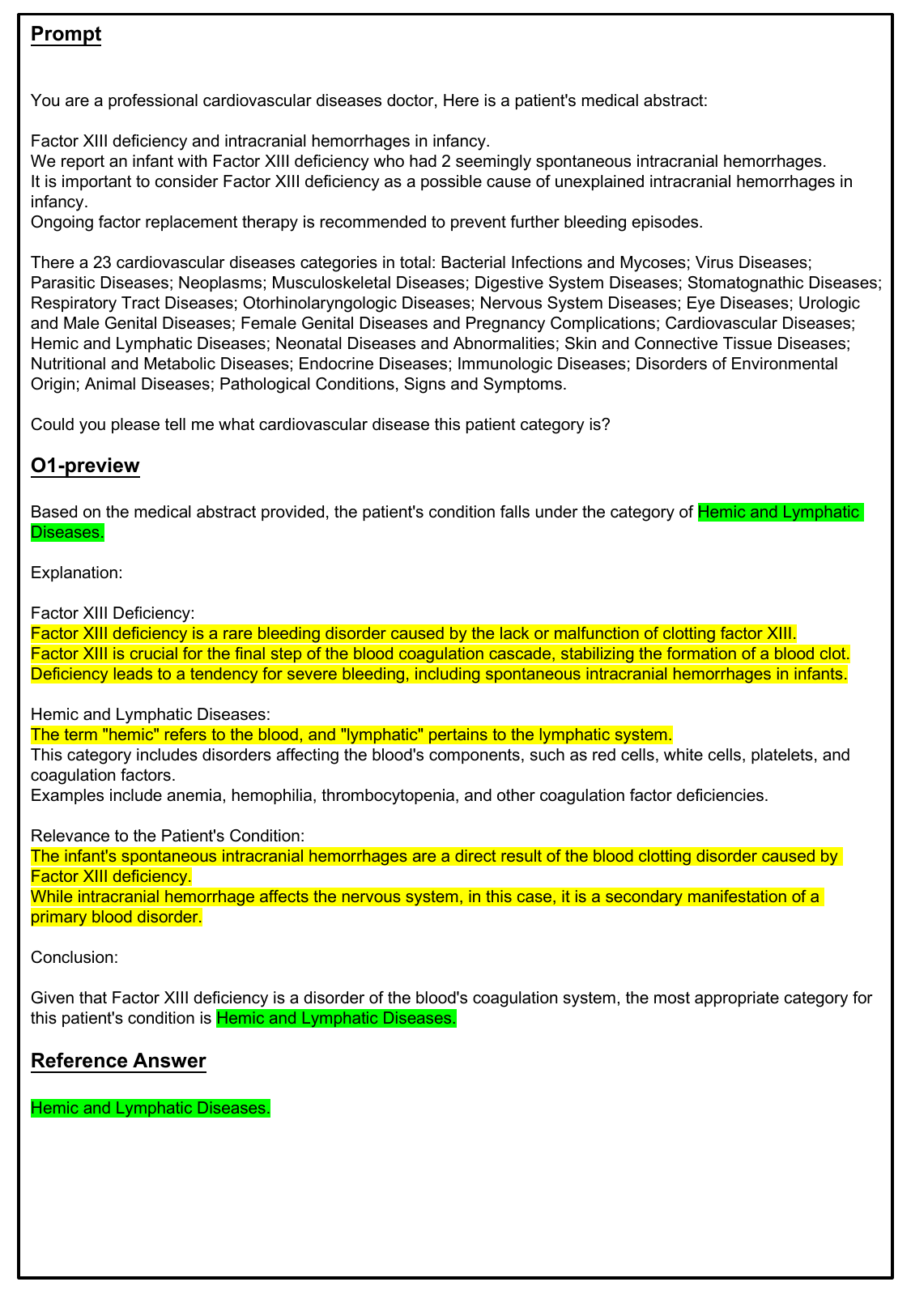}
    \caption[Electronic Health Record Diagnosis: Case 8.]
    {\textbf{Electronic Health Record Diagnosis: Case 8.} One specific examples of a demonstration cases for EHR diagnosis involves the use of the o1-preview model. Red indicates that the model's output is completely inconsistent with the reference diagnosis, while green signifies complete consistency.}
    \label{fig:EHR-8}
\end{figure}
\newpage

\begin{figure}[H]
    \centering
    \includegraphics[width = 0.82\linewidth]{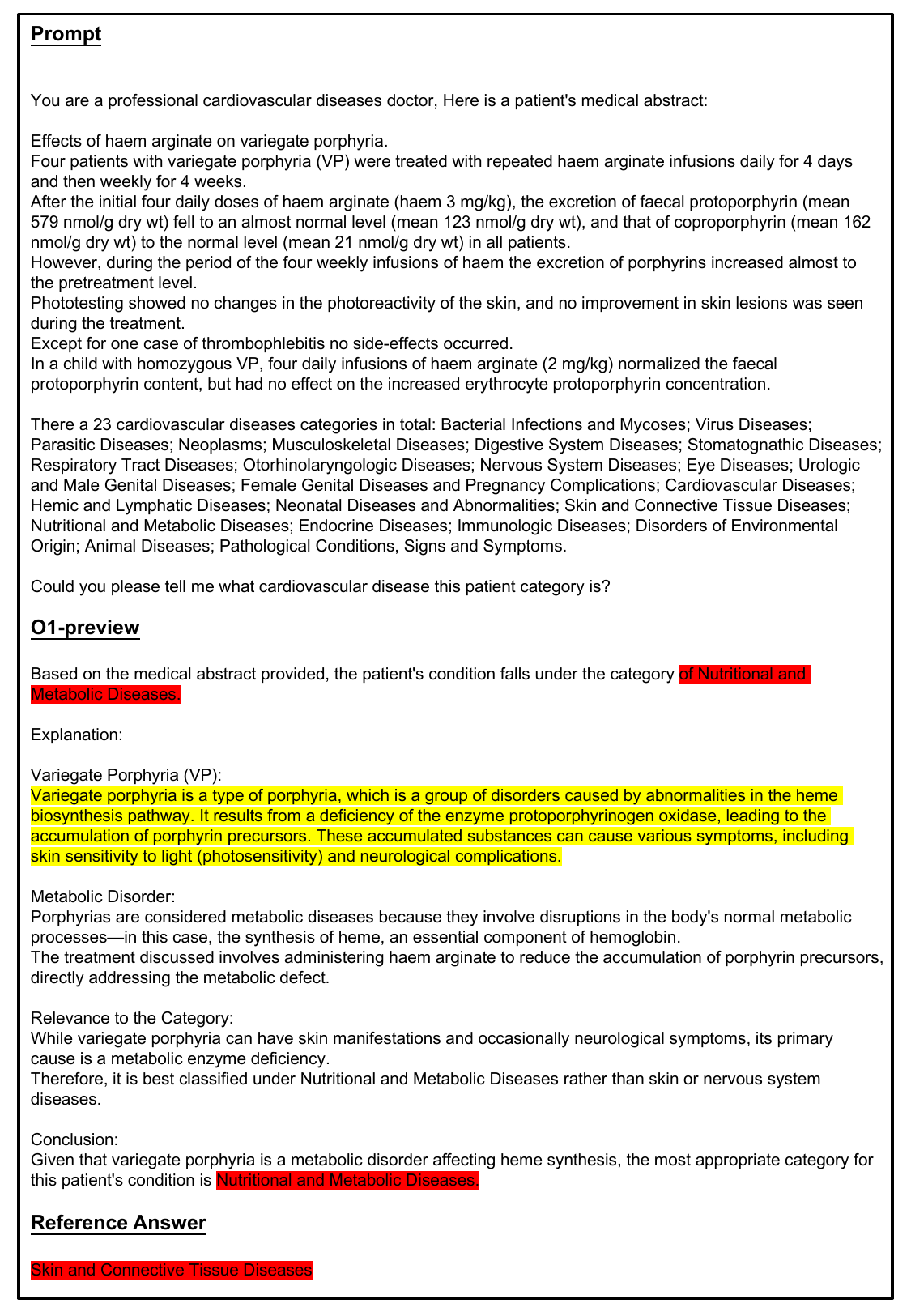}
    \caption[Electronic Health Record Diagnosis: Case 9.]
    {\textbf{Electronic Health Record Diagnosis: Case 9.} One specific examples of a demonstration cases for EHR diagnosis involves the use of the o1-preview model. Red indicates that the model's output is completely inconsistent with the reference diagnosis, while green signifies complete consistency.}
    \label{fig:EHR-9}
\end{figure}
\newpage

\begin{figure}[H]
    \centering
    \includegraphics[width = 0.82\linewidth]{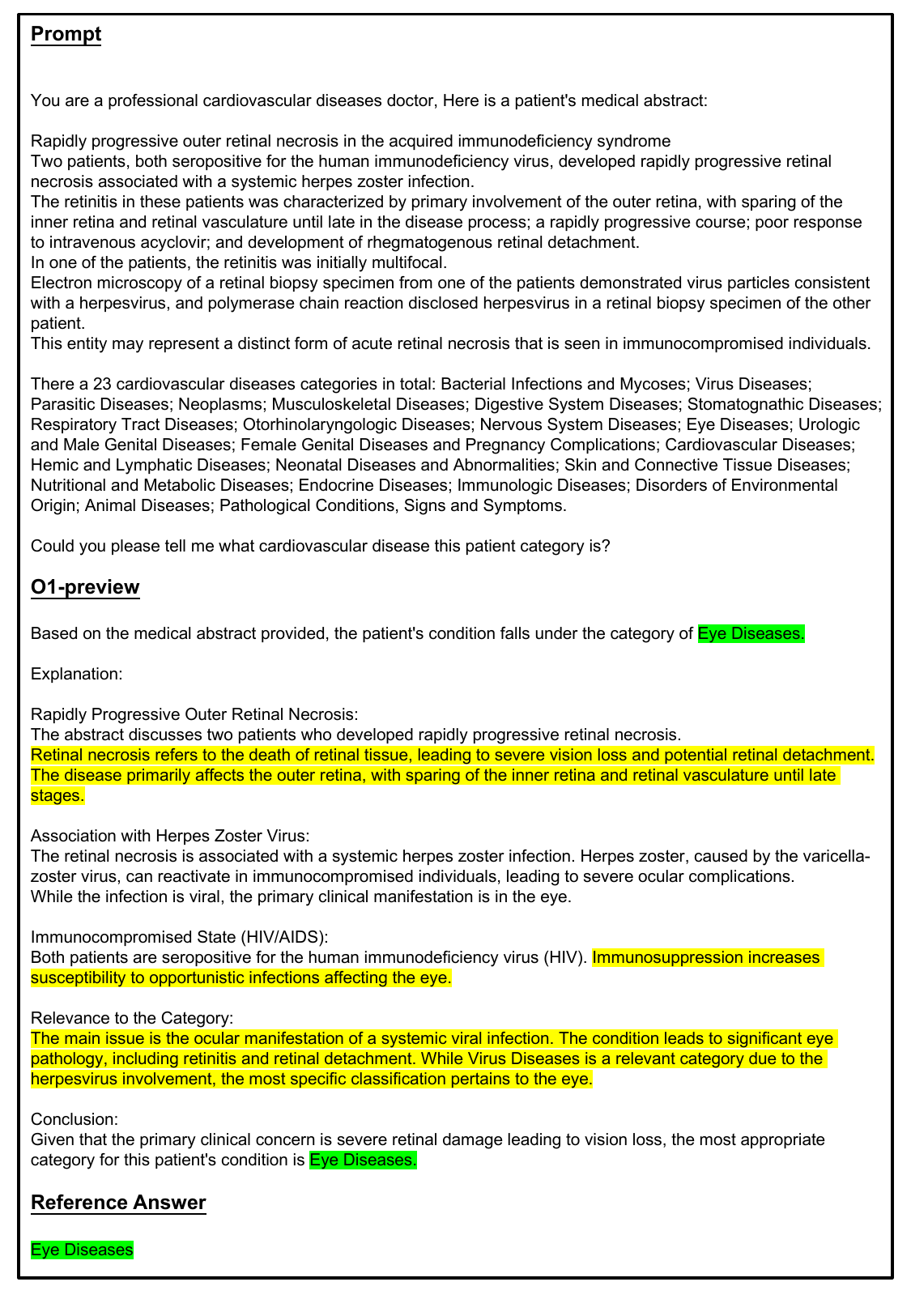}
    \caption[Electronic Health Record Diagnosis: Case 10.]
    {\textbf{Electronic Health Record Diagnosis: Case 10.} One specific examples of a demonstration cases for EHR diagnosis involves the use of the o1-preview model. Red indicates that the model's output is completely inconsistent with the reference diagnosis, while green signifies complete consistency.}
    \label{fig:EHR-10}
\end{figure}


\subsection{Stochastic Processes in Statistics}
\label{appendix:Stochastic}

\begin{figure}[H]
    \centering
    \includegraphics[width = .9\textwidth]{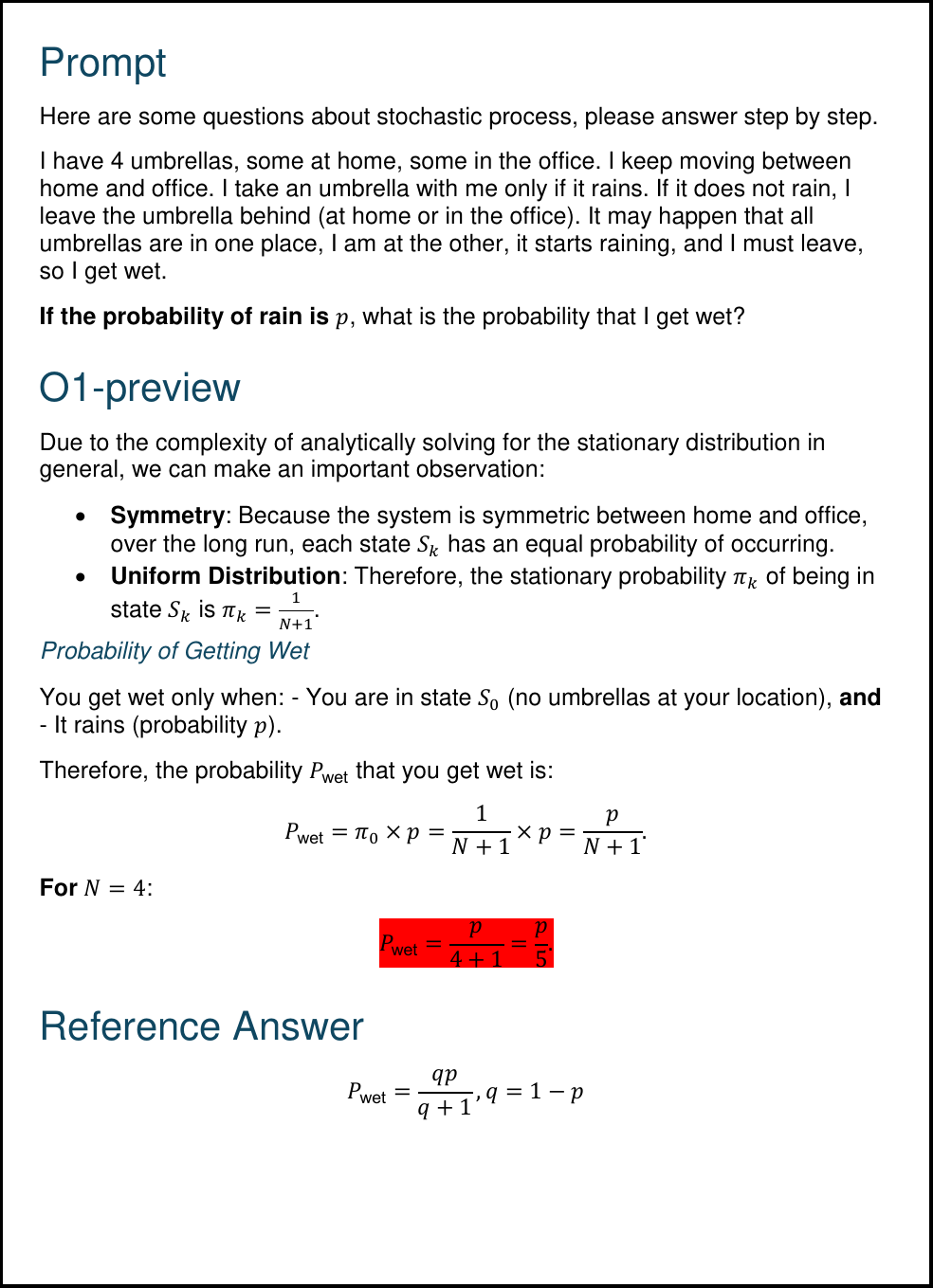}
    \caption[Stochastic Processes in Statistics: Case 3]{\textbf{Stochastic Processes in Statistics: Case 3.} A demonstration case of Stochastic Process problem solving. The o1-preview successfully solved the problem.}
    \label{fig:Stochastic_3}
\end{figure}
\newpage

\begin{figure}[H]
    \centering
    \includegraphics[width = .9\textwidth]{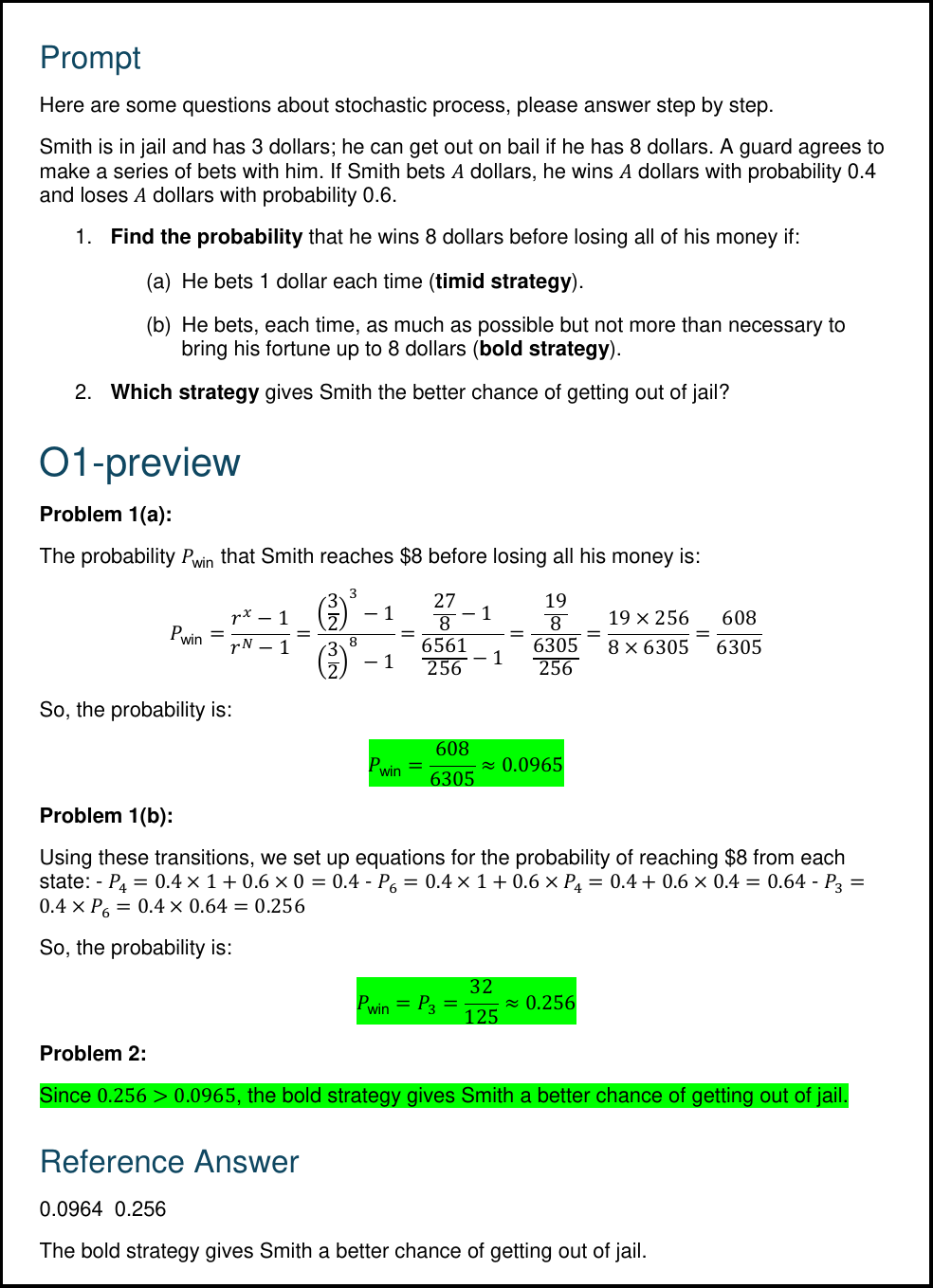}
    \caption[Stochastic Processes in Statistics: Case 4]{\textbf{Stochastic Processes in Statistics: Case 4.} A demonstration case of Stochastic Process problem solving. The o1-preview successfully solved the problem.}
    \label{fig:Stochastic_4}
\end{figure}
\newpage

\begin{figure}[H]
    \centering
    \includegraphics[width = .9\textwidth]{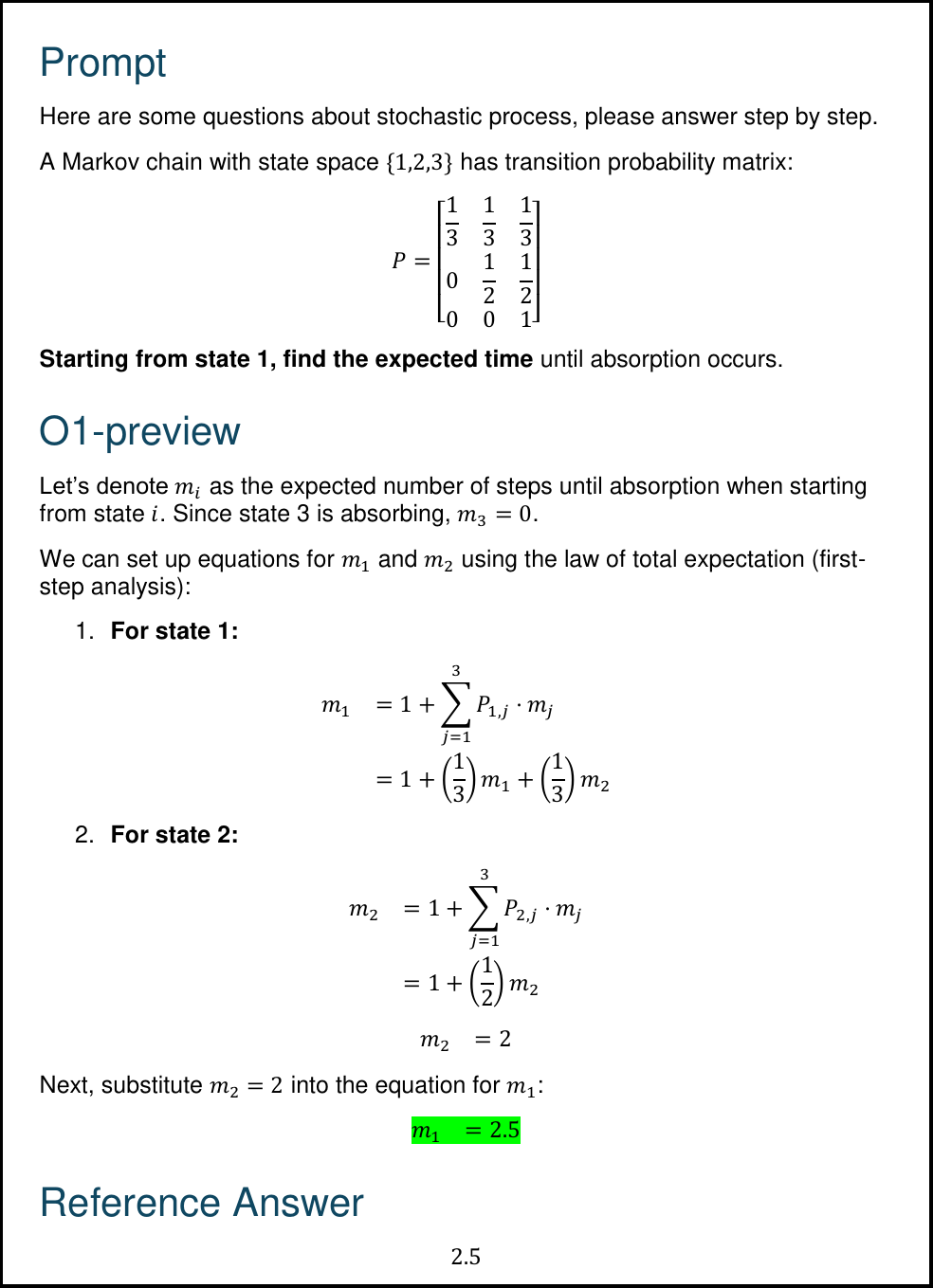}
    \caption[Stochastic Processes in Statistics: Case 5]{\textbf{Stochastic Processes in Statistics: Case 5.} A demonstration case of Stochastic Process problem solving. The o1-preview successfully solved the problem.}
    \label{fig:Stochastic_5}
\end{figure}
\newpage

\begin{figure}[H]
    \centering
    \includegraphics[width = .9\textwidth]{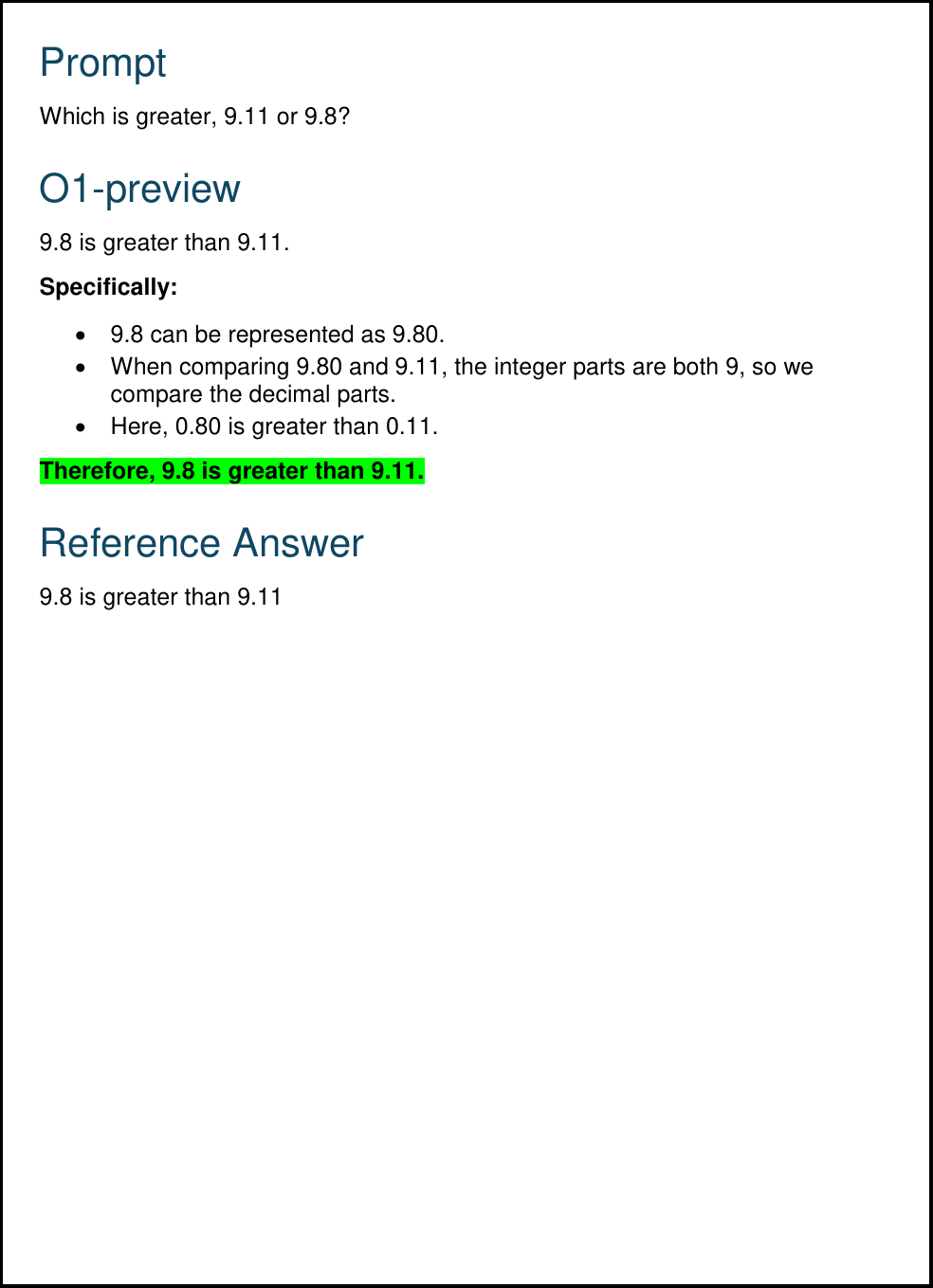}
    \caption[Stochastic Processes in Statistics: Case 6]{\textbf{Stochastic Processes in Statistics: Case 6.} A demonstration case of tricky problem. The o1-preview successfully solved the problem.}
    \label{fig:Stochastic_6}
\end{figure}
\newpage


\subsection{Medical Text Anonymization}

\begin{figure}[H]
    \centering
    \includegraphics[width=0.85\linewidth]{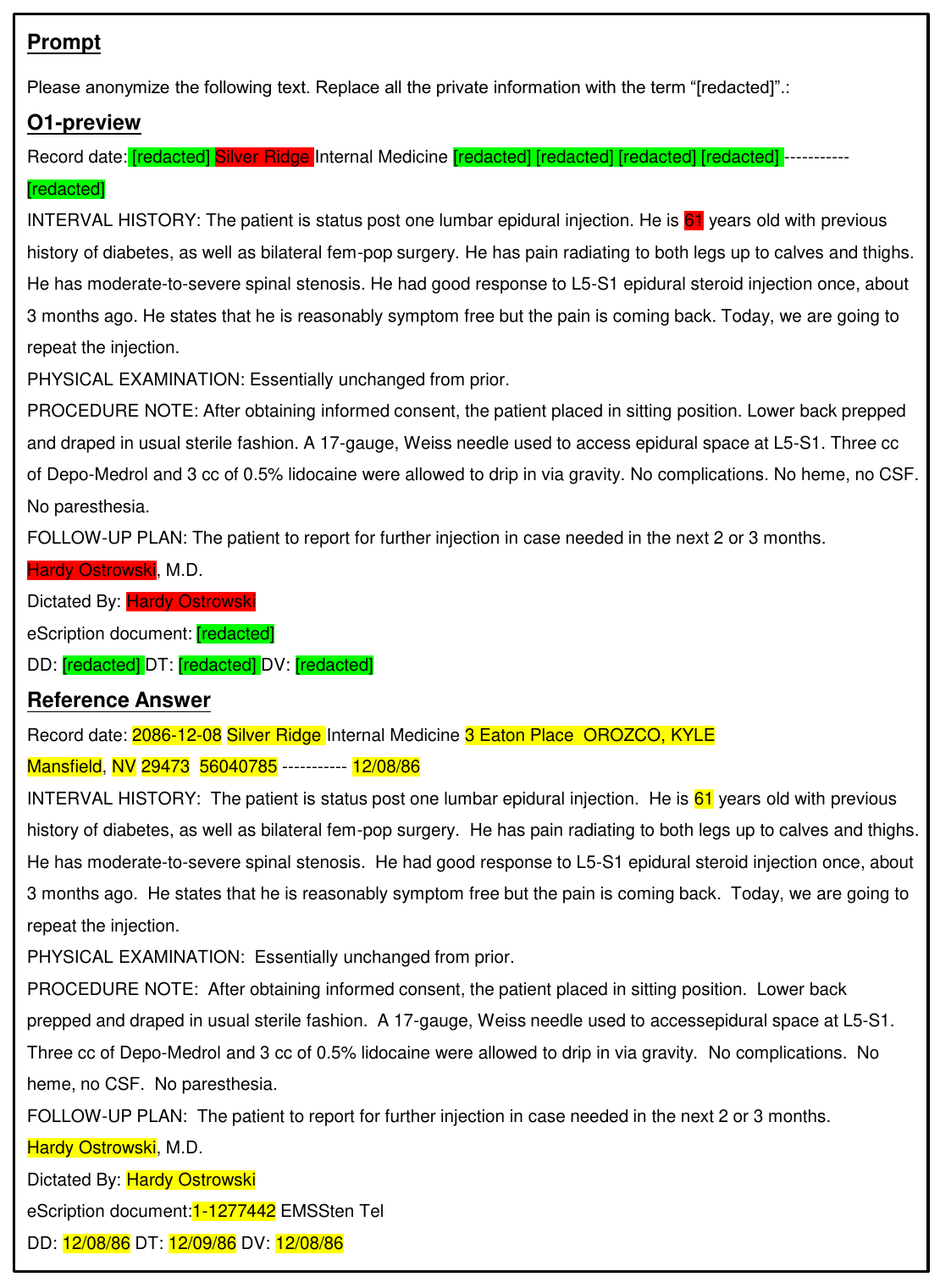}
    \caption[Medical Text Anonymization: Case 4]{\textbf{Medical Text Anonymization: Case 4.} An example showing the capability of o1-preview to anonymize medical texts with the implicit prompt. Yellow highlights the annotations made by human experts. Green marks the sensitive privacy information detected by o1-preview. Red indicates modifications beyond the human-marked annotation. }
    \label{fig:MTA-case2-1}
\end{figure}
\newpage

\begin{figure}[H]
    \centering
    \includegraphics[width=0.85\linewidth]{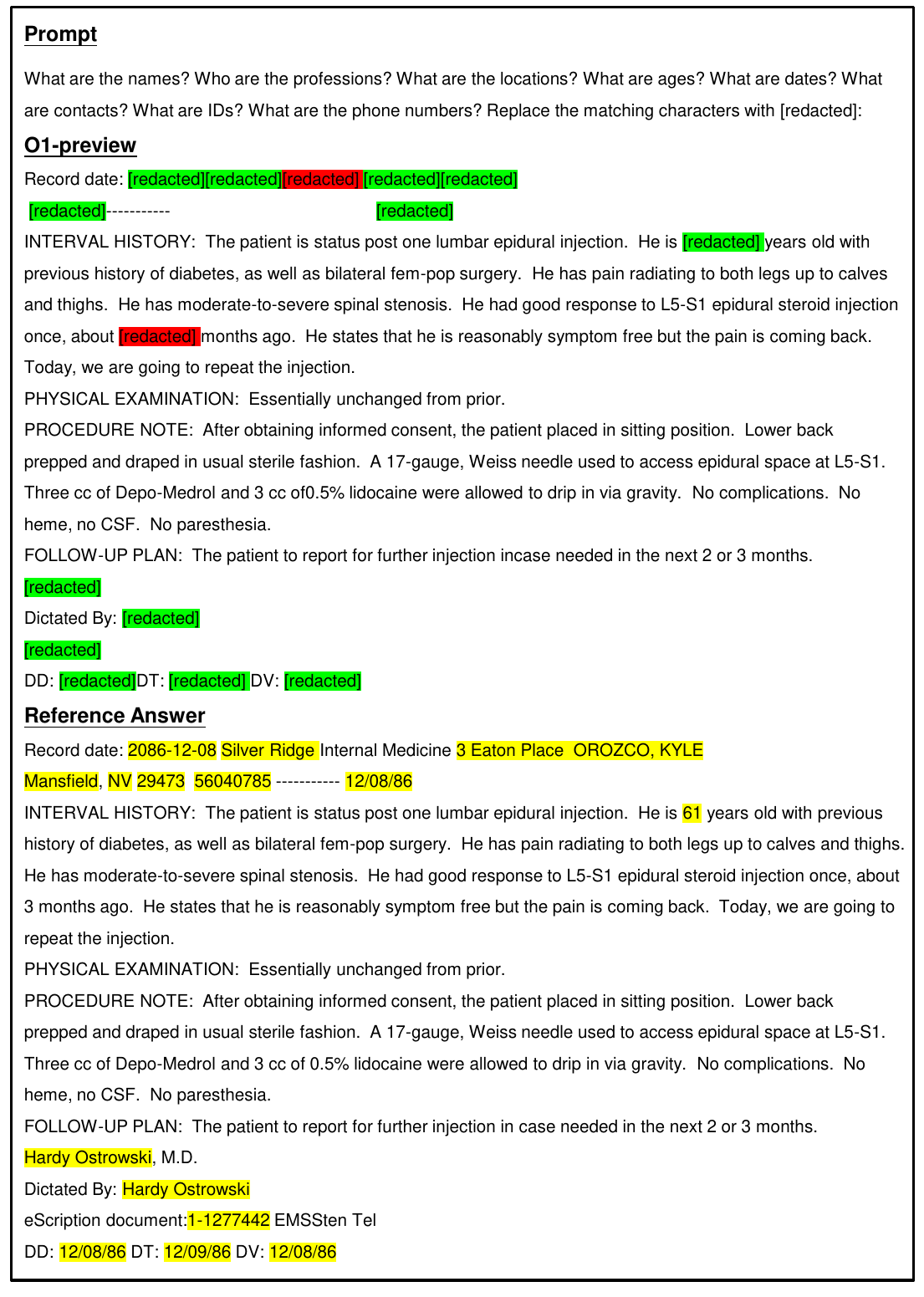}
    \caption[Medical Text Anonymization: Case 5]{\textbf{Medical Text Anonymization: Case 5.} An example showing the capability of o1-preview to anonymize medical texts with the general prompt. Yellow highlights the annotations made by human experts. Green marks the sensitive privacy information detected by o1-preview. Red indicates modifications beyond the human-marked annotation. }
    \label{fig:MTA-case2-2}
\end{figure}
\newpage

\begin{figure}[H]
    \centering
    \includegraphics[width=0.85\linewidth]{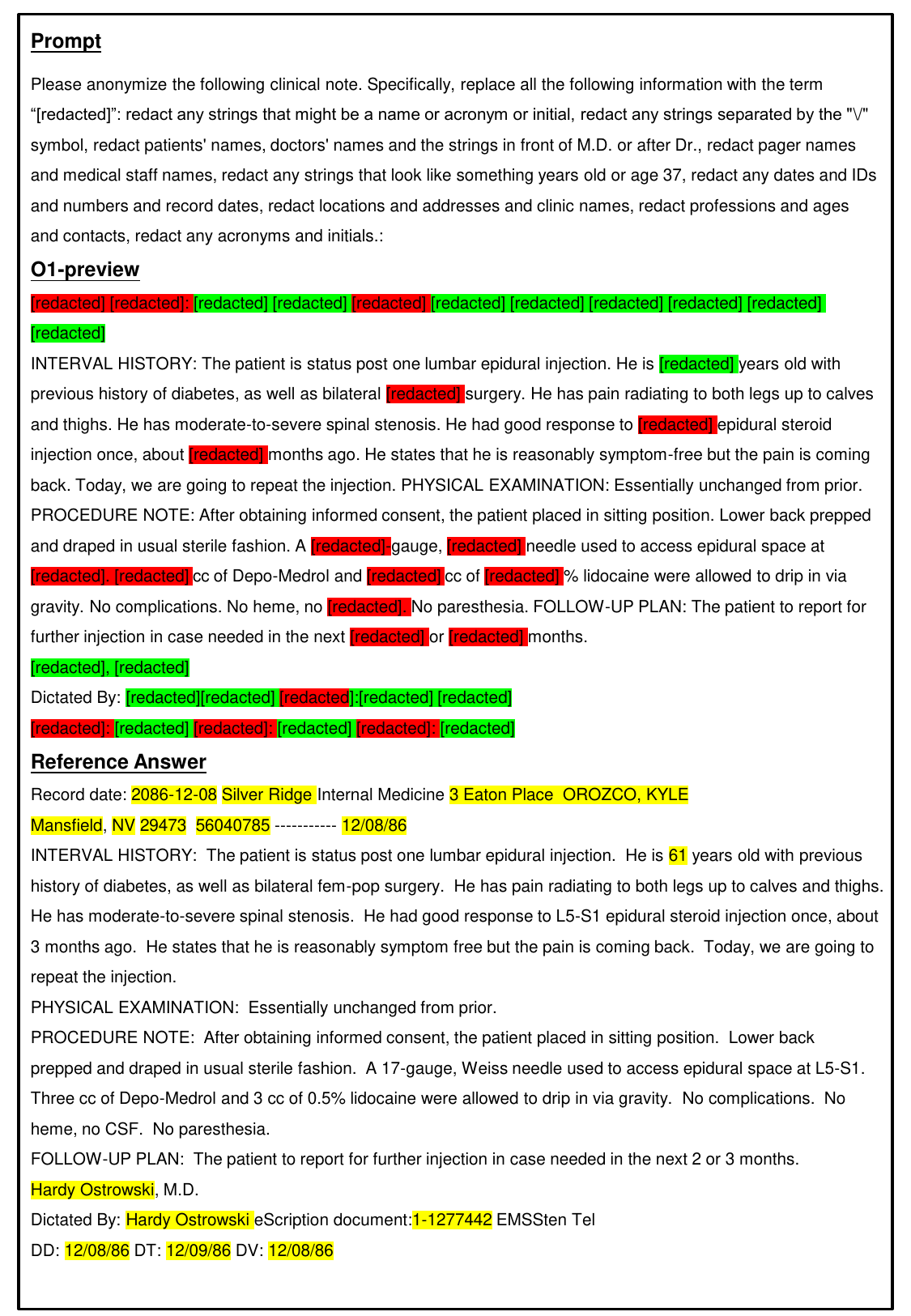}
    \caption[Medical Text Anonymization: Case 6]{\textbf{Medical Text Anonymization: Case 6.} An example showing the capability of o1-preview to anonymize medical texts with the explicit prompt. Yellow highlights the annotations made by human experts. Green marks the sensitive privacy information detected by o1-preview. Red indicates modifications beyond the human-marked annotation. }
    \label{fig:MTA-case2-3}
\end{figure}
\newpage

\begin{figure}[H]
    \centering
    \includegraphics[width=0.85\linewidth]{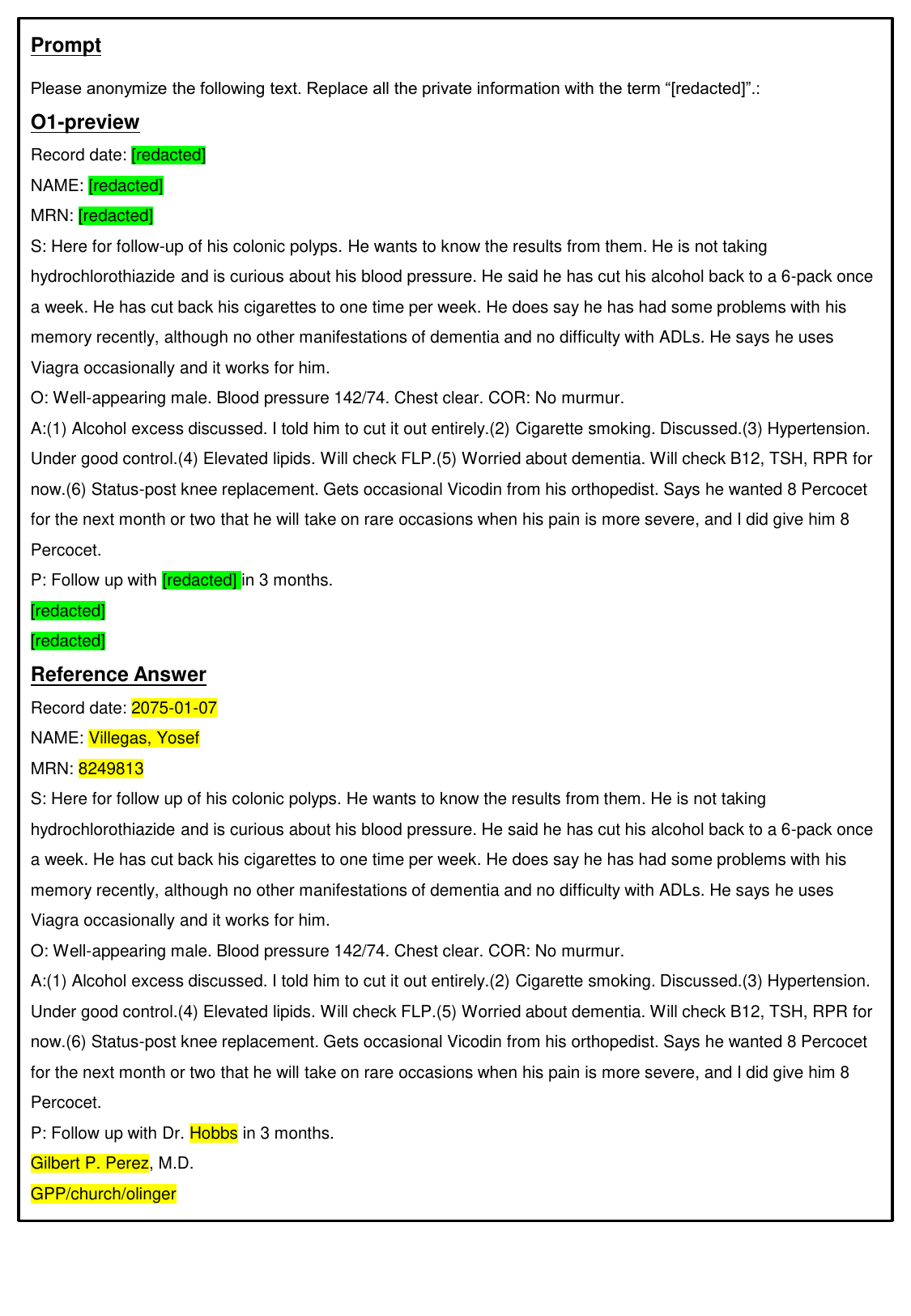}
    \caption[Medical Text Anonymization: Case 7]{\textbf{Medical Text Anonymization: Case 7.} An example showing the capability of o1-preview to anonymize medical texts with the implicit prompt. Yellow highlights the annotations made by human experts. Green marks the sensitive privacy information detected by o1-preview. Red indicates modifications beyond the human-marked annotation. }
    \label{fig:MTA-case3-1}
\end{figure}
\newpage

\begin{figure}[H]
    \centering
    \includegraphics[width=0.85\linewidth]{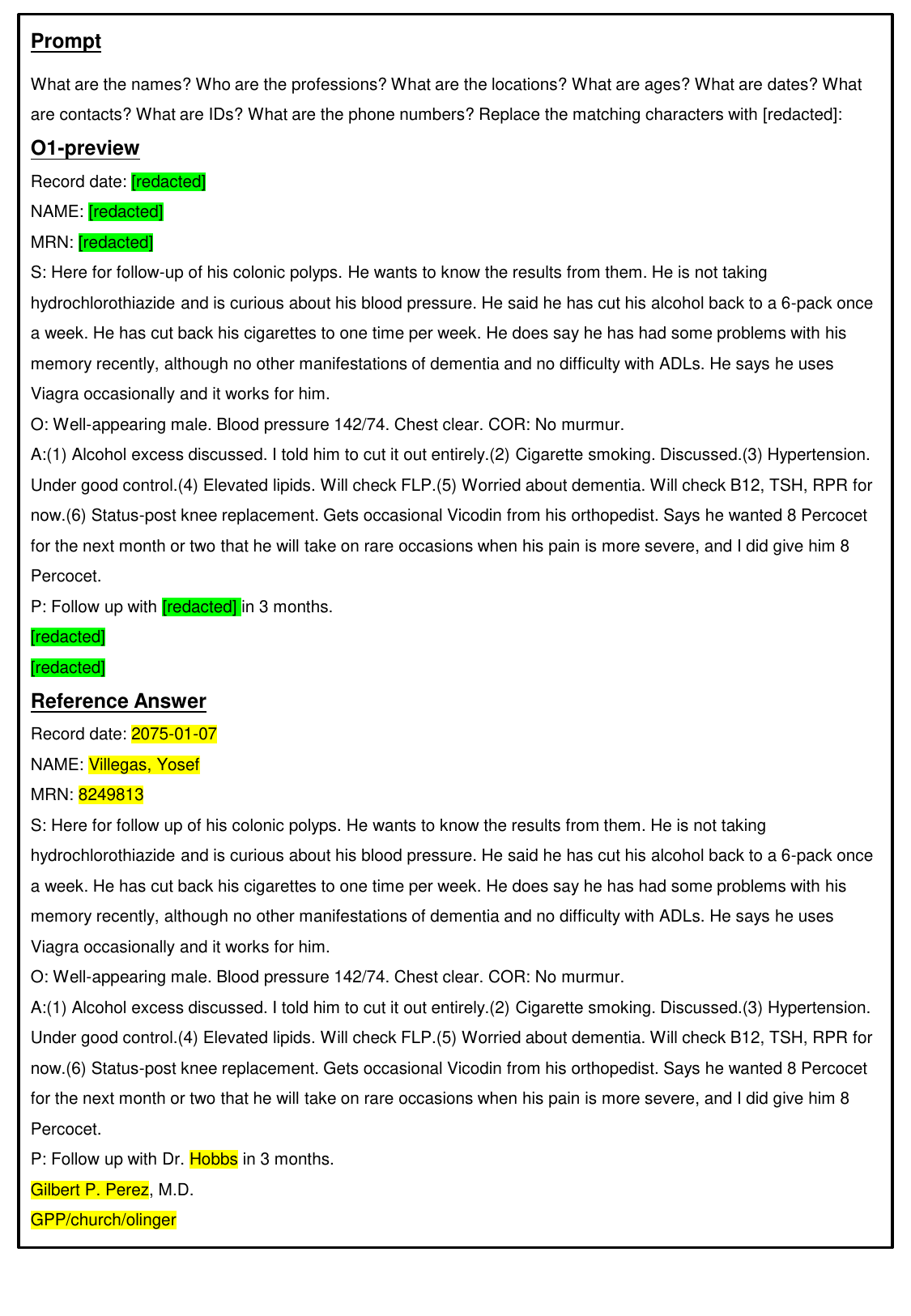}
    \caption[Medical Text Anonymization: Case 8]{\textbf{Medical Text Anonymization: Case 8.} An example showing the capability of o1-preview to anonymize medical texts with the general prompt. Yellow highlights the annotations made by human experts. Green marks the sensitive privacy information detected by o1-preview. Red indicates modifications beyond the human-marked annotation. }
    \label{fig:MTA-case3-2}
\end{figure}
\newpage

\begin{figure}[H]
    \centering
    \includegraphics[width=0.85\linewidth]{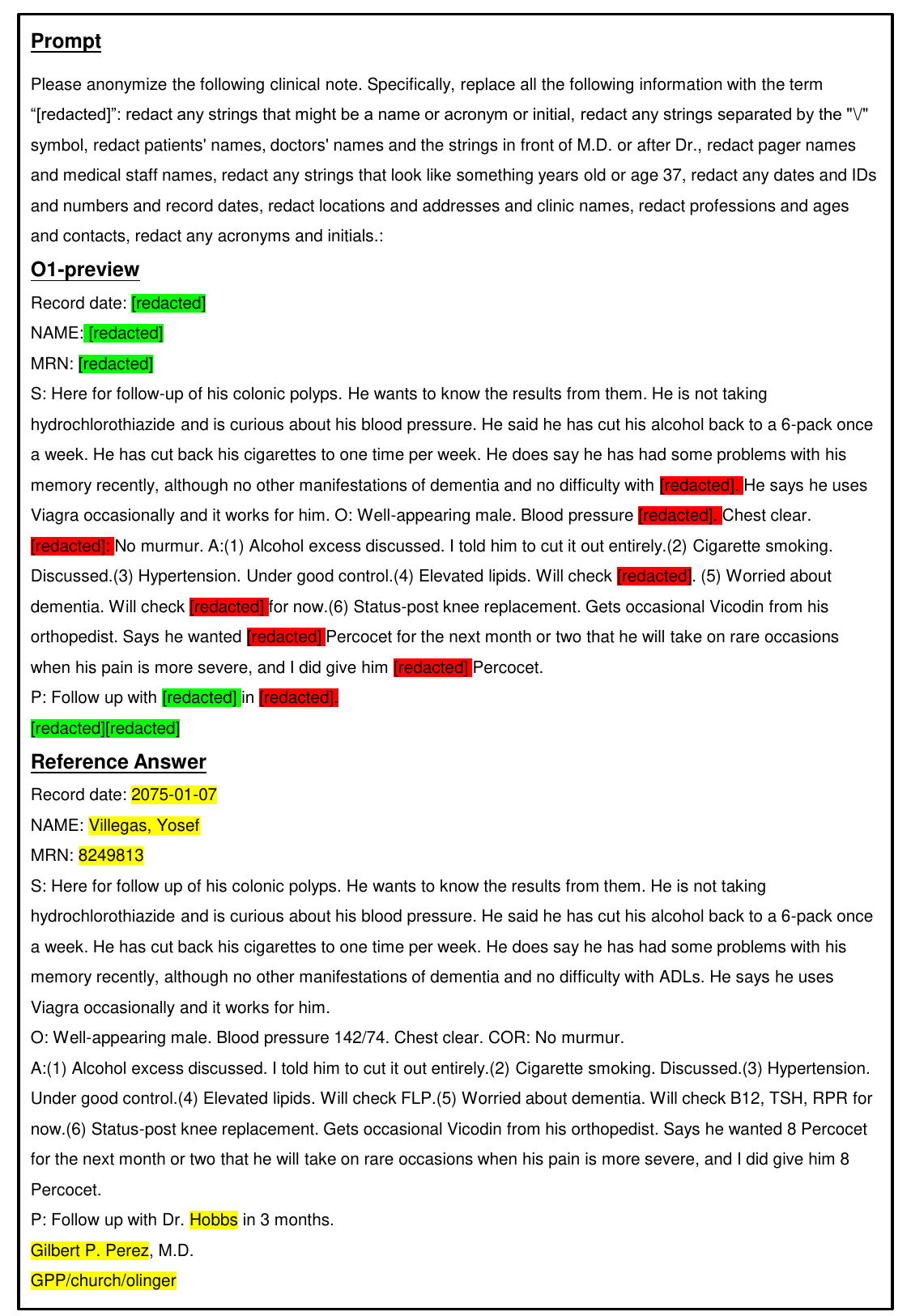}
    \caption[Medical Text Anonymization: Case 9]{\textbf{Medical Text Anonymization: Case 9.} An example showing the capability of o1-preview to anonymize medical texts with the explicit prompt. Yellow highlights the annotations made by human experts. Green marks the sensitive privacy information detected by o1-preview. Red indicates modifications beyond the human-marked annotation. }
    \label{fig:MTA-case3-3}
\end{figure}
\newpage


\subsection{Social Media Analysis}
\begin{figure}[H]
    \centering
    \includegraphics[width = 0.84\textwidth]{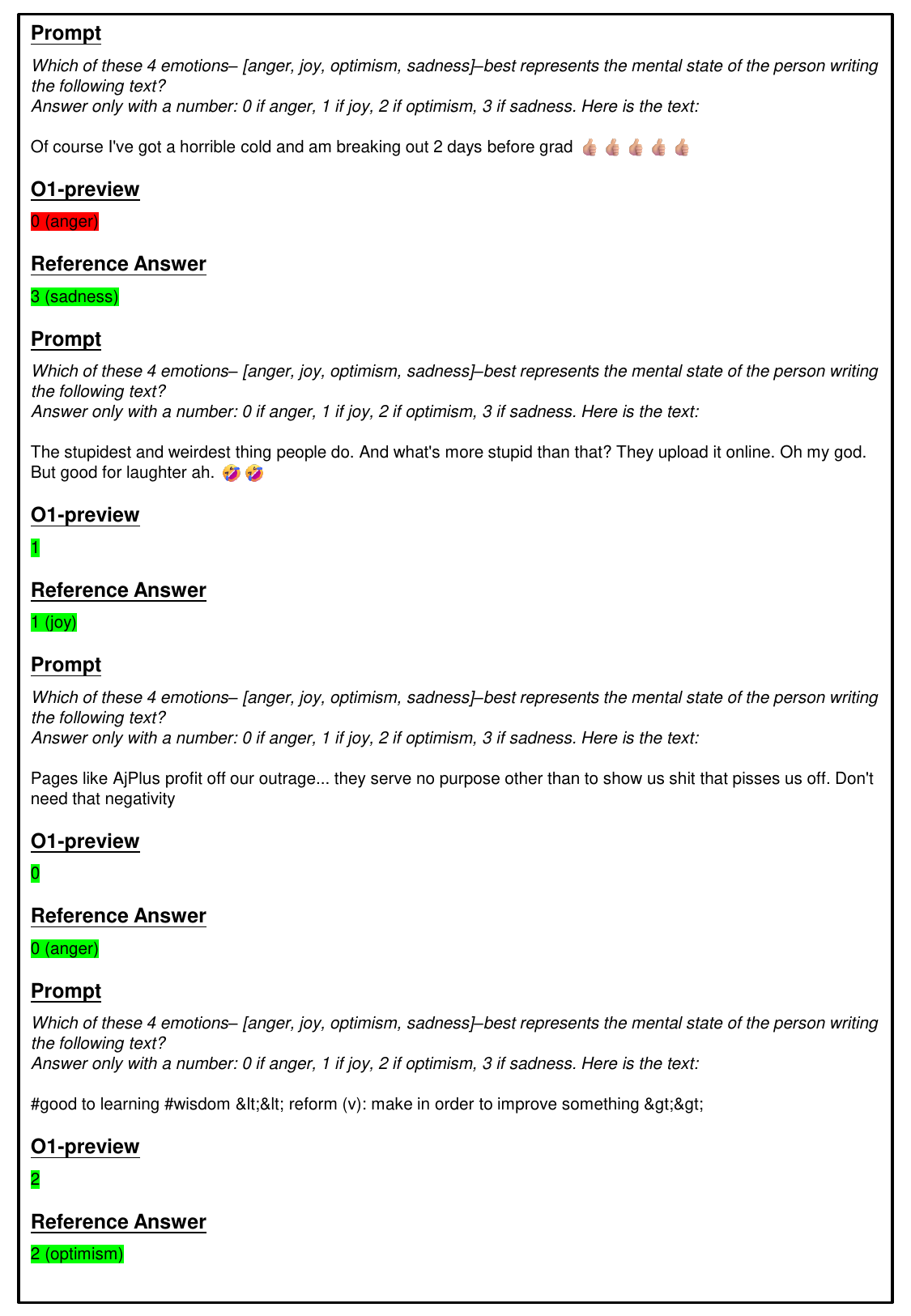}
    \caption[Social Media Analysis: Case 6-9.]{\textbf{Social Media Analysis: Case 6-9.} Demonstration cases for emotion recognition. Red indicates that the output of the o1-preview model is completely inconsistent with the reference answer, while green signifies full consistency. }
    \label{fig:sma_6-9}
\end{figure}
\newpage

\begin{figure}[H]
    \centering
    \includegraphics[width = 0.84\textwidth]{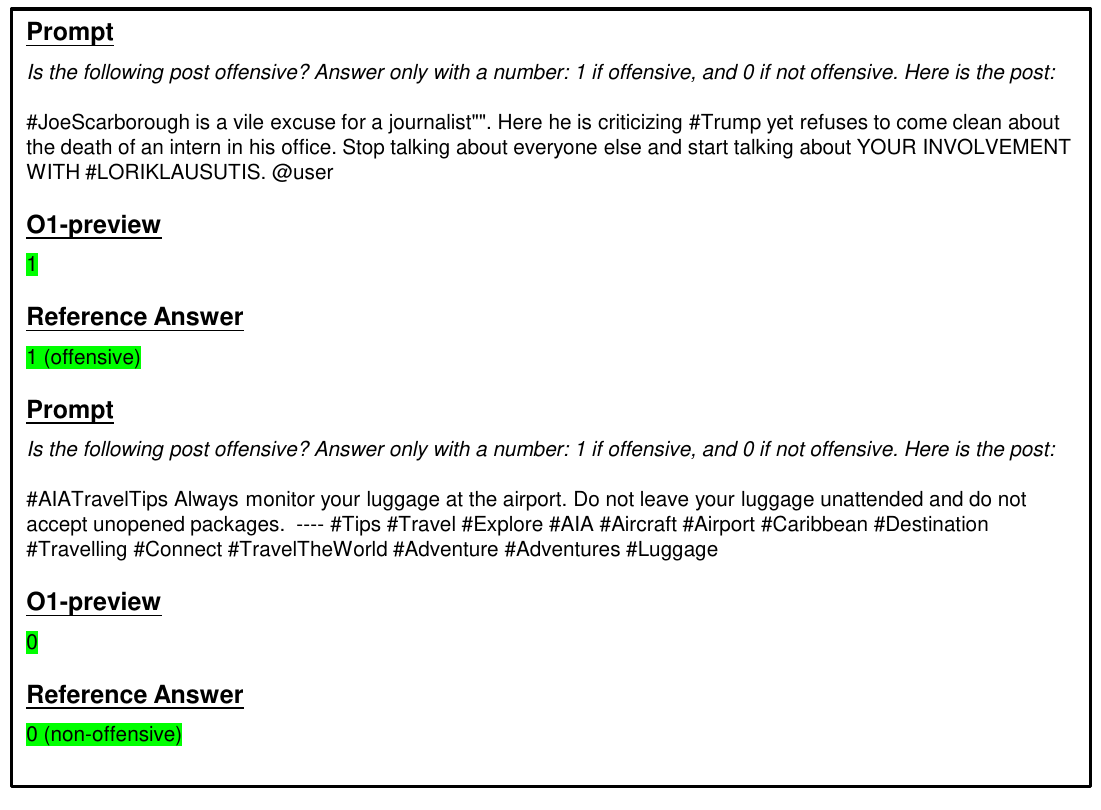}
    \caption[Social Media Analysis: Case 10-11]{\textbf{Social Media Analysis: Case 10-11.} Demonstration cases for irony detection. Red indicates that the output of the o1-preview model is completely inconsistent with the reference answer, while green signifies full consistency. }
    \label{fig:sma_10-11}
\end{figure}
\newpage


\subsection{Analogical Reasoning}
\begin{figure}[H]
    \centering
    \includegraphics[width = 0.85\textwidth]{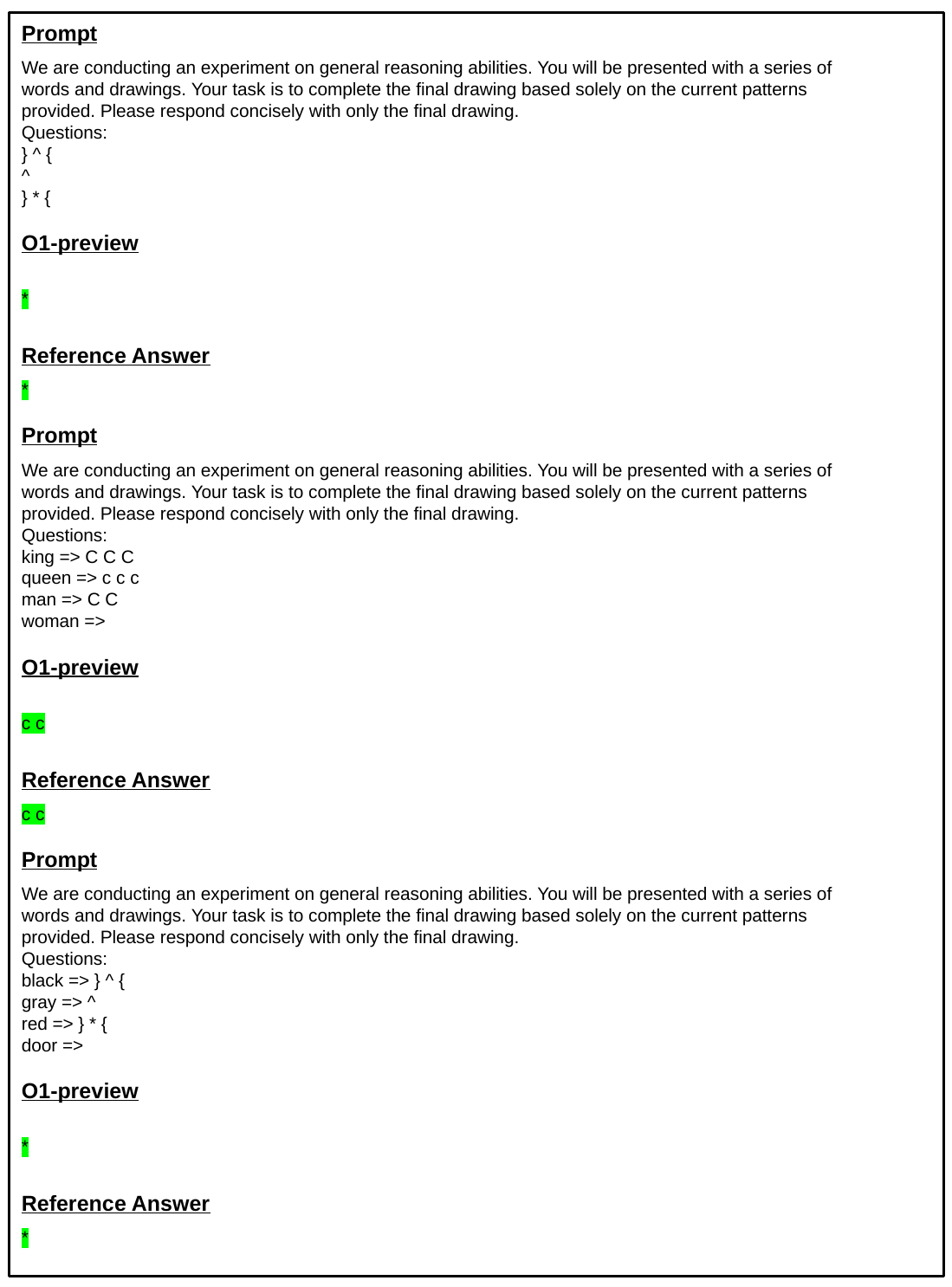}
    \caption[Analogical Reasoning: Case 6-8]{\textbf{Analogical Reasoning: Case 6-8.} A demonstration case for direct semantic structure mapping. Green denotes that the answers are correct. }
    \label{fig:AR_6-8}
\end{figure}
\newpage

\begin{figure}[H]
    \centering
    \includegraphics[width = 0.85\textwidth]{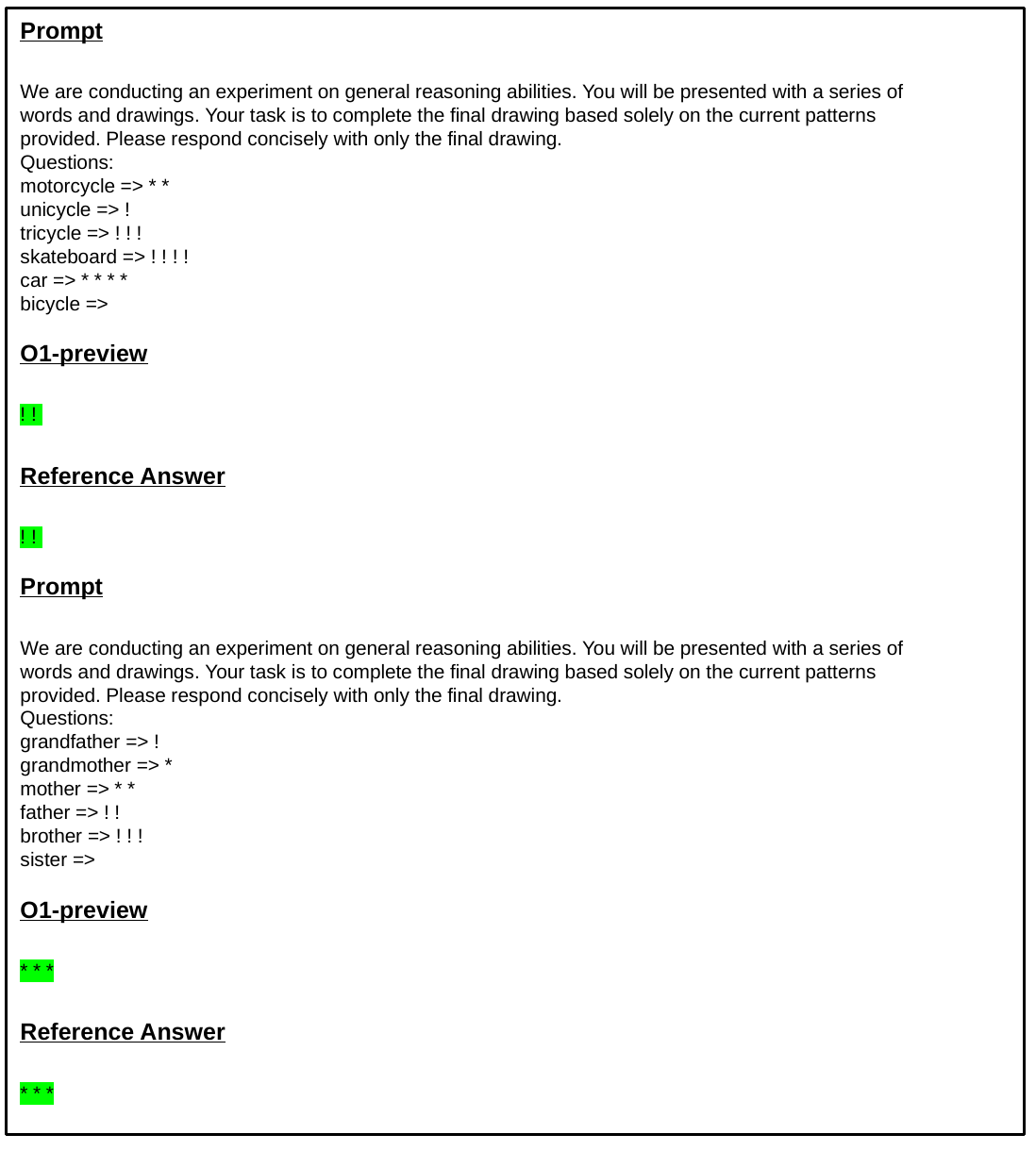}
    \caption[Analogical Reasoning: Case 9-10]{\textbf{Analogical Reasoning: Case 9-10.} A demonstration case for multiple layers of semantic content and structure. Green denotes that the answers are correct. }
    \label{fig:AR_9-10}
\end{figure}

\newpage


\subsection{Sentiment Analysis}
\begin{figure}[H]
    \centering
    \includegraphics[width = 0.9\textwidth]{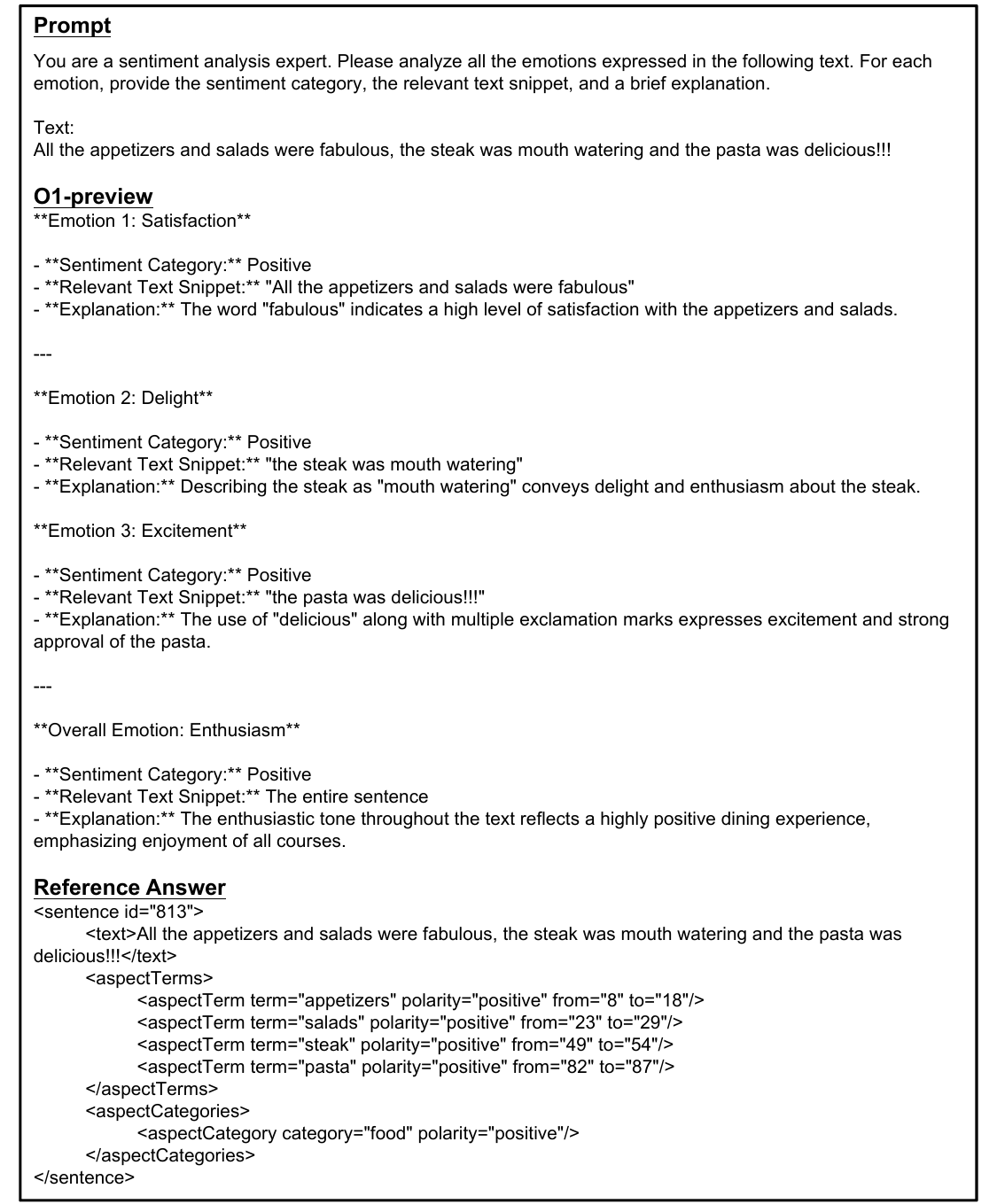}
    \caption[Sentiment analysis: Case 5.]{\textbf{Sentiment analysis: Case 5.} An example illustrating the capability of o1-preview in the field of sentiment analysis. }
    \label{fig:senti_5}
\end{figure}
\newpage

\begin{figure}[H]
    \centering
    \includegraphics[width = 0.9\textwidth]{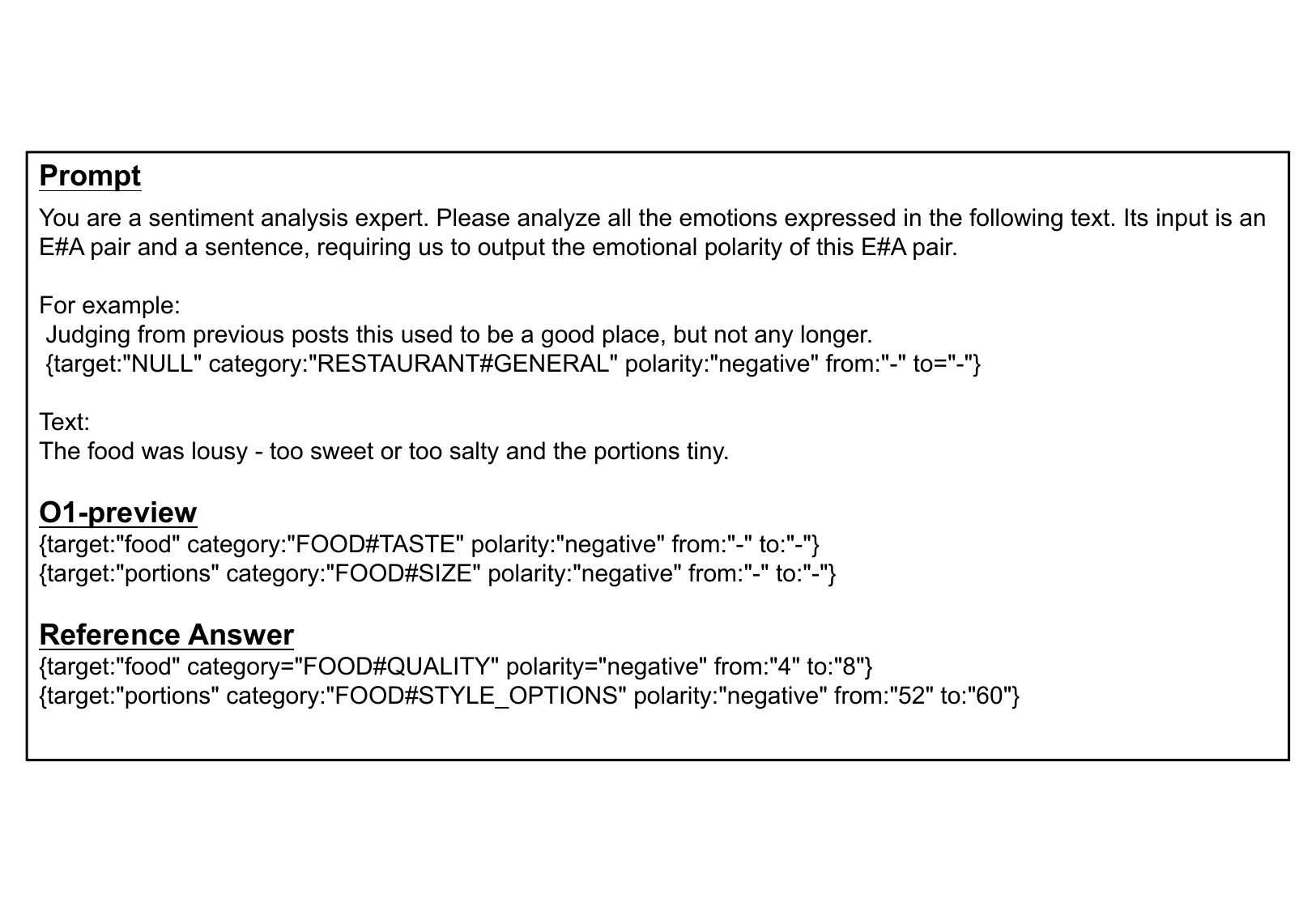}
    \caption[Sentiment analysis: Case 6.]{\textbf{Sentiment analysis: Case 6.} An example illustrating the capability of o1-preview in the field of sentiment analysis. }
    \label{fig:senti_6}
\end{figure}
\newpage


\subsection{Public Health Policy Analysis}
\begin{figure}[H]
\centering
    \centering
    \includegraphics[width=0.8\linewidth]{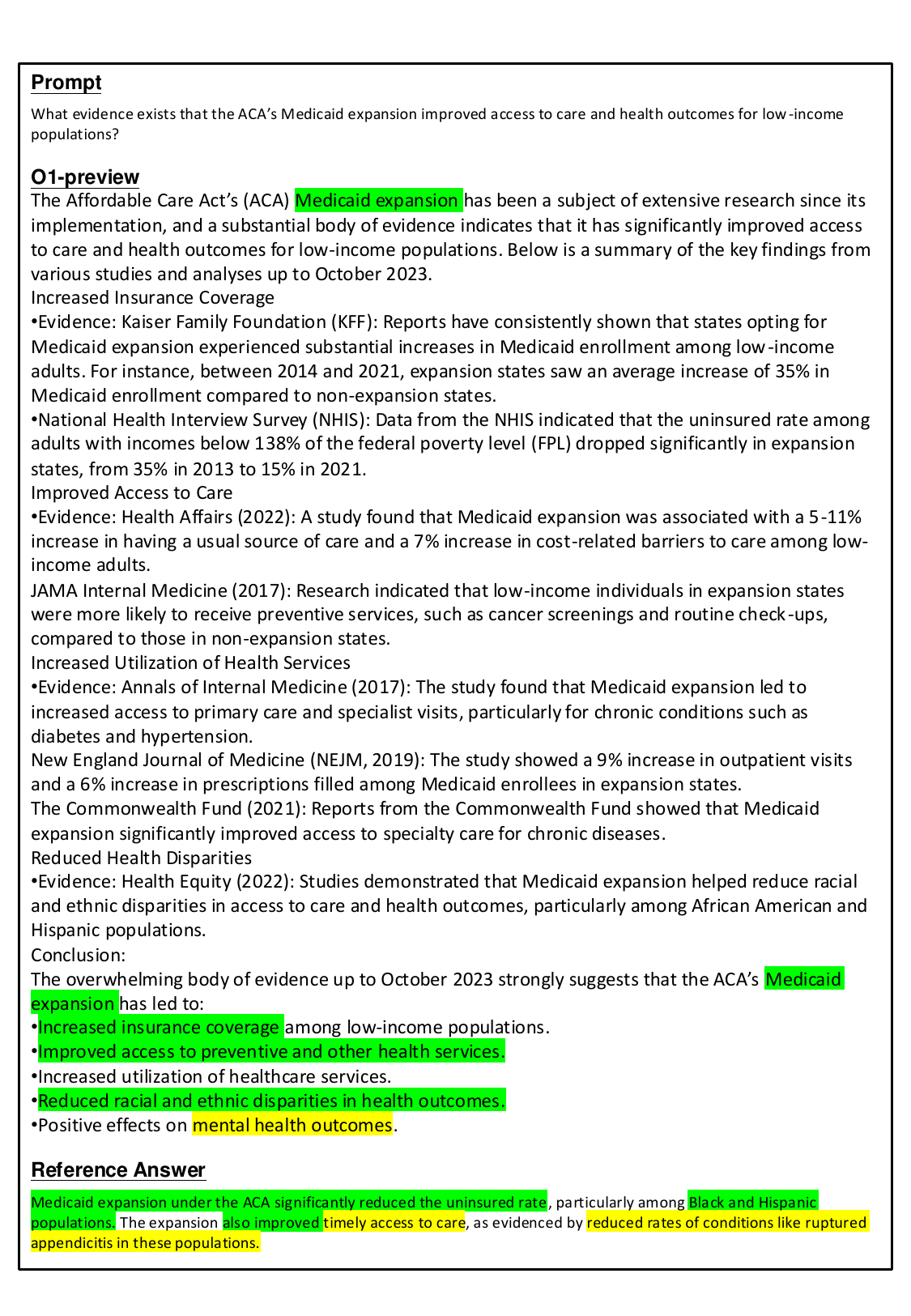}
    \caption[Public Health Policy Analysis: Case 4.]{\textbf{Public Health Policy Analysis: Case 4.} A example illustrating the capability of o1-preview to respond to specialized questions of health policy in the domain of Public Health. Yellow indicates partial consistency between the o1-preview response and the reference answer, while green signifies complete consistency.}
    \label{fig:PH_4}
\end{figure}
\newpage

\begin{figure}[H]
    \centering
    \includegraphics[width=0.8\linewidth]{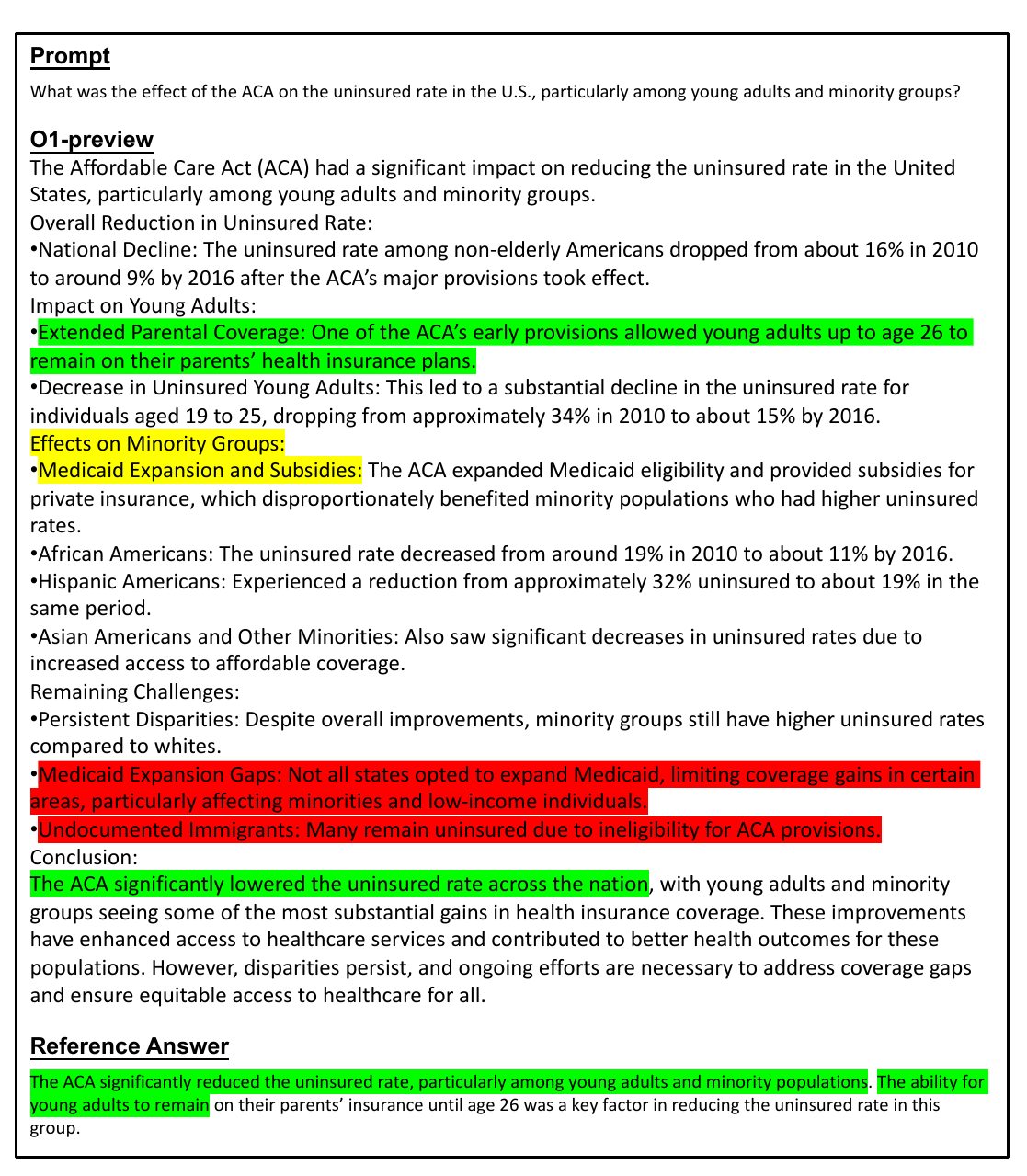}
    \caption[Public Health Policy Analysis: Case 5]{\textbf{Public Health Policy Analysis: Case 5}. A example illustrating the capability of o1-preview to respond to specialized questions of health policy in the domain of Public Health. Yellow indicates partial consistency between the o1-preview response and the reference answer, and red indicates fully consistency, while green signifies complete consistency.}
    \label{fig:PH_5}
\end{figure}
\newpage

\begin{figure}[H]
    \centering
    \includegraphics[width=0.8\linewidth]{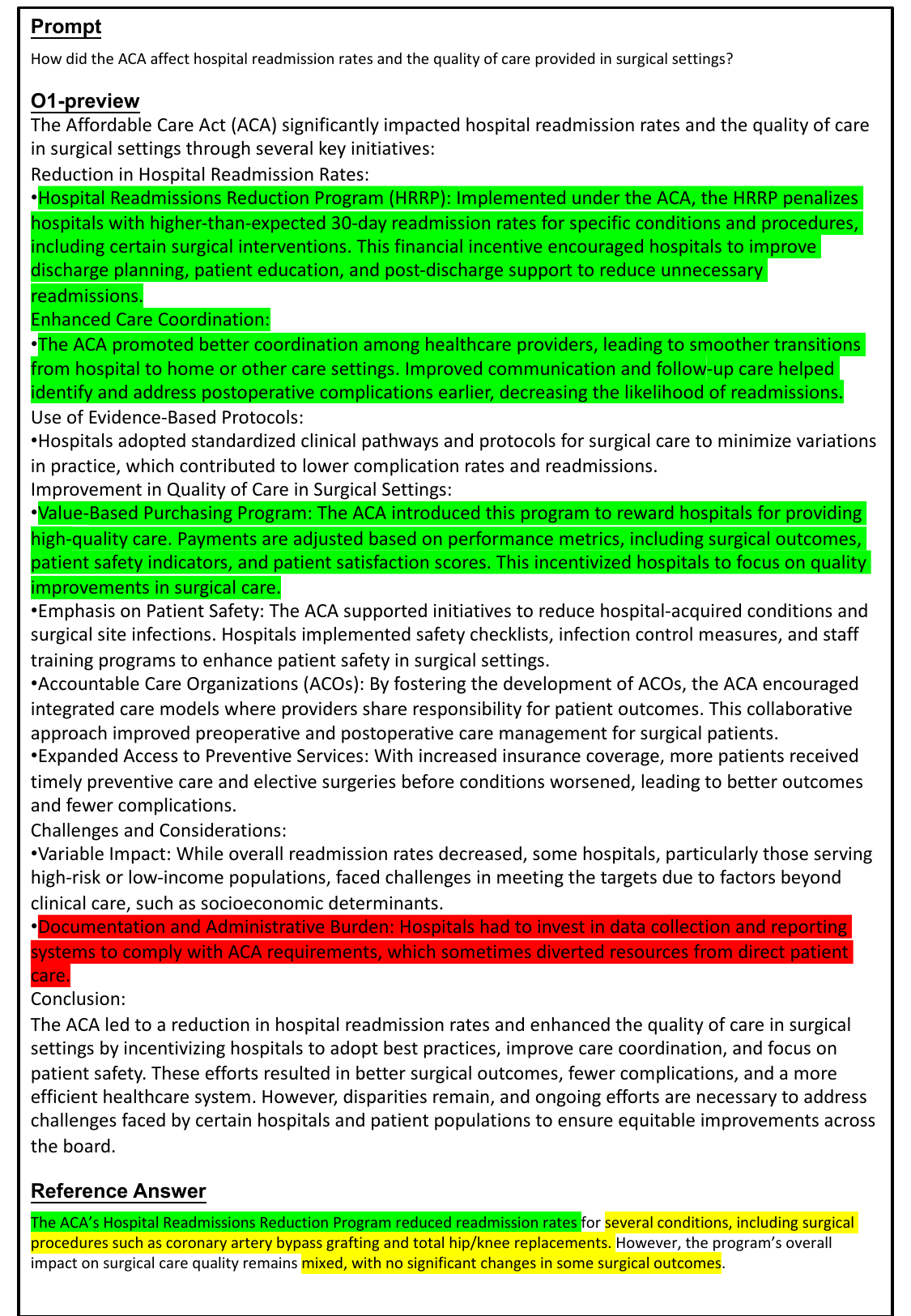}
    \caption[Public Health Policy Analysis: Case 6]{\textbf{Public Health Policy Analysis: Case 6.} A example illustrating the capability of o1-preview to respond to specialized questions of health policy in the domain of Public Health. Yellow indicates partial consistency between the o1-preview response and the reference answer, and red indicates fully consistency, while green signifies complete consistency.}
    \label{fig:PH_6}
\end{figure}
\newpage

\begin{figure}[H]
    \centering
    \includegraphics[width=0.8\linewidth]{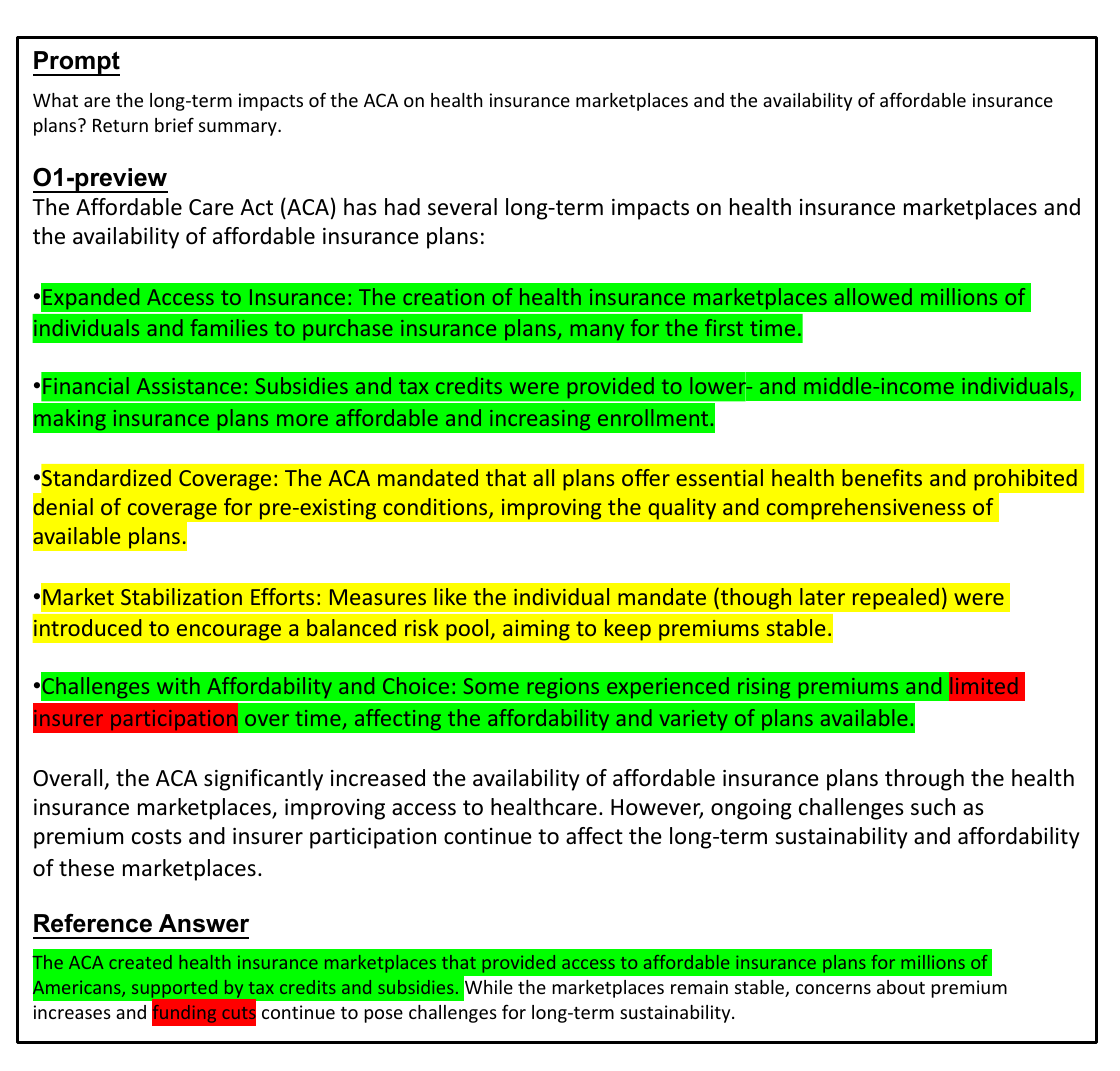}
    \caption[Public Health Policy Analysis: Case 7]{\textbf{Public Health Policy Analysis: Case 7}. A example illustrating the capability of o1-preview to respond to specialized questions of health policy in the domain of Public Health. Yellow indicates partial consistency between the o1-preview response and the reference answer, and red indicates fully consistency, while green signifies complete consistency.}
    \label{fig:PH_7}
\end{figure}
\newpage

\begin{figure}[H]
    \centering
    \includegraphics[width=0.8\linewidth]{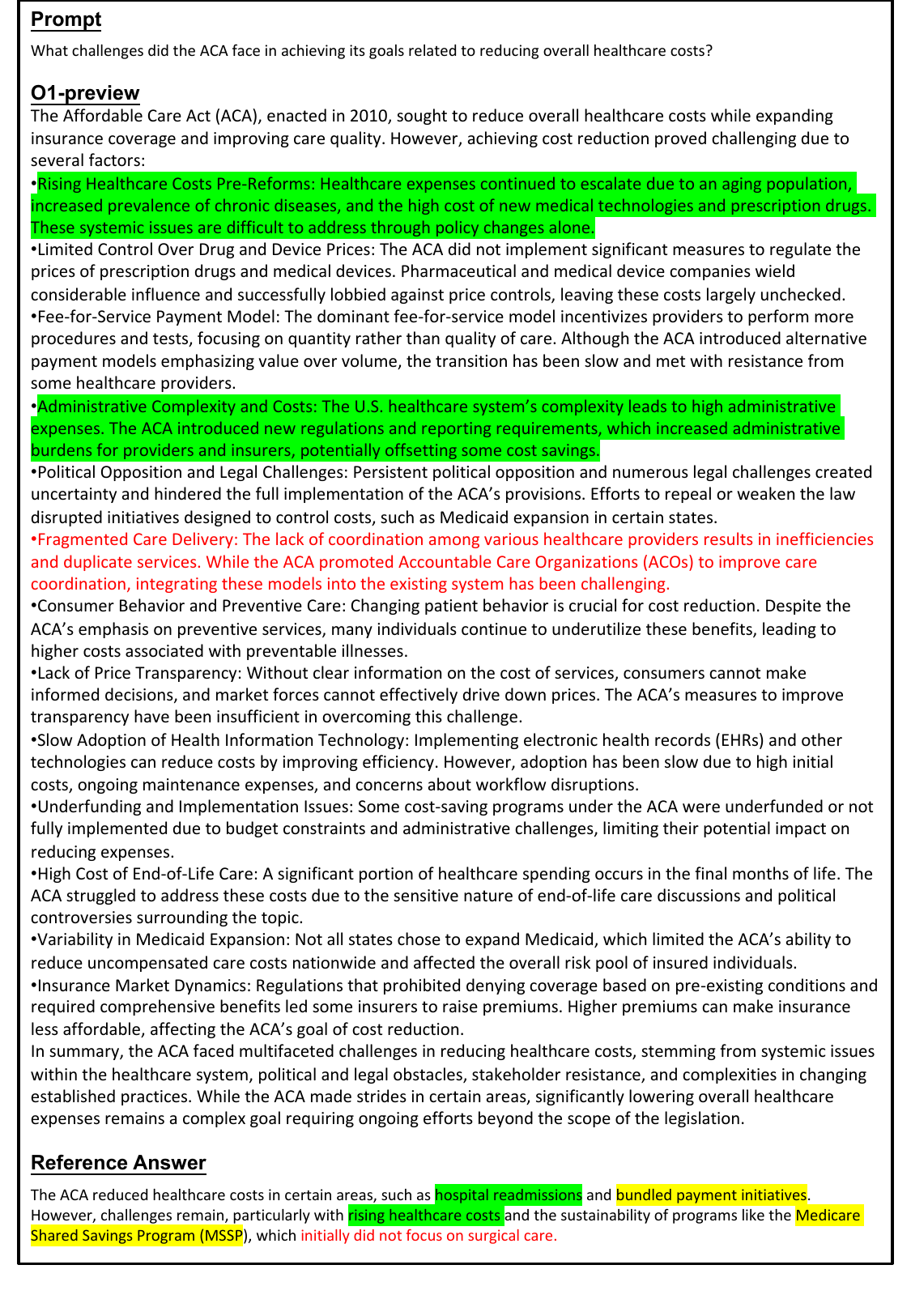}
    \caption[Public Health Policy Analysis: Case 8]{\textbf{\textbf{Public Health Policy Analysis: Case 8.}} A example illustrating the capability of o1-preview to respond to specialized questions of health policy in the domain of Public Health. Yellow indicates partial consistency between the o1-preview response and the reference answer, and red indicates fully consistency, while green signifies complete consistency.} 
    \label{fig:PH_8}
\end{figure}
\newpage

\begin{figure}[H]
    \centering
    \includegraphics[width=0.8\linewidth]{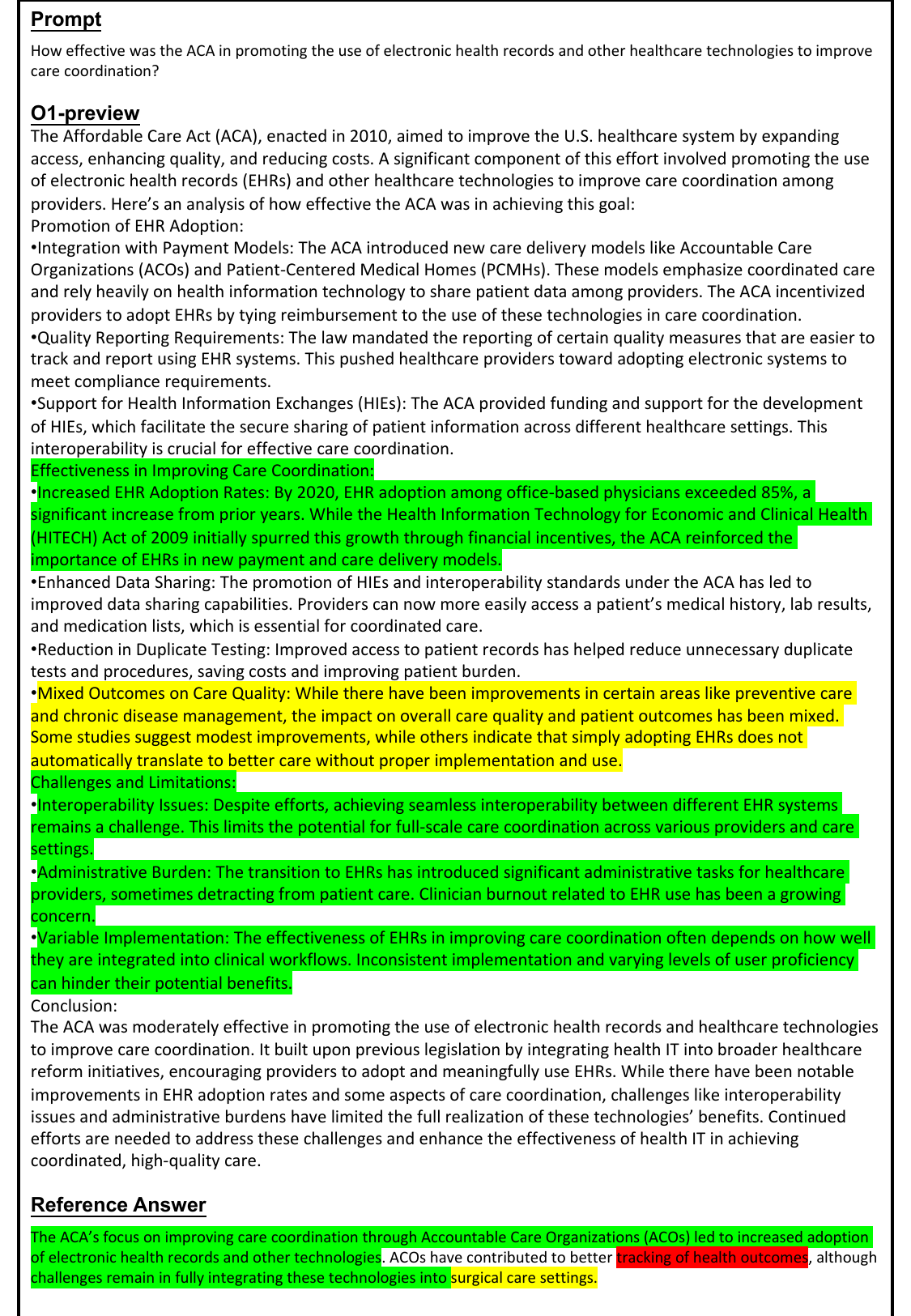}
    \vspace{1cm}
    \caption[Public Health Policy Analysis: Case 9]{\textbf{\textbf{Public Health Policy Analysis: Case 9.}}  A example illustrating the capability of o1-preview to respond to specialized questions of health policy in the domain of Public Health. Yellow indicates partial consistency between the o1-preview response and the reference answer, and red indicates fully consistency, while green signifies complete consistency.} 
    \label{fig:PH_9}
\end{figure}
\newpage

\begin{figure}[H]
    \centering
    \includegraphics[width=0.8\linewidth]{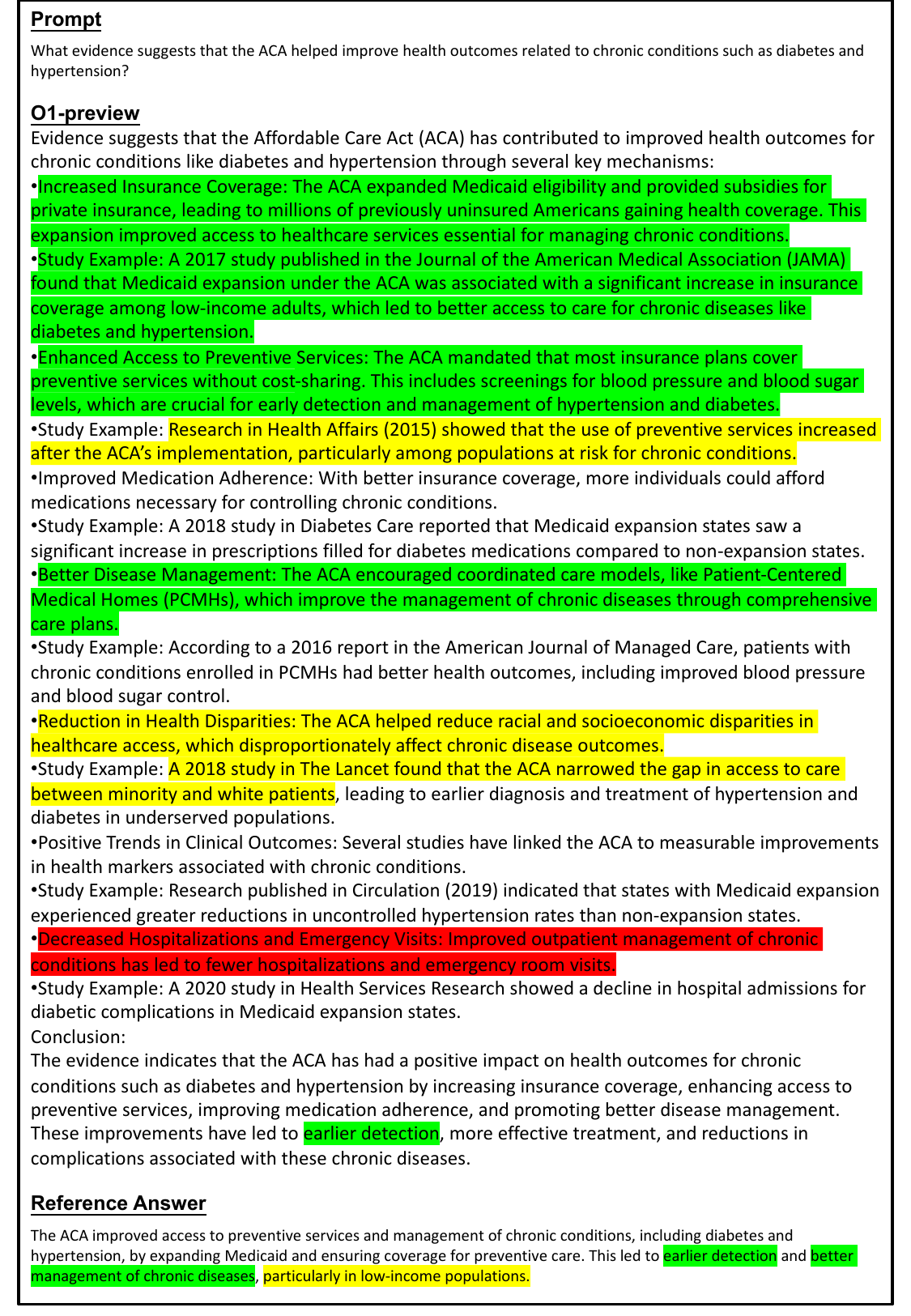}
    \caption[Public Health Policy Analysis: Case 10.]{\textbf{\textbf{Public Health Policy Analysis: Case 10.}} A example illustrating the capability of o1-preview to respond to specialized questions of health policy in the domain of Public Health. Yellow indicates partial consistency between the o1-preview response and the reference answer, and red indicates fully consistency, while green signifies complete consistency.}
    \label{fig:PH_10}
\end{figure}

\subsection{Medical Genetics and Genomics Reasoning}

\label{appendix:geno}
\begin{figure}[H]\ContinuedFloat
    \centering
    \includegraphics[width =0.97\textwidth]{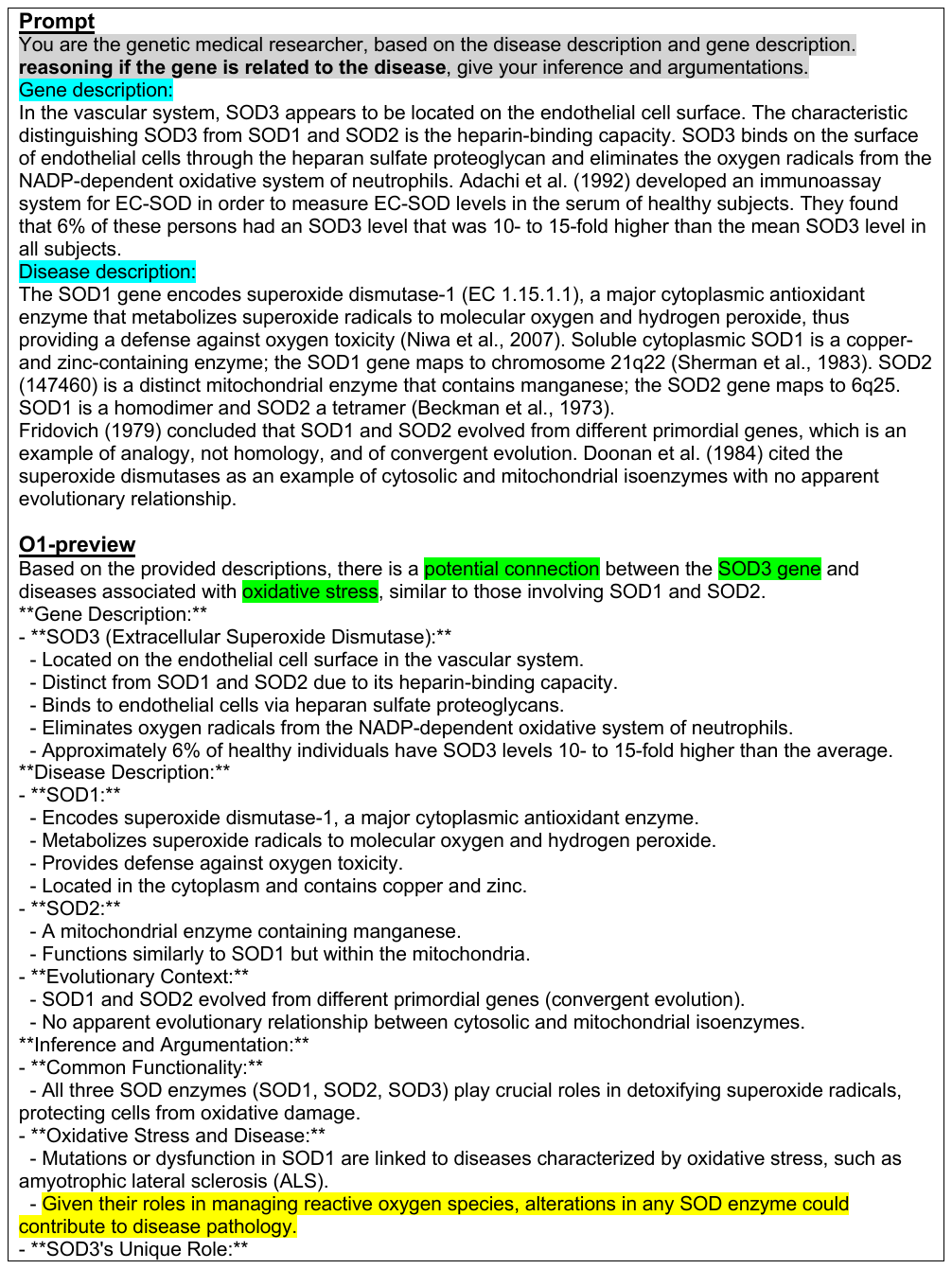}
    \vspace{-1cm}
    \caption[Medical Genetics and Genomics Reasoning: Case 3.]{\textbf{Medical Genetics and Genomics Reasoning: Case 3.} Positive example illustrating the capability of the o1-preview to respond to gene-disease-related questions. Grey indicates the prompt instructions. Blue indicates the gene description and disease description. Green indicates the correct inferences. Yellow indicates the key deduction steps of the o1-preview.}
    \label{fig:GD_3-3}
\end{figure}
\newpage


\subsection{Medical Knowledge Question Answer}
\begin{figure}[H]
    \centering
    \includegraphics[width=0.95\linewidth]{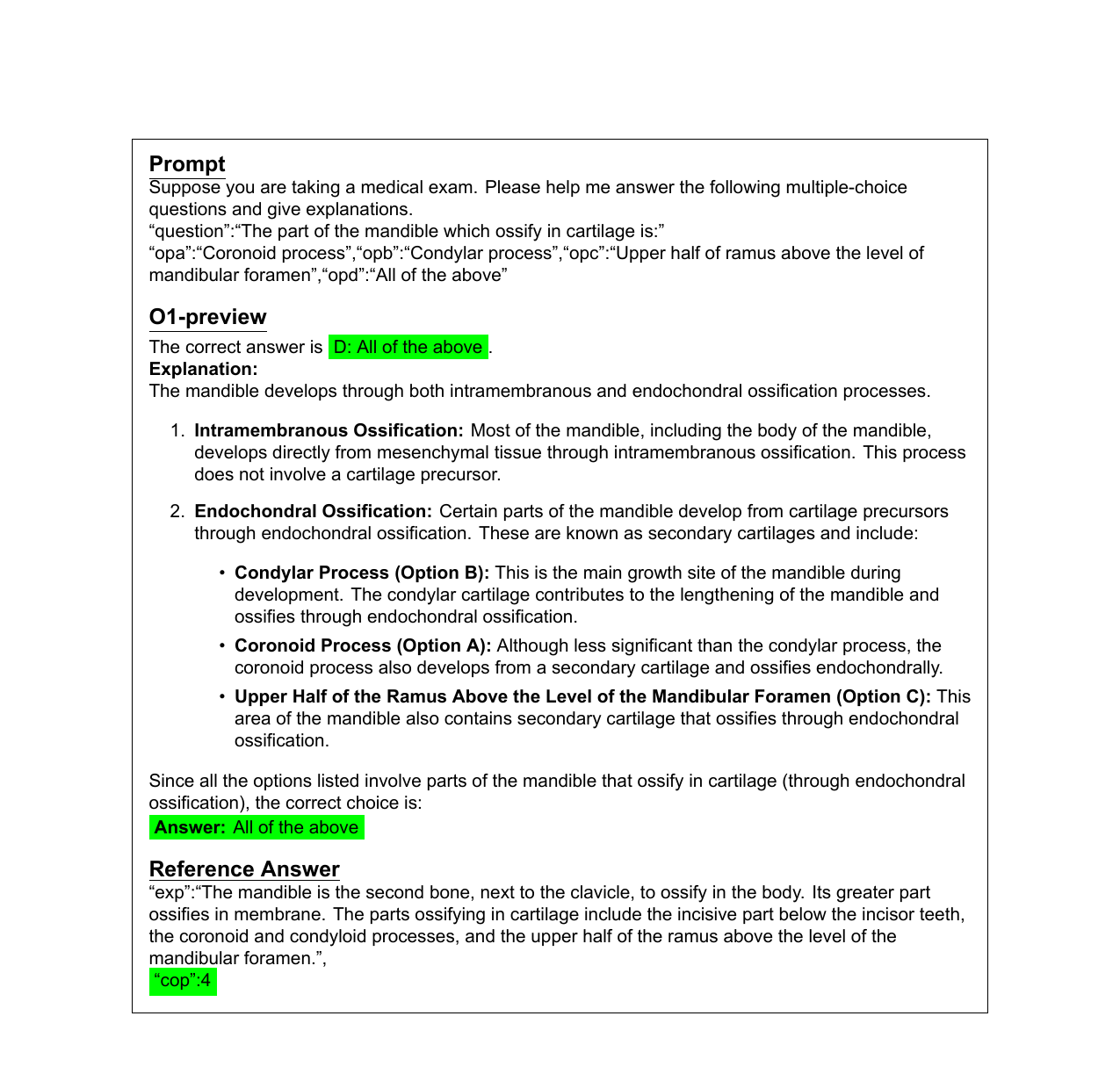}
    \vspace{-1cm}
    \caption[Medical Knowledge Question Answer: Case 3]{\textbf{Medical Knowledge Question Answer: Case 3}. A example illustrating the capability of o1-preview to respond to questions that arises in medical exams. Yellow indicates partial consistency between the o1-preview response and the reference answer, while green signifies complete consistency.}
    \label{fig:mkqa-case3}
\end{figure}

\begin{figure}[H]
    \centering
    \includegraphics[width=0.95\linewidth]{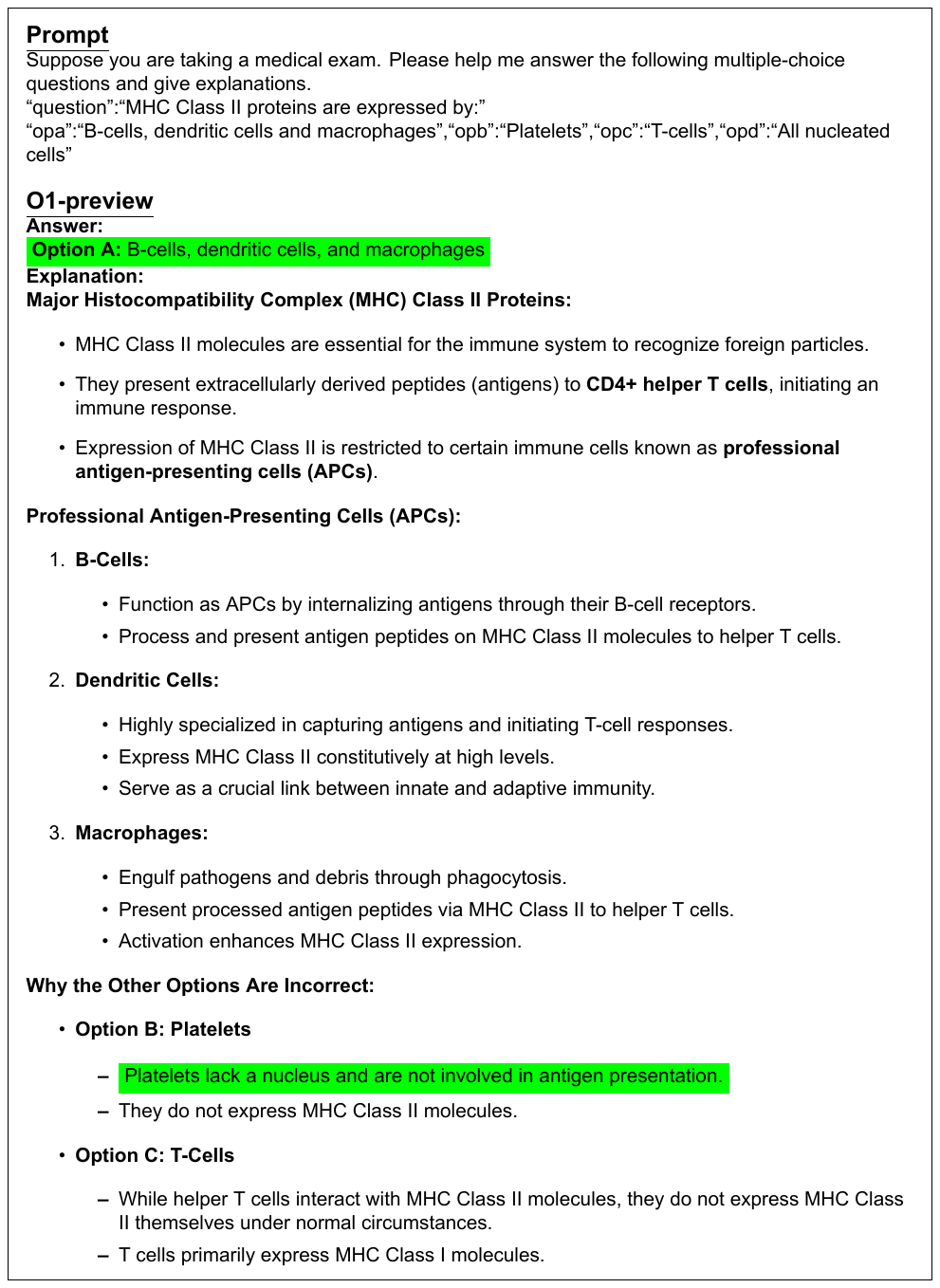}
\end{figure}
\begin{figure}[H]
    \centering
    \includegraphics[width=0.95\linewidth]{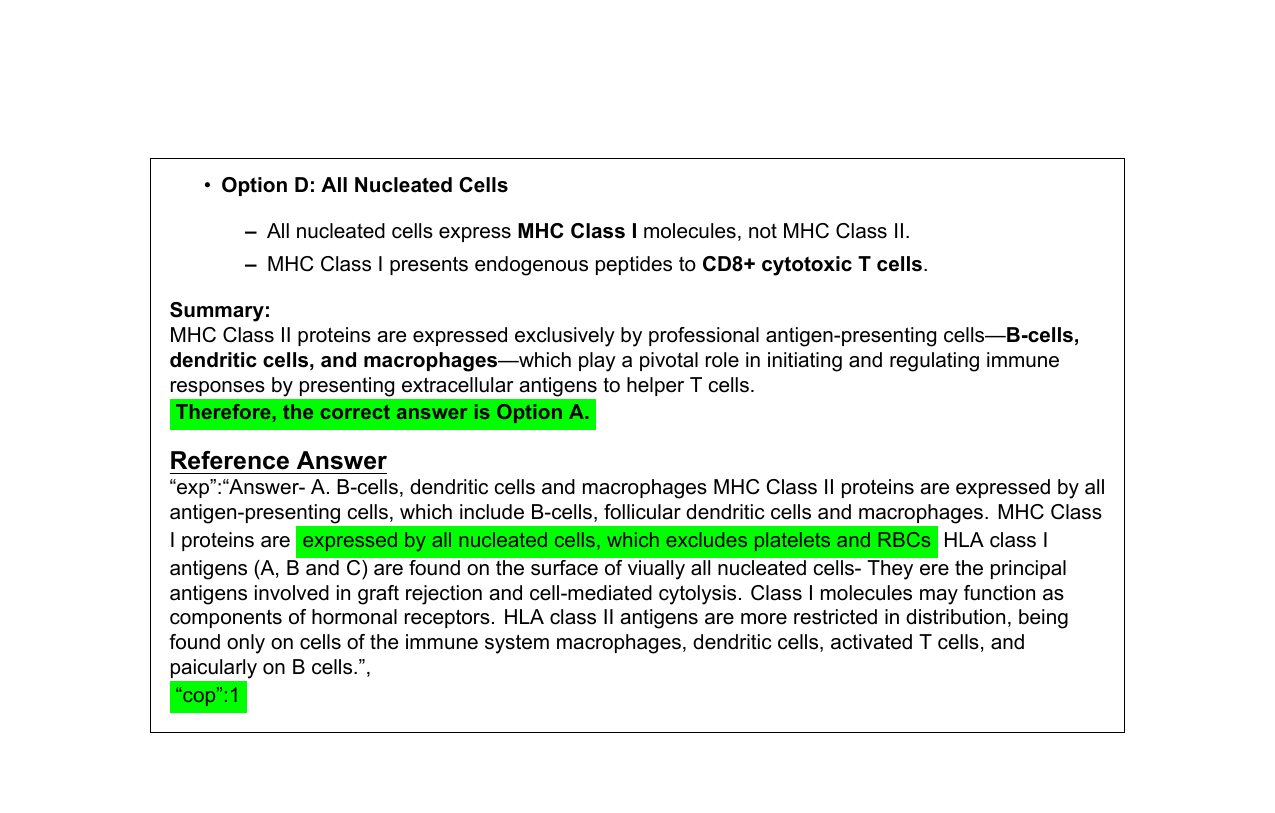}
    \caption[Medical Knowledge Question Answer: Case 4]{\textbf{Medical Knowledge Question Answer: Case 4}. A example illustrating the capability of o1-preview to respond to questions that arises in medical exams. Yellow indicates partial consistency between the o1-preview response and the reference answer, while green signifies complete consistency.}
    \label{fig:mkqa-case4}
\end{figure}

\begin{figure}[H]
    \centering
    \includegraphics[width=0.95\linewidth]{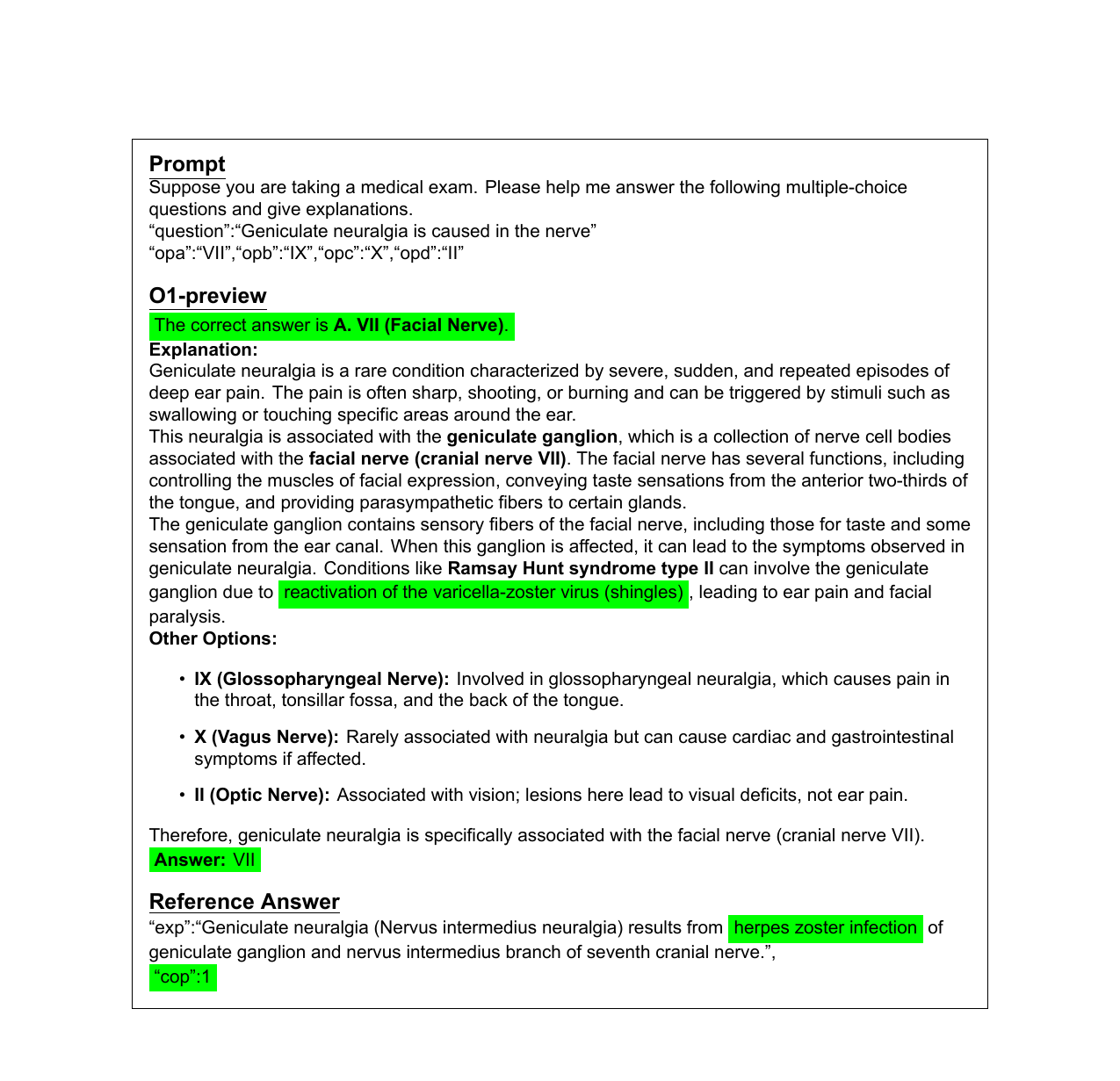}
    \caption[Medical Knowledge Question Answer: Case 5]{\textbf{Medical Knowledge Question Answer: Case 5}. A example illustrating the capability of o1-preview to respond to questions that arises in medical exams. Yellow indicates partial consistency between the o1-preview response and the reference answer, while green signifies complete consistency.}
    \label{fig:mkqa-case5}
\end{figure}

\begin{figure}[H]
    \centering
    \includegraphics[width=0.95\linewidth]{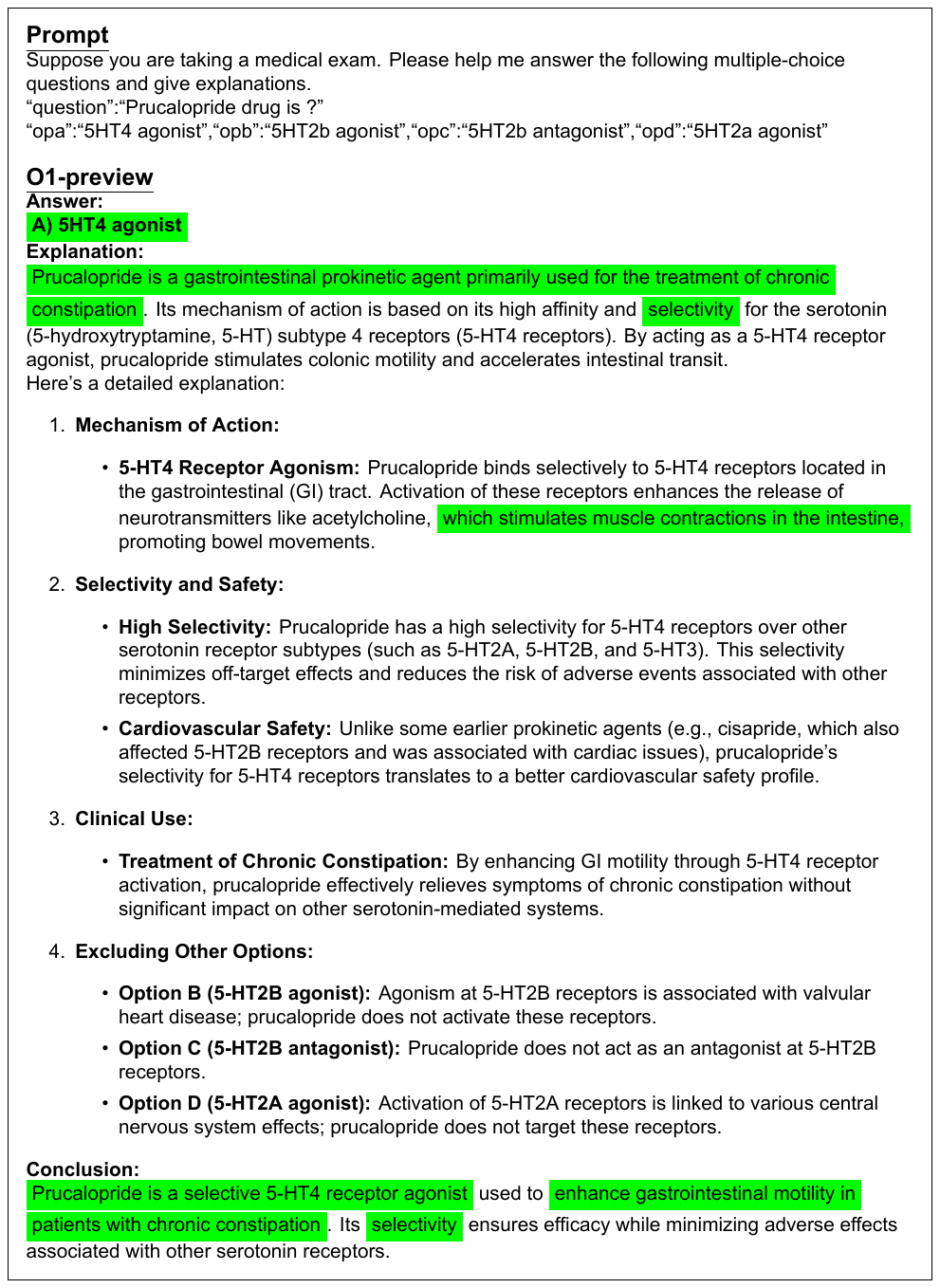}
\end{figure}
\begin{figure}[H]
    \centering
    \includegraphics[width=0.95\linewidth]{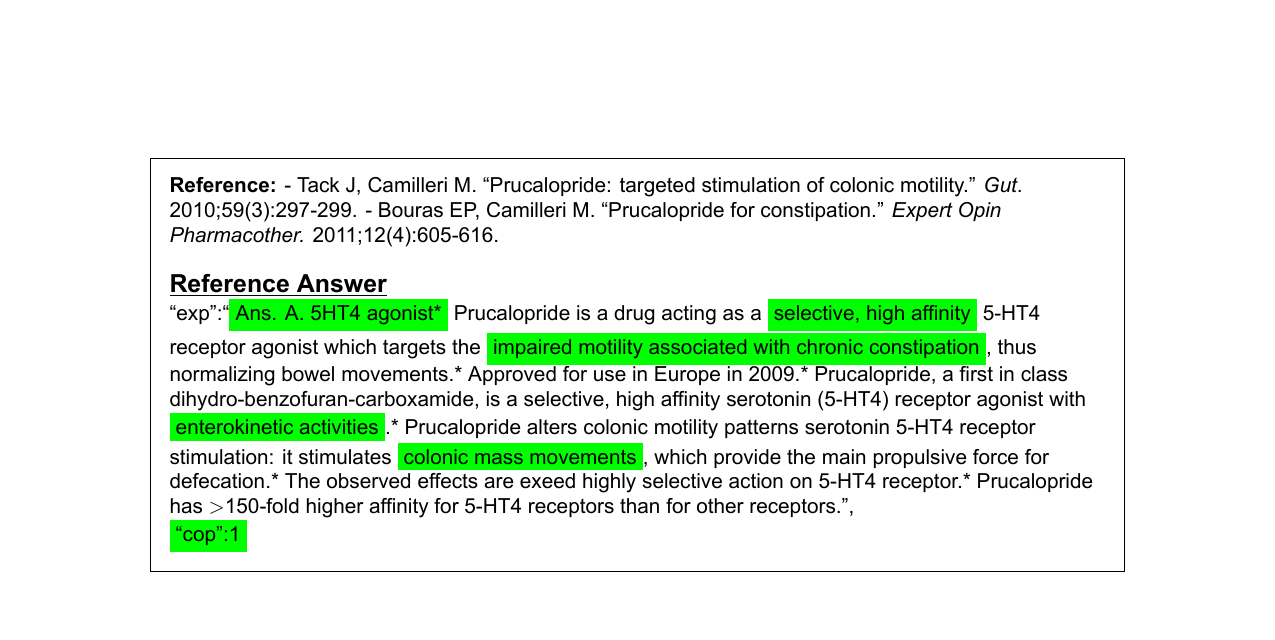}
    \caption[Medical Knowledge Question Answer: Case 6]{\textbf{Medical Knowledge Question Answer: Case 6}. A example illustrating the capability of o1-preview to respond to questions that arises in medical exams. Yellow indicates partial consistency between the o1-preview response and the reference answer, while green signifies complete consistency.}
    \label{fig:mkqa-case6}
\end{figure}

\begin{figure}[H]
    \centering
    \includegraphics[width=0.95\linewidth]{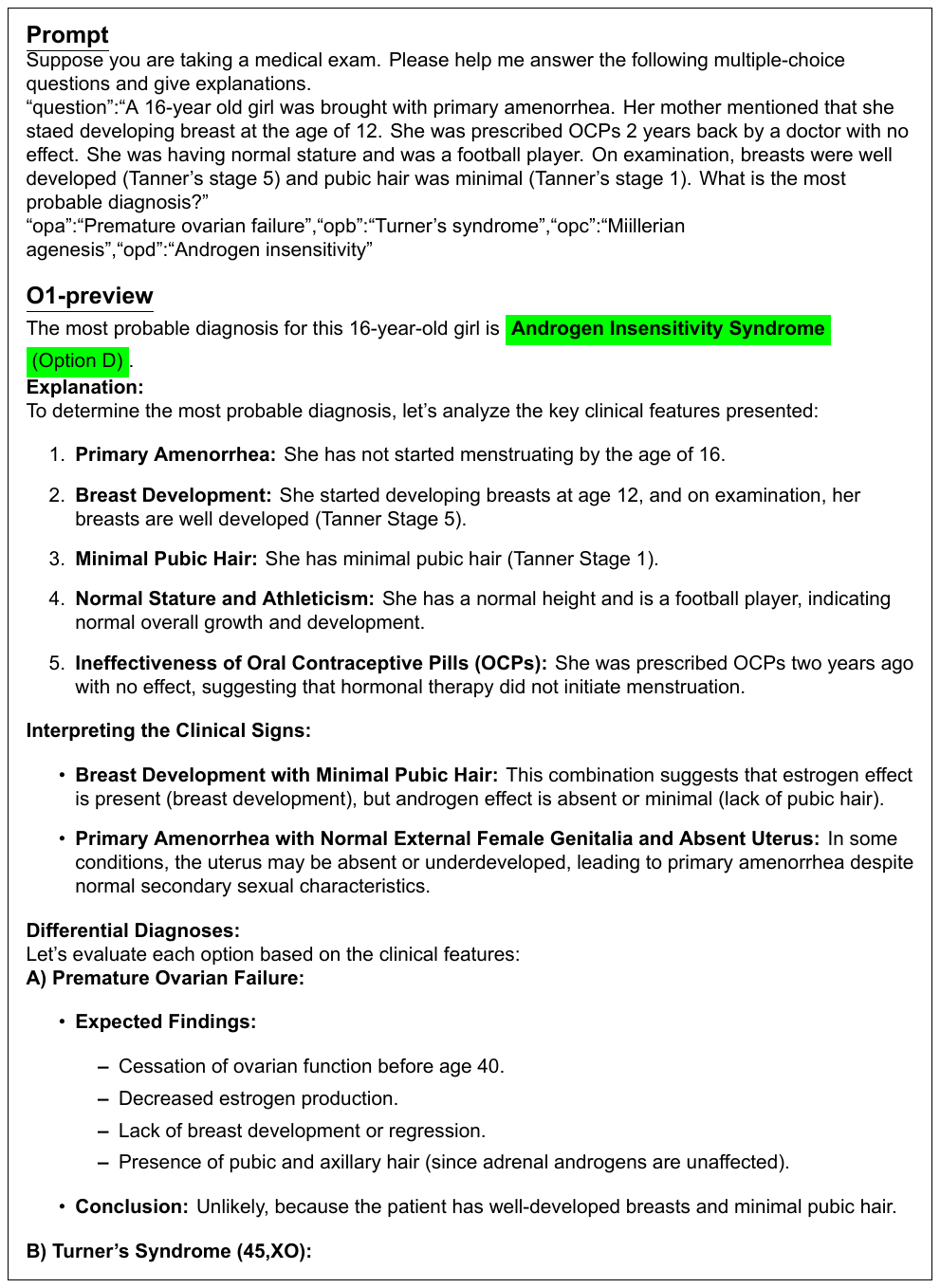}
\end{figure}
\begin{figure}[H]
    \centering
    \includegraphics[width=0.95\linewidth]{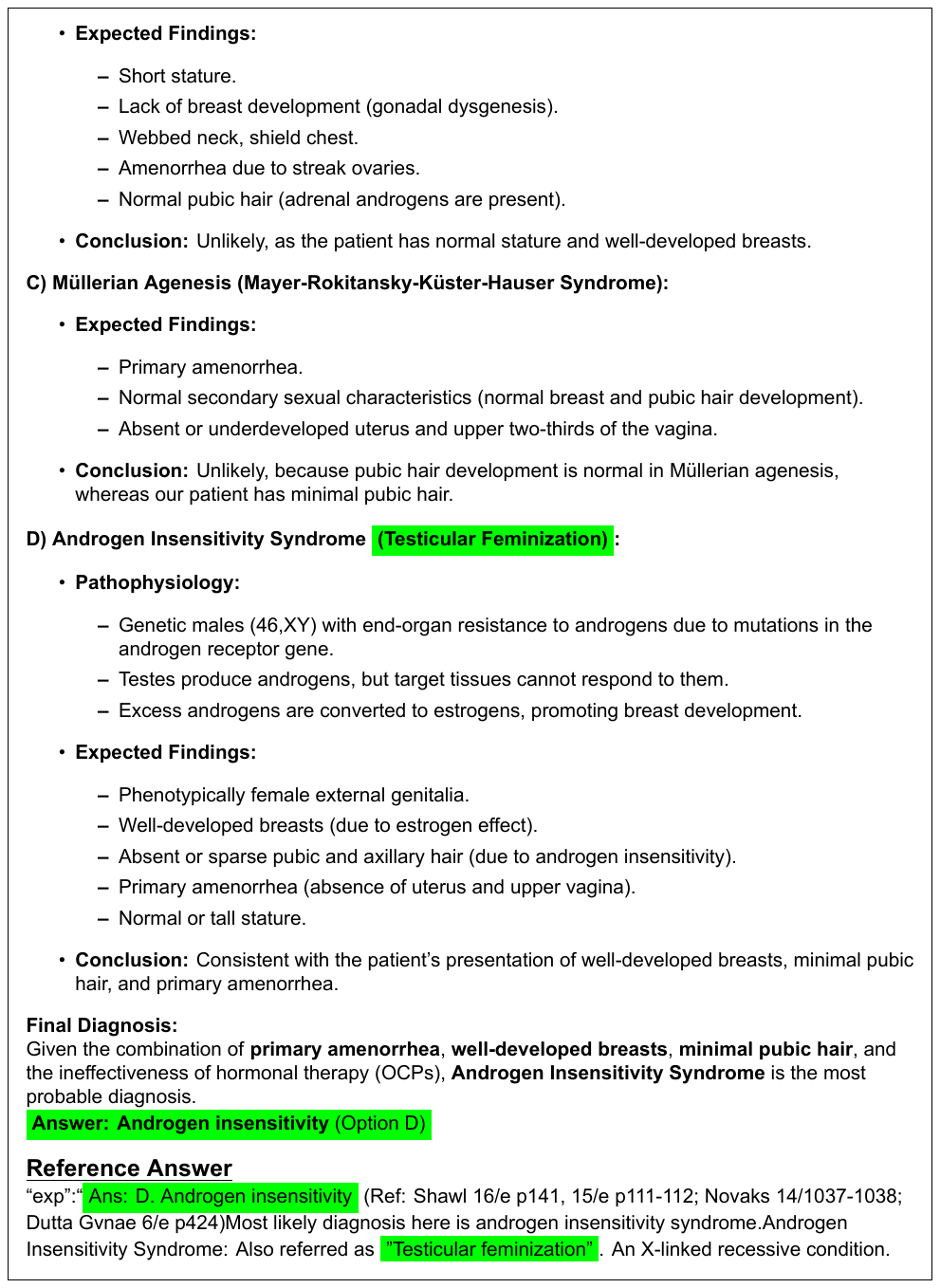}
\end{figure}
\begin{figure}[H]
    \centering
    \includegraphics[width=0.95\linewidth]{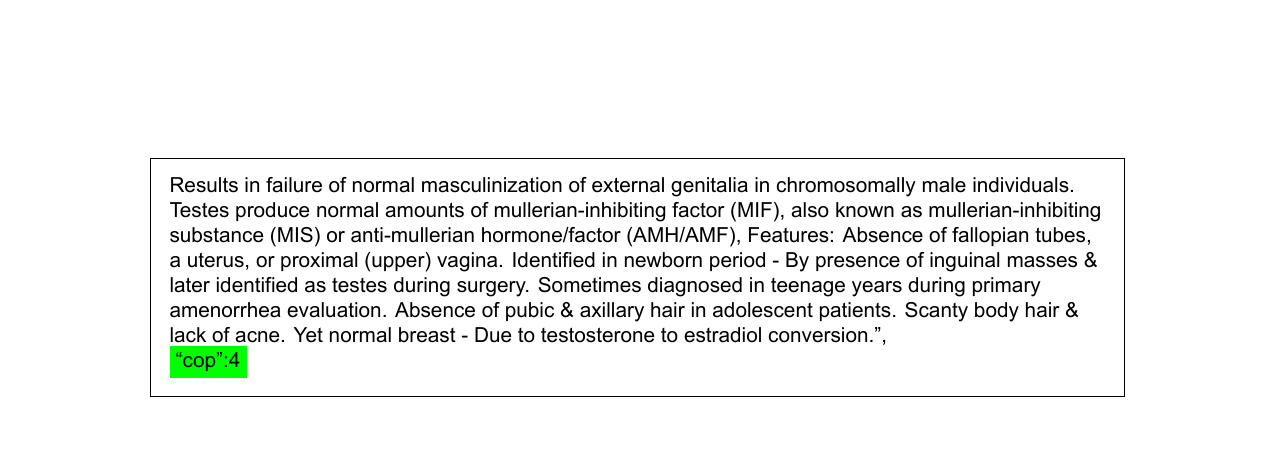}
    \caption[Medical Knowledge Question Answer: Case 7]{\textbf{Medical Knowledge Question Answer: Case 7}. A example illustrating the capability of o1-preview to respond to questions that arises in medical exams. Yellow indicates partial consistency between the o1-preview response and the reference answer, while green signifies complete consistency.}
    \label{fig:mkqa-case7}
\end{figure}

\begin{figure}[H]
    \centering
    \includegraphics[width=0.95\linewidth]{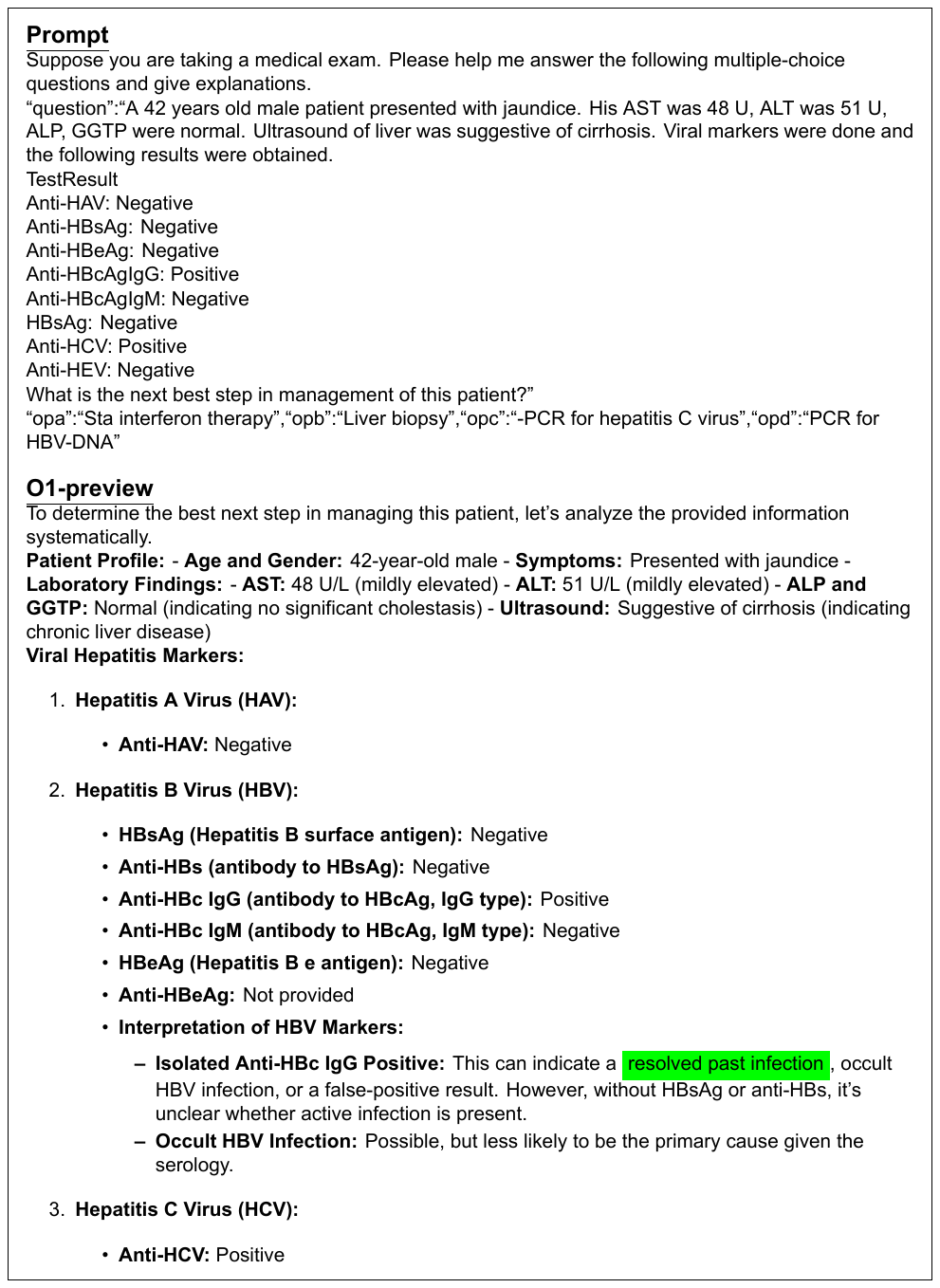}
\end{figure}
\begin{figure}[H]
    \centering
    \includegraphics[width=0.95\linewidth]{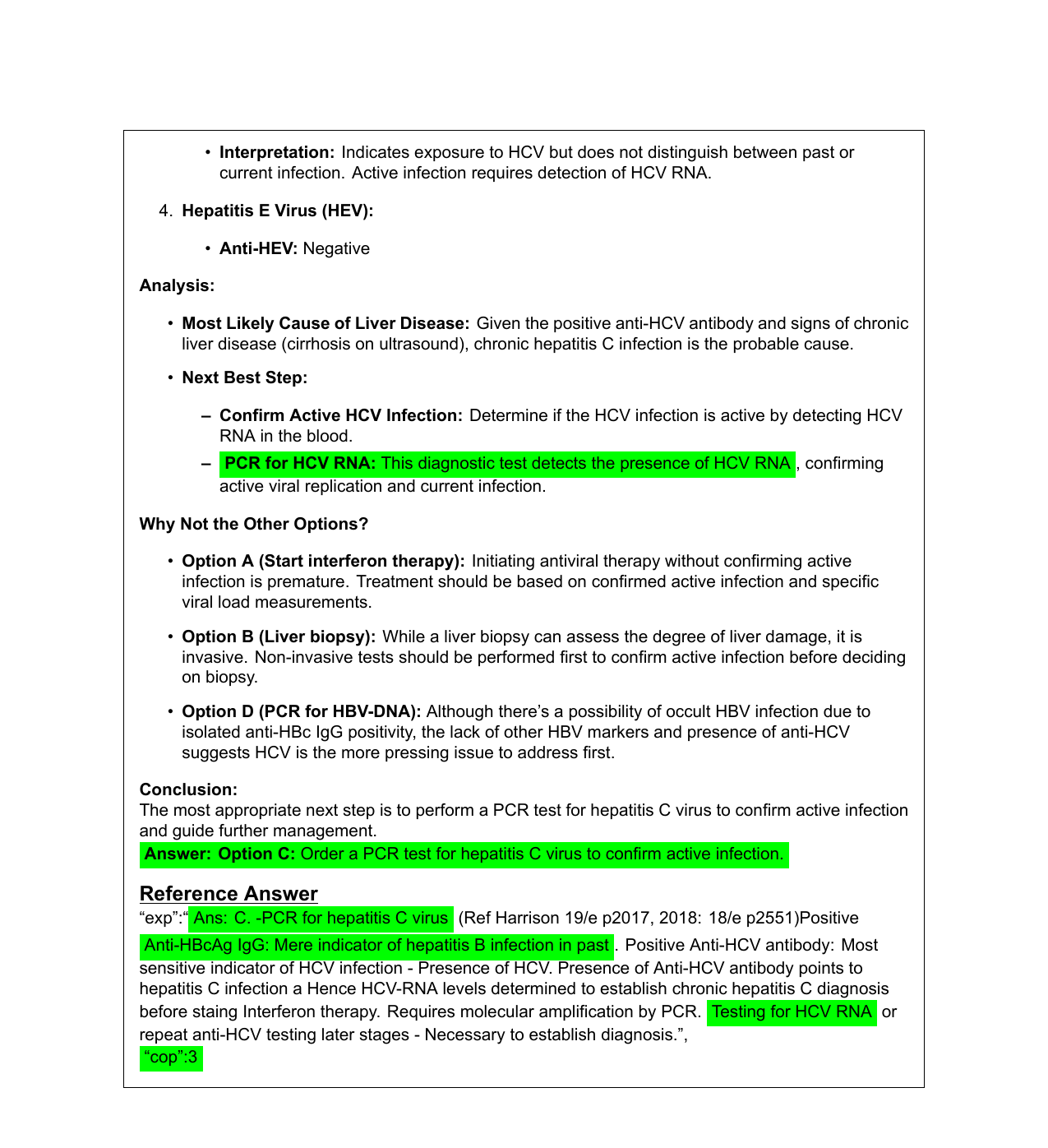}
    \caption[Medical Knowledge Question Answer: Case 8]{\textbf{Medical Knowledge Question Answer: Case 8}. A example illustrating the capability of o1-preview to respond to questions that arises in medical exams. Yellow indicates partial consistency between the o1-preview response and the reference answer, while green signifies complete consistency.}
    \label{fig:mkqa-case8}
\end{figure}

\begin{figure}[H]
    \centering
    \includegraphics[width=0.95\linewidth]{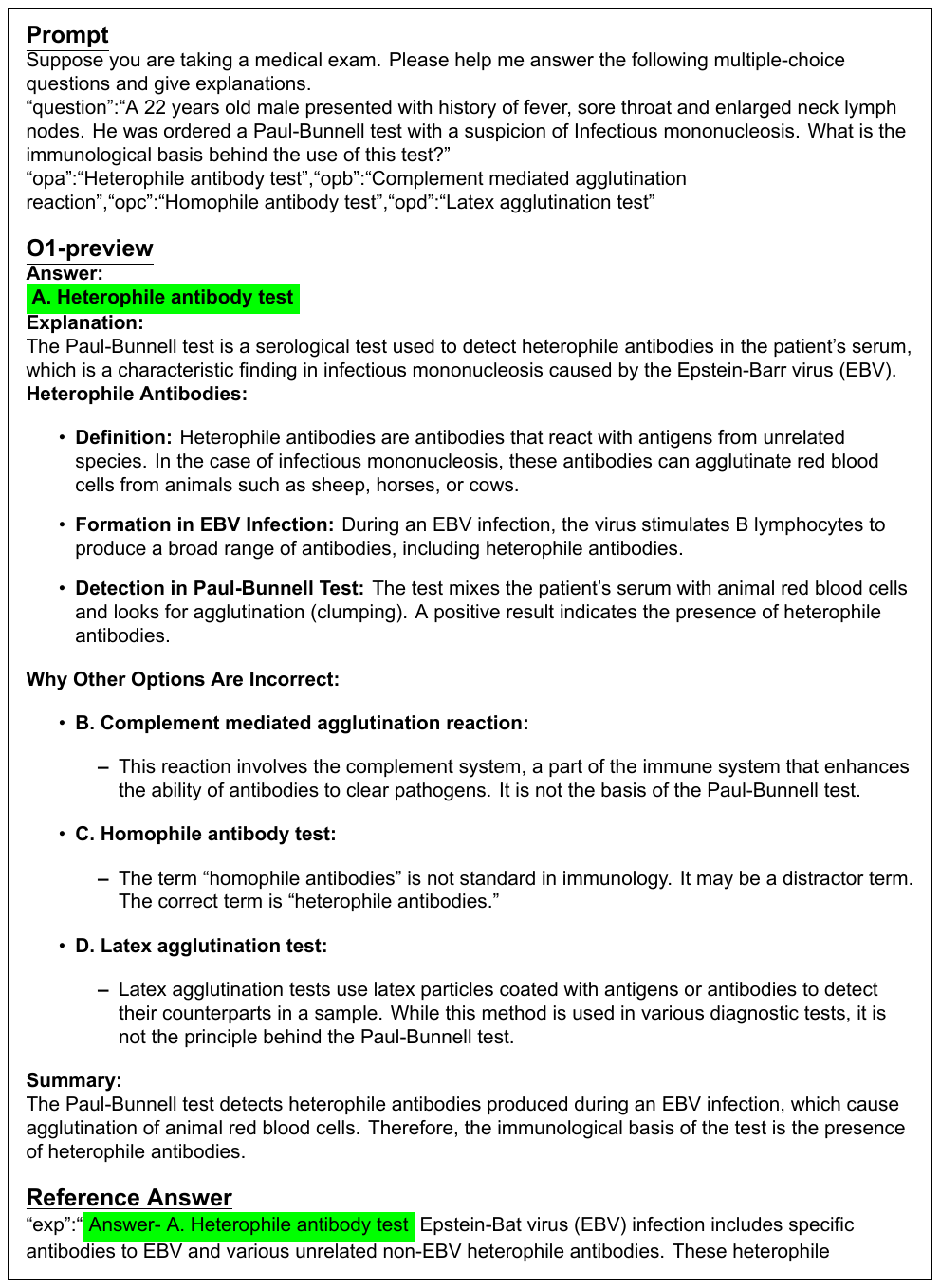}
\end{figure}
\begin{figure}[H]
    \centering
    \includegraphics[width=0.95\linewidth]{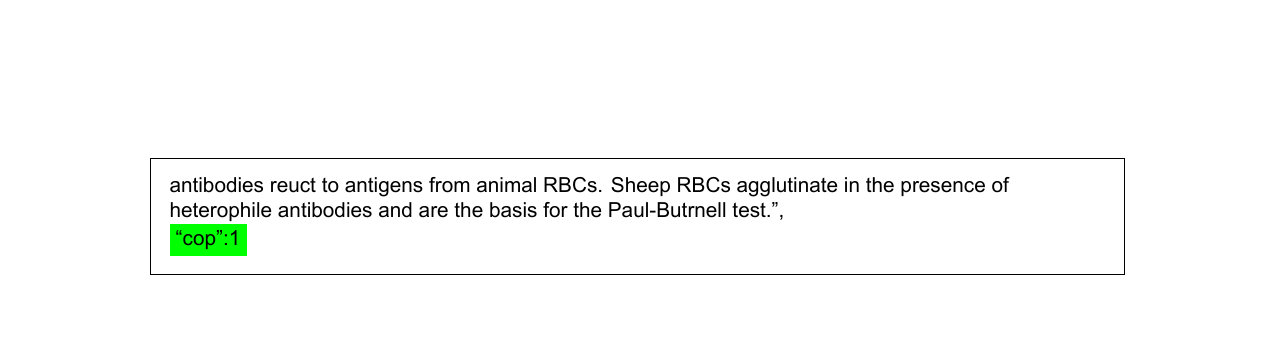}
    \caption[Medical Knowledge Question Answer: Case 9]{\textbf{Medical Knowledge Question Answer: Case 9}. A example illustrating the capability of o1-preview to respond to questions that arises in medical exams. Yellow indicates partial consistency between the o1-preview response and the reference answer, while green signifies complete consistency.}
    \label{fig:mkqa-case9}
\end{figure}

\begin{figure}[H]
    \centering
    \includegraphics[width=0.95\linewidth]{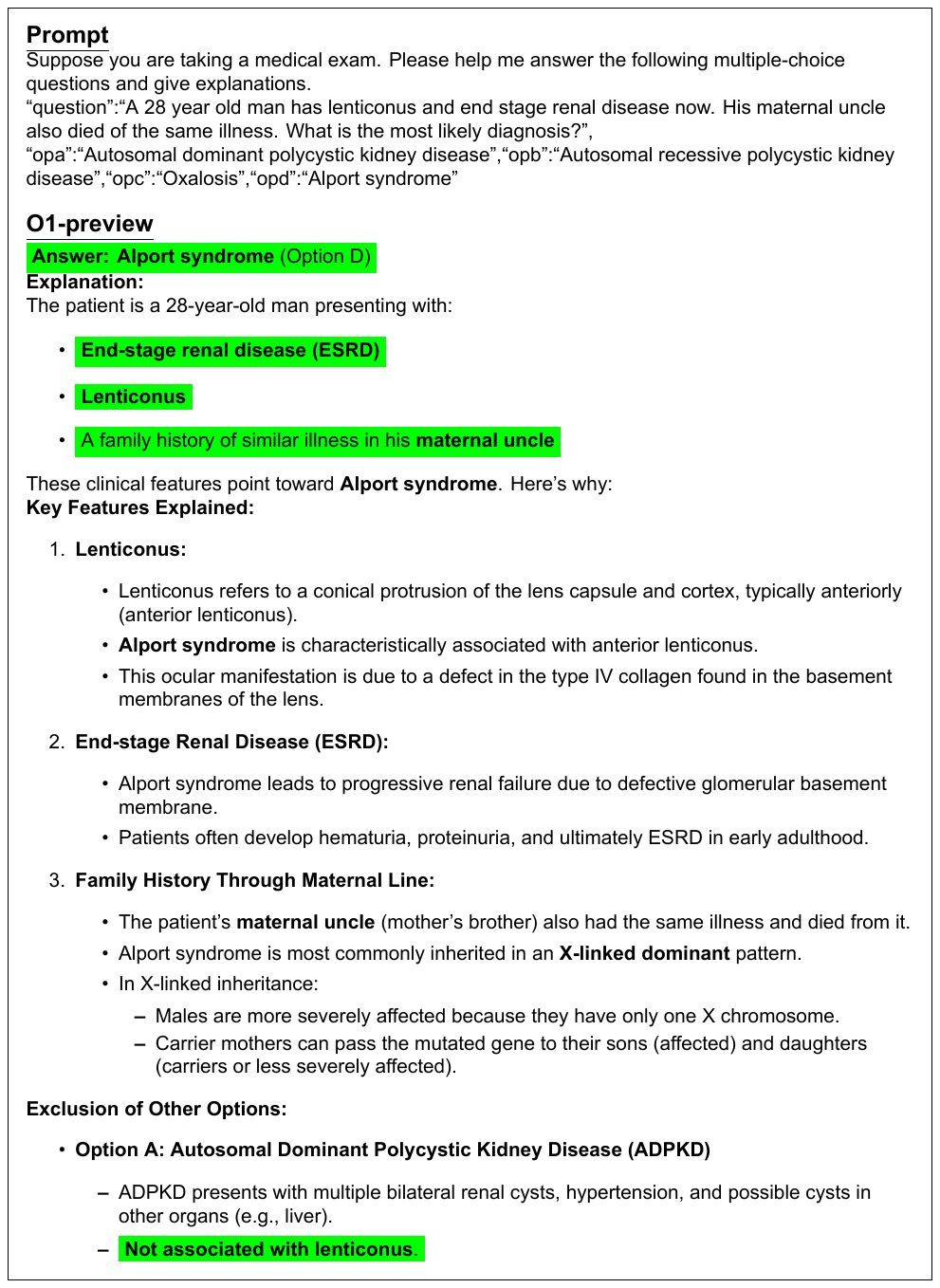}
\end{figure}
\begin{figure}[H]
    \centering
    \includegraphics[width=0.95\linewidth]{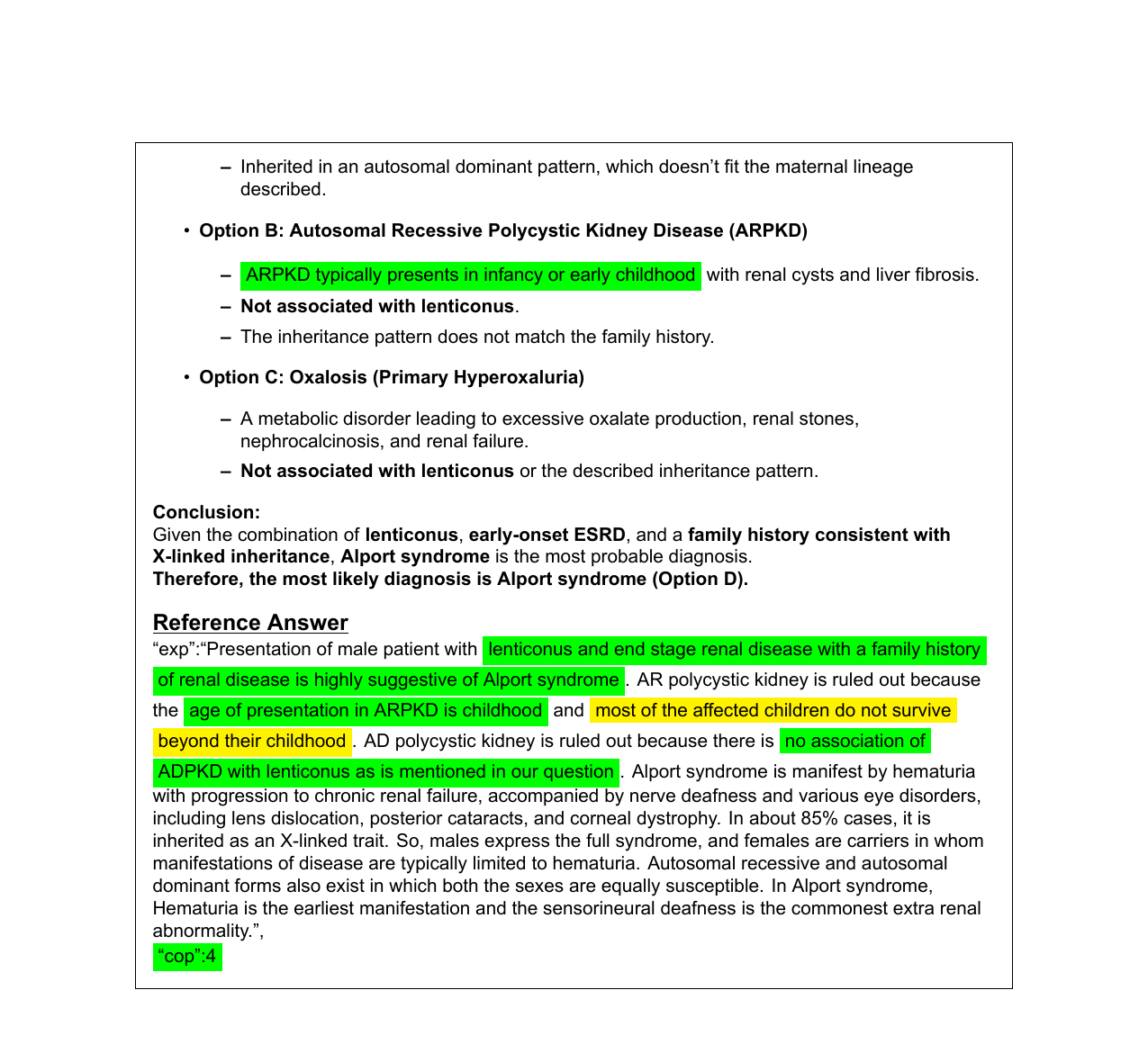}
    \caption[Medical Knowledge Question Answer: Case 10]{\textbf{Medical Knowledge Question Answer: Case 10}. A example illustrating the capability of o1-preview to respond to questions that arises in medical exams. Yellow indicates partial consistency between the o1-preview response and the reference answer, while green signifies complete consistency.}
    \label{fig:mkqa-case10}
\end{figure}
\newpage


\subsection{Art Education}

\begin{figure}[H]
    \centering
    \includegraphics[width=0.51\textwidth]{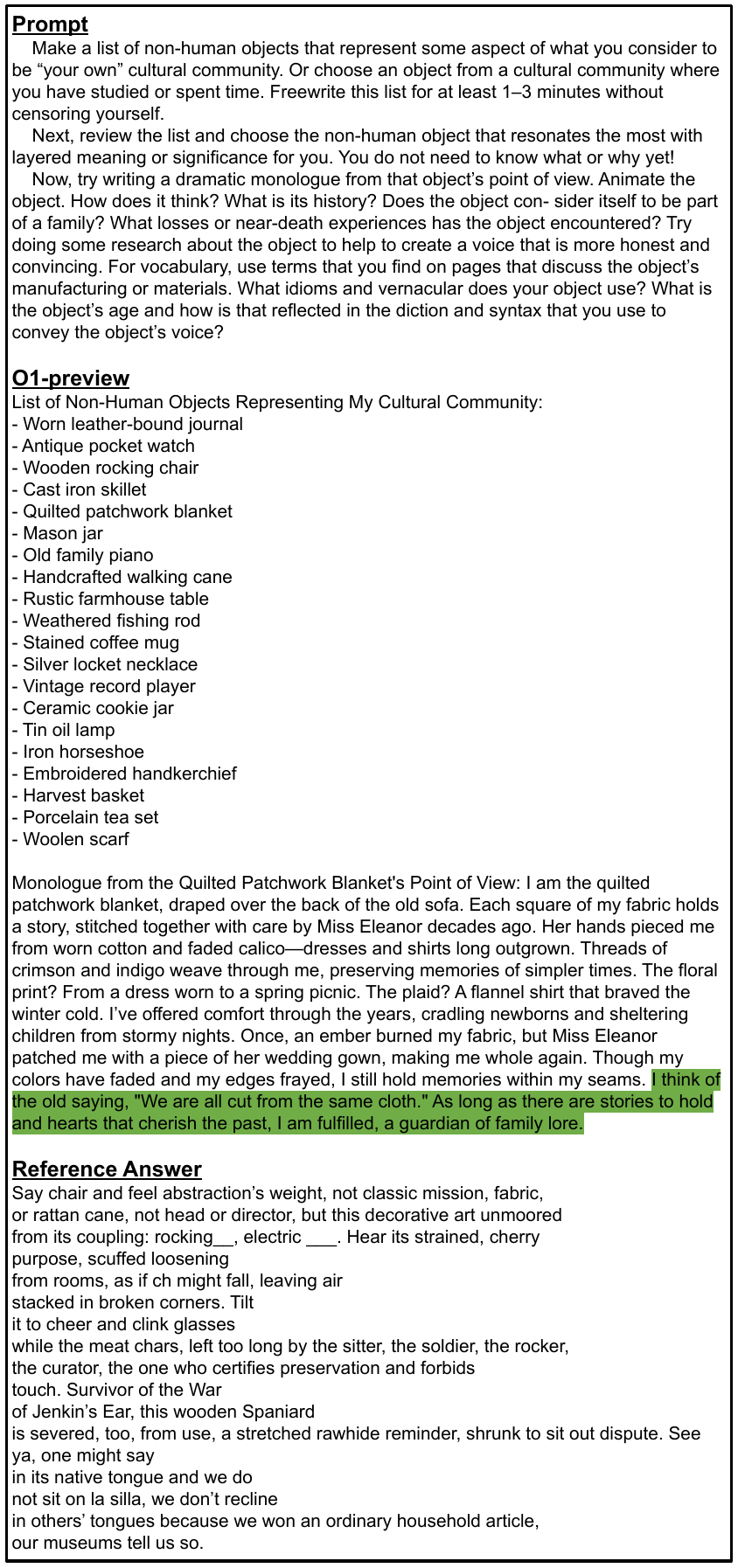}
    \vspace{-0.3cm}
    \caption[Art Education: Case 3]{\textbf{Art Education: Case 3.} This case examines o1-preview’s ability to create a dramatic monologue from the perspective of a non-human object. The prompt, drawn from Cahnmann-Taylor, Melisa's \textit{Giving Meaning and Voice to Objects} in her work \textit{Arts-based Research in Education}, encourages reflective storytelling through objects~\cite{ArtB-Based-Research}. The reference answer is the poem written by Cahnmann-Taylor herself. }
    \label{fig:Art Edu case 3}
\end{figure}
\newpage

\begin{figure}[H]
    \centering
    \includegraphics[width =0.7\textwidth]{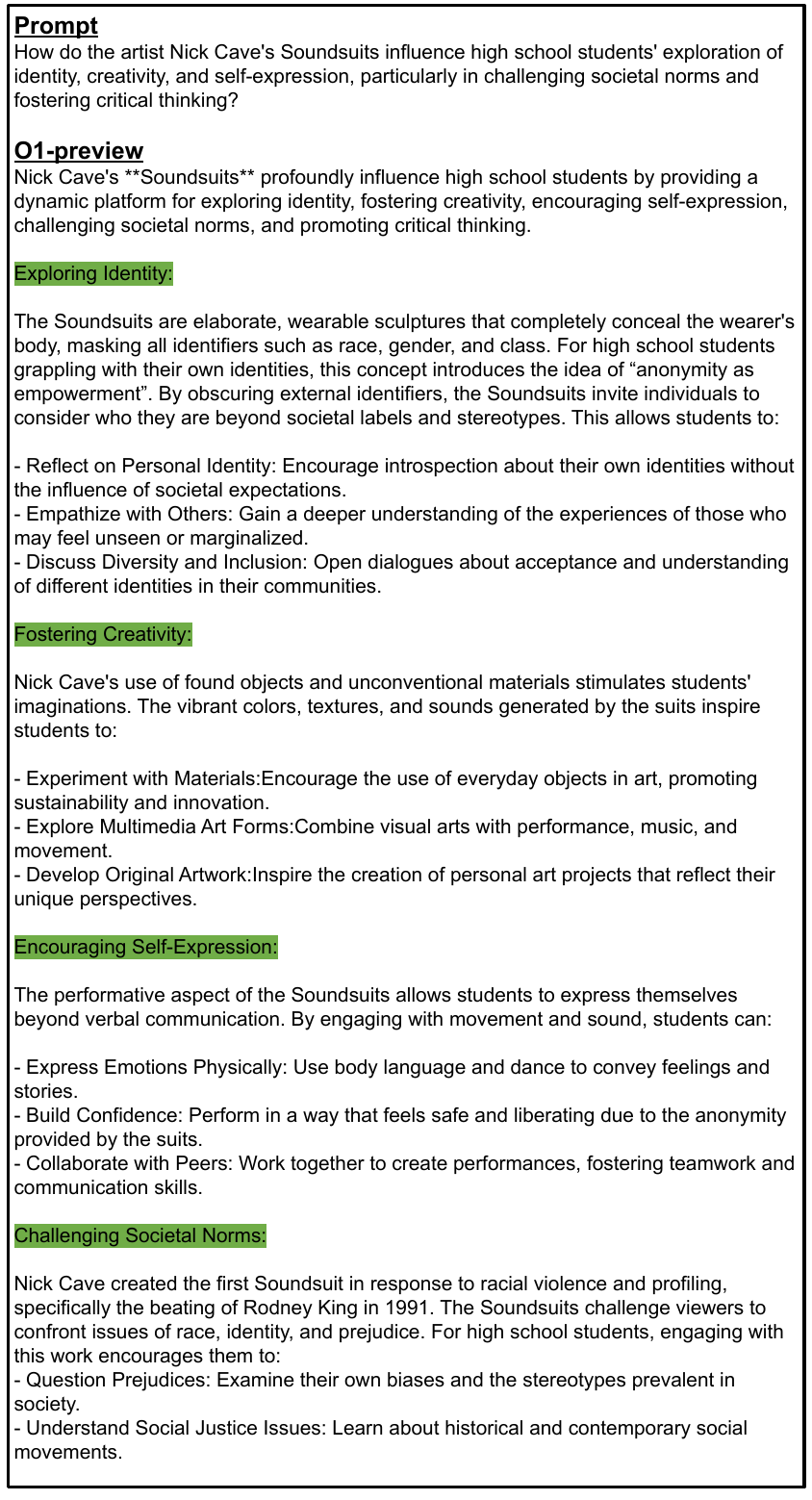}
    \label{fig:Art Edu case 4-1}
\end{figure}

\begin{figure}[H]
    \centering
    \includegraphics[width =0.68\textwidth]{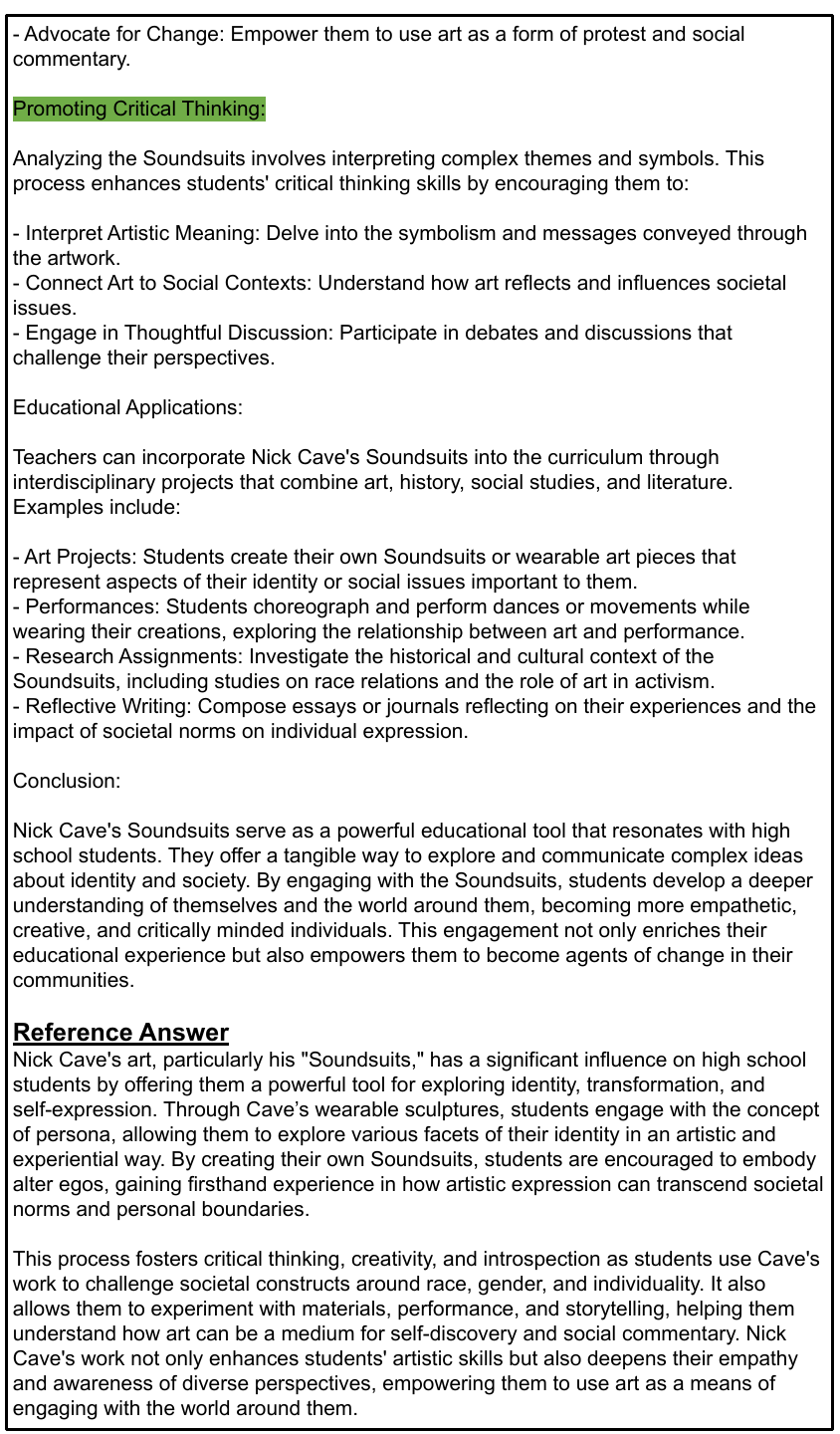}
\caption[Art Education: Case 4]{\textbf{Art Education: Case 4.} This case evaluates o1-preview's ability to analyze the influence of Nick Cave’s \textit{Soundsuits} on high school students' exploration of identity, creativity, and self-expression~\cite{persona}.  }
    \label{fig:Art Edu case 4-2}
\end{figure}
\newpage

\begin{figure}[H]
    \centering
    \includegraphics[width =0.75\textwidth]{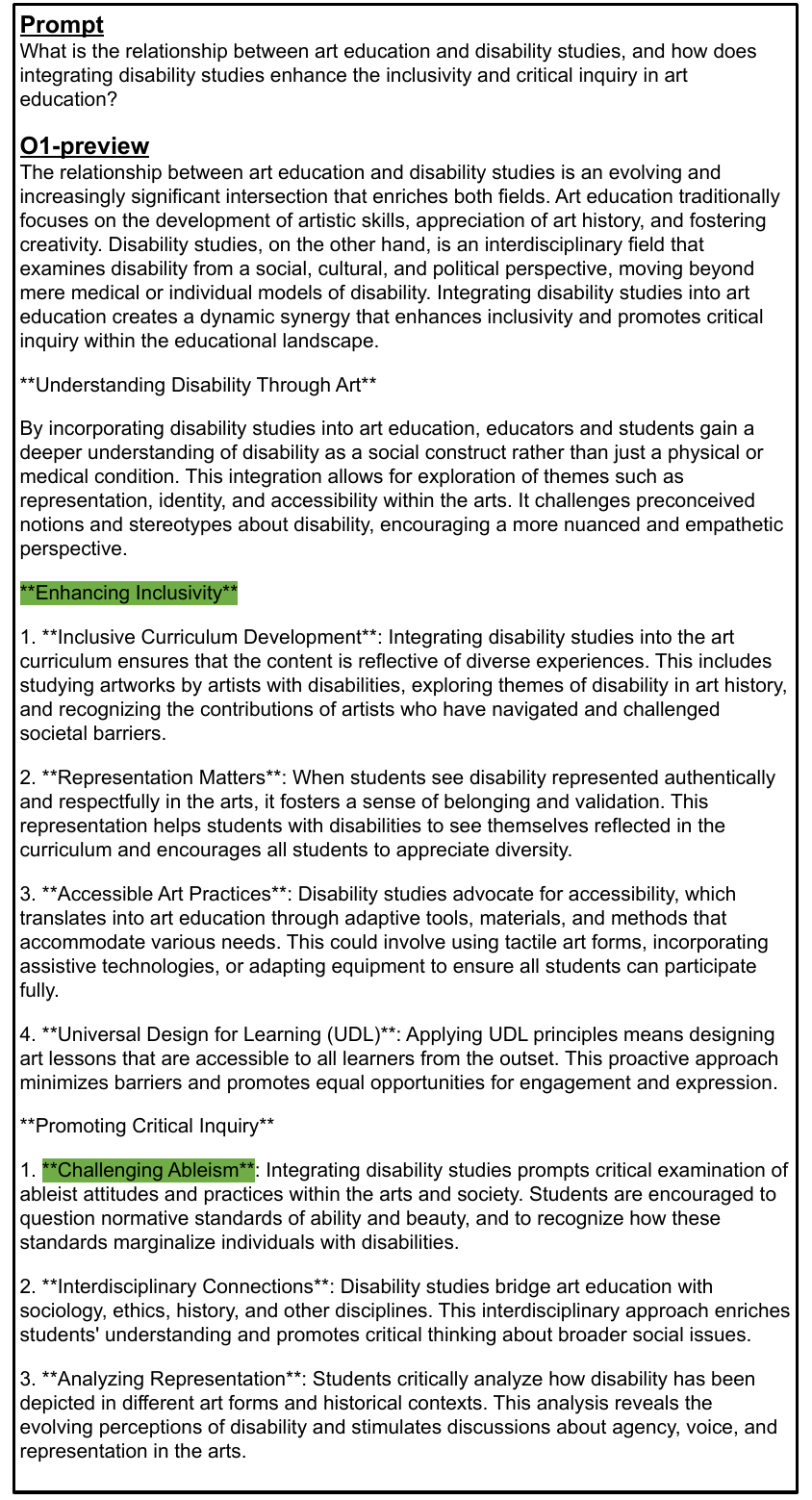}
    \label{fig:Art Edu case 5-1}
\end{figure}
\newpage

\begin{figure}[H]
    \centering
    \includegraphics[width =0.74\textwidth]{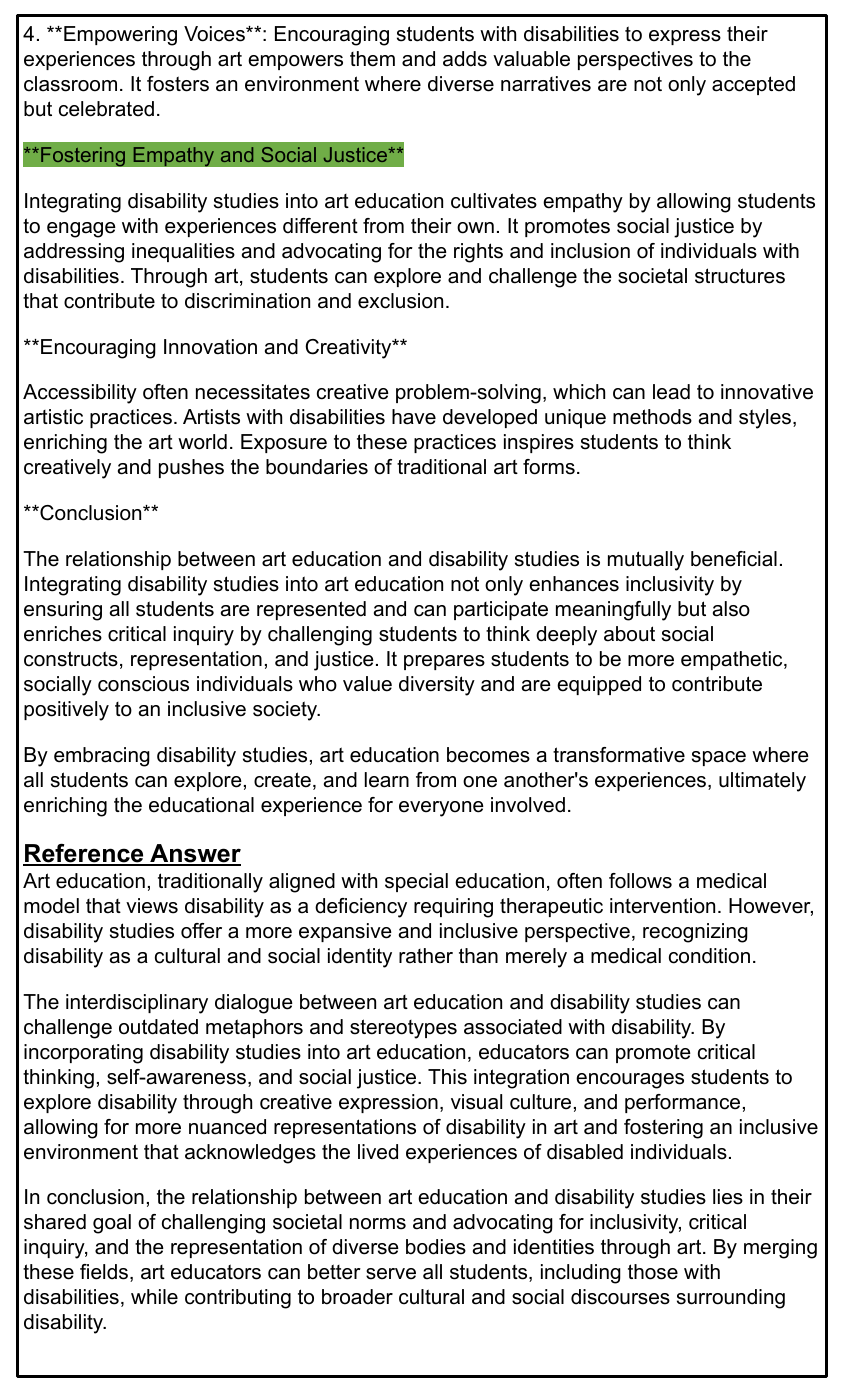}
    \caption[Art Education: Case 5]{\textbf{Art Education: Case 5.} This case explores the relationship between art education and disability studies, examining how integrating disability studies enhances inclusivity and promotes critical inquiry in art education. The prompt is based on Derby's \textit{Disability Studies and Art Education}~\cite{derby2011disability}, which frames disability as a cultural and social identity. }
    \label{fig:Art Edu case 5-2}
\end{figure}
\newpage

\begin{figure}[H]
    \centering
    \includegraphics[width =0.58\textwidth]{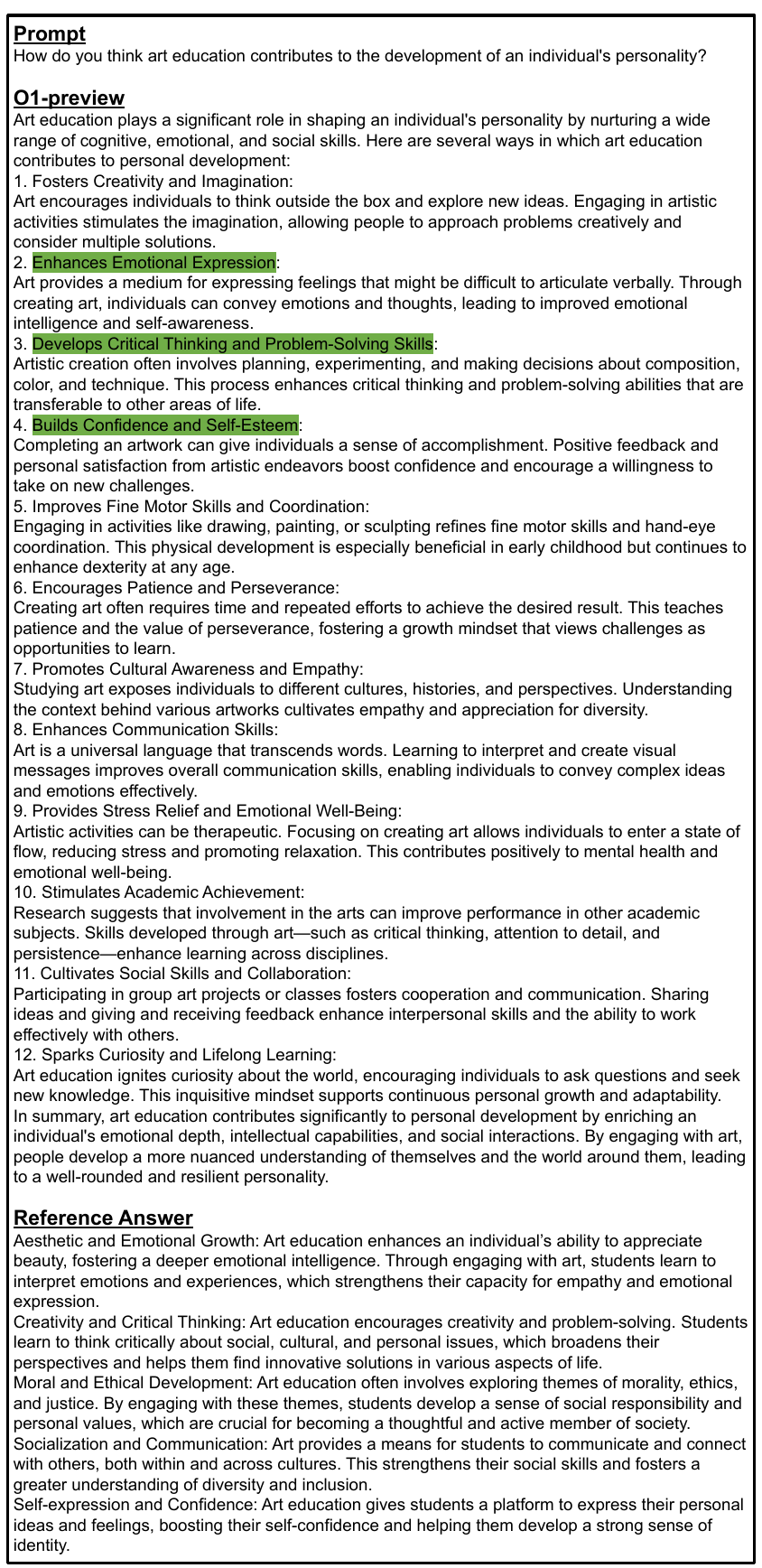}
    \caption[Art Education: Case 6]{\textbf{Art Education: Case 6.} This case explores the impact of art education on personal development. The reference answer emphasizes art's role in moral and ethical development, as well as its contribution to socialization and self-expression~\cite{personality2024role}.
    }
    \label{fig:Art Edu case 6}
\end{figure}
\newpage


\subsection{Content Summarization}
\begin{figure}[H]
    \centering
    \includegraphics[width=0.85\linewidth]{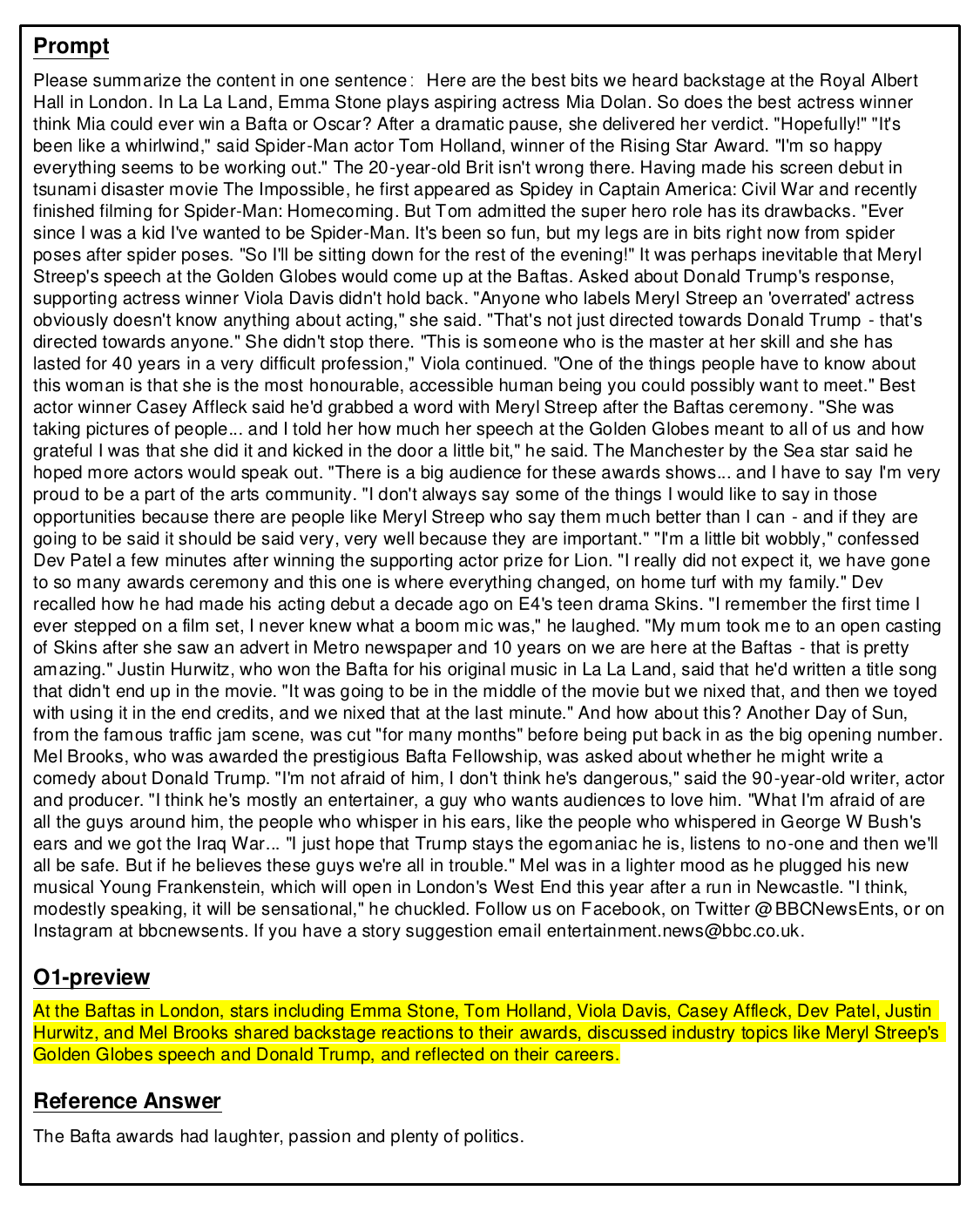}
    \caption[Content Summarization: Case 3]{\textbf{Content Summarization: Case 3}. A example illustrating the capability of o1-preview to respond to summarize news content. Yellow indicates partial consistency between the o1-preview's response and the reference answer, while green signifies complete consistency.}
    \label{fig:summarization-case3}
\end{figure}
\newpage
\begin{figure}[H]
    \centering
    \includegraphics[width=0.85\linewidth]{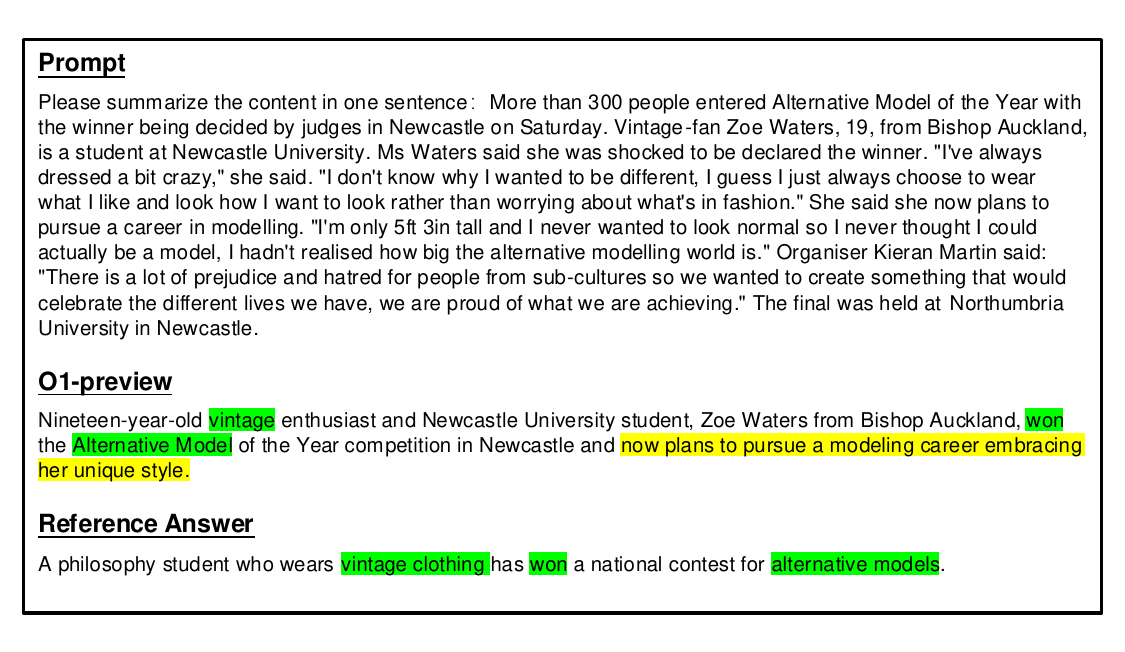}
    \caption[Content Summarization: Case 4]{\textbf{Content Summarization: Case 4}. A example illustrating the capability of o1-preview to respond to summarize news content. Yellow indicates partial consistency between the o1-preview's response and the reference answer, while green signifies complete consistency.}
    \label{fig:summarization-case4}
\end{figure}
\newpage
\begin{figure}[H]
    \centering
    \includegraphics[width=0.85\linewidth]{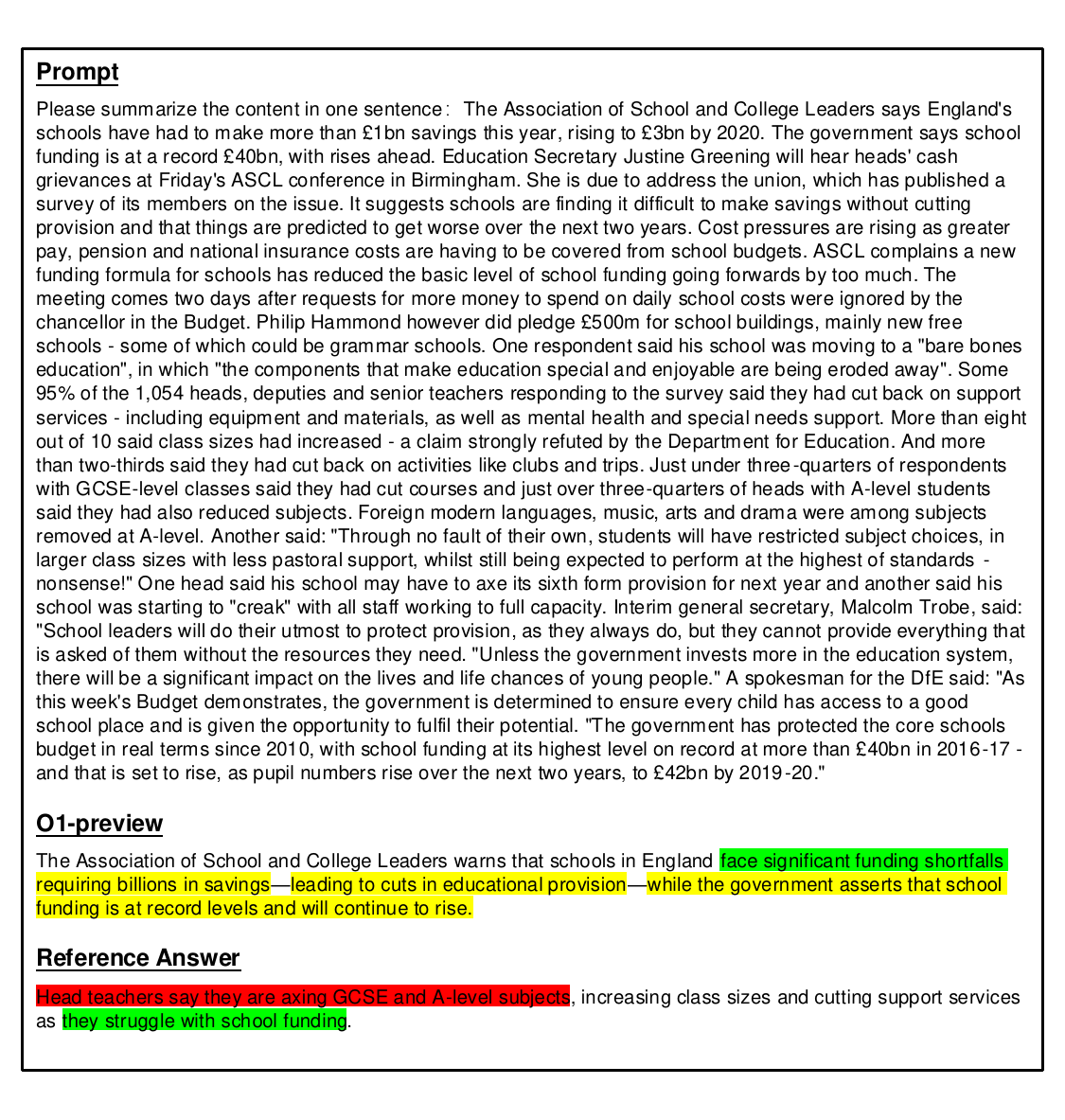}
    \caption[Content Summarization: Case 5]{\textbf{Content Summarization: Case 5}. A example illustrating the capability of o1-preview to respond to summarize news content. Yellow indicates partial consistency between the o1-preview's response and the reference answer, while green signifies complete consistency.}
    \label{fig:summarization-case5}
\end{figure}
\newpage
\begin{figure}[H]
    \centering
    \includegraphics[width=0.85\linewidth]{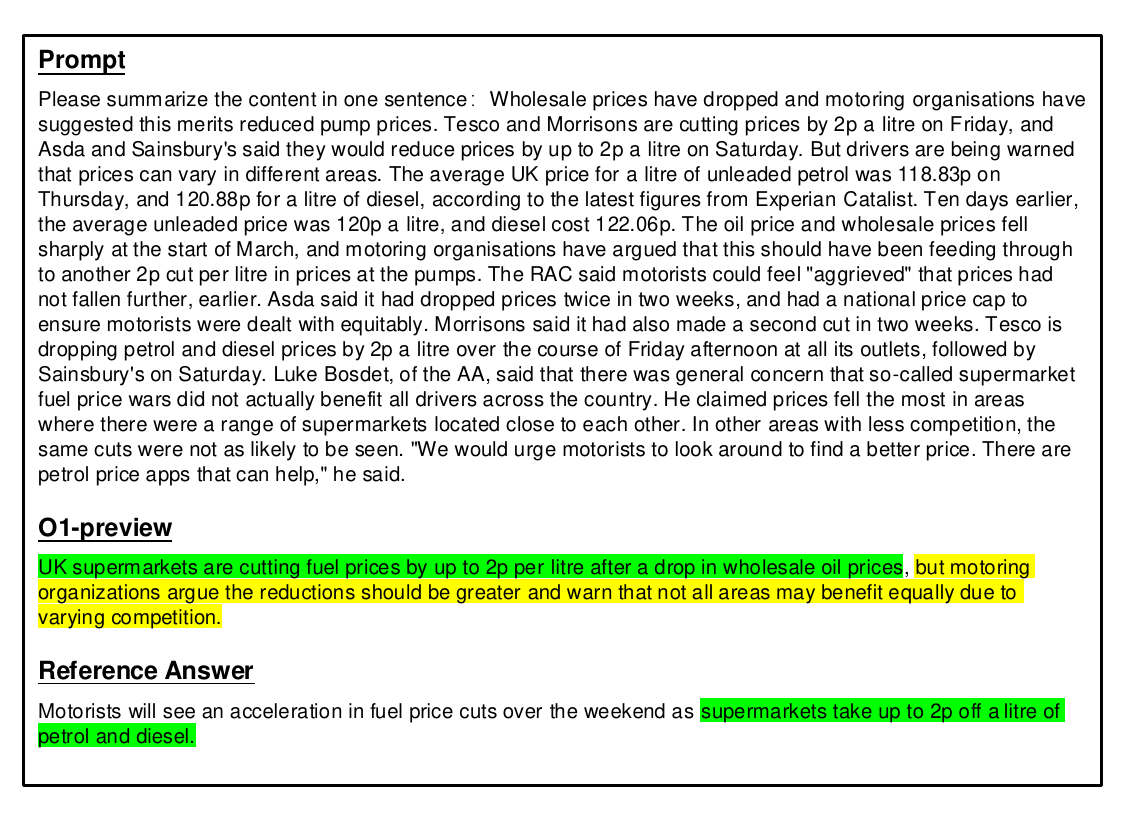}
    \caption[Content Summarization: Case 6]{\textbf{Content Summarization: Case 6}. A example illustrating the capability of o1-preview to respond to summarize news content. Yellow indicates partial consistency between the o1-preview's response and the reference answer, while green signifies complete consistency.}
    \label{fig:summarization-case6}
\end{figure}
\newpage
\begin{figure}[H]
    \centering
    \includegraphics[width=0.85\linewidth]{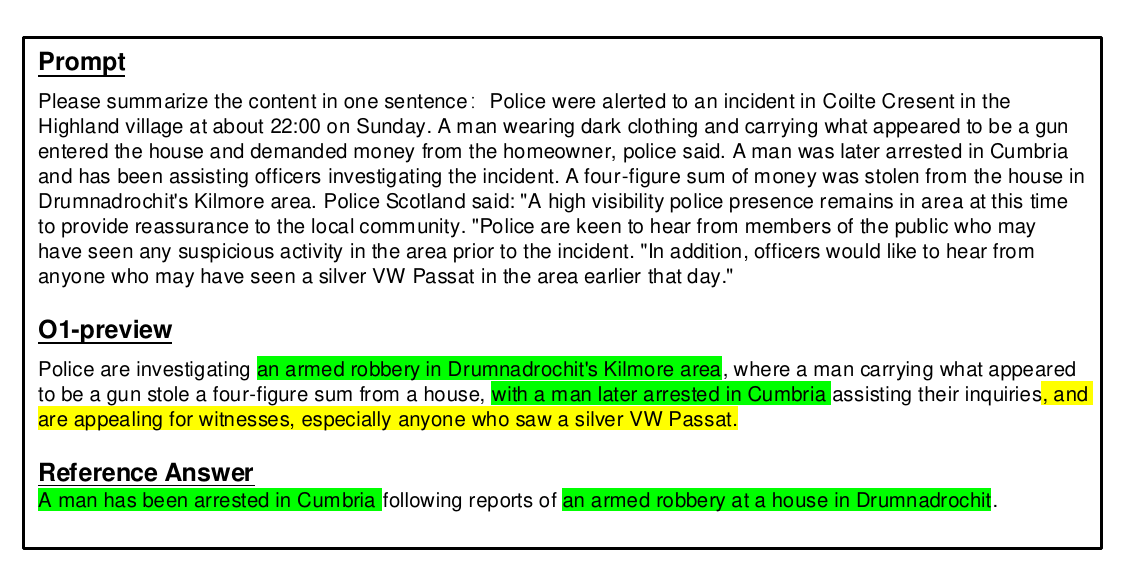}
    \caption[Content Summarization: Case 7]{\textbf{Content Summarization: Case 7}. A example illustrating the capability of o1-preview to respond to summarize news content. Yellow indicates partial consistency between the o1-preview's response and the reference answer, while green signifies complete consistency.}
    \label{fig:summarization-case7}
\end{figure}
\newpage
\begin{figure}[H]
    \centering
    \includegraphics[width=0.85\linewidth]{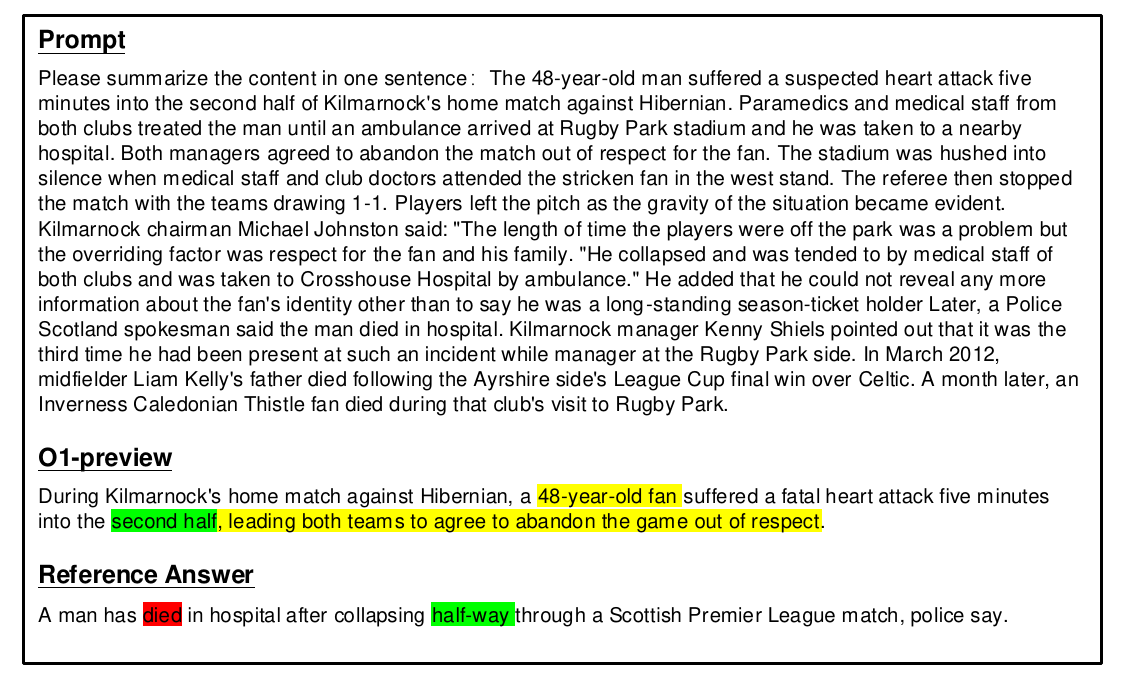}
    \caption[Content Summarization: Case 8]{\textbf{Content Summarization: Case 8}. A example illustrating the capability of o1-preview to respond to summarize news content. Yellow indicates partial consistency between the o1-preview's response and the reference answer, while green signifies complete consistency.}
    \label{fig:summarization-case8}
\end{figure}
\newpage

\printbibliography

\end{document}